%% file: DA_rodler.tex
\documentclass[10pt,a4paper,oneside,onecolumn]{book}

\usepackage{caption}

\usepackage[margin=10pt,font=small,labelfont=bf]{caption}
\usepackage[all]{xy}

\usepackage{makeidx}  
\usepackage{times}
\usepackage{graphicx}
\usepackage{latexsym,footnote}
\usepackage{amsmath}
\DeclareFontFamily{U}{MnSymbolC}{}
\DeclareSymbolFont{MnSyC}{U}{MnSymbolC}{m}{n}    
\DeclareFontShape{U}{MnSymbolC}{m}{n}{
    <-6>  MnSymbolC5
   <6-7>  MnSymbolC6
   <7-8>  MnSymbolC7
   <8-9>  MnSymbolC8
   <9-10> MnSymbolC9
  <10-12> MnSymbolC10
  <12->   MnSymbolC12%
}{}
\DeclareMathSymbol{\powerset}{\mathord}{MnSyC}{180}\usepackage{amsthm}
\usepackage{algorithm}
\usepackage[noend]{algpseudocode}
\algrenewcommand\algorithmicrequire{\textbf{Input:}}
\algrenewcommand\algorithmicensure{\textbf{Output:}}

\usepackage{mathnotation}
\usepackage{array}
\usepackage{dashrule}
\usepackage[T1]{fontenc} 
\usepackage[draft]{fixme}
\usepackage{subfigure}
\usepackage{multirow}
\usepackage{tabularx}
\usepackage{cuted}
\usepackage{amssymb}
\usepackage{changepage}
\usepackage{enumerate}
\usepackage{color}															 
\usepackage[table]{xcolor}
\usepackage{booktabs}
\usepackage{arydshln}
\usepackage{cancel}
\usepackage{rotating}
\usepackage{framed}
\usepackage{float}
\usepackage{ marvosym }
\usepackage[a4paper]{geometry}
\usepackage{pdfpages}
\usepackage{fancyhdr}
\usepackage{hyperref}

\definecolor{light-gray1}{gray}{0.95}
\definecolor{darkgray}{rgb}{0.8,0.8,0.8}
\definecolor{lightgray}{rgb}{0.95,0.95,0.95}

\fxsetup{
nomargin,inline,index,
theme=color
}

\newcounter{examplecounter}
\newenvironment{example}{
    \refstepcounter{examplecounter}%
  
	\vspace{7pt}
	\noindent\textbf{Example \arabic{chapter}.\arabic{examplecounter}}%
  \quad
}{

\vspace{7pt}
}
\numberwithin{examplecounter}{chapter}

\newcounter{remarkcounter}
\newenvironment{remark}{
    \refstepcounter{remarkcounter}%
  
	\vspace{7pt}
	\noindent\textbf{Remark \arabic{chapter}.\arabic{remarkcounter}}%
  \quad
}{

\vspace{7pt}
}
\numberwithin{remarkcounter}{chapter}

\newcommand{\mC}{{\bf{C}}}

\newcommand{\Q}{{\mathit{Q}}}

\newcommand{\orig}{\mathsf{orig}}

\newcommand{\tax}{\mathit{ax}}

\newcommand{\mo}{\mathcal{K}}
\newcommand{\mt}{\mathcal{T}}
\newcommand{\ma}{\mathcal{A}}
\newcommand{\mi}{\mathcal{I}}
\newcommand{\mb}{\mathcal{B}}
\newcommand{\md}{\mathcal{D}}
\newcommand{\me}{\mathcal{E}}
\newcommand{\mc}{\mathcal{C}}

\newcommand{\Tp}{\mathit{P}}
\newcommand{\Tn}{\mathit{N}}
\newcommand{\tp}{\mathit{p}}
\newcommand{\tn}{\mathit{n}}
\newcommand{\dt}{\mathcal{D}_{t}}

\newcommand{\ot}{\mathcal{K}^{*}}
\newcommand{\mS}{{\bf{S}}}

\newcommand{\mD}{{\bf{D}}}
\newcommand{\allC}{{\bf{aC}}}
\newcommand{\minC}{{\bf{mC}}}
\newcommand{\minD}{{\bf{mD}}}
\newcommand{\allD}{{\bf{aD}}}
\newcommand{\dx}[1]{{\bf D}_{#1}^+}
\newcommand{\dnx}[1]{{\bf D}_{#1}^{-}}
\newcommand{\dz}[1]{{\bf D}_{#1}^0}

\newcommand{\scHS}{{\textsc{HS}}}

\newcommand{\RQ}{{\mathit{R}}}

\newcommand{\EX}[2]{{\mathbf{EX}(#1)_{#2}}}
\newcommand{\SO}{{\mathbf{Sol}}}

\newcommand{\Queue}{{\mathbf{Q}}}

\newcommand{\scQX}{{\textsc{QX}}}

\newcommand{\mQ}{{\bf{Q}}}

\newcommand{\FP}{\mathbf{FP}}
\newcommand{\QP}{\mathbf{QP}}
\newcommand{\Pt}{\mathfrak{P}}
\newcommand{\Just}{\mathsf{Just}}

\newtheorem{definition}{Definition}[chapter]{}
\newtheorem{prob_def}{Problem Definition}[chapter]{}
\newtheorem{proposition}{Proposition}[chapter]{}
\newtheorem{lemma}{Lemma}[chapter]{}
\newtheorem{corollary}{Corollary}[chapter]{}

\usepackage[nottoc]{tocbibind}

\begin{document}
\title{A Theory of Interactive Debugging of \\ Knowledge Bases in Monotonic Logics}
\date{}
\author{Patrick Rodler \\
  \multicolumn{1}{p{.7\textwidth}}{\centering\emph{
  Alpen-Adria Universit\"at, Klagenfurt, 9020 Austria}} \\ 
	patrick.rodler@aau.at}
\maketitle


\tableofcontents

\frontmatter

\listoffigures
\clearpage
\listoftables
\clearpage
\addcontentsline{toc}{chapter}{List of Algorithms}
\listofalgorithms
\addtocontents{loa}{\def\string\figurename{Algorithm}}
%

\input{abstract}
\mainmatter
\input{intro}
\input{basic}

\clearpage
\thispagestyle{empty}
\input{related}
\clearpage
\thispagestyle{empty}
\input{conclusion}

\bibliographystyle{splncs03}
\bibliography{library}
\end{document}

%% file: abstract.tex
\chapter{Abstract}
Most artificial intelligence applications rely on knowledge about a relevant real-world domain that is encoded in a knowledge base (KB) by means of some logical knowledge representation language. The most essential benefit of such logical KBs is the opportunity to perform automatic reasoning to derive implicit knowledge or to answer complex queries about the modeled domain. The feasibility of meaningful reasoning requires a KB to meet some minimal quality criteria such as consistency; that is, there must not be any contradictions in the KB. Without adequate tool assistance, the task of resolving such violated quality criteria in a KB can be extremely hard even for domain experts, especially when the problematic KB includes a large number of logical formulas, comprises complicated formalisms, was developed by multiple people or in a distributed fashion or was (partially) generated by means of some automatic systems. 

Non-interactive Debugging systems published in research literature often cannot localize all possible faults (\emph{incompleteness}), suggest the deletion or modification of unnecessarily large parts of the KB (\emph{non-minimality}), return incorrect solutions which lead to a repaired KB not satisfying the imposed quality requirements (\emph{unsoundness}) or suffer from \emph{poor scalability} due to the inherent complexity of the KB debugging problem. Even if a system is complete and sound and considers only minimal solutions, there are generally exponentially many solution candidates to select one from. However, any two repaired KBs obtained from these candidates differ in their semantics in terms of entailments and non-entailments. Selection of just any of these repaired KBs might result in unexpected entailments, the loss of desired entailments or unwanted changes to the KB which in turn might cause unexpected new faults during the further development or application of the repaired KB. Also, manual inspection of a large set of solution candidates can be time-consuming (if not practically infeasible), tedious and error-prone since human beings are normally not capable of fully realizing the semantic consequences of deleting a set of formulas from a KB. Hence there is a need for adequate tools that support a user when facing a faulty KB.

In this work, we account for these issues and propose methods for the interactive debugging of KBs which are complete and sound and compute only minimally invasive solutions, i.e.\ suggest the deletion or modification of just a set-minimal subset of the formulas in the problematic KB. User interaction takes place in the form of queries asked to a person, e.g.\ a domain expert, about intended and non-intended entailments of the correct KB. To construct a query, only a minimal set of two solution candidates must be available. After the answer to a query is known, the search space for solutions is pruned. Iteration of this process until 
there is only a single solution candidate left yields a repaired KB which features exactly the semantics desired and expected by the user. 

The novel contributions of this work are:
\begin{itemize}
	\item \emph{Thorough Theoretical Workup of the Topic of Interactive Debugging of Monotonic KBs:} We evolve the theory of the topic by first elaborating on the theory of non-interactive KB debugging, revealing crucial shortcomings in the application of non-interactive methods and thereby motivating the development and deployment of interactive approaches in KB debugging. Then, we give some important results that guarantee the feasibility of interactive KB debugging, give some precise definitions of the problems interactive KB debugging aims to solve and present algorithms that provably solve these problems.
	\item \emph{A Complete Picture of an Interactive Debugging System Is Drawn:} This is the first work that deals with an entire system of algorithms that are required for the interactive debugging of monotonic KBs, considers and details all algorithms separately, proves their correctness and demonstrates how all these algorithms are orchestrated to make up a full-fledged and provably correct interactive KB debugging system. 
	\item \emph{Two New Algorithms} for the iterative computation of candidate solutions in the scope of interactive KB debugging are proposed. The first one guarantees constant convergence towards the exact solution of the interactive KB problem by the ascertained reduction of the number of remaining solutions after any query is answered. The second one features powerful search tree pruning techniques and might thus be expected to exhibit a more time- and space-saving behavior than existing algorithms, in particular for growing problem instances.
\end{itemize}

%% file: intro.tex
\chapter{Introduction} 
\label{chap:intro}

\paragraph{Motivation.} Most artificial intelligence applications rely on knowledge that is encoded in a knowledge base (KB) by means of some logical knowledge representation language such as propositional logic~\cite{chang1973}, datalog~\cite{Ceri1989a}, first-order logic (FOL)~\cite{chang1973}, The Web Ontology Language (OWL~\cite{patel2004owl}, OWL~2~\cite{Grau2008a,Motik2009a}) or description logic (DL)~\cite{Baader2007}.
Experts in a variety of application domains keep developing KBs of constantly growing size. A concrete example of a repository containing biomedical KBs is the Bioportal\footnote{http://bioportal.bioontology.org}, which comprises vast ontologies with tens or even hundreds of thousands of terms each (e.g.~the SNOMED-CT ontology with currently over 395.000 terms). Such KBs however pose a significant challenge for people as well as tools involved in their evolution, maintenance and application. 

All these activities are based on the most essential benefit of KBs, namely the opportunity to perform automatic reasoning to derive implicit knowledge or to answer complex queries about the modeled domain. 
The feasibility of meaningful reasoning requires a KB to meet the minimum quality criterion \emph{consistency}, i.e.\ there must not be any contradictions in the KB. Because \emph{any} logical formula can be derived from an inconsistent KB. Further on, one might postulate further requirements to be met by a KB. For instance, one might consider faulty a FOL KB entailing $\forall X\, \lnot p(X)$ for some predicate symbol $p$ occurring in the KB. Such a KB would be incoherent, i.e.\ it would violate the requirement \emph{coherency} (which has originally been defined for DL KBs~\cite{Schlobach2007,Parsia2005}. Additionally, test cases can be specified giving information about desired (\emph{positive test cases}) and non-desired (\emph{negative test cases}) entailments a correct KB should feature. This characterization of a KB's intended semantics is a direct analogon to the field of software debugging, where test cases are exploited as a means to verify the correct semantics of the program code.

As KBs are growing in size and complexity, their likeliness of violating one of these criteria increases. Faults in KBs may, for instance, arise because human reasoning is simply overstrained~\cite{Horridge2011b, Horridge2009}. That is, generally a person will not be capable of completely grasping or mentally processing the entire knowledge contained in a (large or complex) KB at once. In fact, a person might fully comprehend some isolated part of a the KB, but might
not be able to determine or understand all implications or non-implications of 
this isolated part combined with other parts of a KB, i.e.\ when new logical formulas are added.

Another reason for the non-compliance with the mentioned quality criteria imposed on KBs might be that multiple (independently working) editors contribute to the development of the KB~\cite{Noy2006a} which may lead to contradictory formulas. The OBO Project\footnote{http://obo.sourceforge.net} and the NCI Thesaurus\footnote{http://nciterms.nci.nih.gov/ncitbrowser} are examples of collaborative KB development projects. Employing automatic tools, e.g.\ \cite{Jimenez-Ruiz2011,Ngo2012,Jean-Mary2009}, to generate (parts of) KBs can further exacerbate the task of KB quality assurance~\cite{meilicke2011,Ferrara2011}. 

Moreover, as studies in cognitive psychology~\cite{Ceraso71,Johnson1999} attest, humans make systematic errors while formulating or interpreting logical formulas. These observations are confirmed by \cite{Rector2004,Roussey2009} which present common faults people make when developing a KB (ontology). Hence, it is essential to devise methods that can efficiently identify and correct faults in a KB. 

\begin{figure*}[htbp]
	\centering
		\includegraphics[width=0.6\textwidth]{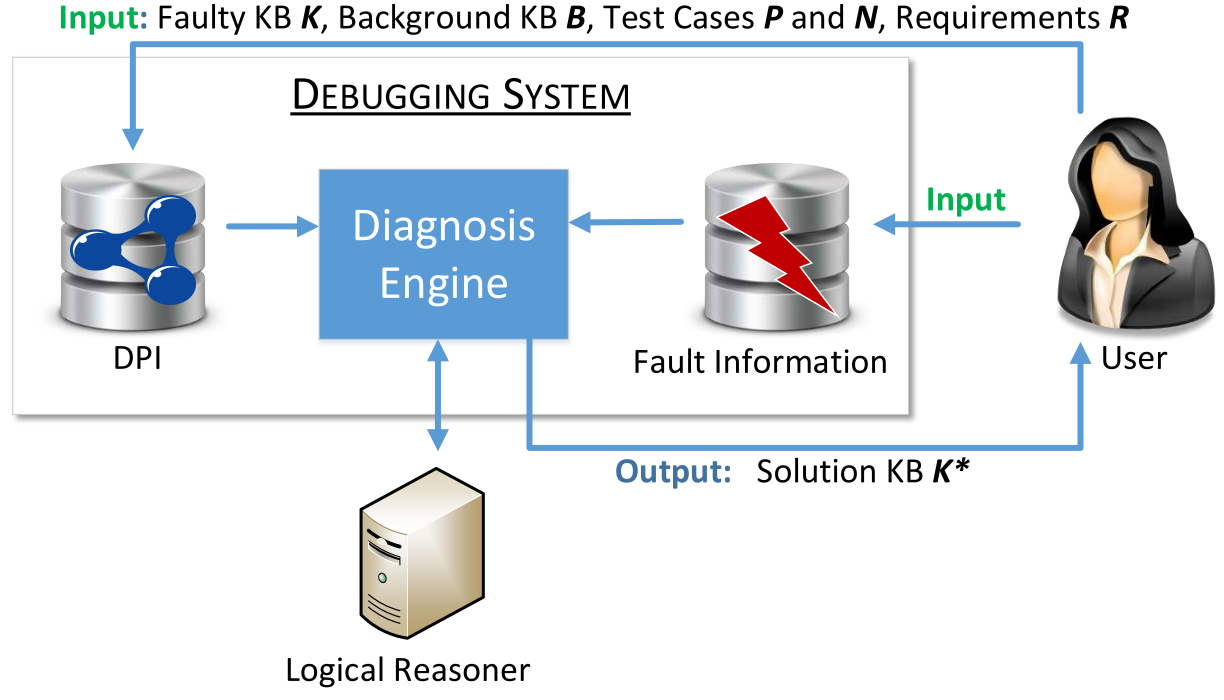}
	\caption[The Principle of Non-Interactive KB Debugging]{The principle of non-interactive KB debugging.}
	\label{fig:non-interactive_debugging_workflow}
\end{figure*}

\paragraph{Non-Interactive KB Debugging.} Given a set of requirements to the KB and sets of test cases, KB debugging methods~\cite{Schlobach2007,Kalyanpur.Just.ISWC07,friedrich2005gdm,Horridge2008} can localize a (potential) fault by computing a subset $\md$ of the formulas in the KB $\mo$ called a \emph{diagnosis}. At least all formulas in a diagnosis must be (adequately) modified or deleted in order to obtain a KB $\ot$ that satisfies all postulated requirements and test cases. Such a KB $\ot$ constitutes the solution to the \emph{KB debugging problem}. Figure~\ref{fig:non-interactive_debugging_workflow}\footnote{Thanks to Kostyantyn Shchekotykhin for making available to me parts of this diagram.} outlines such a KB debugging system. The input to the system is a \emph{diagnosis problem instance (DPI)} defined by 
\begin{itemize}
	\item some KB $\mo$ formulated using some (monotonic) logical language $\mathcal{L}$ (every formula in $\mo$ might be correct or faulty),
	\item (optionally) some KB $\mb$ (over $\mathcal{L}$) formalizing some background knowledge relevant for the domain modeled by $\mo$ (such that $\mb$ and $\mo$ do not share any formulas; all formulas in $\mb$ are considered correct)
	\item a set of requirements $\RQ$ to the correct KB,
	\item sets of positive ($\Tp$) and negative ($\Tn$) test cases (over $\mathcal{L}$) asserting desired semantic properties of the correct KB and
	\item (optionally) some fault information $\FP$, e.g.\ in terms of fault probabilities of logical formulas in $\mo$.
\end{itemize}
Moreover, the system requires a sound and complete logical reasoner for deciding consistency (coherency) and calculating logical entailments of a KB formulated over the language $\mathcal{L}$. Some approaches (including the ones presented in this work) use the reasoner as a black-box (e.g.\ \cite{Shchekotykhin2012,Horridge2011a}) within the debugging system. That is, the reasoner is called as is and serves as an oracle independent from other computations during the debugging process; that is, the internals of the reasoner are irrelevant for the debugging task. On the other hand, glass-box approaches (e.g.\ \cite{Schlobach2007,Horridge2011a,kalyanpur2005}) attempt to exploit internal modifications of the reasoner for debugging purposes; in other words, the sources of problems (e.g.\ contradictory formulas) in the KB are computed as a direct consequence of reasoning~\cite{Horridge2011a}. The advantages of a black-box approach over a glass-box approach are the lower memory consumption and better performance~\cite{kalyanpur2005} of the reasoner and the reasoner independence of the debugging method. The latter benefit is essential for the generality of our approaches and their applicability to various knowledge representation formalisms. 

Given these inputs, the debugging system focuses on (a subset of) all possible fault candidates (usually the set of minimal, i.e.\ irreducible, diagnoses) and usually outputs the most probable one amongst these if some fault information is provided or the minimum cardinality one, otherwise. Alternatively, a debugging system might also be employed to calculate a predefined number of (most probable or minimum cardinality) minimal diagnoses or to determine all minimal diagnoses computable within a predefined time limit. 


\paragraph{Issues with Non-Interactive KB Debugging Systems.} In real-world scenarios, debugging tools often have to cope with large numbers of minimal diagnoses where the trivial application, i.e.\ deletion, of any minimal diagnosis leads to a (repaired) KB with different semantics in terms of entailed and non-entailed formulas. For example, in~\cite{ksgf2010} a sample study of real-world KBs revealed that the number of different minimal diagnoses might exceed thousand by far (1782 minimal diagnoses for a KB with only 1300 formulas). In such situations simple visualization of all these alternative modifications of the ontology is clearly ineffective. Selecting a wrong diagnosis (in terms of its semantics, \emph{not} in terms of fulfillment of test cases and requirements) can lead to unexpected entailments or non-entailments, lost desired entailments and surprising future faults when the KB is further developed. Manual inspection of a large set of (minimal) diagnoses is time-consuming (if not practically infeasible), error-prone and often computationally infeasible due to the complexity of diagnosis computation. 

Moreover, \cite{Stuckenschmidt2008} has put several (non-interactive) debugging systems to the test using a test set of faulty (incoherent OWL) real-world KBs which were partly designed by humans and partly by the application of automatic systems. The result was that most of the investigated systems had serious performance problems, ran out of memory, were not able to locate all the existing faults in the KB (incompleteness), reported parts of a KB as faulty which actually were not faulty (unsoundness), produced only trivial solutions or suggested non-minimal faults (non-minimality). Often, performance problems and incompleteness of non-interactive debugging methods can be traced back to an explosion of the search tree for minimal diagnoses.

\paragraph{The Solution: Interactive KB Debugging.} In this work we present algorithms for interactive KB debugging. These aim at the gradual reduction of compliant minimal diagnoses by means of user interaction, thereby seeking to prevent the search tree for minimal diagnoses from exploding in size by performing regular pruning operations. ``User'' in this case might refer to a single person or multiple persons, usually experts of the particular domain the faulty KB is dealing with such as biology, medicine or chemistry. 
Throughout an interactive debugging session, the user is asked a set of automatically chosen queries about the domain that should be modeled by a given faulty KB. A query can be created by the system after a set $\mD$ of a minimum of two minimal diagnoses has been precomputed (we call $\mD$ the \emph{leading diagnoses}). Each query is a conjunction (i.e.\ a set) of logical formulas that are entailed by some correct subset of the formulas in the KB. With regard to one particular query $Q$, any set of minimal diagnoses for the KB, in particular the set $\mD$ which has been utilized to generate $Q$, can be partitioned into three sets, the first one ($\dx{}$) including all diagnoses in $\mD$ compliant only with a positive answer to $Q$, the second ($\dnx{}$) including all diagnoses in $\mD$ compliant only with a negative answer to $Q$, and the third ($\dz{}$) including all diagnoses in $\mD$ compliant with both answers. A positive answer to $Q$ signalizes that the conjunction of formulas in $Q$ must be entailed by the correct KB wherefore $Q$ is added to the set of positive test cases. Likewise, if the user negates $Q$, this is an indication that at least one formula in $Q$ must not be entailed by the correct KB. As a consequence, $Q$ is added to the set of negative test cases.

Assignment of a query $Q$ to either set of test cases results in a new debugging scenario. In this new scenario, all elements of $\dnx{}$ are no longer minimal diagnoses given that $Q$ has been classified as a positive test case. Otherwise, all diagnoses in $\dx{}$ are invalidated. In this vein, the successive reply to queries generated by the system will lead the user to the single minimal solution diagnosis that perfectly reflects their intended semantics. In other words, after deletion of all formulas in the solution diagnosis from the KB and the addition of the conjunction of all formulas in the specified positive test cases to the KB, the resulting KB meets all requirements and positive as well as negative test cases. In that, the added formulas contained in the positive test cases serve to replace the desired entailments that are broken due to the deletion of the solution diagnosis from the KB. 

Thence, in the interactive KB debugging scenario the user is not required to cope with the understanding of which faults (e.g.\ sources of inconsistency or implications of negative test cases) occur in the faulty initial KB, why they are faults (i.e.\ why particular entailments are given and others not) and how to repair them. All these tasks are undertaken by the interactive debugging system. 

The proposed approaches to interactive KB debugging in this work follow the standard \emph{model-based diagnosis (MBD)} technique~\cite{Reiter87,dekleer1987}. MBD has been successfully applied to a great variety of problems in various fields such as robotics~\cite{Steinbauer2005}, planning~\cite{Steinbauer2009}, debugging of software programs~\cite{wotawa2002}, configuration problems~\cite{Felfernig2004213}, hardware designs~\cite{Friedrich1999}, constraint satisfaction problems and spreadsheets~\cite{Abreu2012}. Given a description (model) of a system, together with an observation of the system's behavior which conflicts with the intended behavior of the system, the task of MBD is to find those components of the system (a diagnosis) which, when assumed to be functioning abnormally, provide an explanation of the discrepancy between the intended and the observed system behavior. 
Translated to the setting of KB debugging, the set of ``system components'' comprises the formulas $\tax_i$ in the given faulty KB $\mo$. The ``system description'' refers to the statement that the KB $\mo$ along with the background KB $\mb$ and the positive test cases $\tp\in\Tp$
must meet all predefined requirements (e.g.\ consistency, coherency) and must not logically entail any of the negative test cases $\tn \in \Tn$, i.e.\ 
\begin{enumerate}[(i)]
	\item $\mo \cup \mb \cup \bigcup_{\tp\in\Tp} \tp$ satisfies requirement $r$ for all $r\in\RQ$ and 
	\item $\mo \cup \mb \cup \bigcup_{\tp\in\Tp} \tp \not\models \tn$ for all $\tn \in \Tn$.
\end{enumerate}
The ``observation which conflicts with the intended behavior of the system'' corresponds to the finding that (i) or (ii) or both are violated. That is, the ``system description'' along with the ``observation'' and the assumption that all components are sound yields an inconsistency. An ``explanation for the discrepancy between observed and intended system behavior'' (i.e.\ a diagnosis) is the assumption $\md$ that all formulas in a subset $\md$ of $\mo$ are faulty (``behave abnormally'') and all formulas in $\mo \setminus \md$ are correct (``do not behave abnormally'') such that the ``system description'' along with the ``observation'' and the assumption $\md$ is consistent.
Computation of (minimal) diagnoses is accomplished 
with the aid of \emph{minimal conflict sets}, i.e.\ irreducible sets of formulas in the KB $\mo$ that preserve the violation of (i) or (ii) or both.

\label{etc:MBD_problem_is_abduction_problem} An MBD problem can be modeled as an abduction problem~\cite{Bylander1991}, i.e.\ finding an explanation for a set of data. It was proven in~\cite{Bylander1991} that the computation of the first explanation (minimal diagnosis) is in $\mathrm{P}$. However, given a set of explanations (minimal diagnoses) it is $\mathrm{NP}$-complete to decide whether there is an additional explanation (minimal diagnosis). Stated differently, the detection of the first explanation can be efficiently accomplished whereas the finding of any further one is intractable (unless $\mathrm{P} = \mathrm{NP}$). When seeing the (interactive) KB debugging problem as an abduction problem, one must additionally take into account the costs for reasoning. Because, a call to a logical reasoner is required in order to decide whether or not a set of hypotheses (a subset of the KB) is an explanation (minimal diagnosis). Incorporating the necessary reasoning costs and assuming consistency a minimal requirement to the correct KB, the finding of the first explanation (minimal diagnosis) is already $\mathrm{NP}$-hard even for propositional KBs~\cite{Selman1989} (since propositional satisfiability checking is $\mathrm{NP}$-complete). The worst case complexity for the debugging of KBs formulated over more expressive logics such as OWL 2 (reasoning is 2-$\mathrm{NExpTime}$-complete~\cite{Grau2008a,Kazakov2008}) will be of course even worse. This seems quite discouraging. However, we have shown in our previous works~\cite{Rodler2013, Shchekotykhin2012, Shchekotykhin2014} that for many real-world KBs interactive KB debugging is feasible in reasonable time, despite high (or intractable) worst case reasoning costs and the intractable complexity of the abduction (i.e.\ minimal diagnosis finding) problem as such. Hence, the goal of this work is amongst others to present algorithms that work well in many \emph{practical} scenarios.

\paragraph{Assumptions about the Interacting User.} About a user $u$ consulting an (interactive) debugging system, we 
make the following plausible assumptions:
\begin{enumerate}[U1]
	\item $u$ is not 
	able to explicitly enumerate a set of logical formulas that express the intended domain that should be modeled in a satisfactory way, i.e.\ without unwanted entailments or non-fulfilled requirements, 
	\item $u$ is able to answer concrete queries about the intended domain that should be modeled, i.e.\ $u$ can classify a given logical formula (or a conjunction of logical formulas) as a wanted or unwanted proposition in the intended domain (i.e.\ an entailment or non-entailment of the correct domain model). 
\end{enumerate}
The first assumption is obviously justified since otherwise $u$ could have never obtained a faulty KB, i.e.\ a KB that violates at least one requirement or test case, and there would be no need for $u$ to employ a debugging system. 

Regarding the second assumption, the first thing to be noted is that any KB (i.e.\ any model of the intended domain) either does entail a certain logical formula $\tax$ or it does not entail $\tax$. Second, if $u$ is assumed to bring along enough expertise in that domain, $u$ should be able to gauge the truth of (at least) some formulas about that domain, especially if these formulas constitute logical entailments of parts of the specified knowledge in KB so far. We want to emphasize that $u$ is not required to be capable of answering all possible queries (or formulas) about the respective domain since $u$ might always skip a particular query in our system without any noticeable disadvantages. In such a case, the system keeps generating further queries, one at a time (usually the next-best one according to some quality measure for queries), until $u$ is ready to answer it. As the number of possible queries is usually exponential in the number of minimal diagnoses exploited to compute it, there will be plenty of different ``surrogate queries'' in most scenarios.

\paragraph{A Motivating Example.} To get a more concrete idea of these assumptions, the reader is invited to think about whether the following first-order KB $\mo$
is consistent (a similar example is discussed in \cite{Horridge2009}):
\begin{align}
\label{ex0:s1}&\forall X (res(X) \leftrightarrow \forall Y (writes(X,Y) \rightarrow paper(Y))) \\
\label{ex0:s2}&\forall X ((\exists Y writes(X,Y)) \rightarrow res(X)) \\
\label{ex0:s3}&\forall X (secr(X) \rightarrow gen(X)) \\
\label{ex0:s4}&\forall X (gen(X) \rightarrow \lnot res(X)) \\
\label{ex0:s5}&secr(pam)
\end{align}
If we assume that the predicate symbols $res$, $secr$ and $gen$ stand for 'researcher', 'secretary' and 'general employee', respectively, and the constant $pam$ stands for the person Pam, the KB says the following:
\begin{itemize}
	\item Formula~\ref{ex0:s1}: ``Somebody is a researcher if and only if everything they write is a paper.''
	\item Formula~\ref{ex0:s2}: ``Everybody who writes something is a researcher.''
	\item Formula~\ref{ex0:s3}: ``Each secretary is a general employee.''
	\item Formula~\ref{ex0:s4}: ``No general employee is a researcher.''
	\item Formula~\ref{ex0:s5}: ``Pam is a secretary.''
\end{itemize}
This KB is indeed inconsistent. The reader might agree that it is not very easy to understand why this is the case. The observations made in \cite{Horridge2009} concerning a slight modification $\mo'$ of the KB $\mo$ extracted from a real-world KB confirm this assumption. Compared to $\mo$, the KB $\mo'$ included only Formulas~\ref{ex0:s1}-\ref{ex0:s3} of $\mo$, was formulated in DL (cf.\ Section~\ref{sec:DL}), and used the terms $A,C,\dots$ instead of $res, paper, \dots$. 
Amongst others, this KB $\mo'$ was used as a sample KB in a study where participants had to find out whether a concrete given formula is or is not entailed by a concrete given KB. In the case of  the KB $\mo'$, the assignment (translated to the terminology in our KB $\mo$) was to find out whether $\forall X (secr(X) \rightarrow res(X))$ is an entailment of formulas~\ref{ex0:s1}-\ref{ex0:s3}. Although $\mo'$ contains only three formulas, the result was that even participants with many years of experience in DL, among them also DL reasoner developers, did not realize that this is in fact the case (the reason for this entailment to hold is that formulas~\ref{ex0:s1}-\ref{ex0:s3} imply that $\forall X\, res(X)$ holds).      

Since $\forall X\, res(X)$ is also necessary for the inconsistency of $\mo$, this suggests that people might also have severe difficulties in comprehending why $\mo$ is inconsistent. Once the validity of this entailment is clear, it is relatively straightforward to see that $\mo$ cannot have any models. For, $res(pam)$ (due to $\forall X\, res(X)$) and $\lnot res(pam)$ (due to formulas~\ref{ex0:s3}-\ref{ex0:s5}) are implications of $\mo$.

Consequently, we might also assume that even experienced knowledge engineers (not to mention pure domain experts) could end up with a contradictory KB like $\mo$, which substantiates our first assumption (U1) about $u$. Probably, the intention of those people who specified formulas~\ref{ex0:s1}-\ref{ex0:s3} was not that $\forall X\, res(X)$ should be entailed. That is, it might be already a too complex task for many people to (mentally) reason even with such a small KB like this and manually derive implicit knowledge from it. 

However, on the other hand, we might well assume $u$ to be able to answer a concrete query about the intended domain they tried to model by $\mo$. For instance, one such query could be whether $Q_1 := \setof{\forall X\, res(X)}$ is a desired entailment of their model (i.e.\ ``should everybody be a researcher in your intended model of the domain?''). If we assume the (seemingly obvious) case that $u$ negates this query, i.e.\ asserts that this is an unwanted entailment, then an interactive debugging system (employing a logical reasoner) can derive that at least one of the formulas~\ref{ex0:s1} and \ref{ex0:s2} must be faulty. This holds because the only set-minimal explanation in terms of formulas in $\mo$ for the entailment $\forall X\, res(X)$ is given by these two formulas. In other words, the set of formulas $\setof{\ref{ex0:s1},\ref{ex0:s2}}$ is the only minimal conflict set in $\mo$ given that $Q_1$ is a negative test case.
Hence, the deletion (or suitable modification) of any of these formulas will break this unwanted entailment.

Before it is known that $Q_1$ must not be entailed by the correct KB, given consistency is the only requirement to the KB postulated by $u$, the complete KB $\mo$ is a minimal conflict set. That is, after the assignment of a (strategically well-chosen) query to the set of positive or, in this case, negative test cases can already shift the focus of potential modifications or deletions to a subset of only two candidate formulas. We would call these two formulas the remaining minimal diagnoses after an answer to the query $Q_1$ has been submitted. 

Initially, there are five minimal diagnoses, each formula in $\mo$ is one. The meaning of a diagnosis is that its deletion from $\mo$ leads to the fulfillment of all requirements and (so-far-)specified positive and negative test cases. As the reader should be easily able to see, the deletion of any formula from $\mo$ yields a consistent KB; e.g.\ removing formula~\ref{ex0:s5} prohibits the entailment $\lnot res(pam)$ whereas discarding formula~\ref{ex0:s2} prohibits the entailment $res(pam)$. The reader should notice that, as soon as the negative test case $Q_1$ is known, removing (only) formula~\ref{ex0:s5} does not yield a correct KB since $\setof{\ref{ex0:s1},\ref{ex0:s2},\ref{ex0:s3},\ref{ex0:s4}}$ still entails $Q_1$ which must not be entailed. 

A second query to $u$ could be, for example, 
$Q_2 : \setof{\exists X ((\exists Y writes(X,Y)) \land \lnot res(X))}$ (i.e.\ ``is there somebody who writes something, but is no researcher?''). 
Again, it is reasonable to suppose that $u$ might know whether or not this should hold in their intended domain model. The (seemingly obvious) answer in this case would be positive, e.g.\ because $u$ intends to model students who write homework, exams, etc., but are no researchers. 
This positive answer leads to the new positive test case $Q_2$. Adding this positive test case, like a set of new formulas, to the KB $\mo$ would result in $\mo_{new} := \mo \cup Q_2$. The debugging system would then figure out that formula~\ref{ex0:s2} is the only minimal conflict set in the KB $\mo_{new}$. The reason for this is that the elimination of formula~\ref{ex0:s2} breaks the entailment $Q_1$ (negative test case) and enables the addition of a new desired entailment $Q_2$ (positive test case) without involving the violation of any requirements (consistency). Therefore, formula~\ref{ex0:s2} is the only minimal diagnosis that is still compliant with the new knowledge in terms of $\Q_1 = \false$ and $Q_2 = \true$ obtained. 

It is important to notice that the solution KB $\mo_{new}$ that is returned to the user as a result of the interactive debugging session includes a new logical formula $Q_2$ that can be seen as a repair of the deleted formula~\ref{ex0:s2}. Since the knowledge after the debugging session is that $\lnot \ref{ex0:s2} \equiv Q_2$ must be true, this new knowledge is incorporated into the KB $\mo_{new}$. This indicates that the fault in KB was simply that the $\lnot$ in front of formula~\ref{ex0:s2} had been forgotten. 

Notice however that the positive test case $Q_2$ is not added to $\mo$ as a usual KB formula, but rather as an \emph{extension} of $\mo$ that has already been approved by the user. 
Should the user at some later point in time commit the same fault again (and explicitly specify some formula $x$ equivalent to formula~\ref{ex0:s2}), then the interactive debugging system, owing to the positive test case $Q_2$, would immediately detect a singleton conflict comprising only formula $x$. As a consequence, each diagnosis considered during this later debugging session would suggest to delete or modify (at least) $x$. 

This scenario should illustrate that, in spite of not being able to specify their domain knowledge in a logically consistent way, the user $u$ might still be able to answer questions about the intended domain, which supports our second assumption made about the user $u$ (the reader might agree that answering $Q_1$ and $Q_2$ is much easier than recognizing the entailment $\forall X\, res(X)$ of the KB). In other words, the availability of an (efficient) debugging system could help $u$ debug their KB, without needing to analyze \emph{which} entailments hold or do not hold, \emph{why} certain entailments hold or do not hold or \emph{why} exactly the KB does not meet certain imposed requirements or test cases, by simply answering queries \emph{whether} a certain entailment \emph{should} or \emph{should not} hold. These queries are automatically generated by the system in a way that they focus on the problematic parts of the KB, i.e.\ the minimal conflict sets, and discriminate between the possible solution candidates, i.e.\ the minimal diagnoses. 

\paragraph{Benefits of the Usage of Conflict Sets.} We want to remark that the usage of minimal conflict sets ``naturally'' forces the system to take into consideration only the smallest relevant (faulty) parts of the problematic KB. This is owed to the property of minimal conflict sets to abstract from what \emph{all} the reasons for a certain entailment or requirements violation are. Instead, only the ``root'' (subset-minimal) causes for such violations are examined and no computation time is wasted to extract ``purely derived'' causes (those which are resolved as a byproduct of fixing all root causes from which it is derived, cf.\ \cite{Horridge2011a,Kalyanpur2006a}). For example, assuming the debugging scenario involving our example KB consisting only of formulas \ref{ex0:s1}-\ref{ex0:s4} which is incoherent and a requirements set including coherency. Then, there are two entailments reflecting the incoherency of this KB, first $\forall X\, \lnot secr(X)$ and second $\forall X\, \lnot gen(X)$ (these entailments hold due to $\forall X\, res(X)$ which follows from formulas~\ref{ex0:s1} and \ref{ex0:s2}). Of these two, only the second one is a ``root'' problem; the first one is a ``purely derived'' problem. That means, the entailment $\forall X\, \lnot secr(X)$ only holds due to the presence of the entailment $\forall X\, \lnot gen(X)$. So, the cause for $\forall X\, \lnot gen(X)$ is given by the set of formulas $\setof{\ref{ex0:s1},\ref{ex0:s2},\ref{ex0:s4}}$ whereas the proper superset $\setof{\ref{ex0:s1},\ref{ex0:s2},\ref{ex0:s3},\ref{ex0:s4}}$ of this set accounts for the entailment $\forall X\, \lnot secr(X)$. The exploitation of minimal conflict sets (the only minimal conflict set for this KB is $\setof{\ref{ex0:s1},\ref{ex0:s2},\ref{ex0:s4}}$) ascertains that such ``purely derived'' causes of requirements or test case violations will not be considered at all.  

\paragraph{The Ability to Incorporate Background Knowledge.} Another feature of the approaches described in this work is their ability to incorporate relevant additional information in terms of a background knowledge KB $\mb$ (which is regarded to be correct). $\mb$ is a (consistent) KB which is usually semantically related with the faulty KB, e.g.\ $\mb$ represents knowledge about the domain modeled by $\mo$ that has already been sufficiently endorsed by domain experts. For instance, a doctor who wants to express their knowledge of dermatology in terms of a KB might resort to an approved background KB that specifies the human anatomy. Taking this background information into account puts the problematic KB into some context with existing knowledge and can thereby help a great deal to restrict the search space for solutions of the (interactive) KB debugging problem. This has also been found in~\cite{Stuckenschmidt2008}. This useful strategy of prior search space restriction is also exploited in the field of ontology matching\footnote{http://www.ontologymatching.org/} where automatic systems are employed to generate an alignment, i.e.\ a set of correspondences between semantically related entities of two different ontologies (KBs). Here, both ontologies are considered correct and diagnoses are only allowed to include elements of the alignment~\cite{Meilicke2007}. 

Applying a strategy like that to our example KB given above, supposing that we know that Pam is not a researcher in the world the KB should model, we might specify the background KB $\mb := \setof{\lnot res(pam)}$ prior to starting the interactive debugging session. This would immediately reduce the initial set of possible minimal diagnoses from five (i.e.\ the entire KB) to two (i.e.\ the first two formulas \ref{ex0:s1} and \ref{ex0:s2}). Reason for this is that the entailment $\forall X\, res(X)$ of formulas \ref{ex0:s1} and \ref{ex0:s2} already conflicts with the background knowledge $\lnot res(pam)$.    


\begin{figure*}[t]
	\centering
		\includegraphics[width=0.85\textwidth]{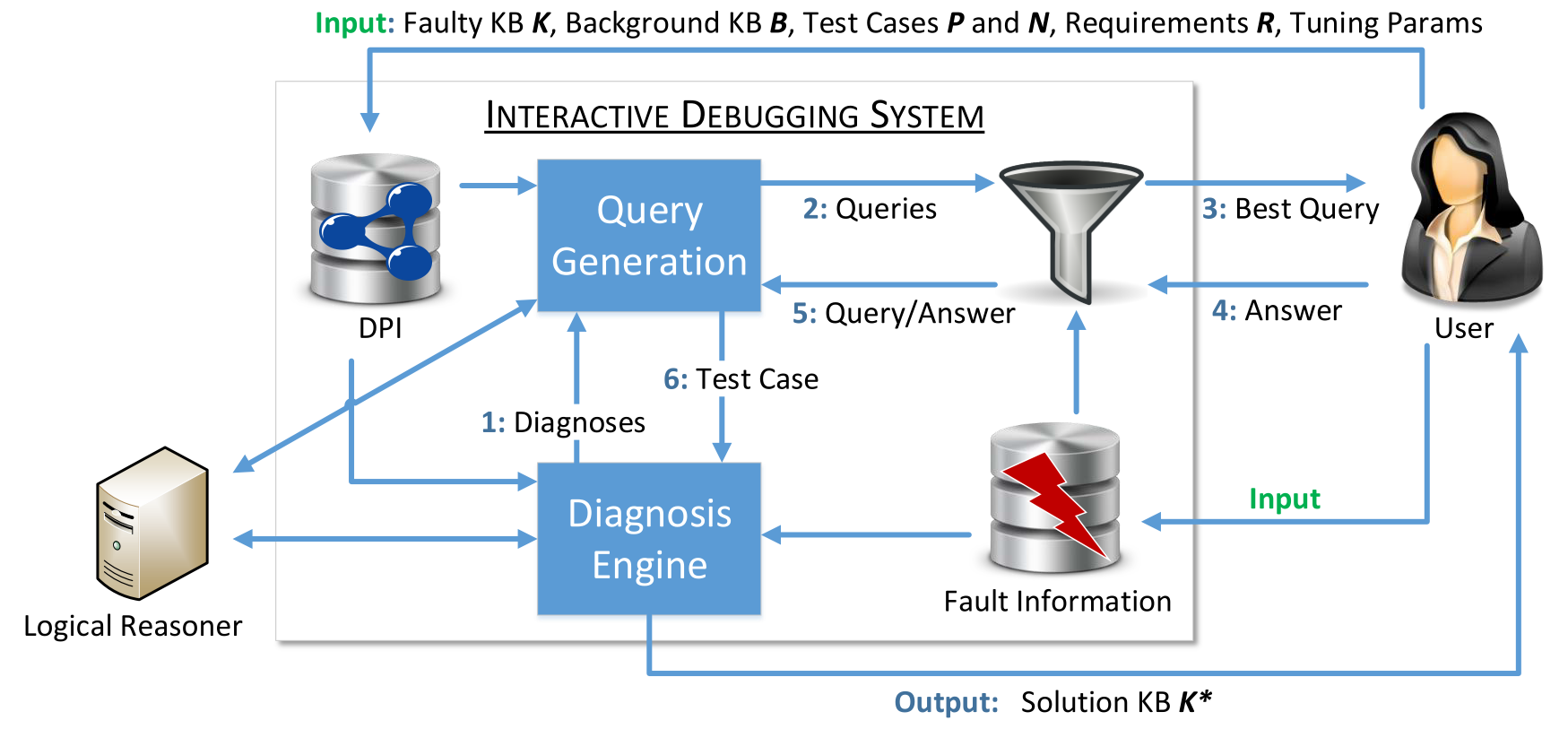}
	\caption[The Principle of Interactive KB Debugging]{The principle of interactive KB debugging.}
	\label{fig:interactive_debugging_workflow}
\end{figure*}

\paragraph{Outline of an Interactive KB Debugging System.} The schema of an interactive debugging system is pictured by Figure~\ref{fig:interactive_debugging_workflow}\footnote{Thanks to Kostyantyn Shchekotykhin for making available to me parts of this diagram.}. As in the case of a non-interactive debugging system (see above), the system receives as input a \emph{diagnosis problem instance (DPI)}.
Further on, a range of additional parameters might be provided to the system. These serve as a means to fine-tune the system's behavior in various aspects. 
Hence, we call these inputs \emph{tuning parameters}. These are (roughly) explained next.

First, some parameters might be specified that take influence on the number of leading diagnoses used for query generation and the necessary computation time invested for leading diagnoses computation. Moreover, some parameter determining the quantity of (pre-)generated queries (of which one is selected to be asked to the user) versus the reaction time (the time it takes the system to compute the next query after the current one has been answered) of the system can be chosen. A further input argument is a query selection measure constituting a notion of query ``goodness'' that is employed to filter out the ``best'' query among the set of generated queries. To give the system a criterion specifying when a solution of the interactive KB debugging problem is ``good enough'', the user is allowed to define a fault tolerance parameter $\sigma$. The lower this parameter is chosen, the better the (possibly ``approximate'') solution that is guaranteed to be found. In case of specifying this parameter to zero, the system will (if feasible) return the ``exact'' solution of the interactive KB debugging problem. Roughly, the exact solution is given in terms of a solution KB obtained by means of a \emph{single} solution candidate (minimal diagnosis) that is left after a sufficient number of queries have been answered (and added to the test cases). On the contrary, an approximate solution is represented by a solution KB obtained by means of a solution candidate with sufficiently high probability (where ``sufficiently high'' is determined by $\sigma$) at some point where there are still multiple solution candidates available. 

Finally, the user may choose between two different modes ($static$ or $dynamic$) of determining the leading diagnoses. The $static$ diagnosis computation strategy guarantees a constant ``convergence'' towards the exact solution by ``freezing'' the set of solution candidates at the very beginning and exploiting answered queries only for the deletion of minimal diagnoses. A possible disadvantage of this approach is the lack of efficient pruning of the used search tree. On the other hand, the $dynamic$ method of calculating leading diagnoses has a primary focus on the preservation of a search tree of small size, thereby aiming at being able to solve diagnosis problem instances which are not soluble by the $static$ approach due to high time and (more critically) space complexity. To this end, more powerful pruning rules are applied in this case which do not permit the algorithm to consider only a fixed set of solution candidates. Rather, the set of minimal diagnoses and minimal conflict sets are generally variable in this case which means that they are subject to change after assignment of an answered query to the test cases.    

Like in the case of a non-interactive debugger, an interactive debugging system requires a sound and complete logical reasoner for deciding consistency (coherency) and calculating logical entailments of a KB formulated over the language $\mathcal{L}$. 

The workflow in interactive KB debugging illustrated by Figure~\ref{fig:interactive_debugging_workflow} is the following:
\begin{enumerate}
	\item A set of leading diagnoses is computed by the diagnosis engine (by means of the fault information, if available) using the logical reasoner and passes it to the query generation module.
	\item The query generation module computes a pool of queries exploiting the set of leading diagnoses and delivers it to the query selection module.
	\item The query selection module filters out the ``best query'' (often by means of the fault information, if available) and shows it to the interacting user.
	\item The user submits an answer to the query.
	\item The query along with the given answer is used to formulate a new test case.
	\item This new test case is transferred back to the diagnosis engine and taken into account in prospective iterations. If the stop criterion (as per $\sigma$, see above) is not met, another iteration starts at step 1. Otherwise, the solution KB $\ot$ constructed from the currently most probable minimal diagnosis is output.
\end{enumerate}

\paragraph{Contributions of this Work.} The contributions of this work are the following:
\begin{itemize}
	\item This work evolves the theory of interactive KB debugging (for monotonic KBs) in a detailed fashion by presupposing a reader to have only some basic knowledge of propositional and first-order logic. To the best of our knowledge, this work provides the most comprehensive and detailed introduction to the field of interactive debugging of (monotonic) KBs. Our previous works on the topic~\cite{Shchekotykhin2012,ksgf2010,Rodler2013,friedrich2005gdm,Shchekotykhin2014} are more application-oriented and thus abstract from some details and omit some of the proofs in favor of comprehensive evaluations of the presented strategies.
	\item This is the first work that gives formal and precise problem statements of the problems addressed in interactive KB debugging and introduces methods that can be proven to solve these problems.
	\item An in-depth discussion of query computation including computational complexity considerations together with an accentuation of potential ways of improving these methods is given. The investigated methods for query computation have been used in~\cite{Shchekotykhin2012,Rodler2013,ksgf2010,Shchekotykhin2014} too, but have not been addressed in depth in these works.
	\item We discuss various ways of exploiting diverse sources of meta information in the KB debugging process from which diagnosis probabilities can be extracted.
	\item We give a formal proof of the soundness of an algorithm $\scQX$ (based on~\cite{junker04}) for the detection of a minimal conflict set in a KB and we show the correctness (completeness, soundness, optimality) of a hitting set tree algorithm \textsc{HS} (based on~\cite{Reiter87}) for finding minimal diagnoses in a KB in best-first order (i.e.\ most probable diagnoses first) which uses $\scQX$ for conflict set computation only on-demand. We are not aware of any other work that comprises such proofs. 
	\item We establish the theoretical relationship between the widely-used notions of a conflict set and a justification. The former is i.a.\ used in~\cite{dekleer1987,Reiter87,Shchekotykhin2012,Rodler2013} and the latter i.a.\ in~\cite{Horridge2008,Horridge2009,Horridge2010,Horridge2011a,Horridge2011b,Horridge2012,Suntisrivaraporn2008,Kalyanpur2006a,MeilickeStuck2009,SattlerSZ09,Nikitina2011}. As a consequence, empirical results concerning the one might be translated to the other. For instance, since each minimal conflict set is a subset of a justification and there is an efficient (polynomial) method for computing a minimal conflict set given a superset of a minimal conflict set, a result manifesting the efficiency of justification computation for a set of KBs (e.g.\ \cite{Horridge2012b}) implies the efficiency of conflict set computation for the same set of KBs. Moreover, we argue that minimal conflict sets are the better choice for our system since these put the focus of the debugger only on the smallest faulty subsets of the KB whereas justifications are better suited in scenarios where exact explanations for the presence of certain entailments are sought.
	\item Two new algorithms for iterative (leading) diagnosis computation in interactive KB debugging are proposed. One that is guaranteed to reduce the number of remaining solutions after a query is answered and one that features more powerful pruning techniques than our previously published algorithms~\cite{Shchekotykhin2012,Rodler2013} (an evaluation that compares the overall efficiency of our previous algorithms with the ones proposed in this work must still be conducted and is part of our future research).
\end{itemize}

\paragraph{Organization of this Work.} The rest of this work is organized as follows: 

In Chapter~\ref{chap:basics}, besides introducing the notation used in this work, we describe the requirements imposed on the logic $\mathcal{L}$ 
that might be used with our approaches, namely monotonicity, idempotency as well as extensiveness. It should be noted that the postulation of these properties does not restrict the applications of our approaches very much. For instance, these might be employed to resolve over-constrained constraint satisfaction problems (CSPs) or repair faulty KBs in PL, FOL, DL, datalog or OWL. Since DL provides the logical underpinning of OWL which has recently received increasing attention due to the extensive research in the field of the Semantic Web, we will also give a short introduction to DL. For, to underline the flexibility of the presented theory in this work, we will illustrate how it can be applied to examples involving PL, FOL as well as DL KBs. 

In Chapter~\ref{chap:OntologyDebugging}, we first give a formal definition of a \emph{KB Debugging} problem and define a diagnosis problem instance (DPI), the input of a KB debugger, and a solution KB, the output of a KB debugger. Further on, we formally characterize a diagnosis and give the notion of KB validity and what it means for a KB to be faulty. We discuss and prove relationships between these notions and specify properties a DPI must satisfy in order to be an admissible (i.e. soluble) input to a KB debugger.  

We motivate why it makes sense to focus on set-minimal diagnoses instead of all diagnoses, i.e.\ to stick to ``The Principle of Parsimony''~\cite{Reiter87,Bylander1991}. This results in the definition of the problem of \emph{Parsimonious KB Debugging}. We prove that solving this problem is equivalent to the computation of minimal diagnoses. Eventually we explain the benefits of using some background KB in (parsimonious) KB debugging. 

In Chapter~\ref{chap:DiagnosisComputation} we describe methods for diagnosis computation. To this end, we first introduce the notion of a (minimal) conflict set, discuss some properties of conflict sets related to the notion of KB validity and give sufficient and necessary criteria for the existence of non-trivial conflict sets w.r.t.\ a DPI. Subsequently, we derive the relationship between a conflict set and the notion of a justification (a minimal set of formulas necessary for a particular entailment to hold) which is well-known and frequently used, especially in the field of DL~\cite{Horridge2008, Horridge2009, Horridge2010, Horridge2011a, Horridge2011b, Horridge2012b}. Concretely, we will demonstrate that a minimal conflict set is a subset of a justification for some negative test case or for some inconsistency (entailment $\false$) or incoherency (entailment $\forall X_1,\dots,X_k\, \lnot p(X_1,\dots,X_k)$ for some predicate symbol $p$ of arity $k$) of the given KB.

Having deduced all relevant characteristics of (minimal) conflict sets, we proceed to give a description of a method ($\scQX$, Algorithm~\ref{algo:qx}) due to~\cite{junker04} which was originally presented as a method for finding preferred explanations (conflicts) in over-constrained CSPs, but can also be employed for an efficient computation of a minimal conflict set w.r.t.\ a DPI in KB debugging. We discuss and exemplify this algorithm in detail, prove its correctness as a routine for minimal conflict set computation and give its complexity.

Having at our disposal a proven sound method for generation of a minimal conflict set, we continue with the delineation of a hitting set tree algorithm similar to the one originally presented in~\cite{Reiter87} which enables the computation of different minimal conflict sets by means of successive calls to $\scQX$, each time given an (adequately) modified DPI. In this manner, a hitting set tree can be constructed (breadth-first) which facilitates the computation of minimal diagnoses (minimum cardinality diagnoses first). We prove the correctness (termination, soundness, completeness, minimum-cardinality-first property) of this hitting set tree algorithm coupled with the $\scQX$ method which serves to solve the problem of parsimonious KB debugging.

In order to be able to incorporate fault information into the diagnoses finding process, we deal with the induction of a probability space over diagnoses in Section~\ref{sec:DiagnosisProbabilitySpace}. We discuss several ways of constructing a probability space including different sources of fault information. 
Hereinafter, we detail how diagnosis probabilities can be determined on the basis of some available fault information and how these can be appropriately updated after new observations (in terms of answered queries) have been made. 
We then outline how fault probabilities can be appropriately incorporated into the hitting set search tree in order to guarantee the discovery of minimal diagnoses in best-first order, i.e.\ most probable ones first. Finally 
we prove the correctness (termination, soundness, completeness, best-first property) of this best-first diagnosis finding algorithm for parsimonious KB debugging.

Section~\ref{sec:non_int_debug_procedure} describes the non-interactive KB debugging procedure (Algorithm~\ref{algo:non_int_debug}) that relies on this best-first diagnosis finding algorithm. Some illustrating examples are provided which at the same time reveal significant shortcomings present in non-interactive KB debugging. This motivates the development of interactive KB debugging algorithms.

Chapter~\ref{chap:InteractiveOntologyDebugging} first states how disadvantages of non-interactive KB debugging procedures can be overcome by allowing a user to take part in the debugging process. We define the problem of \emph{interactive static KB debugging} as well as the problem of \emph{interactive dynamic KB debugging} which ``naturally'' arise from the fact that the DPI in interactive KB debugging is always renewed after a new test case has been specified (a new query has been answered). The former problem searches for a solution KB \emph{w.r.t.\ the DPI given as input} such that this solution KB satisfies all test cases added during the debugging session and there is no other such solution KB. The latter problem searches for a solution KB \emph{w.r.t.\ the current DPI} (i.e.\ the input DPI including all new test cases added throughout the debugging session so far) such that there is no other solution KB w.r.t.\ the current DPI.

Next, in Section~\ref{sec:UserInteraction}, the central term of a \emph{query} is specified which constitutes the medium for user interaction. Queries are generated from a set of \emph{leading diagnoses} which is characterized thereafter. These leading diagnoses are uniquely partitioned into three subsets by each query. The tuple including these subsets is called \emph{q-partition}. Subsequently, the reader is given some explanations how the q-partition can be interpreted, and how it relates to a query. In fact, we will prove that the notion of a q-partition can serve as a criterion for checking whether as set of logical formulas is a query or not. After that, we will learn that a query exists for any set of (at least two) leading diagnoses which grants that the presented algorithms will definitely be able to come up with a query without the need to impose any restrictions on which (minimal) diagnoses are computed by the diagnosis engine in each iteration. 

Section~\ref{sec:QueryGeneration} shows a method for the generation of (a pool of) set-minimal queries (Algorithm~\ref{algo:query_gen}) aiming at stressing the interacting user as sparsely as possible, features in-depth discussions of this method's properties, proves its correctness, provides complexity results and gives some illustrating examples. Further on, drawbacks of this method are pointed out and possible solutions are discussed. 

Subsequently, Section~\ref{sec:WorkflowInInteractiveOntologyDebugging} deals with the presentation of the central algorithm of this work which implements the interactive KB debugger (Algorithm~\ref{algo:inter_onto_debug}). First, this section includes an overview of the workflow of interactive KB debugging (Section~\ref{sec:AlgorithmOverview}), followed by a more comprehensive detailed specification of the algorithm (Section~\ref{sec:DetailedAlgorithmDescription}). Finally, some query selection measures are discussed~\cite{Rodler2013,Shchekotykhin2012} (Section~\ref{sec:query_selection_measures}) and optimization versions of the problems of interactive dynamic and static KB debugging are defined where the goal is to obtain the solution to these problems by asking the user a minimal number of queries. Section~\ref{sec:CorrectnessOfAlgorithmInterOntoDebug} proves the correctness of the interactive KB debugging algorithm and provides a discussion of its complexity.

Chapter~\ref{chap:IterativeDiagnosisComputation} goes into detail w.r.t.\ the two strategies for diagnoses computation introduced in this work that might be plugged into Algorithm~\ref{algo:inter_onto_debug}. Section~\ref{sec:StaticHSTree} describes the $static$ method 
which is a sound and complete method for the iterative computation of minimal diagnoses w.r.t.\ the DPI given as an input to the debugger. In this way, used
as a routine for leading diagnosis computation in Algorithm~\ref{algo:inter_onto_debug}, 
the $static$ method solves the problem of interactive static KB debugging.
Section~\ref{sec:DynamicHSTree} details the $dynamic$ method 
which is a sound and complete method for the iterative computation of minimal diagnoses w.r.t.\ the current DPI, i.e.\ the DPI given as an input to the debugger extended by the information given by all so-far-answered queries. Employed
as a routine for leading diagnosis computation in Algorithm~\ref{algo:inter_onto_debug}, the $dynamic$ method 
solves the problem of interactive dynamic KB debugging.

In Chapter~\ref{chap:RelatedWork} we talk about related work before the concluding Chapter~\ref{chap:conclusion} provides a summary of this work and a discussion of future work topics.

%% file: basic.tex
\chapter{Preliminaries} \label{chap:basics}

\section{Assumptions}
\label{sec:assumptions}
The techniques described in this work are applicable for any logical knowledge representation formalism $\mathcal{L}$ for which 
the entailment relation is
\begin{enumerate}
	\item \label{logic:cond1} \emph{monotonic:} is given when adding a new logical formula to a KB $\mo_{\mathcal{L}}$ cannot invalidate any entailments of the KB, i.e. $\mo_{\mathcal{L}} \models \alpha_{\mathcal{L}}$ implies that $\mo_{\mathcal{L}} \cup \setof{\beta_{\mathcal{L}}} \models \alpha_{\mathcal{L}}$,
	\item \label{logic:cond2} \emph{idempotent:} is given when adding implicit knowledge explicitly to a KB $\mo_{\mathcal{L}}$ does not yield new entailments of the KB, i.e. $\mo_{\mathcal{L}} \models \alpha_{\mathcal{L}}$ and $\mo_{\mathcal{L}} \cup \setof{\alpha_{\mathcal{L}}} \models \beta_{\mathcal{L}}$ implies $\mo_{\mathcal{L}} \models \beta_{\mathcal{L}}$ and
	\item \label{logic:cond3} \emph{extensive:} is given when each logical formula entails itself, i.e. $\{\alpha_{\mathcal{L}}\} \models \alpha_{\mathcal{L}}$ for all $\alpha_{\mathcal{L}}$,
\end{enumerate}
and for which 
\begin{enumerate}
\setcounter{enumi}{3}
	\item \label{logic:cond4} reasoning procedures for \emph{deciding consistency} and \emph{calculating logical entailments} of a KB are available, 
\end{enumerate}
where $\alpha_{\mathcal{L}}, \beta_{\mathcal{L}}$ are logical formulas and $\mo_{\mathcal{L}}$ is a set $\setof{\tax_\mathcal{L}^{(1)},\dots,\tax_\mathcal{L}^{(n)}}$ of logical formulas formulated over the language $\mathcal{L}$. $\mo_{\mathcal{L}}$ is to be understood as the conjunction $\bigwedge_{i=1}^{n} \tax_\mathcal{L}^{(i)}$. Notice that the elements of a KB are called quite differently in literature. Possible denotations are logical formula (e.g.\ \cite{kreuzer2006}), well-formed formula (e.g.\ \cite{chang1973}), (logical) sentence or axiom (e.g.\ \cite{russellnorvig2003}) and axiom (in most of the description logic literature, e.g.\ \cite{Baader2007}). We will mainly stick to the term \emph{formula} (sometimes \emph{axiom}) to refer to the elements of a KB.
As the logic will be clear from the context in the sequel, we will omit the index $\mathcal{L}$ when referring to formulas or KBs over $\mathcal{L}$ throughout the rest of this work.

\section{Considered Logics}
\label{sec:ConsideredLogics}
To underline the general character of this work, we will illustrate our approaches using example diagnosis problem instances expressed in different logical languages. In this section we give notational remarks concerning these different logics used, namely propositional logic (PL), first-order logic (FOL) as well as description logic (DL). Whereas we assume the reader to be familiar with FOL and PL (a good introduction to PL and FOL can be found in~\cite{chang1973}), we will give a short introduction to DL.

\begin{remark}\label{rem:used_logics}
It is important to notice that the usage of DL as well as FOL examples throughout this work should \emph{not} suggest that the Properties \ref{logic:cond1} -- \ref{logic:cond4} stated above are satisfied for any DL or FOL language $\mathcal{L}$. In fact, it is well-known by the theorems of Church and Turing (cf.\ \cite{mendelson2009}; the original works are \cite{church1936,turing1937}) that FOL is not decidable in general, i.e.\ property~\ref{logic:cond4} above is not met. Also in the case of DL, which subsumes a range of different logical languages featuring different expressivity and thus different computational complexity of reasoning procedures, there are languages which are undecidable. For instance, a DL language allowing the formalism of equality role-value-maps which facilitates the expression of concepts like ``persons whose co-workers coincide with their relatives'' can be proven undecidable \cite{Baader2007,schmidt1989}.

Property~\ref{logic:cond4} is satisfied, for example, for the DL language $\mathcal{SROIQ}$ which is the logical underpinning of OWL~2 \cite{Grau2008a}. However, the complexity (2-$\textsc{NExpTime}$-complete~\cite{Kazakov2008}) of logical reasoning is intractable in the worst case for this language which implies the intractability of our methods in the worst case.
Nevertheless, other DL languages applied with similar systems as those described in this paper have been showing reasonable performance~\cite{Shchekotykhin2014, Rodler2013, Shchekotykhin2012}. Also from the theoretical point of view, there are DL languages that allow for efficient reasoning. One example is the OWL 2 EL profile which enables polynomial time reasoning~\cite{baader2005}. For this language, the efficient reasoning service ELK has been presented by~\cite{kazakov2014}.
For FOL, datalog is an example of a decidable sublanguage where reasoning is efficient \cite{russellnorvig2003}. Further, restricted sublanguages of FOL can often be translated to some DL language wherefore DL positive results concerning the decidability of reasoning as well as complexity results can be adopted for these restricted FOL languages~\cite[chapter~4]{Baader2007} \cite{borgida1996}.

Moreover, we want to point out that the practical efficiency of our systems depends strongly on the practical performance (which might be by far better than suggested by the worst case reasoning complexities) of the reasoning services called by our algorithms since the reasoning services are used as a black-box (as mentioned in Chapter~\ref{chap:intro}). 
\qed 
%
%
%
%
\end{remark}

\subsection*{Ontologies and The Semantic Web}
\label{sec:OntologiesAndDescriptionLogic}
Ontologies are KBs that formally and explicitly represent common knowledge about a domain in the form of individuals, concepts (set of individuals) and roles (binary relationships between individuals). As, in the last decade, extensive research has been done in the area of The Semantic Web~\cite{berners2001semantic} making (automatic) ontology development tools and reasoning services more efficient, ontology engineering for the Semantic Web is on the upswing. 
The Semantic Web aims at the enrichment of unstructured information on the web by semantic meta data which should facilitate the usage of the web as structured database of knowledge of all kinds where computers are able to ``understand'' this structured data, establish relationships between different data sources, combine information from different data sources and (most essentially) derive new (implicit) knowledge from the structured data. At this, ontologies are the key to a common vocabulary used for the semantic meta data. Ontologies are employed to precisely define the meaning of different terms, state relationships between different terms and to introduce new terms by means of already specified ones.

The constantly increasing number of people creating ontologies of increasing size (examples were given in Chapter~\ref{chap:intro}) results in more and more (faulty) ontologies which constitute useful application scenarios and test cases for our approaches. For that reason, we also want to use ontology engineering for The Semantic Web as a concrete use case for the presented work. The standard knowledge representation formalism for ontologies is OWL 2~\cite{Motik2009a,Grau2008a} which relies on DL. A short introduction to DL is given next.

\subsubsection*{Description Logic}
\label{sec:DL}
\emph{Description Logic (DL)}~\cite{Baader2007} is a family of knowledge representation languages with a formal logic-based semantics that are designed to represent knowledge about a domain in form of concept descriptions. 
The \emph{syntax} of a description language $\mathcal{L}$ is defined by its signature and a set of constructors. The \emph{signature} of $\mathcal{L}$ corresponds to the union of possibly disjoint sets $N_C$, $N_R$ and $N_I$, where $N_C$ contains all concept names (unary predicates), $N_R$ comprises all role names (binary predicates) and $N_I$ is the set of all individuals (constants) in $\mathcal{L}$. Each concept and role description can be either atomic or complex. The latter ones are composed using constructors defined in the particular language $\mathcal{L}$.
A typical set of DL \emph{constructors} for complex concepts includes conjunction $A \sqcap B$, disjunction $A \sqcup B$, negation $\lnot A$, existential $\exists r.A$ and value $\forall r.A$ restrictions, where $A,B$ are concept descriptions and $r \in N_R$.

\emph{Axioms} are statements of knowledge that must be true in a domain. An \emph{ontology} $\mo$ 
is defined as a tuple $(\mt, \ma)$, where $\mt$ (\emph{TBox}) is a set of terminological axioms and $\ma$ (\emph{ABox}) a set of assertional axioms. Each TBox axiom is expressed by a general concept inclusion $A \sqsubseteq B$, a form of logical implication, or by a definition $A \equiv B$, a kind of logical equivalence, where $A$ and $B$ are concept descriptions or role descriptions. 
ABox axioms are used to assert properties of individuals in terms of the vocabulary defined in the TBox, e.g.\ concept $A(x)$ or role $r(x,y)$ assertions, where $A$ is a concept description, $r$ a role description, and $x,y \in N_I$.

The semantics of a description language is given in terms of \emph{interpretations} $\mi =(\Delta^\mi, \cdot^\mi)$ consisting of a non-empty domain $\Delta^\mi$ and a function $\cdot^\mi$ that assigns to every atomic concept $A \in N_C$ a set $A^\mi \subseteq\Delta^\mi$, to every atomic role $r \in N_R$ a set $r^\mi \subseteq \Delta^\mi \times \Delta^\mi$ and to every individual $x\in N_I$ some value $x^\mi \in \Delta^\mi$. 
The interpretation function is extended to complex concept descriptions by the following inductive definitions: 
\begin{align*}
\top^\mi &= \Delta^\mi\\ 
\bot^\mi &= \emptyset\\ 
(A \sqcap B)^\mi &= A^\mi \cap B^\mi\\ 
(A \sqcup B)^\mi &= A^\mi \cup B^\mi\\ 
(\lnot A)^\mi &= \Delta^\mi \setminus A^\mi\\ 
(\exists r.A)^\mi &= \setof{x\in\Delta^\mi\,|\, \exists y.\, (x,y) \in r^\mi \land y \in A^\mi} \\  
(\forall r.A)^\mi &= \setof{x\in\Delta^\mi\,|\, \forall y.\, (x,y) \in r^\mi \rightarrow y \in A^\mi}
\end{align*}
where $\top$
and $\bot$
are predefined concepts; the former is the universal concept and the latter the bottom concept.

The semantics of axioms is defined as follows for (1) TBox and (2) ABox axioms: 
(1) Interpretation $\mi$ satisfies $A\sqsubseteq B$ iff $A^\mi \subseteq B^\mi$ and it satisfies $A\equiv B$ iff $A^\mi = B^\mi$. (2) $A(x)$ is satisfied by $\mi$ iff $x^\mi \in A^\mi$ and $r(x,y)$ is satisfied iff $(x^\mi,y^\mi) \in r^\mi$. 
An interpretation $\mi$ is a \emph{model} of $\mo = (\mt,\ma)$ iff it satisfies all TBox axioms in $\mt$ and all ABox axioms in $\ma$. An \emph{ontology} $\mo$ \emph{is consistent} iff it has a model. A \emph{concept} $A$ (\emph{role} $r$) \emph{is satisfiable} w.r.t\ $\mo$ iff there is a model $\mi$ of $\mo$ with $A^\mi \neq \emptyset$ ($r^\mi \neq \emptyset$). An \emph{ontology} $\mo$ \emph{is coherent} iff all concepts and roles occurring in $\mo$ are satisfiable. An \emph{axiom $\alpha$ is entailed by $\mo$} iff $\alpha$ is true in all models $\mi$ of $\mo$. For a set of axioms $X$ we write $\mo \models X$ as a shorthand for $\mo \models \alpha$ for all $\alpha\in X$. 

Usually description logic systems provide sound and complete reasoning services to their users. Besides \emph{verification of coherency and consistency of $\mo$} and \emph{satisfiability checking of concepts}, reasoner tasks include classification and realization. \emph{Classification} 
determines, for each concept name $A$ occurring in $\mo$, most specific (general) concepts that subsume (are subsumed by) $A$. A concept $A$ subsumes (is subsumed by) a concept $B$ iff $\mo \models B \sqsubseteq A$ ($\mo \models A \sqsubseteq B$).
Classification is employed to build a taxonomy of concepts in $\mo$.
\emph{Realization}, given an individual name $x$ occurring in $\mo$ and a given set of concepts in $\mo$ (usually all concepts in $\mo$), computes the most specific concepts $A_1,\dots,A_n$ from the set such that $\mo\models A_i(x)$ for all $i=1,\dots,n$. The most specific concepts are those that are minimal w.r.t.\ the subsumption ordering $\sqsubseteq$. 

\begin{example}\label{example:FOL_to_DL}
The example KB given in the Introduction (Chapter~\ref{chap:intro}) can be equivalently represented in DL (cf.\ Remark~\ref{rem:used_logics}) as follows:
\begin{align}
\label{ex0:dl:s1}Res &\equiv \forall writes.Paper \\
\label{ex0:dl:s2}\exists writes.\top &\sqsubseteq Res \\
\label{ex0:dl:s3}Secr &\sqsubseteq Gen \\
\label{ex0:dl:s4}Gen &\sqsubseteq \lnot Res \\
\label{ex0:dl:s5}Secr&(pam)
\end{align}
where $Res$ is the concept symbol with equivalent meaning as the predicate symbol $res$, the role symbol $writes$ corresponds to the equally named binary predicate, $Paper$ to $paper$, and so on. Notice that axiom~\ref{ex0:dl:s2} states that the domain of $writes$ is $Res$. \qed
\end{example}

\section[Notational Remarks]{Notational Remarks
}
\label{sec:NotationalRemarks}

\noindent\emph{General Notational Conventions.} Throughout this work, the nomenclature given by Table~\ref{tab:abbreviations} is used (many of the designators in the table will be explained later in this work). We will mainly refer to an ontology by the term \emph{KB}. 
%
 %

In order to make a clear distinction between scalars and functions, we denote all scalars $g$ by $g$ and all functions $g$ by $g()$. If an ordered list occurs in a set operation, then this list is interpreted as a (non-ordered) set. For example, let $L := [1,3,4,2]$ be an ordered list; then $L \cap \setof{1,2,3}$ yields the \emph{set} $\setof{1,2,3}$.\\

\noindent\emph{Notational Convention for PL (cf.\ \cite{russellnorvig2003}).}
We use uppercase letters $A,B,\dots$ to denote atoms and the standard logical connectives to build PL formulas from atoms. The operator precedence we use is $\lnot$, $\land$, $\lor$, $\rightarrow$, $\leftrightarrow$, from highest to lowest. Given a PL KB $\mo$ and a PL formula $\tax$, we call $\widetilde{\mo}$ and $\widetilde{\tax}$ the \emph{signature of $\mo$} and the \emph{signature of $\tax$}, respectively. The former comprises all atoms occurring in $\mo$ and the latter all atoms occurring in $\tax$.\\

\noindent\emph{Notational Convention for FOL (cf.\ \cite{Ceri1989a}).}
Variables are denoted by uppercase letters; constants and predicate symbols are denoted by strings beginning with a lowercase letter.\footnote{We do not use any function symbols throughout this work.} Recalling the example KB given in Chapter~\ref{chap:intro}, $X, Y$ are variables, $pam$ is a constant and $res$, $writes$, $paper$, $secr$ and $gen$ are predicate symbols. FOL formulas are built from the standard logical connectives described for PL above. The operator precedence we use for FOL formulas is the same as stated above.\footnote{We do not use equality $=$ in FOL formulas throughout this work.} The precedence of quantifiers $\forall$, $\exists$ is such that a quantifier outside of any parenthesized expression holds over everything to the right of it; if occurring in a parenthesized expression, a quantifier holds over everything to the right of it within this expression. For example, 
$\forall X \mathit{prof}(X) \rightarrow \exists Y secr(Y)$ is equivalent to $(\forall X (\mathit{prof}(X) \rightarrow (\exists Y (secr(Y)))))$ (i.e.\ ``for each professor there is at least one secretary'') and not to $(\forall X \mathit{prof}(X)) \rightarrow \exists Y secr(Y)$ (i.e.\ ``if everybody is a professor, then there is at least one secretary''). 

Given a FOL KB $\mo$ and a FOL formula $\tax$, we call $\widetilde{\mo}$ and $\widetilde{\tax}$ the \emph{signature of $\mo$} and the \emph{signature of $\tax$}, respectively. The former comprises all predicate, function and constant symbols occurring in $\mo$ and the latter all predicate, function and constant symbols occurring in $\tax$. The signature of the example KB given in Chapter~\ref{chap:intro} is $\setof{res, writes, paper, secr, gen, pam}$ and the signature of formula~\ref{ex0:s2} of this KB is $\setof{writes,res}$.\vspace{7pt}

\begin{remark}\label{rem:def_incoherency_FOL}
By analogy with the definition of coherency in DL (see Section~\ref{sec:DL}), we call a FOL KB $\mo$ \emph{incoherent} iff $\mo \models \forall X_1,\dots,X_k\, \lnot p(X_1,\dots,X_k)$ for some $k$-place predicate symbol $p$ in the signature of $\mo$ where $k \geq 1$.\qed
\end{remark}
%

\begin{remark}\label{rem:entailment_computation_finite_types_of_entailments}
We want to point out that whenever we will speak of \emph{entailment computation} we address the invocation of a sound reasoning service that is guaranteed to terminate after \emph{finite} execution time and returns a \emph{finite} number of entailments for any KB given as input (cf.\ Remark~\ref{rem:used_logics}). Similarly, when we say that \emph{all entailments} of a KB are computed, we always refer to a \emph{finite} set of entailments of certain types output by such a reasoning service. Examples of such entailment types regarding DL are the (a)~classification and (b)~realization entailments, by which we mean (a)~all the subsumption relationships between concept names appearing in the KB, i.e.\ entailments of the form $C_1 \sqsubseteq C_2$ for concept names $C_1,C_2 \in \widetilde{\mo}$ and (b)~all the concept names instantiated by a given individual for all individuals appearing in the KB, i.e.\ entailments of the form $C(a)$ for concepts names $C \in \widetilde{\mo}$ and individual names $a \in \widetilde{\mo}$.\qed
\end{remark}


\newgeometry{margin=2cm}

\renewcommand{\arraystretch}{1.4}
\begin{table*}[!htbp]
\normalsize
	\centering
	\rowcolors[]{2}{gray!8}{gray!16}
		\begin{tabular}{ll}
		\rowcolor{gray!40}
		\toprule\addlinespace[0pt] 
			Symbol & Meaning \\ \hline
			$2^X$ & the powerset of $X$ where $X$ is a set \\
			$U_X$ & the union of all elements in $X$ where $X$ is a set of sets \\
			$\mathcal{L}$ & a (monotonic, idempotent, extensive) logical knowledge representation language \\
			$\mo_{(i)}$ & a (faulty) KB (optionally with an index)\\
			$\tax_{(i)}$ & a formula in a KB (an axiom in an ontology)\\
			$\mb_{(i)}$ & a (correct) background KB (optionally with an index)\\
			$\Tp$ & the set of positive test cases (each test case is a set of logical formulas) \\
			$\tp_{(i)}$ & a positive test case (optionally with an index) \\
			$\Tn$ & the set of negative test cases (each test case is a set of logical formulas) \\
			$\tn_{(i)}$ & a negative test case (optionally with an index) \\
			$\RQ$ & the set of requirements to the correct KB \\
			$\langle\mo,\mb,\Tp,\Tn\rangle_\RQ$ & a diagnosis problem instance (DPI) \\
			$\allD_{DPI}$ & the set of all diagnoses w.r.t.\ the DPI $DPI$ \\
			$\minD_{DPI}$ & the set of minimal diagnoses w.r.t.\ the DPI $DPI$ \\
			$\md_{(i)}$ & a (minimal) diagnosis (optionally with an index) \\
			$\dt$ & the true diagnosis \\
			$\allC_{DPI}$ & the set of all conflict sets w.r.t.\ the DPI $DPI$ \\
			$\minC_{DPI}$ & the set of minimal conflict sets w.r.t.\ the DPI $DPI$ \\
			$\mc_{(i)}$ & a (minimal) conflict set (optionally with an index) \\
			$\Queue$ & an ordered queue of open nodes in a hitting set tree algorithm \\
			$\mathsf{n}_{(i)},\mathsf{nd}_{(i)},\mathsf{node}_{(i)}$ & nodes in a hitting set tree algorithm (optionally with an index) \\
														& context-dependent (will be clear from the context): \\ 
			\cellcolor{gray!8}		&\cellcolor{gray!8}(1)~an ordered list of the elements $a_1,\dots,a_n$ or\\
			\multirow{-3}{*}{$[a_1,\dots,a_n]$}	& (2)~a (non-ordered) minimal diagnosis comprising formulas $a_1,\dots,a_n$ \\
												& context-dependent (will be clear from the context): \\ 
			\cellcolor{gray!16}	&\cellcolor{gray!16} (1)~a tuple of elements $a_1,\dots,a_n$ or\\
			\multirow{-3}{*}{$\tuple{a_1,\dots,a_n}$}	& (2)~a (non-ordered) minimal conflict set comprising formulas $a_1,\dots,a_n$ \\
			$u$ & the user interacting with the debugging system \\	
			$u()$ & the (user) function that maps queries to answers \\	
			$Q_{(i)}$ & a query (optionally with an index) \\
			$\mQ_{\mD,DPI}$ & the set of all queries w.r.t.\ the leading diagnoses $\mD$ and the DPI $DPI$ \\
			$\Pt(Q)$ & the q-partition of the query $Q$ (abbreviated form) \\
			$\tuple{\dx{}(Q),\dnx{}(Q),\dz{}(Q)}$ & the q-partition of the query $Q$ (written-out form) \\
			$\EX{\md}{DPI}$ & the set of all extensions w.r.t.\ a diagnosis $\md$ and a DPI $DPI$ \\
			$\SO_{DPI}$ & the set of all solution KBs w.r.t.\ the DPI $DPI$ \\
			$\SO^{\max}_{DPI}$ & the set of all maximal solution KBs w.r.t.\ the DPI $DPI$ \\
		\addlinespace[0pt]\bottomrule 	 
		\end{tabular}
		\caption[Symbols and Abbreviations]{Symbols and abbreviations used throughout this work.}
		\label{tab:abbreviations}
\end{table*}

\restoregeometry

\chapter{Knowledge Base Debugging}
\label{chap:OntologyDebugging}
%
KB debugging can be seen as a test-driven procedure comparable to test-driven software development and debugging, where test cases are specified to restrict the possible faults until the user detects the actual fault manually or there is only one (highly probable) fault remaining which is in line with the specified test cases. In this chapter, we want to study the theory of (non-interactive) KB debugging, present and discuss mechanisms that can be employed for the debugging of KBs and reveal drawbacks of such systems. In (non-interactive) KB debugging we assume \emph{test cases fixed during the debugging procedure}. That is, a user might specify a set of test cases offline, run a debugging system and investigate the output solution(s). In case no satisfactory solution has been returned, some additional test cases might be defined offline before the debugger might be invoked again.
 
The inputs to a KB debugging problem can be characterized as follows:
Given is a KB $\mo$ and a KB $\mb$ (background knowledge), both formulated over some logic $\mathcal{L}$ complying with the conditions~\ref{logic:cond1} -- \ref{logic:cond4} given in Chapter~\ref{chap:basics}.  
All formulas in $\mb$ are considered to be correct and all formulas in $\mo$ are considered potentially faulty. $\mo \cup \mb$ does not meet postulated requirements 
$\RQ$ where $\setof{\text{consistency}} \subseteq \RQ \subseteq \setof{\text{coherency, consistency}}$
or does not feature desired semantic properties, called test cases.\footnote{We assume consistency a minimal requirement to a solution KB provided by a debugging system, as inconsistency makes a KB completely useless from the semantic point of view.} Positive test cases (aggregated in the set $\Tp$) correspond to desired entailments and negative test cases ($\Tn$) represent undesired entailments of the correct (repaired) KB (along with the background KB $\mb$). \label{etc:test_cases_are_sets_or_conjuntions_of_formulas} 
Each test case $\tp \in \Tp$ and $\tn \in \Tn$ is \emph{a set of} logical formulas over $\mathcal{L}$. The meaning of a positive test case $\tp \in \Tp$ is that the correct KB integrated with $\mb$ must entail each formula (or the conjunction of formulas) in $\tp$, whereas a negative test case $\tn \in \Tn$ signalizes that some formula (or the conjunction of formulas) in $\tn$ must not be entailed by the correct KB integrated with $\mb$. 
\begin{remark}\label{rem:entailments_as_sets_of_formulas}
In the sequel, we will write $\mo \models X$ for some set of formulas $X$ to denote that $\mo \models \tax$ for all $\tax \in X$ and $\mo \not\models X$ to state that $\mo \not\models \tax$ for some $\tax \in X$.\qed
\end{remark} 
%
The described inputs to the KB debugging problem are captured by the notion of a diagnosis problem instance:
\begin{definition}[Diagnosis Problem Instance]\label{def:dpi}
Let 
\begin{itemize}
	\item $\mo$ be a KB over $\mathcal{L}$,
	\item $\Tp, \Tn$ sets including sets of formulas over $\mathcal{L}$,
	\item $\setof{\text{consistency}}\subseteq \RQ \subseteq \setof{\text{coherency, consistency}}$, 
	\item $\mb$ be a KB over $\mathcal{L}$ such that $\mo \cap \mb = \emptyset$ and $\mb$ satisfies all requirements $r \in \RQ$,
	\item the cardinality of all sets $\mo$, $\mb$, $\Tp$, $\Tn$ be finite.
\end{itemize}
Then we call the tuple $\langle\mo,\mb,\Tp,\Tn\rangle_\RQ$ a \emph{diagnosis problem instance (DPI) over $\mathcal{L}$}.\footnote{
In the following we will often call a DPI over $\mathcal{L}$ simply a DPI for brevity and since the concrete logic will not be relevant in our theoretical analyses as long as it is compliant with the conditions \ref{logic:cond1} -- \ref{logic:cond4} given in Chapter~\ref{chap:basics}. Nevertheless we will mean exactly the logic over which a particular DPI is defined when we use the designator $\mathcal{L}$.
}
\end{definition}
Note that, for now, we do not make any assumptions about the contents of the sets $\mo$, $ \mb$, $\Tp$ and $\Tn$ that go beyond Definition~\ref{def:dpi}.
So, it might be well the case, for example, to specify a DPI according to Definition~\ref{def:dpi} for which there are no solutions or for which only trivial solutions exist. Later on, we will discuss properties a DPI must fulfill to guarantee existence of solutions for it.

We define a solution KB for a DPI as follows:
\begin{definition}[Solution KB]\label{def:target_ont} Let $\langle\mo,\mb,\Tp,\Tn\rangle_\RQ$ be a DPI. Then a KB $\ot$ is called \emph{solution KB w.r.t. $\langle\mo,\mb,\Tp,\Tn\rangle_\RQ$}, written as $\ot \in \SO_{\langle\mo,\mb,\Tp,\Tn\rangle_\RQ}$, iff all the following conditions hold:
\begin{eqnarray}
		 \forall \, r  \in \RQ&:& \;\ot \cup \mb \,\text{ fulfills }\, r  \label{e:1} \\ 
		 \forall \,\tp \in \Tp&:& \;\ot \cup \mb \,\models\, \tp					\label{e:2} \\ 
		 \forall \,\tn \in \Tn&:& \;\ot \cup \mb \,\not\models\, \tn .		\label{e:3}  
\end{eqnarray}
A solution KB $\ot$ w.r.t. a DPI is called \emph{maximal}, written as $\ot \in \SO^{\max}_{\langle\mo,\mb,\Tp,\Tn\rangle_\RQ}$, iff there is no solution KB $\mo'$ such that $\mo' \cap \mo \supset \ot\cap\mo$. 
\end{definition}
Now, the problem of KB debugging can be formalized:\vspace{3pt}

\noindent\fcolorbox{black}{light-gray1}{\parbox[c][2em][c]{0.975\linewidth}{\vspace{-4pt}
\begin{prob_def}[KB Debugging] \label{prob_def:onto_debug}
Given a DPI $\langle\mo,\mb,\Tp,\Tn\rangle_\RQ$, 
find a solution KB w.r.t.\ $\langle\mo,\mb,\Tp,\Tn\rangle_\RQ$.
\end{prob_def}\vspace{-4pt}
}}

\vspace{3pt}
Note that basically any KB $\ot$ that meets conditions~(\ref{e:1}) - (\ref{e:3}) is a solution KB in the sense of Definition~\ref{def:target_ont}. Hence, $\ot$ does not even need to have a non-empty intersection with $\mo$. Only the postulation of maximality of a solution KB (as detailed later in Section~\ref{sec:MinimallyInvasiveOntologyDebugging}) establishes a relationship to the given KB $\mo$.

\begin{remark}\label{rem:reduce_conditions_of_dpi_def}
Let $\mo' := \mo\cup\mb\cup U_{\Tp}$. Then, conditions~(\ref{e:1}) - (\ref{e:3}) can be reduced to conditions~(\ref{e:2}) and (\ref{e:3}) if 
\begin{itemize}
	\item $\Tn := \Tn \cup \setof{\setof{\false}}$ 
given $\RQ = \setof{\mbox{consistency}}$ or
	\item $\Tn := \Tn \cup \{\{\forall X_1,\dots,X_k\, p(X_1,\dots,X_k) \rightarrow\false\}\,|\, p \mbox{ is $k$-place predicate}$ \\
	$ \mbox{symbol in } \widetilde{\mo'}, k \geq 1\} \cup \setof{\setof{\false}}$ 
in case $\RQ = \setof{\mbox{consistency, coherency}}$.
\end{itemize}
This holds because a KB $\mo$ is inconsistent iff $\mo \models \setof{\false}$ and $\mo$ is incoherent iff some predicate symbol in $\mo'$ must be $\false$ for any instantiation. Notice that the latter must hold for all predicate symbols in $\mo'$ and not only in $\mo$ (see Example~\ref{example:illustration_of_rem:reduce_conditions_of_dpi_def}). For PL and DL, the definitions of $\Tn$ are analogous (cf.\ Chapter~\ref{chap:basics}), but for PL coherency is not defined wherefore only the first bullet is relevant for PL. In what follows we will stick to the more explicit characterization of a solution KB given by Definition~\ref{def:target_ont}.\qed
\end{remark}
\begin{example}\label{example:illustration_of_rem:reduce_conditions_of_dpi_def}
Let a DL DPI be defined as 
\begin{align*}
\mo &:= \setof{B \sqsubseteq C} \\
\mb &:= \setof{A \sqsubseteq B, C \sqsubseteq \lnot A} \\
\Tp &:= \emptyset \\
\Tn &:= \emptyset \\
\RQ &:= \setof{\mbox{coherency, consistency}}
\end{align*}
Then, $\widetilde{\mo} = \setof{B,C}$, but there is some concept $A \notin \widetilde{\mo}$, but $A \in \widetilde{\mo'}$, which is unsatisfiable w.r.t.\ $\mo \cup \mb$. Since we want a solution KB \emph{integrated with $\mb$} to meet the conditions~(\ref{e:1}) - (\ref{e:3}), $\mo$ is not a solution KB w.r.t.\ $\langle\mo,\mb,\Tp,\Tn\rangle_\RQ$ despite the fact that it is perfectly consistent and coherent as an isolated KB.\qed 
\end{example}

Whereas the definition of a solution KB refers to the desired properties of the output of a KB debugging system, the following definition can be seen as a characterization of KBs provided as an input to a KB debugger. 
If a KB is valid w.r.t.\ the background knowledge, the requirements and the test cases, then finding a solution KB w.r.t.\ the DPI is trivial. 
Otherwise, obtaining a solution KB from it involves modification of the input KB and subsequent addition of suitable formulas. Usually, the KB $\mo$ part of the DPI given as an input to a debugger is assumed to be invalid w.r.t.\ this DPI.
%
\begin{definition}[Valid KB]\label{def:valid_onto}
Let $\langle\mo,\mb,\Tp,\Tn\rangle_\RQ$ be a DPI.
Then, we say that a KB $\mo'$ is \emph{valid w.r.t.\ $\langle\cdot,\mb,\Tp,\Tn\rangle_\RQ$} iff $\mo' \cup \mb \cup U_\Tp$ does not violate any $r\in\RQ$ and does not entail any $\tn \in \Tn$. A KB is said to be \emph{invalid (or faulty) w.r.t. $\langle\cdot,\mb,\Tp,\Tn\rangle_\RQ$} iff it is not valid w.r.t.\ $\langle\cdot,\mb,\Tp,\Tn\rangle_\RQ$.\footnote{
It would be more precise to call a KB valid \emph{w.r.t.\ the elements $\mb$, $\Tp$, $\Tn$, $\RQ$ of a DPI}. Though, for brevity, we stick to the presented notation where the dot $\cdot$ in $\langle\cdot,\mb,\Tp,\Tn\rangle_\RQ$ signalizes the irrelevance of the first element $\mo$ of a DPI $\langle\mo,\mb,\Tp,\Tn\rangle_\RQ$ for determining validity of a KB $\mo'$ w.r.t.\ this DPI.}
\end{definition} 
Intuitively, if a KB $\mo$ is faulty w.r.t.\ $\langle\cdot,\mb,\Tp,\Tn\rangle_\RQ$, then there is at least one incorrect formula in $\mo$ that needs to be corrected or deleted; 
if a KB $\mo$ is valid w.r.t.\ $\langle\cdot,\mb,\Tp,\Tn\rangle_\RQ$, a solution KB can be \emph{directly} obtained by simply extending $\mo$ by the set $U_\Tp$ of all sentences comprised in positive test cases. 
Note, however, that $\mo$ being valid w.r.t.\ $\langle\cdot,\mb,\Tp,\Tn\rangle_\RQ$ does not necessarily mean that $\mo \cup \mb$ entails any $\tp \in \Tp$. 
\begin{proposition}\label{prop:validonto_targetonto}
Let $\langle\mo,\mb,\Tp,\Tn\rangle_\RQ$ be a DPI. Then, $\mo' \cup U_\Tp \in \SO_{\langle\mo,\mb,\Tp,\Tn\rangle_\RQ}$ 
iff $\mo'$ is valid w.r.t.\ $\langle\cdot,\mb,\Tp,\Tn\rangle_\RQ$.
\end{proposition}
\begin{proof}
``$\Rightarrow$'': If $\mo' \cup U_\Tp$ is a solution KB, then $\mo' \cup U_\Tp \cup \mb$ meets all $r\in\RQ$ as per condition~(\ref{e:1}) and does not entail any $\tn\in\Tn$ as per condition~(\ref{e:3}). Hence, $\mo'$ is valid w.r.t.\ $\langle\cdot,\mb,\Tp,\Tn\rangle_\RQ$.

``$\Leftarrow$'': If $\mo'$ is valid w.r.t.\ $\langle\cdot,\mb,\Tp,\Tn\rangle_\RQ$, then $(\mo' \cup U_\Tp) \cup \mb$ meets all $r\in\RQ$, i.e.\ meets condition~(\ref{e:1}). Moreover, $(\mo' \cup U_\Tp) \cup \mb \not\models \tn$ for all $\tn\in\Tn$, i.e.\ $(\mo' \cup U_\Tp) \cup \mb$ meets condition~(\ref{e:3}). By extensiveness of the used language $\mathcal{L}$, $(\mo' \cup U_\Tp) \cup \mb \models \tp$ for all $\tp\in\Tp$, i.e.\ condition~(\ref{e:2}) is fulfilled by $(\mo' \cup U_\Tp) \cup \mb$. Thus, $\mo' \cup U_\Tp$ is a solution KB.
\end{proof}
\begin{definition}[Extension]\label{def:extension}
Let $\langle\mo,\mb,\Tp,\Tn\rangle_\RQ$ be a DPI over $\mathcal{L}$ and $\mo' \subseteq \mo$. A set of formulas $\me$ over $\mathcal{L}$ is called an \emph{extension w.r.t.\ $\mo'$ and $\langle\mo,\mb,\Tp,\Tn\rangle_\RQ$}, written as $\me \in \EX{\mo'}{\langle\mo,\mb,\Tp,\Tn\rangle_\RQ}$, iff $(\mo\setminus\mo') \cup \me$ is a solution KB w.r.t.\ $\langle\mo,\mb,\Tp,\Tn\rangle_\RQ$.
\end{definition}
\begin{definition}[Diagnosis]\label{def:diagnosis}
Let $\langle\mo,\mb,\Tp,\Tn\rangle_\RQ$ be a DPI. A set of formulas $\md \subseteq \mo$ is called a \emph{diagnosis w.r.t.\ $\langle\mo,\mb,\Tp,\Tn\rangle_\RQ$}, written as $\md \in \allD_{\langle\mo,\mb,\Tp,\Tn\rangle_\RQ}$, iff there exists some $\me \in \EX{\md}{\langle\mo,\mb,\Tp,\Tn\rangle_\RQ}$, i.e.\ $(\mo\setminus\md)\cup \me$ is a solution KB w.r.t.\ $\langle\mo,\mb,\Tp,\Tn\rangle_\RQ$.

A diagnosis $\md$ w.r.t.\ $\langle\mo,\mb,\Tp,\Tn\rangle_\RQ$ is \emph{minimal}, written as $\md \in \minD_{\langle\mo,\mb,\Tp,\Tn\rangle_\RQ}$, iff there is no $\md' \subset \md$ such that 
$\md'$ is a diagnosis w.r.t.\ $\langle\mo,\mb,\Tp,\Tn\rangle_\RQ$. 
A diagnosis $\md$ w.r.t.\ $\langle\mo,\mb,\Tp,\Tn\rangle_\RQ$ is a \emph{minimum cardinality diagnosis w.r.t.\ $\langle\mo,\mb,\Tp,\Tn\rangle_\RQ$} iff there is no diagnosis $\md'$ w.r.t.\ $\langle\mo,\mb,\Tp,\Tn\rangle_\RQ$ such that $|\md'| < |\md|$.
\end{definition}
\begin{proposition}\label{prop:validonto_diag}
Let $\langle\mo,\mb,\Tp,\Tn\rangle_\RQ$ be a DPI. 
Then, $\md \in \allD_{\langle\mo,\mb,\Tp,\Tn\rangle_\RQ}$ 
iff $\mo \setminus \md$ is valid w.r.t.\ $\langle\cdot,\mb,\Tp,\Tn\rangle_\RQ$. 
\end{proposition}
\begin{proof}
``$\Rightarrow$'': If $\md$ is a diagnosis w.r.t.\ $\langle\mo,\mb,\Tp$, $\Tn\rangle_\RQ$, there is some extension $\me$ w.r.t.\ $\md$ and $\langle\mo,\mb,\Tp$, $\Tn\rangle_\RQ$ which implies that $(\mo \setminus \md) \cup \me$ is a solution KB w.r.t.\ $\langle\mo,\mb,\Tp,\Tn\rangle_\RQ$. Now, assume that $\mo \setminus \md$ is not valid w.r.t.\ $\langle\cdot,\mb,\Tp,\Tn\rangle_\RQ$. By Proposition~\ref{prop:validonto_targetonto}, this means that $(\mo \setminus \md) \cup U_\Tp$ is not a solution KB. Hence, $(\mo \setminus \md) \cup U_\Tp \cup \mb$ violates some $r\in\RQ$ or entails some $\tn\in\Tn$. As $(\mo \setminus \md) \cup \me$ is a solution KB, we have that $(\mo \setminus \md) \cup \me \cup \mb \models \tp$ for all $\tp\in\Tp$. So, by idempotency of $\mathcal{L}$, $(\mo \setminus \md) \cup \me \cup \mb \equiv (\mo \setminus \md) \cup \me \cup \mb \cup U_\Tp \supseteq (\mo \setminus \md) \cup U_\Tp \cup \mb$ which violates some $r\in\RQ$ or entails some $\tn\in\Tn$. By monotonicity of $\mathcal{L}$, $(\mo \setminus \md) \cup \me \cup \mb$ also violates some $r\in\RQ$ or entails some $\tn\in\Tn$ whereby $(\mo \setminus \md) \cup \me$ is not a solution KB which is a contradiction.

``$\Leftarrow$'': If $\mo \setminus \md$ is valid w.r.t.\ $\langle\cdot,\mb,\Tp,\Tn\rangle_\RQ$, then $(\mo \setminus \md) \cup \mb \cup U_\Tp$ does not violate any $r\in\RQ$ and does not entail any $\tn \in \Tn$. Since $(\mo \setminus \md) \cup \mb \cup U_\Tp$ also entails each positive test case $\tp \in \Tp$ by extensiveness of $\mathcal{L}$, we can conclude that $(\mo \setminus \md) \cup U_\Tp$ is a solution KB. By Definition~\ref{def:extension}, $U_\Tp \in \EX{\md}{\langle\mo,\mb,\Tp,\Tn\rangle_\RQ}$ and thus $\md$ is a diagnosis w.r.t.\ $\langle\mo,\mb,\Tp,\Tn\rangle_\RQ$.
\end{proof}
In other words, $\md$ is a diagnosis w.r.t.\ $\langle\mo,\mb,\Tp,\Tn\rangle_\RQ$ iff $(\mo \setminus \md) \cup \mb$ meets all requirements, i.e.\ consistency and/or coherency, as per condition~(\ref{e:1}),
does not entail any negative test cases as per condition~(\ref{e:3}), and the positive test cases $\tp \in \Tp$ can be added to $(\mo \setminus \md) \cup \mb$ without violating any of the conditions~(\ref{e:1}) or (\ref{e:3}).

From a given DPI $\langle\mo,\mb,\Tp,\Tn\rangle_\RQ$, a solution KB $\ot$ can be obtained by a deletion and an expansion step. 
The deletion step involves the elimination of a diagnosis $\md \subseteq \mo$ from $\mo$. Note that, due to monotonicity of $\mathcal{L}$, only deletion (and not expansion) of the KB can effectuate a repair of inconsistencies, incoherencies and unwanted entailments. Note, if $\mo$ is already valid w.r.t.\ $\langle\cdot,\mb,\Tp,\Tn\rangle_\RQ$, then $\md$ can be set to $\emptyset$ and the deletion step can be omitted.
%
The expansion step aims at the fulfillment of positive test cases $\Tp$, i.e.\ condition~(\ref{e:2}), which is not necessarily 
the case after the deletion step. In fact, some new logical sentences $\me \in \EX{\md}{\langle\mo,\mb,\Tp,\Tn\rangle_\RQ}$ may need to be added to $(\mo\setminus\md) \cup \mb$ to grant entailment of all positive test cases. 
\begin{corollary}\label{cor:diag_properties}
Let $\md$ be a diagnosis w.r.t.\ $\langle\mo,\mb,\Tp,\Tn\rangle_\RQ$. Then there is a set of logical sentences $\me\in\EX{\md}{\langle\mo,\mb,\Tp,\Tn\rangle_\RQ}$ over $\mathcal{L}$ such that:
\begin{eqnarray*}
		 \forall \, r  \in \RQ&:& \;(\mo\setminus\md) \cup \me \cup \mb \,\text{ fulfills }\, r   \\ 
		 \forall \,\tp \in \Tp&:& \;(\mo\setminus\md) \cup \me \cup \mb \,\models\, \tp					  \\ 
		 \forall \,\tn \in \Tn&:& \;(\mo\setminus\md) \cup \me \cup \mb \,\not\models\, \tn .		  
\end{eqnarray*} 
\end{corollary}
\begin{proof}
The proposition of the corollary is a direct consequence of Definition~\ref{def:target_ont} and Definition~\ref{def:diagnosis}.
\end{proof}
From the point of view of a solution KB $\ot$ w.r.t.\ $\langle\mo,\mb,\Tp,\Tn\rangle_\RQ$, $\mo \setminus \ot$ is a diagnosis w.r.t.\ $\langle\mo,\mb,\Tp,\Tn\rangle_\RQ$ and $\ot \setminus \mo$ is one possible extension w.r.t.\ $\md$ and $\langle\mo,\mb,\Tp,\Tn\rangle_\RQ$.
\begin{proposition}\label{prop:targetonto_diag}
For each solution KB $\ot$ w.r.t.\ $\langle\mo,\mb,\Tp,\Tn\rangle_\RQ$ there is a diagnosis w.r.t.\ $\langle\mo,\mb,\Tp,\Tn\rangle_\RQ$ and an extension $\me$ w.r.t.\ $\md$ and $\langle\mo,\mb,\Tp,\Tn\rangle_\RQ$ such that $\ot = (\mo\setminus\md) \cup \me$ and $\me \cap \md = \emptyset$.
\end{proposition}
\begin{proof}
Let $\ot$ be a solution KB w.r.t. $\langle\mo,\mb,\Tp,\Tn\rangle_\RQ$. Then $\ot$ can be written as $\ot = (\mo \cap \ot) \cup (\ot \setminus \mo) = (\mo \setminus (\mo \setminus \ot)) \cup (\ot \setminus \mo)$. Let $\mo \setminus \ot =: \md$ and $\ot \setminus \mo =: \me$, then $\me \cap \md = \emptyset$. Further on, $\md \subseteq \mo$ holds and $\me$ is a set of logical sentences such that $\ot = (\mo\setminus\md)\cup \me \in \SO_{\langle\mo,\mb,\Tp,\Tn\rangle_\RQ}$. Therefore, $\md \in \allD_{\langle\mo,\mb,\Tp,\Tn\rangle_\RQ}$ and $\me \in \EX{\md}{\langle\mo,\mb,\Tp,\Tn\rangle_\RQ}$.
\end{proof}
\begin{corollary}
The \mbox{(non-)}existence of a diagnosis w.r.t. $\langle\mo,\mb,\Tp,\Tn\rangle_\RQ$ is equivalent to the \mbox{(non-)}existence of a solution KB w.r.t. $\langle\mo,\mb,\Tp,\Tn\rangle_\RQ$.
\end{corollary}
\begin{proof}
Proposition~\ref{prop:targetonto_diag} shows that there is a diagnosis for each solution KB. By Definition~\ref{def:diagnosis}, there is also a solution KB for each diagnosis.
\end{proof}
The next Proposition gives sufficient and necessary criteria for the existence of a solution, i.e.\ a diagnosis or a solution KB, respectively, for a given DPI.
%
\begin{proposition}\label{prop:exist_diag}
Let $\langle\mo,\mb,\Tp,\Tn\rangle_\RQ$ be a DPI. Then, a diagnosis $\md$ w.r.t. $\langle\mo,\mb,\Tp,\Tn\rangle_\RQ$ exists iff 
\begin{itemize}
	\item $\forall \,r\in\RQ \,\;: \; \mb \cup U_P$ fulfills $r$ and
	\item $\forall \,\tn\in\Tn\,: \;\mb \cup U_\Tp \not\models \tn$.
\end{itemize}
\end{proposition}
\begin{proof}

``$\Leftarrow$'': Let us define $\md := \mo$. Then $X := (\mo \setminus \md) \cup \mb \cup U_P = \mb \cup U_P$. Consequently, $X$ satisfies each $r\in\RQ$ as per condition (\ref{e:1}), $X \not\models \tn$ for each $\tn \in \Tn$ as per condition (\ref{e:3}), and finally $X \models \tp$ for each $\tp \in \Tp$ by extensiveness of $\mathcal{L}$ and thus meets condition (\ref{e:2}). So, $X$ is a solution KB w.r.t. $\langle\mo,\mb,\Tp,\Tn\rangle_\RQ$ wherefore $\md$ must be a diagnosis. 

``$\Rightarrow$'': Let $\md \subseteq \mo$ be some diagnosis w.r.t. $\langle\mo,\mb,\Tp,\Tn\rangle_\RQ$. Then, by definition of a diagnosis, there is some solution KB $\ot$ w.r.t. $\langle\mo,\mb,\Tp,\Tn\rangle_\RQ$. Then $\ot \cup \mb \models \tp$ for all $\tp \in \Tp$ by condition~(\ref{e:2}), which implies that $\ot \cup \mb \cup U_\Tp$ does not feature any new entailments compared to $\ot \cup \mb$ by idempotency of $\mathcal{L}$. So, $\ot \cup \mb \equiv \ot \cup \mb \cup U_\Tp$ holds. Now, for arbitrary $\tn \in \Tn$, since $\ot \cup \mb \not\models \tn$ we have that $\ot \cup \mb \cup U_\Tp \not\models \tn$, and, by monotonicity of $\mathcal{L}$, that $\mb \cup U_\Tp \not\models \tn$. Analogously, for any $r\in\RQ$, because $\ot \cup \mb$ satisfies $r$, it must be true that $\ot \cup \mb \cup U_\Tp$ satisfies $r$ and, by monotonicity of $\mathcal{L}$, that $\mb \cup U_\Tp$ satisfies $r$.
\end{proof}
\begin{definition}[Admissible DPI]\label{def:admissible}
We call a DPI $\langle\mo,\mb,\Tp,\Tn\rangle_\RQ$ \emph{admissible} iff there is at least one diagnosis $\md \in \allD_{\langle\mo,\mb,\Tp,\Tn\rangle_\RQ}$.
\end{definition}
A non-admissible DPI may arise in a situation where a user specifies test cases manually. For this procedure a similar error-proneness as for the user's formulation of KB formulas can be assumed. And there are lots of pitfalls to escape, as Proposition~\ref{prop:exist_diag} shows. In particular, 
the specified test cases in $\Tp$ and $\Tn$ must be ``compatible'' with each other, i.e. positive test cases must not contradict negative ones.
For example, adding $\tp_1 := \setof{A \sqsubseteq C, E \equiv B}$ and $\tp_2 := \setof{C \sqsubseteq E}$ to $\Tp$ and $\tn_1 := \setof{A \sqsubseteq B}$ to $\Tn$ leads to a contradiction between $\Tp$ and $\Tn$ and consequently to the non-admissibility of a DPI comprising $\Tp$ and $\Tn$. Furthermore, the background KB $\mb$ which is considered as correct, must indeed be correct, at least in terms of $\RQ$; and negative test cases must be specified in a way not to postulate non-entailment of knowledge specified in $\mb$. A counterexample is $\mb := \setof{\exists r.\top \sqsubseteq A, r(x,y), A \sqsubseteq C}$ and $\Tn := \setof{\setof{C(x)}}$.
And third, the union of positive test cases together with $\mb$ must be in compliance with $\RQ$, particularly the formulas in $\Tp$ must not be inconsistent or incoherent. Because the union of positive test cases $U_\Tp$ can be viewed as an own KB since all logical sentences occurring in some $\tp\in\Tp$ must be true in the solution KB. So, in a setting where test cases are specified manually, faults occur as likely in $U_\Tp$ as they do in $\mo$.

The debugging system presented in this work, however, guarantees by automatic test case generation that 
admissibility of a DPI is satisfied at any time, provided that 
an admissible DPI is given as an initial input to the debugging system. 

\begin{remark}\label{rem:from_non_admissible_DPI_to_admissible_DPI}
In case of a present DPI $\langle\mo,\mb,\Tp,\Tn\rangle_\RQ$ which is non-admissible, the DPI must be properly modified before it can be used with our debugging system. More concretely, the sets $\mb$, $\Tp$ as well as $\Tn$ must be prepared in a way that the two conditions in Proposition~\ref{prop:exist_diag} are satisfied. When supposing that $\mb$ is an already approved and correct KB (which is a reasonable assumption for a KB used as background knowledge during a debugging session), then there are (at least) the following ways to obtain an admissible DPI from a given non-admissible DPI without modifying $\mb$. 

(a)~One straightforward way to achieve that is the deletion of all manually specified test cases from $\Tp$ and $\Tn$. After that, both sets are either the empty set (if no automatic test cases, e.g.\ from former debugging sessions were included in these sets) or comprise only automatically generated test cases. The former case yields an admissible DPI independently of $\mo$ by the property of $\mb$ to not violate any requirements in $\RQ$ (see Definition~\ref{def:dpi}). That the latter case implies the admissibility of the DPI is a property of the debugging system described in this work (as we will show later by Corollary~\ref{cor:query_leaves_valid_diag}).

(b)~Another way to resolve the non-admissibility of a DPI $\langle\mo,\mb,\Tp,\Tn\rangle_\RQ$ is to first check whether $\langle U_{\Tp},\mb,\emptyset,\Tn\rangle_\RQ$ is admissible (verification of Proposition~\ref{prop:exist_diag} by means of a reasoning service). If so, it is clear that $\mb$ does not conflict with $\Tn$. Then, a debugger (like the one presented in this work) can be exploited to find an as small as possible subset of the set of all formulas occurring in the positive test cases, the removal of which causes the DPI to become admissible. This would be accomplished by the computation of a minimal diagnosis $\md_{\Tp}$ w.r.t.\ $\langle U_{\Tp},\mb,\emptyset,\Tn\rangle_\RQ$ and the usage of the modified admissible DPI $\langle\mo,\mb,\setof{U_{\Tp}\setminus\md_{\Tp}},\Tn\rangle_\RQ$ instead of the original one. In this case, only a set-minimal set $\md_{\Tp}$ of formulas that were desired entailments of the user are lost. This modification is possible in polynomial time apart from the reasoning costs, i.e.\ by means of a polynomial number of calls to a reasoner (cf.\ Chapter~\ref{chap:intro}).

(c)~Otherwise, i.e.\ if $\mb$ already conflicts with the negative test cases $\Tn$, then an algorithm similar to Algorithm~\ref{algo:qx} (that will be presented in Section~\ref{sec:cs_comp}) can be employed to determine a maximal subset $\Tn'$ of $\Tn$ w.r.t.\ set inclusion such that $\mb$ will not be in conflict with $\Tn'$. This approach also requires only a polynomial number of calls to a reasoner (cf.\ Proposition~\ref{prop:qx_complexity}). If the resulting modified DPI $\langle\mo,\mb,\Tp,\Tn'\rangle_\RQ$ is not yet admissible, i.e.\ after adding the positive test cases $U_{\Tp}$ to $\mb$ there \emph{are} again conflicts with $\Tn'$, method~(b) must be executed in order to finally obtain an admissible DPI. 

That is, given a non-admissible DPI, there is a transformation achievable in polynomial time which enables the establishment of admissibility involving a set-minimal number of modifications to the given test cases. Thence, in the rest of this work, we will assume that a DPI given as an input to our algorithms is admissible.\qed
\end{remark}

In general, there are multiple (minimal) diagnoses for a DPI, i.e.\ $|\allD_{\langle\mo,\mb,\Tp,\Tn\rangle_\RQ}| \geq$ $|\minD_{\langle\mo,\mb,\Tp,\Tn\rangle_\RQ}|$ $> 1$, and there are multiple, in fact infinitely many, extensions $\me \in \EX{\md}{\langle\mo,\mb,\Tp,\Tn\rangle_\RQ}$ for a fixed diagnosis $\md \in \allD_{\langle\mo,\mb,\Tp,\Tn\rangle_\RQ}$. The task addressed in this work is finding an optimal diagnosis 
for a given DPI, whereas the identification of an \emph{optimal} extension w.r.t.\ that diagnosis and the DPI is not the aim. What we understand by ``optimality'' of a diagnosis will be addressed in more detail in Chapter~\ref{chap:InteractiveOntologyDebugging}. Instead, we will content ourselves with finding any extension that enables to formulate a solution KB given a DPI and a diagnosis for that DPI. In fact, the problem of finding a solution KB for a DPI can be reduced to finding a diagnosis for that DPI since a suitable extension 
can be easily formulated for any diagnosis, as the next proposition shows:
%
\begin{proposition}\label{prop:ex_exist}
Let $\langle\mo,\mb,\Tp,\Tn\rangle_\RQ$ be a DPI and $\md\in\allD_{\langle\mo,\mb,\Tp,\Tn\rangle_\RQ}$. Then $U_\Tp$ is an extension w.r.t.\ $\md$ and $\langle\mo,\mb,\Tp,\Tn\rangle_\RQ$.
\end{proposition}
\begin{proof}
Let us assume that there is some $\md \in \allD_{\langle\mo,\mb,\Tp,\Tn\rangle_\RQ}$ and 
$U_\Tp$ is not an extension w.r.t.\ $\md$ and $\langle\mo,\mb,\Tp,\Tn\rangle_\RQ$.
By the definition of a diagnosis, this is equivalent to stating that $(\mo \setminus \md) \cup U_\Tp$ is not a solution KB which in turn means that at least one condition (\ref{e:1}), (\ref{e:2}) or (\ref{e:3}) of Definition~\ref{def:target_ont} is violated by $(\mo \setminus \md) \cup U_\Tp$. However, the fact that $\md$ is a diagnosis implies the existence of some extension $\me \in \EX{\md}{\langle\mo,\mb,\Tp,\Tn\rangle_\RQ}$ that can be added to $(\mo \setminus \md)$ to obtain a solution KB. This means that conditions (\ref{e:1}) and (\ref{e:3}) must be already valid 
for $(\mo \setminus \md)$, since, by monotonicity of $\mathcal{L}$, addition of logical sentences $\me$ can neither solve inconsistencies or incoherencies necessary for fulfillment of condition (\ref{e:1}) nor invalidate non-desired entailments as per condition (\ref{e:3}). As a consequence, condition (\ref{e:2}) must be violated by $(\mo \setminus \md) \cup U_\Tp$. By extensiveness of $\mathcal{L}$ it holds that $(\mo \setminus \md) \cup U_\Tp \models \tp$ for all $\tp \in \Tp$ whereby we obtain that condition (\ref{e:2}) is fulfilled which yields a contradiction.
\end{proof}
%
Proposition~\ref{prop:ex_exist} claims that the expansion operation, i.e.\ identifying a concrete extension for a diagnosis, is trivial, at least for our purposes, namely formulating an extension reflecting only evident entailments given by the set of positive test cases $\Tp$. Consequently, in order to find a solution KB for some DPI, it is sufficient to concentrate on the deletion step, i.e.\ on the search for diagnoses. 

Note that using $U_\Tp$ as a canonical extension when computing diagnoses does not affect the set of identified diagnoses. In other words, exchanging $\me \in \EX{\md}{\langle\mo,\mb,\Tp,\Tn\rangle_\RQ}$ for $U_\Tp$ in Definition~\ref{def:diagnosis} yields an equivalent definition.
The following corollary proves this statement and summarizes the relationship between the notions \emph{diagnosis}, \emph{solution KB} and \emph{valid KB}.
\begin{corollary}\label{cor:notions_equiv}
The following statements are equivalent:
\begin{enumerate}
\item $\md$ is a diagnosis w.r.t.\ $\langle\mo,\mb,\Tp,\Tn\rangle_\RQ$
\item $(\mo \setminus \md) \cup U_\Tp$ is a solution KB w.r.t.\ $\langle\mo,\mb,\Tp,\Tn\rangle_\RQ$
\item $(\mo \setminus \md)$ is valid w.r.t.\ $\langle\cdot,\mb,\Tp,\Tn\rangle_\RQ$.
\end{enumerate}
\end{corollary}
\begin{proof}
That (1) is equivalent to (2) follows from Definition~\ref{def:diagnosis} which states that $\md$ is a diagnosis w.r.t.\ $\langle\mo,\mb,\Tp,\Tn\rangle_\RQ$ iff there is some set of sentences $\me\in\EX{\md}{\langle\mo,\mb,\Tp,\Tn\rangle_\RQ}$ such that $(\mo \setminus \md) \cup \me$ is a solution KB, and from Proposition~\ref{prop:ex_exist} which proves that $U_\Tp$ is an extension w.r.t.\ \emph{any} diagnosis $\md$ and $\langle\mo,\mb,\Tp,\Tn\rangle_\RQ$. 

That (1) is equivalent to (3) follows directly from Proposition~\ref{prop:validonto_diag} and the equivalence of (2) and (3) has been shown in Proposition~\ref{prop:validonto_targetonto}.
\end{proof}

\section{Parsimonious Knowledge Base Debugging
}
\label{sec:MinimallyInvasiveOntologyDebugging}
Why are \emph{minimal} diagnoses interesting? 
First, the set of minimal diagnoses w.r.t.\ a DPI captures all the information that explains the unwanted properties, i.e.\ violation of requirements or test cases, of the DPI. In other words, the minimal diagnoses represent all subset-minimal possibilities to modify a KB in a way it becomes a valid KB w.r.t.\ the given DPI (e.g.\ by simply deleting a minimal diagnosis from the KB in the trivial case). By monotonicity of the logic $\mathcal{L}$, each superset of a minimal diagnosis w.r.t.\ a DPI is a diagnosis w.r.t.\ this DPI. That is, $\allD_{\langle\mo,\mb,\Tp,\Tn\rangle_\RQ}$ can be easily reconstructed given $\minD_{\langle\mo,\mb,\Tp,\Tn\rangle_\RQ}$.
There is however no evidence (in terms of specified requirements and test cases) in a DPI that would justify the selection of a non-minimal diagnosis. That is, if $\mo$ is a KB and $\md \subseteq\mo$ a minimal diagnosis w.r.t.\ a DPI including $\mo$, $\mo\setminus\md$ does not violate any of the postulated properties that must hold for a KB to be valid w.r.t.\ this DPI. For that reason, there is no \emph{evident} need to delete or modify any other sentences in $\mo$ except for the ones in some \emph{minimal} diagnosis $\md$.
%

Second, usually a setting can be assumed where the author of a KB specifies formulas to the best of their knowledge. Hence, the assumption that a formula is rather correct than faulty, or in other words, that the KB author wants to keep as many formulated sentences as possible in a solution KB obtained from a debugger, is practical. 

This also motivates the importance of a certain subset of minimal diagnoses, namely minimum cardinality diagnoses, which are the solutions of choice in scenarios where no probabilistic information about the KB authors' faults is available, e.g.\ in terms of statistics retrieved from log data of the used IDE (see Section~\ref{sec:DiagnosisProbabilitySpace} for details). In an application where such information is given, minimum cardinality diagnoses might not always be the appropriate choice (for details see Chapter~\ref{chap:InteractiveOntologyDebugging}). In this case the aim is to find a minimal diagnosis with a maximal probability of including only sentences that are actually faulty (which might not necessarily be a minimum cardinality diagnosis).

Third, minimality of diagnoses will be a necessary condition to \emph{guarantee} the possibility of discrimination between different (candidate) diagnoses to formulate a solution KB, as will be seen later in Section~\ref{sec:UserInteraction}.

Fourth, focusing only on minimal diagnoses rather than all diagnoses can greatly reduce the search space for diagnoses and therefore greatly speed up the debugging procedure (cf.~\cite{dekleer1987}).

Projected to the task of KB debugging, namely finding a solution KB w.r.t.\ a given DPI, this means we are interested in minimal invasiveness, that is making as few formula-deletion-modifications to the input KB $\mo$ as possible in the course of the performed debugging actions. That is, the actual goal is to find some \emph{maximal} solution KB $\ot$ for a DPI. Compare with ``The Principle of Parsimony'' in \cite[p.~7]{Reiter87} \cite{Bylander1991}.\vspace{3pt}

\noindent\fcolorbox{black}{light-gray1}{\parbox[c][2em][c]{0.975\linewidth}{\vspace{-4pt}
\begin{prob_def}[Parsimonious KB Debugging] \label{prob_def:evidence_just}
Given a DPI $\langle\mo,\mb,\Tp,\Tn\rangle_\RQ$, the task is to find a maximal solution KB w.r.t.\ $\langle\mo,\mb,\Tp,\Tn\rangle_\RQ$.
\end{prob_def}\vspace{-4pt}
}}

\vspace{3pt}
The next proposition shows that this problem can be reduced to finding a minimal diagnosis. 

\begin{proposition}\label{prop:min_max}

\textbf{(i)} 
$\mo \setminus \ot$ is a minimal diagnosis w.r.t.\ $\langle\mo,\mb,\Tp,\Tn\rangle_\RQ$ for each maximal solution KB $\ot$ w.r.t.\ $\langle\mo,\mb,\Tp,\Tn\rangle_\RQ$. 
 
\textbf{(ii)} If $\md$ is a minimal diagnosis w.r.t.\ $\langle\mo,\mb,\Tp,\Tn\rangle_\RQ$, then $(\mo\setminus\md) \cup \me$ is a maximal solution KB w.r.t.\ $\langle\mo,\mb,\Tp,\Tn\rangle_\RQ$ for all extensions $\me \in \EX{\md}{\langle\mo,\mb,\Tp,\Tn\rangle_\RQ}$.
\end{proposition}
\begin{proof}
\textbf{Ad (i):} Let $\ot$ be an arbitrary maximal solution KB w.r.t.\ $\langle\mo,\mb,\Tp,\Tn\rangle_\RQ$. The first observation is that $\md := \mo \setminus \ot$ is a diagnosis w.r.t.\ $\langle\mo,\mb,\Tp,\Tn\rangle_\RQ$ since 
$\ot \setminus \mo \in \EX{\md}{\langle\mo,\mb,\Tp,\Tn\rangle_\RQ}$ by the fact that $\ot = (\mo \setminus \md) \cup (\ot \setminus \mo)$ is a solution KB by assumption. Let us assume that there is a diagnosis $\md_k \in \allD_{\langle\mo,\mb,\Tp,\Tn\rangle_\RQ}$ such that $\md \supset \md_k$. Since $\md_k$ is a diagnosis, it holds per Definition~\ref{def:diagnosis} that there is an extension $\me \in \EX{\md_k}{\langle\mo,\mb,\Tp,\Tn\rangle_\RQ}$ such that $\ot_k := (\mo \setminus \md_k) \cup \me$ is a solution KB. Further on, $\mo \cap \ot_k = \mo \cap ((\mo \setminus \md_k) \cup \me) = (\mo \setminus \md_k) \cup (\mo \cap \me)$. Since $\mo \cap \ot$ can be written as $\mo \setminus (\mo \setminus \ot) = \mo \setminus \md$ which is a strict subset of $\mo \setminus \md_k$ which in turn is a subset of $(\mo \setminus \md_k) \cup (\mo \cap \me) = \mo \cap \ot_k$. Consequently, $\mo \cap \ot \subset \mo \cap \ot_k$ holds, which is by Definition~\ref{def:target_ont} a contradiction to the maximality of the solution KB $\ot$. Thus, $\md = \mo \setminus \ot$ is a minimal diagnosis w.r.t.\ $\langle\mo,\mb,\Tp,\Tn\rangle_\RQ$.

\textbf{Ad (ii):} Let $\md$ be a minimal diagnosis w.r.t.\ $\langle\mo,\mb,\Tp,\Tn\rangle_\RQ$. Then, by Definition~\ref{def:diagnosis}, there is an extension $\me \in \EX{\md}{\langle\mo,\mb,\Tp,\Tn\rangle_\RQ}$ such that $\ot := (\mo \setminus \md) \cup \me$ is a solution KB. Let us assume that $\me \cap \md \neq \emptyset$. We can rewrite $\ot$ as $\ot = (\mo \setminus \md) \cup (\me \cap \md) \cup (\me \setminus \md)$. Since $\emptyset \subset \me \cap \md \subseteq \md$, we have that $(\mo \setminus \md) \cup (\me \cap \md) \supset \mo \setminus \md$. Thus, there is a $\md' := \md \setminus (\me \cap \md) \subset \md$ 
and an extension $\me' \in \EX{\md'}{\langle\mo,\mb,\Tp,\Tn\rangle_\RQ}$ such that $\me' := \me \setminus \md$ such that $\ot = (\mo \setminus \md') \cup \me'$. As $\ot$ is a solution KB, this is a contradiction to the minimality of $\md$. Therefore, (*) $\me \cap \md = \emptyset$ for all $\me \in \EX{\md}{\langle\mo,\mb,\Tp,\Tn\rangle_\RQ}$ must hold.

Let $\me$ be any extension w.r.t.\ $\md$ and $\langle\mo,\mb,\Tp,\Tn\rangle_\RQ$. Then we can write $\mo \cap \ot = \mo \cap ((\mo \setminus \md) \cup \me) = (\mo \setminus \md) \cup (\mo \cap \me)$ and by (*) also $\mo \cap \me = ((\mo \setminus \md) \cup \md) \cap \me = ((\mo \setminus \md)\cap \me) \cup (\md \cap \me) = (\mo \setminus \md)\cap \me \subseteq \mo \setminus \md$. Consequently, (**) $\mo \cap \ot = \mo\setminus\md$. Now, assume that there is a solution KB $\ot_k$ with the property $\mo \cap \ot_k \supset \mo \cap \ot$. By (**), this implies that $\mo \cap \ot_k \supset \mo \setminus \md$ which means that there is a $\md_k \subset \md \subseteq \mo$ such that $\mo \cap \ot_k = \mo \setminus \md_k \subseteq \ot_k$. Now $\ot_k$ is a solution KB w.r.t.\ $\langle\mo,\mb,\Tp,\Tn\rangle_\RQ$ and can be written as $\ot_k = (\ot_k \cap \mo) \cup (\ot_k \setminus \mo) = (\mo \setminus \md_k) \cup (\ot_k \setminus \mo)$. By $\md_k \subseteq \mo$ and since there is a set of formulas $\me := \ot_k \setminus \mo$ such that $(\mo\setminus\md_k) \cup \me \in \SO_{\langle\mo,\mb,\Tp,\Tn\rangle_\RQ}$ we have that $\me \in \EX{\md_k}{\langle\mo,\mb,\Tp,\Tn\rangle_\RQ}$ must hold wherefore $\md_k$ is a diagnosis by Definition~\ref{def:diagnosis}.
This, however, is a contradiction to the minimality of $\md$. Therefore, $\ot = (\mo \setminus \md) \cup \me$ must be a maximal solution KB for any $\me \in \EX{\md}{\langle\mo,\mb,\Tp,\Tn\rangle_\RQ}$.
\end{proof}
By claim~(i), Proposition~\ref{prop:min_max} assures that each maximal solution KB can be found by investigating all minimal diagnoses w.r.t. a DPI. Claim~(ii) shows that any solution KB built from a minimal diagnosis is indeed maximal. Thus, finding a suitable minimal diagnosis solves the problem of KB debugging completely.

\section{Background Knowledge}
\label{sec:BackgroundKnowledge}
The general debugging setting considered in this work envisions the opportunity for the user to specify some background knowledge $\mb$, i.e.~a set of formulas that are known (or strongly assumed) to be correct in advance.
Note that, in order for the debugging procedure to work soundly, before some background knowledge is incorporated into the DPI, it is necessary to verify its conformance with the postulated requirements $\RQ$ (cf.\ Definition~\ref{def:dpi}).
We can distinguish between two basic scenarios how background knowledge can be leveraged: (1)~We have an initial KB $\mo_{\mathsf{init}}$ and we know or want to assume that a subset of formulas in $\mo_{\mathsf{init}}$ is correct, i.e.\ $\mb \cap \mo_{\mathsf{init}} \neq \emptyset$, and (2)~we have an initial KB $\mo_{\mathsf{init}}$ and some background knowledge disjoint from $\mo_{\mathsf{init}}$, i.e.\ $\mb \cap \mo_{\mathsf{init}} = \emptyset$.

Example use cases for scenario~(1) are situations where a user knows that a subset of formulas $\mb$ in $\mo$ is definitely sound or wants to restrict the scope of debugging to a particular part of the KB. Concretely, this may occur, for instance, when $\mb$ is the result, i.e.\ the finally output solution KB $\ot$, of a former successful debugging session and $\mo$ is a further development of $\ot$, or in a collaborative setting where many users are involved in the development of $\mo$ and one of them may want to debug only formulas authored by herself and not touch foreign formulas, which are thus assumed as correct and assigned to $\mb$. 
In (1), $\mo_{\mathsf{init}}\cap\mb$ and $\mo_{\mathsf{init}}~\setminus~\mb$ partition the original KB $\mo_{\mathsf{init}}$ into a set of correct and a set of possibly incorrect formulas, respectively. The corresponding DPI would thus be $\langle\mo_{\mathsf{init}}~\setminus~\mb,\mb,\Tp,\Tn\rangle_\RQ$ for some sets of test cases $\Tp$ and $\Tn$. Note that this DPI \emph{does} meet the necessary condition (cf. Definition~\ref{def:dpi}) $\mo \cap \mb = \emptyset$ as $(\mo_{\mathsf{init}}~\setminus~\mb) \cap \mb = \emptyset$.
So, in the debugging session, only $\mo := \mo_{\mathsf{init}}~\setminus~\mb$ is used to search for diagnoses, which can reduce the search space substantially. Though, $\mb$ is incorporated in the calculations throughout the KB debugging procedure, but no formula in $\mb$ may take part in a diagnosis. The advantage of this over simply not considering the formulas in $\mb$ at all is, that the semantics of formulas in $\mb$ is not lost and can be exploited, e.g., to grant the desired semantic properties also in the context of existing approved knowledge or to facilitate a greater choice of queries to interact with a user, which can be exploited to ask queries with lower cardinality or involving less complex formulas (see Section~\ref{sec:UserInteraction} for details on queries).

In scenario~(2), the corresponding DPI looks like $\langle\mo_{\mathsf{init}},\mb,\Tp,\Tn\rangle_\RQ$ for some sets of test cases $\Tp$ and $\Tn$. An application of this scenario could be the reuse of an existing KB to support an increase of the fault detection rate and thus more sustainable debugging. For example, when formulating a KB $\mo_{\mathsf{init}}$ about a domain, a reference KB $\mb$ in that domain that is thoroughly curated by experts could be leveraged.  
The use of such a KB $\mb$ is possible both if $\mo_{\mathsf{init}}$ is correct as a standalone KB, i.e.\ $\mo_{\mathsf{init}}$ is already a solution KB for $\langle\mo_{\mathsf{init}},\emptyset,\Tp,\Tn\rangle_\RQ$, or not. In the first case, $\mo_{\mathsf{init}}$ might still contain formulations conflicting with $\mb$. In this vein, in both cases, faults may be detected that would have been missed otherwise.


%

\clearpage
\thispagestyle{empty}

\chapter{Diagnosis Computation}
\label{chap:DiagnosisComputation}
The search space for minimal diagnoses w.r.t.\ $\langle\mo,\mb,\Tp,\Tn\rangle_\RQ$ the size of which is in general $O(2^{|\mo|})$ (if all subsets of the KB $\mo$ are investigated) can be reduced to a great extent by exploiting the notion of a conflict set~\cite{Reiter87,dekleer1987,Shchekotykhin2012}. 
\begin{definition}[Conflict Set]\label{def:cs} Let $\langle\mo,\mb,\Tp,\Tn\rangle_\RQ$ be a DPI. A set of formulas $\mc \subseteq \mo$ is called a \emph{conflict set} w.r.t.\ $\langle\mo,\mb,\Tp,\Tn\rangle_\RQ$, written as $\mc \in \allC_{\langle\mo,\mb,\Tp,\Tn\rangle_\RQ}$, iff $\mc \cup U_\Tp$ is not a solution KB w.r.t.\ $\langle\mo,\mb,\Tp,\Tn\rangle_\RQ$. A conflict set $\mc$ is minimal, written as $\mc \in \minC_{\langle\mo,\mb,\Tp,\Tn\rangle_\RQ}$, iff there is no $\mc' \subset \mc$ such that $\mc'$ is a conflict set.
\end{definition}
Simply put, a (minimal) conflict set is a (minimal) faulty KB that is a subset of $\mo$. That is, a conflict set is one source causing the faultiness of $\mo$ in the context of $\mb \cup U_\Tp$.
In other words, a valid KB may not include all the formulas of any conflict set.
\begin{corollary}\label{cor:validonto_cs}
$\mc \subseteq \mo$ is a conflict set w.r.t.\ $\langle\mo,\mb,\Tp,\Tn\rangle_\RQ$ iff $\mc$ is invalid w.r.t.\ $\langle\cdot,\mb,\Tp,\Tn\rangle_\RQ$.
\end{corollary}
\begin{proof}
If $\mc$ is a conflict set w.r.t.\ $\langle\mo,\mb,\Tp,\Tn\rangle_\RQ$, then $\mc \cup U_\Tp$ is not a solution KB, i.e.\ $\mc \cup \mb \cup U_\Tp$ violates some $r\in\RQ$, some $\tp \in \Tp$ or some $\tn\in\Tn$. By extensiveness of $\mathcal{L}$, $\mc \cup \mb \cup U_\Tp \models \tp$ for all $\tp \in \Tp$, so $\mc \cup \mb \cup U_\Tp$ must violate some $r\in\RQ$ or entail some $\tn\in\Tn$. Thus, by Definition~\ref{def:valid_onto}, $\mc$ is invalid w.r.t.\ $\langle\cdot,\mb,\Tp,\Tn\rangle_\RQ$.

If $\mc \subseteq \mo$ is not valid w.r.t.\ $\langle\cdot,\mb,\Tp,\Tn\rangle_\RQ$, then $\mc \cup \mb \cup U_\Tp$ violates some $r\in\RQ$ or entails some $\tn\in\Tn$, wherefore $\mc \cup U_\Tp \notin \SO_{\langle\mo,\mb,\Tp,\Tn\rangle_\RQ}$. Hence, by Definition~\ref{def:cs}, $\mc$ is a conflict set w.r.t.\ $\langle\mo,\mb,\Tp,\Tn\rangle_\RQ$.
\end{proof}
Consequently, a conflict set $\mc$ along with the background knowledge $\mb$ either violates some $r\in\RQ$, entails some $\tn \in \Tn$, or yields to a violation of some $r\in\RQ$ or entailment of some $\tn \in \Tn$ if all formulas $U_\Tp$ comprised by the positive test cases are added to $\mc$. Any KB $\mo$
that is not valid w.r.t.\ $\langle\cdot,\mb,\Tp,\Tn\rangle_\RQ$ 
is itself a conflict set and includes at least one minimal conflict set. 
%
\begin{proposition}\label{prop:non_validonto_includes_cs}
Let $\langle\mo,\mb,\Tp,\Tn\rangle_\RQ$ be a DPI. Then, $\mo$ is not valid w.r.t.\ $\langle\cdot,\mb,\Tp,\Tn\rangle_\RQ$ iff $\mo$ includes at least one minimal conflict set w.r.t.\ $\langle\mo,\mb,\Tp,\Tn\rangle_\RQ$.
\end{proposition}
\begin{proof}
``$\Rightarrow$'': Let $\mo$ be not valid w.r.t.\ $\langle\cdot,\mb,\Tp,\Tn\rangle_\RQ$. Then $\mo \cup U_\Tp$ is not a solution KB w.r.t.\ $\langle\mo,\mb,\Tp,\Tn\rangle_\RQ$, which means that $\mo$ is a conflict set w.r.t.\ $\langle\mo,\mb,\Tp,\Tn\rangle_\RQ$ by definition~\ref{def:cs}. So, either $\mo$ is a already a minimal conflict set or there must be some subset $\mc \subset \mo$ which is a minimal conflict set w.r.t. $\langle\mo,\mb,\Tp,\Tn\rangle_\RQ$.

``$\Leftarrow$'': Let $\mo$ include at least one minimal conflict set w.r.t.\ $\langle\mo,\mb,\Tp,\Tn\rangle_\RQ$. Then, by Definition~\ref{def:cs}, there is some $\mc \subseteq \mo$ such that $\mc \cup U_\Tp$ is not a solution KB. Hence, by the monotonicity of $\mathcal{L}$, $\mo \cup U_\Tp$ cannot be a solution KB either. So, by Proposition~\ref{prop:validonto_targetonto}, $\mo$ is not valid w.r.t.\ $\langle\cdot,\mb,\Tp,\Tn\rangle_\RQ$.
\end{proof}
As a consequence, a complete and sound method for computing minimal conflict sets w.r.t.\ a DPI $\langle\mo,\mb,\Tp,\Tn\rangle_\RQ$ can be used to decide validity of $\mo$ w.r.t.\ $\langle\cdot,\mb,\Tp,\Tn\rangle_\RQ$. Moreover, such a method can be used to decide whether a given DPI is admissible, i.e.\ has solutions.
For, if a DPI is admissible and the given KB is invalid w.r.t.\ this DPI, then there cannot be an empty conflict set. In other words, if the empty KB is a conflict set -- or, equivalently, an empty conflict set exists w.r.t.\ a DPI --, then the DPI is not admissible.
\begin{proposition}\label{prop:cs_admissible}
Let $\langle\mo,\mb,\Tp,\Tn\rangle_\RQ$ be a DPI and $\mo$ be invalid w.r.t.\ $\langle\cdot,\mb,\Tp,\Tn\rangle_\RQ$. Then, there exists a minimal conflict set $\mc \neq \emptyset$ w.r.t.\ $\langle\mo,\mb,\Tp,\Tn\rangle_\RQ$ iff $\langle\mo,\mb,\Tp,\Tn\rangle_\RQ$ is admissible.
\end{proposition}
\begin{proof}
Since $\mo$ is not valid w.r.t.\ $\langle\cdot,\mb,\Tp,\Tn\rangle_\RQ$, there must be at least one conflict set w.r.t.\ $\langle\mo,\mb,\Tp,\Tn\rangle_\RQ$ by Proposition~\ref{prop:non_validonto_includes_cs}. Assume that there exists a minimal conflict set $\mc \neq \emptyset$ w.r.t.\ $\langle\mo,\mb,\Tp,\Tn\rangle_\RQ$. This can be true iff $\emptyset$ is not a (minimal) conflict set w.r.t.\ $\langle\mo,\mb,\Tp,\Tn\rangle_\RQ$. By Corollary~\ref{cor:validonto_cs} and Definition~\ref{def:valid_onto}, this is equivalent to the fact that $\emptyset \cup \mb \cup U_\Tp \equiv \mb \cup U_\Tp$ does not violate any $r\in\RQ$ and does not entail any $\tn \in \Tn$. By Proposition~\ref{prop:exist_diag}, this holds iff there exists a diagnosis w.r.t.\ $\langle\mo,\mb,\Tp,\Tn\rangle_\RQ$. By Definition~\ref{def:admissible}, this is equivalent to $\langle\mo,\mb,\Tp,\Tn\rangle_\RQ$ being admissible.
\end{proof}
The following proposition provides information about the relationship between (minimal) conflict sets and the background knowledge as well as the positive test cases.
\begin{proposition}\label{prop:cs_properties}
Let $\langle\mo,\mb,\Tp,\Tn\rangle_\RQ$ be a DPI and $\mc$ a conflict set w.r.t.\ $\langle\mo,\mb,\Tp,\Tn\rangle_\RQ$. Then the following holds:
\begin{enumerate}
	\item $\mc \cap \mb = \emptyset$.
	\item If $\mc$ is a minimal conflict set w.r.t.\ $\langle\mo,\mb,\Tp,\Tn\rangle_\RQ$, then $\mc \cap U_\Tp = \emptyset$.
\end{enumerate}
\end{proposition}
\begin{proof}
1): $\mc \cap \mb = \emptyset$ holds since $\mc \subseteq \mo$ (Definition~\ref{def:cs}) and $\mo \cap \mb = \emptyset$ (Definition~\ref{def:dpi}).

2):	Assume that $\mc$ is a minimal conflict set w.r.t.\ $\langle\mo,\mb,\Tp,\Tn\rangle_\RQ$ and $\mc \cap U_\Tp \neq \emptyset$. Since $\mc$ is a conflict set, we have that $\mc \cup \mb \cup U_\Tp$ violates some $r\in\RQ$ or entails some $\tn\in\Tn$ by Corollary~\ref{cor:validonto_cs} and Definition~\ref{def:valid_onto}. Since $(\mc\setminus U_\Tp) \cup \mb \cup U_\Tp = \mc \cup \mb \cup U_\Tp$ and $(\mc\setminus U_\Tp) \subset \mc$, this implies that $(\mc\setminus U_\Tp)$ is a conflict set w.r.t.\ $\langle\mo,\mb,\Tp,\Tn\rangle_\RQ$ which in turn implies that $\mc \notin \minC_{\langle\mo,\mb,\Tp,\Tn\rangle_\RQ}$ which is a contradiction.
\end{proof}

\section{Conflict Sets versus Justifications}
\label{sec:ConflictSetsVersusJustifications}
The notion of a conflict set is closely related to the notion of a justification~\cite{Horridge2008, Horridge2009, Horridge2010, Horridge2011a, Horridge2011b, Horridge2012b} which is frequently adopted in the field of the Semantic Web (cf.\ Section~\ref{sec:OntologiesAndDescriptionLogic}) in order to find minimal explanations for particular entailments in DL ontologies. Thus, the paradigm of a justification can be a useful aid in the debugging of faulty ontologies~\cite{Kalyanpur2006a}. Note that sometimes justifications are referred to as \mbox{MinAs} (Minimal Axiom Sets) \cite{Baader2008} or MUPS (Minimal Unsatisfiability Preserving Sub-TBoxes) \cite{Schlobach2007} where the latter term is mostly used in the context of ontology debugging.
The notion of a (minimal) conflict set, on the other hand, has been mainly adopted in the Diagnosis community~\cite{Reiter87,dekleer1987,Peischl2003,wotawa2002,Felfernig2004213}. In this section we want to establish a relationship between these two widely used instruments used for debugging. It will turn out that both terms are strongly related, but in debugging systems like the ones proposed in our work conflict sets are better suited as they automatically focus \emph{only} on the minimal explanations for \emph{faults} in a KB.

For example, the author of \cite{Kalyanpur2006a} i.a.\ discusses the use of justifications to aid the debugging of incoherent ontologies, i.e.\ ontologies that include unsatisfiable concepts (cf.\ Section~\ref{sec:DL}). If there are multiple unsatisfiable concepts, then some of these might be only unsatisfiable due to the unsatisfiability of another concept. 
Assume, for instance, an incoherent DL KB $\mo := \setof{A \sqsubset B, B \sqsubseteq E \sqcap \lnot E}$. In $\mo$ there are two unsatisfiable concepts $A$ and $B$ where $A$'s unsatisfiability is dependent on $B$'s unsatisfiability. Using the terminology of \cite{Kalyanpur2006a,Horridge2011a}, $A$ would be called a \emph{purely derived} unsatisfiable concept whereas $B$ would be called a \emph{root} unsatisfiable concept. Because the (only) justification for the unsatisfiability of $A$ is $J_A := \mo$ whereas the (only) justification for the unsatisfiability of $B$ is $J_B = \setof{B \sqsubseteq E \sqcap \lnot E} \subset J_A$. Therefore, \cite{Kalyanpur2006a} proposes to resolve root unsatisfiable concepts first since this might resolve some (purely) derived concepts as well, as in this example. However, finding out whether a concept is root or derived involves the computation of justifications for all unsatisfiable concepts in a KB. On the other hand, reliance on minimal conflict sets would implicate a direct focus on the faultiness (in this example: the incoherency) \emph{of the KB} and not necessarily on the exact explanations of all unsatisfiable concepts that cause the incoherency. In this vein, no justification for a purely derived concept can be a minimal conflict set. So, the computation of minimal conflict sets involves only the determination of those justifications for faults that \emph{must necessarily} be resolved. Therefore, for the given example, the only minimal conflict set is $J_B$.

A justification for a given formula (axiom) relative to a KB is a (subset-)minimal subset of the KB that entails the given formula.
\begin{definition}[Justification for a formula]\label{def:just_sentence}\cite{Kalyanpur.Just.ISWC07}
Let $\mo$ be a KB and $\alpha$ a formula, both over $\mathcal{L}$. Then $J \subseteq \mo$ is called a \emph{justification for $\alpha$ w.r.t.\ $\mo$}, written as $J \in \Just(\alpha,\mo)$, iff $J \models \alpha$ and for all $J' \subset J$ it holds that $J' \not\models \alpha$.
\end{definition}
Since we consider test cases which are \emph{sets of formulas} over $\mathcal{L}$, we generalize the definition of a justification as follows:
\begin{definition}[Justification for a set of formulas]\label{def:just_set}
Let $\mo$, $\mo'$ be KBs over $\mathcal{L}$. Then $J \subseteq \mo$ is called a \emph{justification for $\mo'$ w.r.t.\ $\mo$}, written as $J \in \Just(\mo',\mo)$, iff $J \models \mo'$ and for all $J' \subset J$ it holds that $J' \not\models \mo'$.\footnote{Remember that $J \models \mo'$ means that $J \models \tax$ for each $\tax\in\mo'$ (cf.\ Remark~\ref{rem:entailments_as_sets_of_formulas}).}
\end{definition}
In order to express the connection between justifications and conflict sets, we require yet another generalization of this definition. To this end, the following definition characterizes a justification for a set $X$ of KBs relative to a KB $\mo$ as a (subset-)minimal subset of $\mo$ such that this subset entails \emph{some} KB in $X$.
\begin{definition}[Justification for a set of sets of formulas]\label{def:just_set_of_set}
Let $\mo$ be a KB over $\mathcal{L}$ and $X$ a set of KBs over $\mathcal{L}$. Then $J \subseteq \mo$ is called \emph{justification for $X$ w.r.t.\ $\mo$}, written as $J \in \Just(X,\mo)$, iff $J \models \mo'$ for some $\mo' \in X$ and for all $J' \subset J$ it holds that $J' \not\models \mo''$ for all $\mo'' \in X$.
\end{definition}
Based on Definition~\ref{def:just_set_of_set}, the relation between conflict sets and justifications is captured by the following Proposition~\ref{prop:cs_just}. Intuitively, any conflict set w.r.t.\ $\langle\mo,\mb,\Tp,\Tn\rangle_\RQ$ is the part of a justification for a fault that is relevant for the debugging task, where fault refers to an inconsistency (and/or incoherency) and/or a negative test case entailed by $\mo \cup \mb \cup U_\Tp$. Since debugging focuses on the deletion of KB formulas only, ``relevant'' in this context refers to the subset of the justification that does not contain any sentences in $\mb$ and $U_\Tp$, but solely sentences from $\mo$. Importantly, there may be justifications, in general, the relevant subset of which is not a \emph{minimal} conflict set. The reason why this case can arise in spite of the set-minimality of justifications is that the \emph{relevant} part of a justification (for some set of sentences $\mo_1$, e.g.\ a negative test case $\tn_1 \in \Tn$) may be a superset of the \emph{relevant} part of another justification (for some other set of sentences $\mo_2$, e.g.\ another negative test case $\tn_2 \in \Tn$) whereas both justifications are not in a subset-relationship (i.e.\ contain different sentences from $\mb$ and/or $U_\Tp$). This circumstance is illustrated by the following example:
\begin{example}\label{example:min_conflict_set_focuses_on_relevant_part_of_just}
Let a DPI $\langle\mo,\mb,\Tp,\Tn\rangle_\RQ$ be defined as 
\begin{align*}
\mo &:= \setof{B \sqsubseteq E, E \sqsubseteq \exists r.G} \\
\mb &:= \setof{A \sqsubseteq B} \\
\Tn &:= \setof{\setof{A \sqsubseteq E},\setof{B \sqsubseteq \exists r.G}} \\
\Tp &:= \emptyset \\
\RQ &:= \setof{\mbox{consistency}}
\end{align*}
We have that $\mo \cup \mb \cup U_{\Tp}$ is consistent and thus no requirement in $\RQ$ is violated. But, the two negative test cases are both entailed by $\mo \cup \mb \cup U_{\Tp}$ wherefore $\mo$ is invalid w.r.t.\ $\langle\cdot,\mb,\Tp,\Tn\rangle_\RQ$. The set of justifications for the violation of the first negative test case is $J_{\tn_1} = \setof{\setof{A \sqsubseteq B, B \sqsubseteq E}}$; for the second one it is $J_{\tn_2} = \setof{\setof{B \sqsubseteq E, E \sqsubseteq \exists r.G}}$. The relevant subset of the justification $J_1$ in $J_{\tn_1}$ is $J_{1,rel} = \setof{B \sqsubseteq E}$ (since $\setof{A \sqsubseteq B}$ is in $\mb$) whereas the relevant subset of the justification $J_2$ in $J_{\tn_2}$ is $J_{2,rel} = \{B \sqsubseteq E$, $E \sqsubseteq \exists r.G\}$, i.e.\ $J_{1,rel} \subset J_{2,rel}$ despite that there is no set subset-relationship between $J_1$ and $J_2$. Hence, there are two justifications that explain the invalidity of $\mo$ w.r.t.\ $\langle\cdot,\mb,\Tp,\Tn\rangle_\RQ$, but there is only one minimal conflict set $\mc = J_{1,rel}$ w.r.t.\ $\langle\mo,\mb,\Tp,\Tn\rangle_\RQ$.\qed
\end{example}

So, generally, the set of minimal conflict sets w.r.t.\ a DPI is a subset of the set of justifications for faults in $\mo \cup \mb \cup U_\Tp$, which is due to the focus on just the parts of justifications that are relevant for the KB debugging task.
\begin{proposition}\label{prop:cs_just}
Let $\langle\mo,\mb,\Tp,\Tn\rangle_\RQ$ be a DPI. Additionally, let 
\begin{enumerate}[(a)]
	\item $X := \setof{\setof{A_i \sqsubseteq \bot}\,|\,A_i \in N_C} \cup \setof{\setof{r_i \sqsubseteq \bot}\,|\,r_i \in N_R} \cup \setof{\setof{\top \sqsubseteq \bot}} \cup \Tn$ \\ if $\RQ = \setof{\mbox{consistency, coherency}}$ and 
	\item \label{prop:cs_just:bullet_consistency} $X := \setof{\setof{\top \sqsubseteq \bot}} \cup \Tn$ if $\RQ = \setof{\mbox{consistency}}$.\footnote{We use DL notation in this proposition since justifications, as argued, are mostly applied to DL KBs. An equivalent formulation of the proposition for FOL or PL is straightforward (cf.\ Example~\ref{example:FOL_to_DL} and Remark~\ref{rem:reduce_conditions_of_dpi_def}). Note that for PL only (\ref{prop:cs_just:bullet_consistency}) is relevant since coherency is not defined for PL. Further, recall that $N_C$ and $N_R$ are defined in Section~\ref{sec:DL}.} 
\end{enumerate}
Then the following holds: 
\begin{enumerate}
		\item If $\mc$ is a minimal conflict set w.r.t.\ $\langle\mo,\mb,\Tp,\Tn\rangle_\RQ$, then there is some $J \in \Just(X,\mo\cup\mb\cup U_\Tp)$ such that $(J \cap \mo)\setminus U_\Tp = \mc$.
		\item For all $J \in \Just(X,\mo\cup\mb\cup U_\Tp)$ it is true that $\mc := (J \cap \mo)\setminus U_\Tp$ is a conflict set w.r.t.\ $\langle\mo,\mb,\Tp,\Tn\rangle_\RQ$, but not necessarily a minimal one.
\end{enumerate}
\end{proposition}
\begin{proof}
\textbf{1):} Assume that $\mc\in\minC_{\langle\mo,\mb,\Tp,\Tn\rangle_\RQ}$ and for all $J \in \Just(X,\mo\cup\mb\cup U_\Tp)$ it holds that $(J \cap \mo) \setminus U_\Tp \neq \mc$. 
There are two cases to distinguish between: (a)~there is some sentence in $(J \cap \mo) \setminus U_\Tp$ that is not in $\mc$ and (b)~there is some sentence in $\mc$ that is not in $(J \cap \mo) \setminus U_\Tp$.

Let us first assume (a), i.e.\ for all $J\in \Just(X,\mo\cup\mb\cup U_\Tp)$ it holds that there is some sentence $\tax$ in $(J \cap \mo) \setminus U_\Tp$ that is not in $\mc$. 
Additionally, assume there is a $J\in \Just(X,\mo\cup\mb\cup U_\Tp)$ such that $J \subseteq \mc \cup \mb \cup U_\Tp$. We can write $J$ as $J = S_1 \cup S_2 \cup S_3$ for $S_1 := [(J \cap \mo)\setminus U_\Tp]$, $S_2:=[J \cap \mb]$ and $S_3:=[J \cap U_\Tp]$. Since $J = S_1 \cup S_2 \cup S_3 \subseteq \mc \cup \mb \cup U_\Tp$ it must hold in particular that $S_1 \subseteq \mc \cup \mb \cup U_\Tp$ and therefore $\tax \in \mc \cup \mb \cup U_\Tp$. However, $\tax \notin \mc$ by assumption, $\tax \notin \mb$ since $\tax \in \mo$ and $\mb \cap \mo = \emptyset$, and $\tax \notin U_\Tp$ since $\tax \in S_1$ and $S_1 \cap U_\Tp = \emptyset$. This is a contradiction. Hence, for all $J\in \Just(X,\mo\cup\mb\cup U_\Tp)$ it holds that $J \not\subseteq \mc \cup \mb \cup U_\Tp$. Since $X$ captures all $r\in\RQ$ and $\tn\in\Tn$, we can conclude that $\mc$ is not a conflict set w.r.t.\ $\langle\mo,\mb,\Tp,\Tn\rangle_\RQ$ which is a contradiction to $\mc\in\minC_{\langle\mo,\mb,\Tp,\Tn\rangle_\RQ}$.

Let us now assume (b), i.e.\ for all $J\in \Just(X,\mo\cup\mb\cup U_\Tp)$ it holds that there is some sentence $\tax$ in $\mc$ that is not in $(J \cap \mo) \setminus U_\Tp$. Since 
$\mc$ is a conflict set and since $X$ captures all $r\in\RQ$ and $\tn\in\Tn$, we have that $\mc \cup \mb \cup U_\Tp \models \mo'$ for some $\mo' \in X$. So, there must be some $J_0 \in \Just(X,\mo\cup\mb\cup U_\Tp)$ such that $J_0 \subseteq \mc \cup \mb \cup U_\Tp$. As $\mc \in \minC_{\langle\mo,\mb,\Tp,\Tn\rangle_\RQ}$, there cannot be any $J \in \Just(X,\mo\cup\mb\cup U_\Tp)$ with $J \subseteq \mc' \cup \mb \cup U_\Tp$ for arbitrary $\mc' \subset \mc$. This must hold in particular for $J_0$ which implies that $J_0 \cap \mc = \mc$ which is equivalent to $\mc \subseteq J_0$. As (1)~$\mc \subseteq \mo$ (Definition~\ref{def:cs}) and, by Proposition~\ref{prop:cs_properties} and by the fact that $\mc \in \minC_{\langle\mo,\mb,\Tp,\Tn\rangle_\RQ}$, (2)~$\mc \cap U_\Tp = \emptyset$, we can conclude that $\mc \subseteq (J_0 \cap \mo) \setminus U_\Tp$ which is a contradiction since there cannot be a $\tax$ in $\mc$ that is not in $(J_0 \cap \mo) \setminus U_\Tp$.
%

\textbf{2):} If $J \in \Just(X,\mo\cup\mb\cup U_\Tp)$, then, by Definition~\ref{def:just_set_of_set}, $J \models \mo'$ for some $\mo' \in X$ and $J \subseteq \mo\cup\mb\cup U_\Tp$. So, $[(J \cap \mo)\setminus U_\Tp] \cup \mb \cup U_\Tp = (J \cap \mo) \cup \mb \cup U_\Tp \supseteq J$ wherefore $[(J \cap \mo)\setminus U_\Tp] \cup \mb \cup U_\Tp \models \mo'$ by monotonicity of $\mathcal{L}$. As $\mo' \in X$ and $X$ captures all the reasons why some $r\in\RQ$ or some $\tn\in\Tn$ may not be fulfilled (cf. discussion in Chapter~\ref{chap:OntologyDebugging}), we have that $[(J \cap \mo)\setminus U_\Tp] \cup \mb \cup U_\Tp$ violates some $r\in\RQ$ or entails some $\tn \in \Tn$. This implies that $[(J \cap \mo)\setminus U_\Tp] \cup U_\Tp \notin \SO_{\langle\mo,\mb,\Tp,\Tn\rangle_\RQ}$. Since $(J \cap \mo)\setminus U_\Tp \subseteq \mo$ is also true, $(J \cap \mo)\setminus U_\Tp \in \allC_{\langle\mo,\mb,\Tp,\Tn\rangle_\RQ}$ by Definition~\ref{def:cs}.

To see that $(J \cap \mo)\setminus U_\Tp \notin \minC_{\langle\mo,\mb,\Tp,\Tn\rangle_\RQ}$ holds in general, reconsider Example~\ref{example:min_conflict_set_focuses_on_relevant_part_of_just} where $(J_2 \cap \mo) \setminus U_{\Tp} = J_2 \supset \mc$ holds for the justification $J_2$ and the minimal conflict set $\mc$.
\end{proof}

\section{The Relation between Conflict Sets and Diagnoses}
\label{sec:TheRelationBetweenConflictSetsAndDiagnoses}
A minimal conflict set has the property that deletion of any formula in it yields a set of formulas which is correct in the context of $\mb$, $\Tp$, $\Tn$ and $\RQ$.
\begin{proposition}
If $\mc$ is a minimal conflict set w.r.t.\ $\langle\mo,\mb,\Tp,\Tn\rangle_\RQ$, then $\mc'$ is valid w.r.t.\ $\langle\cdot,\mb,\Tp,\Tn\rangle_\RQ$ for each $\mc'\subset\mc$.
\end{proposition}
\begin{proof}
Since $\mc \in \minC_{\langle\mo,\mb,\Tp,\Tn\rangle_\RQ}$, it must hold that $\mc' \notin \allC_{\langle\mo,\mb,\Tp,\Tn\rangle_\RQ}$. Then, by Corollary~\ref{cor:validonto_cs}, $\mc'$ is valid w.r.t.\ $\langle\cdot,\mb,\Tp,\Tn\rangle_\RQ$.
\end{proof}
Hence, by deletion of at least one formula from \emph{each} minimal conflict set w.r.t.\ $\langle\mo,\mb,\Tp,\Tn\rangle_\RQ$, a valid KB can be obtained from $\mo$. Thus, a solution KB $(\mo\setminus\md) \cup U_\Tp$ can be obtained by calculation of a hitting set $\md$ of all minimal conflict sets in $\minC_{\langle\mo,\mb,\Tp,\Tn\rangle_\RQ}$. The Hitting Set problem is defined as follows:
\begin{definition}[Hitting Set]\label{def:hs}
Let $S=\setof{S_1,\dots,S_n}$ be a set of sets. Then, $H$ is called a \emph{hitting set of $S$} 
iff $H \subseteq U_S$ and $H \cap S_i \neq \emptyset$ for all $i=1,\dots,n$. 

A hitting set $H$ of $S$ is \emph{minimal} iff there is no hitting set $H'$ of $S$ such that $H' \subset H$.
\end{definition}
\begin{proposition}\cite{friedrich2005gdm}\label{prop:mindiag_mincs}
A (minimal) diagnosis w.r.t.\ the DPI $\langle\mo,\mb,\Tp,\Tn\rangle_\RQ$ is a (minimal) hitting set of all minimal conflict sets w.r.t.\ $\langle\mo,\mb,\Tp,\Tn\rangle_\RQ$.
\end{proposition}

Now, we want to contemplate two example DPIs and analyze them regarding the their minimal conflict sets and minimal diagnoses:
\begin{example}\label{example:analysis_TabExDpi2}
In this example, we analyze the PL DPI $\tuple{\mo,\mb,\Tp,\Tn}_\RQ$ given by Table~\ref{tab:example2}. There are two minimal conflict sets w.r.t.\ $\tuple{\mo,\mb,\Tp,\Tn}_\RQ$, i.e.\ $\minC_{\tuple{\mo,\mb,\Tp,\Tn}_\RQ} =\setof{\mc_1, \mc_2} = \setof{\tuple{1,2,5},\tuple{1,2,7}}$.\footnote{Please notice that we sometimes write $i$ instead of $\tax_i$ for brevity when it is clear what is meant. We will do so in many other examples as well.}

Why is $\mc_1$ a conflict set w.r.t.\ $\tuple{\mo,\mb,\Tp,\Tn}_\RQ$? We recall Definition~\ref{def:cs} and argue as follows to deduce the entailment $\mc_1 \models \tn_1$ where $\tn_1 \in \Tn$ (left of the colon: the formulas used in the deduction are underlined; right of the colon: the relevant implications are underlined):
\begin{align*}
\underline{\tax_1}:&\quad  \underline{A \;\rightarrow\; E} \\
\underline{\tax_2}:&\quad  X \lor \underline{E \;\rightarrow} \;F \land \underline{Y} \land Z \\
\underline{\tax_5}:&\quad  \underline{Y \;\rightarrow\; \lnot A} \\
\underline{\tax_1, \tax_2, \tax_5}:&\quad  \underline{A \;\rightarrow\; \lnot A} \equiv \underline{\lnot A \lor \lnot A} \equiv \underline{\lnot A} \\
\tn_1 \in \Tn:&\quad  \underline{\lnot A}    \quad\qed
\end{align*}
Minimality of $\mc_2$ is obvious from this argumentation. i.e.\ we cannot deduce $\tn_1$ if any one of the formulas 1, 2 or 5 is omitted, and there is no other fault except for the violation of $\tn_1$.  

Why is $\mc_2$ a conflict set w.r.t.\ $\tuple{\mo,\mb,\Tp,\Tn}_\RQ$? We recall Definition~\ref{def:cs} and argue as follows to deduce the entailment $\mc_2 \cup \mb \models \tn_1$ where $\tn_1 \in \Tn$ (left of the colon: the formulas used in the deduction are underlined; right of the colon: the relevant implications are underlined):
\begin{align*}
\underline{\tax_1}:&\quad  \underline{A \;\rightarrow\; E} \\
\underline{\tax_2}:&\quad  X \lor \underline{E \;\rightarrow} \;F \land Y \land \underline{Z} \\
\underline{\tax_7}:&\quad  \underline{Z \;\rightarrow\; G} \\
(G \;\rightarrow\; \lnot A) \in \mb:&\quad \underline{G \;\rightarrow\; \lnot A} \\ 
\underline{\tax_1, \tax_2, \tax_7},\mb:&\quad  \underline{A \;\rightarrow\; \lnot A} \equiv \underline{\lnot A \lor \lnot A} \equiv \underline{\lnot A} \\
\tn_1 \in \Tn:&\quad  \underline{\lnot A}    \quad\qed
\end{align*}
Minimality of $\mc_2$ is obvious from this argumentation. i.e.\ we cannot deduce $\tn_1$ if any one of the formulas 1, 2 or 7 is omitted, and there is no other fault except for the violation of $\tn_1$.

There are no further minimal conflict sets w.r.t.\ $\tuple{\mo,\mb,\Tp,\Tn}_\RQ$. This is fairly easy to see since
\begin{itemize}
	\item $\mo \cup \mb \cup U_{\Tp} = \mo \cup \mb$ cannot be inconsistent due to the fact that the only negative literal occurring on the righthand side of an implication is $\lnot A$ and $A$ does not occur at the righthand side of any implication in $\mo \cup \mb$,
	\item there is no other way to deduce $\tn_1$ than using a superset of the formulas in $\mc_1$ or $\mc_2$ and
	\item $\tn_1$ is the only negative test case in $\Tn$.
\end{itemize}
Hence, the set of all minimal diagnoses $\minD_{\tuple{\mo,\mb,\Tp,\Tn}_\RQ} =\setof{\md_1, \md_2, \md_3} = \{[1]$, $[2]$, $[5,7]\}$ is obtained by computing all minimal hitting sets of $\minC_{\tuple{\mo,\mb,\Tp,\Tn}_\RQ} =\setof{\mc_1, \mc_2}$ (cf.\ Proposition~\ref{prop:mindiag_mincs}).\qed  
\end{example}

\begin{example}\label{example:analysis_TabExDpi3}
In this example, we analyze the DL DPI $\tuple{\mo,\mb,\Tp,\Tn}_\RQ$ given by Table~\ref{tab:example3}. There are four minimal conflict sets w.r.t.\ $\tuple{\mo,\mb,\Tp,\Tn}_\RQ$, i.e.\ 
\begin{align*}
\minC_{\tuple{\mo,\mb,\Tp,\Tn}_\RQ} =\setof{\mc_1, \mc_2, \mc_3, \mc_4} = \setof{\tuple{1,2,5},\tuple{2,4,6},\tuple{1,3,4},\tuple{1,5,6,8}}
\end{align*}
Why is $\mc_1$ a conflict set w.r.t.\ $\tuple{\mo,\mb,\Tp,\Tn}_\RQ$? We recall Definition~\ref{def:cs} and argue as follows to deduce the entailment $\mc_1 \models \tn_1$ where $\tn_1 \in \Tn$ (left of the colon: the formulas used in the deduction are underlined; right of the colon: the relevant implications are underlined):
\begin{align*}
\underline{\tax_1}:&\quad  \underline{A \;\sqsubseteq\; B} \\
\underline{\tax_2}:&\quad  \underline{B \;\sqsubseteq\; G} \\
\underline{\tax_5}:&\quad  \underline{G \;\sqsubseteq\; K} \\
\underline{\tax_1, \tax_2, \tax_5}:&\quad  \underline{A \;\sqsubseteq\; K} \\
\tn_1  \in \Tn:&\quad  \underline{A \;\sqsubseteq\; K}    \quad\qed
\end{align*}
Minimality of $\mc_1$ is follows from this argumentation. i.e.\ we cannot deduce $\tn_1$ if any one of the formulas 1, 2 or 5 is omitted, and from the fact that we cannot deduce an incoherency ($r_2$), inconsistency ($r_1$) or the entailment of any other negative test case $\tn\in\Tn$ for any KB $\mc'_1 \cup \mb \cup U_{\Tp}$ for any $\mc'_1 \subset \mc_1$.  

Why is $\mc_2$ a conflict set w.r.t.\ $\tuple{\mo,\mb,\Tp,\Tn}_\RQ$? We recall Definition~\ref{def:cs} and argue as follows to deduce 
that $\mc_2 \cup \mb$ is incoherent and thus violates the requirement $r_2 \in \RQ$ (left of the colon: the formulas used in the deduction are underlined; right of the colon: the relevant implications are underlined):
\begin{align*}
\underline{\tax_2}:&\quad  \underline{B \;\sqsubseteq\; G} \\
\underline{\tax_6}:&\quad  \underline{G \;\sqsubseteq\; \exists r.F} \\
(1):\, \underline{\tax_2,\tax_6}:&\quad  \underline{B \;\sqsubseteq\; \exists r.F} \\
\underline{\tax_4}:&\quad  \underline{B \;\sqsubseteq\; \forall r.H} \\
(H \;\sqsubseteq\; \lnot F) \in \mb:&\quad \underline{H \;\sqsubseteq\; \lnot F} \\ 
(2):\, \underline{\tax_4},\mb:&\quad  \underline{B \;\sqsubseteq\; \forall r.\lnot F} \\
(1) \mbox{ and } (2):&\quad  \underline{B \;\sqsubseteq\; \bot} \\
r_1 \in \RQ:&\quad  \underline{B \;\not\sqsubseteq\; \bot}    \quad\qed
\end{align*}
Since we cannot deduce an incoherency ($r_2$), inconsistency ($r_1$) or the entailment of any negative test case $\tn\in\Tn$ for any KB $\mc'_2 \cup \mb \cup U_{\Tp}$ for any $\mc'_2 \subset \mc_2$, the minimality of $\mc_2$ follows.

Why is $\mc_3$ a conflict set w.r.t.\ $\tuple{\mo,\mb,\Tp,\Tn}_\RQ$? We recall Definition~\ref{def:cs} and argue as follows to deduce 
that $\mc_3 \cup \mb \cup U_{\Tp}$ is inconsistent and thus violates the requirement $r_1 \in \RQ$ (left of the colon: the formulas used in the deduction are underlined; right of the colon: the relevant implications are underlined):
\begin{align*}
A(x) \in \mb:&\quad  \underline{A(x)} \\
\underline{\tax_1}:&\quad  \underline{A \;\sqsubseteq\; B} \\
(1):\,\underline{\tax_1},\mb:&\quad  \underline{B(x)} \\
(2):\, \tp_1 \in \Tp:&\quad  \underline{r(x,y)} \\
\underline{\tax_4}:&\quad  \underline{B \;\sqsubseteq\; \forall r.H} \\
(3):\,(1) \mbox{ and } \underline{\tax_4}:&\quad \underline{H(y)} \\
(4):\,\underline{\tax_3}:&\quad  \underline{\lnot H(y)} \\
(3) \mbox{ and } (4):&\quad \mbox{\Lightning} \quad\qed
\end{align*}
No inconsistency ($r_1$) or incoherency ($r_2$) can be derived and no negative test case $\tn \in \Tn$ is entailed from any $\mc'_3 \cup \mb \cup U_{\Tp}$ for $\mc'_3 \subset \mc_3$. Hence, $\mc_3$ is a minimal conflict set w.r.t.\ $\tuple{\mo,\mb,\Tp,\Tn}_\RQ$.

Why is $\mc_4$ a conflict set w.r.t.\ $\tuple{\mo,\mb,\Tp,\Tn}_\RQ$? We recall Definition~\ref{def:cs} and argue as follows to deduce the entailment $\mc_4 \cup \mb \models \tn_2$ where $\tn_2 \in \Tn$ (left of the colon: the formulas used in the deduction are underlined; right of the colon: the relevant implications are underlined):
\begin{align*}
\underline{\tax_8}:&\quad  \underline{L \;\sqsubseteq\; G} \\
\underline{\tax_6}:&\quad  \underline{G \;\sqsubseteq\; \exists r.F} \\
(1):\,\underline{\tax_6,\tax_8}:&\quad  \underline{L \;\sqsubseteq\; \exists r.F} \\
A(x) \in \mb:&\quad  \underline{A(x)} \\
(2):\,\underline{\tax_1},\mb:&\quad  \underline{B(x)} \\
(3):\,\underline{\tax_5}:&\quad  \underline{G \;\sqsubseteq\; K} \\
(1) \mbox{ and } (2) \mbox{ and } (3):&\quad \underline{L \;\sqsubseteq\; \exists r.F,\, B(x),\, G \;\sqsubseteq\; K} \\
\tn_1 \in \Tn:&\quad  \underline{L \;\sqsubseteq\; \exists r.F,\, B(x),\, G \;\sqsubseteq\; K}    \quad\qed
\end{align*}
No inconsistency ($r_1$) or incoherency ($r_2$) can be derived and no negative test case $\tn \in \Tn$ is entailed from any $\mc'_4 \cup \mb \cup U_{\Tp}$ for $\mc'_4 \subset \mc_4$. Thus, $\mc_4$ is a minimal conflict set w.r.t.\ $\tuple{\mo,\mb,\Tp,\Tn}_\RQ$.

Hence, the set of all minimal diagnoses $\minD_{\tuple{\mo,\mb,\Tp,\Tn}_\RQ}$, obtained by computing all minimal hitting sets of $\minC_{\tuple{\mo,\mb,\Tp,\Tn}_\RQ} =\setof{\mc_1, \mc_2, \mc_3, \mc_4}$ (cf.\ Proposition~\ref{prop:mindiag_mincs}), comprises ten minimal diagnoses $\md_i$ for $i = 1,\dots,10$: 
\begin{align*}
\md_1 &= [1,2]   &\md_2 &= [1,4] \\
\md_3 &= [1,6]   &\md_4 &= [2,3,5] \\
\md_5 &= [2,3,6] &\md_6 &= [2,3,8] \\
\md_7 &= [2,4,6] &\md_8 &= [2,4,8] \\
\md_9 &= [3,5,6] &\md_{10} &= [4,5] 
\end{align*}

Although the DPI $\tuple{\mo,\mb,\Tp,\Tn}_\RQ$ is very small in size, i.e.\ number of formulas occurring in it is very small, the reader might agree that it is not trivial on the one hand (1)~to realize which subsets of this KB $\mo$ are (minimal) conflict sets, (2)~to see \emph{that} or \emph{why} a subset of this KB $\mo$ along with the background knowledge $\mb$ and the union of the positive test cases $U_{\Tp}$ is a (minimal) conflict set (cf.\ \cite{Horridge2011b}) and (3)~to assess that there are no further minimal conflict sets w.r.t.\ $\tuple{\mo,\mb,\Tp,\Tn}_\RQ$. This example gives a little bit of an impression that tool assistance in the debugging of KBs is inevitable especially for real-world KBs that are huge in size and/or complex in terms of the expressivity of the used logic or in terms of their ``debugging properties'', i.e.\ large number and/or size of minimal conflict sets and/or minimal diagnoses. 

A means to handle problems~(1) and (3) is provided by some method for the computation of a minimal conflict set (e.g.\ $\scQX$ given by Algorithm~\ref{algo:qx} below, see Section~\ref{sec:cs_comp}) coupled with a hitting set tree algorithm (e.g.\ \textsc{HS} described by Algorithm~\ref{algo:hs} below, see Section~\ref{sec:hs_comp}) for the systematic computation of \emph{different} minimal conflict sets, or other mechanisms such as the ALL\_JUST\_ALG presented in~\cite{Kalyanpur.Just.ISWC07} which computes all justifications for some particular entailment (but, some post-processing of the justifications is necessary to obtain minimal conflict sets, cf.\ Section~\ref{sec:ConflictSetsVersusJustifications}). 

Problem (2) and its complexity for humans has been studied in \cite{Horridge2011b} with a focus on justifications in DL or OWL KBs. Since a minimal conflict set can be regarded as the relevant (i.e.\ potentially faulty) part of a justification for some undesired entailment (i.e.\ a violated requirement or test case) as we analyzed in Section~\ref{sec:ConflictSetsVersusJustifications},
the cognitive complexity model proposed by \cite{Horridge2011b} applies also to minimal conflict sets. Ways to facilitate the understanding of justifications for humans (that might be successfully applied also to conflict sets) have been addressed in \cite{Horridge2010, Horridge2009, Horridge2008}. Moreover, there is an ontology editing browser SWOOP~\cite{Kalyanpur2006b} equipped with a strikeout feature~\cite{Kalyanpur2006a} that highlights parts of justifications that are relevant for the entailment by striking out all irrelevant parts. This is more or less the automation of our analyses of the conflict sets by underlining the relevant parts of the formulas in this example and Example~\ref{example:analysis_TabExDpi2}.\qed   
\end{example}

\renewcommand{\arraystretch}{1.4} 
\begin{table}[h]
	\footnotesize
	\centering
		\rowcolors[]{2}{gray!8}{gray!16} 
		\begin{tabular}{ c c c c } 
			\rowcolor{gray!40}
			\toprule\addlinespace[0pt]
			$i$ & $\tax_i$ & $\mo$ & $\mb$  \\ \addlinespace[0pt]\midrule\addlinespace[0pt]
			1 & $A \rightarrow E$ & $\bullet$ & 	\\
			2 & $X \lor E \rightarrow F \land Y \land Z$ & $\bullet$ &  	\\
			3 & $F \rightarrow B$ & $\bullet$ &  	\\
			4 & $B \rightarrow X$ & $\bullet$ &  	\\
			5 & $Y \rightarrow \lnot A$ & $\bullet$ &  	\\
			6 & $B \rightarrow Z$ & $\bullet$ &  	\\
			7 & $Z \rightarrow G$ & $\bullet$ & 	\\
			8 & $G \rightarrow \lnot A$ &  & $\bullet$  	\\ 
			\addlinespace[0pt]\bottomrule 
			\rowcolor{gray!40}
			$i$ & \multicolumn{3}{c}{$\tp_i\in\Tp$} \\ \addlinespace[0pt]\midrule\addlinespace[0pt]
			$\times$ & \multicolumn{3}{c}{$\times$} 	\\ \addlinespace[0pt]\toprule\addlinespace[0pt]
			\rowcolor{gray!40}
			$i$ & \multicolumn{3}{c}{$\tn_i\in\Tn$} \\ \addlinespace[0pt]\midrule\addlinespace[0pt]
			$1$ & \multicolumn{3}{c}{$\lnot A$} 	\\ \addlinespace[0pt]\toprule\addlinespace[0pt]
			\rowcolor{gray!40}
			$i$ & \multicolumn{3}{c}{$r_i\in\RQ$} \\ \addlinespace[0pt]\midrule\addlinespace[0pt]
			$1$ & \multicolumn{3}{c}{consistency} \\ \addlinespace[0pt]\bottomrule
			\end{tabular}
	\caption{Propositional Logic Example DPI}
	\label{tab:example2}
\end{table}


\renewcommand{\arraystretch}{1.4} 
\begin{table}[h]
\footnotesize
	\centering
		\rowcolors[]{2}{gray!8}{gray!16} 
		\begin{tabular}{ c c c c } 
			\rowcolor{gray!40}
			\toprule\addlinespace[0pt]
			$i$ & $\tax_i$ & $\mo$ & $\mb$  \\ \addlinespace[0pt]\midrule\addlinespace[0pt]
			1 & $A \sqsubseteq B$ & $\bullet$ & 	\\
			2 & $B \sqsubseteq G$ & $\bullet$ &  	\\
			3 & $\lnot H(y)$ & $\bullet$ &  	\\
			4 & $B \sqsubseteq \forall r.H$ & $\bullet$ &  	\\
			5 & $G \sqsubseteq K$ & $\bullet$ &  	\\
			6 & $G \sqsubseteq \exists r.F$ & $\bullet$ &  	\\
			7 & $A(x)$ &  & $\bullet$	\\
			8 & $L \sqsubseteq G$ & $\bullet$ &   	\\ 
			9 & $H \sqsubseteq \lnot F$ &  &  $\bullet$ 	\\ 
			\addlinespace[0pt]\bottomrule 
			\rowcolor{gray!40}
			$i$ & \multicolumn{3}{c}{$\tp_i\in\Tp$} \\ \addlinespace[0pt]\midrule\addlinespace[0pt]
			1 & \multicolumn{3}{c}{$r(x,y)$} 	\\ \addlinespace[0pt]\toprule\addlinespace[0pt]
			\rowcolor{gray!40}
			$i$ & \multicolumn{3}{c}{$\tn_i\in\Tn$} \\ \addlinespace[0pt]\midrule\addlinespace[0pt]
			1 & \multicolumn{3}{c}{$A \sqsubseteq K$} 	\\ 
			2 & \multicolumn{3}{c}{$L \sqsubseteq \exists r.F, B(x), G \sqsubseteq K$} 	\\ \addlinespace[0pt]\toprule\addlinespace[0pt]
			\rowcolor{gray!40}
			$i$ & \multicolumn{3}{c}{$r_i\in\RQ$} \\ \addlinespace[0pt]\midrule\addlinespace[0pt]
			$1$ &\multicolumn{3}{c}{consistency} \\ 
			$2$ &\multicolumn{3}{c}{coherency} \\ \addlinespace[0pt]\bottomrule
			\end{tabular}
	\caption{Description Logic Example DPI}
	\label{tab:example3}
\end{table}


\section{Methods for Diagnosis Computation}
\label{sec:MethodsForDiagnosisComputation}
Two common methods employed for the computation of (minimal) diagnoses~\cite{Shchekotykhin2012,Rodler2013} are the
QuickXPlain algorithm~\cite{junker04} (in short $\scQX$) and a hitting set search tree~\cite{Reiter87,greiner1989correction} (in short $\scHS$). Thereby, $\scQX$ serves as a deterministic method for computing one minimal conflict set w.r.t.\ a given DPI $\langle\mo,\mb,\Tp,\Tn\rangle_\RQ$ per call. Since a diagnosis is a hitting set of \emph{all} minimal conflict sets, more than one minimal conflict set is generally required to compute a diagnosis. Due to its determinism, however, $\scQX$ always computes the same minimal conflict set for the same input DPI. Thus, in order to compute different (or all) minimal conflict sets, the input to $\scQX$ needs to be varied accordingly. This can be done by means of $\scHS$ which serves as a search tree to systematically and successively explore all minimal conflict sets w.r.t.\ an initially given DPI. Note that often not all minimal conflict sets w.r.t.\ a DPI are necessary to obtain a minimal diagnosis w.r.t.\ this DPI. This is the case when different minimal conflict sets overlap, i.e.\ have a non-empty intersection. In the extreme case, when all minimal conflict sets w.r.t.\ a DPI share some formulas, then the computation of any single minimal conflict set can suffice to obtain a minimal diagnosis, which is actually even a minimum cardinality diagnosis.

Another approach for computing a minimal conflict set (or justification) is the ``expand-and-shrink'' algorithm presented in \cite{Kalyanpur.Just.ISWC07}. However, empirical evaluations and a theoretical analysis of the best and worst case complexity of the ``expand-and-shrink'' method compared to $\scQX$ performed in \cite{Shchekotykhin2008} revealed that the latter is preferable over the former. 

Also, alternative strategies for the computation of minimal diagnoses have been suggested. One common method is to avoid the indirection of diagnosis computation via minimal conflict sets and use algorithms that determine diagnoses \emph{directly}~\cite{satoh2006}, i.e.\ without the necessity to compute conflict sets. This approach has been applied for the non-interactive debugging of ontologies~\cite{Du2011} and constraints~\cite{Felfernig2011}. In our previous work, we adopted such a direct technique for the interactive debugging of KBs~\cite{Shchekotykhin2014}. The reason why we stick to the conflict-based approach in this work is that we want to present best-first algorithms that figure out minimal diagnoses in descending order of their probability. This is not (systematically) realizable with a direct approach. 

\subsection{Computation of a Minimal Conflict Set}
\label{sec:cs_comp}
The $\scQX$ algorithm takes a DPI $\langle\mo_\orig,\mb_\orig,\Tp,\Tn\rangle_\RQ$ over some monotonic logic $\mathcal{L}$ as input and returns a minimal conflict set $\mc\subseteq\mo_\orig$ w.r.t.\ $\langle\mo_\orig,\mb_\orig,\Tp,\Tn\rangle_\RQ$ as output, if some conflict set exists for the DPI, and 'no conflict' otherwise.
\paragraph{Monotonic Properties.} Basically, $\scQX$ can be employed to find for an input set $X$ a set-minimal subset $X_{\min} \subseteq X$ that has a certain property $prop$ 
for problems of completely different nature such as propositional unsatisfiability or over-constrainedness of constraint satisfaction problems. The only postulated prerequisite for $\scQX$ to work correctly is that $prop$ is a monotonic property. A property is monotonic if and only if the binary function that returns 1 if the property holds for the input set and 0 otherwise is a monotonic function.
\begin{definition}[Binary monotonic function]\label{def:monotonic}
Let $X$ be a set and $f:2^X \rightarrow \setof{0,1}$ be a binary function defined for all subsets of $X$. Then, $f$ is monotonic iff 
\begin{align*}
\forall X', X'' \subseteq X:\;\, X' \subset X'' \land f(X') = 1 \implies f(X'')=1
\end{align*}
\end{definition}
So, $prop$ is monotonic iff, given that $prop$ holds for some set $X'$, it follows that $prop$ also holds for any superset $X''$ of $X'$. Note that, by simple logical transformation, an equivalent statement can be derived from Definition~\ref{def:monotonic}; namely that, given that $prop$ does not hold for some set $X''$, it follows that $prop$ does not hold for any subset $X'$ of $X''$ either. 

As inconsistency and incoherency as well as the entailment of some $\tn\in\Tn$ over some monotonic language $\mathcal{L}$ are clearly monotonic properties, the following proposition holds.
\begin{proposition}
Let $\langle\mo,\mb,\Tp,\Tn\rangle_\RQ$ be a DPI. Then, the invalidity of $\mo' \subseteq \mo$ w.r.t.\ $\langle\cdot,\mb,\Tp,\Tn\rangle_\RQ$ (as per Definition~\ref{def:valid_onto}) is a monotonic property.
\end{proposition}
By Corollary~\ref{cor:validonto_cs}, a (minimal) conflict set w.r.t.\ $\langle\mo,\mb,\Tp,\Tn\rangle_\RQ$ is a (minimal) invalid sub-KB of $\mo$ w.r.t.\ $\langle\cdot,\mb,\Tp,\Tn\rangle_\RQ$. Therefore:
\begin{corollary}
Let $\langle\mo,\mb,\Tp,\Tn\rangle_\RQ$ be a DPI. Then, being a conflict set w.r.t.\ $\langle\mo,\mb,\Tp,\Tn\rangle_\RQ$ is a monotonic property.
\end{corollary}
Thus, $\scQX$ is applicable for the problem of finding a minimal conflict set w.r.t.\ a DPI.  
As we shall see later in Section~\ref{sec:QueryGeneration}, another monotonic property will enable us to apply $\scQX$ also for the minimization of queries asked to an interacting user in the interactive debugging of KBs.

\paragraph{How $\scQX$ (Algorithm~\ref{algo:qx}) Works.} After verifying that the trivial cases, i.e.\ $\mo_\orig$ is already a valid KB w.r.t.\ $\langle\cdot,\mb_\orig,\Tp,\Tn\rangle_\RQ$ or $\mo_\orig = \emptyset$, are not met, a non-empty minimal conflict set w.r.t.\ $\langle\mo_\orig,\mb_\orig$, $\Tp,\Tn\rangle_\RQ$ must exist. So, the algorithm enters the recursive procedure $\scQX'(\emptyset,\langle\mo_\orig,\mb_\orig,\Tp,\Tn\rangle_\RQ)$. Note that the parameters $\Tp, \Tn, \RQ$ of $\scQX'$ are used for validity tests (\textsc{isKBValid}, line~\ref{algoline:validitytest2}) only and are maintained invariant during the entire recursive execution. 
In case $\mo_\orig$ is not a singleton, i.e.\ it does not hold for sure that $\mo_\orig$ is an element of a minimal conflict set w.r.t.\ $\langle\mo_\orig,\mb_\orig,\Tp,\Tn\rangle_\RQ$, the idea is to apply a divide-and-conquer strategy to reduce $\mo_\orig$ into two subproblems and solve one subproblem first, i.e.\ find a minimal conflict set for this subproblem, and then the second subproblem. The union of the minimal conflict sets found for the subproblems is then a minimal conflict set for the original problem. This division into smaller problems is recursively executed for each subproblem until the trivial case, i.e.\ the KB of the subproblem that is analyzed includes only one element, occurs. Then this element is an element of a minimal conflict set w.r.t.\ the original problem.

Simply put, one can imagine that $\scQX$ takes $\mo_\orig$, partitions it into $\mo_1$ and $\mo_2$ and first considers the DPI with KB $\mo_2$ and background knowledge $\mb\cup\mo_1$ (line~\ref{algoline:recursive_call1}). If the latter already includes a conflict set (second condition in line~\ref{algoline:validitytest2}), then $\mo_2$ can be safely discarded and does not need to be further considered. 
Instead, $\mo_1$ is further investigated, i.e.\ the DPI with KB $\mo_{1,2}$ and background knowledge $\mb \cup \mo_{1,1}$ where $\mo_{1,1}$ and $\mo_{2,2}$ partition $\mo_1$. Notice that, in this way, $|\mo_2|$ sentences can be dismissed by a single call to \textsc{isKBValid} which is the only function in Algorithm~\ref{algo:qx} that calls a reasoner.

If, on the other hand, $\mb\cup\mo_1$ includes no conflict set, $\mo_2$ is partitioned into $\mo_{2,1}$ and $\mo_{2,2}$ and the two DPIs, the first with KB $\mo_{2,2}$ and background knowledge $\mb\cup\mo_1\cup\mo_{2,1}$ and the second with KB $\mo_{2,1}$ and background knowledge $\mb \cup \mo_1 \cup \mc_{2,2}$, are recursively analyzed where $\mc_{2,2}$ is the result computed for the first DPI.

This recursion is executed until encountering a trivial case, i.e.\ a leaf node of the recursion tree, along each path. Then, the recursion unwinds by building the union of all leaf nodes, i.e.\ the union of all returned sets for subproblems where a trivial case occurred.

\begin{algorithm}
\small
\caption{$\scQX$: Computation of a Minimal Conflict Set} \label{algo:qx}
\begin{algorithmic}[1]
\Require a DPI $\langle\mo_\orig,\mb_\orig,\Tp,\Tn\rangle_\RQ$
\Ensure a minimal conflict set w.r.t.\ $\langle\mo_\orig,\mb_\orig,\Tp,\Tn\rangle_\RQ$ 

\Procedure{$\scQX$}{$\langle\mo_\orig,\mb_\orig,\Tp,\Tn\rangle_\RQ$}
\If{$\Call{isKBValid}{\mo_\orig, (\mb_\orig, \Tp, \Tn, \RQ)}$} \label{algoline:validitytest1}
    \State \Return~`no conflict'
\ElsIf{$\mo_\orig = \emptyset$}\label{algoline:O=0}
		\State \Return~$\emptyset$\label{algoline:emptyset}
\Else \State \Return \Call{$\scQX'$}{$\emptyset, \langle\mo_\orig,\mb_\orig,\Tp,\Tn\rangle_\RQ$} \label{algoline:call_QX'}
\EndIf
\EndProcedure

\vspace{10pt}

\Procedure{$\scQX'$}{$\mc,\langle\mo,\mb,\Tp,\Tn\rangle_\RQ$}
\If{$\mc \neq \emptyset \land \neg \Call{isKBValid}{\mb,\langle\cdot,\emptyset,\Tp,\Tn\rangle_\RQ}$}\label{algoline:validitytest2} 
	\State \Return $\emptyset$ \label{algoline:return_emptyset}
\EndIf
\If{$|\mo| = 1$}  \label{algoline:test_singleton}              
  \State \Return $\mo$ \label{algoline:return_O}
\EndIf
\State $k \gets \Call{split}{|\mo|}$\label{algoline:split}
\State $\mo_1 \gets \Call{get}{\mo, 1, k}$\label{algoline:get1} 
\State $\mo_2 \gets \Call{get}{\mo, k + 1, |\mo|}$\label{algoline:get2}
\State $\mc_2 \gets \Call{$\scQX'$}{\mo_1, \langle\mo_2,\mb\cup\mo_1,\Tp,\Tn\rangle_\RQ}$ \label{algoline:recursive_call1}
\State $\mc_1 \gets \Call{$\scQX'$}{\mc_2, \langle\mo_1,\mb\cup\mc_2,\Tp,\Tn\rangle_\RQ}$ \label{algoline:recursive_call2}
\State \Return $\mc_1 \cup \mc_2$ \label{algoline:return_upwards}
\EndProcedure

\vspace{10pt}

\Procedure{\textsc{isKBValid}}{$\mo, \langle\cdot,\mb,\Tp,\Tn\rangle_\RQ$}
\State $\mo' \gets \mo \cup \mb \cup \bigcup_{\tp\in\Tp} \tp$
\If{$\neg \Call{verifyReq}{\mo',\RQ}$}
	\State \Return \false
\EndIf
\For{$\tn \in \Tn$}
\If{$\Call{entails}{\mo', \tn}$}
\State \Return \false
\EndIf
\EndFor
\State \Return \true
\EndProcedure

\end{algorithmic}
\normalsize
\end{algorithm}


\renewcommand{\arraystretch}{1.4} 
\begin{table}[h]
\footnotesize
	\centering
		\rowcolors[]{2}{gray!8}{gray!16} 
		\begin{tabular}{ c c c c } 
			\rowcolor{gray!40}
			\toprule\addlinespace[0pt]
			$i$ & $\tax_i$ & $\mo$ & $\mb$  \\ \addlinespace[0pt]\midrule\addlinespace[0pt]
			1 & $A \sqsubseteq B$ & $\bullet$ & 	\\
			2 & $B \sqsubseteq E$ & $\bullet$ &  	\\
			3 & $B \sqsubseteq D \sqcap \lnot \exists s.C$ & $\bullet$ &  	\\
			4 & $C \sqsubseteq \lnot(D \sqcup E)$ & $\bullet$ &  	\\
			5 & $D \sqsubseteq \lnot B$ & $\bullet$ &  	\\
			6 & $A(w)$ &  & $\bullet$   	\\
			7 & $A(v)$ &  & $\bullet$	\\
			8 & $s(v,w)$ & & $\bullet$   	\\ 
			\addlinespace[0pt]\bottomrule 
			\rowcolor{gray!40}
			$i$ & \multicolumn{3}{c}{$\tp_i\in\Tp$} \\ \addlinespace[0pt]\midrule\addlinespace[0pt]
			1 & \multicolumn{3}{c}{$B(w)$} 	\\ \addlinespace[0pt]\toprule\addlinespace[0pt]
			\rowcolor{gray!40}
			$i$ & \multicolumn{3}{c}{$\tn_i\in\Tn$} \\ \addlinespace[0pt]\midrule\addlinespace[0pt]
			1 & \multicolumn{3}{c}{$\lnot C(w)$} 	
			\\ \addlinespace[0pt]\toprule\addlinespace[0pt]
			\rowcolor{gray!40}
			$i$ & \multicolumn{3}{c}{$r_i\in\RQ$} \\ \addlinespace[0pt]\midrule\addlinespace[0pt]
			$1$ &\multicolumn{3}{c}{consistency} \\ 
			$2$ &\multicolumn{3}{c}{coherency} \\ \addlinespace[0pt]\bottomrule
			\end{tabular}
	\caption{Description Logic Example DPI 2}
	\label{tab:example4}
\end{table}

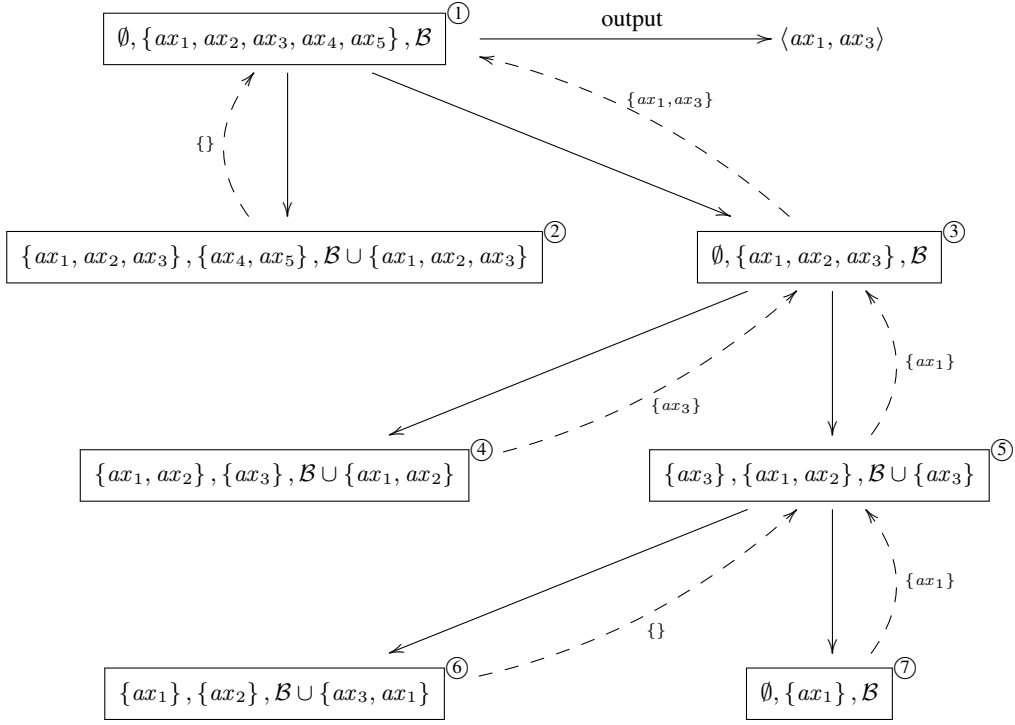
\begin{figure*}[tb]
\centering
\begin{minipage}[c]{0.99\linewidth} 
\small
\begin{displaymath}
\xymatrix{
         \boxed{\emptyset , \setof{\tax_1,\tax_2,\tax_3,\tax_4,\tax_5} , \mb}^{\textcircled{\scriptsize 1}} \ar[dd] \ar[ddr] \ar[r]^{\qquad\qquad\qquad\mbox{output}} &  \tuple{\tax_1,\tax_3}     \\ \\
\boxed{\setof{\tax_1,\tax_2,\tax_3}, \setof{\tax_4,\tax_5}, 
				\mb \cup\setof{\tax_1,\tax_2,\tax_3}}^{\textcircled{\scriptsize 2}} \ar@/^2pc/@{-->}[uu]^-{\setof{}}   & 	
\boxed{\emptyset, \setof{\tax_1,\tax_2,\tax_3}, \mb}^{\textcircled{\scriptsize 3}} \ar[ddl] \ar[dd] \ar@/_2pc/@{-->}[uul]_-{\setof{\tax_1,\tax_3}}  \\ \\
\boxed{\setof{\tax_1,\tax_2}, \setof{\tax_3}, \mb\cup\setof{\tax_1,\tax_2}}^{\textcircled{\scriptsize 4}} \ar@/_2pc/@{-->}[uur]_-{\setof{\tax_3}}  & 
\boxed{\setof{\tax_3}, \setof{\tax_1,\tax_2}, \mb \cup \setof{\tax_3}}^{\textcircled{\scriptsize 5}} \ar[ddl] \ar[dd] \ar@/_2pc/@{-->}[uu]_-{\setof{\tax_1}}  \\ \\
\boxed{\setof{\tax_1}, \setof{\tax_2}, \mb\cup\setof{\tax_3,\tax_1}}^{\textcircled{\scriptsize 6}} \ar@/_2pc/@{-->}[uur]_-{\setof{}} & 
\boxed{\emptyset, \setof{\tax_1}, \mb}^{\textcircled{\scriptsize 7}}  \ar@/_2pc/@{-->}[uu]_-{\setof{\tax_1}}  \\ 
}
\end{displaymath}
\end{minipage}
\caption[Recursion Tree for the Computation of a Minimal Conflict Set]{Recursion tree produced during the computation of the minimal conflict set $\tuple{\tax_1,\tax_3}$ w.r.t.\ the DPI shown by Table~\ref{tab:example4} using Algorithm~\ref{algo:qx}. Nodes in the depicted tree represent calls $\scQX'(\mc,\langle\mo,\mb,\Tp,\Tn\rangle_\RQ)$ and are written in format $\boxed{\mc,\mo,\mb}^{\textcircled{\scriptsize $k$}}$ where $k$ is a counter starting from 1 that indicates when the respective call is made. 
A recursive call to $\scQX'$ (left branch = call in line~\ref{algoline:recursive_call1}; right branch = call in line~\ref{algoline:recursive_call2}) is denoted by a normal arrow whereas the return of a set is visualized by a dashed arrow. 
} 
\label{fig:qx_example}
\end{figure*}

\noindent The next example illustrates one execution of $\scQX$ which computes one minimal conflict set:
\begin{example}\label{example:qx} Let us consider the DL example DPI depicted by Table~\ref{tab:example4}. We will now demonstrate how a minimal conflict set is computed by Algorithm~\ref{algo:qx} (see Fig.~\ref{fig:qx_example}).
Since $\mo$ is not the empty set and not a valid KB w.r.t.\ the DPI (conditions in lines~\ref{algoline:O=0} and \ref{algoline:validitytest1} are false), $\scQX'(\emptyset,\langle\mo,\mb,\Tp,\Tn\rangle_{\RQ})$ is called in line~\ref{algoline:call_QX'}. This call is illustrated by the root node (node \textcircled{\scriptsize 1}) of the recursion tree given in Fig.~\ref{fig:qx_example} (whereas the evaluations made by $\scQX$ prior to this call are not depicted in the figure). Notice that each node in the tree shows only the values of $\mc$, $\mo$ and $\mb$ since all other parameters $\Tp$, $\Tn$ and $\RQ$ are invariant throughout the entire execution of Algorithm~\ref{algo:qx}. 

Due to the fact that $\mc = \emptyset$ and $\mo$ includes five formulas and is thus not a singleton, $\mo=\setof{\tax_1,\dots,\tax_5}$ is partitioned into $\mo_1=\setof{\tax_1,\tax_2,\tax_3}$ and $\mo_2=\setof{\tax_4,\tax_5}$ and $\scQX'$ is recursively called in line~\ref{algoline:recursive_call1} with parameters $\mc = \mo_1$, $\mo = \mo_2$ and $\mb = \mb\cup\setof{\tax_1,\tax_2,\tax_3}$ which is expressed in the figure by a left branch to node \textcircled{\scriptsize 2}. 
This call, however, returns $\emptyset$ directly since $\mb\cup\setof{\tax_1,\tax_2,\tax_3}$ is already invalid w.r.t.\ $\langle\cdot,\emptyset,\Tp,\Tn\rangle_\RQ$ because $\mb\cup\setof{\tax_1,\tax_2,\tax_3}\cup U_\Tp = \setof{A(w), \underline{A(v)}, \underline{s(v,w)}} \cup \setof{\underline{A \sqsubseteq B}, B \sqsubseteq E, \underline{B \sqsubseteq} D \sqcap \underline{\lnot \exists s.C}} \cup \setof{\setof{B(w)}} \models \setof{\lnot C(w)}$ which is a negative test case, i.e.\ must not be entailed by a solution KB w.r.t.\ the input DPI (the parts of the formulas relevant for the entailment to hold are underlined). Returning $\emptyset$ in this case means discarding $\mo_2 = \setof{\tax_4,\tax_5}$. 

So, the algorithm opens a right branch from the root to node \textcircled{\scriptsize 3} by calling $\scQX'$ (line~\ref{algoline:recursive_call2}) with parameters $\mc = \emptyset$ (result of left branch), $\mo=\mo_1=\setof{\tax_1,\tax_2,\tax_3}$ and $\mb = \mb$. During the execution of this call $\mo_1$ is partitioned into $\setof{\tax_1,\tax_2}$ (left branch to node \textcircled{\scriptsize 4}) and $\setof{\tax_3}$ (right branch to node \textcircled{\scriptsize 5}). In node \textcircled{\scriptsize 4}, it holds that $\mb \cup \setof{\tax_1,\tax_2}$ can be extended to a solution KB by adding $U_\Tp$, i.e.\ $\mb \cup \setof{\tax_1,\tax_2}$ is valid. As it is already an established fact since the execution of node \textcircled{\scriptsize 2} that $\mb \cup \setof{\tax_1,\tax_2,\tax_3}$ is invalid, it must be the case that $\tax_3$ is an element of a minimal conflict set w.r.t.\ the input DPI (as there is a conflict set w.r.t.\ the input DPI in $\setof{\tax_1,\tax_2,\tax_3}$, but there is none in $\setof{\tax_1,\tax_2}$). The algorithm accounts for that by checking whether $\mo$ is a singleton (line~\ref{algoline:test_singleton}) in which case it is guaranteed that $\mo$ is a subset of a minimal conflict set w.r.t.\ the input DPI. So, node \textcircled{\scriptsize 4} returns $\setof{\tax_3}$. This procedure is continued until each path from the root node reaches a node where a trivial case is met. Then the recursion unwinds and, when arrived at the root node, the minimal conflict set $\tuple{\tax_1,\tax_3}$ is returned.

That $\mc := \tuple{\tax_1,\tax_3}$ is indeed a conflict set can be recognized easily by the underlinings in the formulas given before. Minimality is given since $\mb\cup\mc\cup U_\Tp$ is neither inconsistent nor incoherent and the deletion of any formula from $\mc$ breaks the entailment of $\tn_1$. Hence, $\scQX$ has returned a sound output.
\qed

\end{example}

The complexity of Algorithm~\ref{algo:qx} in terms of the number of calls to the function \textsc{isKBValid}, which is the only place in the algorithm where a reasoning service is consulted, is captured by the following proposition.
\begin{proposition}[Complexity of $\scQX$]\label{prop:qx_complexity}\cite{junker04}
Let $\langle\mo,\mb,\Tp,\Tn\rangle_\RQ$ be a DPI and the function \textsc{split} (line~\ref{algoline:split} of Algorithm~\ref{algo:qx}) be defined as $\textsc{split}(n) = \lfloor \frac{n}{2}\rfloor$ where $n$ is a natural number. Then, the worst case number of calls to \textsc{isKBValid} during one call to $\scQX(\langle\mo,\mb,\Tp,\Tn\rangle_\RQ)$ is in $O(|\mc|\log \frac{|\mo|}{|\mc|})$ where $\mc$ is the output of $\scQX(\langle\mo,\mb,\Tp,\Tn\rangle_\RQ)$.

For any other definition of the function \textsc{split}, the worst case number of {\textsc{isKBValid}} invocations gets larger.
\end{proposition}

\subsection{Correctness of Conflict Set Computation}
\label{sec:cs_comp_correctness}
This section is dedicated to the proof of correctness of Algorithm~\ref{algo:qx}. First, we show some essential properties of $\scQX$ by various Lemmata which will finally be exploited to demonstrate the overall soundness of $\scQX$. 

The $\scQX$ algorithm accepts a DPI $\langle\mo_\orig,\mb_\orig,\Tp,\Tn\rangle_\RQ$ over some monotonic language $\mathcal{L}$ 
as input and returns a minimal conflict set $\mc\subseteq\mo_\orig$ w.r.t.\ $\langle\mo_\orig,\mb_\orig,\Tp,\Tn\rangle_\RQ$ as output. 
First, the algorithm checks whether $\mo_\orig$ is a valid KB w.r.t.\ the input DPI $\langle\cdot,\mb_\orig,\Tp,\Tn\rangle_\RQ$ (line~\ref{algoline:validitytest1}). If so, there is no conflict set for the DPI by Proposition~\ref{prop:non_validonto_includes_cs} and the algorithm returns 'no conflict'. Otherwise, the test $\mo_\orig = \emptyset$ is performed (line~\ref{algoline:O=0}). If so, then the negative outcome of the validity test executed in line~\ref{algoline:validitytest1} actually means that one of the two criteria of Proposition~\ref{prop:exist_diag} is violated which, by Definition~\ref{def:admissible}, implies that the DPI is not admissible. Invalidity of $\mo_\orig$ w.r.t.\ $\langle\cdot,\mb_\orig,\Tp,\Tn\rangle_\RQ$ and non-admissiblity of $\langle\mo_\orig,\mb_\orig,\Tp,\Tn\rangle_\RQ$ mean that there is only one minimal conflict set $\mc = \emptyset$ by Proposition~\ref{prop:cs_admissible}. Thus, $\emptyset$ is returned in line~\ref{algoline:emptyset}.

\begin{lemma}\label{lem:qx_recursion_finds_cs}
Let $\langle\mo,\mb,\Tp,\Tn\rangle_\RQ$ be an admissible DPI and $\mo$ be invalid w.r.t.\ $\langle\cdot,\mb,\Tp,\Tn\rangle_\RQ$. Then, there is a minimal conflict set $\mc \supset \emptyset$ w.r.t.\ $\langle\mo,\mb,\Tp,\Tn\rangle_\RQ$. 
\end{lemma}
\begin{proof}
The proposition is a direct consequence of Proposition~\ref{prop:cs_admissible}.
\end{proof}

So, if both initial tests (lines~\ref{algoline:validitytest1} and \ref{algoline:O=0}) are negative, then, by Lemma~\ref{lem:qx_recursion_finds_cs}, there is a non-trivial minimal conflict set w.r.t.\ $\langle\mo_\orig,\mb_\orig,\Tp,\Tn\rangle_\RQ$ 
wherefore the algorithm enters the recursion by a call to the procedure $\scQX'$. 

The argumentation so far proves the following lemma.
\begin{lemma}\label{lem:qx_start_conditions}
\leavevmode
\begin{itemize}
\item \label{lem_enum:no_conflict} $\scQX(\langle\mo,\mb,\Tp,\Tn\rangle_\RQ)$ returns 'no conflict' iff there is no (minimal) conflict w.r.t.\ $\langle\mo,\mb,\Tp,\Tn\rangle_\RQ$.
\item \label{lem_enum:emptyset} $\scQX(\langle\mo,\mb,\Tp,\Tn\rangle_\RQ)$ returns $\emptyset$ iff $\emptyset$ is the only (minimal) conflict w.r.t.\ $\langle\mo,\mb,\Tp,\Tn\rangle_\RQ$.
\item \label{lem_enum:qx'} $\scQX(\langle\mo,\mb,\Tp,\Tn\rangle_\RQ)$ returns $\scQX'(\emptyset,\langle\mo,\mb,\Tp,\Tn\rangle_\RQ)$ iff there is some minimal conflict $\mc \supset \emptyset$ w.r.t.\ $\langle\mo,\mb,\Tp,\Tn\rangle_\RQ$.
\end{itemize}
\end{lemma}

\begin{corollary}\label{cor:call_QX'_admissible}
$\scQX(\langle\mo,\mb,\Tp,\Tn\rangle_\RQ)$ returns $\scQX'(\emptyset,\langle\mo,\mb,\Tp,\Tn\rangle_\RQ)$ iff $\langle\mo,\mb,\Tp,\Tn\rangle_\RQ$ is an admissible DPI.
\end{corollary}
\begin{proof}
By the third proposition of Lemma~\ref{lem:qx_start_conditions} and Proposition~\ref{prop:non_validonto_includes_cs} we have that $\scQX(\langle\mo,\mb,\Tp,\Tn\rangle_\RQ)$ returns $\scQX'(\emptyset,\langle\mo,\mb,\Tp,\Tn\rangle_\RQ)$ iff $\mo$ is invalid w.r.t.\ $\langle\cdot,\mb,\Tp,\Tn\rangle_\RQ$. By Proposition~\ref{prop:cs_admissible}, we can then conclude that $\scQX(\langle\mo,\mb,\Tp,\Tn\rangle_\RQ)$ returns $\scQX'(\emptyset,\langle\mo,\mb,\Tp,\Tn\rangle_\RQ)$ iff $\langle\mo,\mb,\Tp,\Tn\rangle_\RQ$ is an admissible DPI.
\end{proof}

The input arguments (at any call) to $\scQX'$ are (a)~some subset $\mc$ of the original input KB $\mo_\orig$ to $\scQX$ and (b)~a DPI $\langle\mo,\mb,\Tp,\Tn\rangle_\RQ$ where $\mo \subseteq \mo_\orig$ and $\mb \supseteq \mb_\orig$. 

The principle of $\scQX'$ relies on the following fact.
\begin{lemma}\label{lem:qx_recursion_principle}\cite{junker04}
Let $\mo_1, \mo_2$ be a partition of $\mo$. If $\mc_2$ is a minimal conflict set w.r.t.\ $\langle\mo_2, \mb \cup \mo_1,\Tp,\Tn\rangle_\RQ$ and $\mc_1$ is a minimal conflict set w.r.t.\ $\langle\mo_1, \mb \cup \mc_2,\Tp,\Tn\rangle_\RQ$, then $\mc_1 \cup \mc_2$ is a minimal conflict set w.r.t.\ $\langle\mo_1 \cup \mo_2, \mb,\Tp,\Tn\rangle_\RQ = \langle\mo, \mb,\Tp,\Tn\rangle_\RQ$.
\end{lemma}
\begin{proof}
Since $\mc_1$ is a minimal conflict set w.r.t.\ $\langle\mo_1, \mb \cup \mc_2,\Tp,\Tn\rangle_\RQ$, we have that $\mc_1$ is invalid w.r.t.\ $\langle\cdot, \mb \cup \mc_2,\Tp,\Tn\rangle_\RQ$. From that we obtain that $\mc_1 \cup \mc_2$ must be invalid w.r.t.\ $\langle\cdot, \mb,\Tp,\Tn\rangle_\RQ$. Further on, by the fact that $\mo_1,\mo_2$ partition $\mo$ we have that $\mc_1 \subseteq \mo_1 \subseteq \mo$ since $\mc_1$ is a minimal conflict set w.r.t.\ $\langle\mo_1, \mb \cup \mc_2,\Tp,\Tn\rangle_\RQ$ and $\mc_2 \subseteq \mo_2 \subseteq \mo$ since $\mc_2$ is a minimal conflict set w.r.t.\ $\langle\mo_2, \mb \cup \mo_1,\Tp,\Tn\rangle_\RQ$. Consequently, $\mc_1\cup\mc_2 \subseteq\mo$ must be true. So, by Corollary~\ref{cor:validonto_cs}, $\mc_1 \cup \mc_2$ is a conflict set w.r.t.\ $\langle\mo, \mb,\Tp,\Tn\rangle_\RQ$. 

To show the minimality of $\mc_1 \cup \mc_2$, assume that $\mc \subset \mc_1 \cup \mc_2$ is a minimal conflict set w.r.t.\ $\langle\mo$, $\mb,\Tp,\Tn\rangle_\RQ$. Due to $\mo_1 \cap \mo_2 = \emptyset$ and $\mc_1 \subseteq \mo_1$ and $\mc_2 \subseteq \mo_2$, it must hold that $\mc_1\cap\mc_2 = \emptyset$. Thus, (1)~$\mc \cap \mc_1 \subset \mc_1$ or (2)~$\mc \cap \mc_2 \subset \mc_2$. 

Let us assume (1) holds. Then, $\mc$ is invalid w.r.t.\ $\langle\cdot, \mb,\Tp,\Tn\rangle_\RQ$, i.e.\ $\mc \cup \mb \cup U_\Tp = (\mc'_1 \cup \mc_2) \cup \mb \cup U_\Tp = \mc'_1 \cup (\mb \cup \mc_2) \cup U_\Tp$ violates some $r\in\RQ$ or some $\tn\in\Tn$ where $\mc'_1 \subset \mc_1$. This, however, is a contradiction to the minimality of the conflict set $\mc_1$ w.r.t.\ $\langle\mo_1, \mb \cup \mc_2,\Tp,\Tn\rangle_\RQ$.

Now, let us assume (2) holds. Then, $\mc$ is invalid w.r.t.\ $\langle\cdot, \mb,\Tp,\Tn\rangle_\RQ$, i.e.\ $\mc \cup \mb \cup U_\Tp = (\mc_1 \cup \mc'_2) \cup \mb \cup U_\Tp$ violates some $r\in\RQ$ or some $\tn\in\Tn$ where $\mc'_2 \subset \mc_2$. By monotonicity of $\mathcal{L}$ and $\mc_1 \subseteq \mo_1$, this implies $\mc'_2 \cup (\mo_1 \cup \mb) \cup U_\Tp$ violates some $r\in\RQ$ or some $\tn\in\Tn$, i.e.\ $\mc'_2 \subset \mo_2$ is a conflict set w.r.t.\ 
$\langle\mo_2, \mb \cup \mo_1,\Tp,\Tn\rangle_\RQ$ which is a contradiction due to $\mc'_2 \subset \mc_2$ and the minimality of the conflict set $\mc_2$ w.r.t.\ 
$\langle\mo_2, \mb \cup \mo_1,\Tp,\Tn\rangle_\RQ$.
\end{proof}

$\scQX'(\mc,\langle\mo,\mb,\Tp,\Tn\rangle_\RQ)$ computes a minimal conflict set w.r.t.\ $\langle\mo,\mb,\Tp,\Tn\rangle_\RQ$ in a divide-and-conquer fashion whereby the argument $\mc$ is the set of sentences of $\mo_\orig$ that has been added to $\mb$ in the current iteration. That is, in this iteration $\scQX'$ will output either (1)~$\emptyset$ if the current $\mb$ (which includes $\mc$) already contains a minimal conflict set w.r.t.\ the original DPI $\langle\mo_\orig,\mb_\orig,\Tp,\Tn\rangle_\RQ$ or (2)~a minimal conflict set w.r.t.\ the current DPI $\langle\mo,\mb,\Tp,\Tn\rangle_\RQ$ (i.e.\ a subset of a minimal conflict set w.r.t.\ the original DPI) which does not include any sentence from $\mc$.

\begin{lemma}\label{lem:qx'_termination_etc}
\leavevmode
\begin{enumerate}
	\item \label{lem_enum:c_sub_b} For each call $\scQX'(\mc,\langle\mo,\mb,\Tp,\Tn\rangle_\RQ)$ within Algorithm~\ref{algo:qx} it holds that $\mc \subseteq \mb$.
	\item \label{lem_enum:c_neq_emptyset} If $\scQX'(\mc,\langle\mo,\mb,\Tp,\Tn\rangle_\RQ)$ is called in line~\ref{algoline:recursive_call1} of Algorithm~\ref{algo:qx}, $\mc \neq \emptyset$ holds.
	\item \label{lem_enum:return_emptyset} If $\scQX'(\mc,\langle\mo,\mb,\Tp,\Tn\rangle_\RQ)$ returns $\emptyset$, then there is some non-empty minimal conflict set w.r.t.\ $\langle\mc,\mb\setminus\mc,\Tp,\Tn\rangle_\RQ$.
	\item \label{lem_enum:emptyset_only_cs} If $\scQX'(\mc,\langle\mo,\mb,\Tp,\Tn\rangle_\RQ)$ returns $\emptyset$, then $\emptyset$ is the only minimal conflict set w.r.t.\ $\langle\mo,\mb,\Tp,\Tn\rangle_\RQ$.
	\item \label{lem_enum:terminates} $\scQX'(\mc,\langle\mo,\mb,\Tp,\Tn\rangle_\RQ)$ terminates.
	
\end{enumerate}
\end{lemma}
\begin{proof}
\leavevmode\\
\indent 1): There are three situations when $\scQX'(\mc,\langle\mo,\mb,\Tp,\Tn\rangle_\RQ)$ is called within Algorithm~\ref{algo:qx}, namely in lines~\ref{algoline:call_QX'}, \ref{algoline:recursive_call1} and \ref{algoline:recursive_call2}. In line~\ref{algoline:call_QX'}, $\mc := \emptyset \subseteq \mb$ holds. In line~\ref{algoline:recursive_call1}, $\mc:=\mo_1 \subseteq \mb \cup \mo_1 =: \mb$ holds. In line~\ref{algoline:recursive_call2}, $\mc:=\mc_2 \subseteq \mb \cup \mc_2 =: \mb$ holds.

2): In line~\ref{algoline:recursive_call1}, $\scQX'$ is called with $\mc:=\mo_1$, which is always not the empty set due to the definition of the \textsc{split} function in line~\ref{algoline:split} that is used to extract $\mo_1$ from $\mo$.

3): The first observation is that $\scQX'(\mc,\langle\mo,\mb,\Tp,\Tn\rangle_\RQ)$ cannot return $\emptyset$ if $\mc = \emptyset$ as in this case the first condition in line~\ref{algoline:validitytest2} is not met. Thus, in particular, $\scQX'$ cannot return $\emptyset$ if called in line~\ref{algoline:call_QX'}.

So, $\emptyset$ can be returned by $\scQX'(\mc,\langle\mo,\mb,\Tp,\Tn\rangle_\RQ)$ only if it is called (1)~in line~\ref{algoline:recursive_call1} or (2)~in line~\ref{algoline:recursive_call2}. 

If $\scQX'(\mc,\langle\mo,\mb,\Tp,\Tn\rangle_\RQ)$ returns $\emptyset$, then $\mc \neq \emptyset$ and $\mb$ is invalid w.r.t.\ $\langle\cdot,\emptyset,\Tp,\Tn\rangle_\RQ$ (line~\ref{algoline:validitytest2}), i.e.\ $\mb$ contains a minimal conflict set w.r.t.\ $\tuple{\mb,\emptyset, \Tp, \Tn}_\RQ$ which is non-empty by Proposition~\ref{prop:cs_admissible} since $\tuple{\mb,\emptyset, \Tp, \Tn}_\RQ$ is an admissible DPI by admissibility of the input DPI and the invariance of $\Tp, \Tn, \RQ$ throughout $\scQX'$. Additionally, $\mc \subseteq \mb$ holds by the first proposition of this lemma. Now, assume that there is no non-empty (minimal) conflict set w.r.t.\ $\langle\mc,\mb\setminus\mc,\Tp,\Tn\rangle_\RQ$. Then, for each minimal conflict set $\mc'$ (which we know is non-empty) w.r.t.\ $\langle\mb,\emptyset,\Tp,\Tn\rangle_\RQ$ it must hold that $\mc \cap \mc' = \emptyset$, i.e.\ there is already a non-empty minimal conflict set w.r.t.\ $\langle\mb\setminus\mc,\emptyset,\Tp,\Tn\rangle_\RQ$. 

Case~(1): Let us assume first that the call to $\scQX'$ was made in line~\ref{algoline:recursive_call1}. Then, before this call to $\scQX'$, $\mb$ was exactly $\mb\setminus\mc$. By the second proposition of this lemma, $\mc\neq\emptyset$ as $\scQX'$ was called in line~\ref{algoline:recursive_call1}. Thus, before the current call to $\scQX'$, the algorithm must have already returned $\emptyset$ (both conditions in line~\ref{algoline:validitytest2} are met) in line~\ref{algoline:return_emptyset} which is a contradiction to the assumption that $\scQX'(\mc,\langle\mo,\mb,\Tp,\Tn\rangle_\RQ)$ was called in line~\ref{algoline:recursive_call1}.

Case~(2): Now, assume that the call to 
$\scQX'(\mc_2,\langle\mo_1,\mb\cup\mc_2,\Tp,\Tn\rangle_\RQ)$ 
was made in line~\ref{algoline:recursive_call2}. Then $\mc_2$ is the result of the call to $\scQX'(\mo_1,\langle\mo_2,\mb\cup\mo_1,\Tp,\Tn\rangle_\RQ)$ in line~\ref{algoline:recursive_call1}. By the argumentation above, we have that $\mc_2 \neq \emptyset$ and there is a non-empty minimal conflict set w.r.t.\ $\langle\mb\cup\mc_2,\emptyset,\Tp,\Tn\rangle_\RQ$. Moreover, we have that there is a non-empty minimal conflict set w.r.t.\ $\langle\mb,\emptyset,\Tp,\Tn\rangle_\RQ$. However, as $\scQX'(\mo_1,\langle\mo_2,\mb\cup\mo_1,\Tp,\Tn\rangle_\RQ)$ in line~\ref{algoline:recursive_call1} did not return $\emptyset$ and $\mo_1 \neq \emptyset$ by the second proposition of this lemma, it must hold that $\mb\cup\mo_1$ is valid w.r.t.\ $\langle\cdot,\emptyset,\Tp,\Tn\rangle_\RQ$, i.e.\ there is no (minimal) conflict set w.r.t.\ $\langle\mb\cup\mo_1,\emptyset,\Tp,\Tn\rangle_\RQ$. By monotonicity of $\mathcal{L}$, this is a contradiction to the fact that there is a non-empty minimal conflict set w.r.t.\ $\langle\mb,\emptyset,\Tp,\Tn\rangle_\RQ$.

4): Assume $\scQX'(\mc,\langle\mo,\mb,\Tp,\Tn\rangle_\RQ)$ returns $\emptyset$ and there is some non-empty minimal conflict set w.r.t.\ $\langle\mo,\mb,\Tp,\Tn\rangle_\RQ$. Since $\emptyset$ is returned, both conditions in line~\ref{algoline:validitytest1} must be met, i.e.\ 
in particular $\mb$ must be invalid w.r.t.\ $\langle\cdot,\emptyset,\Tp,\Tn\rangle_\RQ$ which means that $\langle\mo,\mb,\Tp,\Tn\rangle_\RQ$ is not admissible. By Proposition~\ref{prop:cs_admissible}, there cannot be a non-empty (minimal) conflict set w.r.t.\ $\langle\mo,\mb,\Tp,\Tn\rangle_\RQ$. This yields a contradiction.

5): $\scQX'(\mc,\langle\mo,\mb,\Tp,\Tn\rangle_\RQ)$ either returns $\emptyset$ in line~\ref{algoline:return_emptyset} iff the conditions in line~\ref{algoline:validitytest2} are met or otherwise returns $\mo$ in line~\ref{algoline:return_O} iff $|\mo|=1$ or otherwise calls itself recursively in lines~\ref{algoline:recursive_call1} and \ref{algoline:recursive_call2}. However, for each recursive call $\scQX'(\mc',\langle\mo',\mb',\Tp,\Tn\rangle_\RQ)$ within $\scQX'(\mc,\langle\mo,\mb,\Tp,\Tn\rangle_\RQ)$ it holds that $\mo' \subset \mo$ as $\mo' \in \setof{\mo_1,\mo_2}$ and $\mo_1, \mo_2 \subset \mo$ due to the definition of the \textsc{split} function in line~\ref{algoline:split} that is used to compute $\mo_1$ and $\mo_2$ from $\mo$ in lines~\ref{algoline:get1} and \ref{algoline:get2}. Hence, each recursive call must finally reach the stopping criterion $|\mo|=1$ and return $\mo$ if it does not reach the stopping criterion in line~\ref{algoline:validitytest2} before.
\end{proof}
\begin{lemma}\label{lem:one_rec_call_adm}
Let $\langle\mo,\mb,\Tp,\Tn\rangle_\RQ$ be an admissible DPI. If $\scQX'(\mc,\langle\mo,\mb,\Tp,\Tn\rangle_\RQ)$ is called, then at least one of the immediate recursive calls of $\scQX'$ in line~\ref{algoline:recursive_call1} or line~\ref{algoline:recursive_call2} is given an admissible DPI as argument. 
\end{lemma}
\begin{proof}
Let us assume that $\langle\mo,\mb,\Tp,\Tn\rangle_\RQ$ is an admissible DPI. Within $\scQX'(\mc$, $\langle\mo,\mb$, $\Tp,\Tn\rangle_\RQ)$, the immediate recursive call is $\scQX'(\mo_1,\langle\mo_2,\mb\cup\mo_1,\Tp,\Tn\rangle_\RQ)$ in line~\ref{algoline:recursive_call1} and $\scQX'(\mc_2,\langle\mo_1,\mb\cup\mc_2,\Tp,\Tn\rangle_\RQ)$ in line~\ref{algoline:recursive_call2} where $\mo_1,\mo_2$ is a partition of $\mo$ and $\mc_2$ is the result of $\scQX'(\mo_1,\langle\mo_2,\mb\cup\mo_1,\Tp,\Tn\rangle_\RQ)$. If $\langle\mo_2,\mb\cup\mo_1,\Tp,\Tn\rangle_\RQ$ is admissible, then the proposition of the lemma is fulfilled. So, assume that that $\langle\mo_2,\mb\cup\mo_1,\Tp,\Tn\rangle_\RQ$ is not admissible. Due to this non-admissibility, it must hold that $\mb\cup\mo_1$ is invalid w.r.t.\ $\langle\cdot,\emptyset,\Tp,\Tn\rangle_\RQ$, so the second condition in line~\ref{algoline:validitytest1} is met. As the call to $\scQX'(\mo_1,\langle\mo_2,\mb\cup\mo_1,\Tp,\Tn\rangle_\RQ)$ was made in line~\ref{algoline:recursive_call1}, it must be true by Lemma~\ref{lem:qx'_termination_etc}, prop.~\ref{lem_enum:c_neq_emptyset} that $\mo_1 \neq \emptyset$ wherefore the first condition in line~\ref{algoline:validitytest1} is met as well. Thus, the result of the call of $\scQX'$ in line~\ref{algoline:recursive_call1} must be $\emptyset$. So, the call of $\scQX'$ in line~\ref{algoline:recursive_call2} looks like $\scQX'(\emptyset,\langle\mo_1,\mb,\Tp,\Tn\rangle_\RQ)$. However, the DPIs $\langle\mo_1,\mb,\Tp,\Tn\rangle_\RQ$ and $\langle\mo,\mb,\Tp,\Tn\rangle_\RQ$ are identical except for the first entries, i.e.\ $\mo_1$ and $\mo$. We know that the latter DPI is admissible. Due to the fact that admissibility of a DPI is defined independently of the KB (the first entry of the DPI tuple), we have that $\langle\mo_1,\mb,\Tp,\Tn\rangle_\RQ$ must be admissible. This completes the proof.
\end{proof}
As long as the algorithm goes downwards in the recursion tree (and has never gone upwards), (1)~the invariant that a minimal conflict set exists for each recursive call to $\scQX'$ holds, (2)~each call to $\scQX'$ that returns, returns a singleton or empty set and (3)~the two calls to $\scQX'$ immediately before going upwards in the recursion tree for the first time must both return either a singleton or an empty set.

\begin{lemma}[QX: Downwards Correctness]\label{lem:qx_downwards}
Let $\langle\mo,\mb,\Tp,\Tn\rangle_\RQ$ be an admissible DPI and let
	there be a non-empty minimal conflict set w.r.t.\ $\langle\mo,\mb,\Tp,\Tn\rangle_\RQ$. Then, the following propositions hold: 
\begin{enumerate}	
	\item \label{lem_enum:downwards_1} Before line~\ref{algoline:return_upwards} has ever been reached during the execution of $\scQX'(\mc,\langle\mo$, $\mb,\Tp$, $\Tn\rangle_\RQ)$, the following holds: If some call to $\scQX'(\mc',\langle\mo',\mb',\Tp,\Tn\rangle_\RQ)$ returns a set $S$, then $S = \emptyset$ or $|S| = 1$.
	\item \label{lem_enum:downwards_2} Before line~\ref{algoline:return_upwards} has ever been reached during the execution of $\scQX'(\mc,\langle\mo$, $\mb,\Tp$, $\Tn\rangle_\RQ)$, the following holds: If $\scQX'(\mc',\langle\mo',\mb',\Tp,\Tn\rangle_\RQ)$ is recursively called, then there is some non-empty minimal conflict set w.r.t.\ $\langle\mo'\cup\mc',\mb'\setminus\mc',\Tp,\Tn\rangle_\RQ$.
	\item \label{lem_enum:downwards_3} Before line~\ref{algoline:return_upwards} has ever been reached during the execution of $\scQX'(\mc,\langle\mo,\mb,\Tp$, $\Tn\rangle_\RQ)$, the following holds: If some call to $\scQX'(\mc',\langle\mo',\mb',\Tp,\Tn\rangle_\RQ)$ returns a set $S$, then 
		$S$ is a minimal conflict set w.r.t.\ $\langle\mo,\mb,\Tp,\Tn\rangle_\RQ$.
	\item \label{lem_enum:downwards_4} When line~\ref{algoline:return_upwards} is reached for the first time, each of the calls to $\scQX'$ immediately before in lines~\ref{algoline:recursive_call1} and \ref{algoline:recursive_call2} must have returned $\emptyset$ or some $\mo$ with $|\mo|=1$.
\end{enumerate}
\end{lemma}
\begin{proof}
\leavevmode\\
\indent 1): Assume the opposite, i.e.\ some call to $\scQX'(\mc',\langle\mo',\mb',\Tp,\Tn\rangle_\RQ)$ returns a set $S$ with $|S| > 1$ before line~\ref{algoline:return_upwards} has ever been reached. There are three places where $\scQX'$ can return, namely in line~\ref{algoline:return_emptyset}, in line~\ref{algoline:return_O} or in line~\ref{algoline:return_upwards}. However, in line~\ref{algoline:return_emptyset}, only $\emptyset$ and in line~\ref{algoline:return_O} only a singleton set can be returned. That is, $S$ must be returned in line~\ref{algoline:return_upwards} which is a contradiction to the assumption that line~\ref{algoline:return_upwards} has not yet been reached.

2): \emph{Induction Base:} The first recursive call $\scQX'(\mc',\langle\mo',\mb',\Tp,\Tn\rangle_\RQ)$ can only occur at line~\ref{algoline:recursive_call1} where $\mc' = \mo_1$, $\mo'=\mo_2$ and $\mb' = \mb \cup \mo_1$ and $\mo_1,\mo_2$ is a partition of $\mo$ as per the definition of the \textsc{split} and \textsc{get} functions in lines~\ref{algoline:split}-\ref{algoline:get2}. So, $\mo'\cup\mc' = \mo$ and $\mb'\setminus\mc' = \mb$. The latter holds since $\mc' \subseteq \mo$ and for each DPI $\mo \cap \mb =\emptyset$ holds by Definition~\ref{def:dpi}. As there is a non-empty minimal conflict set w.r.t.\ $\langle\mo,\mb,\Tp,\Tn\rangle_\RQ$ we have that there is a non-empty minimal conflict set w.r.t.\ $\langle\mo'\cup\mc',\mb'\setminus\mc',\Tp,\Tn\rangle_\RQ$ by the fact that $\langle\mo,\mb,\Tp,\Tn\rangle_\RQ=\langle\mo'\cup\mc',\mb'\setminus\mc',\Tp,\Tn\rangle_\RQ$. Thus, the existence of a non-empty minimal conflict set w.r.t.\ $\langle\mo'\cup\mc',\mb'\setminus\mc',\Tp,\Tn\rangle_\RQ$ is given 
during the execution of the first
recursive call to $\scQX'$.

\emph{Induction Assumption:} Now, let us assume that the existence of a non-empty minimal conflict set w.r.t.\ $\langle\mo\cup\mc,\mb\setminus\mc,\Tp,\Tn\rangle_\RQ$ is given during some call $\scQX'(\mc,\langle\mo,\mb,\Tp$, $\Tn\rangle_\RQ)$. The goal is now to show that the existence of a non-empty minimal conflict set w.r.t.\ $\langle\mo'\cup\mc',\mb'\setminus\mc',\Tp,\Tn\rangle_\RQ$ is given during any recursive call $\scQX'(\mc',\langle\mo',\mb',\Tp,\Tn\rangle_\RQ)$ that is invoked during execution of $\scQX'(\mc,\langle\mo,\mb,\Tp,\Tn\rangle_\RQ)$.

\emph{Induction Step:} Now, there are three cases where this recursive call to $\scQX'$ can take place, namely (1)~in line~\ref{algoline:recursive_call1}, (2)~in line~\ref{algoline:recursive_call2} where the result of $\scQX'$ in line~\ref{algoline:recursive_call1} is $\mc_2 = \emptyset$ and (3)~in line~\ref{algoline:recursive_call2} where the result of $\scQX'$ in line~\ref{algoline:recursive_call1} is some $\mc_2$ with $|\mc_2|=1$. The case where some $\mc_2$ with $|\mc_2|>1$ is returned by $\scQX'$ in line~\ref{algoline:recursive_call1}, is impossible due to the assumption that line~\ref{algoline:return_upwards} has not yet been reached and the first proposition of this lemma.

Case (1): Let us assume that the call $\scQX'(\mc',\langle\mo',\mb',\Tp,\Tn\rangle_\RQ)$ is made in line~\ref{algoline:recursive_call1}. Since that call is made within $\scQX'(\mc,\langle\mo,\mb,\Tp,\Tn\rangle_\RQ)$, it must hold that some condition in line~\ref{algoline:validitytest1} during $\scQX'(\mc,\langle\mo,\mb,\Tp$, $\Tn\rangle_\RQ)$ is violated, as otherwise a return would have taken place in line~\ref{algoline:return_emptyset} which is a contradiction to the assumption that $\scQX'(\mc',\langle\mo',\mb',\Tp,\Tn\rangle_\RQ)$ is called in line~\ref{algoline:recursive_call1}. 

Let us first assume that $\mc = \emptyset$ holds. In this case, the first condition in line~\ref{algoline:validitytest1} is violated and, by the \emph{Induction Assumption}, it is true that there is a non-empty minimal conflict set w.r.t.\ the DPI $\langle\mo\cup\mc,\mb\setminus\mc,\Tp,\Tn\rangle_\RQ$ which is equal to the DPI $\langle\mo,\mb,\Tp,\Tn\rangle_\RQ$ by $\mc = \emptyset$. So, an equal argumentation to the one of the \emph{Induction Base} can be applied to derive that there is a non-empty minimal conflict set w.r.t.\ $\langle\mo'\cup\mc',\mb'\setminus\mc',\Tp,\Tn\rangle_\RQ$.

If $\mc \neq \emptyset$ holds, on the other hand, then the first condition in line~\ref{algoline:validitytest1} is satisfied wherefore the second condition in line~\ref{algoline:validitytest1} must be violated. That is, there is no conflict set w.r.t.\ $\langle\mb,\emptyset,\Tp,\Tn\rangle_\RQ$. As there is a non-empty minimal conflict set w.r.t.\ $\langle\mo\cup\mc,\mb\setminus\mc,\Tp,\Tn\rangle_\RQ$ by the \emph{Induction Assumption}, $\mc \subseteq \mb$ by Lemma~\ref{lem:qx'_termination_etc}, prop.~\ref{lem_enum:c_sub_b} and $|\mo|\geq 2$ by the fact that there was no return in line~\ref{algoline:return_O}, there must be a non-empty minimal conflict set w.r.t.\ $\langle\mo,\mb,\Tp,\Tn\rangle_\RQ$. Again, an equal argumentation to the one of the \emph{Induction Base} can be applied to derive that there is a non-empty minimal conflict set w.r.t.\ $\langle\mo'\cup\mc',\mb'\setminus\mc',\Tp,\Tn\rangle_\RQ$.
%
%

Case (2): Here, we assume that the recursive call $\scQX'(\mc',\langle\mo',\mb',\Tp,\Tn\rangle_\RQ)$ is made in line~\ref{algoline:recursive_call2} and the result of $\scQX'$ in line~\ref{algoline:recursive_call1} is $\mc_2 = \emptyset$. So, it holds that $\mc' = \mc_2 = \emptyset$, $\mo' = \mo_1$ and $\mb' = \mb$, i.e.\ the recursive call can be written as $\scQX'(\emptyset,\langle\mo_1,\mb,\Tp,\Tn\rangle_\RQ)$. By the fact that $\scQX'(\mo_1,\langle\mo_2,\mb\cup\mo_1,\Tp,\Tn\rangle_\RQ)$ called in line~\ref{algoline:recursive_call1} returned $\emptyset$, both conditions in line~\ref{algoline:validitytest1} during $\scQX'(\mo_1,\langle\mo_2,\mb\cup\mo_1,\Tp,\Tn\rangle_\RQ)$ must have been met. Thus, in particular the existence of a non-empty minimal conflict set w.r.t.\ $\langle\mb\cup\mo_1,\emptyset,\Tp,\Tn\rangle_\RQ$ must be given. Further on, by the \emph{Induction Assumption} there is a non-empty minimal conflict set w.r.t.\ $\langle\mc\cup\mo,\mb\setminus\mc,\Tp,\Tn\rangle_\RQ$.

Let us first assume $\mc = \emptyset$. In this case $\langle\mc\cup\mo,\mb\setminus\mc,\Tp,\Tn\rangle_\RQ$ can be written as $\langle\mo,\mb,\Tp,\Tn\rangle_\RQ$ and it holds that there is a non-empty minimal conflict set w.r.t.\ $\langle\mo,\mb,\Tp,\Tn\rangle_\RQ$, i.e.\ $\mo$ is invalid w.r.t.\ $\langle\cdot,\mb,\Tp,\Tn\rangle_\RQ$. By Proposition~\ref{prop:cs_admissible}, this implies that $\langle\mo,\mb,\Tp,\Tn\rangle_\RQ$ is admissible. In other words, there is no conflict set w.r.t.\ $\langle\mb,\emptyset,\Tp,\Tn\rangle_\RQ$. Consequently, there must be a non-empty minimal conflict set w.r.t.\ $\langle\mo_1,\mb,\Tp,\Tn\rangle_\RQ$.

If $\mc \neq \emptyset$, on the other hand, then the second condition in line~\ref{algoline:validitytest1} during $\scQX'(\mc,\langle\mo,\mb,\Tp,\Tn\rangle_\RQ)$ must be invalid, i.e.\ there is no conflict set w.r.t.\ $\langle\mb,\emptyset,\Tp,\Tn\rangle_\RQ$. Consequently, there must be a non-empty minimal conflict set w.r.t.\ $\langle\mo_1,\mb,\Tp,\Tn\rangle_\RQ$.

Case (3): Here, we assume that the recursive call $\scQX'(\mc',\langle\mo',\mb',\Tp,\Tn\rangle_\RQ)$ is made in line~\ref{algoline:recursive_call2} and the result of $\scQX'$ in line~\ref{algoline:recursive_call1} is $\mc_2 \neq \emptyset$. As $\mc_2 \neq \emptyset$ and line~\ref{algoline:return_upwards} has never been reached by assumption, $\mc_2$ must have been returned in line~\ref{algoline:return_O} of $\scQX'(\mo_1,\langle\mo_2,\mb\cup\mo_1,\Tp,\Tn\rangle_\RQ)$ (which was called in line~\ref{algoline:recursive_call1}) wherefore $\mc_2 = \mo_2$ must hold.
So, it holds that $\mc' = \mo_2$, $\mo' = \mo_1$ and $\mb' = \mb \cup \mo_2$, i.e.\ the recursive call can be written as $\scQX'(\mo_2,\langle\mo_1,\mb\cup\mo_2,\Tp,\Tn\rangle_\RQ)$. 
%
 %
By the \emph{Induction Assumption}, there is a non-empty minimal conflict set w.r.t.\ $\langle\mc\cup\mo,\mb\setminus\mc,\Tp,\Tn\rangle_\RQ$. Moreover, $\mc\subseteq\mb$ by Lemma~\ref{lem:qx'_termination_etc}, prop.~\ref{lem_enum:c_sub_b} and (*) there is a non-empty minimal conflict set w.r.t.\ the DPI $\langle\mo,\mb,\Tp,\Tn\rangle_\RQ$ which is equal to the DPI $\langle\mo_1\cup\mo_2,\mb,\Tp,\Tn\rangle_\RQ$ by the fact that $\mo_1,\mo_2$ partition $\mo$ as per the definition of the \textsc{split} and \textsc{get} functions in lines~\ref{algoline:split}-\ref{algoline:get2}. 

What must still be proven, is (*):
Let us first assume that $\mc = \emptyset$ holds. In this case, $\langle\mc\cup\mo,\mb\setminus\mc,\Tp,\Tn\rangle_\RQ = \langle\mo,\mb,\Tp,\Tn\rangle_\RQ$ and thus there is a non-empty minimal conflict set w.r.t.\ $\langle\mo,\mb,\Tp,\Tn\rangle_\RQ$.

If $\mc \neq \emptyset$, on the other hand, then the second condition in line~\ref{algoline:validitytest1} during $\scQX'(\mc,\langle\mo,\mb,\Tp,\Tn\rangle_\RQ)$ must be invalid as otherwise $\emptyset$ would have been returned which is a contradiction to the assumption that the recursive call $\scQX'(\mc',\langle\mo',\mb',\Tp,\Tn\rangle_\RQ)$ was invoked in line~\ref{algoline:recursive_call2}. So, there is no conflict set w.r.t.\ $\langle\mb,\emptyset,\Tp,\Tn\rangle_\RQ$. Consequently, there must be a non-empty minimal conflict set w.r.t.\ $\langle\mo,\mb,\Tp,\Tn\rangle_\RQ$ due to $\mc\subseteq\mb$ by Lemma~\ref{lem:qx'_termination_etc}, prop.~\ref{lem_enum:c_sub_b}.

3): Case $S\neq\emptyset$: By $S \neq \emptyset$ and the fact that line~\ref{algoline:return_upwards} has not yet been reached, we obtain by the first proposition of this lemma that $|S|=1$ must hold.

There are two cases that can trigger $\scQX'(\mc,\langle\mo,\mb,\Tp,\Tn\rangle_\RQ)$ to return $\mo$ with $|\mo|=1$, i.e.\ case~1 involving $\mc \neq \emptyset$ and case~2 involving $\mc =\emptyset$. 

In case~1, $\mb$ must be valid w.r.t.\ $\langle\cdot,\emptyset,\Tp,\Tn,\rangle_\RQ$ as otherwise $\emptyset$ would be returned in line~\ref{algoline:return_emptyset}. So, there is no (minimal) conflict set w.r.t.\ $\langle\mb,\emptyset,\Tp,\Tn\rangle_\RQ$. 

As $|\mo|=1$ by assumption and by the fact that $\mc \subseteq\mb$ (holds by Lemma~\ref{lem:qx'_termination_etc}, prop.~\ref{lem_enum:c_sub_b}) and there is some non-empty minimal conflict set w.r.t.\ $\langle\mo\cup\mc,\mb\setminus\mc,\Tp,\Tn\rangle_\RQ$ (holds by the second proposition of this lemma), $\mo$ must include a non-empty minimal conflict set w.r.t.\ $\langle\mo,\mb,\Tp,\Tn\rangle_\RQ$. Since the only proper subset of $\mo$ is the empty set, $\mo$ must be a minimal conflict set w.r.t.\ $\langle\mo,\mb,\Tp,\Tn\rangle_\RQ$.

Case~2 can arise only when $\scQX'(\mc,\langle\mo,\mb,\Tp,\Tn\rangle_\RQ)$ is called in line~\ref{algoline:call_QX'} or line \ref{algoline:recursive_call2}. In line~\ref{algoline:recursive_call1} $\scQX'$ is called with $\mc\neq\emptyset$ by Lemma~\ref{lem:qx'_termination_etc}, prop.~\ref{lem_enum:c_neq_emptyset}.

In line~\ref{algoline:call_QX'} $\scQX'$ is called with $\mc=\emptyset$ and, by Corollary~\ref{cor:call_QX'_admissible}, with an admissible DPI $\langle\mo,\mb,\Tp,\Tn\rangle_\RQ$ for which a non-empty minimal conflict set exists as arguments. By the second proposition of this lemma, there is some non-empty minimal conflict set w.r.t.\ $\langle\mo\cup\emptyset,\mb\setminus\emptyset,\Tp,\Tn\rangle_\RQ = \langle\mo,\mb,\Tp,\Tn\rangle_\RQ$, and, by admissibility of $\langle\mo,\mb,\Tp,\Tn\rangle_\RQ$, there is no (minimal) conflict set w.r.t.\ $\langle\mb,\emptyset,\Tp,\Tn\rangle_\RQ$. By $|\mo|=1$, $\mo$ must be a minimal conflict set w.r.t.\ $\langle\mo,\mb,\Tp,\Tn\rangle_\RQ$.

A necessary condition for $\scQX'$ to be called with $\mc =\emptyset$ in line~\ref{algoline:recursive_call2} is obviously that $\scQX'(\mo_1,\langle\mo_2,\mb\cup\mo_1,\Tp,\Tn\rangle_\RQ)$ called in line~\ref{algoline:recursive_call1} returns $\emptyset$. By the Lemma~\ref{lem:qx'_termination_etc}, prop.~\ref{lem_enum:return_emptyset}, there is some non-empty minimal conflict set w.r.t.\ $\langle\mo_1,\mb,\Tp,\Tn\rangle_\RQ$. In line~\ref{algoline:recursive_call2}, the call $\scQX'(\emptyset,\langle\mo_1,\mb,\Tp,\Tn\rangle_\RQ)$ is made which, by assumption, returns $\mo_1$ with $|\mo_1|=1$. That means
$\mo_1$ is a minimal conflict set w.r.t.\ $\langle\mo_1,\mb,\Tp,\Tn\rangle_\RQ$.

Case $S=\emptyset$: Here, both conditions in line~\ref{algoline:validitytest1} must be met, i.e.\ in particular $\mb$ is invalid w.r.t.\ $\langle\cdot,\emptyset,\Tp,\Tn\rangle_\RQ$ which implies that $\mo$ is invalid w.r.t.\ $\langle\cdot,\mb,\Tp,\Tn\rangle_\RQ$ and $\langle\mo,\mb,\Tp,\Tn\rangle_\RQ$ is admissible. Therefore, by Proposition~\ref{prop:cs_admissible}, there is no non-empty minimal conflict set w.r.t.\ $\langle\mo,\mb,\Tp,\Tn\rangle_\RQ$. However, since $\mo$ is invalid w.r.t.\ $\langle\cdot,\mb,\Tp,\Tn\rangle_\RQ$, there must be a conflict set w.r.t.\ $\langle\mo,\mb,\Tp,\Tn\rangle_\RQ$. So, there is only the empty minimal conflict set w.r.t.\ $\langle\mo,\mb,\Tp,\Tn\rangle_\RQ$.

4): This proposition is an immediate consequence of the first proposition of this lemma.
\end{proof}

\begin{lemma}\label{lem:non_adm_return_emptyset}
Let $\langle\mo,\mb,\Tp,\Tn\rangle_\RQ$ be a non-admissible DPI. Then, $\emptyset$ is the only minimal conflict set w.r.t.\ $\langle\mo,\mb,\Tp,\Tn\rangle_\RQ$ and $\scQX'(\mc,\langle\mo,\mb,\Tp,\Tn\rangle_\RQ)$ with $\mc \neq \emptyset$ returns $\emptyset$ immediately in line~\ref{algoline:return_emptyset}.
%
\end{lemma}
\begin{proof}
Since $\langle\mo,\mb,\Tp,\Tn\rangle_\RQ$ is non-admissible, $\mb\cup U_\Tp$ violates some $r\in\RQ$ or $\mb\cup U_\Tp \models \tn$ for some $\tn\in\Tn$. Therefore, $\emptyset$ is invalid w.r.t.\ $\langle\cdot,\mb,\Tp,\Tn\rangle_\RQ$, which, by Corollary~\ref{cor:validonto_cs}, implies that $\emptyset$ is a (minimal) conflict set w.r.t.\ $\langle\mo,\mb,\Tp,\Tn\rangle_\RQ$.

$\scQX'(\mc,\langle\mo,\mb,\Tp,\Tn\rangle_\RQ)$ returns $\emptyset$ in line~\ref{algoline:return_emptyset} as both conditions in line~\ref{algoline:validitytest2} are satisfied due to $\mc \neq \emptyset$ and the non-admissibility of $\langle\mo,\mb,\Tp,\Tn\rangle_\RQ$.
\end{proof}

\begin{lemma}\label{lem:adm_not_return_emptyset}
Let $\langle\mo,\mb,\Tp,\Tn\rangle_\RQ$ be an admissible DPI. Then $\scQX'(\mc,\langle\mo,\mb,\Tp,\Tn\rangle_\RQ)$ does not return in line~\ref{algoline:return_emptyset}.
\end{lemma}
\begin{proof}
By Definition~\ref{def:admissible}, $\mb$ must be valid w.r.t.\ $\langle\cdot,\emptyset,\Tp,\Tn\rangle_\RQ$. Hence, the second condition in line~\ref{algoline:validitytest2} is not satisfied wherefore a return cannot take place in line~\ref{algoline:return_emptyset}.
\end{proof}
\begin{lemma}\label{lem:qx_result_non-empty}
Let $\langle\mo,\mb,\Tp,\Tn\rangle_\RQ$ be an admissible DPI and let there be a non-empty minimal conflict set w.r.t.\ $\langle\mo,\mb,\Tp,\Tn\rangle_\RQ$. Then the following holds:
When $\scQX'(\mc,\langle\mo,\mb,\Tp,\Tn\rangle_\RQ)$
reaches line~\ref{algoline:return_upwards} for the first time, $\mc_1\cup\mc_2$ is a non-empty minimal conflict set w.r.t.\ $\langle\mo,\mb,\Tp,\Tn\rangle_\RQ$.
\end{lemma}
\begin{proof}
The premises of this lemma are the same as those of Lemma~\ref{lem:qx_downwards}.
By Lemma~\ref{lem:qx_downwards}, prop.~\ref{lem_enum:downwards_4} we know that for $\mc_2$ and $\mc_1$ that are returned by the the calls to $\scQX'$ in lines~\ref{algoline:recursive_call1} and \ref{algoline:recursive_call2} $|\mc_1|\leq 1$ and $|\mc_2|\leq 1$ holds. Moreover, we know by Lemma~\ref{lem:qx_recursion_principle} that $\mc_1 \cup \mc_2$ is a minimal conflict set w.r.t.\ $\langle\mo,\mb,\Tp,\Tn\rangle_\RQ$.

What remains open is to show that $\mc_1 \cup \mc_2 \neq \emptyset$. To this end, we first assume that $\mc \neq \emptyset$. Then, by Lemma~\ref{lem:non_adm_return_emptyset}, $\langle\mo,\mb,\Tp,\Tn\rangle_\RQ$ must be an admissible DPI since it does not return in line~\ref{algoline:return_emptyset}, but only in line~\ref{algoline:return_upwards}.

If, on the other hand, $\mc = \emptyset$ holds, we can apply Lemma~\ref{lem:qx_downwards}, prop.~\ref{lem_enum:downwards_2} to obtain that there is a non-empty minimal conflict set w.r.t.\ $\langle\mo,\mb,\Tp,\Tn\rangle_\RQ$. This implies that $\mo$ is invalid w.r.t.\ $\langle\cdot,\mb,\Tp,\Tn\rangle_\RQ$. Therefore, we can conclude by means of Proposition~\ref{prop:cs_admissible} that $\langle\mo,\mb,\Tp,\Tn\rangle_\RQ$ is an admissible DPI.

Thus, in both cases we have that $\langle\mo,\mb,\Tp,\Tn\rangle_\RQ$ is an admissible DPI. Applying Lemma~\ref{lem:one_rec_call_adm} yields that at least one recursive call to $\scQX'$ in lines~\ref{algoline:recursive_call1} and \ref{algoline:recursive_call2} is given an admissible DPI as argument. By Lemma~\ref{lem:adm_not_return_emptyset}, this call cannot return in line~\ref{algoline:return_emptyset}. So, it must return in line~\ref{algoline:return_O} by the assumption that line~\ref{algoline:return_upwards} has not yet been reached before, wherefore it must return a set of cardinality 1. This completes the proof.
\end{proof}
As long as the algorithm goes upwards after going upwards for the first time, a non-empty minimal conflict set is propagated upwards.
\begin{lemma}[QX: Upwards Correctness]\label{lem:qx_upwards}
Let $\langle\mo,\mb,\Tp,\Tn\rangle_\RQ$ be an admissible DPI and let there be a non-empty minimal conflict set w.r.t.\ $\langle\mo,\mb,\Tp,\Tn\rangle_\RQ$. Then:
%
After $\scQX'(\mc,\langle\mo,\mb,\Tp,\Tn\rangle_\RQ)$ has reached line~\ref{algoline:return_upwards} for the first time, the following holds: As long as line~\ref{algoline:recursive_call1} is not reached, each return in line~\ref{algoline:return_upwards} returns a minimal conflict set w.r.t.\ $\langle\mo,\mb,\Tp,\Tn\rangle_\RQ$.
\end{lemma}
\begin{proof}
The premises of this lemma are the same as those of Lemma~\ref{lem:qx_downwards}.
By Lemma~\ref{lem:qx_result_non-empty} we know that a non-empty minimal conflict $\mc$ set is returned at the first return that is made in line~\ref{algoline:return_upwards}. As, by assumption, $\mc$ is not the result $\mc_2$ of a prior call to $\scQX'$ in line~\ref{algoline:recursive_call1}, it must be the result $\mc_1$ of a prior call to $\scQX'$ in line~\ref{algoline:recursive_call2}. 
%
%
Since the premises of Lemma~\ref{lem:qx_downwards} are fulfilled, Lemma~\ref{lem:qx_downwards} can be applied.
Since the call $\scQX'(\mo_1,\langle\mo_2,\mb\cup\mo_1,\Tp,\Tn\rangle)$ (that returned $\mc_2$) in line~\ref{algoline:recursive_call1} took place before line~\ref{algoline:return_upwards} was first reached, we have that $\mc_2$ is a minimal conflict set w.r.t.\ $\langle\mo_2,\mb\cup\mo_1,\Tp,\Tn\rangle$ by Lemma~\ref{lem:qx_downwards}, prop.~\ref{lem_enum:downwards_3}. By Lemma~\ref{lem:qx_recursion_principle}, we have that $\mc_2 \cup \mc$ is a minimal conflict set w.r.t.\ $\langle\mo,\mb,\Tp,\Tn\rangle$. As long as line~\ref{algoline:recursive_call1} is not reached, the same argumentation can be used to show that a minimal conflict set is returned in line~\ref{algoline:return_upwards}.
\end{proof}
When the algorithm goes downwards again after going upwards for the first time, the invariant that that a minimal conflict set exists for each recursive downwards call to $\scQX'$ holds.
\begin{lemma}[QX: Downwards-after-upwards Correctness]\label{lem:qx_downwards_after_upwards}
Let $\langle\mo,\mb,\Tp,\Tn\rangle_\RQ$ be an admissible DPI and let there be a non-empty minimal conflict set w.r.t.\ $\langle\mo,\mb,\Tp,\Tn\rangle_\RQ$. Then:
After $\scQX'(\mc,\langle\mo,\mb,\Tp$, $\Tn\rangle_\RQ)$ has reached line~\ref{algoline:return_upwards} for the first time, the following holds: If line~\ref{algoline:recursive_call1} is reached for the first time, then, if the DPI $\langle\mo_1,\mb\cup\mc_2,\Tp,\Tn\rangle_\RQ$ which is the argument to the immediate call $\scQX'(\mc_2,\langle\mo_1,\mb\cup\mc_2,\Tp,\Tn\rangle_\RQ)$ 
in line~\ref{algoline:recursive_call2} is admissible, then there is a non-empty minimal conflict set w.r.t.\ $\langle\mo_1,\mb\cup\mc_2,\Tp,\Tn\rangle_\RQ$.
\end{lemma}
\begin{proof}
The premises of this lemma are the same as those of Lemma~\ref{lem:qx_downwards}.
Since line~\ref{algoline:recursive_call1} is first reached after line~\ref{algoline:return_upwards} has been reached for the first time, it must hold that $\scQX'(\mo_1,\langle\mo_2,\mb\cup\mo_1,\Tp,\Tn\rangle_\RQ)$ in line~\ref{algoline:recursive_call1} was called before line~\ref{algoline:return_upwards} has been reached. The reason for this to hold is the fact that only returns and no new calls to $\scQX'$ can have been made between the first occurrence of line~\ref{algoline:return_upwards} and the next occurrence of line~\ref{algoline:recursive_call1}.  

Therefore, the result $\mc_2$ of the call $\scQX'(\mo_1,\langle\mo_2,\mb\cup\mo_1,\Tp,\Tn\rangle_\RQ)$ in line~\ref{algoline:recursive_call1} is a minimal conflict set w.r.t.\ $\langle\mo_2,\mb\cup\mo_1,\Tp,\Tn\rangle_\RQ$ due to Lemma~\ref{lem:qx_downwards}, prop.~\ref{lem_enum:downwards_3}. As a consequence, $\mc_2 \cup \mb \cup \mo_1 \cup U_\Tp$ violates some $r\in\RQ$ or some $\Tn\in\Tn$. As the DPI $\langle\mo_1,\mb\cup\mc_2,\Tp,\Tn\rangle_\RQ$ is admissible by assumption, it holds that $\mc_2 \cup \mb \cup U_\Tp$ does not violate any $r\in\RQ$ or $\Tn\in\Tn$. Hence, $\mo_1$ must be invalid w.r.t.\ $\langle\cdot,\mb\cup\mc_2,\Tp,\Tn\rangle_\RQ$ which implies that there must be a non-empty minimal conflict set $S$ w.r.t.\ $\langle\mo_1,\mb\cup\mc_2,\Tp,\Tn\rangle_\RQ$.
\end{proof}
By applying the argumentation of Lemmas~\ref{lem:qx_downwards}, \ref{lem:qx_upwards} and \ref{lem:qx_downwards_after_upwards} 
recursively on the entire recursion tree, we can prove the correctness of $\scQX'$.
\begin{lemma}\label{lem:qx'_correctness}
If $\scQX'(\mc,\langle\mo_{\mathsf{orig}},\mb_{\mathsf{orig}},\Tp,\Tn\rangle_\RQ)$ is called in line~\ref{algoline:call_QX'} by Algorithm~\ref{algo:qx}, it returns a non-empty minimal conflict set w.r.t.\ $\langle\mo_{\mathsf{orig}},\mb_{\mathsf{orig}},\Tp,\Tn\rangle_\RQ$.
\end{lemma}
\begin{proof}
If $\scQX'(\mc,\langle\mo_{\mathsf{orig}},\mb_{\mathsf{orig}},\Tp,\Tn\rangle_\RQ)$ is called in line~\ref{algoline:call_QX'} of Algorithm~\ref{algo:qx}, it must be true, by Lemma~\ref{lem:qx_start_conditions}, prop.~\ref{lem_enum:qx'} and Corollary~\ref{cor:call_QX'_admissible}, that $\langle\mo_{\mathsf{orig}},\mb_{\mathsf{orig}},\Tp,\Tn\rangle_\RQ$ is an admissible DPI for which a non-empty minimal conflict set exists.
As a consequence, the premises of Lemma~\ref{lem:qx_downwards} are met for $\langle\mo_{\mathsf{orig}},\mb_{\mathsf{orig}},\Tp,\Tn\rangle_\RQ$.

There are two cases to consider: Either (a)~$|\mo_{\mathsf{orig}}| \leq 1$ or (b)~$|\mo_{\mathsf{orig}}| > 1$ for the initial call to $\scQX'(\mc,\langle\mo_{\mathsf{orig}},\mb_{\mathsf{orig}},\Tp,\Tn\rangle_\RQ)$ in line~\ref{algoline:call_QX'}. In case (a), $0=|\mo_{\mathsf{orig}}| < 1$ cannot hold as there must be a non-empty minimal conflict set $\mc$ w.r.t.\ $\langle\mo_{\mathsf{orig}},\mb_{\mathsf{orig}},\Tp,\Tn\rangle_\RQ$ due to Lemma~\ref{lem:qx_start_conditions}, prop.~\ref{lem_enum:qx'}. Since $\emptyset \subset \mc \subseteq \mo_{\mathsf{orig}}$ must hold for $\mc$, this would be a contradiction to $|\mo_{\mathsf{orig}}|=0$.

So, $|\mo_{\mathsf{orig}}| = 1$ holds in case (a). In this case, $\scQX'$ returns $\mo_{\mathsf{orig}}$ immediately in line~\ref{algoline:return_O}, since $\mc=\emptyset$ and thus the conditions checked in line~\ref{algoline:validitytest2} cannot be met. In this case, $\mo_{\mathsf{orig}}$ is indeed a non-empty minimal conflict set since
for the DPI $\langle\mo_{\mathsf{orig}},\mb_{\mathsf{orig}},\Tp,\Tn\rangle_\RQ$ given as argument 
there is a non-empty minimal conflict set by Lemma~\ref{lem:qx_start_conditions}, prop.~\ref{lem_enum:qx'}. Therefore $\emptyset$ cannot be a conflict set w.r.t.\ this DPI whereby $\mo_{\mathsf{orig}}$ is the only possible minimal conflict set due to $|\mo_{\mathsf{orig}}| = 1$.

Case (b): In this case, a direct return can neither take place in line~\ref{algoline:return_emptyset} by $\mc = \emptyset$ nor in line~\ref{algoline:return_O} by $|\mo_{\mathsf{orig}}| > 1$. So, $\scQX'$ is called recursively in lines~\ref{algoline:recursive_call1} and \ref{algoline:recursive_call2}. Since $\scQX'$ terminates due to Lemma~\ref{lem:qx_start_conditions}, prop.~\ref{lem_enum:terminates}, $\scQX'$ must reach line~\ref{algoline:return_upwards}. The first time some recursive call $\scQX'(\mc,\langle\mo,\mb,\Tp,\Tn\rangle_\RQ)$ reaches line~\ref{algoline:return_upwards}, it returns a non-empty minimal conflict set w.r.t.\ $\langle\mo,\mb,\Tp,\Tn\rangle_\RQ$ due to Lemma~\ref{lem:qx_result_non-empty}.

By Lemma~\ref{lem:qx_upwards}, as long as line~\ref{algoline:recursive_call1} is not reached, i.e.\ no ``left branch'' (call to $\scQX'$ in line~\ref{algoline:recursive_call1}) but only ``right branches'' (calls to $\scQX'$ in line~\ref{algoline:recursive_call2}) return, a minimal conflict set $S$ is returned for each call to $\scQX'$ that ``wraps'' (is higher in the recursion tree than) the call that was the first to reach line~\ref{algoline:return_upwards}. It holds that $S \neq \emptyset$ since $S$ is a union of sets including the non-empty set returned when line~\ref{algoline:return_upwards} was first reached.

When it comes to an execution of line~\ref{algoline:recursive_call1}, i.e.\ the left branch returns, then the algorithm will take the right branch by executing line~\ref{algoline:recursive_call2}, i.e.\ calling $\scQX'(\mc_2,\langle\mo_1,\mb\cup\mc_2,\Tp,\Tn\rangle_\RQ)$, and go downwards in the recursion tree. 

Now, there are two cases. First, $\langle\mo_1,\mb\cup\mc_2,\Tp,\Tn\rangle_\RQ$ is non-admissible. Then, by Lemma~\ref{lem:non_adm_return_emptyset}, there is only one minimal conflict set w.r.t.\ $\langle\mo_1,\mb\cup\mc_2,\Tp,\Tn\rangle_\RQ$, namely $\emptyset$, and $\scQX'(\mc_2,\langle\mo_1,\mb\cup\mc_2,\Tp,\Tn\rangle_\RQ)$ directly returns $\emptyset$. As also the result $\mc_2$ of the call to $\scQX'(\mo_1,\langle\mo_2,\mb\cup\mo_1,\Tp,\Tn\rangle_\RQ)$ immediately before in line~\ref{algoline:recursive_call1} is a minimal conflict set w.r.t.\ $\langle\mo_2,\mb\cup\mo_1,\Tp,\Tn\rangle_\RQ$, as established above, we can apply Lemma~\ref{lem:qx_recursion_principle} to derive that indeed a minimal conflict set w.r.t.\ $\langle\mo,\mb,\Tp,\Tn\rangle_\RQ$ is returned in line~\ref{algoline:return_upwards}. Thus, Lemma~\ref{lem:qx_upwards} can be further applied to move upwards in the recursion tree until line~\ref{algoline:recursive_call1} occurs again.

Second, $\langle\mo_1,\mb\cup\mc_2,\Tp,\Tn\rangle_\RQ$ is admissible. Then, by Lemma~\ref{lem:qx_downwards_after_upwards}, there is a non-empty minimal conflict set w.r.t.\ $\langle\mo_1,\mb\cup\mc_2,\Tp,\Tn\rangle_\RQ$. Hence, Lemma~\ref{lem:qx_downwards} can be used again for the subtree of the recursion tree rooted at the call $\scQX'(\mc_2,\langle\mo_1,\mb\cup\mc_2,\Tp,\Tn\rangle_\RQ)$. That is, it can be used to show that each call to $\scQX'$ within this subtree returns a minimal conflict set w.r.t.\ the DPI given as argument as long as the algorithm moves downwards in the tree. Having reached line~\ref{algoline:return_upwards} for the first time, Lemma~\ref{lem:qx_result_non-empty} lets us conclude again that a non-empty conflict set w.r.t.\ the respective argument DPI is actually returned at this place. Subsequently, Lemma~\ref{lem:qx_upwards} 
can be applied to show that each return gives back a minimal conflict set w.r.t.\ the argument DPI of the respective call, as long as the algorithm moves upwards in the recursion tree.

What is still open is to show that the call $\scQX'(\mc_2,\langle\mo_1,\mb\cup\mc_2,\Tp,\Tn\rangle_\RQ)$ in line~\ref{algoline:recursive_call2} that is made immediately after the algorithm first reached line~\ref{algoline:recursive_call1} after moving upwards after reaching line~\ref{algoline:return_upwards} for the first time returns a minimal conflict set w.r.t.\ $\langle\mo_1,\mb\cup\mc_2,\Tp,\Tn\rangle_\RQ$, indeed. This holds by the fact that Lemmas~\ref{lem:qx_downwards} and ~\ref{lem:qx_upwards} guarantee that a left branch always returns a minimal conflict set, Lemma~\ref{lem:qx_downwards_after_upwards} guarantees that Lemmas~\ref{lem:qx_downwards} and ~\ref{lem:qx_upwards} can be applied after making a single right branch. However, as $\scQX'$ terminates the recursion tree is finite and thus the case must arise where the right branch directly returns. In case the DPI $\langle\mo,\mb,\Tp,\Tn\rangle_\RQ$ given as argument for this right branch is non-admissible, the only minimal conflict set $\emptyset$ is returned, as established above. If the DPI $\langle\mo,\mb,\Tp,\Tn\rangle_\RQ$ given as argument for this right branch is admissible, on the other hand, then we have already shown above that there is a non-empty minimal conflict set w.r.t.\ this DPI. Moreover, $|\mo|=1$ must hold due to the fact that this right branch directly returns (without entering a further recursion). Therefore, $\mo$ is returned which is actually a minimal conflict set w.r.t.\ $\langle\mo,\mb,\Tp,\Tn\rangle_\RQ$ as $\mo$ is the only non-empty subset of $\mo$.  
\end{proof}

\begin{proposition}\label{prop:qx_correctness}
Let $\langle\mo,\mb,\Tp,\Tn\rangle_\RQ$ be a DPI. Then, $\scQX(\langle\mo,\mb,\Tp,\Tn\rangle_\RQ)$ terminates and returns 
\begin{itemize}
	\item 'no conflict' iff there is no conflict w.r.t.\ $\langle\mo,\mb,\Tp,\Tn\rangle_\RQ$ $\quad$ \\ ($\mo$ is valid w.r.t.\ $\langle\cdot,\mb,\Tp,\Tn\rangle_\RQ$)
	\item $\emptyset$ iff $\emptyset$ is the only minimal conflict set w.r.t.\ $\langle\mo,\mb,\Tp,\Tn\rangle_\RQ$ $\quad$ \\(DPI is non-admissible)
	\item a non-empty minimal conflict set w.r.t.\ $\langle\mo,\mb,\Tp,\Tn\rangle_\RQ$ iff there is a non-empty minimal conflict set w.r.t.\ $\langle\mo,\mb,\Tp,\Tn\rangle_\RQ$ $\quad$ \\(DPI is admissible and $\mo$ is invalid w.r.t.\ $\langle\cdot,\mb,\Tp,\Tn\rangle_\RQ$).
\end{itemize}
\end{proposition}
\begin{proof}
The proposition is a direct consequence of Lemma~\ref{lem:qx_start_conditions} and Lemma~\ref{lem:qx'_correctness}.
\end{proof}


\section{Hitting Set Tree Based Diagnosis Computation}
\label{sec:hs_comp}
One way to compute minimal diagnoses from minimal conflict sets is to use a hitting set tree algorithm which was originally proposed by Reiter~\cite{Reiter87}. In this work we describe methods for non-interactive and interactive diagnosis computation based on the ones used in~\cite{friedrich2005gdm, ksgf2010, Shchekotykhin2012} which are closely related to the original hitting set tree algorithm. Differences of the described non-interactive algorithm to the original one of Reiter are 
\begin{enumerate} 
\item the usage of different edge weights (probabilities) inducing an order of node generation (uniform-cost) different to breadth-first and 
\item the opportunity to specify an execution time threshold $t$ as well as a minimal ($n_{\min}$) and maximal ($n_{\max}$) desired number of minimal diagnoses to be computed by the algorithm. 
\end{enumerate}
In this vein, the algorithm computes at least the $n_{\min}$ most-probable minimal diagnoses w.r.t.\ the given probabilities and goes on computing further next most-probable minimal diagnoses until either overall computation time reaches the time limit $t$ or $n_{\max}$ diagnoses have been computed.
 
Such a time threshold and an interval of minimal and maximal number of diagnoses is particularly relevant in settings where not all potential minimal faulty sets need to be computed, such as iterative, interactive settings where reaction time is crucial (since a user is waiting to interact with the system). Instead, in such settings only a ``representative'' set of minimal diagnoses is exploited to decide which question to ask a user such that the answer to that question allows the constructed partial tree to be pruned. After pruning, the tree is expanded again to compute another ``representative'' set of minimal diagnoses.

\paragraph{Inputs.} The non-interactive version of the algorithm is delineated by Algorithm~\ref{algo:hs}. The algorithm takes as input an admissible DPI $\tuple{\mo,\mb,\Tp,\Tn}_\RQ$, some computation timeout $t$, a
desired minimal ($n_{\min}$) and maximal ($n_{\max}$) number of minimal diagnoses to be returned, and a function $p:\mo \rightarrow (0,0.5)$ that assigns to each formula $\tax \in \mo$ a weight that represents the (estimated) likeliness of $\tax$ to be faulty and thereby determines the search strategy, e.g.\ breadth-first or uniform-cost. 
Within the algorithm, $p()$ is used to impose an order on open nodes that tells the algorithm which node to expand next. Details concerning the function $p()$ will be discussed in Section~\ref{sec:DiagnosisProbabilitySpace}. Until the end of the currect section (Section~\ref{sec:hs_comp}) we assume that $p()$ implies a first-in-first-out sorting of open nodes, i.e.\ a breadth-first search strategy as described in~\cite{Reiter87}.


\paragraph{Algorithm Overview and Implementation Remarks.} To compute minimal diagnoses w.r.t.\ $\langle\mo,\mb,\Tp$, $\Tn\rangle_\RQ$ from minimal conflict sets w.r.t.\ $\langle\mo,\mb,\Tp,\Tn\rangle_\RQ$, the algorithm produces a labeled tree where a non-closed node is labeled by a minimal conflict set and a closed node is labeled by either $valid$ or $closed$. From a non-closed node labeled by a minimal conflict set $\mc=\setof{\tax_p, \dots, \tax_q}$ there are $|\mc|$ outgoing edges, each labeled by one $\tax\in\mc$ and each leading to a new node that needs to be labeled. Closed nodes are leaf nodes of the produced tree, i.e.\ they have no successor nodes, and correspond to non-minimal or duplicate hitting sets (label $closed$) or to minimal hitting sets (label $valid$) of all minimal conflict sets w.r.t.\ the input DPI $\langle\mo,\mb,\Tp,\Tn\rangle_\RQ$. Conflict sets to label nodes are computed only on-demand for time efficiency after the attempt to reuse an already computed one fails. In case an appropriate order of node labeling (e.g.\ breadth-first tree construction) is used, the complete tree given when all nodes in the tree are closed contains all minimal diagnoses w.r.t.\ the DPI $\langle\mo,\mb,\Tp,\Tn\rangle_\RQ$ provided as input. In this complete tree, the set of edge labels on each path from the root node to a node labeled by $valid$ is a minimal diagnosis. 

What Algorithm~\ref{algo:hs} actually does is building up a \emph{pruned HS-tree} for a given DPI. So, we next provide formal definitions of a \emph{(partial) HS-tree} and a \emph{(partial) pruned HS-tree} based on the definitions given in~\cite{Reiter87}.
\begin{definition}[HS-Tree]\label{def:hs_tree}
Let $\langle\mo,\mb,\Tp,\Tn\rangle_\RQ$ be an admissible DPI. An edge-labeled and node-labeled tree $T$ is called an \emph{HS-tree w.r.t.\ $\langle\mo,\mb,\Tp,\Tn\rangle_\RQ$} iff it is a smallest tree with the following properties: 
\begin{enumerate}
\item The root of $T$ is labeled by $valid$ if $\mo$ is valid w.r.t.\ $\langle\cdot,\mb,\Tp,\Tn\rangle_\RQ$. Otherwise, the root is labeled by a conflict set w.r.t.\ $\langle\mo,\mb,\Tp,\Tn\rangle_\RQ$.
\item If $\mathsf{n}$ is a node of $T$, define $H(\mathsf{n})$ to be the set of edge labels on the path in
$T$ from the root node to $\mathsf{n}$. 
If $\mathsf{n}$ is labeled by $valid$, it has no successor nodes in $T$. If $\mathsf{n}$ is labeled by a conflict set $\mc$ w.r.t.\ $\langle\mo,\mb,\Tp,\Tn\rangle_\RQ$, then for each $\tax \in \mc$, $\mathsf{n}$ has a successor node $\mathsf{n}_{\tax}$ joined to $\mathsf{n}$ by an edge labeled by $\tax$. The label for $\mathsf{n}_{\tax}$ is a conflict set $\mc'$ w.r.t.\ $\langle\mo,\mb,\Tp,\Tn\rangle_\RQ$ such that $\mc' \cap H(\mathsf{n}_{\tax}) = \emptyset$ if such a set $\mc'$ exists. Otherwise, $\mathsf{n}_{\tax}$ is labeled by $valid$.
\end{enumerate}
$T$ is called a \emph{partial HS-tree w.r.t.\ $\langle\mo,\mb,\Tp,\Tn\rangle_\RQ$} iff $T$ is a HS-tree w.r.t.\ $\langle\mo,\mb,\Tp,\Tn\rangle_\RQ$ where not all nodes in $T$ are labeled and non-labeled nodes have no successors. 
\end{definition}
\begin{definition}[Pruned HS-Tree]\label{def:pruned_hs_tree}
Let $\langle\mo,\mb,\Tp,\Tn\rangle_\RQ$ be an admissible DPI. An edge-labeled and node-labeled tree $T$ is called a \emph{pruned HS-tree (pHS-tree) w.r.t.\ $\langle\mo,\mb,\Tp,\Tn\rangle_\RQ$} iff $T$ is the result of constructing an HS-tree w.r.t.\ $\langle\mo,\mb,\Tp,\Tn\rangle_\RQ$ with due regard to the following rules: 
\begin{enumerate}
\item Label nodes in the HS-tree in \emph{breadth-first order}.
\item Use only \emph{minimal} conflict sets w.r.t.\ $\langle\mo,\mb,\Tp,\Tn\rangle_\RQ$ to label nodes in $T$.\label{def:pruned_hs_tree:use_only_min_cs}
\item \emph{Reusing node labels:} If node $\mathsf{n}$ is labeled by $\mc$ and $\mathsf{n}'$ is a node such that $H(\mathsf{n}') \cap \mc = \emptyset$, label $\mathsf{n}'$ by $\mc$. 
\item\label{def:pruned_hs_tree:non_min_pruning_rule} \emph{Non-minimality pruning rule:} If node $\mathsf{n}$ is labeled by $valid$ and node $\mathsf{n}'$ is such that $H(\mathsf{n}) \subseteq H(\mathsf{n}')$, label $\mathsf{n}'$ by $closed$.
\item If node $\mathsf{n}$ is labeled by $closed$, it has no successors.
\item\label{def:pruned_hs_tree:rule6} \emph{Duplicate pruning rule:} If node $\mathsf{n}$ is next to be labeled and there is some node $\mathsf{n}'$ such that $H(\mathsf{n}') = H(\mathsf{n})$, then label
$\mathsf{n}$ by $closed$. 
\end{enumerate}
$T$ is called a \emph{partial pruned HS-tree} iff $T$ is a pruned HS-tree where not all nodes in $T$ have been labeled yet and non-labeled nodes have no successors. 
\end{definition}

\begin{remark}\label{rem:pruned_hs_tree}
Notice that we use a definition of a pruned HS-tree that slightly differs from the definition given in~\cite{Reiter87} in that we inherently assume that only \emph{minimal} conflict sets w.r.t.\ the given DPI are used to label nodes in the tree. Therefore we could omit the last rule in the definition of \cite{Reiter87}. Namely, such a situation where some node has been labeled by a subset of the label of another node cannot arise in our definition since no minimal conflict set can be a subset of another different minimal conflict set w.r.t.\ the same DPI.

In general, there are multiple different pHS-trees w.r.t.\ one and the same DPI~\cite{greiner1989correction}. Reason for this is that 
\begin{itemize}
\item the order of adding successor nodes (on the same tree level) to the queue $\Queue$ and
\item which of generally multiple minimal conflict sets to (re)use to label a node 
\end{itemize} 
is not determined by Definition~\ref{def:pruned_hs_tree}.\qed
\end{remark} 

By~\cite[Theorem~4.8]{Reiter87} and Proposition~\ref{prop:mindiag_mincs}, the following holds:
\begin{proposition}\label{prop:pruned_hs_tree_finds_all_min_diags}
Let $\langle\mo,\mb,\Tp,\Tn\rangle_\RQ$ be an admissible DPI and $T$ a pHS-tree w.r.t.\ $\langle\mo,\mb,\Tp,\Tn\rangle_\RQ$. Then, $\setof{H(\mathsf{n})\,|\, \mathsf{n} \mbox{ is a node of } T \mbox{ labeled by } valid} = \minD_{\langle\mo,\mb,\Tp,\Tn\rangle_\RQ}$, i.e.\ the set of all minimal diagnoses w.r.t.\ $\langle\mo,\mb,\Tp,\Tn\rangle_\RQ$.
\end{proposition}

\begin{remark}\label{rem:algo_hs:internal_representation} A \emph{node} $\mathsf{nd}$ in Algorithm~\ref{algo:hs} is defined as the set of formulas that label the edges on the path from the root node to $\mathsf{nd}$. In other words, we associate a node $\mathsf{n}$ with $H(\mathsf{n})$.
In this vein, Algorithm~\ref{algo:hs} \emph{internally} does not store a labeled tree, but only ``relevant'' sets of nodes 
and conflict sets. That is, it does not store any
\begin{itemize}
	\item non-leaf nodes,
	\item labels of non-leaf nodes, i.e.\ it does not store which minimal conflict set labels which node,
	\item edges between nodes, 
	\item labels of edges and
	\item leaf nodes labeled by $closed$.
\end{itemize}
%
Let $T$ denote the (partial) pHS-tree produced by Algorithm~\ref{algo:hs} at some point during its execution (Corollary~\ref{cor:algo_hs_returns_relevant_data_of_(partial)_pruned_hs_tree} will show that Algorithm~\ref{algo:hs} using breadth-first search	in fact produces a (partial) pHS-tree). Then, Algorithm~\ref{algo:hs} only stores 
\begin{itemize}
	\item a set of nodes $\mD_{calc}$ where each node corresponds to the edge labels along a path in $T$ leading to a leaf node that has been labeled by $valid$ (minimal diagnoses w.r.t.\ $\langle\mo,\mb,\Tp,\Tn\rangle_\RQ$), 
	\item a list of open (non-closed) nodes $\Queue$ where each node in $\Queue$ corresponds to the edge labels along a path in $T$ leading from the root node to a leaf node that has been generated, but has not yet been labeled and
	\item the set $\mC_{calc}$ of already computed minimal conflict sets w.r.t.\ $\langle\mo,\mb,\Tp,\Tn\rangle_\RQ$ that have been used to label non-leaf nodes in $T$.
\end{itemize}
We call $\tuple{\mD_{calc}, \Queue, \mC_{calc}}$ the \emph{relevant data of $T$}. If $T$ is a pHS-tree, then $\Queue$ is the empty list.

This internal representation of the constructed (partial) pHS-tree by its relevant data does not constrain the functionality of the algorithm. 
This holds as diagnoses are paths from the root, i.e.\ nodes in the internal representation, and the goal of a (partial) pHS-tree is to determine minimal diagnoses w.r.t.\ the given DPI. 
The node labels or edge labels along a certain path and their order along this path is completely irrelevant when it comes to finding a label for the leaf node of this path. Instead, only the set of edge labels is required for the computation of the label for a leaf node. Also, to rule out nodes corresponding to non-minimal diagnoses, it is sufficient to know the set of already found diagnoses $\mD_{calc}$. No already closed nodes are needed for the correct functionality of Algorithm~\ref{algo:hs}.\qed
%
\end{remark}

\paragraph{Initialization.} First, Algorithm~\ref{algo:hs} initializes the variable $t_{start}$ with the current system time (\textsc{getTime}), the set of calculated minimal diagnoses $\mD_{calc}$ to the empty set and the ordered queue of open nodes $\Queue$ to a list including the empty set only (i.e.\ only the unlabeled root node). 

\paragraph{The Main Loop.} Within the loop (line~\ref{algoline:hs:repeat}) the algorithm gets the node to be processed next, namely the first node $\mathsf{node}$ (\textsc{getFirst}, line~\ref{algoline:hs:getfirst}) in the list of open nodes $\Queue$ ordered by the function $p_{nodes}()$ and removes $\mathsf{node}$ from $\Queue$ (\textsc{deleteFirst}, line~\ref{algoline:hs:remove_from_queue}). Note that $p_{nodes}()$ can be directly obtained from $p()$. As mentioned before, for the moment the reader should simply suppose that $p_{nodes}()$ imposes an order on $\Queue$ which effectuates a breadth-first labeling of open nodes in the tree. A definition of $p_{nodes}()$ will be given by Definition~\ref{def:p_node()} after a motivation and detailed explanation of $p_{nodes}()$ will have been given in Section~\ref{sec:DiagnosisProbabilitySpace}. 

\paragraph{Computation of Node Labels.} Then, a label is computed for $\mathsf{node}$ in line~\ref{algoline:hs:label}. Nodes are labeled by $valid$, $closed$ or a minimal conflict set w.r.t.\ $\langle\mo,\mb,\Tp,\Tn\rangle_\RQ$ by the procedure \textsc{label} (line~\ref{algoline:hs:procedure_label} ff.). This procedure gets as inputs the DPI $\tuple{\mo,\mb,\Tp,\Tn}_\RQ$, the current node $\mathsf{node}$, the set of already computed minimal conflicts ($\mC_{calc}$) and minimal diagnoses ($\mD_{calc}$) and the queue $\Queue$ of open nodes, and it returns an updated set of computed minimal conflicts $\mC_{calc}$ and a label for $\mathsf{node}$. It works as follows:

A node $\mathsf{node}$ is labeled by $closed$ iff (a)~there is an already computed minimal diagnosis $\md$ in $\mD_{calc}$ that is a subset of this node, i.e.\ $\md \subseteq \mathsf{node}$, which means that $\mathsf{node}$ cannot be a minimal diagnosis (non-minimality criterion, lines~\ref{algoline:hs:non_min_crit_start}-\ref{algoline:hs:non_min_crit_end}) or (b)~there is some node $\mathsf{nd}$ in the queue of open nodes $\Queue$ such that $\mathsf{node} = \mathsf{nd}$ which means that one of the two tree branches with an equal set of edge labels can be closed, i.e.\ removed from $\Queue$ (duplicate criterion, lines~\ref{algoline:hs:duplicate_crit_start}-\ref{algoline:hs:duplicate_crit_end}). 

If none of these $closed$-criteria is met, the algorithm searches for some $\mc$ in $\mC_{calc}$, the set of already computed minimal conflict sets, such that $\mc \cap \mathsf{node} = \emptyset$ and returns the label $\mc$ for $\mathsf{node}$ (reuse criterion, lines~\ref{algoline:hs:reuse_crit_start}-\ref{algoline:hs:reuse_crit_end}). This means that the path represented by $\mathsf{node}$ cannot be a diagnosis as there is (at least) one minimal conflict set, namely $\mc$, that is not hit by $\mathsf{node}$. 

If the reuse criterion does not apply, a call to $\scQX(\tuple{\mo \setminus \mathsf{node}, \mb, \Tp, \Tn}_\RQ)$ is made (line~\ref{algoline:hs:qx}) in order to check whether there is a not-yet-computed minimal conflict set that is not hit by $\mathsf{node}$. Note that the KB $\mo \setminus \mathsf{node}$ that is given to $\scQX$ as part of the argument DPI ensures that only minimal conflict sets $\mc \subseteq \mo \setminus \mathsf{node}$ can be computed, i.e.\ ones that do not share any single formula with $\mathsf{node}$ (cf.\ Section~\ref{sec:cs_comp}).
 
\begin{remark}\label{rem:qx_with_O_setminus_node_yields_cs_wrt_O}
A minimal conflict set computed by $\scQX(\tuple{\mo \setminus \mathsf{node}, \mb, \Tp, \Tn}_\RQ)$ is a minimal conflict set w.r.t.\ $\tuple{\mo, \mb, \Tp, \Tn}_\RQ$ indeed since (i)~$\scQX(\tuple{\mo \setminus \mathsf{node}, \mb, \Tp, \Tn}_\RQ)$ returning a set $\mc$ means that $\mc$ is a minimal conflict set w.r.t.\ $\tuple{\mo \setminus \mathsf{node}, \mb, \Tp, \Tn}_\RQ$ by Proposition~\ref{prop:qx_correctness} and (ii)~the ``$\Rightarrow$'' direction of Corollary~\ref{cor:validonto_cs} implies that $\mc$ is not valid w.r.t.\ $\tuple{\cdot, \mb, \Tp, \Tn}_\RQ$ and (iii)~the ``$\Leftarrow$'' direction of Corollary~\ref{cor:validonto_cs} lets us conclude that $\mc$ is a minimal conflict w.r.t.\ $\tuple{X, \mb, \Tp, \Tn}_\RQ$ where $X$ is any superset of $\mc$, in particular $X := \mo$. 
\qed
\end{remark}

$\scQX$ may then return 
(a)~'no conflict', i.e.\ $\mo \setminus \mathsf{node}$ is already valid w.r.t.\ $\tuple{\cdot,\mb,\Tp,\Tn}_\RQ$, 
or (b)~a new conflict set $L \neq \emptyset$ such that $L \notin \mC_{calc}$. 
Note that the case of the output $L = \emptyset$ of $\scQX$ cannot arise since (i)~the DPI provided as input to the algorithm is assumed to be admissible, (ii)~no other DPI for which $\scQX$ is called can be non-admissible since admissibility is defined only by the sets $\mb,\Tp,\Tn,\RQ$ which remain unmodified throughout the execution of Algorithm~\ref{algo:hs}, and (iii)~as per Proposition~\ref{prop:qx_correctness}, $\scQX$ returns $\emptyset$ only if the DPI given to it as an argument is non-admissible.
Further on, we point out that the conflict set $L$ in case (b) must be a \emph{new} conflict set since the reuse criterion is always checked \emph{before} the call to $\scQX$ and thus must be negative. That is, each $\mc\in\mC_{calc}$ is hit by $\mathsf{node}$ and $L$ is not hit by $\mathsf{node}$ wherefore $L \neq \mc$ must hold for all $\mc\in\mC_{calc}$. 

In each of the described cases, the \textsc{label} procedure returns a tuple including the respective label as explained and the set $\mC_{calc}$ where $\mC_{calc}$ is equal to the input argument $\mC_{calc}$ in all cases except for the case where a new minimal conflict set is computed by $\scQX$. In this case, the newly computed conflict set is added to $\mC_{calc}$ (line~\ref{algoline:hs:add_conflict}) before the procedure returns.

\paragraph{Processing of a Node Label.} Back in the main procedure, $\mC_{calc}$ is updated (line~\ref{algoline:hs:update_Ccalc}) and then the label $L$ returned by procedure \textsc{label} is processed as follows: 

If $L = valid$, then there is no minimal conflict set w.r.t.\ $\langle\mo,\mb,\Tp,\Tn\rangle_\RQ$ that is not hit by (i.e.\ has an empty intersection with) the current node $\mathsf{node}$. 
Thus, $\mathsf{node}$ is added to the set of calculated minimal diagnoses $\mD_{calc}$. Minimality of diagnoses added to $\mD_{calc}$ is guaranteed by the pruning rule (lines~\ref{algoline:hs:non_min_crit_start}-\ref{algoline:hs:non_min_crit_end}) which eliminates non-minimal nodes (paths) and the way the tree is built level by level by the used breadth-first strategy. In case a uniform-cost variant of tree construction is used, certain properties of the function $p()$ need to be postulated to preserve this minimality guarantee. We discuss these properties in Section~\ref{sec:DiagnosisProbabilitySpace}.

If, on the other hand, $L=closed$ is the returned label of the procedure $\textsc{label}$, then there is either a minimal diagnosis in $\mD_{calc}$ that is a subset of the current node $\mathsf{node}$ or a duplicate of $\mathsf{node}$ is already included in $\Queue$. Consequently, $\mathsf{node}$ must simply be removed from $\Queue$ which has already been executed in line~\ref{algoline:hs:remove_from_queue}.

In the third case, if a minimal conflict set $L$ is returned in line~\ref{algoline:hs:label}, then $L$ is a label for $\mathsf{node}$ meaning that $|L|$ successor nodes of $\mathsf{node}$ need to be added to $\Queue$ in sorted order using the function $p_{nodes}()$ (\textsc{insertSorted}, line~\ref{algoline:hs:generate_nodes}), as will be explained in more detail in Section~\ref{sec:DiagnosisProbabilitySpace}.

\paragraph{Recap.} To summarize, in each iteration, the node $\mathsf{node}$ that is the first element of the queue $\Queue$ is deleted from $\Queue$ and,
\begin{enumerate}
	\item if $\mathsf{node}$ is a diagnosis, it is added to the set $\mD_{calc}$
	\item if there is some diagnosis in $\mD_{calc}$ that is a proper subset of $\mathsf{node}$ or $\mathsf{node}$ is equal to some other node in $\Queue$, no action is performed, i.e.\ the algorithm deletes $\mathsf{node}$ without substitution
	\item if there is some minimal conflict set that $\mathsf{node}$ does not hit, then such a conflict set $\mc$ is computed and for each $\tax \in \mc$ a new node $\mathsf{node} \cup \setof{\tax}$ is added to $\Queue$.
\end{enumerate}
We call each node $\mathsf{nd}$ that is added to $\Queue$ in the latter case a \emph{successor of the node $\mathsf{node}$}. 

\subsection*{Correctness of Breadth-first Diagnosis Computation}
\label{sec:CorrectnessOfBreadthFirstDiagnosisComputation}
For the discussion of the output of Algorithm~\ref{algo:hs} we will exploit the following result saying that Algorithm~\ref{algo:hs} computes all and only minimal diagnoses, if it executes until the queue of open nodes becomes the empty set. 
\begin{proposition}[Soundness and Completeness of Algorithm~\ref{algo:hs} using Breadth-First Search]\label{prop:hs-tree_bfs_soundness_completeness}
Let $\langle\mo$, $\mb,\Tp,\Tn\rangle_\RQ$ be an admissible DPI given as input to Algorithm~\ref{algo:hs}. If Algorithm~\ref{algo:hs} using a breadth-first tree construction strategy terminates due to $\Queue = []$, 
then the algorithm returns exactly the set of all minimal diagnoses w.r.t.\ $\langle\mo,\mb,\Tp,\Tn\rangle_\RQ$. 
\end{proposition}
\begin{proof}
This proposition is a consequence of Proposition~\ref{prop:pruned_hs_tree_finds_all_min_diags} and the following Lemma~\ref{lem:algo_hs_produces_pruned_hs_tree} which witnesses that Algorithm~\ref{algo:hs} using a breadth-first tree construction strategy produces a pHS-tree as per Definition~\ref{def:pruned_hs_tree}.
\end{proof}
%
\begin{lemma}\label{lem:algo_hs_produces_pruned_hs_tree} Algorithm~\ref{algo:hs} with the admissible input DPI $\langle\mo,\mb,\Tp,\Tn\rangle_\RQ$ using a breadth-first tree construction strategy is a procedure for producing a pHS-tree $T$ w.r.t.\ $\langle\mo,\mb,\Tp,\Tn\rangle_\RQ$.
\end{lemma}
\begin{proof}
We verify whether all rules given by Definitions~\ref{def:hs_tree} and \ref{def:pruned_hs_tree} are satisfied by Algorithm~\ref{algo:hs}.
\begin{itemize}
\item Definition~\ref{def:hs_tree}, rule 1: The root node $\emptyset$ which is the only element of the initial list $\Queue$ is labeled by the first call to \textsc{label} for $\mathsf{node} := \emptyset$ in line~\ref{algoline:hs:label}. If $valid$ is returned, then $\scQX(\langle\mo,\mb,\Tp,\Tn\rangle_\RQ)$ must have returned 'no conflict' which is the case if $\mo$ is valid w.r.t.\ $\langle\cdot,\mb,\Tp,\Tn\rangle_\RQ$. 

Otherwise, if $valid$ is not returned by \textsc{label}, then some minimal conflict set $L$ w.r.t.\ $\langle\mo,\mb,\Tp,\Tn\rangle_\RQ$ must have been returned in line~\ref{algoline:hs:return_computed_cs}. $L$ is a minimal conflict set w.r.t.\ $\langle\mo,\mb,\Tp,\Tn\rangle_\RQ$ by Proposition~\ref{prop:qx_correctness} and since $\scQX(\langle\mo,\mb,\Tp,\Tn\rangle_\RQ)$ has not returned 'no conflict' as otherwise $valid$ would have been returned contradicting our assumption and since $\langle\mo,\mb,\Tp,\Tn\rangle_\RQ$ is an admissible DPI by assumption. \textsc{label} cannot have returned earlier in line~\ref{algoline:hs:non_min_crit_end} or line~\ref{algoline:hs:duplicate_crit_end}, since $\mD_{calc}$ is the empty set and $\Queue$ the empty list at this time. The former holds since $\mD_{calc}$ is only extended in line~\ref{algoline:hs:update_Dcalc} which cannot ever have been reached before the first call to \textsc{label} has returned. The latter holds as $\Queue$ initially contained only $\emptyset$ and as $\emptyset$ was deleted from $\Queue$ in line~\ref{algoline:hs:remove_from_queue} before the call to \textsc{label} was made in line~\ref{algoline:hs:label}.

\item Definition~\ref{def:hs_tree}, rule 2: Suppose a node $\mathsf{node}$ is labeled by $valid$, then it is added to $\mD_{calc}$ in line~\ref{algoline:hs:update_Dcalc}. Since $\mathsf{node}$ can only get a label different from $closed$ if it is the only exemplar of this node in $\Queue$ due to the duplicate criterion (lines~\ref{algoline:hs:duplicate_crit_start}-\ref{algoline:hs:duplicate_crit_end}), it must be the case that $\mathsf{node}\notin\Queue$ (line~\ref{algoline:hs:remove_from_queue}) after $\mathsf{node}$ has been labeled by $valid$. Only nodes that get labeled by a conflict set can have successor nodes added to $\Queue$ in line~\ref{algoline:hs:generate_nodes}. Only nodes in $\Queue$ can get a label (cf.\ lines~\ref{algoline:hs:getfirst} and \ref{algoline:hs:label}). For $\mathsf{node}$ to be added to $\Queue$ at some later point in time there must be a proper subset of $\mathsf{node}$ that is still in $\Queue$ as each node newly added to $\Queue$ is a proper superset of some node in $\Queue$ (cf.\ line~\ref{algoline:hs:generate_nodes} which is the only position in the algorithm where nodes are added to $\Queue$). This is impossible due to the breadth-first tree construction strategy which implies that all nodes of cardinality $|\mathsf{node}|-1$ have already been labeled (and thus deleted from $\Queue$ in line~\ref{algoline:hs:remove_from_queue}) when $\mathsf{node}$ is being labeled. Hence, if $\mathsf{node}$ is labeled by $valid$, then it has no successors.

If $\mathsf{node}$ is labeled by some conflict set $L$, then Algorithm must come to line~\ref{algoline:hs:generate_nodes}, where a successor $\mathsf{node} \cup \setof{e}$ is added to $\Queue$ for all $e \in L$.

How node $\mathsf{node}_e := \mathsf{node} \cup \setof{e}$ must be labeled is overridden by the rules 3, 4 and 6 of Definition~\ref{def:pruned_hs_tree} (see below). 

\item Definition~\ref{def:pruned_hs_tree}, rule 1: This is true by our assumption about $p()$ and $p_{nodes}()$.

\item Definition~\ref{def:pruned_hs_tree}, rule 2: This holds since $\scQX(\langle\mo\setminus\mathsf{node},\mb,\Tp,\Tn\rangle_\RQ)$ computes only minimal conflict sets w.r.t.\ $\langle\mo,\mb,\Tp,\Tn\rangle_\RQ$ (cf.\ Remark~\ref{rem:qx_with_O_setminus_node_yields_cs_wrt_O}).

\item Definition~\ref{def:pruned_hs_tree}, rule 3: All minimal conflict sets that have been used to label nodes so far are stored in $\mC_{calc}$. Before a minimal conflict to label $\mathsf{node}$ might be computed by a call to $\scQX$ in line~\ref{algoline:hs:qx}, the reuse criterion in lines~\ref{algoline:hs:reuse_crit_start}-\ref{algoline:hs:reuse_crit_end} checks whether there is a set $\mc$ in $\mC_{calc}$ with $\mc \cap \mathsf{node}$. If positive, $\mc$ is returned as a label for $\mathsf{node}$.

\item Definition~\ref{def:pruned_hs_tree}, rule 4: This is accomplished by the non-minimality criterion in lines~\ref{algoline:hs:non_min_crit_start}-\ref{algoline:hs:non_min_crit_end} which checks for existence of a node already labeled by $valid$ which is a subset of the node to be labeled right now. All nodes labeled by $valid$ are stored in $\mD_{calc}$ (cf. lines~\ref{algoline:hs:L=valid} and \ref{algoline:hs:update_Dcalc}).

\item Definition~\ref{def:pruned_hs_tree}, rule 5: If some node $\mathsf{node}$ is labeled by $closed$, then no action is performed (cf.\ line~\ref{algoline:hs:do_nothing}). Before each node is labeled in line~\ref{algoline:hs:label}, it is deleted from $\Queue$ in line~\ref{algoline:hs:remove_from_queue}. That $\mathsf{node}$ cannot be inserted into $\Queue$ at some later point in time follows from the argumentation used above to demonstrate that Definition~\ref{def:hs_tree}, rule 2 is met.

\item Definition~\ref{def:pruned_hs_tree}, rule 6: This is achieved by the duplicate criterion in lines~\ref{algoline:hs:duplicate_crit_start}-\ref{algoline:hs:duplicate_crit_end} where $\Queue$ is browsed for some node equal to the one that is to be labeled right now. 
When some node $\mathsf{node}$ is next to be labeled, then all duplicates of $\mathsf{node}$ must already be in $\Queue$ as reasoned above in the argumentation to show that Definition~\ref{def:hs_tree}, rule 2 is satisfied. Thus, the criterion must search for duplicates in no other collections than $\Queue$. Indeed, only one (i.e.\ the last non-deleted) exemplar of these duplicates of $\mathsf{node}$ in $\Queue$ can get a label other than $closed$ due to the duplicate criterion which closes duplicates as long as there are any.  
\end{itemize}
We conclude that Algorithm~\ref{algo:hs} is a procedure for constructing a pHS-tree.
\end{proof}
By Proposition~\ref{prop:hs-tree_bfs_soundness_completeness} and the fact that there is no place in Algorithm~\ref{algo:hs} where nodes are removed from $\mD_{calc}$ (which
implies that only \emph{minimal} diagnoses can be added to $\mD_{calc}$), the following corollary is obvious.
\begin{corollary}\label{cor:algo_hs_returns_relevant_data_of_(partial)_pruned_hs_tree}
Algorithm~\ref{algo:hs} with the admissible input DPI $\langle\mo,\mb,\Tp,\Tn\rangle_\RQ$ using a breadth-first tree construction strategy stores by $\tuple{\mD_{calc},\Queue,\mC_{calc}}$ the relevant data of 
\begin{itemize}
\item a pHS-tree w.r.t.\ $\langle\mo,\mb,\Tp,\Tn\rangle_\RQ$ if Algorithm~\ref{algo:hs} stops due to $\Queue = []$,
\item a partial pHS-tree w.r.t.\ $\langle\mo,\mb,\Tp,\Tn\rangle_\RQ$ otherwise.
\end{itemize}
\end{corollary}
If a pHS-tree is computed in breath-first order, minimal diagnoses are generated with increasing cardinality, as the following Corollary~\ref{cor:algo_hs_finds_min_card_diags_first} attests. Consequently, for the generation of all minimum cardinality diagnoses, only the first level of the tree has to be generated, where a node is labeled.
\begin{corollary}\label{cor:algo_hs_finds_min_card_diags_first}
If Algorithm~\ref{algo:hs} using breadth-first search returns a set $\mD$ of cardinality $n$, then $\mD$ is the set of diagnoses of minimum cardinality w.r.t.\ the DPI $\langle\mo,\mb,\Tp,\Tn\rangle_\RQ$ given as input to the algorithm. That is, if $\mD$ contains some diagnosis of cardinality $k$, then it includes all diagnoses w.r.t.\ $\langle\mo,\mb,\Tp,\Tn\rangle_\RQ$ of cardinality lower than $k$.
\end{corollary}
\begin{proof}
By Proposition~\ref{prop:hs-tree_bfs_soundness_completeness}, it is a fact that the algorithm computes all and only minimal diagnoses w.r.t.\ $\langle\mo,\mb,\Tp,\Tn\rangle_\RQ$. As these are computed in breadth-first order, the first computed diagnoses must be the minimum cardinality ones. To see this, assume that $\md$ with $|\mD| = n$ is returned which includes one non-minimum cardinality diagnosis $\md$ and does not comprise a minimum cardinality diagnosis $\md'$, i.e.\ $|\md| > |\md'|$. By breadth-first search, 
nodes are labeled in ascending order of their cardinality. And, if the first node of cardinality $k$ is labeled, no more nodes of cardinality $k-1$ can be in $\Queue$ (cf.\ proof of Lemma~\ref{lem:algo_hs_produces_pruned_hs_tree}).
So, we have that the pHS-tree obtained by further execution of the algorithm until $\Queue = []$ can never label $\md'$ since $|\md| > |\md'|$ and $\md$ has already been labeled. Hence, the algorithm would not return $\md'$ in its final output $\mD$. 
Since each minimum cardinality diagnosis is a minimal diagnosis, $\md'$ is a minimal diagnosis. Thus, we have a contradiction to the fact that the algorithm computes all minimal diagnoses.
\end{proof}

\paragraph{Output.} The repeat-loop is iterated until the stop criterion (line~\ref{algoline:hs:until}) applies. 
In case at least $n_{\min}$ minimal diagnoses w.r.t.\ $\langle\mo,\mb,\Tp,\Tn\rangle_\RQ$ exist, there are two cases:
\begin{itemize}
\item If the finding of the $n_{\min}$-th minimal diagnosis happens after $t' < t$ time has passed since the start of Algorithm~\ref{algo:hs}, then the algorithm will continue iterating and terminate only if execution time amounts to at least $t$ time or $|\mD|=n_{\max}$ at the time line~\ref{algoline:hs:until} is processed.
\item Otherwise, if the detection of the $n_{\min}$-th minimal diagnosis takes place after processing longer than $t$ time, then the algorithm will terminate immediately after having determined the $n_{\min}$-th minimal diagnosis.
\end{itemize}
In both cases, the output is a set $\mD$ of minimal diagnoses w.r.t.\ $\langle\mo,\mb,\Tp,\Tn\rangle_\RQ$ such that $n_{\min} \leq |\mD| \leq n_{\max}$ and $\mD$ is the set of best minimal diagnoses as per $p()$, in this case the set of minimal diagnoses with minimum cardinality since $p()$ is assumed to be specified as to cause a breadth-first tree construction. 

If fewer than $n_{\min}$ minimal diagnoses exist w.r.t.\ $\langle\mo,\mb,\Tp,\Tn\rangle_\RQ$, then $\Queue = []$ will be the cause for the algorithm to terminate. In this case, the pHS-tree w.r.t.\ $\langle\mo,\mb,\Tp,\Tn\rangle_\RQ$ has been built up and all minimal diagnoses w.r.t.\ $\langle\mo,\mb,\Tp,\Tn\rangle_\RQ$ are stored in $\mD_{calc}$. Thus, the output is the set $\minD_{\langle\mo,\mb,\Tp,\Tn\rangle_\RQ}$ of all minimal diagnoses w.r.t.\ $\langle\mo,\mb,\Tp,\Tn\rangle_\RQ$.


\paragraph{Termination.} The next proposition shows that Algorithm~\ref{algo:hs} must yield a set of minimal diagnoses after finite time.
\begin{proposition}\label{prop:hs:termination}
Algorithm~\ref{algo:hs} always terminates.
\end{proposition}
\begin{proof}
This is due to the fact that minimal conflict sets used to label non-leaf nodes are subsets of $\mo$ and that nodes in $\Queue$ are subsets of $\mo$, which is a finite set by Definition~\ref{def:dpi}. Moreover, 
a node in $\Queue$ is either deleted without substitution from $\Queue$ if $valid$ or $closed$ (line~\ref{algoline:hs:remove_from_queue}) or deleted (line~\ref{algoline:hs:remove_from_queue}) and replaced by proper supersets of it (\textsc{insertSorted} in line~\ref{algoline:hs:generate_nodes}). This means that the cardinality of all nodes in $\Queue$ is strictly monotonically increasing. Thus each node (path) $\mathsf{node}$ is guaranteed to be closed ($valid$ or $closed$) when $\mathsf{node} = \mo$ as in this case $\mathsf{node}$ must hit all possible (minimal) conflict sets $\mc_i$ w.r.t.\ $\langle\mo,\mb,\Tp,\Tn\rangle_\RQ$ since $\mc_i \subseteq\mo$ holds by Definition~\ref{def:cs}. So, after finite time the queue $\Queue$ definitely becomes the empty list which is a stop criterion (line~\ref{algoline:hs:until}).
\end{proof}

The argumentation so far proves the following 
\begin{proposition}\label{prop:hs_bfs_correct}
Let $\langle\mo,\mb,\Tp,\Tn\rangle_\RQ$ be an admissible DPI, $t,n_{\min},n_{\max}\in\mathbb{N}$ 
and $p:\mo \rightarrow (0,0.5)$ defined in a way that $\Queue$ is always ordered first-in-first-out.
For these inputs, Algorithm~\ref{algo:hs} always terminates and returns a set $\mD$ of minimal diagnoses w.r.t.\ $\langle\mo,\mb,\Tp,\Tn\rangle_\RQ$ which is 
\begin{itemize}
\item the set of the $|\mD|$ minimal diagnoses of minimum cardinality w.r.t.\ $\langle\mo,\mb,\Tp,\Tn\rangle_\RQ$ (i.e.\ the first $|\mD|$ elements in $\minD_{\langle\mo,\mb,\Tp,\Tn\rangle_\RQ}$ if $\minD_{\langle\mo,\mb,\Tp,\Tn\rangle_\RQ}$ is assumed to be sorted in ascending order by cardinality) such that $n_{\min} \leq |\mD| \leq n_{\max}$, if at least $n_{\min}$ minimal diagnoses exist w.r.t.\ $\langle\mo,\mb,\Tp,\Tn\rangle_\RQ$, or
\item the set of all minimal diagnoses w.r.t.\ $\langle\mo,\mb,\Tp,\Tn\rangle_\RQ$, 
otherwise.
\end{itemize}
\end{proposition}

\begin{algorithm*}
\small
\caption{$\scHS$: Computation of Minimal Diagnoses} \label{algo:hs}
\begin{algorithmic}[1]
\Require an admissible DPI $\langle\mo,\mb,\Tp,\Tn\rangle_\RQ$, 
a desired computation timeout $t$, a desired minimal ($n_{\min}$) and maximal ($n_{\max}$) number of diagnoses to be returned, a function $p:\mo \rightarrow (0,0.5)$ 
\Ensure a set $\mD$ which is \newline
(a)~a set of most probable (according to $p()$) minimal diagnoses w.r.t.\ $\langle\mo,\mb,\Tp,\Tn\rangle_\RQ$ such that $n_{\min} \leq |\mD| \leq n_{\max}$, if at least $n_{\min}$ minimal diagnoses exist w.r.t.\ $\langle\mo,\mb,\Tp,\Tn\rangle_\RQ$, or \newline
(b)~the set of all minimal diagnoses w.r.t.\ $\langle\mo,\mb,\Tp,\Tn\rangle_\RQ$ otherwise
\vspace{10pt}
\Procedure{$\scHS$}{$\langle\mo,\mb,\Tp,\Tn\rangle_\RQ, t, n_{\min}, n_{\max}, p()$}
\State $t_{start} \gets \Call{getTime}{ }$
\State $\mD_{calc}, \mC_{calc} \gets \emptyset$
\State $\Queue \gets [\emptyset]$							
\Repeat 																							\label{algoline:hs:repeat}
\State $\mathsf{node} \gets \Call{getFirst}{\Queue}$	\label{algoline:hs:getfirst}
\State $\Queue \gets \Call{deleteFirst}{\Queue}$ 	\label{algoline:hs:remove_from_queue}
\State $\tuple{L,\mathbf{C}} \gets \Call{label}{\langle\mo,\mb,\Tp,\Tn\rangle_\RQ, \mathsf{node}, \mathbf{C}_{calc}, \mD_{calc}, \Queue}$\label{algoline:hs:label}
\State $\mathbf{C}_{calc} \gets \mathbf{C}$    				\label{algoline:hs:update_Ccalc}
\If{$L = valid$}\label{algoline:hs:L=valid}
	\State $\mD_{calc} \gets \mD_{calc} \cup \setof{\mathsf{node}}$		\label{algoline:hs:update_Dcalc}
\ElsIf{$L = closed$}\label{algoline:hs:do_nothing}  \Comment{do nothing}
\Else 	\Comment{$L$ must be a minimal conflict set}
	\For{$e \in L$}						
		\State $\Queue \gets \Call{insertSorted}{ \mathsf{node} \cup \setof{e}, \Queue, p_{nodes}()}$\label{algoline:hs:generate_nodes}
\EndFor
\EndIf
\Until{$\Queue = [] \lor [|\mD_{calc}| \geq n_{\min} \land ( |\mD_{calc}| = n_{\max} \lor \Call{getTime}{ } - t_{start} > t)]$}  \label{algoline:hs:until}
\State \Return $\mD_{calc}$
\EndProcedure
\vspace{10pt}
\Procedure{\textsc{label}}{$\langle\mo,\mb,\Tp,\Tn\rangle_\RQ,\mathsf{node},\mathbf{C}_{calc},\mD_{calc}, \Queue$}    \label{algoline:hs:procedure_label}
\For{$\mathsf{nd} \in \mD_{calc}$}																								\label{algoline:hs:non_min_crit_start}
	\If{$\mathsf{node} \supseteq \mathsf{nd}$}  \Comment{non-minimality}
			\State \Return $\tuple{closed,\mathbf{C}_{calc}}$														\label{algoline:hs:non_min_crit_end}
	\EndIf
\EndFor
\For{$\mathsf{nd} \in \Queue$}																										\label{algoline:hs:duplicate_crit_start}
	\If{$\mathsf{node} = \mathsf{nd}$}  \Comment{remove duplicates}
			\State \Return $\tuple{closed,\mathbf{C}_{calc}}$														\label{algoline:hs:duplicate_crit_end}
	\EndIf
\EndFor
\For{$\mc \in \mathbf{C}_{calc}$}																									\label{algoline:hs:reuse_crit_start}
	\If{$\mc \cap \mathsf{node} = \emptyset$}\label{algoline:hs:test_cs_not_hit}  \Comment{reuse $\mc$}
		\State \Return $\tuple{\mc,\mathbf{C}_{calc}}$																\label{algoline:hs:reuse_crit_end}
	\EndIf
\EndFor
\State $L\gets \Call{QX}{\langle\mo\setminus\mathsf{node},\mb,\Tp,\Tn\rangle_\RQ}$	\label{algoline:hs:qx}
\If{$L$ = \text{'no conflict'}}   \Comment{$\mathsf{node}$ is a diagnosis}
	\State \Return $\tuple{valid,\mathbf{C}_{calc}}$														\label{algoline:hs:return_valid}
\Else										\Comment{$L$ is \emph{new} minimal conflict set ($\notin \mathbf{C}_{calc}$)}
	\State $\mathbf{C}_{calc} \gets \mathbf{C}_{calc} \cup \setof{L}$							\label{algoline:hs:add_conflict}  
	\State \Return $\tuple{L,\mathbf{C}_{calc}}$\label{algoline:hs:return_computed_cs}
\EndIf
\EndProcedure
\end{algorithmic}
\normalsize
\end{algorithm*}

\section{Diagnosis Probability Space}
\label{sec:DiagnosisProbabilitySpace}
The induction of a probability space~\cite{durrett2010} over diagnoses facilitates incorporation of well-established probability theoretic methods into the process of KB debugging; for example, a Bayesian approach~\cite{Shchekotykhin2012,Rodler2013,dekleer1987} for identifying the \emph{true diagnosis}, i.e.\ the one which leads to a solution KB with the desired semantics, by repeated measurements (see Chapter~\ref{chap:InteractiveOntologyDebugging}). Let the true diagnosis be denoted as $\dt$ in the sequel. 

\paragraph{The Probability Space of All Diagnoses.} From the point of view of probability theory, a diagnosis can be viewed as an atomic event in a probability space $\langle\Omega,\mathcal{E},p\rangle$ defined as follows:
\begin{itemize}
	\item $\Omega$ is the sample space consisting of all possible diagnoses w.r.t. a DPI $\tuple{\mo,\mb,\Tp,\Tn}_\RQ$, i.e. $\Omega = \allD_{\tuple{\mo,\mb,\Tp,\Tn}_\RQ}$,
	\item $\mathcal{E}$ is a sigma-algebra on $\Omega$, in our case 
	the powerset $2^{\Omega}$ of $\Omega$, and
	\item $p$ is a probability measure assigning a probability to each event in $\mathcal{E}$, i.e. $p:\mathcal{E} \rightarrow [0,1]$ such that $\sum_{\omega\in\Omega} p(\setof{\omega}) = 1$ which means $\sum_{\md\in\allD_{\tuple{\mo,\mb,\Tp,\Tn}_\RQ}} p(\setof{\md}) = 1$.
\end{itemize}
So, $p(\setof{\md})$ for $\md\in\allD_{\tuple{\mo,\mb,\Tp,\Tn}_\RQ}$ can be seen as the probability that $\md$ is the true diagnosis, i.e.\ the probability of the event $\dt = \md$ (or $\dt \in \setof{\md}$). Consequently, $p(\setof{\md})$ for $\md\in\allD_{\tuple{\mo,\mb,\Tp,\Tn}_\RQ}$ is the probability distribution of the random variable $\dt$, i.e.\ the probability distribution of the true diagnosis.
In this vein, the probability of a set $\setof{\md_i,\dots,\md_j} \in \mathcal{E}$ is interpreted as the likeliness of this set to comprise the true diagnosis $\dt$. That is, $p(\{\md_i,\dots,\md_j\}) = p(\dt\in\{\md_i,\dots,\md_j\}) = p(\dt=\md_i \vee \dots \vee \dt=\md_j) = 0.3$ means that $\dt$ is an element of $\setof{\md_i,\dots,\md_j}$ with 30\% probability. Note that singletons are often written without curly braces, i.e.\ $p(\setof{\md_i})$ is usually written as $p(\md_i)$; we will also do so in the rest of this work. 

The elements of the sample space $\Omega$ of a probability space are often called atomic events because they must be \emph{mutually exclusive} (i.e.\ two atomic events cannot ``happen'' at the same time as an outcome of the fictive experiment a probability space describes) and \emph{exhaustive} (i.e.\ for each ``execution'' of the experiment the probability space describes one atomic event must ``happen''). Since the true diagnosis $\dt$ must be a diagnosis w.r.t.\ $\tuple{\mo,\mb,\Tp,\Tn}_\RQ$ and $\Omega$ by definition comprises all such diagnoses, exhaustiveness is clearly fulfilled.
Mutual exclusiveness is a consequence of the fact that each diagnosis $\md$ 
gives complete information about the correctness of each formula $\tax_k \in \mo$. In other words, $\dt \in \setof{\md}$ is a shorthand for the statement that all $\tax_i \in \md$ are faulty and all $\tax_j \in \mo\setminus\md$ are correct. Thus, any two different diagnoses are mutually exclusive events, i.e. $\dt = \md_i$ implies $\dt \neq \md_j$ for all $\md_j \in \allD$ such that $\md_i \neq \md_j$. 

The probability measure $p$ is completely defined if a probability $p(\md)$ for each diagnosis $\md \in \Omega$ is given. Then, by the mutual exclusiveness of events $\dt \in \setof{\md_i}$ and $\dt \in \setof{\md_j}$ for $\md_i \neq \md_j$, the probability 
\begin{align}
p(E) = \sum_{\md\in E} p(\md)
\end{align}
for each event $E \in \mathcal{E}$.

\paragraph{Restricted Probability Spaces of Diagnoses.} 
In many cases, only a restricted set of diagnoses w.r.t.\ a DPI is considered relevant for the debugging task. That is, the focus is on locating the true diagnosis among a predefined subset of all diagnoses $\allD_{\tuple{\mo,\mb,\Tp,\Tn}_\RQ}$.
This involves an adaptation of the probability space, in particular of the set $\Omega$. For instance, if not the set of all, but only the set of minimal diagnoses $\minD_{\tuple{\mo,\mb,\Tp,\Tn}_\RQ}$ w.r.t.\ $\tuple{\mo,\mb,\Tp,\Tn}_\RQ$ should be considered by a debugging system -- as motivated in Section~\ref{sec:MinimallyInvasiveOntologyDebugging} --  then $\Omega := \minD_{\tuple{\mo,\mb,\Tp,\Tn}_\RQ}$. The other properties $\mathcal{E} = 2^{\Omega}$ and $\sum_{\omega\in\Omega} p(\setof{\omega}) = 1$ remain the same for each restricted probability space, but depend on $\Omega$. Thus, for example, a probability $p(\md)$ for $\md \in \minD_{\tuple{\mo,\mb,\Tp,\Tn}_\RQ} \subseteq \allD_{\tuple{\mo,\mb,\Tp,\Tn}_\RQ}$ must be generally defined differently, i.e.\ assigned a higher value, when $\Omega = \minD_{\tuple{\mo,\mb,\Tp,\Tn}_\RQ}$ instead of $\Omega = \allD_{\tuple{\mo,\mb,\Tp,\Tn}_\RQ}$. This is due to the condition that all probabilities of atomic events in $\Omega$ must sum up to 1. In practice, because of the computational complexity of diagnosis computation, the used probability space will usually need to be restricted even further in that $\Omega$ comprises only a set of ``leading diagnoses'' which is a subset of all minimal diagnoses w.r.t.\ a DPI (see Section~\ref{sec:UserInteraction}).

\subsection{Construction of a Probability Space} 
\label{sec:prob_space_construction}
Since a diagnosis constitutes an assumption about the correctness of each formula in the KB, the probability of a diagnosis $\md$ (to be the true diagnosis $\dt$) can be computed by means of fault probabilities of formulas. In other words, computing the probability of the event $\md = \dt$ corresponds to computing the probability of the event that exactly all formulas in $\md$ are faulty and all other formulas in the KB are correct.

\subsubsection*{Estimating Fault Probabilities of Formulas in the KB}
Next we discuss various possibilities of how the probability of an $\tax\in\mo$ might be assessed. To this end, we first make a distinction between situations where some useful empirical data is available or not and then we differentiate between different sorts of such available data and how to take advantage of it. 

\paragraph{Empirical Data is Accessible.} Let us first reflect on how to utilize different empirical data sources in order to compute formula probabilities. Data can be of the following kinds (enumeration may not be complete): 
\begin{enumerate}[(a)]
	\item Regarding formulas: Change logs of formulas in the KB
	\item Regarding the user: Data about common mistakes of the user who has formulated the KB
\end{enumerate}

\noindent\emph{Ad (a):} Prerequisite for the availability of change logs of formulas in the KB is the usage of some KB engineering software with integrated
logging or change management. Examples of such KB (ontology) developing environments are Prot\'{e}g\'{e}~\cite{Noy2000}, Web Prot\'{e}g\'{e}~\cite{Tudorache2013}, SWOOP~\cite{Kalyanpur2006b}, OntoEdit~\cite{Sure2002} or KAON2\footnote{http://kaon2.semanticweb.org/}. 
Given a formula $\tax\in\mo$ and its change log, the fault probability $p(\tax)$ of this formula can be estimated by counting the number of modifications accomplished for $\tax$ in the change log. The intuition is, the more often $\tax$ has been altered, the more uncertain the (set of) author(s) might be about its correctness. This method of probability computation however suffers from a cold-start problem. If a KB is completely newly created, then such information is not available at all. On the other hand, for KBs that are being developed over a long period of time, this method can be assumed to be a rather reliable way of assessing the likeliness of formulas to be faulty.

\noindent\emph{Ad (b):} Clearly, data about common mistakes of a user has to be related to some type of entity that is recurrent and not dependent on a particular KB. Formulas are therefore not suitable and too coarse-grained since one and the same formula will rarely occur in many KBs. More adequate entities to relate a user fault to are predicates (terms) and logical connectives -- these usually (re-)appear in many different KBs. In this way, the extrapolation and reusability of collected personal fault information of a user within one KB and between different KBs is granted.    

One way of obtaining data about common mistakes of user $u$ on this syntactical level is, for instance, the examination of diagnoses got as a result of past debugging sessions performed on KBs authored by $u$. Another way is, again, to use the change logs (if available) of formulas in KBs user $u$ has created in the past. 

Given such a past diagnosis $\md$, we know that all formulas $\tax\in\md$ \emph{that had been written by $u$} have been confirmed to be faulty by a user. So, these formulas could be analyzed for contained predicates (terms) and logical connectives and the probability of being faulty of those syntactical constructs could be raised relative to those constructs that do not occur in formulas in $\md$. At this, the following assumptions could be made:
\begin{itemize}
\item If a formula has been confirmed to be faulty by the user, then the meaning of all predicates (terms) appearing in this formula is not correct (because in the domain that should be modeled the relationship between the predicates (terms) occurring in the formula stated by the formula must not hold). So, all predicates (terms) in $\tax$ get more suspicious of being faulty \emph{in general} if $\tax\in\md$ for some past solution diagnosis $\md$.
\item If a formula including some logical connective is part of some past solution diagnosis, then this type of logical connective gets more suspicious of being faulty \emph{in general}.
\end{itemize}

When exploiting change logs of formulas \emph{authored by $u$}, the following assumptions could be made: 
\begin{itemize}
\item If a formula has been modified, then a user has changed the meaning of all predicates (terms) appearing in this formula. So, all predicates (terms) in $\tax$ get more suspicious of being faulty \emph{in general} if $\tax$ has been edited at least once. The more often it has been altered, the more suspicious the predicates (terms) get.
\item If some logical connective in a formula is modified, i.e.\ deleted or added, then this type of logical connective gets more suspicious of being faulty \emph{in general}.
\end{itemize}

The following example should give an intuition of these assumptions:
\begin{example}
Imagine the situation where the author of formula $\tax := \forall X\, pet(X) \leftrightarrow animal(X) \land (\exists Y hasOwner(X,Y) \land person(Y))$ is known to have only vague knowledge about the predicate $pet$ and to frequently interchange $\land$ and $\lor$ when formulating logical formulas. This could be reflected by the assignment of higher fault probability to the predicate $pet$ than to the predicates $animal, hasChild$ and $person$ and by raising the fault probability of $\land$ as well as $\lor$ compared to other logical connectives available in the used logic $\mathcal{L}$. Then, formula $\tax$ should intuitively have a higher probability of being faulty than, e.g., formula $\tax' = \forall X\, animal(X) \rightarrow \lnot person(X)$ since $\tax'$ does not include any of the ``suspicious'' terms or connectives as $\tax$ does.\qed
\end{example}

A probability of $0.25$ of some predicate (term) $a$ occurring in $\mo$ could then account for the observation made in the logs that, in past debugging sessions (not necessarily related to the current KB $\mo$), every fourth formula formulated by user $u$ which includes the term $a$ was modified at least once. Similarly, another term $b$ could be assigned fault probability $0.5$ which could reflect that formulas formulated by $u$ including $b$ have been altered twice as often as formulas formulated by $u$ comprising $a$. Given additionally that $a$ occurred in two formulas formulated by $u$ of past diagnoses whereas $b$ did not occur in any, the probability of $a$ could be increased by some addend or factor to take account of this. 

Concerning some logical connective, say $\exists$, the observation that all past diagnosis formulas contained $\exists$ and in $80\%$ of formulas formulated by this user including $\exists$ the $\exists$ connective has been modified at least once, the fault probability of $\exists$ might be assigned rather high. In comparison, the probability of some other connective, say $\lnot$, occurring in no diagnosis and having been altered only in $10\%$ of the formulas comprising $\lnot$, the probability of the $\lnot$ connective might be estimated rather low.

A shortcoming of this approach is again a cold-start problem. If a user is new to conceptualizing knowledge in a structured logical manner or at least in the given logical language $\mathcal{L}$, then no such (personalized) past diagnoses or change logs will be available. So, this issue especially concerns beginners who are usually anyhow more prone to errors than expert-users. On the positive side, utilization of such empirical data can yield to fault information that is very well tailored for the user and that can imply a significant reduction of computation time and user effort necessary for debugging of the KB at hand~\cite{Shchekotykhin2012}. 

\paragraph{No Empirical Data is Available.} If no data of the kinds (a) and (b) discussed above is available to a debugging system, then we have the following possibilities:
\begin{enumerate}[(a)]
	\setcounter{enumi}{+2}
	\item Common fault patterns
	\item Subjective self-assessment of a user
	\item Examination of structural complexity of logical formulas
	\item Using no probabilities
\end{enumerate}

\noindent\emph{Ad (c):} A common fault pattern~\cite{Rector2004,Corcho09,Kalyanpur2006}, also called anti-pattern, refers to a set of formulas that either leads to an inconsistency (logical anti-pattern) or corresponds to a potential modeling error that -- alone -- does not lead to a inconsistency or incoherency (non-logical anti-pattern), but still might become a source of inconsistency if merged with other formulas (cf.\ Section~\ref{sec:BackgroundKnowledge}).
Although most of these patterns incorporate more than one formula which makes the individual consideration of a formula in terms of fault probability calculation difficult, an idea to incorporate knowledge about anti-patterns to probability estimation of formulas could be to count for each $\tax\in\mo$ in how many different (logical or non-logical) anti-patterns it occurs. The higher this count, the more likely a formula might be involved in a conflict set and thus in the true diagnosis. 

A drawback of this method could be that most of the formulas involved in a KB might not correspond to any formula occurring in an anti-pattern. Thus, one might end up with no probability estimate for most of the formulas in a KB $\mo$. Besides that, the information provided by these anti-patterns is not personalized at all 
and therefore might significantly diverge from the true fault probabilities for a user and lead to a false bias in the used fault data. This justifies to basically rely on another approach to get a first estimate of a formula's likeliness of being faulty and use this method only to make adaptations to already established probabilities.

\noindent\emph{Ad (d):} The method of a user's self-assessment of own fault probabilities supposes a user to be able to specify fault probabilities of predicates (terms), logical connectives or complete formulas by themselves. Since users not always have a clear picture of own strengths and weaknesses, this variant must be regarded with suspicion. Furthermore, in settings where several persons are involved in the engineering of the KB, a reasonable rating of fault probabilities of terms, connectives or formulas authored by other persons might be difficult or impossible for a user.

\noindent\emph{Ad (e):} Here the idea is to examine ``grammatical'' (i.e.\ syntactical) aspects of formulas such as the ``nesting depth'' of subordinate clauses or the mere ``length'' of a formula. The underlying assumption can be that higher length and/or deeper nesting  means higher complexity and cognitive difficulty in understanding of the formula's semantics -- as it does in natural language. For instance, it is reasonable to expect formulas like $\tax_1 := \forall X\, a(X) \rightarrow (\exists Y\, r_1(X,Y) \land (\forall Z\, r_2(Y,Z) \rightarrow b(Z)))$ 
to tend to be more error-prone and more likely to be faulty than $\tax_2 := \forall X\, g(X) \rightarrow b(X)$. 
This intuition is modeled by the maximum nesting depth as well as by the length of $\tax_1$ in comparison to $\tax_2$. Using the analogy to natural language, the maximum nesting depth of a formula could roughly be defined as the maximum number of encapsulated subordinate clauses that cannot be ``flattened'' occurring in the natural language translation of the formula. For formula $\tax_1$, this would imply a maximum nesting depth of two; for $\tax_2$ it would amount to zero. The reason is that $\tax_1$ stated in natural language would sound ``if somebody $X$ is $a$, then there is somebody $Y$, who satisfies property $r_1$ with $X$ and for whom anybody, who satisfies property $r_2$ with $Y$ is $b$''. In this natural language formulation, there are two subordinate clauses, i.e.\ the clauses beginning with the word ``who''; the first is at nesting depth one and the second at depth two. These subordinate clauses cannot be flattened, i.e.\ be brought to some lower depth, because the $Z$ is related to the $Y$ which in turn is related to the $X$.
The length of formulas could be defined similarly as in~\cite{Horridge2008} which provides such a definition for DL languages. 
In this case the length of $\tax_1$ and $\tax_2$ would be four (roughly: four predicates in $\tax_1$) and two (two predicates in $\tax_2$), respectively. 

A disadvantage of such a ``grammatical'' approach gets evident when most of the formulas in a KB are rather ``simple'', i.e.\ have a low nesting depth and a short length. In such case this method will give little differentiation between different formulas and should thus be combined with another method of probability estimation in general.

\noindent\emph{Ad (f):} In a situation where all the aforementioned ways of gauging probabilities do not apply or are believed to have a too high risk of introducing a false bias into the debugging system, the solution is to define all formulas to be equally probably faulty. The obvious pro of this is that the system cannot get misled by unreasonable fault probabilities whereas the con is that possibly well-suited probabilistic information cannot be exploited. Moreover, experiments in our previous work \cite{Shchekotykhin2012} have manifested that fault information of only ``average'' quality most often leads to a better performance than no fault information. Apart from that, we have suggested a reinforcement learning ``plug-in'' to a debugger which could successfully mitigate the negative effect of low-quality fault information and in many cases, in spite of the low-quality fault information, even led to lower resource consumption (user, time) than a debugger without this plug-in using good fault information \cite{Rodler2013}.

\paragraph{Collaborative KB Development.} In a collaborative development scenario involving several authors, provenance information could be additionally leveraged to refine probability estimates (cf.\ \cite{Kalyanpur2006}). At this point, user skills could come into play; that is, formulas authored by more experienced authors get a lower overall fault probability as opposed to beginners concerning KB engineering or logic skills or expertise in the modeled domain. This probability adaptation can also affect syntactical elements in that one and the same predicate (term) or logical connective can get a different probability depending on in which formula it occurs and who authored that formula.

\begin{remark}\label{rem:problems_with_probs_of_syntactical_elements}
Of course, these assumptions and methods of obtaining fault probabilities of syntactical elements and formulas are only \emph{some} possible ways of doing so. For example, one might argue that the ``authorship'' of a formula is somewhat not clearly defined. What if user $u_1$ has originally written formula $\tax$ and then user $u_2$ alters the formula to become $\tax'$? Who is the author of $\tax'$? $u_1$, $u_2$ or both? For whose fault probability computation should the renewed modification of $\tax'$ to $\tax''$ count? Questions like this one need to be discussed and maybe evaluations using real data need to be accomplished in order to find a practical answer; or perhaps to find out that completely different approaches turn out to be reasonable. This is a topic of our future work.\qed
\end{remark}

\begin{remark}\label{rem:ax_prob_not_zero}
By the definition of a DPI (Definition~\ref{def:dpi}) stating that the KB $\mo$ must be disjoint with the background knowledge $\mb$ and the role $\mb$ has within a DPI, namely to comprise all formulas that are definitely correct, we postulate that no formula $\tax \in \mo$ must have a probability of zero. 
In a situation when this is not the case, a modified DPI must be used where such formulas have been moved from $\mo$ to $\mb$.\qed 
\end{remark}

\paragraph{Computation of Diagnosis Probabilities.} In the following, we denote by $\overline{\tax}$ ($\overline{\mo}$) the set of logical connectives and quantifiers occurring in a formula $\tax$ (in the KB $\mo$) and by $\widetilde{\tax}$ ($\widetilde{\mo}$) the signature of $\tax$ (of $\mo$). 
\begin{example}
Considering the DL formula $\tax := \mathsf{Pet} \equiv \mathsf{Animal} \sqcap \exists \mathsf{hasOwner}.\mathsf{Person}$, we have that $\overline{\tax} = \setof{\equiv,\sqcap,\exists}$ and $\widetilde{\tax}=\setof{\mathsf{Pet, Animal, hasOwner, Person}}$.\qed
\end{example}
We now suppose that either a fault probability $p(e) := p(``e \mbox{ is faulty''})$ of each element $e\in\overline{\mo}\cup\widetilde{\mo}$ or the fault probability $p(\tax) := p(``\tax \mbox{ is faulty''})$ of each formula $\tax\in\mo$ is given. For estimation of these probabilities any (combination) of the methods mentioned above might be employed. In case formula probabilities are given, diagnosis probabilities can be directly computed by Formula~\ref{eq:diag_prob_calc}. Otherwise, the following pre-computations must be performed.

The fault probability $p(\tax)$ of $\tax$ can be calculated as the probability that at least one (occurrence of a) syntactical element in $\tax$ is faulty. So, $p(\tax)$ is equal to 1 minus the probability that none of the syntactical elements occurring in $\tax$ is faulty. Hence, under the assumption of mutual independence of syntactical faults concerning elements $e\in\overline{\tax}\cup\widetilde{\tax}$,
\begin{align}
p(\tax) = 1 - \prod_{e \in \overline{\tax} \cup \widetilde{\tax}} (1-p(e))^{n(e)} \label{eq:ax_prob_calc}
\end{align}
where $n(e)$ is the number of occurrences of syntactical element $e$ in $\tax$. 

If $p(\tax)$ for all $\tax \in \mo$ is known, the fault probability $p(\md)$ of any diagnosis $\md \in \Omega \subseteq \allD_{\langle\mo,\mb,\Tp,\Tn\rangle_\RQ}$ can be determined as the probability that each formula in $\md$ is faulty whereas each formula in $\mo \setminus \md$ is correct, i.e.\ not faulty. Thence, 
\begin{align}
p(\md) = \prod_{\tax_r \in \md} p(\tax_r) \prod_{\tax_s \in \mo\setminus\md} (1-p(\tax_s))  \label{eq:diag_prob_calc}
\end{align}
Recall that probabilities of all atomic events in a well-defined probability space must sum up to $1$. As not every subset of $\mo$ is a diagnosis, this is in general not the case. Therefore, diagnosis probabilities need to be normalized, i.e.\ each diagnosis probability $p(\md)$ must be divided by the sum of all diagnosis probabilities for diagnoses in $\Omega$. That is, the following adjustment is necessary:
\begin{align}
p(\md) \quad\gets\quad \frac{p(\md)}{\sum_{\md_k\in\Omega} p(\md_k)}  \label{eq:diag_prob_norm}
\end{align}
We want to emphasize that the probability measures $p(e)$ of syntactical elements $e$ and $p(\tax)$ of formulas $\tax$ are not required to satisfy any conditions except for $p(e) \in (0,1]$ and $p(\tax) \in (0,1]$ for all $e \in \overline{\tax} \cup \widetilde{\tax}$ and all $\tax \in \mo$ (see Remark~\ref{rem:ax_prob_not_zero} why the intervals $(0,1]$ are open). In particular, no normalization is needed. The reason for this is that ``$e$ is faulty'' and ``$\tax$ is faulty'' are assumptions about a \emph{single} logical connective and a \emph{single} logical formula, respectively. ``$\md$ is the true diagnosis'', to the contrary, is an assumption about \emph{each} formula in the KB $\mo$. So, the probabilities of two different syntactical elements $e_i\neq e_j$ 
are computed on the basis of two different probability spaces, namely 
$\Omega_{e_i} = \setof{``e_i \mbox{ is faulty''}, ``e_i \mbox{ is not faulty''}}$ and $\Omega_{e_j} = \setof{``e_j \mbox{ is faulty''}, ``e_j \mbox{ is not faulty''}}$ 
which clearly do not depend on each other at all. The same argumentation holds for probabilities of formulas.

\paragraph{More Reliable Probabilities through Observations.} 
As we argued before, the basic fault information from which diagnosis probabilities are deduced might be rather vague. A usual way of dealing with scenarios of that kind, is to regard the initial probabilities as a first (a-priori) estimation and to gather additional information, e.g.\ by making measurements or observations, and exploit this information to adapt the a-priori estimation in order to obtain a more reliable a-posteriori estimation. The more additional information has been accumulated and incorporated, the more realistic is the resulting updated estimation of probabilities. 

A well-known technique enabling computation of a-posteriori probabilities from a-priori probabilities is Bayes' Theorem. Let $p(\md)$ be the a-priori probability of some $\md\in\Omega \subseteq \allD_{\tuple{\mo,\mb,\Tp,\Tn}_\RQ}$ and $Obs$ be a new observation. Then, the a-posteriori probability $p(\md\,|\,Obs)$ of $\md$, i.e.\ the probability that the true diagnosis $\dt = \md$ taking into account the new information $Obs$, is computed according to Bayes' Theorem as
\begin{align}
p(\md\,|\,Obs) = \frac{p(Obs\,|\,\md)\;\,p(\md)}{p(Obs)} \label{eq:bayes}
\end{align}
where $p(Obs)$ is the (a-priori) probability that observation $Obs$ is made and $p(Obs\,|\,\md)$ is the (a-priori) probability that the observation $Obs$ is made under the assumption that $\md$ is the true diagnosis, i.e.\ $\dt = \md$. That is, the a-priori probability $p(\md)$, i.e.\ the probability that $\dt = \md$ without any additional knowledge, must be multiplied by $p(Obs\,|\,\md) / p(Obs)$ which is often referred to as the \emph{support $Obs$ provides for $\md$}. If the support is greater than 1, then the a-posteriori probability of $\md$ is greater than its a-priori probability, otherwise the a-posteriori probability gets smaller after incorporating the new information $Obs$. Note that Bayes' Theorem is only applicable to KB debugging if a suitable class of observations can be defined such that $p(Obs)$ and $p(Obs\,|\,\md)$ can be computed for observations $Obs$ of this class. As we shall see in Section~\ref{sec:UserInteraction}, the assignment of test cases to either $\Tp$ or $\Tn$ is one such class of observations. For instance, $t_i \in \Tp$ and $t_j \in \Tn$ for sets of formulas $t_i,t_j$ over $\mathcal{L}$ are two such observations.  

\subsection{Using Probabilities for Diagnosis Computation}
\label{sec:probs_diag_comp}

If available, formula fault probabilities can be exploited during construction of the pHS-tree (Algorithm~\ref{algo:hs}) in that most probable instead of minimum cardinality diagnoses are calculated first. To achieve that, breadth-first construction of the tree must be replaced by uniform-cost order of node expansion by means of the function $p()$ that assigns a fault probability to each formula $\tax\in\mo$. Thereby, the ``probability'' $p(\mathsf{nd})$ of a node 
$\mathsf{nd}= \setof{\tax_s,\dots, \tax_t}$ in Algorithm~\ref{algo:hs} is defined through $p(\tax), \tax\in\mo$ as 
\begin{align}\label{eq:path_prob_calc}
p(\mathsf{nd}) = \prod_{\tax_i\in\mathsf{nd}} p(\tax_i) \prod_{\tax_j \in \mo\setminus\mathsf{nd}} (1-p(\tax_j))
\end{align}
Notice that this formula extends the definition of Formula~\ref{eq:diag_prob_calc} to arbitrary subsets of $\mo$, not only diagnoses. Thus, Formula~\ref{eq:diag_prob_calc} is a special case of Formula~\ref{eq:path_prob_calc}.
 
First, note that we put ``probability'' of a node in quotation marks as, to be concise, each node (path) which is not yet a diagnosis, i.e.\ needs to be further expanded to become one, has probability zero (of being the true diagnosis $\dt$). For, a probability space is defined on a set of diagnoses and not on a set of arbitrary subsets $\mathsf{nd}$ of the KB. However, we misuse the diagnosis probability space in this case to determine the probability of ``pseudo-diagnoses'' in order to impose an order on the queue of open nodes in the tree. This will guarantee the finding of the most probable diagnoses first, as we shall see below (Proposition~\ref{prop:hs_prob_correct}).

Second, note that no normalization, i.e.\ application of Formula~(\ref{eq:diag_prob_norm}), is necessary within the scope of the non-interactive Algorithm~\ref{algo:hs} since the aim here is only the expansion of nodes $\mathsf{nd}$ in the order of $p(\mathsf{nd})$ and the return of the most probable identified diagnoses at a certain point in time. For this, the comparison of the probability of one node $\mathsf{nd}$ with the probability of another node $\mathsf{nd}'$ suffices.
Thus, no other calculations using the properties of a probability space are performed by Algorithm~\ref{algo:hs}. We shall recognize in Section~\ref{sec:WorkflowInInteractiveOntologyDebugging} that this will not hold for the interactive Algorithm~\ref{algo:inter_onto_debug} where Formula~(\ref{eq:diag_prob_norm}) is essential.

So, nodes $\mathsf{nd}$ are inserted into $\Queue$ in a way descending order of node probabilities in $\Queue$ is always maintained. Consequently, nodes with highest fault probability are processed first. This is practical since a user will usually be most interested in seeing those possible faults first that have the highest (estimated) probability to be the actual fault they seek.

However, one needs to be careful when using probabilities as weights in order not to lose the property of Algorithm~\ref{algo:hs} to compute \emph{minimal} diagnoses only. To this end,
the formula probabilities $p(\tax)$ for all $\tax\in\mo$ must be adapted as
\begin{align}\label{eq:adapt_ax_prob_to_get_min_diags}
p(\tax) \quad\gets\quad c\,\;\;p(\tax)
\end{align}
where the factor $c$ is an arbitrary positive real number smaller than $0.5$, e.g.\ $c := 0.49 / \max_{\setof{\tax\in\mo}}(p(\tax))$. This transformation effects that all probabilities $p(\tax)$ become smaller than $50\%$. In other words, each formula must be more likely to be correct than faulty which in turn means that a minimal diagnosis is more likely than any of its supersets.
\begin{definition}\label{def:p_node()}
Let $p:\mo \rightarrow [0,1]$ be some function that assigns to each $\tax\in\mo$ some $p(\tax) \in [0,1]$. Then, we denote by $p_{nodes}: 2^{\mo} \rightarrow [0,1]$ the function that assigns to each node $\mathsf{nd} \subseteq \mo$ some $p_{nodes}(\mathsf{nd})\in [0,1]$ which is obtained by means of Formula~\ref{eq:path_prob_calc} and $p()$. 
\end{definition}
\begin{lemma}\label{lem:superset_lower_prob}
Let $\mathsf{nd},\mathsf{nd}'\subseteq\mo$ where $\mathsf{nd}\subset \mathsf{nd}'$ and $p: \mo \rightarrow (0,0.5)$ a function which assigns to each $\tax\in\mo$ some probability $p(\tax) \in (0,0.5)$. 
Then $p_{nodes}(\mathsf{nd}) > p_{nodes}(\mathsf{nd}')$ holds. 
\end{lemma}
\begin{proof}
According to Formula~\ref{eq:path_prob_calc} and Definition~\ref{def:p_node()} we have that 
\begin{align*}
p_{nodes}(\mathsf{nd}) = \prod_{\tax_i\in \mathsf{nd}} p(\tax_i) \prod_{\tax_j \in \mo \setminus \mathsf{nd}} (1-p(\tax_j))
\end{align*}
Then the probability $p_{nodes}(\mathsf{nd}')$ can be computed from $p_{nodes}(\mathsf{nd})$ in that, for each formula $\tax$ in $\mathsf{nd}'\setminus \mathsf{nd} \subseteq \mo \setminus \mathsf{nd}$, we multiply $p_{nodes}(\mathsf{nd})$ by a factor $f_\tax := p(\tax) / (1-p(\tax))$ because $\tax$ ``moves'' from $\mo\setminus \mathsf{nd}$ to $\mathsf{nd}$. However, $f_\tax < 1$ holds due to $p(\tax) < 0.5$ and thus $1-p(\tax) > 0.5$.
\end{proof}
This result will be a key to proving the completeness, soundness and correctness of Algorithm~\ref{algo:hs} in the next section.

The next definition characterizes a (partial) weighted pHS-tree, the type of hitting set tree constructed by Algorithm~\ref{algo:hs} given any function $p(\tax) \in (0,0.5)$ for all $\tax\in\mo$ as input which is not necessarily specified in a way a breadth-first tree construction is forced.
\begin{definition}[Weighted Pruned HS-Tree]\label{def:weighted_pruned_hs_tree}
Let $\langle\mo,\mb,\Tp,\Tn\rangle_\RQ$ be an admissible DPI and let $w:\mo\rightarrow [0,1]$ be a weight function which assigns a weight to each node $\mathsf{n} \subseteq \mo$ with the property that $w(\mathsf{n}_1) > w(\mathsf{n}_2)$ if $\mathsf{n}_1 \subset \mathsf{n}_2$. An edge-labeled and node-labeled tree $T$ is called a \emph{weighted pruned HS-tree (wpHS-tree) w.r.t.\ $\langle\mo,\mb,\Tp,\Tn\rangle_\RQ$ and $w()$} iff $T$ is the result of constructing an HS-tree w.r.t.\ $\langle\mo,\mb,\Tp,\Tn\rangle_\RQ$ with due regard to the following rule  
\begin{enumerate}
\item Label open nodes in the HS-tree in order of descending $w()$,
\end{enumerate}
and the rules 2 to 6 as per Definition~\ref{def:pruned_hs_tree}. 

$T$ is called a \emph{partial weighted pruned HS-tree w.r.t.\ $\langle\mo,\mb,\Tp,\Tn\rangle_\RQ$ and $w()$} iff $T$ is a weighted pruned HS-tree w.r.t.\ $\langle\mo,\mb,\Tp,\Tn\rangle_\RQ$ and $w()$ where not all nodes in $T$ have been labeled yet and non-labeled nodes have no successors. 
\end{definition}
Then, we have the following relationship between a (partial) pHS-tree
and a (partial) wpHS-tree. An explanation why this holds will be given in Section~\ref{sec:UsingProbabilitiesToComputeMinimumCardinalityDiagnoses}.
\begin{proposition}
A (partial) pHS-tree w.r.t.\ $\langle\mo,\mb,\Tp,\Tn\rangle_\RQ$ is a (partial) wpHS-tree w.r.t.\ $\langle\mo,\mb,\Tp,\Tn\rangle_\RQ$ and $w()$ where $w()$ is a weight function which, additionally to the property postulated in Definition~\ref{def:weighted_pruned_hs_tree}, satisfies $w(\mathsf{n}_1) = w(\mathsf{n}_2)$ if $|\mathsf{n}_1| = |\mathsf{n}_2|$.

In general, a (partial) wpHS-tree w.r.t.\ $\langle\mo,\mb,\Tp,\Tn\rangle_\RQ$ and $w()$ is not a (partial) pHS-tree w.r.t.\ $\langle\mo,\mb,\Tp,\Tn\rangle_\RQ$.
\end{proposition}

\begin{lemma}\label{lem:algo_hs_produces_weighted_pruned_hs_tree} Algorithm~\ref{algo:hs} 
is a procedure for producing a wpHS-tree $T$ w.r.t.\ $\langle\mo,\mb,\Tp,\Tn\rangle_\RQ$ and $p_{nodes}()$. 
%
\end{lemma}
\begin{proof}
First, the property $p_{nodes}(\mathsf{n}_1) > p_{nodes}(\mathsf{n}_2)$ if $\mathsf{n}_1 \subset \mathsf{n}_2$ postulated by Definition~\ref{def:weighted_pruned_hs_tree} holds by Lemma~\ref{lem:superset_lower_prob} and the fact that the function $p$ given as input to Algorithm~\ref{algo:hs} satisfies $p(\tax)\in(0,0.5)$ for all $\tax\in\mo$. Moreover, the DPI $\langle\mo,\mb,\Tp,\Tn\rangle_\RQ$ provided as an input to Algorithm~\ref{algo:hs} is admissible, as postulated by Definition~\ref{def:weighted_pruned_hs_tree}.

The compliance with rule 1 of Definition~\ref{def:hs_tree} as well as with rules 2 to 6 of Definition~\ref{def:pruned_hs_tree} is a simple consequence of Lemma~\ref{lem:algo_hs_produces_pruned_hs_tree}. In the following we prove that rule 2 of Definition~\ref{def:hs_tree} and rule 1 of Definition~\ref{def:weighted_pruned_hs_tree} are satisfied.
\begin{itemize}
	\item Definition~\ref{def:hs_tree}, rule 2: Suppose a node $\mathsf{nd}$ is labeled by $valid$. Then it is added to $\mD_{calc}$ in line~\ref{algoline:hs:update_Dcalc}. Since $\mathsf{nd}$ can only get a label different from $closed$ if it is the only exemplar of this node in $\Queue$ due to the duplicate criterion (lines~\ref{algoline:hs:duplicate_crit_start}-\ref{algoline:hs:duplicate_crit_end}), it must be the case that $\mathsf{nd}\notin\Queue$ (line~\ref{algoline:hs:remove_from_queue}) after $\mathsf{nd}$ has been labeled by $valid$. Only nodes that get labeled by a conflict set can have successor nodes added to $\Queue$ in line~\ref{algoline:hs:generate_nodes}. Only nodes in $\Queue$ can get a label (cf.\ lines~\ref{algoline:hs:getfirst} and \ref{algoline:hs:label}). For $\mathsf{nd}$ to be added to $\Queue$ at some later point in time there must be a proper subset of $\mathsf{nd}$ that is still in $\Queue$ as each node newly added to $\Queue$ is a proper superset of some node in $\Queue$ (cf.\ line~\ref{algoline:hs:generate_nodes} which is the only position in the algorithm where nodes are added to $\Queue$). This is impossible since $\Queue$ is ordered descending by $p_{nodes}()$. 
Hence, each proper subset of $\mathsf{nd}$ must have been ranked before $\mathsf{nd}$ in $\Queue$ and thus must have already been labeled because
$\mathsf{nd}$ is already labeled by assumption.
Hence, if $\mathsf{nd}$ is labeled by $valid$, then it has no successors.
	\item Definition~\ref{def:weighted_pruned_hs_tree}, rule 1: 
That nodes are processed and labeled in order of descending $p_{nodes}()$ follows from the fact that new nodes are inserted into $\Queue$ only in a way that the order of $\Queue$ by descending $p_{nodes}()$ is maintained (\textsc{insertSorted} in line~\ref{algoline:hs:generate_nodes}) and by the fact that always the first element of $\Queue$ is selected to be labeled next (\textsc{getFirst} in line~\ref{algoline:hs:getfirst}).
\end{itemize}
This completes the proof.
\end{proof}
Let the relevant data of a wpHS-tree be defined as for a pHS-tree (cf.\ Remark~\ref{rem:algo_hs:internal_representation}). By the correctness of Lemma~\ref{lem:algo_hs_produces_weighted_pruned_hs_tree}, we have:
\begin{corollary}\label{cor:algo_hs_returns_relevant_data_of_weighted_(partial)_pruned_hs_tree}
Algorithm~\ref{algo:hs} 
stores by $\tuple{\mD_{calc},\Queue,\mC_{calc}}$ the relevant data of 
\begin{itemize}
\item a wpHS-tree w.r.t.\ $\langle\mo,\mb,\Tp,\Tn\rangle_\RQ$ and $p_{nodes}()$ if Algorithm~\ref{algo:hs} stops due to $\Queue = []$, and
\item a partial wpHS-tree w.r.t.\ $\langle\mo,\mb,\Tp,\Tn\rangle_\RQ$ and $p_{nodes}()$ otherwise.
\end{itemize}
\end{corollary}

\subsection{Correctness of Weighted Diagnosis Computation}
\label{sec:CorrectnessOfAlgorithmHs}
First, we show the completeness of Algorithm~\ref{algo:hs} regarding minimal diagnoses, i.e.\ that it computes \emph{all} minimal diagnoses w.r.t.\ the DPI it is given as input.
\begin{lemma}\label{lem:algo_hs_only_diags_in_Dcalc}
Only diagnoses w.r.t.\ $\tuple{\mo,\mb,\Tp,\Tn}_\RQ$ can be added to $\mD_{calc}$ by Algorithm~\ref{algo:hs}.
\end{lemma}
\begin{proof}
A node $\mathsf{nd}$ can be added to $\mD_{calc}$ only in line~\ref{algoline:hs:update_Dcalc}. To reach this line, \textsc{label} must have returned $valid$ for $\mathsf{nd}$. For this to hold, $\scQX(\tuple{\mo\setminus\mathsf{nd},\mb,\Tp,\Tn}_\RQ)$ must have returned 'no conflict' which implies that $\mathsf{nd}$ is a diagnosis w.r.t.\ $\tuple{\mo,\mb,\Tp,\Tn}_\RQ$ by Propositions~\ref{prop:qx_correctness} and \ref{prop:validonto_diag}.
\end{proof}
\begin{lemma}\label{lem:algo_hs_path+cs}
Let $T$ denote a (partial) wpHS-tree produced by Algorithm~\ref{algo:hs}. Further, let $\Queue$ be the queue of open nodes in $T$ maintained by Algorithm~\ref{algo:hs} and let $\mathsf{nd}$ be some node which occurs only once in $\Queue$ and which is a proper subset of some minimal diagnosis w.r.t.\ $\tuple{\mo,\mb,\Tp,\Tn}_\RQ$. Then:
\begin{enumerate}[(1)]
\item The nodes $\emptyset=\mathsf{nd}_1,\dots,\mathsf{nd}_k$ along any path from the root node $\emptyset$ to $\mathsf{nd}_k$ in $T$ satisfy $\mathsf{nd}_i \subset \mathsf{nd}_{i+1}$ and $|\mathsf{nd}_i|+1 = |\mathsf{nd}_{i+1}|$ and $\mathsf{nd}_i \subseteq \mo$ for $1 \leq i \leq k$. 
\item  If the \textsc{label} function is called for $\mathsf{nd}$, then it yields some minimal conflict set $\mc$ w.r.t.\ $\tuple{\mo,\mb,\Tp,\Tn}_\RQ$ with $\mathsf{nd} \cap \mc = \emptyset$.
\end{enumerate}
\end{lemma}
\begin{proof}
(1): In the representation used by Algorithm~\ref{algo:hs}, a node $\mathsf{nd}$ in the (partial) wpHS-tree $T$ produced by Algorithm~\ref{algo:hs} is defined as the set of all edge labels on the path from the root node 
to $\mathsf{nd}$ (see Remark~\ref{rem:algo_hs:internal_representation}) and the successor of a node is defined as a node added to $\Queue$ after $\mathsf{nd}$ has been labeled by a minimal conflict set.
After the \textsc{label} function for node $\mathsf{nd}$ has returned some minimal conflict set $L$ as a label for $\mathsf{nd}$, Algorithm~\ref{algo:hs} goes to line~\ref{algoline:hs:generate_nodes} since $L\neq closed$ and $L\neq valid$ and adds an element $\mathsf{nd} \cup \setof{e}$ to $\Queue$ for each $e \in L$. Therefore, it holds that $|\mathsf{nd} \cup \setof{e}| = |\mathsf{nd}|+1$ for each successor of $\mathsf{nd}$. Hence,  $\mathsf{nd}_i \subset \mathsf{nd}_{i+1}$ and $|\mathsf{nd}_i|+1 = |\mathsf{nd}_{i+1}|$ holds for any path of nodes $\emptyset=\mathsf{nd}_1,\dots,\mathsf{nd}_k$ in $T$ starting from the root node.

The argumentation why each node must be a subset of $\mo$ is as follows:
Suppose $\mathsf{node}\cup\setof{e}$ is added to $\Queue$ in line~\ref{algoline:hs:generate_nodes} which is the only place in Algorithm~\ref{algo:hs} where nodes are added to $\Queue$. So, \textsc{label} must have returned neither $valid$ nor $closed$ for $\mathsf{node}$. Hence, $\mathsf{node}$ cannot be a diagnosis w.r.t.\ $\langle\mo,\mb,\Tp,\Tn\rangle_\RQ$ as otherwise \textsc{label} with argument $\mathsf{node}$ must have returned $valid$ in line~\ref{algoline:hs:return_valid}. Due to the fact that $\mathsf{node} = \mo$ is definitely a diagnosis w.r.t.\ $\langle\mo,\mb,\Tp,\Tn\rangle_\RQ$ as it must hit all minimal conflict sets w.r.t.\ $\langle\mo,\mb,\Tp,\Tn\rangle_\RQ$ which must all be subsets of $\mo$ (Definition~\ref{def:cs}), $\mathsf{node} \subset \mo$ must hold. 

(2): Suppose the \textsc{label} function is called for a node $\mathsf{nd}\in\Queue$ where $\mathsf{nd} \subset \md$ for some minimal diagnosis $\md$. 

First, there cannot be any $\mathsf{nd}' \in \mD_{calc}$ with $\mathsf{nd}' \subseteq \mathsf{nd}$ since $\mD_{calc}$ includes only diagnoses w.r.t.\ $\langle\mo,\mb,\Tp,\Tn\rangle_\RQ$ and $\mathsf{nd} \subset \md$ wherefore there would be a diagnosis $\mathsf{nd}' \subset \md$, contradiction. Due to the fact that $\mathsf{nd}$ is present only once in $\Queue$, there cannot be some $\mathsf{nd}' = \mathsf{nd}$ in $\Queue$. Thus, $closed$ cannot be returned for $\mathsf{nd}$ by \textsc{label}.

By the facts that a diagnosis must hit all minimal conflict sets (Proposition~\ref{prop:mindiag_mincs}) and that $\mathsf{nd}$ is a proper subset of a diagnosis, either the criterion checked in line~\ref{algoline:hs:test_cs_not_hit} must be true or $\scQX(\tuple{\mo\setminus\mathsf{nd},\mb,\Tp,\Tn}_\RQ)$ must return a minimal conflict set $L$, i.e.\ $L \neq$ 'no conflict'. In both cases, a minimal conflict set is returned by \textsc{label}.

There are no other labels that can be returned by \textsc{label}.
\end{proof}

\begin{lemma}\label{lem:each_min_diag_occurs_in_queue}
Each minimal diagnosis w.r.t.\ 
$\langle\mo,\mb,\Tp,\Tn\rangle_\RQ$ 
occurs as a node in $\Queue$ during the execution of Algorithm~\ref{algo:hs}, if the execution stops due to $\Queue = []$.
\end{lemma}
\begin{proof}
For Algorithm~\ref{algo:hs} it holds that 
\begin{enumerate}[(i)]
\item if $\mathsf{nd}$ is the last exemplar of some node in $\Queue$ which is a proper subset of some minimal diagnosis w.r.t.\ $\tuple{\mo,\mb,\Tp,\Tn}_\RQ$ and the \textsc{label} function is called for $\mathsf{nd}$, then it yields some minimal conflict set $\mc$ w.r.t.\ $\tuple{\mo,\mb,\Tp,\Tn}_\RQ$ with $\mathsf{nd} \cap \mc = \emptyset$ by Lemma~\ref{lem:algo_hs_path+cs} and 
\item each node $\mathsf{nd}$ that has been labeled by some minimal conflict set $\mc$ is deleted from $\Queue$ (line~\ref{algoline:dyn:delete_from_queue}) whereupon one successor node $\mathsf{nd}_{\tax} = \mathsf{nd} \cup \setof{\tax}$ for each element $\tax\in\mc$ is added to $\Queue$ (\textsc{insertSorted} in line~\ref{algoline:dyn:generate_nodes}) and
\item each minimal diagnosis w.r.t.\ $\tuple{\mo,\mb,\Tp,\Tn}_\RQ$ is a superset of $\emptyset$ and a subset of $\mo$ (Definition~\ref{def:diagnosis}) which includes one element of each minimal conflict set w.r.t.\ $\tuple{\mo,\mb,\Tp,\Tn}_\RQ$ and includes only elements of minimal conflict sets (Proposition~\ref{prop:mindiag_mincs}).
\end{enumerate}

Let $\md$ be some minimal diagnosis w.r.t.\ $\langle\mo,\mb,\Tp,\Tn\rangle_\RQ$. Then, there is a path of nodes from the root node $\emptyset$ to $\md$ in the pHS-tree produced by Algorithm~\ref{algo:hs}, if the execution stops due to $\Queue = []$. 

This holds by the following argumentation: If $\md = \emptyset$, then the path is $\tuple{\emptyset}$. Now, suppose $\md \supset \emptyset$. Since $\md$ is a minimal diagnosis wherefore no other diagnosis can be equal to $\emptyset$, the root node $\mathsf{n}_0 := \emptyset$ of the constructed tree must be labeled by some minimal conflict set $\mc_1$.  
Then, by (iii), there must be some $\tax_1\in\mc_1$ that is an element of $\md$. So, we define $\mathsf{n}_1 := \setof{\tax_1}$. If $\mathsf{n}_1 = \md$, then the path is $\tuple{\emptyset,\mathsf{n}_1}$. Otherwise, due to $\md \supset \mathsf{n}_1$ and (i), 
node $\mathsf{n}_1$ in the pHS-tree must be labeled by some minimal conflict set $\mc_2$. Then, by (iii), there must be some $\tax_2\in\mc_2$ that is an element of $\md$. So, we define $\mathsf{n}_2 := \mathsf{n}_1 \cup \setof{\tax_2}$. If $\mathsf{n}_2 = \md$, then the path is $\tuple{\emptyset,\mathsf{n}_1,\mathsf{n}_2}$. Otherwise, due to $\md \supset \mathsf{n}_2$ and (i), 
node $\mathsf{n}_2$ in the pHS-tree must be labeled by some minimal conflict set $\mc_3$. This reasoning can be continued until $\mathsf{n}_k = \md$ for some $k$. By (iii), $\md\subseteq\mo$ holds wherefore such $k$ must exist. 

Algorithm~\ref{algo:hs} cannot stop executing before $\mathsf{n}_k$ has been in $\Queue$ since each node $\mathsf{n}_i$ labeled by a minimal conflict set $\mc_{i+1}$ involves the addition of $|\mc_{i+1}|$ successor nodes to $\Queue$ by (ii). In particular, the successor node $\mathsf{n}_i \cup \setof{\tax_{i+1}}$ must be added to $\Queue$. As the execution stops due to $\Queue = []$, all nodes $\mathsf{n}_i$ for $i\leq k$ must be labeled before termination. Thus, $\md$ must be in $\Queue$ sometime. 
%
%
%
%
%
\end{proof}
\begin{proposition}[Completeness of Algorithm~\ref{algo:hs}]\label{prop:hs-tree_completeness}
If Algorithm~\ref{algo:hs} terminates due to $\Queue = []$,
then the algorithm returns a set $\mD$ including all minimal diagnoses w.r.t.\ 
$\langle\mo,\mb,\Tp,\Tn\rangle_\RQ$. 
\end{proposition}
\begin{proof}
Assume some minimal diagnosis $\md$ w.r.t.\ $\langle\mo,\mb,\Tp,\Tn\rangle_\RQ$ where $\md\notin\mD$ after Algorithm~\ref{algo:hs} has returned due to $\Queue = []$. First, each minimal diagnosis will occur in $\Queue$ throughout the execution of Algorithm~\ref{algo:hs} because it executes until $\Queue = []$ wherefore Lemma~\ref{lem:each_min_diag_occurs_in_queue} applies. 
Any node $\mathsf{nd}$ in $\Queue$ can only be deleted from $\Queue$ if \textsc{label} is called with the argument node $\mathsf{nd}$ (lines~\ref{algoline:hs:remove_from_queue} and \ref{algoline:hs:label}). There is no other point in Algorithm~\ref{algo:hs} where elements are removed from $\Queue$.
Since at the end $\Queue = []$, each minimal diagnosis, in particular $\md$, must be labeled.

Suppose $\md$ is the last exemplar of possibly multiple duplicates of it in $\Queue$. Then, the \textsc{label} function cannot return $closed$ for $\md$. 
This holds, on the one hand, because the duplicate criterion (lines~\ref{algoline:hs:duplicate_crit_start}-\ref{algoline:hs:duplicate_crit_end}) only removes possible duplicate nodes from $\Queue$, but never the last exemplar of a node in $\Queue$. On the other hand, $\md$ can never be closed due to the non-minimality criterion (lines~\ref{algoline:hs:non_min_crit_start}-\ref{algoline:hs:non_min_crit_end}) as $\mD_{calc}$ can only include diagnoses w.r.t.\ $\langle\mo,\mb,\Tp,\Tn\rangle_\RQ$ by Proposition~\ref{lem:algo_hs_only_diags_in_Dcalc}. Thus, due to the minimality of $\md$, $\mD_{calc}$ cannot comprise any diagnosis $\md'$ with $\md' \subseteq \md$, except for some $\md'$ which is equal to $\md$. This would however be a contradiction to the assumption that $\md\notin\mD$.

The reuse criterion (lines~\ref{algoline:hs:reuse_crit_start}-\ref{algoline:hs:reuse_crit_end}) cannot apply for $\md$ either since a minimal diagnosis is a hitting set of all minimal conflict sets (Proposition~\ref{prop:mindiag_mincs}) wherefore there cannot be a minimal conflict set in $\mC_{calc}$ which has an empty intersection with $\md$. So, the algorithm will come to line~\ref{algoline:hs:qx} where $\scQX(\tuple{\mo\setminus\md,\mb,\Tp,\Tn}_\RQ)$ will return 'no conflict' (Propositions~\ref{prop:qx_correctness} and \ref{prop:validonto_diag}). Therefore, $\md$ will be labeled by $valid$ and will be added to $\mD_{calc}$ in line~\ref{algoline:hs:update_Dcalc}.
\end{proof}
Next, we show the soundness of Algorithm~\ref{algo:hs} w.r.t.\ minimal diagnoses, i.e.\ that it computes \emph{only} minimal diagnoses w.r.t.\ the DPI it is given as input.
\begin{proposition}[Soundness of Algorithm~\ref{algo:hs}]\label{prop:hs-tree_soundness}
If 
an element $\md$ is added to the set $\mD_{calc}$ during the execution of Algorithm~\ref{algo:hs}, $\md$ is a minimal diagnosis w.r.t.\ 
$\langle\mo,\mb,\Tp,\Tn\rangle_\RQ$. 
\end{proposition}
\begin{proof}
Assume that some element $\mathsf{nd}$ is added to $\mD_{calc}$ which is not a diagnosis w.r.t.\ $\langle\mo,\mb,\Tp,\Tn\rangle_\RQ$. This immediately yields a contradiction due to Lemma~\ref{lem:algo_hs_only_diags_in_Dcalc}.
%

Assume now that some element $\mathsf{nd}$ is added to $\mD_{calc}$ which is a diagnosis w.r.t.\ $\langle\mo,\mb,\Tp,\Tn\rangle_\RQ$, but not a minimal one. Now, since $\mathsf{nd}$ is a non-minimal diagnosis, there is some $\md \subset \mathsf{nd}$ which is a minimal diagnosis w.r.t.\ $\langle\mo,\mb,\Tp,\Tn\rangle_\RQ$.

Then, there are three cases to distinguish: (a)~$\md$ is in $\Queue$ and (b)~$\md$ is in $\mD_{calc}$ and (c)~$\md$ is neither in $\Queue$ nor in $\mD_{calc}$, i.e.\ the node $\md$ has not yet been generated.

Note that these are all possible cases as $\md$ is a \emph{minimal} diagnosis by assumption. So, $\md$ cannot have been ruled out, i.e.\ labeled by $closed$, by the non-minimality criterion (lines~\ref{algoline:hs:non_min_crit_start}-\ref{algoline:hs:non_min_crit_end}) before since only diagnoses can be added to $\mD_{calc}$ as argued in the first paragraph of this proof and there cannot be a diagnosis $\md'\in\mD_{calc}$ such that $\md' \subset \md$. The case $\md' = \md$ is already considered by case (b). The duplicate criterion (lines~\ref{algoline:hs:duplicate_crit_start}-\ref{algoline:hs:duplicate_crit_end}) does not need to be taken into account since it deletes duplicate nodes only. 

(a): To be added to $\mD_{calc}$, $\mathsf{nd}$ must have been the first element of the queue $\Queue$ by \textsc{getFirst} in line~\ref{algoline:hs:getfirst}. Since $\md \in \Queue$ by assumption and since $\Queue$ is sorted in descending order of node probability (\textsc{insertSorted} in line~\ref{algoline:hs:generate_nodes}), we conclude that $p_{nodes}(\md) \leq p_{nodes}(\mathsf{nd})$. However, as $p_{nodes}(X)$ for a node $X\subseteq \mo$ is defined by means of $p(\tax)$ where $p(\tax) \in (0,0.5)$ for all $\tax\in\mo$ as per Formula~\ref{eq:path_prob_calc} (Definition~\ref{def:p_node()}), Lemma~\ref{lem:superset_lower_prob} applies and establishes the truth of $p_{nodes}(S_1) > p_{nodes}(S_2)$ if $S_1 \subset S_2$ for $S_1,S_2 \subseteq \mo$. By $\md \subset \mathsf{nd}$, this implies $p_{nodes}(\md) > p_{nodes}(\mathsf{nd})$, contradiction.
%

(b): Assuming case (b), we can derive a contradiction as follows. By the fact that $\mathsf{nd}$ is added to $\mD_{calc}$, it must hold that the \textsc{label} procedure called for $\mathsf{nd}$ in line~\ref{algoline:hs:label} returned $valid$ as part of its output in line~\ref{algoline:hs:return_valid}. However, as $\md \subset \mathsf{nd}$ is already an element of $\mD_{calc}$ by assumption, the \textsc{label} procedure must have already returned in line~\ref{algoline:hs:non_min_crit_end} wherefore it cannot have reached line~\ref{algoline:hs:return_valid}, contradiction.

(c): Suppose that $\md$ has not yet been generated as a node in $\Queue$. 
By Lemma~\ref{lem:algo_hs_path+cs}, the nodes $\emptyset=\mathsf{nd_1},\dots,\mathsf{nd_k}$ along a path from the root node in the pHS-Tree produced by Algorithm~\ref{algo:hs} 
satisfy $\mathsf{nd}_i \subset \mathsf{nd}_{i+1}$ and $|\mathsf{nd}_i|+1 = |\mathsf{nd}_{i+1}|$.
So, by Lemma~\ref{lem:superset_lower_prob}, the node probabilities along any path from the root node are strictly monotonically decreasing.  Since $p_{nodes}(\md) > p_{nodes}(\mathsf{nd})$ holds by the same argumentation as in (a), we have that all nodes on the path from the root node to $\md$ have a higher probability than $\mathsf{nd}$. As $\Queue$ is sorted in descending order of node probability and in each iteration the first element in $\Queue$ is processed as explained in (a), we infer that $\md$ must have already been generated at the time $\mathsf{nd}$ is processed, contradiction. 
%
%
%
\end{proof}
Next, we argue that Algorithm~\ref{algo:hs} computes minimal diagnoses in descending order of diagnosis probability according to the parameter $p()$ given as input to the algorithm.  
\begin{corollary}\label{cor:hs_tree_finds_most-prob_diags_first}
Let the probability $p(\md)$ of a diagnosis $\md$ in Algorithm~\ref{algo:hs} be computed from the given function $p(\tax),\tax\in\mo$ as per Formula~\ref{eq:diag_prob_calc}.
\begin{enumerate}
	\item At any point in time during the execution of Algorithm~\ref{algo:hs}, $\mD_{calc}$ comprises the $|\mD_{calc}|$ most probable minimal diagnoses w.r.t.\ 
	$\langle\mo,\mb,\Tp,\Tn\rangle_\RQ$.
	\item 
If Algorithm~\ref{algo:hs} 
returns a set $\mD$ of cardinality $n$, then $\mD$ is the set of the $n$ most-probable minimal diagnoses w.r.t.\ $\langle\mo,\mb,\Tp,\Tn\rangle_\RQ$.
%
\end{enumerate}
\end{corollary}
\begin{proof}
(1): By Propositions~\ref{prop:hs-tree_completeness} and \ref{prop:hs-tree_soundness}, it is a fact that Algorithm~\ref{algo:hs} 
computes all and only minimal diagnoses w.r.t.\ $\langle\mo,\mb,\Tp,\Tn\rangle_\RQ$. 
What must still be shown is that minimal diagnoses are added to $\mD_{calc}$ in descending order of their probability $p()$ as per Formula~\ref{eq:diag_prob_calc}. 
The probability $p(\md)$ of some diagnosis $\md$ is equal to $p_{nodes}(\md)$ since a each diagnosis is a node and Formula~\ref{eq:diag_prob_calc} is a special case of Formula~\ref{eq:path_prob_calc} by which the probability $p_{nodes}(\mathsf{nd})$ of a node $\mathsf{nd}$ is calculated.

Let us denote by $\md_{pmax}$ the minimal diagnosis with maximum probability that has not yet been added to $\mD_{calc}$ and by $\md_{\lnot pmax}$ an arbitrary minimal diagnosis with non-maximal probability. That is, $p_{nodes}(\md_{\lnot pmax}) < p_{nodes}(\md_{pmax})$. So, we need to demonstrate that each node $\mathsf{nd} \subset \md_{pmax}$ on a path from the root node to node $\md_{pmax}$ is processed before $\md_{\lnot pmax}$ is treated. By Lemma~\ref{lem:algo_hs_path+cs}, a path from the root node in the pHS-Tree produced by Algorithm~\ref{algo:hs} is a set of nodes $\emptyset=\mathsf{nd_1},\dots,\mathsf{nd_k}$ where $\mathsf{nd}_i \subset \mathsf{nd}_{i+1}$ and $|\mathsf{nd}_i|+1 = |\mathsf{nd}_{i+1}|$. Further recall that the probability $p_{nodes}(X)$ of a node $X\subseteq \mo$ in Algorithm~\ref{algo:hs} is defined as per Formula~\ref{eq:path_prob_calc}. So, by Lemma~\ref{lem:superset_lower_prob}, the node probabilities along any path from the root node are strictly monotonically decreasing. Hence, each node $\mathsf{nd}$ on a path from the root node to $\md_{pmax}$ has a probability $p_{nodes}(\mathsf{nd}) > p_{nodes}(\md_{pmax}) > p_{nodes}(\md_{\lnot pmax})$. By the insertion of new nodes into $\Queue$ (\textsc{insertSorted} in line~\ref{algoline:hs:generate_nodes}) in a way descending order of $\Queue$ as per $p_{nodes}()$ is always maintained, and by the selection of the first element of $\Queue$ (\textsc{getFirst} in line~\ref{algoline:hs:getfirst}) as next node to be processed, each node $\mathsf{nd}$ on a path to $\md_{pmax}$ must be processed before $\md_{\lnot pmax}$ is processed. Consequently, minimal diagnoses are added to $\mD_{calc}$ in descending order of their probability $p()$ as per Formula~\ref{eq:diag_prob_calc}. 

(2): This proposition follows directly from (1).
\end{proof}
\begin{proposition}\label{prop:hs_prob_correct}
Algorithm~\ref{algo:hs} 
always terminates and returns a set $\mD$ of minimal diagnoses w.r.t.\ $\langle\mo$, $\mb,\Tp$, $\Tn\rangle_\RQ$ which is 
\begin{itemize}
\item the set of the $|\mD|$ most probable (w.r.t.\ $p()$ and Formula~\ref{eq:diag_prob_calc}) minimal diagnoses w.r.t.\ $\langle\mo,\mb,\Tp,\Tn\rangle_\RQ$ such that $n_{\min} \leq |\mD| \leq n_{\max}$, if at least $n_{\min}$ minimal diagnoses exist w.r.t.\ $\langle\mo,\mb,\Tp,\Tn\rangle_\RQ$, or 
\item the set of all minimal diagnoses w.r.t.\ $\langle\mo,\mb,\Tp,\Tn\rangle_\RQ$, otherwise.
\end{itemize}
\end{proposition}
\begin{proof}
The proposition is a direct consequence of Propositions~\ref{prop:hs:termination}, \ref{prop:hs-tree_completeness} and \ref{prop:hs-tree_soundness} and Corollary~\ref{cor:hs_tree_finds_most-prob_diags_first}.
\end{proof}

\subsection{Using Probabilities to Compute Minimum Cardinality Diagnoses}
\label{sec:UsingProbabilitiesToComputeMinimumCardinalityDiagnoses}

The function $p:\mo\rightarrow (0,0.5)$ can be defined in a way 
that minimum cardinality instead of maximum probability diagnoses 
are identified first. To this end, $p()$ is specified as a fixpoint function that maps each formula $\tax\in\mo$ to \emph{one and the same} constant value $p(\tax) := c$ where $c$ is an arbitrary real number such that $0 < c < 0.5$, e.g.\ $c := 0.3$. That in this setting diagnoses are found in order of ascending cardinality is a simple consequence of Corollary~\ref{cor:hs_tree_finds_most-prob_diags_first}.

\begin{example}\label{example:ax_prob_calc}
Let us now study how such formula and diagnosis probabilities would be constructed for the example DPI depicted by Table~\ref{tab:example2}. Let us suppose that the KB $\mo$ in the DPI was formulated by a single user $u$ for whom the personal fault probabilities of syntactical elements $\widetilde{\mo}\cup\overline{\mo}$ given by the first row of Table~\ref{tab:example:ax_prob} have been extracted from log data of the KB editing software applied by $u$. Then, the resulting probabilities of formulas $\tax\in\mo$ as per Formula~\ref{eq:ax_prob_calc} are as presented in the rightmost column of Table~\ref{tab:example:ax_prob}. The entries in the table from the second to the last but two column display the number of occurrences of the syntactical element given by the column label in the formula given by the row label. These values are required to compute the formula probabilities listed in the last but one column as per Formula~\ref{eq:ax_prob_calc}. The final probabilities that can ``safely'' be incorporated into Algorithm~\ref{algo:hs} under a guarantee that only minimal diagnoses will be output are shown in the last column. These result from an application of Formula~\ref{eq:adapt_ax_prob_to_get_min_diags} to the probabilities given in the last but one column with an adaptation parameter $c := 0.49$.

Notice that, for example, $p(\tax_5)$ is rather high since the predicates $A$ and $Y$ as well as the connective $\lnot$ occurring in $\tax_5$ have a comparably high fault probability in relation to syntactical elements appearing in other formulas. Formula $\tax_3$, on the other hand, comprises only two predicates which should be well-understood by $u$ and no connectives except for $\rightarrow$ which is not problematic for $u$ either. Therefore, its fault probability is rather low.\qed 
\end{example}

\begin{table*}
\small
\centering
\setlength\tabcolsep{2pt}
\begin{tabular}{c @{\kern10pt } c c c c c c c c @{\kern15pt } c c c c @{\kern15pt} c @{\kern12pt} c}
\toprule
fault prob.   & 0.25 & 0.01 & 0.03 & 0.05 & 0.4 & 0.1 & 0.6 & 0.6 & 0.01 & 0.25 & 0.05 & 0.05 & & \\ \midrule
& \multicolumn{8}{c}{\hspace{-18pt}terms $\widetilde{\mo}$ } & \multicolumn{4}{c}{\hspace{-18pt} logical conn.\ $\overline{\mo}$} & after Eq.~\ref{eq:ax_prob_calc} & after Eq.~\ref{eq:adapt_ax_prob_to_get_min_diags}\\ 
\cmidrule(l{0em}r{1.5em}){2-9} \cmidrule(r{1.7em}){10-13} \cmidrule(l{-0.2em}r{1.3em}){14-14} \cmidrule(l{-0.2em}r{2pt}){15-15}
$\tax\in\mo$	& $A$ & $B$ & $E$ & $F$ & $G$ & $X$ & $Y$ & $Z$ & $\rightarrow$ & $\lnot$ & $\land$ & $\lor$ & $p(\tax)$ & $p(\tax)$\\ \midrule
						$\tax_1$ & 1 &   & 1 &   &   &   &   &   &       1       &         &         &        & 0.28	 &	0.14		  \\
						$\tax_2$ &   &   & 1 & 1 &   & 1 & 1 & 1 &       1       &         &     2   &    1   & 0.89  &	0.43			\\
						$\tax_3$ &   & 1 &   & 1 &   &   &   &   &       1       &         &         &        & 0.07  &  0.03			\\
						$\tax_4$ &   & 1 &   &   &   & 1 &   &   &       1       &         &         &        & 0.12	 &	0.06			\\
						$\tax_5$ & 1 &   &   &   &   &   & 1 &   &       1       &     1   &         &        & 0.78	 &	0.38			\\
						$\tax_6$ &   & 1 &   &   &   &   &   & 1 &       1       &         &         &        & 0.61	 &	0.30			\\
						$\tax_7$ &   &   &   &   & 1 &   &   & 1 &       1       &         &         &        & 0.76	 &	0.37			\\ 
						\bottomrule
\end{tabular}
\caption[Computation of Fault Probabilities]{Computing fault probabilities of formulas in $\mo$ given fault probabilities of syntactical elements $e \in\widetilde{\mo}\cup\overline{\mo}$ for the DPI given by Table~\ref{tab:example2}.} \label{tab:example:ax_prob}
\end{table*}

\section{Non-Interactive Debugging Algorithm}
\label{sec:non_int_debug_procedure}

Algorithm~\ref{algo:non_int_debug} describes the procedure for non-interactive debugging of KBs. The algorithm requires as input all the parameters that are required by Algorithm~\ref{algo:hs} and an additional parameter $auto \in \setof{\true,\false}$ indicating either automatic ($\true$) or manual ($\false$) mode. If $auto=\false$, Algorithm~\ref{algo:non_int_debug} calls $\scHS$ (Algorithm~\ref{algo:hs}) with the parameters as provided. The set of minimal diagnoses $\mD$ returned by $\scHS$ is then presented to the user who can select a diagnosis manually after inspecting the diagnoses in $\mD$. Alternatively, in case of $auto = \true$, the system calls $\scHS$ with the parameters as provided, but with $n_{\min} = n_{\max} = 1$. Hence, only the most probable minimal diagnosis is computed by $\scHS$ and returned as an output of Algorithm~\ref{algo:non_int_debug} to the user.

If a user wants the algorithm to output the set of all minimal diagnoses w.r.t.\ $\langle\mo,\mb,\Tp,\Tn\rangle_\RQ$, then the parameter setting $auto = \false$ and $n_{\min} = \infty$ must be chosen. If, on the other hand, a fixed number $n$ of leading diagnoses should be computed (as long as there are at least $n$ minimal diagnoses for the DPI), then $n_{\min} := n =: n_{\max}$ are the correct parameter settings. Note that in both cases the specification of $t$ has no effect.

Of course, the user can also apply Algorithm~\ref{algo:non_int_debug} several times with varying parameters $t$, $n_{\min}$, $n_{\max}$ and $p()$. Or they can specify a test case, i.e.\ add a set of formulas $X$ either to $\Tp$ (if each $\tax\in X$ should be entailed by the correct KB) or to $\Tn$ (if \emph{the conjunction of} all formulas in $X$ must not be implied by the correct KB), and rerun the algorithm with this modified DPI. 

Anyway, the user must either find the correct diagnosis (if it is an element of the output set $\mD$ at all) by hand or be convinced that the returned minimum cardinality or respectively maximum probability diagnosis is indeed the one that yields a solution KB with the intended semantics. Moreover, when formulating test cases by hand, a user can be assumed to be as likely to specify something contradictory or faulty as during creation of the KB itself.

Unsurprisingly, application of Algorithm~\ref{algo:non_int_debug} will often lead to unsatisfying solution ontologies. Remedy for this is provided by Interactive KB Debugging which on the one hand requires higher effort of one (or several) user(s), but on the other hand ensures a high quality solution in terms of its semantics to the problem of Parsimonious KB Debugging (Problem Definition~\ref{prob_def:evidence_just}). 

\begin{algorithm*}
\small
\caption{Non-Interactive KB Debugging} \label{algo:non_int_debug}
\begin{algorithmic}[1]
\Require a tuple $\tuple{ \langle\mo,\mb,\Tp,\Tn\rangle_\RQ, t, n_{\min}, n_{\max}, p(), auto}$ consisting of
\begin{itemize}
	\item an admissible DPI $\langle\mo,\mb,\Tp,\Tn\rangle_\RQ$,
	\item some computation timeout $t$,
	\item a desired minimal ($n_{\min}$) and maximal ($n_{\max}$) number of diagnoses to be returned,
	\item a function $p:\mo \rightarrow (0, 0.5)$ and
	\item a boolean parameter $auto \in \setof{\true,\false}$.
\end{itemize}
\Ensure a set $\mD$ which is 
\begin{enumerate}[(a)]
	\item the set of the $|\mD|$ most probable minimal diagnoses w.r.t.\ $\langle\mo,\mb,\Tp,\Tn\rangle_\RQ$ such that $n_{\min} \leq |\mD| \leq n_{\max}$, if at least $n_{\min}$ minimal diagnoses exist w.r.t.\ $\langle\mo,\mb,\Tp,\Tn\rangle_\RQ$, or
	\item the set of all minimal diagnoses w.r.t.\ $\langle\mo,\mb,\Tp,\Tn\rangle_\RQ$ otherwise
\end{enumerate}
where ``most-probable'' refers to the probability measure $p_{nodes}()$ (cf.\ Definition~\ref{def:p_node()}) obtained from the given function $p()$. 
\vspace{10pt}
\If{$auto = \true$}
	\State $\mD \gets \Call{\scHS}{\langle\mo,\mb,\Tp,\Tn\rangle_\RQ, t, 1, 1, p()}$  \Comment{see Algorithm~\ref{algo:hs}}
\Else 
	\State $\mD \gets \Call{\scHS}{\langle\mo,\mb,\Tp,\Tn\rangle_\RQ, t, n_{\min}, n_{\max}, p()$}  \Comment{see Algorithm~\ref{algo:hs}}
\EndIf
\State \Return $\mD$
\end{algorithmic}
\normalsize
\end{algorithm*}

\begin{figure*}
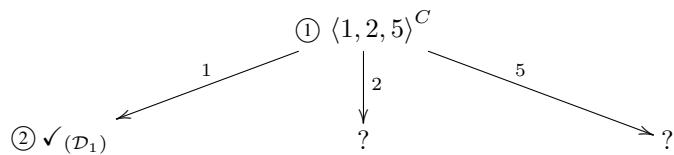

\centering
\begin{minipage}[c]{0.99\textwidth}
\xygraph{
!{<0cm,0cm>;<2cm,0cm>:<0cm,1.5cm>::}
!{(1,4)}*+{{\textcircled{\scriptsize 1}}\,\tuple{1,2,5}^C}="c1c"
!{(-1,3) }*+{{\textcircled{\scriptsize 2}}\,\checkmark_{(\md_1)}}="d1"
!{(1,3) }*+{{\textcircled{\scriptsize 3}}\,\checkmark_{(\md_2)}}="d2" 
!{(3,3) }*+{{\textcircled{\scriptsize 4}}\,\tuple{1,2,7}^C}="c2c"
!{(1,2) }*+{{\textcircled{\scriptsize 5}}\,\times_{(\supset\md_1)}}="nonmin"
!{(3,2) }*+{{\textcircled{\scriptsize 6}}\,\times_{(\supset\md_2)}}="nonmin1"
!{(4,2) }*+{{\textcircled{\scriptsize 7}}\,\checkmark_{(\md_3)}}="d3"
"c1c":"d1"_{1}
"c1c":"d2"^{2}
"c1c":"c2c"^{5}
"c2c":"nonmin"_{1}
"c2c":"nonmin1"^{2}
"c2c":"d3"^{7}
}
\end{minipage}

\vspace{5pt}
\begin{flushleft}
\scriptsize $auto=\false$
\end{flushleft}
\vspace{-10pt} 
\hdashrule{\textwidth}{0.5pt}{2mm}
\vspace{-15pt} 
\begin{flushleft}
\scriptsize $auto=\true$
\end{flushleft} 

\vspace{5pt}
\centering
\begin{minipage}[c]{0.99\textwidth} 
\xygraph{
!{<0cm,0cm>;<2cm,0cm>:<0cm,1.5cm>::}
!{(1,4)}*+{{\textcircled{\scriptsize 1}}\,\tuple{1,2,5}^C}="c1c"
!{(-1,3) }*+{{\textcircled{\scriptsize 2}}\,\checkmark_{(\md_1)}}="d1"
!{(1,3) }*+{?}="d2" 
!{(3,3) }*+{?}="c2c"
"c1c":"d1"_{1}
"c1c":"d2"^{2}
"c1c":"c2c"^{5}
}
\end{minipage}

\vspace{10pt}
\caption[Non-Interactive KB Debugging Process without Fault Information]{Non-interactive KB debugging process without any given fault information applied to the DPI given by Table~\ref{tab:example2} with settings $auto = \false$ and $n_{\min} = \infty$ (above) and $auto = \true$ (below).} \label{fig:example:non-interactive_onto_debug_auto=false+nmin=infty_and_auto=true}
\end{figure*}

\begin{figure*}
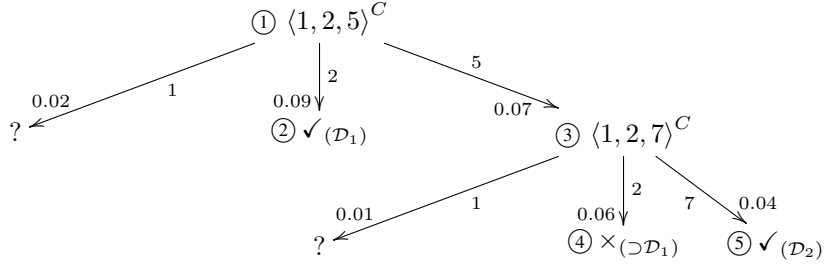

\centering
\begin{minipage}[c]{0.99\textwidth} 
\xygraph{
!{<0cm,0cm>;<2cm,0cm>:<0cm,1.5cm>::}
!{(1,4)}*+{{\textcircled{\scriptsize 1}}\,\tuple{1,2,5}^C}="c1c"
!{(-1,3) }*+{?}="d1"
!{(1,3) }*+{{\textcircled{\scriptsize 2}}\,\checkmark_{(\md_1)}}="d2" 
!{(3,3) }*+{{\textcircled{\scriptsize 3}}\,\tuple{1,2,7}^C}="c2c"
!{(1,2) }*+{?}="nonmin"
!{(3,2) }*+{{\textcircled{\scriptsize 4}}\,\times_{(\supset\md_1)}}="nonmin1"
!{(4,2) }*+{{\textcircled{\scriptsize 5}}\,\checkmark_{(\md_2)}}="d3"
"c1c":"d1"^{1}_(0.85){0.02}
"c1c":"d2"^{2}_(0.72){0.09}
"c1c":"c2c"^{5}_(0.67){0.07}
"c2c":"nonmin"^{1}_(0.85){0.01}
"c2c":"nonmin1"^{2}_(0.72){0.06}
"c2c":"d3"_{7}^(0.75){0.04}
}
\end{minipage}

\vspace{10pt}
\caption[Non-Interactive KB Debugging Process with Fault Information]{Non-interactive KB debugging process with given fault information applied to the DPI given by Table~\ref{tab:example2} with settings $auto = \false$, $n_{\min} = 2$, $n_{\max} = 4$ and $t = 1$.} \label{fig:example:non-interactive_onto_debug_auto=false+nmin=2+nmax=4_with_probs}
\end{figure*}

\begin{example}\label{example:non_interactive_debugging_with_tabExDpi2_and_without_probs}
Assume a user wants to find a maximal solution KB for the example DPI $\tuple{\mo,\mb,\Tp,\Tn}_{\RQ}$ provided by Table~\ref{tab:example2} and that no data giving information about fault probabilities of syntactical constructs or formulas in $\mo$ is available. Therefore, let $p(\tax) := c$ for some fixed $c \in (0,0.5)$ (see Section~\ref{sec:probs_diag_comp} for an explanation of this choice of $c$). The non-interactive KB debugging algorithm presented by Algorithm~\ref{algo:non_int_debug} called with $\tuple{\mo,\mb,\Tp,\Tn}_{\RQ}$, the function $p()$, $n_{\min}=\infty$ and $auto = \false$ as inputs results in the hitting set tree given by the upper picture in Figure~\ref{fig:example:non-interactive_onto_debug_auto=false+nmin=infty_and_auto=true}. By $n_{\min}=\infty$ and $auto = \false$, the user signalizes that \emph{inspection of} \emph{all} minimal diagnoses w.r.t.\ the input DPI is desired. Hence, the (complete) breadth-first pHS-tree as per Algorithm~\ref{algo:hs} is constructed. So, the output is the set of all minimal diagnoses $\minD_{\tuple{\mo,\mb,\Tp,\Tn}_{\RQ}} = \setof{[1],[2],[5,7]}$.

In the shown hitting set tree, minimal diagnoses are indicated by nodes labeled by $\checkmark_{(\md)}$ where $\md$ is a name given to this diagnosis. A node closed due to non-minimality is denoted by $\times_{(\supset \md)}$ where $\md$ is some minimal diagnosis that is a subset of the set of edge labels along the path leading from the root node to this node. 
The label $\mc^C$ means that the minimal conflict set $\mc$ has been freshly computed by a call to $\scQX$. The label $\mc^R$, on the other hand, means that the minimal conflict set $\mc$ has been reused from the set of already computed minimal conflict sets. In this example, both minimal conflict sets are computed by $\scQX$ and no conflict sets are reused. The order of node labeling is indicated by the numbers $\textcircled{\scriptsize i}$ starting from 1. Open nodes, i.e.\ generated nodes that have not yet been labeled, are indicated by a question mark.

In case $auto = \true$ was given as an input to the algorithm instead, the partial pHS-tree depicted by the lower picture in Figure~\ref{fig:example:non-interactive_onto_debug_auto=false+nmin=infty_and_auto=true} would be constructed and the output would be $\mD = \setof{\md_1} = \setof{[1]}$ containing just the first found and thus most probable minimal diagnosis w.r.t.\ the input DPI. Note that $\md_1 = [1]$ and $\md_2 = [2]$ (which is not computed) have equal probability and whether the one or the other is computed first depends only on the ordering of equally probable (in this case: equal cardinality) nodes in $\Queue$. As already mentioned in Section~\ref{sec:probs_diag_comp}, in this example the most probable diagnosis is equivalent to a minimum cardinality diagnosis since all formula probabilities are equal.

Please notice that the internal ``flat'' representation used by Algorithm~\ref{algo:hs} which does not store a tree but only the set of open and closed nodes differs from the standard tree representation~\cite{Kalyanpur2006a, friedrich2005gdm, Suntisrivaraporn2008, Reiter87} we use to depict the hitting set tree \emph{graphically} in Figure~\ref{fig:example:non-interactive_onto_debug_auto=false+nmin=infty_and_auto=true}. Whereas within Algorithm~\ref{algo:hs} a node $\mathsf{node}$ stores the set of all the edge labels on the path leading from the root node to $\mathsf{node}$, in the figure we label each node in the tree by the respective label that is computed for this node by the \textsc{label} function, i.e.\ either by a minimal conflict set, by $\checkmark$ or by $\times$.\qed
\end{example}

\begin{example}\label{example:non_interactive_debugging_with_tabExDpi2_and_probs}
Recall Example~\ref{example:ax_prob_calc} which demonstrated how formula fault probabilities are constructed from fault probabilities of syntactical elements for the example DPI depicted by Table~\ref{tab:example2}. Now we want to show how the non-interactive KB debugging algorithm given by Algorithm~\ref{algo:non_int_debug} works when these formula probabilities are incorporated. 

Suppose the inputs to the algorithm are the DPI $\tuple{\mo,\mb,\Tp,\Tn}_{\RQ}$, the function $p(\tax)$ for $\tax\in\mo$ displayed by the rightmost column of Table~\ref{tab:example:ax_prob} and $auto = \false$. Further on, let the user of the debugging algorithm be willing to wait a maximum of one second for an output and let them postulate a minimum of two most probable minimal diagnoses to be returned, e.g.\ to have at least a second choice if the employed formula probabilities are not perfectly suitable and the most probable diagnosis is not the desired solution. These postulations are expressed by specifying the parameters $n_{\min} = 2$ and $t = 1$ (second). Additionally, assume the user expects the provided probabilities to be sufficiently reasonable such that the desired diagnosis will be among the best four diagnoses wherefore $n_{\max} = 4$ is chosen. Moreover, let us imagine that the time for each fresh computation of a minimal conflict plus generation of the (unlabeled) successor nodes of this node is $0.4$ seconds 
and the cost of computing any other label of a node is $0.1$ seconds. 

Then the partial wpHS-tree produced by Algorithm~\ref{algo:non_int_debug} initialized in this way is illustrated by Figure~\ref{fig:example:non-interactive_onto_debug_auto=false+nmin=2+nmax=4_with_probs}. The used notation is as described in Example~\ref{example:non_interactive_debugging_with_tabExDpi2_and_without_probs} with one additional attribute. Namely, each edge is not only labeled by one element of the conflict set from which it goes out, but also by a label $p \in (0,1)$ that is placed near the arrow head of the arrow that expresses the edge. This label $p$ gives 
the probability as per $p_{nodes}()$ (cf.\ Definition~\ref{def:p_node()}) of the (partial) diagnosis that corresponds to the union of the 
edge labels along the path from the root to and including the edge that is labeled by $p$. 
For example, the label $0.06$ of the edge directed at the node number $\textcircled{\scriptsize 4}$ means that the probability of $\setof{2,5}$ is $0.06$. Further on, open, i.e.\ generated, but not yet labeled nodes, are designated by a question mark.

As outlined by the circled numbers $\textcircled{\scriptsize i}$, as a first action the root node is labeled by the newly computed minimal conflict set $\tuple{1,2,5}$, the computation time of which amounts to $0.4$. Then, the tree construction proceeds according to the (partial) diagnosis probabilities according to $p_{nodes}()$ computed from the formula probabilities $p(\tax), \tax\in\mo$ provided by the last column of Table~\ref{tab:example:ax_prob}. 
Therefore, the most probable edge leading away from the root node is labeled next. This already leads to the finding of the first minimal diagnosis $\md_1 = [2]$ after overall computation time of $0.5$ seconds. Since $n_{\min} = 2$ diagnoses have not yet been computed and there are still unlabeled open nodes, namely those corresponding to paths $\setof{1}$ and $\setof{5}$, the algorithm continues the execution by labeling the next best node $\setof{5}$ with a probability of $0.07$ -- as opposed to $0.02$ for the other open node $\setof{1}$. Since $\setof{5}$ is neither a superset of an already computed minimal diagnosis nor a duplicate of another open node nor a diagnosis itself, it must be labeled by some minimal conflict set. Because the already established minimal conflict set $\tuple{1,2,5}$ is not disjoint with $\setof{5}$, no reuse is possible and $\scQX$ is called to determine a new minimal conflict set $\tuple{1,2,7}$ w.r.t.\ $\tuple{\mo,\mb,\Tp,\Tn}_{\RQ}$. All successor nodes of the newly labeled node $\textcircled{\scriptsize 3}$, i.e.\ the nodes corresponding to the paths $\setof{1,5}, \setof{2,5}$ and $\setof{5,7}$, are added to the list $\Queue$ of open nodes such that descending order of probabilities is maintained. The resulting queue is then $\Queue = [\setof{2,5},\setof{5,7}, \setof{1}, \setof{1,5}]$. As a next step, again the first and thus best open node $\setof{2,5}$ is chosen from $\Queue$ and labeled by $\times_{(\supset \md_1)}$ which means that the corresponding path is closed since it is a superset of an already found minimal diagnosis, namely $\md_1 = [2]$. At this point, the overall computation time amounts to $1$ second which corresponds to the time limit $t$. For that reason, the algorithm will go ahead searching for minimal diagnoses only until a minimal number $n_{\min}$ thereof is detected. 
The node processed next, corresponding to the path $\setof{5,7}$, is then determined to be a minimal diagnosis by the \textsc{label} procedure.
 
Thus, the output of the algorithm after $1.1$ seconds execution time is the set of minimal diagnoses $\mD = \setof{[2],[5,7]}$ which is a proper subset of all minimal diagnoses $\mD_{\tuple{\mo,\mb,\Tp,\Tn}_{\RQ}} = \setof{[1],[2],[5,7]}$. However, if we assume that the user's intended KB should entail $E \rightarrow G$, for instance, then none of the returned diagnoses can be used to compute a solution KB featuring this entailment when integrated with the background knowledge $\mb$. Hence, the true diagnosis $\dt$ would be missed in this case.

Also, when computing all minimal diagnoses w.r.t.\ a DPI -- if this is even possible in a concrete case due to the computational complexity -- and showing them to the user, a user might review just the most probable ones and make a decision on which one to choose only based on these. For instance, \cite{ksgf2010} reported on one DPI where computation of all minimal diagnoses, 1782 in number, is feasible. In such a case it is hard to expect that a user will be willing or will have the time to inspect more than a small fraction of these 1782 diagnoses. The consequence will be a wrong choice of diagnosis in many cases, also because a simple view on a diagnosis will often not lead to the certainty of a user that this one is or is not the desired one. The reason for this is that usually it is too complex for a human brain to perform the necessary mental reasoning to make oneself a picture of the implications of choosing one diagnosis as opposed to another one.

For our example DPI, a user getting the output $\mD = \minD_{\tuple{\mo,\mb,\Tp,\Tn}_{\RQ}} = \setof{[1],[2],[5,7]}$ with the computed probabilities $p([1]) = 12\%$, $p([2]) = 60\%$ and $p([5,7]) = 28\%$ might decide to just inspect the diagnoses that make the most probable $80\%$ fraction of diagnoses. In this case, either $[2]$ or $[5,7]$ would be selected, which corresponds to a wrong choice in case $E \rightarrow G$ should be entailed be the resulting solution KB after integration with the background KB $\mb$.\qed  
\end{example}

\clearpage
\thispagestyle{empty}

\chapter{Interactive Knowledge Base Debugging}
\label{chap:InteractiveOntologyDebugging}
So far, we have learned that the problem of (parsimonious) KB debugging as defined in Problem Definitions~\ref{prob_def:onto_debug} and \ref{prob_def:evidence_just} in Chapter~\ref{chap:OntologyDebugging} can be solved by investigating minimal diagnoses w.r.t.\ a given DPI $\langle\mo,\mb,\Tp,\Tn\rangle_\RQ$. We have seen how minimal diagnoses can be computed, we have introduced a probability space over diagnoses and we have discussed how a-priori probability estimates for diagnoses can be established. Now, assume the situation where a DPI with say $100$ minimal diagnoses is given, among which there is one diagnosis $\md$ with highest estimated probability $p(\md) = 10\%$. By the definitions of a diagnosis and a solution KB (Definitions~\ref{def:target_ont} and \ref{def:diagnosis}), each of the $100$ diagnoses can be used to formulate a solution KB w.r.t.\ the DPI $\langle\mo,\mb,\Tp,\Tn\rangle_\RQ$. So, should the system output the solution KB $(\mo \setminus \md) \cup U_{\Tp}$ obtained from $\md$ as the optimal solution? Will a user be satisfied with a likeliness of $90\%$ of being offered a suboptimal solution? What if the diagnoses probabilities are bad estimates and another diagnosis $\md'$ should actually have a probability of $20\%$? 

Why not simply apply Algorithm~\ref{algo:non_int_debug} to show all $100$ minimal diagnoses to the user and let them select the preferred one by hand? First, due to the complexity of diagnosis calculation algorithms (cf.\ Chapter~\ref{chap:intro}), pre-computation of $100$ (or, generally, all) minimal diagnoses is usually not tractable within reasonable time. This makes such an approach quite unattractive in an interactive setting. Second, going through large sets of diagnoses can be time-consuming, tedious and error-prone. Third, human beings are normally not capable of (fully) realizing the semantic consequences of deleting a diagnosis from a KB, especially if the KB is large, complex and/or has been created by multiple engineers or automatic systems. Thus, applying a suboptimal diagnosis can result in unexpected entailments or unwanted changes, and thus an incorrect solution KB (incorrect in the sense of the semantics, \emph{not} in the sense of violating given requirements or test cases), which might cause unexpected new faults and contradictions when augmented by new formulas. Consequently, a solution diagnosis is only acceptable if the user has sufficiently scrutinized and approved its semantic effect to the KB.

This leads to the definition of two types of Interactive KB Debugging problems. First, there is the problem of \emph{Interactive Dynamic KB Debugging} which, given an input DPI, aims at the extension of this DPI by new test cases confirmed by a user such that there is only one minimal diagnosis left w.r.t.\ the extended DPI. Second, we specify the problem of \emph{Interactive Static KB Debugging} which, given an input DPI, aims at the formulation of new test cases confirmed by a user such that these new test cases rule out all but one minimal diagnosis w.r.t.\ the input DPI.\vspace{3pt}

\noindent\fcolorbox{black}{light-gray1}{\parbox[c][7.1em][c]{0.975\linewidth}{\vspace{-4pt}
\begin{prob_def}[Interactive Dynamic KB Debugging]\label{prob_def:dynamic}
Given a DPI $\langle\mo,\mb,\Tp,\Tn\rangle_\RQ$, the task is to find a maximal solution KB $(\mo\setminus\md) \cup U_{\Tp\cup\Tp'}$ w.r.t.\ a DPI $\langle\mo,\mb,\Tp\cup\Tp',\Tn\cup\Tn'\rangle_\RQ$ such that 
\begin{itemize}
\item $\md$ is the only minimal diagnosis w.r.t.\ $\langle\mo,\mb,\Tp\cup\Tp',\Tn\cup\Tn'\rangle_\RQ$ and 
\item a user has confirmed that each $\tp'\in\Tp'$ is a positive test case and that each $\tn'\in\Tn'$ is a negative test case.
\end{itemize}
\end{prob_def}\vspace{-4pt}
}}

\vspace{3pt}
\begin{remark}\label{rem:ad_int_dyn_KB_debug_problem}
The solution of an Interactive Dynamic KB Debugging problem given the DPI $\langle\mo,\mb$, $\Tp,\Tn\rangle_\RQ$ solves the problem of KB Debugging (Problem Defnition~\ref{prob_def:onto_debug}) as well as the problem of Parsimonious KB Debugging (Problem Defnition~\ref{prob_def:evidence_just}) for the DPI $\langle\mo,\mb,\Tp\cup\Tp',\Tn\cup\Tn'\rangle_\RQ$, but in general \emph{not} for the original DPI $\langle\mo,\mb,\Tp,\Tn\rangle_\RQ$. This is the reason why we term it ``dynamic'', since a solution is found for a version of the initial DPI that has been extended by test cases.\qed
\end{remark}
\vspace{4pt}

\noindent\fcolorbox{black}{light-gray1}{\parbox[c][8.5em][c]{0.975\linewidth}{\vspace{-4pt}
\begin{prob_def}[Interactive Static KB Debugging]\label{prob_def:static}
Given a DPI $\langle\mo,\mb,\Tp,\Tn\rangle_\RQ$, the task is to find a maximal solution KB $(\mo\setminus\md) \cup U_\Tp$ w.r.t.\ $\langle\mo,\mb,\Tp,\Tn\rangle_\RQ$ such that 
\begin{itemize}
\item there are sets of positive test cases $\Tp'$ and negative test cases $\Tn'$ where a user has confirmed that each $\tp'\in\Tp'$ is a positive test case and that each $\tn'\in\Tn'$ is a negative test case, and 
\item $\md$ is the only minimal diagnosis w.r.t.\ $\langle\mo,\mb,\Tp,\Tn\rangle_\RQ$ that satisfies all positive and negative test cases $\Tp'$ and $\Tn'$, respectively.
\end{itemize}
\end{prob_def}\vspace{-4pt}
}}

\vspace{3pt}
\begin{remark}\label{rem:ad_int_static_KB_debug_problem}
The solution of an Interactive Static KB Debugging problem given the DPI $\langle\mo,\mb,\Tp,\Tn\rangle_\RQ$ constitutes a solution to the problem of KB Debugging (Problem Defnition~\ref{prob_def:onto_debug}) as well as to the problem of Parsimonious KB Debugging (Problem Defnition~\ref{prob_def:evidence_just}) \emph{for the original DPI} $\langle\mo,\mb,\Tp,\Tn\rangle_\RQ$, therefore the term ``static''.\qed
\end{remark}
Now, we give a more formal definition of a true diagnosis (an informal characterization of which was given in Section~\ref{sec:DiagnosisProbabilitySpace}). If sufficiently many new test cases are specified and added to a given DPI such that there is only one remaining minimal diagnosis w.r.t.\ the input DPI (the input DPI extended by the new test cases) left, then this diagnosis is referred to as the true diagnosis w.r.t.\ Interactive Static (Dynamic) KB Debugging.
\begin{definition}[True Diagnosis]\label{def:true_diagnosis}
Let $\dt$ be equal to $\md$ in Problem Definition~\ref{prob_def:minimize_user_interact_static} (\ref{prob_def:minimize_user_interact_dynamic}). Then $\dt$ is called the \emph{true diagnosis w.r.t.\ Interactive Static KB Debugging (Interactive Dynamic KB Debugging)}.
\end{definition}

\section{User Interaction}
\label{sec:UserInteraction}
The idea in interactive KB debugging is to iteratively consult a user asking them to give additional information as regards desired and undesired entailments of the correct KB. Thus, the principle of interactive KB debugging is based on that of \emph{Sequential Diagnosis} which has been suggested by \cite{dekleer1987} as an iterative way to localize the faulty components (among an initially large set of possibilities) in malfunctioning digital circuits by performing repeated (most informative) measurements. We have shown in our previous works \cite{ksgf2010,Shchekotykhin2012} how sequential diagnosis can be applied to KBs (ontologies). 

In our approach, for the selection of which question (of a pool of possible ones) to ask a user next, an active learning~\cite{settles2012} approach is applied.\footnote{Note that the minimal a-posteriori expected entropy of solution candidate probabilities as a means to select the best next measurement as used in \cite{dekleer1987} is only one of many possible active learning strategies \cite{settles2012}.} \emph{Active Learning} is an iterative supervised machine learning technique in which a learning algorithm is able to interactively query the user to obtain a label for a desired unlabeled instance. In the case of a KB debugging system, an unlabeled instance is a set of logical formulas and the label is whether the conjunction of these formulas should or should not be entailed by the correct KB. Since the learner can choose the instances to be labeled, the number of consultations of an interacting user required to learn a concept (in this case the one solution KB with the desired semantics w.r.t.\ a given DPI) can often be much lower than the number required in a standard supervised learning setting since the risk that the algorithm must deal with lots of uninformative examples is reduced. 

We suppose the user of an interactive KB debugger to be a single person or multiple persons, usually experts of the particular domain the faulty KB is dealing with or authors of the faulty KB.
Moreover, we 
assume the interacting user to be able to answer concrete queries about the intended domain that should be modeled. Otherwise put, we suppose that a user can classify a given logical formula (or a conjunction of logical formulas) as a wanted or unwanted proposition in the intended domain, i.e.\ as an entailment or non-entailment of the correct domain model. We have already argued in Chapter~\ref{chap:intro} why this assumption is plausible. 

\subsection{Queries}
\label{sec:Queries}
In interactive KB debugging, a set of logical formulas $Q$ is presented to the user who should decide whether to assign $Q$ to the set of positive ($\Tp$) or negative ($\Tn$) test cases w.r.t. a given DPI $\langle\mo,\mb,\Tp,\Tn\rangle_\RQ$. In other words, the system asks the user ``should the KB you intend to model entail all formulas in $Q$?''. In that, $Q$ is generated by the debugging algorithm in a way that \emph{any} decision of the user 
\begin{enumerate}
	\item invalidates at least one minimal diagnosis (\emph{search space restriction}) and
	\item preserves validity of at least one minimal diagnosis (\emph{solution preservation}).
\end{enumerate}
We call a set of logical formulas $Q$ with these properties a \emph{query}. Successive classification of queries as entailments (all formulas in $Q$ must be entailed) or non-entailments (at least one formula in $Q$ must not be entailed) of the correct KB enables gradual restriction of the search space for (minimal) diagnoses. Further on, classification of sufficiently many queries guarantees the detection of a single correct solution diagnosis which can be used to determine a solution KB with the correct semantics w.r.t.\ a given DPI.\footnote{Correctness of the diagnosis must not be understood as a guarantee that all formulas in the KB which are not in the diagnosis are definitely correct. Instead, correctness must be seen with regard to other diagnoses and with the ``Principle of Parsimony'' in mind (cf.\ Section~\ref{sec:MinimallyInvasiveOntologyDebugging}). That is, all other possible diagnoses are ruled out by a present set of test cases wherefore the single remaining diagnosis is the one that is correct (in comparison with all other incorrect ones). And, there is no evidence (at the time the correct diagnosis is found) that any other formulas in the KB might be faulty. This might change however after new formulas are added to the KB.} 

\begin{definition}[Query]\label{def:query}
Let $\langle\mo,\mb,\Tp,\Tn\rangle_\RQ$ over $\mathcal{L}$ and $\mD \subseteq \minD_{\langle\mo,\mb,\Tp,\Tn\rangle_\RQ}$. 
Then a set of logical formulas $Q\neq\emptyset$ over $\mathcal{L}$ is called a \emph{query w.r.t.\ $\mD$} iff there are diagnoses $\md, \md' \in \mD$ such that $\md \notin \minD_{\langle\mo,\mb,\Tp \cup \setof{Q},\Tn\rangle_\RQ}$ and $\md'\notin\minD_{\langle\mo,\mb,\Tp,\Tn \cup \setof{Q}\rangle_\RQ}$. The set of all queries w.r.t.\ $\mD$ and $\langle\mo,\mb,\Tp,\Tn\rangle_\RQ$ is denoted by $\mQ_{\mD,\langle\mo,\mb,\Tp,\Tn\rangle_\RQ}$.
\end{definition}
\begin{remark}
Although Definition~\ref{def:query} only postulates that at least one diagnosis in $\mD$ is invalidated for whatever answer is given to the query, this implies that, for each answer to the query, there is also a diagnosis that remains valid after adding the corresponding test case to the DPI, as will be shown by Proposition~\ref{prop:query_dx_dnx}.\qed
\end{remark}

So, w.r.t. a set of minimal diagnoses $\mD \subseteq \minD_{\langle\mo,\mb,\Tp,\Tn\rangle_\RQ}$, a query $Q$ is a set of logical formulas that rules out at least one diagnosis in $\mD$ (and therefore in $\minD_{\langle\mo,\mb,\Tp,\Tn\rangle_\RQ}$) as a candidate to formulate a solution KB, regardless of whether $Q$ is classified as a positive or negative test case.

\subsection{Leading Diagnoses}
\label{sec:LeadingDiagnoses}
Query generation requires a precalculated set of minimal diagnoses $\mD \subseteq \minD_{\langle\mo,\mb,\Tp,\Tn\rangle_\RQ}$ 
that serves as a representative for all minimal diagnoses $\minD_{\langle\mo,\mb,\Tp,\Tn\rangle_\RQ}$. As already mentioned, computation of the entire set $\minD_{\langle\mo,\mb,\Tp,\Tn\rangle_\RQ}$ is generally not tractable within reasonable time. Usually, $\mD$ is defined as a set of most probable or minimum cardinality diagnoses (cf.\ Chapter~\ref{chap:DiagnosisComputation}). Therefore, $\mD$ is called the set of \emph{leading diagnoses w.r.t. $\langle\mo,\mb,\Tp,\Tn\rangle_\RQ$} \cite{Shchekotykhin2012}. 

The leading diagnoses $\mD$ are then exploited to determine a query $Q$ the answering of which enables a discrimination between the diagnoses in $\minD_{\langle\mo,\mb,\Tp,\Tn\rangle_\RQ}$. That is, a subset of $\minD_{\langle\mo,\mb,\Tp,\Tn\rangle_\RQ}$ which is not ``compatible'' with the new information obtained by adding the test case $Q$ to $\Tp$ or $\Tn$ is ruled out (see Proposition~\ref{prop:dpi_update} below). For the computation of the subsequent query only a leading diagnoses set $\mD_{new}$ w.r.t.\ the minimal diagnoses still compliant with the new sets of test cases $\Tp'$ and $\Tn'$ is taken into consideration, i.e.\ $\mD_{new} \subseteq \mD_{\langle\mo,\mb,\Tp',\Tn'\rangle_\RQ}$.

The number of precomputed leading diagnoses $\mD$ affects the quality of the obtained query. The higher $|\mD|$, the more representative is $\mD$ w.r.t. $\minD_{\langle\mo,\mb,\Tp,\Tn\rangle_\RQ}$, the more options there are to specify a query in a way that a user can easily comprehend and answer it, and the higher is the chance that a query that eliminates a high rate of diagnoses w.r.t. $\mD$ will also eliminate a high rate of all minimal diagnoses $\minD_{\langle\mo,\mb,\Tp,\Tn\rangle_\RQ}$. The selection of a lower $|\mD|$ on the other hand means better timeliness regarding the interaction with a user, first because fewer leading diagnoses might be computed much faster and second because the search space for an ``optimal'' query is smaller.\footnote{Roughly, a query $Q$ is ``optimal'' if the number of queries that still need to be answered to identify the desired solution KB after $Q$ is added to the (positive or negative) test cases is minimal. ``Optimality'' of a query can be captured by quantitative information theoretic measures studied in the field of active learning \cite{settles2012} that can be used to estimate the quality of a query beforehand, i.e.\ before an answer to it is known. See Section~\ref{sec:query_selection_measures} and \cite{Rodler2013, ksgf2010, Shchekotykhin2012} for details.}
So, the optimal number of leading diagnoses depends on the complexity of the particular DPI considered. One way to determine a suitable $|\mD|$ can be to first define an interval $[n_{\min},n_{\max}]$ that must comprise $|\mD|$ where the upper bound defines the desired number of leading diagnoses and the lower bound the minimally postulated number. Second, the search for minimal diagnoses is run at least as long as it takes to compute $n_{\min}$ diagnoses and at the longest until $n_{\max}$ diagnoses have been found or a timeout $t$ expires that is specified in a manner it enables frequent user interaction. Note that such parameters have already been taken into account in the non-interactive KB debugging Algorithm~\ref{algo:hs} (see Section~\ref{sec:non_int_debug_procedure}).\label{etc:leading_diag_params}

%

\subsection{Q-Partitions}
\label{sec:QPartitions}
Now we introduce the notion of a \emph{q-partition}, a partition of the leading diagnoses set $\mD$ induced by a query w.r.t.~$\mD$. A q-partition will be a helpful instrument in deciding whether a set of logical formulas is a query or not. It will facilitate an estimation of the impact a query answer has in terms of invalidation of minimal diagnoses. And, given fault probabilities, it will enable us to gauge the probability of getting a positive or negative answer to a query. 

From now on, given a DPI $\langle\mo,\mb,\Tp,\Tn\rangle_\RQ$ and some minimal diagnosis $\md_i$ w.r.t.\ $\langle\mo,\mb,\Tp,\Tn\rangle_\RQ$, we will use the following abbreviation for the solution KB obtained by deletion of $\md_i$ along with the given background knowledge $\mb$:
\begin{align} 
\mo^{*}_i \; := \; (\mo \setminus \md_i) \cup \mb \cup U_\Tp \label{eq:sol_ont_candidate} 
\end{align}
\begin{definition}[q-Partition\footnote{In existing literature, e.g.\ \cite{Shchekotykhin2012,Rodler2013,ksgf2010}, a q-partition is often simply referred to as partition. We call it q-partition to emphasize that not each partition of $\mD$ into three sets is necessarily a q-partition.}]\label{def:q-partition}
Let $\langle\mo,\mb,\Tp,\Tn\rangle_\RQ$ be a DPI over $\mathcal{L}$, $\mD\subseteq \minD_{\langle\mo,\mb,\Tp,\Tn\rangle_\RQ}$. 
Further, let $Q$ be a set of logical formulas over $\mathcal{L}$ and
\begin{itemize}
\item $\dx{}(Q):=\setof{\md_i \in \mD\,|\,\mo^{*}_i \models Q}$, 
\item $\dnx{}(Q):=\setof{\md_i \in \mD\,|\,\exists x\in\RQ\cup\Tn: \mo^{*}_i \cup Q \text{ violates } x}$,
\item $\dz{}(Q) := \mD \setminus (\dx{j} \cup \dnx{j})$. 
\end{itemize}
Then $\langle \dx{}(Q), \dnx{}(Q), \dz{}(Q) \rangle$ is called a \emph{q-partition} iff $Q$ is a query w.r.t.\ $\mD$ and $\langle\mo,\mb,\Tp,\Tn\rangle_\RQ$.
\end{definition}
\begin{remark}\label{rem:dnx_contains_exactly_these_diagnoses...}
The set $\dnx{}(Q)$ contains exactly those diagnoses $\md_i\in\mD$ where $\mo \setminus \md_i$ is invalid w.r.t.\ $\tuple{\cdot,\mb,\Tp\cup\setof{Q},\Tn}$ (cf.~Definition~\ref{def:valid_onto}).\qed
\end{remark}
\begin{proposition}\label{prop:q-partition_is_partition}
For each query $Q$ w.r.t. some $\mD \subseteq \minD_{\langle\mo,\mb,\Tp,\Tn\rangle_\RQ}$ it holds that $\langle \dx{}(Q)$, $\dnx{}(Q)$, $\dz{}(Q) \rangle$ is a partition of $\mD$.
\end{proposition}
\begin{proof}
First, by definition of $\dz{}(Q)$, we have that $\dx{}(Q) \cup \dnx{}(Q) \cup \dz{}(Q) = \mD$, $\dx{}(Q) \cap \dz{}(Q) = \emptyset$ and $\dnx{}(Q) \cap \dz{}(Q) = \emptyset$. Second, $\dx{}(Q) \cap \dnx{}(Q) = \emptyset$ since $\mo^{*}_i \models Q_j$ and $\exists x\in\RQ\cup\Tn: (\mo^{*}_i \cup Q_j \text{ violates } x)$ imply by idempotency of $\mathcal{L}$ that $\mo^{*}_i$ violates some $x\in\RQ\cup\Tn$ which is a contradiction to $\md_i$ being a diagnosis w.r.t. $\langle\mo,\mb,\Tp,\Tn\rangle_\RQ$. Thus, each diagnosis in $\mD$ is an element of exactly one set of $\dx{}(Q), \dnx{}(Q), \dz{}(Q)$ which is equivalent to the statement of the proposition.
\end{proof}
\begin{remark}\label{rem:query_partitions_any_set_of_diagnoses_into_dx_dnx_dz}
In fact, Proposition~\ref{prop:q-partition_is_partition} holds for any set $\mD \subseteq \allD_{\langle\mo,\mb,\Tp,\Tn\rangle_\RQ}$, i.e.\ for any subset of \emph{all} diagnoses w.r.t.\ $\langle\mo,\mb,\Tp,\Tn\rangle_\RQ$. This can be easily seen from the proof of Proposition~\ref{prop:q-partition_is_partition} which does not require minimality of diagnoses. That is, any set of diagnoses w.r.t.\ a DPI is partitioned into the three sets $\dx{}(Q)$, $\dnx{}(Q)$ and $\dz{}(Q)$ as per Definition~\ref{def:q-partition} by a query $Q$ w.r.t.\ this DPI.\qed 
\end{remark}
\begin{proposition}\label{prop:unique_q-partition}
For each query $Q$ w.r.t. some $\mD \subseteq \minD_{\langle\mo,\mb,\Tp,\Tn\rangle_\RQ}$ there is one and only one partition $\langle \dx{}(Q), \dnx{}(Q), \dz{}(Q) \rangle$.
\end{proposition}
\begin{proof}
The existence of a partition $\dx{}(Q), \dnx{}(Q), \dz{}(Q)$ follows directly from Proposition~\ref{prop:q-partition_is_partition}. Assume there are two different partitions $\langle \dx{1}(Q), \dnx{1}(Q), \dz{1}(Q) \rangle$ and $\langle \dx{2}(Q), \dnx{2}(Q), \dz{2}(Q) \rangle$. Then, (a)~$\dx{1}(Q) \neq \dx{2}(Q)$ or (b)~$\dnx{1}(Q)\neq\dnx{2}(Q)$ or (c)~$\dz{1}(Q)\neq\dz{2}(Q)$ must hold. If (a) is true, then there is one diagnosis $\md_i \in \mD$ such that $\mo^{*}_{i} \models Q$ and $\mo^{*}_{i} \not\models Q$ -- a contradiction. If (b) is true, then there is one diagnosis $\md_i \in \mD$ such that $\mo^{*}_{i} \cup Q$ violates some $x\in\RQ\cup\Tn$ and $\mo^{*}_{i} \cup Q$ does not violate any $y\in\RQ\cup\Tn$ -- a contradiction. If (c) is true, then $(\dx{1}(Q)\cup\dnx{1}(Q)) \neq (\dx{2}(Q)\cup\dnx{2}(Q))$ which implies that either (a) or (b) must be true. 
\end{proof}
Due to the uniqueness of a q-partition $\langle \dx{}(Q), \dnx{}(Q), \dz{}(Q) \rangle$ for a query $Q$, we denote this q-partition by $\Pt(Q)$.
As a consequence of Definition~\ref{def:q-partition} and Proposition~\ref{prop:unique_q-partition}, a query $Q$ is a set of common entailments of KBs $\mo^{*}_i$, each resulting from the deletion of a single minimal diagnosis $\md_i \in \dx{}(Q)$ from $\mo$. 
\begin{corollary}
For each query $Q\in\mQ_{\mD,\langle\mo,\mb,\Tp,\Tn\rangle_\RQ}$ there is a set of minimal diagnoses $\dx{}(Q) \subseteq \minD_{\langle\mo,\mb,\Tp,\Tn\rangle_\RQ}$ as defined by Definition~\ref{def:q-partition} such that $Q \subseteq \setof{e \,|\,\forall \md_i \in \dx{}(Q): \mo^{*}_i \models e}$.
\end{corollary}

\subsection{Interpretation of Q-Partitions}
\label{sec:InterpretationOfQPartitions}
Since $\mo^{*}_i$ corresponds to the solution KB (along with $\mb$) obtained under the assumption that $\dt = \md_i$, i.e.\ the true diagnosis (cf.\ Definition~\ref{def:true_diagnosis}) corresponds to $\md_i$, the sets $\dx{}(Q)$ and $\dnx{}(Q)$ can be interpreted as those leading diagnoses that predict the classification of $Q$ as a positive and negative test case, respectively. In other words, if the true diagnosis $\dt$ is in $\dx{}(Q)$, then the true solution KB $\mo^{*}_t$ entails $Q$ by Definition~\ref{def:q-partition}. Therefore the user will answer $Q$ positively (cf.\ Definition~\ref{def:true_diagnosis}). If, conversely, $\dt$ is in $\dnx{}(Q)$, then the true solution KB $\mo^{*}_t$ would be invalidated if $Q$ was answered positively, since $\mo^{*}_t \cup Q = (\mo\setminus\dt)\cup\mb\cup U_{\Tp\cup\setof{Q}}$ violates some $x\in\RQ\cup\Tn$ and thus $\mo\setminus\dt$ is invalid w.r.t.\ $\tuple{\cdot,\mb,\Tp\cup\setof{Q},\Tn}_\RQ$, which implies that $\dt$ is not a diagnosis w.r.t.\ $\tuple{\mo,\mb,\Tp\cup\setof{Q},\Tn}_\RQ$ according to Proposition~\ref{prop:validonto_diag}. Hence, the user will answer $Q$ negatively (cf.\ Definition~\ref{def:true_diagnosis}).
Diagnoses in $\dz{}(Q)$ on the other hand neither predict $Q \in \Tp$ nor $Q \in \Tn$. This means that we do not know how the user will answer a query $Q$ for which the true diagnosis $\dt$ is in $\dz{}(Q)$. In this case, for any answer to $Q$, the true diagnosis $\dt$ is in the set of minimal diagnoses w.r.t.\ the new DPI including $Q$ as a test case. To summarize: If the true diagnosis $\dt$ is an element of $\dx{}(Q)$ ($\dnx{}(Q)$), then $Q$ will be answered positively (negatively). 

Conversely, this means that a q-partition $\Pt(Q)$ gives a prior indication which leading diagnoses would be invalidated by a user's answer. Diagnoses in $\dx{}(Q)$ are invalidated by the classification $Q \in \Tn$, and diagnoses in $\dnx{}(Q)$ in case of $Q \in \Tp$. Diagnoses in $\dz{}(Q)$ can never be invalidated by an answer to $Q$. Thus, intuitively, queries with $\dz{}(Q) = \emptyset$ are preferable over other queries (as per the information provided by the set of leading diagnoses $\mD$) as the number of (definitely) eliminated diagnoses in $\minD_{\langle\mo,\mb,\Tp,\Tn\rangle_\RQ}$ should be maximized.

The following proposition is a direct consequence of Corollary~\ref{cor:notions_equiv} and explicates the impact of the addition of a test case to a DPI regarding the set of minimal diagnoses for this DPI.
\begin{proposition}\label{prop:dpi_update}
Let $Q$ be a query w.r.t. $\mD \subseteq \minD_{\langle\mo,\mb,\Tp,\Tn\rangle_\RQ}$ and let the answer of a user to $Q$ be $u(Q) \in \setof{\true,\false}$.

If $u(Q) = \true$, then $\md_i \in \minD_{\langle\mo,\mb,\Tp,\Tn\rangle_\RQ}$ is a diagnosis w.r.t.\ $\langle\mo,\mb,\Tp\cup\setof{Q},\Tn\rangle_\RQ$ iff $\mo\setminus \md_i$ is valid w.r.t.\ $\langle\cdot,\mb,\Tp\cup\setof{Q},\Tn\rangle_\RQ$. 

In other words, both of the following conditions must hold:
\begin{align*}
\forall r \in \RQ & \;:\; \mo_i^* \cup Q \; \emph{does not violate } r\\
\forall \tn \in \Tn & \;:\; \mo_i^* \cup Q \not\models \tn
\end{align*}

If $u(Q) = \false$, then $\md_i \in \minD_{\langle\mo,\mb,\Tp,\Tn\rangle_\RQ}$ is a diagnosis w.r.t.\ $\langle\mo,\mb,\Tp,\Tn\cup\setof{Q}\rangle_\RQ$ iff $\mo\setminus \md_i$ is valid w.r.t.\ $\langle\cdot,\mb,\Tp,\Tn\cup\setof{Q}\rangle_\RQ$. 

In other words, both of the following conditions must hold:
\begin{align*}
\forall r \in \RQ & \;:\; \mo_i^* \; \emph{does not violate } r\\
\forall \tn \in (\Tn \cup \setof{Q}) & \;:\; \mo_i^* \not\models \tn
\end{align*}
\end{proposition}
\begin{remark}\label{rem:invalidated_sets_of_q-partition_for_query_answer}
From Proposition~\ref{prop:dpi_update} and Definition~\ref{def:q-partition} it is easy to see that at least $\md_i \in \dnx{}(Q) \subset \minD_{\langle\mo,\mb,\Tp,\Tn\rangle_\RQ}$ are eliminated by a positive answer to $Q$. Namely, $\dnx{}(Q)$
comprises exactly those diagnoses $\md_i$ that imply the violation of some $r\in\RQ$ or the entailment of some $\tn\in\Tn$ if $Q$ is added to $\mo_i^*$. On the other hand, at least $\md_i \in \dx{}(Q) \subset \minD_{\langle\mo,\mb,\Tp,\Tn\rangle_\RQ}$ are discarded if $u(Q) = \false$ as all diagnoses in $\dx{}(Q)$ entail $Q$ which must not be entailed. 

Note that, in general, the addition of a query to the test cases of a DPI causes not only an invalidation of some leading minimal diagnoses in $\mD$, but also the elimination of minimal diagnoses that have not even been computed yet. On the other hand, an added test case might also introduce new \emph{minimal} diagnoses, i.e.\ ones that were no minimal diagnoses before this test case was added. However, the newly obtained DPI after the addition of any new test case can only exhibit a reduced set of \emph{all} (i.e.\ minimal and non-minimal) diagnoses compared with the DPI before the test case was added. 
\qed
\end{remark}

\subsection{The Relation between a Query and its Q-Partition}
\label{sec:TheRelationBetweenAQueryAndItsQPartition}
The following proposition shows the relationship between a query and its q-partition and provides a criterion that enables to check whether a set of logical formulas is a query w.r.t.\ some set of leading diagnoses or not.
\begin{proposition}\label{prop:query_dx_dnx}
Let $\langle\mo,\mb,\Tp,\Tn\rangle_\RQ$ be a DPI over $\mathcal{L}$ and $\mD \subseteq \minD_{\langle\mo,\mb,\Tp,\Tn\rangle_\RQ}$. Then a set of logical formulas $Q\neq\emptyset$ over $\mathcal{L}$ is a query w.r.t. $\mD$ iff $\dx{}(Q) \neq \emptyset$ and $\dnx{}(Q)~\neq~\emptyset$.
\end{proposition}
\begin{proof}
``$\Leftarrow$'': If $\dx{}(Q) \neq \emptyset$ and $\dnx{}(Q) \neq \emptyset$ holds, then a non-empty set of diagnoses $\dnx{}(Q)$ ($\dx{}(Q)$) becomes invalid for positive (negative) answer to $Q$. So, $Q$ is a query.

``$\Rightarrow$'': If $Q$ is a query, then there are diagnoses $\md, \md' \in \mD$ such that $\md \notin \minD_{\langle\mo,\mb,\Tp \cup \setof{Q},\Tn\rangle_\RQ}$ and $\md'\notin\minD_{\langle\mo,\mb,\Tp,\Tn \cup \setof{Q}\rangle_\RQ}$. Consequently, $\md \in \mD \setminus \minD_{\langle\mo,\mb,\Tp \cup \setof{Q},\Tn\rangle_\RQ}$ and $\md' \in \mD \setminus \minD_{\langle\mo,\mb,\Tp,\Tn \cup \setof{Q}\rangle_\RQ}$ holds. But, as the diagnoses in $\mD \setminus \minD_{\langle\mo,\mb,\Tp \cup \setof{Q},\Tn\rangle_\RQ}$ are exactly the diagnoses in $\mD$ that become invalid by the positive answer to $Q$, we obtain $\md\in\dnx{}(Q)$. The argumentation for $\md' \in \dx{}(Q)$ is analogous. Hence, $\dx{}(Q) \neq \emptyset$ and $\dnx{}(Q) \neq \emptyset$.  
\end{proof}
\begin{corollary}\label{cor:q-partition_dx_dnx}
Let $\mD\subseteq \minD_{\langle\mo,\mb,\Tp,\Tn\rangle_\RQ}$. Then, for each q-partition $\Pt(Q) = \langle \dx{}(Q), \dnx{}(Q), \dz{}(Q)\rangle$ w.r.t. $\mD$ it holds that $\dx{}(Q) \neq \emptyset$ and $\dnx{}(Q)~\neq~\emptyset$.
\end{corollary}
\begin{proof}
Follows from Definition~\ref{def:q-partition} which grants the existence of a query for any q-partition and Proposition~\ref{prop:query_dx_dnx} which states that neither $\dx{}(Q)$ nor $\dnx{}(Q)$ must be empty sets for any query.
\end{proof}
So, by Proposition~\ref{prop:query_dx_dnx}, a query not only eliminates at least one leading diagnosis, but also leaves at least one leading diagnosis valid. Therefore, an admissible DPI can never get non-admissible by adding a query to the positive or negative test cases.
\begin{corollary}\label{cor:query_leaves_valid_diag}
Let $\langle\mo,\mb,\Tp,\Tn\rangle_\RQ$ be an admissible DPI, $\mD \subseteq \minD_{\langle\mo,\mb,\Tp,\Tn\rangle_\RQ}$ and $Q\in\mQ_{\mD,\langle\mo,\mb,\Tp,\Tn\rangle_\RQ}$. Then $\langle\mo,\mb,\Tp\cup\setof{Q},\Tn\rangle_\RQ$ as well as $\langle\mo,\mb,\Tp,\Tn\cup\setof{Q}\rangle_\RQ$ are admissible DPIs.
\end{corollary}
\begin{proof}
Assume that $\langle\mo,\mb,\Tp\cup\setof{Q},\Tn\rangle_\RQ$ is non-admissible. Then there is no valid diagnosis for this DPI. Since $\langle\mo,\mb,\Tp,\Tn\rangle_\RQ$ is an admissible DPI, this means that $Q$ invalidates each diagnosis $\md\in\allD_{\langle\mo,\mb,\Tp,\Tn\rangle_\RQ} \supseteq \minD_{\langle\mo,\mb,\Tp,\Tn\rangle_\RQ}\supset\mD$. By Proposition~\ref{prop:query_dx_dnx}, this is a contradiction to the fact that $Q$ is a query. The argumentation for $\langle\mo,\mb,\Tp,\Tn\cup\setof{Q}\rangle_\RQ$ is analogue.
\end{proof}
This means in particular that a query can never contain a conflict set or result in a violation of some requirement $r\in\RQ$ when added to $\mb\cup U_\Tp$ (cf. Proposition~\ref{prop:exist_diag}).

\subsection{Existence of Queries}
\label{sec:ExistenceOfQueries}
For any set of at least two leading minimal diagnoses the existence of a query is guaranteed, as the next proposition and corollary show. 
In particular, this implies that for arbitrary two minimal diagnoses $\md,\md'$ w.r.t. a DPI there is a query $Q$ that enables to differentiate between $\md$ and $\md'$, i.e.\ exactly one of these diagnoses is invalidated by each answer to $Q$.
\begin{proposition}\label{prop:q1}
Let $\mD \subseteq \minD_{\langle\mo,\mb,\Tp,\Tn\rangle_\RQ}$ with $|\mD|\geq 2$ and $U_\mD$ be the union of all diagnoses in $\mD$. Then 
\begin{enumerate}[(I)]
	\item $Q:=(U_{\mD}\setminus \md_i)$ is a query w.r.t. $\mD$ for arbitrary $\md_i\in\mD$ and
	\item $\Pt(Q)=\langle\setof{\md_i},\mD\setminus\setof{\md_i},\emptyset\rangle$.
\end{enumerate}
\end{proposition}
\begin{proof}
\textbf{Ad (I):} Assume that $Q$ is not a query. Then either (1)~$Q = \emptyset$ or (2)~$\dx{}(Q) = \emptyset$ or (3)~$\dnx{}(Q) = \emptyset$. In the following we prove that neither (1) nor (2) nor (3) can hold.

(1): $Q = \emptyset$ means that $\md_i \supseteq U_{\mD}$. Since any diagnosis $\md$ in $\mD$ is a subset of $U_{\mD}$, this implies that for each $\md \in \mD$, $\md \subseteq \md_i$ holds. As $|\mD| \geq 2$ is assumed, there is a $\md_k \neq \md_i \in \mD$ for which this property holds. This, however, is a contradiction to the 
minimality of diagnosis $\md_i$.

(2): $\dx{}(Q) = \emptyset$ cannot hold, since $(\mo \setminus \md_i) \supseteq (U_{\mD}\setminus \md_i)$ and $U_{\mD}\setminus \md_i \models Q$ by monotonicity of description logics imply that $\mo^{*}_i = (\mo \setminus \md_i) \cup \mb \cup U_P \models Q$. Hence, there is at least one diagnosis, namely $\md_i$, in $\dx{}(Q)$.

(3): To prove that $\dnx{}(Q) \neq \emptyset$, we must show that there is a diagnosis $\md\in\mD$ such that $Y:=(\mo\setminus\md) \cup \mb \cup U_P \cup Q = (\mo\setminus\md) \cup \mb \cup U_P \cup (U_{\mD}\setminus \md_i)$ is incoherent. 
However, $(\mo\setminus\md) \cup (U_{\mD}\setminus \md_i) = \mo \setminus (\md\cap\md_i)$ by distributive and De Morgan laws which yields $Y = \mo \setminus (\md\cap\md_i)\cup \mb \cup U_P$.
But, $\md\cap\md_i \subset \md$ must hold as $\md \not\subseteq \md_i$ by the subset-minimality of $\md_i$ whereby $\md$ must comprise a formula $\tax \notin \md_i$. Hence, $Y \supset (\mo\setminus\md) \cup \mb \cup U_P$ is incoherent by subset-minimality of $\md$.

\textbf{Ad (II):} We already know that $\md_i \in \dx{}(Q)$ by (2). Since $\md\in\mD$ in (3) can be chosen arbitrarily, we obtain that $\md\in\dnx{}(Q)$ for all diagnoses $\md\in\mD\setminus\setof{\md_i}$.
\end{proof}
We immediately obtain a lower bound for the number of queries by Proposition~\ref{prop:q1}:
\begin{corollary}\label{cor:query_num_lower_bound}
Let $\mD \subseteq \minD_{\langle\mo,\mb,\Tp,\Tn\rangle_\RQ}$ with $|\mD|>1$. Then a lower bound for the number of queries w.r.t. $\mD$ is $|\mD|$.
\end{corollary}
\begin{remark}\label{rem:existence_of_queries_requires_minimal_leading_diagnoses}
Notice that the preceding proposition and corollary require a set of \emph{minimal} diagnoses. This means that subset-minimality of diagnoses is a necessary prerequisite for guaranteeing the possibility of discrimination between diagnoses. In other words, interactive debugging by means of (some or only) non-minimal diagnoses cannot be proven to work correctly (without making any further assumptions).\qed
\end{remark} 

\section{Query Generation}
\label{sec:QueryGeneration}
In this section, we want to describe, discuss and prove the correctness of methods for the generation of queries which takes place at each iteration of an interactive KB debugging algorithm after a set of leading diagnoses has been determined.
With Algorithm~\ref{algo:query_gen}, similar versions of which can be found in~\cite{Shchekotykhin2012, Rodler2013}, we present a way to compute a pool $\QP$ of queries and associated q-partitions w.r.t.\ a set of leading diagnoses $\mD$ and a DPI $\tuple{\mo,\mb,\Tp,\Tn}_\RQ$. The generation of this pool $\QP$ is the first stage of the query computation function used in the interactive debugging algorithm (Algorithm~\ref{algo:inter_onto_debug}) presented below. In a second stage, one particular query that meets certain criteria such as maximum expected information gain is selected from $\QP$ (see Section~\ref{sec:query_selection_measures}).

Before we give a description of Algorithm~\ref{algo:query_gen}, let us have a look at some example by which we want to demonstrate the principle how a query w.r.t.\ some set of leading diagnoses for a DPI can be constructed. This should give the reader a first idea and an intuition of how the presented algorithm works.

\renewcommand{\arraystretch}{1.4} 
\begin{table*}
\footnotesize
	\centering
		\rowcolors[]{2}{gray!8}{gray!16} 
		\begin{tabular}{ c c c c	} 
			\rowcolor{gray!40}
			\toprule\addlinespace[0pt]
			$i$ & $\tax_i$ & $\mo$ & $\mb$  \\ \addlinespace[0pt]\midrule\addlinespace[0pt]
			1 & $\forall X a_1(X) \;\rightarrow\; a_2(X) \land m_1(X) \land m_2(X)$ & $\bullet$ & 	\\
			2 & $\forall X a_2(X) \;\rightarrow\; \lnot(\exists Y s(X,Y) \land m_3(Y)) \land \exists Z s(X,Z) \land m_2(Z)$ & $\bullet$ &  	\\
			3 & $\forall X m_1(X) \;\rightarrow\; \lnot a(X) \land b(X)$ & $\bullet$ &  	\\
			4 & $\forall X m_2(X) \;\rightarrow\; (\forall Y s(X,Y) \rightarrow a(Y)) \land d(X)$ & $\bullet$ & 	\\
			5 & $\forall X m_3(X) \;\leftrightarrow\; b(X) \lor c(X)$ & $\bullet$ & 	\\
			6 & $a_1(w)$ &  & $\bullet$   \\
			7 & $a_1(u)$ &  & $\bullet$  	\\
			8 & $s(u,w)$ &  & $\bullet$  \\ 
			\addlinespace[0pt]\bottomrule
			\rowcolor{gray!40}
			$i$ & \multicolumn{3}{c}{$\tp_i\in\Tp$} \\ \addlinespace[0pt]\midrule\addlinespace[0pt]
			$\times$ & \multicolumn{3}{c}{$\times$} 	\\ \addlinespace[0pt]\toprule\addlinespace[0pt]
			\rowcolor{gray!40}
			$i$ & \multicolumn{3}{c}{$\tn_i\in\Tn$} \\ \addlinespace[0pt]\midrule\addlinespace[0pt]
			$\times$ & \multicolumn{3}{c}{$\times$} 	\\ \addlinespace[0pt]\toprule\addlinespace[0pt]
			\rowcolor{gray!40}
			$i$ & \multicolumn{3}{c}{$r_i\in\RQ$} \\ \addlinespace[0pt]\midrule\addlinespace[0pt]
			1 & \multicolumn{3}{c}{consistency} \\ 
			2 & \multicolumn{3}{c}{coherency} \\ \addlinespace[0pt]\bottomrule
		\end{tabular}
	\caption{First-Order Logic Example DPI}
	\label{tab:example1}
\end{table*}

\begin{example}\label{example:query_computation}
Consider the example FOL DPI given by Table~\ref{tab:example1}. The set of minimal conflict sets $\minC_{\tuple{\mo,\mb,\Tp,\Tn}_\RQ} = \setof{\mc_1,\mc_2} = \setof{\tuple{1,3,4},\tuple{1,2,3,5}}$ (like in previous examples, formulas $\tax_i$ in Table~\ref{tab:example1} are sometimes referred to just by their number $i$ if it is clear from the context what is meant). Let the set of leading diagnoses be the set of all minimal diagnoses, i.e.\ $\mD = \minD_{\tuple{\mo,\mb,\Tp,\Tn}_\RQ} = \setof{\md_1,\md_2,\md_3,\md_4} = \setof{[1],[3],[4,5],[2,4]}$. To enable a better understanding of this example, we first analyze why $\mc_1$ and $\mc_2$ are minimal conflict sets w.r.t.\ $\tuple{\mo,\mb,\Tp,\Tn}_\RQ$.

Why is $\mc_1$ a conflict set w.r.t.\ $\tuple{\mo,\mb,\Tp,\Tn}_\RQ$? In the following we underline the formulas $\tax_i$ and relevant parts of these formulas used in the derivation of the conflict set. First, there is the background KB $\mb$ including $\underline{a_1(w)}$ and $a_1(u)$. Due to $\underline{\tax_1}$, by substitution of $X$ by $w$ (written as $X/w$), we obtain $a_2(w), \underline{m_1(w)}$ and $m_2(w)$ from $a_1(w)$. Likewise, we can derive $a_2(u), m_1(u)$ and $\underline{m_2(u)}$ from $a_1(u)$ by $X/u$. Substituting $X$ by $w$ in $\underline{\tax_3}$ yields $\underline{m_1(w) \rightarrow \lnot a(w)} \land b(w)$. Thus, we obtain $\underline{\lnot a(w)}$. A substitution of $X$ by $u$ in $\underline{\tax_4}$ results in $m_2(u) \rightarrow (\forall Y s(u,Y) \rightarrow a(Y)) \land d(u)$. By $Y/w$, we have $\underline{m_2(u) \rightarrow (s(u,w) \rightarrow a(w))} \land d(u)$. Since $m_2(u)$ has already been deduced from the background formula $a_1(u)$ and $\underline{s(u,w)}$ is a background formula as well, we can conclude $\underline{a(w)}$ from $\tax_4$. All in all, we have derived $\lnot a(w)$ and $a(w)$, i.e.\ an inconsistency, by means of $\mb$ and $\mc_1$ (and $U_\Tp$ which is the empty set) wherefore $\mc_1$ is a conflict set w.r.t.\ $\tuple{\mo,\mb,\Tp,\Tn}_\RQ$ by Definition~\ref{def:cs}. The minimality of $\mc_1$ can be easily verified by the way we derived that it is a conflict set; namely, leaving out any of the formulas $\tax_1$, $\tax_3$ or $\tax_4$ does not allow to derive an inconsistency or incoherency (note that the set of negative test cases $\Tn$ is empty). 

Why is $\mc_2$ a conflict set w.r.t.\ $\tuple{\mo,\mb,\Tp,\Tn}_\RQ$? We argue as follows to deduce the inconsistency responsible for $\mc_2$ to be a conflict set (the relevant implications and used formulas are again underlined):
\begin{align*}
(1):\,a_1(w) \in \mb:&\quad \underline{a_1(w)}  \\
(2):\,X/w \mbox{ in } \underline{\tax_1}:&\quad  \underline{a_1(w) \;\rightarrow}\; a_2(w) \land \underline{m_1(w)} \land m_2(w) \\
(3):\,X/w \mbox{ in } \underline{\tax_3}:&\quad  \underline{m_1(w) \;\rightarrow}\; \lnot a(w) \land \underline{b(w)} \\
(4):\,\underline{\tax_5} \mbox{ and } X/w:& \quad \underline{b(w) \;\rightarrow\; m_3(w)} \\
(5):\,(1) - (4):& \quad \underline{m_3(w)} \\
(6):\,a_1(u) \in \mb:&\quad \underline{a_1(u)}  \\
(7):\,X/u \mbox{ in } \underline{\tax_1}:&\quad  \underline{a_1(u) \;\rightarrow\; a_2(u)} \land m_1(u) \land m_2(u) \\
(8):\,X/u \mbox{ in } \underline{\tax_2}:&\quad  \underline{a_2(u) \;\rightarrow\; \lnot (\exists Y s(u,Y) \land m_3(Y))} \\ 											 &\quad	\land (\exists Z s(u,Z) \land m_2(Z))  \\
(9):\,(6) - (8):& \quad \underline{\lnot (\exists Y s(u,Y) \land m_3(Y))} \\
(10):\,s(u,w) \in \mb:&\quad \underline{s(u,w)}  \\
(11):\,(5)\mbox{ and } (10):&\quad 	\underline{\exists Y s(u,Y) \land m_3(Y)}	\\
(9)\mbox{ and } (11):&\quad \mbox{\Lightning} \quad\qed
\end{align*}
Minimality of $\mc_2$ can again be verified by observing that, given any formula of $\mc_2$ is left out, no inconsistency or incoherency can be derived.

Now we show how to construct a query manually. As suggested by Definition~\ref{def:q-partition} and Proposition~\ref{prop:query_dx_dnx} and discussed in Section~\ref{sec:TheRelationBetweenAQueryAndItsQPartition},
an obvious way of generating a query w.r.t.\ $\mD$ and $\tuple{\mo,\mb,\Tp,\Tn}_\RQ$ is via the notion of a q-partition. Definition~\ref{def:q-partition} states that $Q$ is a set of common entailments of KBs $\mo_i^*$ (Formula~\ref{eq:sol_ont_candidate}) where $\md_i \in \dx{}(Q)$, a subset of $\mD$. Hence, a first step towards query computation is to choose some non-empty subset $\mS$ of the leading diagnoses $\mD$ which we will call the \emph{seed} for query generation. For our manual construction, let $\mS = \setof{\md_3,\md_4} = \setof{[4,5],[2,4]}$. For each of the diagnoses $\md_i$ in $\mS$, we assemble the KB $\mo_i^*$ and use a reasoning engine to obtain a set of entailments $E_{\md_i}$ of $\mo_i^*$. For $\md_3$ we obtain $\mo_3^* := \setof{1,2,3,4,5}\setminus\setof{4,5} \cup \setof{6,7,8} \cup \setof{} = \setof{1,2,3,6,7,8}$. Similarly, we compute $\mo_4^* = \setof{1,3,5,6,7,8}$. 

Suppose that the reasoner invoked by the used \textsc{getEntailments} function produces only entailments of the type $\forall X p_1(X) \rightarrow p_2(X)$ for predicate names $p_1, p_2$ and of the type $p(a)$ where $p$ is a predicate name and $a$ is a constant (cf.\ Remark~\ref{rem:entailment_computation_finite_types_of_entailments}). For this purpose, DL and OWL reasoners, respectively, such as Pellet \cite{sirin2007pellet}, HermiT \cite{Shearer2008}, FaCT++ \cite{Tsarkov06} or KAON2\footnote{http://kaon2.semanticweb.org/} could be used with their classification and realization reasoning services. The reason why this is possible can be realized after a short analysis of the DPI $\tuple{\mo,\mb,\Tp,\Tn}_\RQ$ given by Table~\ref{tab:example1}. For, this DPI can be translated to DL similarly as demonstrated in Example~\ref{example:FOL_to_DL}. All the mentioned reasoners can deal with the expressivity of the resulting DL language.

Then, we obtain the sets $E_{\md_3}$ and $E_{\md_4}$, i.e.\ the sets of entailments of $\mo_3^*$ and $\mo_4^*$, respectively, as depicted by Table~\ref{tab:example:query_construction_entailments}. The set of common entailments $Q$, i.e.\ $Q = E_{\md_3} \cap E_{\md_4}$ is then the set containing all elements in the rows of Table~\ref{tab:example:query_construction_entailments} that are above the dashed line. 

Notice at this point that the set $\setof{a_1(w),a_1(u),s(u,w)} = \mb$ does not need to be computed or, respectively, included in $Q$ since none of these formulas can serve to discriminate between diagnoses (which is the only aim of a query). The simple reason for this is that $\mo_i^*$ for \emph{each} $\md_i \in \mD$ comprises these formulas and thus each $\mo_i^*$ entails these formulas by the extensiveness of FOL (cf.\ Chapter~\ref{chap:basics}). Since entailed by each potential solution KB $\mo_i^*$, these formulas cannot yield a violation of any requirements or test cases since none of the KBs $\mo_i^*$ violates any requirements or test cases (follows from Definitions~\ref{def:diagnosis} and \ref{def:target_ont}).   

Continuing with our query construction, we know by Proposition~\ref{prop:query_dx_dnx} that $Q$ is a query w.r.t.\ $\mD$ and $\tuple{\mo,\mb,\Tp,\Tn}_\RQ$ iff $\dx{}(Q) \neq \emptyset$ and $\dnx{}(Q) \neq \emptyset$. Whereas it is trivial that the former condition is met since $\dx{}(Q)$ contains (at least) the two diagnoses $\md_3$ and $\md_4$ that we used to compute $Q$ (cf.\ Definition~\ref{def:q-partition}), we still need to verify whether the latter condition is actually satisfied for $Q$. To this end, as per Definition~\ref{def:q-partition}, we must simply find some diagnosis $\md_j$ in $\mD \setminus \mS = \setof{\md_1,\md_2,\md_3,\md_4} \setminus \setof{\md_3,\md_4} = \setof{\md_1,\md_2}$ such that $\mo_j^* \cup Q$ violates some $x\in\Tn\cup\RQ$, i.e.\ whether some negative test case is entailed or whether this KB is incoherent or inconsistent. So, we start with $\md_1$, i.e.\ we examine $(\mo\setminus\md_1) \cup \mb \cup \Tp \cup Q = \setof{1,2,3,4,5}\setminus \setof{1} \cup \setof{6,7,8} \cup \setof{} \cup Q = \setof{2,3,4,5,6,7,8} \cup Q$. 

And, indeed, we are able to prove an inconsistency for this KB. To see that, verify that by $X/w$ in $e_2\in Q$ (see Table~\ref{tab:example:query_construction_entailments}) and $a_1(w) = \tax_6 \in \mo_1^*$ we can derive $m_1(w)$ which lets us conclude $\lnot a(w)$ by the substitution of $X$ by $w$ in $\tax_3\in \mo_1^*$. On the other hand, we obtain $a(w)$ by $X/u$ in $e_3 \in Q$, $\setof{X/u,Y/w}$ in $\tax_4 \in \mo_1^*$ and $s(u,w) = \tax_8 \in \mo_1^*$ as shown in the explanation for conflict set $\mc_1$ above. Thus, $\md_1 \in \dnx{}(Q)$.

That is, we have just proven that $Q$ is de facto a query w.r.t.\ $\mD$ and $\tuple{\mo,\mb,\Tp,\Tn}_\RQ$. And this, although we have not yet assigned each leading diagnosis to the respective set of the q-partition of $Q$. In a situation where just any query shall be asked to the user, this would suffice, and the query could be presented to the interacting user.

However, in case a ``best'' query according to some criterion shall be determined from a set of different competing queries, usually the computation of the full q-partition of each competing query is required. This is due to the fact that the q-partition provides information about several properties of queries that are considered by common query selection techniques (for details see Section~\ref{sec:query_selection_measures}). So, let us complete the q-partition for our query $Q$ by investigating $\mo_2^* \cup Q = \setof{1,2,4,5,6,7,8} \cup Q$. Also in this case we can derive an inconsistency which can be easily realized by reconsidering the argumentation why $\mc_2$ is a conflict set above and by using $e_4 \in Q$ instead of $\tax_3 \notin \mo_2^* \cup Q$. That means, the final q-partition $\Pt(Q)$ for $Q$ is given by $\tuple{\setof{\md_3,\md_4},\setof{\md_1,\md_2},\emptyset}$.

The next question that arises directly from the proofs that $\md_3, \md_4 \in \dnx{}(Q)$ is whether there is a (set-minimal) subset $Q_{\min}$ of $Q$ such that $Q_{\min}$ preserves the discrimination properties of $Q$, i.e.\ the q-partition $\Pt(Q_{\min}) = \Pt(Q)$. In fact, the answer is yes for the query $Q$ we computed, but also for the majority of other cases. This is a simple consequence of using the reasoning engine as a black-box which suggests a strategy we pursued in our query construction which relies on a precomputation of entailments and a final minimization part. Sticking to this black-box concept however does not allow to use some customized reasoning procedure that pointedly returns a set of common entailments $Q$ for a set of diagnoses $\mS \subset \mD$ where all formulas in $Q$ are necessary for a requirement or test case violation, respectively, of KBs $\mo_j^*$ for diagnoses in $\mD \setminus \mS$.

What militates for such a black-box approach is the generality and independence of a particular logic (for which an adequate glass-box reasoner exists), the easier implementation of the debugging system and potential performance issues with a glass-box approach~\cite{kalyanpur2005}. For a black-box algorithm to work, only a reasoner implementing a sound and complete inference procedure for the used logic $\mathcal{L}$ must be available.

In general, there is more than one minimized version of a query that preserves the q-partition. Theoretically, the number of such minimal queries w.r.t.\ one q-partition can be exponential in the size of the initially computed query that is provided as an input to the minimization procedure. For our query $Q$, for instance, 
\begin{align*}
Q_{\min,1}=\{&a_2(u),b(w)\} = \{e_7,e_{12}\}, \\
Q_{\min,2}=\{&\forall X a_1(X)\rightarrow a_2(X),b(w)\} = \setof{e_1,e_{12}}, \\
Q_{\min,3}=\{ &\forall X a_1(X)\rightarrow a_2(X),\\
					    &\forall X a_1(X)\rightarrow m_1(X), \\ 					
					    &\forall X m_1(X)\rightarrow b(X)\}  
					= \{e_1,e_2,e_4\} \quad\mbox{   and} \\ 
Q_{\min,4}= \{& \forall X a_1(X)\rightarrow m_1(X), \\
									 &\forall X a_1(X)\rightarrow m_2(X), \\				 
									 & \forall X m_1(X)\rightarrow b(X)\} 
									= \{e_2,e_3,e_4\} 
\end{align*}
are set-minimal, q-partition preserving subqueries. Namely, each of the sets $Q_{\min,1}$, $Q_{\min,2}$ and $Q_{\min,3}$ together with $\setof{2,5,6,7,8}$ implies an inconsistency since $m_3(w)$ and $\lnot m_3(w)$ can be derived and $\setof{2,5,6,7,8}\subseteq \mo_1^*$ and $\setof{2,5,6,7,8}\subseteq \mo_2^*$. $\setof{e_2,e_3} \subset Q_{\min,4}$ yields an inconsistency when added to $\mo_1^*$, i.e.\ $a(w)$ and $\lnot a(w)$ are entailed, and $\setof{e_4} \subset Q_{\min,4}$ merged with $\mo_2^*$ yields an inconsistency, i.e.\ the derivation of $m_3(w)$ and $\lnot m_3(w)$. In order not to overwhelm the user we would of course ask them such a minimized version of a query rather than the full query that contains plenty of irrelevant formulas.

An example of a seed $\mS$ that does not lead to the discovery of a query is $\mS = \setof{\md_1,\md_2,\md_3}$ since the set of common entailments $E_{\md_1}\cap E_{\md_2}\cap E_{\md_3} = \emptyset$. Note that this holds when all $E_{\md_i}$ contain only entailments of the types we specified above. For other types of entailments, i.e.\ a different specification of the \textsc{getEntailments} function, this might no longer hold.
%
%
\qed
\end{example}

\renewcommand{\arraystretch}{1.4} 
\begin{table}
\small
	\centering
		\rowcolors[]{2}{gray!8}{gray!16} 
		\begin{tabular}{ c | c | c } 
			\rowcolor{gray!40}
			\toprule\addlinespace[0pt]
			   & $E_{\md_3}$ & $E_{\md_4}$   \\ \addlinespace[0pt]\midrule\addlinespace[0pt]
			$e_1$ & $\forall X a_1(X)\rightarrow a_2(X)$  &  $\forall X a_1(X)\rightarrow a_2(X)$\\ 
			$e_2$ & $\forall X a_1(X)\rightarrow m_1(X)$  &  $\forall X a_1(X)\rightarrow m_1(X)$ \\
			$e_3$ & $\forall X a_1(X)\rightarrow m_2(X)$  &  $\forall X a_1(X)\rightarrow m_2(X)$ \\
			$e_4$ & $\forall X m_1(X)\rightarrow b(X)$    &  $\forall X m_1(X)\rightarrow b(X)$ \\
			$e_5$ & $\forall X a_1(X) \rightarrow b(X)$   &  $\forall X a_1(X) \rightarrow b(X)$\\
			$e_6$ & $a_2(w)$  &  $a_2(w)$ \\
			$e_7$ &  $a_2(u)$ &  $a_2(u)$\\
			$e_8$ &  $m_1(w)$ &  $m_1(w)$\\
			$e_9$ &  $m_1(u)$ &  $m_1(u)$\\
			$e_{10}$ &  $m_2(w)$ &  $m_2(w)$\\
			$e_{11}$ &  $m_2(u)$ &  $m_2(u)$ \\
			$e_{12}$ &  $b(w)$   &  $b(w)$\\
			$e_{13}$ &  $b(u)$   &  $b(u)$\\
			\hdashline
			$e_{14}$ &  & $\forall X b(X) \rightarrow m_3(X)$ \\
			$e_{15}$ &  & $\forall X c(X) \rightarrow m_3(X)$ \\
			$e_{16}$ &  & $\forall X m_1(X) \rightarrow m_3(X)$ \\
			$e_{17}$ &  & $\forall X a_1(X) \rightarrow m_3(X)$ \\
			$e_{18}$ &  & $m_3(w)$\\
			$e_{19}$ &  & $m_3(u)$ \\
			\addlinespace[0pt]\bottomrule
		\end{tabular}
	\caption[(Example~\ref{example:query_computation}) Computing Entailments for Query Generation]{(Example~\ref{example:query_computation}) Entailments computed for KBs $\mo_3^*$ and $\mo_4^*$.}
	\label{tab:example:query_construction_entailments}
\end{table}

\begin{algorithm*}\label{algo:query_gen}
\small
\caption{Generation of Queries and Q-Partitions \normalsize} \label{algo:query_gen}
\begin{algorithmic}[1]
\Require an admissible DPI $\tuple{\mo,\mb,\Tp,\Tn}_\RQ$, a set of minimal diagnoses $\mD \subseteq \minD_{\tuple{\mo,\mb,\Tp,\Tn}_\RQ}$ such that $|\mD| \geq 2$, a desired number $q \in \mathbb{N} \cup \setof{\infty}, q \geq 1$ of queries w.r.t.\ $\tuple{\mo,\mb,\Tp,\Tn}_\RQ$ to be returned
\Ensure a set $\QP$ including tuples $\tuple{Q,\tuple{\dx{}(Q),\dnx{}(Q),\dz{}(Q)}}$ such that: 
If $q \geq |\QP_{\max}|$, 
then 
\begin{enumerate}
\item there are no two tuples $\tuple{Q,\Pt(Q)}, \tuple{Q',\Pt(Q')}$ in $\QP$ such that $Q = Q'$ \emph{or} $\Pt(Q)=\Pt(Q')$, and
\item $\QP$ includes a tuple $\tuple{Q,\tuple{\dx{}(Q),\dnx{}(Q),\dz{}(Q)}}$ only if $Q\in\mQ_{\mD,\tuple{\mo,\mb,\Tp,\Tn}_\RQ}$, and  
\item $\QP$ includes at most one tuple where $\dx{}(Q) = Y$ for each $Y \subset \mD$, and 
\item for each $Y \subset \mD$ for which a query $Q$ w.r.t.\ $\mD$ and $\tuple{\mo,\mb,\Tp,\Tn}_\RQ$ exists such that 
	(a)~$Q$ includes only entailments computed by the used \textsc{getEntailments} function and
	(b)~$\Pt(Q)$ is such that $\dx{}(Q) = Y$,
$\QP$ includes a tuple $\tuple{Q',\Pt(Q')}$ such that $\dx{}(Q') = Y$, and
\item $\QP \neq \emptyset$.
\end{enumerate}
If $q < |\QP_{\max}|$, then $\QP$ includes $q$ tuples satisfying (1), (2) and (3). 
($|\QP_{\max}| \geq 0$ is the maximum number of tuples $\tuple{Q,\Pt(Q)}$ that can be computed by \textsc{getPoolOfQueries} by the used \textsc{getEntailments} function)

\vspace{10pt}

\Procedure{$\textsc{getPoolOfQueries}$}{$\langle\mo,\mb,\Tp,\Tn\rangle_\RQ, \mD, q$}
\State $E_\mD \gets \emptyset$
\For{$\md\in\mD$} \label{algoline:query:ent_start}
	\State $E_\md \gets \Call{getEntailments}{\md,\mo,\mb,\Tp}$     \Comment{$E_{\md_r}$ is the set of entailments of $\mo_r^*$}
	\State $E_\mD \gets E_\mD \cup \setof{\tuple{\md,E_\md}}$ \label{algoline:query:ent_end}
\EndFor
\For{$\emptyset\subset \mS \subset \mD$}  \label{algoline:query:seed}
	\State $isQuery \gets \false$
	\State $Q \leftarrow \Call{getCommonEntailments}{\mS, E_\mD}$  \label{algoline:query:common_ent}
	\If {$Q \neq \emptyset$} 
		\For{$\md_r \in \mD\setminus\mS$} 		\label{algoline:query:verify_CQ3_start}
			\If{$Q \subseteq E_{\md_r}$}					\Comment{Does $\mo^{*}_r \,\models Q$ ?}
					\State $\dx{} \leftarrow \dx{} \cup \left\{\md_r\right\}$     \label{algoline:query:dx}
			\ElsIf{$\lnot\Call{isKBValid}{\mo^{*}_r \cup Q, \tuple{\cdot,\emptyset,\emptyset,\Tn}_{\RQ}}$}  \label{algoline:query:is_ont_valid}  \Comment{\textsc{isKBValid} (see Algorithm~\ref{algo:qx})}
					\State $\dnx{} \leftarrow \dnx{} \cup \left\{\md_r\right\}$    \label{algoline:query:dnx}
					\State $isQuery \gets \true$    \label{algoline:query:is_query_true}
			\Else
					\State $\dz{} \leftarrow \dz{} \cup \left\{\md_r\right\}$     \label{algoline:query:dz}
			\EndIf
		\EndFor		\label{algoline:query:verify_CQ3_end}
		\If {$isQuery \land \lnot \Call{inclQPart}{\QP,\tuple{\dx{}, \dnx{}, \dz{}}}$} \label{algoline:query:add_QP_start}
				\State $Q' \gets \Call{minQ}{\emptyset,Q,\emptyset, \tuple{\dx{}, \dnx{}, \dz{}}, \tuple{\mo,\mb,\Tp,\Tn}_\RQ}$   \label{algoline:query:minQ}
				\State $\QP \leftarrow \QP \cup \setof{\tuple{Q', \tuple{\dx{}, \dnx{}, \dz{}}}}$  \label{algoline:query:add_QP_end}
				\If {$|\QP| = q$}					\label{algoline:query:test_QP}
						\State \Return $\QP$ \label{algoline:query:return_QP_1}
				\EndIf
		\EndIf
	\EndIf
\EndFor
\If{$|\QP| = 0$} \label{algoline:query:check_QP_empty}
	\State $\QP \gets \Call{addTrivialQueries}{\mD,\QP}$\label{algoline:query:addTrivialQueries}
\EndIf
\State \Return $\QP$
\EndProcedure

\vspace{10pt}

\Procedure{$\textsc{minQ}$}{$X,Q,QB,\tuple{\dx{}, \dnx{}, \dz{}},\langle\mo,\mb,\Tp,\Tn\rangle_\RQ$}
\If{$\ X \neq \emptyset \land \Call{isQPartConst}{QB,\tuple{\dx{}, \dnx{}, \dz{}}, \tuple{\mo,\mb,\Tp,\Tn}_\RQ}$}  \label{algoline:query:validitytest2}  
	\State \Return $\emptyset$    \label{algoline:query:return_emptyset}
\EndIf
\If{$|Q| = 1$}  \label{algoline:query:test_singleton}              
  \State \Return $Q$ \label{algoline:query:return_Q}
\EndIf
\State $k \gets \Call{split}{|Q|}$     \label{algoline:query:split}
\State $Q_1 \gets \Call{get}{Q, 1, k}$
\State $Q_2 \gets \Call{get}{Q, k + 1, |Q|}$
\State $Q^{\min}_2 \gets \Call{\textsc{minQ}}{Q_1,Q_2,QB\cup Q_1,\tuple{\dx{}, \dnx{}, \dz{}},\langle\mo,\mb,\Tp,\Tn\rangle_\RQ}$ \label{algoline:query:recursive_call1}
\State $Q^{\min}_1 \gets \Call{\textsc{minQ}}{Q^{\min}_2,Q_1,QB\cup Q^{\min}_2	,\tuple{\dx{}, \dnx{}, \dz{}},\langle\mo,\mb,\Tp,\Tn\rangle_\RQ}$ \label{algoline:query:recursive_call2}
\State \Return $Q^{\min}_1 \cup Q^{\min}_2$  
\EndProcedure

\vspace{10pt}

\Procedure{\textsc{isQPartConst}}{$Q,\tuple{\dx{}, \dnx{}, \dz{}}, \tuple{\mo,\mb,\Tp,\Tn}_\RQ$}
\For{$\md_r \in \dnx{}$} 
			\If{$\Call{isKBValid}{\mo^{*}_r \cup Q, \tuple{\cdot,\emptyset,\emptyset,\Tn}_{\RQ}}$} \Comment{\textsc{isKBValid} (see Algorithm~\ref{algo:qx})}
					\State \Return \false
			\EndIf
\EndFor
\For{$\md_r \in \dz{}$} 
			\If{$\mo^{*}_r \models Q$}
					\State \Return \false
			\EndIf
\EndFor
\State \Return \true
\EndProcedure
\end{algorithmic}
\normalsize
\end{algorithm*}
%
%
%

\subsection{Generation of a Pool of Queries}
\label{sec:GenerationOfAPoolOfQueries}
The main function \textsc{getPoolOfQueries} of Algorithm~\ref{algo:query_gen} gets as inputs an admissible DPI $\tuple{\mo,\mb,\Tp,\Tn}_\RQ$ over $\mathcal{L}$, a set of leading (minimal) diagnoses $\mD \subseteq \minD_{\tuple{\mo,\mb,\Tp,\Tn}_\RQ}$ such that $|\mD|\geq 2$ and a parameter $q \in \mathbb{N}\cup\setof{\infty}, q\geq 1$ that indicates the number of queries in $\mQ_{\mD,\tuple{\mo,\mb,\Tp,\Tn}_\RQ}$ the algorithm is supposed to return (where $q := \infty$ signalizes that a maximum number of queries should be output). The way of generating a pool of queries is guided by Proposition~\ref{prop:query_dx_dnx} which says that a non-empty set $Q$ of formulas over $\mathcal{L}$ is a query w.r.t.\ $\mD$ and $\tuple{\mo,\mb,\Tp,\Tn}_\RQ$ if and only if $\dx{}(Q)$ as well as $\dnx{}(Q)$ are non-empty sets of diagnoses. That is, the necessary and sufficient criteria for $Q$ to be a query are 
\begin{enumerate}[(CQ1)]
	\item $\quad Q \neq \emptyset$ and
	\item $\quad\dx{}(Q) \neq \emptyset$ and
	\item $\quad\dnx{}(Q)\neq \emptyset$.
\end{enumerate}
Note, since the disjoint sets of diagnoses $\dx{}(Q) \subseteq \mD$ and $\dnx{}(Q) \subseteq \mD$ must not be empty, $|\mD|\geq 2$ must be postulated in order for any queries to exist w.r.t.\ $\mD$ and $\tuple{\mo,\mb,\Tp,\Tn}_\RQ$ (cf.\ Corollary~\ref{cor:query_num_lower_bound}).  

As a first action (lines \ref{algoline:query:ent_start}-\ref{algoline:query:ent_end}), the algorithm computes a set of entailments $E_{\md_i}$ 
for each $\mo_i^*$ (cf.\ Formula~\ref{eq:sol_ont_candidate})
where $\md_i \in \mD$ and stores these entailments along with the respective diagnosis as a tuple $\tuple{\md_i,E_{\md_i}}$ in a set $E_\mD$. \label{etc:definition_of_getEntailments_function} This is accomplished by the function \textsc{getEntailments} which gets a tuple $\tuple{X,Y,Z,W}$ of arguments where $X,Y,Z$ are sets of formulas over some logic $\mathcal{L}$ and $W$ is a set including sets of formulas over $\mathcal{L}$. Then, \textsc{getEntailments} computes a \emph{finite} (cf.\ Remark~\ref{rem:entailment_computation_finite_types_of_entailments}) set of entailments of certain types (cf.\ Examples~\ref{example:query_computation} and \ref{example:ad_Table_of_queries_partitions}) of the KB $(Y\setminus X) \cup Z \cup U_{W}$. 

Then, the algorithm runs through all proper non-empty subsets $\mS$ of the leading diagnoses $\mD$ and, for each $\mS$, it computes the set of common entailments $Q$ of all KBs $\mo_i^*$ where $\md_i \in \mS$ (function \textsc{getCommonEntailments}) by means of the precomputed set $E_\mD$. That is, $Q := \bigcap_{\md \in \mS} E_{\md}$. If $Q$ is non-empty, then CQ1 and CQ2 are fulfilled for $Q$. CQ2 is met since $\mS \neq \emptyset$ and thus there is a diagnosis $\md_i\in\mD$ such that $\mo_i^* \models Q$ which implies that $\dx{}(Q) \neq \emptyset$. So, the algorithm proceeds to verify CQ3 (lines~\ref{algoline:query:verify_CQ3_start}-\ref{algoline:query:verify_CQ3_end}) in that it assigns the remaining diagnoses in $\mD$ that are not in $\mS$ to the according sets $\dx{}(Q)$, $\dnx{}(Q)$ or $\dz{}(Q)$ as per Definition~\ref{def:q-partition}. Note that the function \textsc{isKBValid} has been specified in Algorithm~\ref{algo:qx} on page~\pageref{algo:qx}. With the parameters given when called in line~\ref{algoline:query:is_ont_valid}, \textsc{isKBValid} checks whether $\mo_r^* \cup Q = (\mo\setminus\md_r) \cup\mb\cup U_{\Tp\cup\setof{Q}}$ does not violate any requirement in $\RQ$ and does not entail any test case in $\Tn$. Once the call to this function returns $\false$ for one diagnosis $\md_r \in \mD\setminus\mS$, it holds that $\md_r \in \dnx{}(Q)$ thus CQ3 is definitely met. Therefore, $isQuery$ is set to $\true$ in line~\ref{algoline:query:is_query_true}. If, on the other hand, $isQuery$ is not set to $\true$ for any diagnosis in $\mD \setminus \mS$, then the set $\dnx{}(Q) =\emptyset$ and thus $Q$ is not in $\mQ_{\mD,\tuple{\mo,\mb,\Tp,\Tn}_\RQ}$.

So far, we have proven the following proposition.
\begin{proposition}\label{prop:query_gen_isQuery_correct}
Let a DPI $\tuple{\mo,\mb,\Tp,\Tn}_\RQ$, a set of diagnoses $\mD\subseteq\minD_{\tuple{\mo,\mb,\Tp,\Tn}_\RQ}$ and a natural number $q \geq 1$ be the input to the function \textsc{getPoolOfQueries}. Then, a value stored in variable $Q$ at the time \textsc{getPoolOfQueries} executes line~\ref{algoline:query:add_QP_start} is a query w.r.t.\ $\mD$ and $\tuple{\mo,\mb,\Tp,\Tn}_\RQ$ iff the variable $isQuery$ stores the value $\true$.
\end{proposition}

If the purpose was only to find queries (and not q-partitions), the algorithm could stop processing for the current $Q$ and go to the next set $\mS$, given that $isQuery$ is set to $\true$ for some diagnosis. However, as the q-partition provides meaningful information to assess a query, e.g.\ it gives the number of diagnoses invalidated for each answer or the estimated probability of each answer (cf.\ Section~\ref{sec:UserInteraction}), the q-partition is a necessary input to the subsequently called function \textsc{selectBestQuery} (line~\ref{algoline:inter_onto_debug_continued:selectBestQuery} in Algorithm~\ref{algo:inter_onto_debug_continued}, see later in Sections~\ref{sec:Walkthrough} and \ref{sec:query_selection_measures}) that selects a query from the pool of queries $\QP$. For this reason, the algorithm continues until the computation of the q-partition for $Q$ is complete.

In a last step (lines~\ref{algoline:query:add_QP_start}-\ref{algoline:query:add_QP_end}), given that $isQuery$ is $\true$ and there is not yet a query with the same q-partition in $\QP$, the algorithm computes a set-minimal subset $Q_{\min}$ of $Q$ such that the q-partition of $Q_{\min}$ is the same as the one of $Q$ (function \textsc{minQ}). Finally, the tuple $\tuple{Q_{\min},\tuple{\dx{}, \dnx{}, \dz{}}}$ including the minimized query $Q_{\min}$ along with its q-partition $\tuple{\dx{}, \dnx{}, \dz{}}$ is added to $\QP$. If $|\QP|=q$, then $\QP$ is returned; otherwise, a further iteration for another $\mS$ is executed. If $|\QP|=q$ is not met until all seeds $\mS$ have been processed, the set $\QP$ is checked for emptiness in line~\ref{algoline:query:check_QP_empty}. If $\QP = \emptyset$, then the function \textsc{addTrivialQueries} (line~\ref{algoline:query:addTrivialQueries}) adds $|\mD| \geq 2$ queries as defined by $Q$ in Proposition~\ref{prop:q1} to $\QP$ (cf. Corollary~\ref{cor:query_num_lower_bound}) and then returns $\QP$; otherwise, $\QP$ is directly returned.

\begin{remark}\label{rem:guaranteeing_non-empty_QP_by_getPoolOfQueries}
Notice that lines~\ref{algoline:query:check_QP_empty} and \ref{algoline:query:addTrivialQueries} in Algorithm~\ref{algo:query_gen} aim at ensuring the non-emptiness of the pool of queries $\QP$ returned by \textsc{getPoolOfQueries} for \emph{any} \textsc{getEntailments} function (see Example~\ref{example:ad_Table_of_queries_partitions} for different specifications of the \textsc{getEntailments} function). This is a necessary criterion for the interactive KB debugging system (Algorithm~\ref{algo:inter_onto_debug}) to work in a sound way since it guarantees that the \textsc{calcQuery} function (line~\ref{algoline:inter_onto_debug:calc_query} in Algorithm~\ref{algo:inter_onto_debug}) \emph{always} returns a query w.r.t.\ the current set of leading diagnoses $\mD$ and the given DPI. 
%
Note that the $|\mD|$ queries generated and added to $\QP$ by \textsc{addTrivialQueries} can be \emph{trivially} obtained without the consultation of a reasoning service by extraction of the respective formulas from the KB $\mo$, as prescribed by Proposition~\ref{prop:q1}.\qed 
%
%
\end{remark}

\subsection{Discussion of Query Pool Generation}
\label{sec:remarks_query_gen}

\paragraph{Multiple Equal Q-Partitions.} In the general case there is more than one query w.r.t.\ one and the same q-partition. For that reason alone that a minimized query is a set-minimal subset of an initially computed one where multiple such subsets may exist. 

\begin{example}\label{example:multiple_minimized_queries}
An example for such a query resulting in multiple minimized subqueries with identical q-partition can be found in Example~\ref{example:query_computation}.\qed
\end{example}

However, note that \textsc{getPoolOfQueries} is designed to compute a pool $\QP$ that includes at most one query with one and the same q-partition. The idea behind this is (1)~to minimize the calls to the expensive function \textsc{minQ} and (2)~that two queries with the same q-partition have exactly the same properties w.r.t.\ common query selection criteria such as maximum expected information gain or maximum worst case invalidation rate of diagnoses after the query answer is known. Such criteria have been shown to often lead to a reduction of debugging effort for the interacting user (cf. \cite{Shchekotykhin2012,Rodler2013}).
As the purpose of the computation of the pool of queries $\QP$ is to constitute an input to the query selection function that uses exactly such selection measures, the inclusion of only one query with a particular q-partition is reasonable, also (3)~to minimize computation time of the query selection function which needs to go through all elements of $\QP$ in order to pick the ``best'' one in the worst case.
 
On the other hand, regarding the comprehensibility of the query, i.e.\ the cognitive load on the user when it comes to understanding the meaning of the query, two queries with the same q-partition may well be significantly different. This however is beyond the scope of this work and considered a topic for future research. 

The following proposition gives evidence that the set $\QP$ returned by \textsc{getPoolOfQueries} is indeed duplicate-free w.r.t.\ the q-partitions in $\QP$. 
\begin{proposition}\label{prop:query_gen_duplicate_free}
Let a DPI $\tuple{\mo,\mb,\Tp,\Tn}_\RQ$, a set of diagnoses $\mD\subseteq\minD_{\tuple{\mo,\mb,\Tp,\Tn}_\RQ}$ and $q \in \mathbb{N} \cup \setof{\infty}, q \geq 1$ be the input to the function \textsc{getPoolOfQueries}. Then, the function \textsc{getPoolOfQueries} returns a set $\QP$ including tuples of the form $\tuple{Q,\Pt(Q)}$ where $Q\in\mQ_{\mD,\tuple{\mo,\mb,\Tp,\Tn}_\RQ}$ is a query and $\Pt(Q) = \tuple{\dx{}(Q),\dnx{}(Q),\dz{}(Q)}$ is the q-partition of $Q$ such that $\QP$ does not include any two equal queries and does not include any two equal q-partitions.  
\end{proposition}
\begin{proof}
The test of the criterion $\lnot \textsc{inclQPart}$ tested before the call to \textsc{minQ} will always return false for the q-partition $\tuple{\dx{},\dnx{},\dz{}}$ if $\tuple{\dx{},\dnx{},\dz{}}$ is already included in a tuple in $\QP$. Since \textsc{minQ} is q-partition-preserving, no q-partition that does not occur in a tuple in $\QP$ can become equal to some q-partition in $\QP$ by a call to \textsc{minQ}. Therefore, $\QP$ cannot include any two equal q-partitions. Since two equal queries have equal q-partitions, any two different q-partitions cannot be q-partitions of equal queries. Thus, $\QP$ cannot include any two equal queries either. 
\end{proof}
Note that, on account of the q-partition preserving property of \textsc{minQ}, only such q-partitions are ruled out by the criterion in line~\ref{algoline:query:add_QP_start} that would lead to duplicates at the time they should be added to $\QP$ in line~\ref{algoline:query:add_QP_end}.

\paragraph{Computation of Entailments.} Generally, the (\emph{theoretical}) number of entailments of a set of formulas is not finite. However, the entailments (of a certain type) returned by a reasoner are finite. For instance, asked for entailments of $\setof{A \sqsubseteq B \sqcap C}$, a reasoner performing the classification reasoning service would give back $A \sqsubseteq B$ and $A \sqsubseteq C$, but not entailments like $A \sqsubseteq B \sqcup C$ or $A \sqsubseteq C \sqcap C \sqcap C$. That is, when we speak of entailments, then we mean entailments in the \emph{practical} sense (cf.\ Remark~\ref{rem:entailment_computation_finite_types_of_entailments}), i.e.\ w.r.t.\ a reasoning service such as classification for DL KBs which computes all and only subsumptions $X \sqsubseteq Y$ such that $Y$ is the most specific concept that subsumes $X$, or forward-chaining for Datalog KBs which computes all and only atoms that are entailed by the KB.

\begin{example}\label{example:entailment_number} 
If we recall Example~\ref{example:query_computation}, we see that the number of computed entailments of $\mo_4^*$ and $\mo_3^*$ was 19 and 13 respectively, which are rather high numbers in the light of the small KBs, but importantly these numbers are necessarily finite. For, there cannot be more than $|Pred|^2$ entailments of the $\forall X p_1(X) \rightarrow p_2(X)$ type and not more than $|Pred|\,|Const|$ entailments of the $p(a)$ type for a KB whose signature includes the unary predicate symbols $Pred$ and constant symbols $Const$ and does not include any function symbols. In case of KB $\mo_3^*$, for example, the set $Pred = \setof{a_1,a_2,m_1,m_2,m_3,a,b}$ and $Const=\setof{u,w}$ which means that upper bounds for the number of entailments of the first and second type are $49$ and $14$, respectively.\qed
\end{example}

Further, note that the number of existing different q-partitions and which q-partitions there are at all w.r.t.\ some set of leading diagnoses $\mD$ and a DPI depends on the function \textsc{getEntailments}, i.e.\ on the set of entailments calculated by it. 


\begin{example}
Recall Example~\ref{example:query_computation} where we constructed a query $Q$ w.r.t.\ the set of all minimal diagnoses for the DPI given by Table~\ref{tab:example1}. Assume now that only entailments of the first type, i.e.\ those of the form $\forall X p_1(X) \rightarrow p_2(X)$, and none of the second type $p(a)$ are computed by \textsc{getEntailments} and denote the set of entailments of this form of $\mo_i^*$ by $E'_{\md_i}$. Then, $Q' = E'_{\md_3}\cup E'_{\md_4} = \setof{e_1,\dots,e_5}$ (cf.\ Table~\ref{tab:example:query_construction_entailments}), i.e.\ a subset of the query $Q$ computed for a \textsc{getEntailments} function producing entailments of both types. The q-partition of $Q'$ is the same as the q-partition of $Q$, namely $\tuple{\setof{\md_3,\md_4},\setof{\md_1,\md_2},\emptyset}$. However, the queries $Q_{\min,1}$ and $Q_{\min,2}$ are no longer obtained as minimized versions of $Q'$, unlike $Q_{\min,3}$ and $Q_{\min,4}$ which are subqueries of $Q'$, too.  \qed
\end{example}

\paragraph{Minimizing the Set $\dz{}$ in Q-Partitions.} Recall that $\dz{} = \emptyset$ is a desirable property of a q-partition since a query with such q-partition may invalidate any leading diagnosis, depending on the answer to the query (cf.\ Section~\ref{sec:UserInteraction}). In other words, no leading diagnosis is guaranteed to be still valid \emph{for any answer} after the query is added as a test case to the DPI. 

In general, \textsc{getPoolOfQueries} computes q-partitions where $\dz{}$ may be a non-empty set. However, if the \textsc{getEntailments} function is specified to compute certain explicit entailments of $\mo$, then $\dz{} = \emptyset$ can be guaranteed.
\begin{definition}[Explicit entailment]\label{def:explicit_entailment}
Let $\mo$ be a KB. Then, $\alpha$ is an explicit entailment of $\mo$ iff $\alpha \in \mo$.
\end{definition}
Now, if each set of entailments $E_\md$ computed by \textsc{getEntailments} includes all the formulas that occur in some diagnosis in $\mD$, but do not occur in $\md$, then \textsc{getPoolOfQueries} definitely returns a set $\QP$ of queries and associated q-partitions where $\dz{}(Q) = \emptyset$ holds for each tuple in $\QP$.
\begin{proposition}\label{prop:query_gen_explicit_entailments_dz_empty}
Let $\langle\mo,\mb,\Tp,\Tn\rangle_\RQ$ be a DPI and $\mD \subseteq \minD_{\langle\mo,\mb,\Tp,\Tn\rangle_\RQ}$. If the set $E_{\md}$ computed by \textsc{getEntailments} meets $E_{\md} \supseteq U_{\mD} \setminus \md$ for all $\md \in \mD$, then \textsc{getPoolOfQueries} computes only queries $Q$ with $\dz{}(Q) = \emptyset$.
\end{proposition}
\begin{proof}
Assume that $Q$ is some query computed by \textsc{getPoolOfQueries}. As \textsc{minQ} is a q-partition preserving transformation of $Q$, we can assume w.l.o.g.\ that $Q$ is a query computed by \textsc{getPoolOfQueries} \emph{before} \textsc{minQ} is called for $Q$. 
We have to show that for an arbitrary diagnosis $\md_i \in \mD$ either $\md_i$ is assigned to $\dx{}(Q)$ or to $\dnx{}(Q)$. 

So, let us assume that there is a diagnosis $\md_k$ which is assigned to $\dz{}(Q) = \mD\setminus (\dx{}(Q)\cup\dnx{}(Q))$ in line~\ref{algoline:query:dz}. Then, $Q \not\subseteq E_{\md_k}$ and $\mo_k^* \cup Q$ does not violate any $x \in \RQ\cup\Tn$ must hold, otherwise $\md_k$ would have already been assigned to $\dx{}(Q)$ in line~\ref{algoline:query:dx} or to $\dnx{}(Q)$ in line~\ref{algoline:query:dnx}. 
But $Q \not\subseteq E_{\md_k}$ implies $Q \not\subseteq U_{\mD} \setminus \md_k$ since $E_{\md_k} \supseteq U_{\mD} \setminus \md_k$ by precondition.
This in turn means that there is some formula $\tax$ in $Q$ which is not in $U_{\mD} \setminus \md_k$. Then $\tax\in\md_k$ must hold, as otherwise for all formulas $\tax'\in Q$ it would hold that $\tax'$ is an entailment of $\mo_k^* = (\mo\setminus\md_k)\cup \mb \cup U_\Tp$, i.e.\ an entailment of all formulas in $\mo \cup \mb \cup U_{\Tp}$ except for those in $\md_k$. However, all entailments of $\mo_k^*$ are stored in $E_{\md_k}$ by the implementation of the function \textsc{getEntailments}. Thus $Q \subseteq E_{\md_k}$ would hold which cannot be the case as shown before. Consequently, we have derived that $Q \cap \md_k \neq \emptyset$ which means by set-minimality of diagnoses in $\mD$, in particular of $\md_k$, that $\mo_k^* \cup Q$ must violate some $x\in\RQ\cup\Tn$ which is a contradiction to the assumption that $\md_k \in \dz{}(Q)$.
\end{proof}

\begin{example}
Let us come back to the example DPI given by Table~\ref{tab:example1}. The possibility of a query $Q$ constructed by Algorithm~\ref{algo:query_gen} with $\dz{}(Q) \neq \emptyset$ is witnessed by the selection of seed $\mS = \setof{\md_1}$ and the assumption that entailments of the two types given in Example~\ref{example:query_computation} are produced by \textsc{getEntailments}. The set of entailments $Q = E_{\md_1} = \setof{e_4,e_{14},e_{15},\forall X m_2(X) \rightarrow d(X)}$ (for $e_i$ cf.\ Table~\ref{tab:example:query_construction_entailments}). Then, $\md_2$ as well as $\md_3$ are assigned to $\dnx{}(Q)$ as both KBs $\mo_3^*\cup Q$, $\mo_4^* \cup Q$ entail $m_3(w)$ and $\lnot m_3(w)$ wherefore they are both inconsistent and thus violate $r_1 \in \RQ$. However, $\md_4 \in \dz{}(Q)$ since $\mo_i^* \not\models \forall X m_2(X) \rightarrow d(X)$ and hence does not entail $Q$ and since $\mo_i^* \cup Q$ does not violate consistency or coherency (recall that the set of negative test cases is empty in the DPI and thus must not be considered), i.e.\ does not contain a conflict set.

Applying Proposition~\ref{prop:query_gen_explicit_entailments_dz_empty}, we could use a modified \textsc{getEntailments} function that returns a minimal set of entailments just that the precondition of the proposition is met, i.e.\ $E'_{\md} = U_{\mD} \setminus \md$ for all $\md\in\mD$. With this function, for the seed $\mS = \setof{\md_1}$ we would get $Q' = E'_{\md_1} = \setof{2,3,4,5}$ (again, formulas in Table~\ref{tab:example1} are referred to just by their number). Let us now check whether $\dz{}(Q')$ is indeed empty. As explicit entailments are stronger than non-explicit ones, we must still have that $\md_2,\md_3 \in \dnx{}(Q')$. For $\md_4$, we have $\mo_4^* \cup Q' = \setof{1,3,5,6,7,8} \cup \setof{2,3,4,5} = \setof{1,2,3,4,5,6,7,8}$ which corresponds to the entire KB plus background knowledge of the given DPI and includes conflict sets $\mc_1 = \setof{1,3,4}$ and $\mc_2= \setof{1,2,3,5}$ wherefore it is inconsistent. Therefore, diagnosis $\md_4$ must also be an element of $\dnx{}(Q')$. 

Please note that making the entailments $Q = E_{\md_1}$ computed by the unmodified \textsc{getEntailments} function only slightly stronger would already suffice to force inclusion of $\md_4$ in $\dz{}(Q)$. In fact, including $\tax_4 := \forall X m_2(X) \rightarrow (\forall Y s(X,Y) \rightarrow a(Y)) \land d(X)$ in $Q$ instead of $\forall X m_2(X) \rightarrow d(X)$ would make $Q$ non-disjoint with $\md_4$ as both comprise $\tax_4$. Consequently, in line with the proof of Proposition~\ref{prop:query_gen_explicit_entailments_dz_empty}, $\mo_4^* \cup Q$ must include a conflict set ($\setof{1,3,4}$) wherefore $\md_4 \in \dnx{}(Q)$. 

Another point we want to mention is that empty $\dz{}$ \emph{could} also be achieved by making the query slightly weaker. For our concrete query $Q = E_{\md_1}$, this means that leaving out $\forall X m_2(X) \rightarrow d(X)$ would lead to empty $\dz{}(Q)$. However, the difference to the scenario above where we made $Q$ sightly stronger is that $\md_4$ would be an element of $\dx{}(Q)$ instead of $\dnx{}(Q)$ in this case. i.e.\ the q-partition would be $\tuple{\setof{\md_1,\md_4},\setof{\md_2,\md_3},\emptyset}$. 

A shortcoming of the strategy of making the query weaker is that it can be computationally expensive as perhaps a large number of subsets of $Q$ might need to be considered and tested for fulfillment of $\dz{}(Q)=\emptyset$. Each such test would involve calls to the reasoner which are usually expensive. A second drawback is that no guarantee is given to finally end up with an empty set $\dz{}(Q)$ since weakening of $Q$ might also involve the ``shift'' of some diagnosis from $\dnx{}(Q)$ to $\dz{}(Q)$. On the other hand, the strategy of computing stronger entailments is computationally more resource-saving as (trivially obtained) explicit entailments can be added to make the query stronger. Furthermore, making the query stronger -- in a controlled way, by adding formulas from $U_{\mD} \setminus U_{\dx{}(Q)}$ to $Q$ as suggested by Proposition~\ref{prop:query_gen_explicit_entailments_dz_empty} -- can never lead to non-empty $\dz{}(Q)$ as Proposition~\ref{prop:query_gen_explicit_entailments_dz_empty} substantiates. \qed
\end{example}

\paragraph{(Non-)Completeness of Query Pool $\QP$.} Note that specifying $q:=\infty$ causes \textsc{getPoolOfQueries} to run through all $\mS\subset\mD$ and to compute a maximum number of queries. However, in general, not all theoretically possible queries are computed by \textsc{getPoolOfQueries}. One trivial reason for this is that only minimized, i.e.\ set-minimal, queries are contained in the returned set $\QP$. 

But, also queries $Q'$ with $\dx{}(Q') = Y \subset \mD$ will not be included in $\QP$ if there is some query $Q$ with $\dx{}(Q) = Y$ such that $|\dnx{}(Q)| > |\dnx{}(Q')|$ (and, equivalently, $|\dz{}(Q)| < |\dz{}(Q')|$). As we will learn in a moment, both mentioned reasons for the incompleteness of the output of \textsc{getPoolOfQueries} will even be desirable for reasons of efficiency. That is, the mentioned types of queries that are not taken into account in $\QP$ are ``non-preferred'' as non-set-minimal queries demand a non-necessary amount of user interaction and the answering of queries $Q$ with a non-necessarily large set $\dz{}(Q)$ involves a worse discrimination between leading minimal diagnoses (and, if these are ``good'' representatives of all minimal diagnoses, then of all minimal diagnoses) than other queries $Q'$ with $|\dz{}(Q')| < |\dz{}(Q)|$ and $\dx{}(Q) = \dx{}(Q')$. 


Still, \textsc{getPoolOfQueries} meets a completeness criterion for a subset of all queries $\mQ_{\mD,\tuple{\mo,\mb,\Tp,\Tn}_\RQ}$, elements of which cannot be \emph{trivially} detected to be ``non-preferred''. That is, \textsc{getPoolOfQueries} is complete w.r.t.\ the set $\dx{}$, as the following proposition states. In other words, for each subset $X \subset \mD$ it detects a q-partition with $\dx{} = X$, if one exists. 

\begin{proposition} \label{prop:query:dx_complete}
Let a DPI $\tuple{\mo,\mb,\Tp,\Tn}_\RQ$, $\mD \subseteq \minD_{\tuple{\mo,\mb,\Tp,\Tn}_\RQ}$ such that $|\mD| \geq 2$ and some $q \in \mathbb{N} \cup \setof{\infty}, q \geq 1$  be the inputs to \textsc{getPoolOfQueries} and let $|\QP_{\max}| \geq 0$ be the maximum number of tuples $\tuple{Q,\Pt(Q)}$ that can be computed by \textsc{getPoolOfQueries} by means of the used \textsc{getEntailments} function. Further, let $Y$ be an arbitrary subset of $\mD$. If there is some query $Q \in \mQ_{\mD,\tuple{\mo,\mb,\Tp,\Tn}_\RQ}$ that (1)~includes only entailments that are computed by \textsc{getEntailments} and (2)~has a q-partition such that $\dx{}(Q) = Y$, then \textsc{getPoolOfQueries} with parameter $q\geq |\QP_{\max}|$ returns a set $\QP$ including a query $Q'$ with $\dx{}(Q') = Y$. 
Moreover, this query $Q'$ is found in the iteration where the seed $\mS = Y$.
\end{proposition}
\begin{proof}
Since $q\geq |\QP_{\max}|$, \textsc{getPoolOfQueries} will arrive at a step where it selects the seed $\mS = Y$ in line~\ref{algoline:query:seed}. Now, let us assume that in this iteration no query $Q$ with $\dx{}(Q)=Y$ is found. Then, either (a)~no query is found at all, i.e.\ CQ1 or CQ2 or CQ3 are violated, or (b)~a query $Q$ with $\dx{}(Q) \neq Y$ is found. 

(a): Assume first that CQ1 is violated, i.e.\ \textsc{getCommonEntailments} called with argument $\mS$ returns $\emptyset$. This implies that the KBs $\mo_r^*$ for $\md_r \in Y$ have no common entailments, if entailments are computed by \textsc{getEntailments}. This however means that there cannot be a q-partition with $\dx{} \supseteq Y$ which is a contradiction to the precondition that there is some query $Q \in \mQ_{\mD,\tuple{\mo,\mb,\Tp,\Tn}_\RQ}$ that includes only entailments computed by \textsc{getEntailments} and has a q-partition such that $\dx{}(Q) = Y$.

Second, assume that CQ2 is violated, i.e.\ $\dx{}(Q) = \emptyset$. If \textsc{getCommonEntailments} with argument $\mS$ returned $Q \neq \emptyset$, then $\dx{}(Q) \supseteq \mS \supset \emptyset$ would hold. Thus, $Q = \emptyset$, i.e.\ CQ1 is violated. So, as shown before, this leads to a contradiction.

In case any of CQ1 or CQ2 is violated, we already derived a contradiction. So, we make the assumption that CQ1 and CQ2 are met. So, finally, let us assume that CQ3 is violated, i.e.\ that $\dnx{}(Q) = \emptyset$. That is, if $Q$ (which must be a non-empty set by CQ1) denotes \emph{all} common entailments (computable with \textsc{getEntailments}) of $\mo_r^*$ for $\md_r\in Y$, then $\mo_i^* \cup Q$ does not violate any $x\in\RQ\cup\Tn$ for any $\md_i \in \mD \setminus \mS$. Consequently, for all diagnoses $\md_i$ in $\mD$ we have that $\mo_i^* \cup Q$ does not violate any $x\in\RQ\cup\Tn$. But, as there is, by precondition, a query with $\dx{} = Y$, this query must be a subset of all possible common entailments (computable with \textsc{getEntailments}) of KBs $\mo_i^*$ for diagnoses in $Y$, i.e.\ this query must be a subset of $Q$. But, by monotonicity of $\mathcal{L}$, no $\mo_i^* \cup Q'$ for a subset $Q'$ of $Q$ can violate $x\in\RQ\cup\Tn$ if $Q$ does not. Again, we have a contradiction to the precondition as above.

(b): Here, a query $Q$ is found with $\dx{}(Q) \neq Y$ and $\dnx{}(Q) \neq \emptyset$. Since $Q$ is a query, $Q\neq \emptyset$ must hold. Since the seed $\mS = Y$, this means that $Q$ is the set of all common entailments (computable with \textsc{getEntailments}) of $\mo_i^*$ for $\md_i \in Y$, i.e.\ $\dx{}(Q) \supseteq Y$. By $\dx{}(Q) \neq Y$, we conclude that $\dx{}(Q) \supset Y$ must be true. The only way of achieving a smaller set $\dx{}(Q)$, namely $\dx{}(Q) = Y$, is to add some formulas to $Q$ as making $Q$ smaller can only increase $\dx{}(Q)$. This holds because postulating that, instead of $Q$, only a subset $Q'$ of $Q$ must be entailed by $\mo_i^*$, can cause a new KB $\mo_j^*$ for diagnosis $\md_j \notin\dx{}(Q)$ to entail $Q'$. However, as $Q$ is the set of \emph{all} entailments computable with \textsc{getEntailments} of KBs $\mo_i^*$ for $\md_i\in Y$, a superset $Q''$ of $Q$ computed by \textsc{getEntailments} with $\dx{}(Q'') = Y$ can never be obtained. Therefore, we have a contradiction to the precondition.

We have now proven the following: If there exists a q-partition as described in the proposition, then this q-partition is found in the iteration where the seed $\mS = Y$.  
%
\end{proof}
\begin{remark}\label{rem:ad_prop:query:dx_complete}
Regarding Proposition~\ref{prop:query:dx_complete}, note the following:
\begin{enumerate}[(a)]
	\item In fact, as one and the same q-partition must occur at most once in $\QP$, \textsc{getPoolOfQueries} must only keep assigning diagnoses in $\mD \setminus \mS$ to the respective sets of the q-partition as long as $\dx{} = \mS$. Because for $\dx{} = Z \supset \mS$, we know to find a query (if one exists) for the seed $\mS = Z$.
	\item A statement equivalent to the proposition is: If there is no query (including only entailments computed by the \textsc{getEntailments} function) with $\dx{} = Y$ found for seed $\mS = Y$, then such a query and q-partition, respectively, does not exist.\qed
\end{enumerate}
\end{remark}
The following proposition states that if a q-partition with one and the same set $\dx{}$ is found twice during the execution of \textsc{getPoolOfQueries}, then the queries for both q-partitions and thus both q-partitions must be equal. That is, for one set $\dx{}$, there is at most one tuple in $\QP$. 
\begin{proposition}\label{prop:query_gen_dx_unique}
Let $Q_i$ be a query with $\dx{}(Q_i) = Y$ in the set $\QP$ returned by \textsc{getPoolOfQueries} and found for seed $\mS_i = Y$ and
let $Q_j$ be a query with $\dx{}(Q_j) = Y$ in the set $\QP$ returned by \textsc{getPoolOfQueries} and found for some seed $\mS_j \subset Y$. Then $Q_i = Q_j$.  
\end{proposition}
\begin{proof} Let $Q'_i, Q'_j$ be the queries stored in the variable $Q$ in line~\ref{algoline:query:add_QP_start} for seeds $\mS_i$ and $\mS_j$, respectively; i.e. the supersets of the queries $Q_i,Q_j$ \emph{before} the minimization function \textsc{minQ} is called for each of them. $Q'_j \subseteq Q'_i$ holds by the fact that $Q'_i$ is the set of \emph{all} common entailments computable with \textsc{getEntailments} of $\mo_r^*$ for $\md_r \in Y$ and by the fact that $Q'_j$ must be a set of common entailments computed by \textsc{getEntailments} of exactly these KBs, because of $\dx{}(Q'_j) = Y$ and Definition~\ref{def:q-partition}. $Q'_j \supseteq Q'_i$ holds by the fact that $Q'_j$ is computed as intersection of $E_{\md_r}$ where $\md_r \in \mS_j$ and $Q'_i$ is computed as intersection of $E_{\md_s}$ where $\md_s \in \mS_i \supset \mS_j$. Thus, we can conclude that $Q'_i = Q'_j$.

As $Q'_i = Q'_j$, also $\Pt(Q'_i) = \Pt(Q'_j)$ must hold for the q-partitions by Proposition~\ref{prop:unique_q-partition}. That the minimized versions $Q_i, Q_j$ of $Q'_i, Q'_j$ output by \textsc{minQ} are equal, follows from the determinism of the \textsc{minQ} function, wherefore equal inputs, i.e.\ $(\emptyset,Q'_i,\emptyset,\Pt(Q'_i),\tuple{\mo,\mb,\Tp,\Tn}_\RQ) = (\emptyset,Q'_j,\emptyset,\Pt(Q'_j),\langle\mo,\mb,\Tp$, $\Tn\rangle_\RQ)$, must yield equal outputs.
\end{proof}
%
%
\begin{remark}\label{rem:improvements_of_algo_QP_generation}
Proposition~\ref{prop:query_gen_dx_unique} hints at a possible improvement 
of Algorithm~\ref{algo:query_gen}, namely to check in line~\ref{algoline:query:seed} whether the seed $\mS$ already occurs as a set $\dx{}$ in some tuple in $\QP$ and only continue the execution for $\mS$ if this does not hold (not shown in Algorithm~\ref{algo:query_gen}). In this vein, time and reasoning costs (line~\ref{algoline:query:dnx}) can be saved. 

Another improvement regarding line~\ref{algoline:query:seed} is to delete all remaining seeds $\mS'$ with the property $\mS' \supset \mS$ if $Q$ in line~\ref{algoline:query:common_ent} is the empty set (not shown in Algorithm~\ref{algo:query_gen}). Namely, all seeds $\mS'$ must also lead to $Q=\emptyset$ since the intersection of $E_{\md}$ for $\md\in\mS$ already returned $\emptyset$ wherefore the intersection of $E_{\md}$ for $\md\in\mS'$ must also return $\emptyset$.\qed
\end{remark}

By now, we know from Proposition~\ref{prop:query_gen_dx_unique} that, given a query with $\dx{}$ exists, one and only one q-partition with $\dx{}$ will be added to $\QP$, but which one?

W.r.t.\ one and the same set $\dx{}$, queries with a set $\dnx{}$ with higher cardinality are preferable over others as the cardinality of $\dz{}$ should be minimized (cf. Section~\ref{sec:UserInteraction}). So, preferable queries among those with equal set $\dx{}$ are those for which $\dnx{}$ is a set-maximal set. Exactly such a query is added to $\QP$ for each $\dx{}$ for which a query exists, as the following proposition shows.
\begin{proposition}\label{prop:given_dx_the_q-partition_with_min_dz_is_found}
If the set $\QP$ returned by \textsc{getPoolOfQueries} comprises a query $Q$ with $\dx{}(Q) = Y$, then $Q$ is a query with minimal $|\dz{}(Q)|$ among all queries $Q'$ with $\dx{}(Q') = Y$ computable with the function \textsc{getEntailments}.
\end{proposition} 
\begin{proof}
Assume that \textsc{getPoolOfQueries} finds a query $Q$ with $\dx{}(Q) = Y$ and $|\dz{}(Q)| = k$ and assume there is a query $Q'$ (consisting only of entailments computed by function \textsc{getEntailments}) with $\dx{}(Q') = Y$ and with $|\dz{}(Q')| < k$. This means that $|\dnx{}(Q)| < |\dnx{}(Q')|$. However, as $Q$ is computed for seed $\mS = Y$, $Q$ is a maximal set of entailments computable with \textsc{getEntailments} of $\mo_i^*$ for $\md_i \in Y$. Because $Q'$ is also a common entailment of $\mo_i^*$ for $\md_i \in Y$, we have that $Q' \subseteq Q$ must be true. 
Since the fact that $\mo_i^* \cup Q$ does not violate any $x\in\RQ\cup\Tn$, i.e.\ the fact that $\md_i \notin \dnx{}(Q)$, implies by monotonicity of $\mathcal{L}$ that $\mo_i^* \cup Q'$ for the subset $Q'$ of $Q$ cannot violate any $x\in\RQ\cup\Tn$ either, i.e.\ $\md_i \notin \dnx{}(Q')$, we conclude that $|\dnx{}(Q')| \leq |\dnx{}(Q)|$ must hold. This is a contradiction.
\end{proof}
%
%

\subsection{Minimization of Queries}
\label{sec:minimization_of_queries}
\noindent\textbf{\textsc{minQ}.} The minimization of the query $Q$ by \textsc{minQ} (see Algorithm~\ref{algo:query_gen}) while preserving the q-partition
aims at simplifying the job of the answering user who only needs to go through a smaller set of logical formulas $Q_{\min}$ in order to come up with an answer to the query. Since the q-partition reflects the properties of a query w.r.t.\ the invalidation of (leading) diagnoses and two queries have equal such properties, then of course the one that is a subset of the other should be asked.

The concept of the function \textsc{minQ} is similar to the one of $\scQX$ (Algorithm~\ref{algo:qx}). Like $\scQX$, \textsc{minQ} carries out a divide-and-conquer strategy to find a set-minimal set with a monotonic property. In this case, the monotonic property is not the invalidity of a subset of the KB w.r.t.\ a DPI (as per Definition~\ref{def:valid_onto}) as it is for the computation of minimal conflict sets using $\scQX$, but the property of some $Q_{\min}\subset Q$ having the same q-partition as $Q$. So, the crucial difference between $\scQX$ and \textsc{minQ} is the function that checks this monotonic property. For \textsc{minQ}, this function -- that checks a subset of a query for constant q-partition -- is \textsc{isQPartConst}.	 

\paragraph{\textsc{minQ} -- Input Parameters.} \textsc{minQ} gets five parameters as input. The first three, namely $X, Q$ and $QB$, are relevant for the divide-and-conquer execution, whereas the last two, namely the original q-partition $\tuple{\dx{}, \dnx{}, \dz{}}$ of the query (i.e.\ the parameter $Q$) that should be minimized, and the DPI $\langle\mo,\mb,\Tp,\Tn\rangle_\RQ$ are both needed as an input to the function \textsc{isQPartConst}. Besides the latter two, another argument $QB$ is passed to this function where $QB$ is a subset of the original query $Q$. \textsc{isQPartConst} then checks whether the q-partition for the (potential) query $QB$ is equal to the q-partition $\tuple{\dx{}, \dnx{}, \dz{}}$ of the original query given as argument. The DPI is required as the parameters $\mo, \mb,\Tp,\Tn$ and $\RQ$ are necessary for these checks. 

\paragraph{\textsc{minQ} -- Testing Sub-Queries for Constant Q-Partition.} In particular, \textsc{isQPartConst} tests for each $\md_r\in\dnx{}$ whether $\mo_r^*\cup QB$ is valid (w.r.t.\ $\tuple{\cdot,\emptyset,\emptyset,N}_\RQ$). If so, this means that $\md_r \notin \dnx{}(QB)$ and thus that the q-partition of $QB$ is different to the one of $Q$ wherefore $\false$ is immediately returned. If $\true$ for all $\md_r\in\dnx{}$, it is tested for $\md_r\in\dz{}$ whether $\mo_r^* \models QB$. If so, this means that $\md_r \notin \dz{}(QB)$ and thus that the q-partition of $QB$ is different to the one of $Q$ wherefore $\false$ is immediately returned. If $\false$ is not returned for any $\md_r\in\dnx{}$ or $\md_r\in\dz{}$, then the conclusion is that $QB$ is a query w.r.t.\ to $\mD$ and $\langle\mo,\mb,\Tp,\Tn\rangle_\RQ$ and has the same q-partition as $Q$ wherefore the function returns $\true$. 

Note that, instead of calling a reasoner to answer whether $\mo_r^* \models QB$, the set of precalculated entailments $E_{\md_r}$ of $\mo_r^*$ for each $\md_r\in\mD$ can be given as an argument to \textsc{minQ} as well as to \textsc{isQPartConst} (not shown in Algorithm~\ref{algo:query_gen}). In this case an equivalent test is $QB \subseteq E_{\md_r}$. Such a strategy is particularly appropriate if reasoning is expensive for the DPI at hand.

Soundness of \textsc{isQPartConst} is proven by the following lemma.
\begin{lemma}\label{lem:isqpartconst_correct}
Let $\langle\mo,\mb,\Tp,\Tn\rangle_\RQ$ be a DPI, $\mD \subseteq \minD_{\langle\mo,\mb,\Tp,\Tn\rangle_\RQ}$, $Q\in\mQ_{\mD,\langle\mo,\mb,\Tp,\Tn\rangle_\RQ}$ with q-partition $\Pt(Q) = \tuple{\dx{}(Q),\dnx{}(Q),\dz{}(Q)}$. Then a non-empty set $QB \subset Q$ is a query in $\mQ_{\mD,\langle\mo,\mb,\Tp,\Tn\rangle_\RQ}$ with $\Pt(QB) = \Pt(Q)$ if 
\begin{enumerate}
	\item $\forall \md_r\in\dnx{}(Q): \; \mo_r^* \cup QB$ violates some $r\in\RQ$ or entails some $\tn\in\Tn$ and
	\item $\forall \md_r\in\dz{}(Q): \; \mo_r^* \not\models QB$.
\end{enumerate}
\end{lemma}
\begin{proof}
Let $Q \in \mQ_{\mD,\langle\mo,\mb,\Tp,\Tn\rangle_\RQ}$ and $QB$ be an arbitrary proper subset of $Q$.
If criterion 1) of this lemma is met, then we know that each diagnosis in $\dnx{}(Q)$ is in $\dnx{}(QB)$ as well, i.e.\ (I):~$\dnx{}(QB) \supseteq \dnx{}(Q)$ holds.

Assume a minimal diagnosis $\md_r\in\dz{}(Q)$. Then, $\mo_r^* \cup Q$ does not violate any $r\in\RQ$ and does not entail any $\tn\in\Tn$ and $\mo_r^*$ does not entail $Q$. This however implies that $\mo_r^* \cup QB$ cannot violate any $r\in\RQ$ and cannot entail any $\tn\in\Tn$ either by monotonicity of $\mathcal{L}$. But it is possible that $\mo_r^* \models QB$.
So, validity of criterion 2) of this lemma is sufficient to guarantee that each diagnosis in $\dz{}(Q)$ is in $\dz{}(QB)$ as well, i.e.\ (II):~$\dz{}(QB) \supseteq \dz{}(Q)$ holds.

As all diagnoses in $\dx{}(Q)$ entail all formulas in $Q$ by Definition~\ref{def:q-partition}, all diagnoses in $\dx{}(Q)$ must entail $QB$ as well. Consequently, due to deletion of some formulas from $Q$, no $\md_r \in \dx{}(Q)$ can ``move'' to any set $\dnx{}(QB)$ or $\dz{}(QB)$.
That is, (III):~$\dx{}(QB) \supseteq \dx{}(Q)$ must hold.

So, the overall conclusion is that, if criterion 1) and 2) are met, then (I), (II) and (III) hold. Assume that some $\supseteq$-relation in $i\in\setof{\text{(I), (II), (III)}}$ is a $\supset$-relation. This leads to a violation of some $j \in\setof{\text{(I), (II), (III)}}$ with $j\neq i$ since $\tuple{\dx{}(Q),\dnx{}(Q),\dz{}(Q)}$ and $\tuple{\dx{}(QB),\dnx{}(QB),\dz{}(QB)}$ are partitions of $\mD$. Therefore, all $\supseteq$-relations must be $=$-relations and we can derive that $\Pt(Q) = \Pt(QB)$.

Moreover, we have that $QB$ must be a query. This is due to the facts that $QB$ is non-empty, $Q$ is a query and the q-partitions of $Q$ and $QB$ are equal. Therefore, $\dx{}(QB) = \dx{}(Q) \geq 1$ and $\dnx{}(QB) = \dnx{}(Q) \geq 1$ which lets us conclude by Proposition~\ref{prop:query_dx_dnx} that $QB$ is a query.
\end{proof}

\paragraph{\textsc{minQ} -- The Divide-and-Conquer Strategy.} Intuitively, \textsc{minQ} partitions the given query $Q$ in two parts $Q_1$ and $Q_2$ and first analyzes $Q_2$ while $Q_1$ is part of $QB$ (line~\ref{algoline:query:recursive_call1}). Note that in each iteration $QB$ is the subset of $Q$ that is currently assumed to be part of the sought minimized query (i.e.\ the one query that will finally be output by \textsc{minQ}). In other words, analysis of $Q_2$ while $Q_1$ is part of $QB$ means that all irrelevant formulas in $Q_2$ should be located and removed from $Q_2$ resulting in $Q^{\min}_2 \subseteq Q_2$. That is, $Q^{\min}_2$ must include only relevant formulas which means that $Q^{\min}_2$ along with $QB$ is a query with an equal q-partition as $Q$, but the deletion of any further formula from $Q^{\min}_2$ changes the q-partition.

After the relevant subset $Q^{\min}_2$ of $Q_2$, i.e.\ the subset that is part of the minimized query, has been returned, $Q_1$ is removed from $QB$, $Q^{\min}_2$ is added to $QB$ and $Q_1$ is analyzed for a relevant subset that is part of the minimized query (line~\ref{algoline:query:recursive_call2}). This relevant subset, $Q^{\min}_1$, together with $Q^{\min}_2$, then builds a set-minimal subset of the input $Q$ that is a query and has a q-partition equal to that of $Q$. Note that the argument $X$ of \textsc{minQ} is the subset of $Q$ that has most recently been added to $QB$.

For each call in line~\ref{algoline:query:recursive_call1} or line~\ref{algoline:query:recursive_call2}, the input $Q$ to \textsc{minQ} is recursively analyzed until a trivial case arises, i.e.\ (a)~until $Q$ is identified to be irrelevant for the computed minimized query wherefore $\emptyset$ is returned (lines~\ref{algoline:query:validitytest2} and \ref{algoline:query:return_emptyset}) or (b)~until $|Q|=1$ and $Q$ is not irrelevant for the computed minimized query wherefore $Q$ is returned (lines~\ref{algoline:query:test_singleton} and \ref{algoline:query:return_Q}).

\begin{example}\label{example:ad_Table_of_queries_partitions}
Let us reconsider the FOL DPI depicted by Table~\ref{tab:example1} on page~\pageref{tab:example1}. 
We recall that sets of minimal conflict sets and minimal diagnoses w.r.t.\ this DPI were given by $\minC_{\tuple{\mo,\mb,\Tp,\Tn}_\RQ} = \setof{\mc_1,\mc_2} = \setof{\tuple{1,3,4},\tuple{1,2,3,5}}$ as well as $\minD_{\tuple{\mo,\mb,\Tp,\Tn}_\RQ} = \setof{\md_1,\md_2,\md_3,\md_4} = \setof{[1],[3],[4,5],[2,4]}$. For this DPI, a set of minimized queries computed by \textsc{getPoolOfQueries} is presented by Table~\ref{tab:queries_partitions}. Note that these queries have been produced by different \textsc{getEntailments} functions (as indicated by the dashed lines in Table~\ref{tab:queries_partitions}). That is, $Q_i$ for $i \in \setof{1,\dots,5}$ have been produced by the same \textsc{getEntailments} function that is described in Example~\ref{example:query_computation}. For $i \in \setof{6,\dots,9}$, $Q_i$ has been computed from a \textsc{getEntailments} function that outputs only explicit entailments (cf.\ Definition~\ref{def:explicit_entailment}) and $Q_{10}$ from a \textsc{getEntailments} function that returns a finite set of entailments where each entailment is some FOL formula. This could be accomplished, for example, by some resolution-based reasoning procedure \cite{chang1973}.

It is important to realize that the results regarding Algorithm~\ref{algo:query_gen} established so far, most of which depend on the particular used \textsc{getEntailments} function, must only hold within one part of Table~\ref{tab:queries_partitions} (where different parts are separated by the dashed lines). For example, for $Q_2$ and $Q_9$ it holds that $\dx{}(Q_2) = \dx{}(Q_9)$, but $\dnx{}(Q_2) \neq \dnx{}(Q_9)$ and $\dz{}(Q_2) \neq \dz{}(Q_9)$. By application of one and the same \textsc{getEntailments} function, this case would be prohibited by Proposition~\ref{prop:query_gen_dx_unique}. Furthermore, by Proposition~\ref{prop:given_dx_the_q-partition_with_min_dz_is_found}, only $Q_9$ would be an element of the query pool $\QP$ in this case since $\dz{}(Q_9) \subset \dz{}(Q_2)$.

Moreover, we want to remark that $Q_7$, $Q_8$ and $Q_9$ can be seen as a proof that $Q_6$ is indeed set-minimal. Each $Q_i, i \in \setof{7,8,9}$ is a result of the removal of a single formula from $Q_6$. And, each such $Q_i$ features a q-partition different from the one of $Q_6$. This illustrates quite well the principle of \textsc{minQ} which performs tests of exactly this kind to verify minimality of a query or detect formulas that might be deleted from it under preservation of the q-partition, respectively.

Another essential note is that it is guaranteed that $\dz{}(Q_6) = \emptyset$. This holds due to the construction of $Q_6$ as $U_{\mD}\setminus \md_4 = \setof{1,2,3,4,5}\setminus [2,4] = \setof{1,3,5}$ (recall that we use squared brackets to denote diagnoses in spite of the fact that these are sets, cf.\ Table~\ref{tab:abbreviations}). So, $Q_6$ comprises all formulas occurring in minimal diagnoses except for the ones contained in $\md_4$. We have that for any two different minimal diagnoses $\md_i, \md_j$ w.r.t.\ one and the same DPI it must be true that $\md_i \setminus \md_j \neq \emptyset$ as well as $\md_j \setminus \md_i \neq \emptyset$ as otherwise one would be necessarily a subset of the other. From this, we can easily derive that $\mo^*_i \cup Q_6$ for $i \in \setof{1,\dots,3}$, i.e.\ for all minimal diagnoses $\md_i$ w.r.t.\ this DPI other than $\md_4$ which was used to build the query $Q_6$, must comprise a conflict set. This must be valid by the minimality of $\md_i$ and since by $Q_6$ at least one formula of $\md_i$ is readded to the KB. Note that a similar argumentation was used in the proof of Proposition~\ref{prop:query_gen_explicit_entailments_dz_empty}.\qed
\end{example}

\renewcommand{\arraystretch}{1.4} 
\begin{table*}[!htb]
\footnotesize
\centering
\begin{tabular}{@{\extracolsep{2pt}}lllll}
$i$ & Query $Q_i$ & $\dx{}(Q_i)$  & $\dnx{}(Q_i)$ &  $\dz{}(Q_i)$ \\ \hline
$1$ & $\{\forall X b(X) \rightarrow m_3(X)\}$ & $\{\md_1,\md_2,\md_4\}$ & $\{\md_3\}$ & $\emptyset$\\
$2$ & $\{b(w)\}$ & $\{\md_3, \md_4\}$ & $\{\md_2\}$ & $\{\md_1\}$ \\
$3$ & $\{\forall X m_1(X) \rightarrow b(X)\}$ & $\{\md_1,\md_3,\md_4\}$ & $\{\md_2\}$ & $\emptyset$\\
$4$ & $\{m_1(w), m_2(u)\}$ & $\{\md_2,\md_3,\md_4\}$ & $\{\md_1\}$ & $\emptyset$  \\
$5$ & $\{a(w)\}$ & $\{\md_2\}$ & $\{\md_3, \md_4\}$ & $\{\md_1\}$\\ \hdashline
$6$ & $\{\tax_1,\tax_3,\tax_5\}$ & $\{\md_4\}$ & $\{\md_1,\md_2, \md_3\}$ & $\emptyset$\\
$7$ & $\{\tax_3,\tax_5\}$ & $\{\md_1, \md_4\}$ & $\{\md_2,\md_3\}$ & $\emptyset$\\
$8$ & $\{\tax_1,\tax_5\}$ & $\{\md_2, \md_4\}$ & $\{\md_1,\md_3\}$ & $\emptyset$\\
$9$ & $\{\tax_1,\tax_3\}$ & $\{\md_3, \md_4\}$ & $\{\md_1,\md_2\}$ & $\emptyset$\\ \hdashline
\multirow{2}{*}{$10$} & $\{\forall X m_1(X) \rightarrow \lnot a(X),$ & \multirow{2}{*}{$\{\md_1\}$} & \multirow{2}{*}{$\{\md_2,\md_3, \md_4\}$} & \multirow{2}{*}{$\emptyset$}\\
    & $\forall X m_2(X) \rightarrow (\forall Y s(X,Y) \rightarrow a(Y))\}$ & \\
\hline
\end{tabular}
\caption[Queries and Associated Q-Partitions]{Some queries and associated q-partitions for the DPI given by Table~\ref{tab:example1}.}
\label{tab:queries_partitions}
\end{table*}

\subsection{Soundness of Query Minimization}
\label{sec:SoundnessOfQueryMinimization}
The following lemma shows that the function \textsc{isQPartConst} used by \textsc{minQ} is indeed a monotonic function (cf.\ Definition~\ref{def:monotonic}), which is a necessary prerequisite for versions of the $\scQX$ algorithm to work in a sound way. 
\begin{lemma}\label{lem:isqpartconst_monotonic}
Let $\langle\mo,\mb,\Tp,\Tn\rangle_\RQ$ be a DPI, $\mD \subseteq \minD_{\langle\mo,\mb,\Tp,\Tn\rangle_\RQ}$, $Q\in\mQ_{\mD,\langle\mo,\mb,\Tp,\Tn\rangle_\RQ}$ with q-partition $\Pt(Q)$. Further, let $f: 2^Q \rightarrow \setof{0,1}$ be a function that maps a subset $QB$ of $Q$ to 1 if $QB$ has q-partition $\Pt(QB) = \Pt(Q)$, to 0 otherwise. Then, $f$ is a monotonic function (as per Definition~\ref{def:monotonic}).
\end{lemma}
\begin{proof}
Assume a subset $Q'$ of $Q$ with $f(Q') = 1$, i.e.\ $Q'$ has q-partition $\Pt(Q') = \Pt(Q)$. Let $Q' \subset Q'' \subseteq Q$ and assume that $f(Q'') = 0$, i.e.\ $Q''$ has a q-partition $\Pt(Q'') \neq \Pt(Q)$. 

As shown in the proof of Lemma~\ref{lem:isqpartconst_correct}, $\dx{}(X_1) \supseteq \dx{}(X_2)$ holds for any $X_1 \subseteq X_2$. Therefore, we have $\dx{}(Q') \supseteq \dx{}(Q'') \supseteq \dx{}(Q)$ and by $\Pt(Q') = \Pt(Q)$ that $\dx{}(Q') = \dx{}(Q)$ and thus that all $\supseteq$-relations are $=$-relations. So, either $\dnx{}(Q'') \neq \dnx{}(Q)$ or $\dz{}(Q'') \neq \dz{}(Q)$ must hold.

First, assume that $\dnx{}(Q'') \neq \dnx{}(Q)$. Then, as $\mo_r^* \cup Q'' \subset \mo_r^* \cup Q$ and by monotonicity of $\mathcal{L}$, it can only be the case that for some $\md_r \in \mD$ some $x \in \RQ \cup \Tn$ that is violated for $\mo_r^* \cup Q$ is \emph{not} violated for $\mo_r^* \cup Q''$. Hence, $\dnx{}(Q'') \subset \dnx{}(Q)$ must hold. By a similar argumentation -- \emph{without} the assumption that $\dnx{}(Q') \neq \dnx{}(Q'')$ holds -- we have that $\dnx{}(Q') \subseteq \dnx{}(Q'')$ and thus, altogether, that $\dnx{}(Q') \subset \dnx{}(Q)$ must be true. Due to $\Pt(Q') = \Pt(Q)$ we know that $\dnx{}(Q') = \dnx{}(Q)$ which is a contradiction.

Finally, assume that $\dz{}(Q'') \neq \dz{}(Q)$. Since $\mo_r^* \cup Q$ does not violate any $x \in \RQ \cup \Tn$ for $\md_r\in\dz{}(Q)$, $\mo_r^* \cup Q''$ cannot violate any $x \in \RQ \cup \Tn$ by monotonicity of $\mathcal{L}$. As a conclusion, the only possibility for $\dz{}(Q'') \neq \dz{}(Q)$ is that $\mo_r^* \models Q''$ for some $\md_r\in\dz{}(Q)$, i.e.\ that $\md_r \in \dx{}(Q'')$ which implies that $\dz{}(Q'') \subset \dz{}(Q)$. By a similar argumentation -- \emph{without} the assumption that $\dz{}(Q') \neq \dz{}(Q'')$ holds -- we have that $\dz{}(Q') \subseteq \dz{}(Q'')$ and thus, altogether, that $\dz{}(Q') \subset \dz{}(Q)$ must be true. Due to $\Pt(Q') = \Pt(Q)$ we know that $\dz{}(Q') = \dz{}(Q)$ which is a contradiction.  

This completes the proof for monotonicity of the given function $f$.
\end{proof} 
\begin{proposition}[Correctness of \textsc{minQ}]\label{prop:minQ_correctness}
Given a query $Q\in\mQ_{\mD,\tuple{\mo,\mb,\Tp,\Tn}_\RQ}$ as input, \textsc{minQ} computes a subset $Q_{\min} \subseteq Q$ such that $\Pt(Q_{\min}) = \Pt(Q)$ and there is no $Q' \subset Q_{\min}$ such that $\Pt(Q') = \Pt(Q)$. 
\end{proposition}
\begin{proof}
This proposition is a consequence of the correctness of $\scQX$ shown by Proposition~\ref{prop:qx_correctness}, of the correctness of function \textsc{isQPartConst} established by Lemma~\ref{lem:isqpartconst_correct} and of the monotonicity of the property tested by the function \textsc{isQPartConst} guaranteed by Lemma~\ref{lem:isqpartconst_monotonic}.
\end{proof}

\subsection{Complexity of Query Pool Generation}
\label{sec:query_gen_complexity}
The complexity of query minimization, i.e.\ one call to \textsc{minQ}, in terms of calls to the \textsc{isQPartConst} function is directly obtained from the complexity results for the standard $\scQX$ algorithm given by Proposition~\ref{prop:qx_complexity}.
\begin{proposition}[Complexity of \textsc{minQ}]\label{prop:minQ_complexity}
Let $\langle\mo,\mb,\Tp,\Tn\rangle_\RQ$ be a DPI, $\mD\subseteq\minD_{\langle\mo,\mb,\Tp,\Tn\rangle_\RQ}$, $Q \in \mQ_{\mD,\langle\mo,\mb,\Tp,\Tn\rangle_\RQ}$ with $\Pt(Q) = \tuple{\dx{}(Q), \dnx{}(Q), \dz{}(Q)}$ and the function \textsc{split} (line~\ref{algoline:query:split} of Algorithm~\ref{algo:query_gen}) be defined as $\textsc{split}(n) = \lfloor \frac{n}{2}\rfloor$ where $n$ is a natural number. Then, the worst case number of calls to \textsc{isQPartConst} during one call to $\textsc{minQ}(\emptyset,Q,\emptyset,\Pt(Q),\langle\mo,\mb,\Tp,\Tn\rangle_\RQ)$ is in 
\begin{align*}
O\left(\left|Q_{\min}\right|\log \frac{|Q|}{\left|Q_{\min}\right|}\right)
\end{align*}
where $Q_{\min}$ is the output of $\textsc{minQ}(\emptyset,Q,\emptyset,\Pt(Q),\langle\mo,\mb,\Tp,\Tn\rangle_\RQ)$.

For any other definition of the function \textsc{split}, the worst case number of calls to \textsc{isQPartConst} gets larger.
\end{proposition}
%
The overall complexity of \textsc{getPoolOfQueries} in terms of calls to functions that call the reasoner, i.e.\ functions \textsc{getEntailments}, \textsc{isKBValid} and \textsc{isQPartConst}, is established by the following proposition.
\begin{proposition}[Complexity of \textsc{getPoolOfQueries}]\label{prop:query_gen_complexity}
Let $\tuple{\mo,\mb,\Tp,\Tn}_\RQ$ be a DPI, $q$ a natural number and $\mD \subseteq \minD_{\langle\mo,\mb,\Tp,\Tn\rangle_\RQ}$. Then, the worst case number of calls to functions that call a reasoner during one call to $\textsc{getPoolOfQueries}(\langle\mo,\mb,\Tp,\Tn\rangle_\RQ, \mD, q)$ is in 
\begin{align*}
O\left(\left(|\mD|+\left|Q_{\min}^{(\max)}\right|\log \frac{\left|Q^{(\max)}\right|}{\left|Q_{\min}^{(\max)}\right|}\right)2^{|\mD|}\right)
\end{align*}
where $\left|Q^{(\max)}\right|$ is the maximum size of a query before minimization, i.e.\ the size of the set of maximum cardinality that is stored in variable $Q$ in line~\ref{algoline:query:minQ} throughout all iterations, and $\left|Q_{\min}^{(\max)}\right|$ is the maximum size of a minimized query, i.e.\ the size of the set of maximum cardinality that is stored in variable $Q'$ in line~\ref{algoline:query:minQ} throughout all iterations.
\end{proposition}
\begin{proof}
During the execution of the for-loop over lines~\ref{algoline:query:ent_start}-\ref{algoline:query:ent_end} the function \textsc{getEntailments} is called $|\mD|$ times. During the execution of the for-loop over lines~\ref{algoline:query:seed}-\ref{algoline:query:return_QP_1} which may be executed at most $2^{|\mD|}-2$ times, \textsc{isKBValid} is called at most $|\mD|-1$ times since $|\mS| \geq 1$ and $\mS \subset \mD$ and thus $|\mD\setminus\mS| \leq |\mD|-1$ holds; furthermore, \textsc{minQ} may be called once, namely if the condition tested by the if-statement in line~\ref{algoline:query:add_QP_start} is true. During one execution of \textsc{minQ}, by Proposition~\ref{prop:minQ_complexity}, at most 
\begin{align*}
|Q_{\min}|\log \frac{|Q|}{|Q_{\min}|}
\end{align*}
calls to \textsc{isQPartConst} are made where $Q_{\min}$ is the output of the call to $\textsc{minQ}$. So, an upper bound of the number of calls to \textsc{isQPartConst} performed by one call to \textsc{minQ} among all calls to \textsc{minQ} throughout the execution of \textsc{getPoolOfQueries}, is 
\begin{align*}
\left|Q_{\min}^{(\max)}\right|\log \frac{\left|Q^{(\max)}\right|}{\left|Q_{\min}^{(\max)}\right|}
\end{align*}
where $\left|Q_{\min}^{(\max)}\right|$ is the set of maximum cardinality that is stored in variable $Q'$ in line~\ref{algoline:query:minQ} throughout all iterations and $\left|Q^{(\max)}\right|$ is the set of maximum cardinality that is stored in variable $Q$ in line~\ref{algoline:query:minQ} throughout all iterations. 

So, all in all we know that functions that call a reasoner are invoked at most 
\begin{align*}
|\mD| + \left(|\mD|-1 + \left|Q_{\min}^{(\max)}\right|\log \frac{\left|Q^{(\max)}\right|}{\left|Q_{\min}^{(\max)}\right|}\right) (2^{|\mD|} - 2)
\end{align*}
times during the execution of \textsc{getPoolOfQueries}. Since 
\begin{align*}
\left(|\mD|+\left|Q_{\min}^{(\max)}\right|\log \frac{\left|Q^{(\max)}\right|}{\left|Q_{\min}^{(\max)}\right|}\right)2^{|\mD|}
\end{align*}
is an upper bound of this number, the proposition holds.
\end{proof}
Note that none of the parameters that affect the complexity of the function \textsc{getPoolOfQueries} grows with the size of the DPI provided as an input to the interactive KB debugging problem. Merely the costs for reasoning, where a black-box debugging approach has no influence on, are affected by a higher complexity or larger size of the input DPI. Moreover, the size of the most relevant parameter influencing the worst case complexity, namely the exponent $|\mD|$, can be specified by the user to any value greater or equal to 2. In other words, minus reasoning time, the generation of a pool of queries is a fixed parameter tractable problem~\cite{downey1995} in the context of interactive KB debugging.

\subsection{Shortcomings of Query Pool Generation}
\label{sec:query_gen_shortcomings}
First, the exponential time complexity regarding the parameter $|\mD|$ is a problem arising from the paradigm of computing an optimal query w.r.t.\ a certain quantitative measure $qsm()$ such as information gain~\cite{Shchekotykhin2012, Rodler2013} by calculating a (generally exponentially large) pool $\QP$ of queries in a first stage, whereupon $qsm(Q) \in \mathbb{R}$ is evaluated for $Q\in\QP$ until the one $Q^*$ with optimal $qsm(Q^*)$ is found and selected as the query to be asked to the user.
 
A key to solving this issue is the use of a different paradigm that does not rely on the computation of the pool $\QP$. Instead, qualitative measures can be derived from quantitative measures that have been used in interactive debugging scenarios~\cite{Shchekotykhin2012, Rodler2013, ksgf2010}. These qualitative measures provide a way to estimate the $qsm()$ value of \emph{partial q-partitions}, i.e.\ ones where not all leading diagnoses have been assigned to the respective set in the q-partition yet. That way a \emph{direct} search for a query with (nearly) optimal properties is possible. A similar strategy called CKK has been employed in~\cite{Shchekotykhin2012} for the information gain measure (see Section~\ref{sec:query_selection_measures}). From such a technique we can expect to save a high number of reasoner calls. Because only a usually small subset of q-partitions included in the pool computed by \textsc{getPoolOfQueries} is required to find a query with desirable properties if the search is implemented by means of a heuristic that involves the exploration of seemingly favorable (potential) queries and (partial) q-partitions, respectively, first. This is a topic of future work.

Another shortcoming of \textsc{getPoolOfQueries} is the extensive use of reasoning services which may be computationally expensive (depending on the given DPI). Instead of computing a set of common entailments $Q$ of a set of KBs $\mo_i^*$ first and consulting a reasoner to fill up the (q-)partition for $Q$ in order to test whether $Q$ is a query at all, the idea enabling a significant reduction of reasoner dependence is to compute some kind of \emph{canonical query} without a reasoner and use simple set comparisons to decide whether the associated partition is a q-partition. Guided by qualitative properties mentioned before, a search for such q-partition with desirable properties can be accomplished \emph{without reasoning at all}. Also, a set-minimal version of the optimal canonical query can be computed without reasoning aid. Only for the optional enrichment of the identified optimal canonical query by additional entailments and for the subsequent minimization of the enriched query, the reasoner may be employed. This is also a topic of future work.

Another aspect that can be improved is that \emph{only one} minimized version of each query is computed by Algorithm~\ref{algo:query_gen}. That is, per q-partition $\Pt$, there might be some set-minimal queries which do not occur in the output set $\QP$. From the point of view of how well a query might be understood by an interacting user, of course not all minimized queries can be assumed equally good in general. Hence, in order to avoid a situation where a potentially best-understood query w.r.t.\ $\Pt$ is not included in $\QP$, the query minimization process (see Section~\ref{sec:minimization_of_queries}) might be adapted to take into account some information about faults the interacting user is prone to. This could be exploited to estimate how well this user might be able to understand and answer a query. For instance, given that the user frequently has problems to apply $\exists$ in a correct manner to express what they intend to express, but has never made any mistakes in formulating implications $\rightarrow$, then the query $Q_1 = \setof{\forall X\,p(X) \rightarrow q(X), r(a)}$ might be better comprehended than $Q_2 = \setof{\forall X \exists Y s(X,Y)}$. One way to achieve the finding of a well-understood query for some q-partition $\Pt$ is to run the query minimization \textsc{minQ} more than once, each time with a modified input (using a hitting set tree to accomplish this in a systematic manner -- cf.\ Chapter~\ref{chap:DiagnosisComputation}, where an analogue idea is used to compute different minimal conflict sets w.r.t.\ a DPI). In this way, different set-minimal queries for $\Pt$ can be identified and the process can be stopped when a suitable query is found.     
\subsection{Correctness of Query Pool Generation}
\label{sec:query_gen_correctness}
The following proposition confirms the correctness of Algorithm~\ref{algo:query_gen}, i.e.\ of the function \textsc{getPoolOfQueries}. Roughly, it states that the output of $\QP$ of the function is duplicate-free, i.e.\ no query or q-partition occurs twice in $\QP$, that $\QP$ includes \emph{only queries} and q-partitions, that tuples in $\QP$ are unique w.r.t.\ the set $\dx{}$ of a q-partition and that, given $q > |\QP|$, there is no subset $Y$ of $\mD$ for which a q-partition with $\dx{} = Y$ exists and for which no q-partition with $\dx{} = Y$ is an element of $\QP$. 
\begin{proposition}\label{prop:getPoolOfQueries_correctness}
Let a DPI $\tuple{\mo,\mb,\Tp,\Tn}_\RQ$, $\mD \subseteq \minD_{\tuple{\mo,\mb,\Tp,\Tn}_\RQ}$ such that $|\mD| \geq 2$ and some $q \in \mathbb{N} \cup \setof{\infty}, q \geq 1$  be the inputs to \textsc{getPoolOfQueries} and let $|\QP_{\max}| \geq 0$ be the maximum number of tuples $\tuple{Q,\Pt(Q)}$ that can be computed by \textsc{getPoolOfQueries} by means of the used \textsc{getEntailments} function. If $q \geq |\QP_{\max}|$ (in particular $q=\infty$), then 
\begin{enumerate}
\item there are no two tuples $\tuple{Q,\Pt(Q)}, \tuple{Q',\Pt(Q')}$ in $\QP$ such that $Q = Q'$ \emph{or} $\Pt(Q)=\Pt(Q')$, and
\item $\QP$ includes a tuple $\tuple{Q,\tuple{\dx{}(Q),\dnx{}(Q),\dz{}(Q)}}$ only if $Q\in\mQ_{\mD,\tuple{\mo,\mb,\Tp,\Tn}_\RQ}$, and  
\item $\QP$ includes at most one tuple where $\dx{}(Q) = Y$ for each $Y \subset \mD$, and 
\item for each $Y \subset \mD$ for which a query $Q$ w.r.t.\ $\mD$ and $\tuple{\mo,\mb,\Tp,\Tn}_\RQ$ exists such that 
\begin{enumerate}
	\item $Q$ includes only entailments computed by the used \textsc{getEntailments} function and
	\item $\Pt(Q)$ is such that $\dx{}(Q) = Y$,
\end{enumerate}
$\QP$ includes a tuple $\tuple{Q',\Pt(Q')}$ such that $\dx{}(Q') = Y$, and
\item $\QP \neq \emptyset$.
\end{enumerate}
If $q < |\QP_{\max}|$, then $\QP$ includes $q$ tuples satisfying (1), (2) and (3).
\end{proposition}
\begin{proof}
Statement~(1) is a consequence of Proposition~\ref{prop:query_gen_duplicate_free}. Statement~(2) is an implication of Proposition~\ref{prop:query_gen_isQuery_correct} and Proposition~\ref{prop:minQ_correctness}. The former says that only sets $Q$ that are actually queries w.r.t.\ $\mD$ and $\tuple{\mo,\mb,\Tp,\Tn}_\RQ$ can pass line~\ref{algoline:query:add_QP_start}. Thus, only queries are passed to \textsc{minQ} as parameter $Q$. By the latter which states that \textsc{minQ} is correct, i.e.\ outputs a query if the input is a query, statement~(2) follows. Statement~(3) follows from Proposition~\ref{prop:query_gen_dx_unique}. 
If $q \geq |\QP_{\max}|$, the truth of statement~(4) is witnessed by Proposition~\ref{prop:query:dx_complete}. Statement~(5) is true by lines~\ref{algoline:query:check_QP_empty} and \ref{algoline:query:addTrivialQueries} and by Proposition~\ref{prop:q1} as well as Corollary~\ref{cor:query_num_lower_bound} and the premise that $|\mD| \geq 2$ which guarantee that the function \textsc{addTrivialQueries} always adds at least $|\mD| \geq 2 > 0$ queries to $\QP$. In case $q < |\QP_{\max}|$, only statements (1), (2) and (3) are satisfied in general (for the same reasons as given above for the case $q \geq |\QP_{\max}|$) and $\QP$ is returned in line~\ref{algoline:query:return_QP_1} by the definition of $|\QP_{\max}|$. Thence, the condition $|\QP| = q \geq 1$ tested in line~\ref{algoline:query:test_QP} must be valid for $\QP$.   
\end{proof}

\begin{algorithm*}
\small
\caption{Interactive KB Debugging}\label{algo:inter_onto_debug}
\begin{algorithmic}[1]
\Require a tuple $\tuple{ \langle\mo,\mb,\Tp,\Tn\rangle_\RQ, n_{\min}, n_{\max}, t, p_{\widetilde{\mo}\cup\overline{\mo}}, q, qsm(),\sigma, mode}$ consisting of
\begin{itemize}
\item an admissible DPI $\langle\mo,\mb,\Tp,\Tn\rangle_\RQ$, 
\item leading diagnoses computation parameters, natural numbers $n_{\min} \geq 2, n_{\max}, t$, 
\item a function $p_{\widetilde{\mo}\cup\overline{\mo}}: \widetilde{\mo}\cup\overline{\mo} \rightarrow (0,1]$,
\item a parameter $q \in \mathbb{N}\cup\setof{\infty}, q \geq 1$ that determines the size of the computed query pool,
\item a function $qsm(Q) \in \mathbb{R}$ used for query selection that assigns a real number to a query $Q$ to express the ``goodness'' of $Q$, 
\item a maximum fault tolerance $\sigma \in [0,1]$ and 
\item a mode $mode \in \setof{static,dynamic}$ that determines the used method for diagnosis computation. 
\end{itemize}
\Ensure The output depends on $mode$ and $\sigma$:
\begin{itemize}
\item $mode = static$: 
a maximal solution KB w.r.t.\ the input DPI $\langle\mo,\mb,\Tp,\Tn\rangle_\RQ$ which is
\begin{itemize}
	\item an approximation of the solution to Interactive Static KB Debugging (Problem Def.~\ref{prob_def:static}) if $\sigma > 0$. 
	\item the (exact) solution to Interactive Static KB Debugging if $\sigma = 0$. 
\end{itemize}
\item $mode = dynamic$: 
a maximal solution KB w.r.t.\ the current DPI $\langle\mo,\mb,\Tp\cup\Tp',\Tn\cup\Tn'\rangle_\RQ$ which is
\begin{itemize}
	\item an approximation of the solution to Interactive Dynamic KB Debugging (Problem Def.~\ref{prob_def:dynamic}) if $\sigma > 0$. 
	\item the (exact) solution to Interactive Dynamic KB Debugging if $\sigma = 0$. 
\end{itemize}
\end{itemize}
(for a more formal and precise characterization of the output see Proposition~\ref{prop:correctness_of_interactive_KB_debugging_algo} on page~\pageref{prop:correctness_of_interactive_KB_debugging_algo}) 
%
\newline

\State $\Tp', \Tn', \mathbf{C}_{calc}, \mD_{\checkmark}, \mD_{\times}, \mD_{out}, \mD_{\supset}, qData \gets \emptyset$ \label{algoline:inter_onto_debug:var_inst_start}
\State $\Queue_{dup}, QA \gets []$
\State $\Queue \gets [\emptyset]$ \label{algoline:inter_onto_debug:Q={{}}}
\State $answer \gets \false$ \label{algoline:inter_onto_debug:var_inst_end}
\State $p_{\mo}() \gets \Call{getFormulaProbs}{\mo, p_{\widetilde{\mo}\cup\overline{\mo}}()}$  \label{algoline:inter_onto_debug:getAxiomProbs}	\Comment{application of Formulas~\ref{eq:ax_prob_calc} and \ref{eq:adapt_ax_prob_to_get_min_diags}}
\While{\true}  \label{algoline:inter_onto_debug:while}
\If{$mode = static$} \Comment{see Algorithm~\ref{algo:inter_stat_hs}}
	\State $\tuple{\mD_{\checkmark},\Queue, \mathbf{C}_{calc}, \mD_{\times}} \gets \Call{staticHS}{}(\langle\mo,\mb,\Tp,\Tn\rangle_\RQ, \Queue, t, n_{\min}, n_{\max}, $ 
	\Statex \qquad\qquad\qquad\qquad\quad\qquad\qquad\qquad\qquad\quad\; $\mathbf{C}_{calc}, \mD_{\checkmark}, \mD_{\times}, p_{\mo}(), \Tp', \Tn')$\label{algoline:inter_onto_debug:staticHS}
\Else \Comment{see Algorithms~\ref{algo:inter_dyn_hs}, \ref{algo:update_tree} and \ref{algo:prune}}
\State $\tuple{\mD_{\checkmark},\Queue, \mathbf{C}_{calc}, \mD_{\times}, \mD_{\supset},\Queue_{dup}} \gets
	\Call{dynamicHS}{}(\langle\mo,\mb,\Tp,\Tn\rangle_\RQ, \Queue, \Queue_{dup}, t, n_{\min}, n_{\max},$  
	\Statex \qquad\qquad\qquad\qquad\quad\qquad\quad\qquad\qquad\qquad\qquad\qquad\qquad\; $\mathbf{C}_{calc},\mD_{\checkmark}, \mD_{\times}, p_{\mo}(), \Tp', \Tn', \mD_{\supset})$ \label{algoline:inter_onto_debug:dynamicHS}
\EndIf
\State $p_{\mD}() \gets \Call{getProbDist}{\mD_{\checkmark},p_{\mo}(),\langle\mo,\mb,\Tp,\Tn\rangle_\RQ,QA}$\label{algoline:inter_onto_debug:getProbDist}  \Comment{see Algorithm~\ref{algo:inter_onto_debug_continued}}
\State $\md_{\max} \gets \Call{getMode}{\mD_{\checkmark},p_{\mD}()}$  \label{algoline:inter_onto_debug:getMode}
\If{$p_{\mD}(\md_{\max}) \geq 1-\sigma$}     \label{algoline:inter_onto_debug:stop_crit} \Comment{stop criterion}
	\State \Return $\Call{getSolKB}{\md_{\max},\langle\mo,\mb,\Tp\cup\Tp',\Tn\cup\Tn'\rangle_\RQ, \Tp', mode}$\label{algoline:inter_onto_debug:return} \Comment{return solution KB}
\Else
	\State $\tuple{Q,\Pt(Q)} \gets \Call{calcQuery}{}(\mD_{\checkmark},qData,p_{\mD}(),p_{\widetilde{\mo}\cup\overline{\mo}}(), qsm(),$
	\Statex \qquad\qquad\qquad\qquad\qquad\qquad\qquad\quad\;$\langle\mo,\mb,\Tp\cup\Tp',\Tn\cup\Tn'\rangle_\RQ, q)$\label{algoline:inter_onto_debug:calc_query} \Comment{see Algorithm~\ref{algo:inter_onto_debug_continued}}
	\State $answer \gets u(Q)$\label{algoline:inter_onto_debug:user_interaction}    \Comment{user interaction}
	\State $QA \gets \Call{append}{\tuple{Q,answer},QA} $\label{algoline:inter_onto_debug:param_update_start}
	\State $\mD_{out} \gets \Call{getInvalidDiags}{\Pt(Q), answer}$\label{algoline:inter_onto_debug:get_invalid_diags}
	\State $qData \gets \Call{updateQData}{\mD_{out}, \mD_{\checkmark}, answer}$
	\State $\mD_{\checkmark} \gets \mD_{\checkmark} \setminus \mD_{out}$\label{algoline:inter_onto_debug:update_D_checkmark}
	\State $\mD_{\times} \gets \mD_{\times} \cup \mD_{out}$\label{algoline:inter_onto_debug:update_D_times}
	\If{$answer = \true$}
		\State $\Tp' \gets \Tp' \cup \setof{Q}$\label{algoline:inter_onto_debug:add_pos_tc}
	\Else
		\State $\Tn' \gets \Tn' \cup \setof{Q}$\label{algoline:inter_onto_debug:param_update_end}
	\EndIf
\EndIf
\EndWhile
\algstore{save_algo_inter_onto_debug}
\end{algorithmic}
\normalsize
\end{algorithm*}
\restoregeometry

\begin{algorithm*}
\small
\caption{Interactive KB Debugging (continued)}\label{algo:inter_onto_debug_continued}
\begin{algorithmic}[1]
\algrestore{save_algo_inter_onto_debug}
\Procedure{\textsc{getProbDist}}{$\mD_{\checkmark},p_{\mo}(),\langle\mo,\mb,\Tp,\Tn\rangle_\RQ,QA$}
\State $\Tp'', \Tn'' \gets \emptyset$
\State $p_{\mD,prio}() \gets \Call{getPrioDiagProbs}{\mD_{\checkmark},p_{\mo}(),\langle\mo,\mb,\Tp,\Tn\rangle_\RQ}$ \label{algoline:get_prob_dist:getPrioDiagProbs} \Comment{application of Formula~\ref{eq:diag_prob_calc}}
\For{$\tuple{Q,u(Q)} \in QA$}\label{algoline:get_prob_dist:for-loop_start}  \Comment{run through chronologically sorted query-answer pairs}
	\If{$u(Q) = \true$}
		\For{$\md_r \in \mD_{\checkmark}$}   \Comment{function \textsc{getEntailments} is defined on page~\pageref{etc:definition_of_getEntailments_function}}
			\State $E_{\md_r} \gets \Call{getEntailments}{\md_r,\mo,\mb,\Tp\cup\Tp''}$   \Comment{$E_{\md_r}$ is a set of entailments of 
			 $\mo^{*}_r$}
			\If{$Q \not\subseteq E_{\md_r}$}   \Comment{$\md_r \in \dz{}(Q)$} \label{algoline:get_prob_dist:check_dz_1}
				\State $p_{\mD,prio}(\md_r) \gets \frac{1}{2}\, p_{\mD,prio}(\md_r)$
			\EndIf
		\EndFor
		\State $\Tp'' \gets \Tp'' \cup \setof{Q}$    \label{algoline:get_prob_dist:update_Tp''}
	\Else
		\For{$\md_r \in \mD_{\checkmark}$}   \Comment{\textsc{isKBValid} (see Algorithm~\ref{algo:qx})}
			\If{$\Call{isKBValid}{(\mo\setminus\md_r) \cup Q, \tuple{\cdot,\mb,\Tp\cup\Tp'',\Tn\cup\Tn''}_{\RQ}}$}   \Comment{$\md_r \in \dz{}(Q)$}     \label{algoline:get_prob_dist:check_dz_2}
				\State $p_{\mD,prio}(\md_r) \gets \frac{1}{2}\, p_{\mD,prio}(\md_r)$
			\EndIf
		\EndFor
		\State $\Tn'' \gets \Tn'' \cup \setof{Q}$   \label{algoline:get_prob_dist:update_Tn''}
	\EndIf
\EndFor\label{algoline:get_prob_dist:for-loop_end}
\State $sum \gets \sum_{\md_r \in \mD_{\checkmark}} p_{\mD,prio}(\md_r)$\label{algoline:get_prob_dist:summation} 
\For{$\md_r \in \mD_{\checkmark}$}\label{algoline:get_prob_dist:for-loop2_start}
	\State $p_{\mD,prio}(\md_r) \gets \frac{1}{sum}\, p_{\mD,prio}(\md_r)$ \label{algoline:get_prob_dist:normalize} \Comment{normalization}
\EndFor
\State \Return $p_{\mD,prio}()$
\EndProcedure
\vspace{10pt}
\Procedure{\textsc{calcQuery}}{$\mD_{\checkmark},qData,p_{\mD}(),p_{\widetilde{\mo}\cup\overline{\mo}}(),qsm(),\langle\mo,\mb,\Tp\cup\Tp',\Tn\cup\Tn'\rangle_\RQ, q$}
\State $\QP \gets \Call{getPoolOfQueries}{}(\langle\mo,\mb,\Tp\cup\Tp',\Tn\cup\Tn'\rangle_\RQ, \mD_{\checkmark}, q)$ \Comment{see Algorithm~\ref{algo:query_gen}}
\State \Return $\Call{selectBestQuery}{}(\QP, qData,p_{\mD}(),p_{\widetilde{\mo}\cup\overline{\mo}}(),qsm())$ \label{algoline:inter_onto_debug_continued:selectBestQuery} \Comment{see Section~\ref{sec:query_selection_measures}}
\EndProcedure
\end{algorithmic}
\normalsize
\end{algorithm*}

\section{An Algorithm for Interactive Knowledge Base Debugging}
\label{sec:WorkflowInInteractiveOntologyDebugging}
In this section we will give a description of an algorithm for interactive KB debugging (Algorithm~\ref{algo:inter_onto_debug}) which implements the entire functionality required by an interactive debugging system. All other algorithms presented so far will be subroutines of Algorithm~\ref{algo:inter_onto_debug} which are either directly or indirectly called by it. Before we explain and discuss Algorithm~\ref{algo:inter_onto_debug} in detail, we give the reader a rough and informal overview of the algorithm's input, output and actions in the following section in order to make the details of the algorithm easier to digest. 

\begin{remark}\label{rem:what_we_mean_by_currentDPI_inputDPI_intermediateDPI}
Note, in the following, when we speak of \emph{the input DPI} we refer to the DPI $\tuple{\mo,\mb,\Tp,\Tn}_\RQ$ that is provided as an input to Algorithm~\ref{algo:inter_onto_debug}, by \emph{the current DPI} we mean the DPI $\tuple{\mo,\mb,\Tp\cup\Tp',\Tn\cup\Tn'}_\RQ$ where $\Tp'$ and $\Tn'$, respectively, are all positive and negative test cases added to the input DPI from the start of the algorithm's execution until the current point in time. Further on, \emph{an intermediate (or previous) DPI} denotes a DPI $\tuple{\mo,\mb,\Tp\cup\Tp'',\Tn\cup\Tn''}_\RQ$ which is not the current DPI and where $\emptyset \subseteq \Tp'' \subseteq \Tp'$ and $\emptyset \subseteq \Tn'' \subseteq \Tn'$. Finally, \emph{the last-but-one DPI} corresponds to an intermediate DPI $\tuple{\mo,\mb,\Tp\cup\Tp'',\Tn\cup\Tn''}_\RQ$ where 
either $|\Tp'| = |\Tp''|+1$ or $|\Tn'| = |\Tn''|+1$ is true, but not both.\qed
\end{remark}

\subsection[Overview]{Interactive Debugging Algorithm: Overview}
\label{sec:AlgorithmOverview}
\textbf{Input:}\\
\indent An admissible DPI and some meta information where the latter consists of 
\begin{itemize}
	\item fault probabilities of syntactical elements occurring in the KB,
	\item a minimal and desired number of leading diagnoses,
	\item a desired maximum reaction time (time between two successive queries presented to the user), 
	\item a maximum fault tolerance (roughly, the probability of being presented a non-desired solution KB as output),
	\item a measure for query selection (determines which query is the best query within a given set of queries),
	\item a parameter that determines the size of the computed pool of queries in each iteration and
	\item a parameter specifying the way the hitting set tree for computation of leading diagnoses is constructed and updated.
\end{itemize}
\vspace{3pt}

\noindent\textbf{Output:}
\begin{adjustwidth}{\parindent}{}
A solution KB such that the diagnosis used to formulate the solution KB has a probability (w.r.t.\ the current leading diagnoses) greater than or equal to 1 minus the given maximum fault tolerance.
\end{adjustwidth}
\vspace{6pt}

\noindent\textbf{Procedure:}
\begin{enumerate}
		\item \emph{Initialization:} Compute the fault probability of each formula in the KB by means of the given fault probabilities.
		\item \emph{Leading Diagnoses Computation:} Use a hitting set tree constructed and updated in a manner as specified in the input coupled with $\scQX$ to calculate a set of leading diagnoses. In that, the cardinality and computation time of the set of leading diagnoses is determined by the corresponding input parameters specifying minimal and desired number of leading diagnoses and desired reaction time. 
    \item \emph{Probability Update and Stop Criterion:} Use the formula fault probabilities and the new information obtained by already specified test cases (answered queries) to compute updated (posterior) probabilities of the current leading diagnoses.
		If one diagnosis probability is greater than or equal to 1 minus the maximum fault tolerance, return the solution KB obtained by deletion of this diagnosis from the KB and subsequent addition of the union of all positive test cases.
		\item \emph{Query Generation and Selection:} Use the set of leading diagnoses (and possibly their fault probabilities) to generate a pool of queries, the size of which depends on the respective parameter provided as input. Given the pool of queries, select the best query according to the given query selection measure.
		\item \emph{User Interaction and Incorporation of New Information:} Ask the user the selected query and add it to the positive test cases in case of a positive answer and to the negative test cases otherwise.
    \item \emph{Hitting Set Tree Update:} Update the hitting set tree based on the new information given by the classification of the test case resulting from the query answer. In particular, this involves the deletion of all those minimal diagnoses that conflict with the new test case.
    \item Repeat from Step 2.
\end{enumerate}

\subsection[Detailed Description]{Interactive Debugging Algorithm: Detailed Description}
\label{sec:DetailedAlgorithmDescription}
To describe the detailed process of Algorithm~\ref{algo:inter_onto_debug}, we first characterize the input arguments, the output and the meaning of the variables used and then provide a step-by-step textual description of the actions taken by the algorithm.

\subsubsection{Input Arguments}
\label{sec:InputArguments}
The input parameters of Algorithm~\ref{algo:inter_onto_debug} are the following:
\begin{itemize}
\item An admissible DPI $\langle\mo,\mb,\Tp,\Tn\rangle_\RQ$ (cf.\ Definition~\ref{def:admissible}).  
\item Natural numbers $n_{\min} \geq 2, n_{\max}, t$ for leading diagnoses calculation (see description in Section~\ref{sec:UserInteraction} on page~\pageref{etc:leading_diag_params}). 

\emph{Remark:} The postulation $n_{\min} \geq 2$ is necessary in order for the existence of queries w.r.t.\ any computed set of leading minimal diagnoses $\mD$ and $\langle\mo,\mb,\Tp,\Tn\rangle_\RQ$ to be guaranteed (see Proposition~\ref{prop:q1}).
\item A	 function $p_{\widetilde{\mo}\cup\overline{\mo}}: \widetilde{\mo}\cup\overline{\mo} \rightarrow (0,1]$ that assigns a fault probability $p_{\widetilde{\mo}\cup\overline{\mo}}(e)$ to each $e \in \widetilde{\mo}\cup\overline{\mo}$ reflecting the degree of belief that (one occurrence of) a syntactical element $e$ appearing in $\mo$ is faulty (see Section~\ref{sec:DiagnosisProbabilitySpace}). 

\emph{Remarks:} Forbidding a probability of zero for syntactical elements assures that no formula in $\mo$ can have a probability of zero (cf.\ Remark~\ref{rem:ax_prob_not_zero}).

Recall from Section~\ref{sec:prob_space_construction} that $\widetilde{\mo}$ refers to the signature of $\mo$ (cf.\ Chapter~\ref{chap:basics})
and $\overline{\mo}$ denotes the set of all logical connectives occurring in $\mo$. From probabilities of logical connectives and elements of the signature, probabilities of formulas in $\mo$ and from those in turn probabilities of diagnoses w.r.t.\ the DPI can be derived as shown by Formulas~\ref{eq:ax_prob_calc} and \ref{eq:diag_prob_calc}.

Further note that in the description of the algorithms in this section, unlike in Section~\ref{sec:DiagnosisProbabilitySpace}, we use different denotations for probabilities of syntactical elements ($p_{\widetilde{\mo}\cup\overline{\mo}}$), formulas ($p_{\mo}()$) and diagnoses ($p_{\mD}()$) in order to make a clear distinction between these different functions.

\item A natural number $q \geq 1$ that denotes the number of queries that should be precomputed, i.e.\ the preferred size of the query pool $\QP$ (see Section~\ref{sec:QueryGeneration}), before the ``best'' tuple $\tuple{Q^*,\Pt(Q^*)}$ is selected from $\QP$. 

\emph{Remark:} In general, higher $q$ implies better quality of the selected query in terms of the query selection measure $qsm()$ (see next bullet point). The chance of locating a good query in a larger set of queries is higher. On the other hand, higher $q$ involves a worse reaction time, i.e.\ time between two successive queries. The more queries are computed, the more time the function \textsc{getPoolOfQueries} consumes.	
\item A query selection measure $qsm()$ where $qsm: \QP \rightarrow \mathbb{R}$ is a function that assigns a real-valued number $qsm(\tuple{Q,\Pt(Q)})$ to each tuple in $\QP$, often called the score of $\tuple{Q,\Pt(Q)}$. 

\emph{Remark:} $qsm()$ defines what is considered the ``best'' query in the set $\QP$, namely the query $Q^*$ in the tuple $\tuple{Q^*,\Pt(Q^*)}$ with best score among all tuples in the pool $\QP$.
Diverse measures that can be used as a $qsm()$ function in this algorithm have been discussed and evaluated within the scope of interactive KB debugging in literature~\cite{Shchekotykhin2012,Rodler2013} (for details see Section~\ref{sec:query_selection_measures}). 
\item A maximum fault tolerance $\sigma$ that defines the stop criterion of the algorithm.
That is, for a current set of leading diagnoses, the stop criterion is satisfied iff the most probable leading diagnosis has an (updated) probability of at least $1 - \sigma$ (see below for a precise definition of what ``updated'' means). 

\emph{Remark:} The smaller $\sigma$ is chosen, the higher is the chance that a desired diagnosis is found. Selecting $\sigma:=0$, i.e.\ admitting zero fault tolerance, is the safest (but also most time-consuming) way to run a debugging session with Algorithm~\ref{algo:inter_onto_debug}, as in this case the session will stop only after all but one diagnosis have been invalidated by test cases. 
\item A mode $mode \in \setof{static,dynamic}$ that determines 
\begin{enumerate}[(i)]
	\item which type of leading diagnoses are computed, i.e.\ only minimal diagnoses w.r.t.\ the input DPI ($static$) or minimal diagnoses w.r.t.\ the current DPI 
	($dynamic$),
	\item the hitting set tree pruning strategy after a query has been answered, i.e.\ conservative pruning ($static$) or invasive pruning ($dynamic$),
	\item the space and time complexity of diagnosis computation, i.e.\ not much affected by the asked queries ($static$) -- tree is almost monotonically growing, but cannot get larger in size than the complete non-interactive hitting set tree (the tree produced by Algorithm~\ref{algo:hs} with input $n_{\min} = \infty$) -- or significantly influenced by the asked queries ($dynamic$) -- tree may shrink significantly if new test cases do not introduce ``completely new'' minimal conflict sets (that are in no subset-relation with an existing one), or lead to a tree that is significantly larger than the complete non-interactive hitting set tree 
if many ``completely new'' minimal conflict sets result from the addition of new test cases. For an in-depth discussion and comparison of both strategies the reader may consult Chapter~\ref{chap:IterativeDiagnosisComputation}.
\end{enumerate}
\end{itemize}


\subsubsection{Output}
\label{sec:Output}
The output of Algorithm~\ref{algo:inter_onto_debug} can be explained as follows by making a distinction between the two modes of the algorithm specified by input parameter $mode$: 

\begin{proposition}\label{prop:correctness_of_interactive_KB_debugging_algo}
If $mode = static$, then Algorithm~\ref{algo:inter_onto_debug} returns the (exact) solution of the Interactive Static KB Debugging problem (Problem Definition~\ref{prob_def:static}) if $\sigma = 0$ and an approximate solution of the problem if $\sigma > 0$ where the likeliness of finding the (exact) solution increases with decreasing $\sigma$. 

More concretely, a maximal solution KB $\ot = (\mo \setminus \md_{\max}) \cup U_\Tp$ w.r.t.\ the input DPI $\langle\mo,\mb,\Tp,\Tn\rangle_\RQ$ is returned such that 
\begin{enumerate}
\item \label{prop:correctness_of_interactive_KB_debugging_algo:stat1} $\md_{\max} \in \mD$ \hfill ($\md_{\max}$ is an element of the current set of leading diagnoses) 
\item \label{prop:correctness_of_interactive_KB_debugging_algo:stat2} $\md_{\max} = \argmax_{\md\in\mD} p_{\mD}(\md)$ \hfill ($\md_{\max}$ is the a-posteriori most probable leading diagnosis)
\item \label{prop:correctness_of_interactive_KB_debugging_algo:stat3} $p_{\mD}(\md_{\max}) \geq 1 - \sigma$ \hfill (the a-posteriori probability of $\mD_{\max}$ exceeds the predefined threshold)
\item \label{prop:correctness_of_interactive_KB_debugging_algo:stat4} $\mD\subseteq\minD_{\langle\mo,\mb,\Tp,\Tn\rangle_\RQ}\cap\minD_{\langle\mo,\mb,\Tp\cup\Tp',\Tn\cup\Tn'\rangle_\RQ}$ comprises the $|\mD|$ most probable minimal diagnoses w.r.t.\ $\langle\mo,\mb,\Tp,\Tn\rangle_\RQ$ as per the diagnosis probability measure $p_{\mD,prio}()$ \newline 
(the set of leading diagnoses corresponds to the a-priori most probable minimal diagnoses w.r.t.\ the input DPI that satisfy all specified test cases),
\item \label{prop:correctness_of_interactive_KB_debugging_algo:stat5}  a-priori probability measure $p_{\mD,prio}()$ is computed from $p_{\widetilde{\mo}\cup\overline{\mo}}()$ as per 
\begin{enumerate}
	\item Formula~\ref{eq:ax_prob_calc} \hfill (computation of formula fault probabilities)
	\item Formula~\ref{eq:adapt_ax_prob_to_get_min_diags} \hfill (adaptation of formula fault probabilities)
	\item Formula~\ref{eq:diag_prob_calc} \hfill (computation of diagnoses probabilities from formula fault probabilities)
\end{enumerate}
\item \label{prop:correctness_of_interactive_KB_debugging_algo:stat6} the a-posteriori probability measure $p_{\mD}()$ is computed from $p_{\mD,prio}()$ as per Bayes' Theorem (Formula~\ref{eq:bayes}, for details see below) taking into account the new information given by the set of all answered queries so far, i.e.\ the collected sets of positive ($\Tp'$) and negative ($\Tn'$) test cases.
\end{enumerate}
If $mode = dynamic$, then Algorithm~\ref{algo:inter_onto_debug} returns the (exact) solution of the Interactive Dynamic KB Debugging problem (Problem Definition~\ref{prob_def:dynamic}) if $\sigma = 0$ and an approximate solution of the problem if $\sigma > 0$ where the likeliness of finding the (exact) solution increases with decreasing $\sigma$. 

More concretely, a maximal solution KB $\ot = (\mo \setminus \md_{\max}) \cup U_{\Tp\cup\Tp'}$ w.r.t.\ the current DPI $\langle\mo,\mb,\Tp\cup\Tp',\Tn\cup\Tn'\rangle_\RQ$ is returned such that 
\begin{enumerate}
\item \label{prop:correctness_of_interactive_KB_debugging_algo:dyn1} $\md_{\max} \in \mD$ \hfill ($\md_{\max}$ is an element of the current set of leading diagnoses) 
\item \label{prop:correctness_of_interactive_KB_debugging_algo:dyn2} $\md_{\max} = \argmax_{\md\in\mD} p_{\mD}(\md)$ \hfill ($\md_{\max}$ is the a-posteriori most probable leading diagnosis)
\item \label{prop:correctness_of_interactive_KB_debugging_algo:dyn3} $p_{\mD}(\md_{\max}) \geq 1 - \sigma$ \hfill (the a-posteriori probability of $\mD_{\max}$ exceeds the predefined threshold)
\item \label{prop:correctness_of_interactive_KB_debugging_algo:dyn4} $\mD\subseteq\minD_{\langle\mo,\mb,\Tp\cup\Tp',\Tn\cup\Tn'\rangle_\RQ}$ comprises the $|\mD|$ most probable minimal diagnoses w.r.t.\ $\langle\mo,\mb,\Tp\cup\Tp',\Tn\cup\Tn'\rangle_\RQ$ as per the diagnosis probability measure $p_{\mD,prio}()$ \newline 
(the set of leading diagnoses corresponds to the a-priori most probable minimal diagnoses w.r.t.\ the current DPI),
\item \label{prop:correctness_of_interactive_KB_debugging_algo:dyn5} the a-priori probability measure $p_{\mD,prio}()$ is computed from $p_{\widetilde{\mo}\cup\overline{\mo}}()$ as per 
\begin{enumerate}
	\item Formula~\ref{eq:ax_prob_calc} \hfill (computation of formula fault probabilities)
	\item Formula~\ref{eq:adapt_ax_prob_to_get_min_diags} \hfill (adaptation of formula fault probabilities)
	\item Formula~\ref{eq:diag_prob_calc} \hfill (computation of diagnoses probabilities from formula fault probabilities)
\end{enumerate}
\item \label{prop:correctness_of_interactive_KB_debugging_algo:dyn6} the a-posteriori probability measure $p_{\mD}()$ is computed from $p_{\mD,prio}()$ as per Bayes' Theorem (Formula~\ref{eq:bayes}, for details see below) taking into account the new information given by the set of all answered queries so far, i.e.\ the collected sets of positive ($\Tp'$) and negative ($\Tn'$) test cases.
\end{enumerate}
\end{proposition}
\begin{remark}\label{rem:approximate_solution}
We still need to explain what we mean by ``approximate solution'' of the Interactive Static (Dynamic) KB Debugging problem. 
Roughly, an approximate solution is one constructed from a diagnosis which is not the only remaining minimal diagnosis.
More precisely, an \emph{approximate solution} of
\begin{itemize}
	\item the Interactive Static KB Debugging problem is a maximal solution KB $(\mo \setminus \md)\cup U_{\Tp}$ such that 
\begin{itemize}
	\item $\md$ is a minimal diagnosis w.r.t.\ the input DPI and w.r.t.\ the current DPI and 
	\item there is some $\md' \neq \md$ which is a minimal diagnosis w.r.t.\ the input DPI and w.r.t.\ the current DPI  
\end{itemize}
	\item the Interactive Dynamic KB Debugging problem is a maximal solution KB $(\mo \setminus \md)\cup U_{\Tp\cup\Tp'}$ such that
\begin{itemize}
	\item $\md$ is a minimal diagnosis w.r.t.\ the current DPI and 
	\item there is some $\md' \neq \md$ which is a minimal diagnosis w.r.t.\ the current DPI  
\end{itemize}
\end{itemize}
where the input DPI is given by $\langle\mo,\mb,\Tp,\Tn\rangle_\RQ$ and the currect DPI by $\langle\mo,\mb,\Tp\cup\Tp',\Tn\cup\Tn'\rangle_\RQ$.

So, as long as not all but one diagnosis candidate that enables the formulation of a solution KB has been ruled out by the classification of test cases, we speak of an approximate solution. Now, the lower a value for $\sigma$ is predefined, the longer Algorithm~\ref{algo:inter_onto_debug} will usually need to iterate and the more test cases will usually need to be specified until one diagnosis has a probability greater than or equal to $1 - \sigma$. Thence, at the time a diagnosis exceeds the probability $1-\sigma$ there will be usually fewer minimal diagnoses left than in case of the selection a higher value for $\sigma$. Therefore, the likeliness of picking the (exact) solution will usually be the higher, the lower $\sigma$ is.\qed 
\end{remark}

\begin{remark}\label{rem:absolute_vs_relative_fault_tolerance_AND_initial_vs_updated_probabilities}
Note that granting a maximum \emph{absolute} fault tolerance $\sigma$ that is independent of a set of leading diagnoses is generally computationally infeasible due to the high complexity of diagnosis computation (see Chapter~\ref{chap:intro}). Since, for an absolute fault tolerance to hold, \emph{all} minimal diagnoses w.r.t.\ the current DPI have to be computed in order to determine their probability and to decide whether the most probable diagnosis has a probability greater than or equal to $1-\sigma$. 

In fact, the fault tolerance used by Algorithm~\ref{algo:inter_onto_debug} which is \emph{relative} to the set of leading diagnoses, i.e.\ the (a-priori) most probable minimal diagnoses $\mD$ 
w.r.t.\ a DPI can be interpreted as follows. Under the assumption that the true diagnosis $\dt$ is included in $\mD$, the chance that the most probable minimal diagnosis $\md_{\max} \in \mD$ which satisfies the stop criterion is not equal to $\dt$ is smaller than the predefined threshold $\sigma$ (cf.\ Section~\ref{sec:DiagnosisProbabilitySpace}). Thus, under this assumption, the (a-posteriori) probability of being presented a non-desired solution KB as output of Algorithm~\ref{algo:inter_onto_debug} is smaller than $\sigma$. 

The a-priori diagnoses probability measure $p_{\mD,prio}()$ refers to the one that is computed \emph{directly} from the fault information provided as an input to Algorithm~\ref{algo:inter_onto_debug} whereas the a-posteriori diagnoses probability measure $p_{\mD}()$ is the one obtained from $p_{\mD,prio}()$ after incorporating the information given by the new test cases specified so far during the debugging session.
So, $p_{\mD,prio}()$ and $p_{\mD}()$ might differ in terms of the probability order of diagnoses. Incorporation of updated probabilities \emph{directly} into the hitting set tree algorithms to be used for the determination of leading diagnoses in the order prescribed by an updated probability measure is only possible if there is an additional update operator (besides Bayes' Theorem for adapting \emph{diagnoses} probabilities) that can be applied to \emph{formula} probabilities. For, the latter are exploited in the hitting set tree to assign probability weights to paths that are \emph{not yet diagnoses} (cf.\ $p_{nodes}()$ specified by Definition~\ref{def:p_node()} and the discussion of Formula~\ref{eq:path_prob_calc}) in order to guide the search for minimal diagnoses in best-first order. Updated \emph{diagnosis} probabilities are not helpful at all for this purpose. Devising a reasonable mechanism of updating formula probabilities seems to be hard mostly due to the lack of suitable data that might be collected during the debugging session to accomplish that. What would be imaginable during the debugging session is to try to learn something about the fault probability of \emph{syntactical elements} by examining the positive (all formulas are \emph{definitely} correct) and singleton negative (the single formula is \emph{definitely} incorrect) test cases. However, a drawback of such a strategy comes into effect when only syntactically very simple queries are used which is, for instance, the case in Example~\ref{example:query_computation} (see the definition of the \textsc{getEntailments} function there). From such queries not many useful insights concerning faulty syntactical elements might be gained. On the other hand, such queries are absolutely desirable from the point of view of how well a user might comprehend the formulas asked by the system. Hence, these two aspects seem to contradict each other. Still, it is a topic for future research to attempt to elaborate a solution for that issue.

A way to achieve that $p_{\mD}()$ coincides with $p_{\mD,prio}()$, at least in case $mode=static$, is to exclude queries $Q$ with $\dz{}(Q) \neq \emptyset$ (see Remark~\ref{rem:prob_update_obsolete_if_dz(Q)={}_for_all_queries}).
How this might be accomplished is stated by Proposition~\ref{prop:query_gen_explicit_entailments_dz_empty}. 
Please notice that ignorance of queries with non-empty $\dz{}$ does not implicate any disadvantages for interactive debugging. On the contrary, it is even a desirable feature of a debugger and brings along higher computational efficacy of query generation and stronger test cases from the logical point of view (cf.\ Section~\ref{sec:remarks_query_gen}). For the scenario $mode=dynamic$, it is not possible in general to bypass the probability update by means of such queries (see Remark~\ref{rem:prob_update_obsolete_if_dz(Q)={}_for_all_queries}).  
%
\qed
\end{remark} 

\subsubsection{Variables}
\label{sec:Variables}
The variables used by Algorithm~\ref{algo:inter_onto_debug} that are not input arguments to the algorithm are the following:
\begin{itemize}
\item $\Tp', \Tn'$ are the sets of positive and negative test cases, respectively, collected \emph{during} the execution of Algorithm~\ref{algo:inter_onto_debug} so far. That is, $\Tp'$ stores all positively answered queries, whereas $\Tn'$ stores all negatively answered ones.
\item $\mC_{calc}$ is the 
set of all conflict sets computed by $\scQX$ during the execution of Algorithm~\ref{algo:inter_onto_debug} so far.

\emph{Remark:} In case of static debugging ($mode=static$), $\mC_{calc}$ includes exclusively minimal conflict sets w.r.t.\ the input DPI, whereas, in case of dynamic debugging ($mode=dynamic$), $\mC_{calc}$ may comprise minimal conflict sets w.r.t.\ the current or any intermediate DPI. 
\item $\mD_{\checkmark}$ is the set of leading diagnoses returned by a call of \textsc{staticHS} in case of static debugging ($mode=static$) and by a call of \textsc{dynamicHS} in case of dynamic debugging ($mode=dynamic$). 

\emph{Remarks:} In case of dynamic debugging, $\mD_{\checkmark} \subseteq \minD_{\langle\mo,\mb,\Tp\cup\Tp',\Tn\cup\Tn'\rangle_\RQ}$ is the set of most probable minimal diagnoses w.r.t.\ the current DPI $\langle\mo,\mb,\Tp\cup\Tp',\Tn\cup\Tn'\rangle_\RQ$ as per the diagnosis probability measure $p_{\mD,prio}()$ computed from $p_{\widetilde{\mo}\cup\overline{\mo}}()$ by Formulas~\ref{eq:ax_prob_calc}, \ref{eq:adapt_ax_prob_to_get_min_diags}, \ref{eq:diag_prob_calc} and \ref{eq:diag_prob_norm} (cf.\ Sections~\ref{sec:DiagnosisProbabilitySpace} and \ref{sec:Output}).

In case of static debugging, $\mD_{\checkmark} \subseteq \minD_{\langle\mo,\mb,\Tp,\Tn\rangle_\RQ} \cap \minD_{\langle\mo,\mb,\Tp\cup\Tp',\Tn\cup\Tn'\rangle_\RQ}$, i.e.\ $\mD_{\checkmark}$ includes only diagnoses that are minimal diagnoses w.r.t.\ the input DPI $\langle\mo,\mb,\Tp,\Tn\rangle_\RQ$ as well as w.r.t.\ the current DPI $\langle\mo,\mb,\Tp\cup\Tp',\Tn\cup\Tn'\rangle_\RQ$. Moreover, $\mD_{\checkmark}$ comprises the most probable minimal diagnoses \emph{w.r.t.\ the input DPI} according to the diagnosis probability measure $p_{\mD,prio}()$ computed from $p_{\widetilde{\mo}\cup\overline{\mo}}()$ by Formulas~\ref{eq:ax_prob_calc}, \ref{eq:adapt_ax_prob_to_get_min_diags}, \ref{eq:diag_prob_calc} and \ref{eq:diag_prob_norm} (cf.\ Sections~\ref{sec:DiagnosisProbabilitySpace} and \ref{sec:Output}).
%
\item $\mD_{\times}$ stores all minimal diagnoses w.r.t.\ the input DPI that have been invalidated by one of the collected positive and negative test cases $\Tp'$ and $\Tn'$, respectively ($mode=static$). 
$\mD_{\times}$ stores the minimal diagnoses w.r.t.\ the last-but-one DPI that have been invalidated by the most recently added test case ($mode=dynamic$). 
%
\item $\mD_{out}$ is the subset of the set of current leading diagnoses $\md_{\checkmark}$ that has been invalidated by the most recently added test case.
\item $\mD_{\supset}$ stores all diagnoses that are non-minimal w.r.t.\ the current DPI, i.e.\ for each diagnosis $\mathsf{nd} \in \mD_{\supset}$ there is some $\mathsf{nd}'\in\mD_{\checkmark}$ such that $\mathsf{nd} \supset \mathsf{nd}'$ ($mode=dynamic$). 

\emph{Remark:} $\mD_{\supset}$ is solely needed for dynamic and not for static debugging as the latter does not need to store non-minimal diagnoses (cf.\ rule~\ref{def:pruned_hs_tree:non_min_pruning_rule} of Definition~\ref{def:pruned_hs_tree} on page~\pageref{def:pruned_hs_tree}). Reason for this is the fact that only minimal diagnoses w.r.t.\ the input DPI are searched for. 
%
On the other hand, in case of dynamic debugging, non-minimal diagnoses might become minimal ones after some new test cases are specified since minimal diagnoses w.r.t.\ the (changing) current DPI are considered.
\item\label{etc:var:metrics} $qData$ is an informal variable that comprehends any kind of data that might be taken into account by the query selection measure $qsm()$ and that might need to be adapted after a query has been answered (and diagnoses have been invalidated) in order to take the obtained new information into account. One can imagine $qData$ as a log specific to the particular function $qsm()$ that is used which records data of prior (query answering) iterations executed by the algorithm such as certain performance measures. An example of a $qsm()$ strategy using one such metric, namely the ratio of leading diagnoses invalidated by a test case, can be found in~\cite{Rodler2013}. 
\item \label{etc:QA} $QA := [\tuple{Q,u(Q)}]_{Q\in\Tp'\cup\Tn'}$ where $u(Q) \in \setof{\true,\false}$ is the chronologically ordered list of queries and user answers collected so far during the execution of Algorithm~\ref{algo:inter_onto_debug}.
\item $\Queue$ is the current queue of open nodes in the 
hitting set tree maintained by Algorithm~\ref{algo:inter_onto_debug}.
\item The list $\Queue_{dup}$ roughly stores all duplicate nodes (that is, nodes for each of which there is a node in the hitting set tree that corresponds to an equal set of edge labels) computed so far during the execution of Algorithm~\ref{algo:inter_onto_debug}.

\emph{Remark:} The list $\Queue_{dup}$ is only relevant in case $mode = dynamic$ and not needed if $mode=static$. The purpose of this set is to enable the ``replacement'' of pruned nodes which is necessary to guarantee the completeness of \textsc{dynamicHS} in terms of not missing any minimal diagnoses (for a detailed explanation, see Section~\ref{sec:DynamicHSTree}).
\newline
\end{itemize}

\subsubsection{Algorithm Walkthrough}
\label{sec:Walkthrough}
\paragraph{Initialization.} In the first \ref{algoline:inter_onto_debug:var_inst_end} lines,
variable declarations take place. First, all variables that store sets of conflict sets, diagnoses or test cases, and $qData$ are initialized to the empty set. Further on, $\Queue_{dup}$ and $QA$ are initialized to an empty list. Finally, the queue $\Queue$ of open nodes used for the hitting set tree construction by \textsc{staticHS} ($mode=static$) or \textsc{dynamicHS} ($mode=dynamic$), respectively, is set to $[\emptyset]$ since it initially includes only a non-labeled root node.
\begin{remark}\label{rem:queue_root_node_emptyset_emptylist}
The non-labeled root node is denoted by $\emptyset$ since nodes in \textsc{staticHS} are associated with the set of edge labels along the path in the hitting set tree from the root node to this node (cf.\ Chapter~\ref{chap:DiagnosisComputation} and Section~\ref{sec:StaticHSTree}). Hence, the root node itself corresponds to the empty path which includes no edges.

Notice that in case of \textsc{dynamicHS}, nodes will be (ordered) lists instead of (non-ordered) sets like in \textsc{staticHS} (cf.\ Section~\ref{sec:DynamicHSTree}). That is, to be precise, the unlabeled root node in this case corresponds to the empty list $[]$. For the ease of representation of Algorithm~\ref{algo:inter_onto_debug}, only one set $\Queue$ is initialized to be used with either \textsc{staticHS} or \textsc{dynamicHS}. Thence, by abuse of notation, we associate $\emptyset$ in this case with the empty list $[]$.\qed
\end{remark}

\paragraph{Computing Formula Probabilities.} Then, \textsc{getFormulaProbs} is called in line~\ref{algoline:inter_onto_debug:getAxiomProbs} with the KB $\mo$ and the function $p_{\widetilde{\mo}\cup\overline{\mo}}: \widetilde{\mo}\cup\overline{\mo} \rightarrow (0,1]$ as inputs. The function first applies Formula~\ref{eq:ax_prob_calc} to compute probabilities for each formula in $\mo$, then applies Formula~\ref{eq:adapt_ax_prob_to_get_min_diags} to these probabilities leading to the output $p_{\mo}: \mo \rightarrow (0,0.5)$, a function that assigns a value $p_{\mo}(\tax) \in (0,0.5)$ to each $\tax\in\mo$.

\paragraph{Computing Leading Diagnoses.} At this point, all input arguments required by for the hitting set tree construction are instantiated. So, the algorithm enters the while loop in line~\ref{algoline:inter_onto_debug:while}. As a first step within the loop, either \textsc{staticHS}, if $mode = static$, or \textsc{dynamicHS}, otherwise, is called in order to obtain a tuple including a set of leading diagnoses along with variables that store the ``state'' of the (partial) hitting set tree constructed so far and facilitate the reuse of this tree in the next iteration.

In concrete terms, \textsc{staticHS} accepts the arguments $\langle\mo,\mb,\Tp,\Tn\rangle_\RQ$, $\Queue$, $t$, $n_{\min}$, $n_{\max}$, $\mathbf{C}_{calc}$, $\mD_{\checkmark}$, $\mD_{\times}$, $p_{\mo}()$, $\Tp'$ and $\Tn'$ and returns a tuple $\tuple{\mD,\Queue, \mathbf{C}_{calc}, \mD_{\times}}$ the elements of which are defined as follows:
\begin{itemize}
\item $\mD$ is the current set of leading diagnoses such that
\begin{enumerate}[(a)]
\item \label{etc:staticHS_output_bullet_a} $\mD \subseteq \minD_{\langle\mo,\mb,\Tp,\Tn\rangle_\RQ} \cap \minD_{\langle\mo,\mb,\Tp\cup\Tp',\Tn\cup\Tn'\rangle_\RQ}$ is the set of most probable 
 minimal diagnoses w.r.t.\ $\langle\mo,\mb,\Tp,\Tn\rangle_\RQ$ that satisfy all test cases $\Tp'$ and $\Tn'$ such that 
\begin{enumerate}[(i)]
\item $n_{\min} \leq |\mD| \leq n_{\max}$ and 
\item \label{etc:staticHS_must_compute_at_least_one_further_min_diag_per_call} $\mD\supset\mD_{\checkmark}$,
\end{enumerate} 
if such a set $\mD$ 
 exists;
or 
\item $\mD$ is equal to the set of all minimal diagnoses $\minD_{\langle\mo,\mb,\Tp,\Tn\rangle_\RQ} \cap \minD_{\langle\mo,\mb,\Tp\cup\Tp',\Tn\cup\Tn'\rangle_\RQ}$, otherwise;
\end{enumerate}
where ``most-probable'' refers to the diagnosis probability measure $p_{\mD,prio}()$ obtained from $p_{\mo}()$ 
by application of Formulas~\ref{eq:diag_prob_calc} and \ref{eq:diag_prob_norm}.
\item $\Queue$ is the current queue of open nodes of the hitting set tree.
\item $\mathbf{C}_{calc} \subseteq \minC_{\langle\mo,\mb,\Tp,\Tn\rangle_\RQ}$ is the set of all computed minimal conflict sets w.r.t.\ the input DPI throughout all calls of \textsc{staticHS} during the execution of Algorithm~\ref{algo:inter_onto_debug} so far.
\item $\mD_{\times}$ comprises all computed minimal diagnoses throughout all calls of \textsc{staticHS} during the execution of Algorithm~\ref{algo:inter_onto_debug} so far where each $\md\in\mD_{\times}$ has been invalidated by some test case in $\Tp'$ or $\Tn'$.
\end{itemize} 

Similarly, \textsc{dynamicHS} accepts the arguments $\langle\mo,\mb,\Tp,\Tn\rangle_\RQ$, $\Queue$, $\Queue_{dup}$, $t$, $n_{\min}$, $n_{\max}$, $\mathbf{C}_{calc}$, $\mD_{\checkmark}$, $\mD_{\times}$, $p_{\mo}()$, $\Tp'$, $\Tn'$ and $\mD_{\supset}$ and returns a tuple $\tuple{\mD,\Queue, \mathbf{C}_{calc}, \mD_{\times}, \mD_{\supset}, \Queue_{dup}}$ the elements of which are defined as follows:
\begin{itemize}
\item $\mD$ is the current set of leading diagnoses such that
\begin{enumerate}[(a)]
\item \label{etc:dynamicHS_output_bullet_a} $\mD \subseteq \minD_{\langle\mo,\mb,\Tp\cup\Tp',\Tn\cup\Tn'\rangle_\RQ}$ is the set of most probable minimal diagnoses w.r.t.\ $\langle\mo,\mb,\Tp\cup\Tp',\Tn\cup\Tn'\rangle_\RQ$ such that 
\begin{enumerate}[(i)]
\item $n_{\min} \leq |\mD| \leq n_{\max}$ and 
\item \label{etc:dynamicHS_must_compute_at_least_one_further_min_diag_per_call} $\mD \setminus\mD_{\checkmark} \neq \emptyset$,
\end{enumerate}
if such a set $\mD$ exists, or 
\item $\mD$ is equal to the set of all minimal diagnoses $\minD_{\langle\mo,\mb,\Tp\cup\Tp',\Tn\cup\Tn'\rangle_\RQ}$, otherwise,
\end{enumerate}
where ``most-probable'' refers to the diagnosis probability measure $p_{\mD,prio}()$ obtained from $p_{\mo}()$ by application of Formulas~\ref{eq:diag_prob_calc} and \ref{eq:diag_prob_norm}.
\item $\Queue$ is the current queue of open (non-labeled) nodes of the hitting set tree,
\item $\mathbf{C}_{calc}$ is a set of conflict sets w.r.t.\ the current DPI $\langle\mo,\mb,\Tp\cup\Tp',\Tn\cup\Tn'\rangle_\RQ$,
\item $\mD_{\times} = \emptyset$, 
\item $\mD_{\supset}$ is the set of all processed nodes so far throughout the execution of Algorithm~\ref{algo:inter_onto_debug} that are non-minimal diagnoses w.r.t.\ the current DPI $\langle\mo,\mb,\Tp\cup\Tp',\Tn\cup\Tn'\rangle_\RQ$ and
\item $\Queue_{dup}$ includes all duplicate nodes found so far throughout the execution of Algorithm~\ref{algo:inter_onto_debug} (for a detailed explanation see Section~\ref{sec:DynamicHSTree} and Algorithm~\ref{algo:inter_dyn_hs}).
\end{itemize}

\begin{remark}\label{rem:p_D,prio()_same_order_for_diags_as_p_nodes()}
It is very important to notice that the function $p_{nodes}()$ for $p() := p_{\mo}()$ as specified by Definition~\ref{def:p_node()} on page~\pageref{def:p_node()} imposes the same order on a set of minimal diagnoses as the a-priori probability measure $p_{\mD,prio}()$. That is $p_{nodes}(\md) = c \cdot p_{\mD,prio}(\md)$ for all minimal diagnoses $\md$ w.r.t.\ a DPI where $c$ is a constant (which is the same for all diagnoses $\md$). The difference between both functions is that $p_{nodes}()$ is defined for all $X \subseteq \mo$ whereas $p_{\mD,prio}()$ is only defined for (leading) minimal diagnoses $\md \subseteq \mo$. Further on $p_{\mD,prio}()$ is normalized whereas $p_{nodes}()$ is not which accounts for the (normalization) constant $c$. The function $p_{nodes}()$ is essential for the best-first construction of the hitting set tree in \textsc{staticHS} and \textsc{dynamicHS} since it allows for the assignment of a ``probability'' to non-diagnoses (cf.\ the discussion of Formula~\ref{eq:path_prob_calc} on page~\pageref{eq:path_prob_calc}). 
Since the input argument $p()$ (which is the same for all calls) to \textsc{staticHS} as well as \textsc{dynamicHS} is equal to $p_{\mo}()$ by lines~\ref{algoline:inter_onto_debug:staticHS} and \ref{algoline:inter_onto_debug:dynamicHS} in Algorithm~\ref{algo:inter_onto_debug}, the set $\mD$ returned by \textsc{staticHS} (\textsc{dynamicHS}) is also the set of most probable minimal diagnoses w.r.t.\ $\langle\mo,\mb,\Tp,\Tn\rangle_\RQ$ ($\langle\mo,\mb,\Tp\cup\Tp',\Tn\cup\Tn'\rangle_\RQ$) \emph{as per the function $p_{nodes}()$} (cf.\ Propositions~\ref{prop:static_hs_correctness} and \ref{prop:dynamic_hs_correctness}).\qed
\end{remark}

\begin{remark}\label{rem:return_params_of_stat+dynHS}
Notice that the return parameter that is relevant for the main purpose of Algorithm~\ref{algo:inter_onto_debug}, namely to compute a query and thereby obtain a new test case classified by the user, is solely the set of leading diagnoses $\mD$. The other return parameters serve as a means to store the state of the hitting set tree that is gradually built up by successive calls of \textsc{staticHS} (if $mode=static$) and \textsc{dynamicHS} (if $mode=dynamic$), respectively. Whereas $\Queue$ and $\mC_{calc}$ (and $\mD_{\supset}$ and $\Queue_{dup}$ in case of \textsc{dynamicHS}) are never modified until the next call to \textsc{staticHS} or \textsc{dynamicHS}, the sets $\mD_{\checkmark}$ and $\mD_{\times}$ are only changed once, after the subset of invalidated leading diagnoses $\mD_{out}$ is known, in lines~\ref{algoline:inter_onto_debug:update_D_checkmark} and \ref{algoline:inter_onto_debug:update_D_times}.\qed
\end{remark}

At this moment, we do not go into detail regarding the way how leading diagnoses are computed by \textsc{staticHS} and \textsc{dynamicHS}. We simply suppose that both functions act in a manner that the outputs just specified are returned for the given inputs. An in-depth delineation of both functions will be given in Sections~\ref{sec:StaticHSTree} and \ref{sec:DynamicHSTree} in Chapter~\ref{chap:IterativeDiagnosisComputation}. Further note that the return parameter $\mD$ is stored in variable $\mD_{\checkmark}$ from line~\ref{algoline:inter_onto_debug:dynamicHS} on. 

\paragraph{Computing a Probability Distribution of Leading Diagnoses.}\label{etc:computing_prob_dist_of_leading_diags} After the set of leading diagnoses $\mD_{\checkmark}$ has been computed, the variables $\mD_{\checkmark}$, $p_{\mo}()$, $\langle\mo,\mb,\Tp,\Tn\rangle_\RQ$ and $QA$ are used as arguments to the function \textsc{getProbDist} (see Algorithm~\ref{algo:inter_onto_debug_continued}) which computes a probability distribution of the leading diagnoses, i.e.\ a probability measure $p_{\mD}()$ for the probability space with sample space $\Omega = \mD_{\checkmark}$ (cf.\ Section~\ref{sec:DiagnosisProbabilitySpace}).
As a first action to achieve this, the (a-priori) probabilities $p_{\mD,prio}(\md)$ for $\md\in\mD_{\checkmark}$ are computed from the (a-priori) probabilities $p_{\mo}(\tax)$ for formulas $\tax\in\mo$ as per Formula~\ref{eq:diag_prob_calc} (\textsc{getPrioDiagProbs} in line~\ref{algoline:get_prob_dist:getPrioDiagProbs}). Application of Formula~\ref{eq:diag_prob_norm} is not necessary at this point as probabilities are anyhow normalized at the end of \textsc{getProbDist} (line~\ref{algoline:get_prob_dist:normalize}). Notice that the function $p_{\mo}()$ remains constant, i.e.\ unmodified, throughout the entire execution of Algorithm~\ref{algo:inter_onto_debug}.  

Now, since a-priori diagnosis probabilities assigned by $p_{\mD,prio}()$ directly rely upon $p_{\mo}()$ which in turn is computed directly from the initially given fault probabilities $p_{\widetilde{\mo}\cup\overline{\mo}}()$, the probability measure $p_{\mD,prio}()$ is adapted to yield \emph{a-posteriori diagnosis probabilities} $p_{\mD}()$ in order to reflect the new evidence provided by the collected test cases $\Tp'$ and $\Tn'$.

The a-posteriori probability of a current leading diagnosis $\md$ in $\mD_{\checkmark}$ is $p_{\mD}(\md\,|\,QA)$ and can be computed by means of Bayes' Theorem (Formula~\ref{eq:bayes}) from $p_{\mD,prio}()$ as follows.
\begin{align*} 
p_{\mD}(\md\,|\,QA) = \frac{p_{\mD,prio}(QA\,|\,\md)\;\,p_{\mD,prio}(\md)}{p_{\mD,prio}(QA)} 
\end{align*}
where $QA$ is the chronologically ordered list of queries and user answers collected so far during the execution of Algorithm~\ref{algo:inter_onto_debug} (see page~\pageref{etc:QA}).
We point out that $p_{\mD,prio}(QA)$ is only a normalization factor that is equal for each diagnosis and thus does not need to be explicitly computed. The crucial factor is 
\begin{align*}
p_{\mD,prio}(QA\,|\,\md) = p_{\mD,prio}(\forall \tuple{Q,u(Q)} \in QA: Q = u(Q) \,|\,\md)
\end{align*}
which describes the probability of getting exactly the answer $u(Q)$ for each query $Q\in \Tp'\cup\Tn'$ under the assumption that $\md$ corresponds to the true diagnosis $\dt$, i.e.\ $\dt = \md$. In other words, $p_{\mD,prio}(QA\,|\,\md)$ is the probability of $QA$ under the assumption that the user answers in a way that $u(Q) = \true$ if $\md \in \dx{}(Q)$ and $u(Q) = \false$ if $\md \in \dnx{}(Q)$. 

For a single query $Q_i$, the probability $p_{\mD,prio}(Q_i=u(Q_i)\,|\,\md)$ is defined as (cf. \cite{dekleer1987})
\begin{equation}\label{eq:cond_query_prob1}
p_{\mD,prio}(Q_i=u(Q_i)\,|\,\md) = 
\begin{cases}
1, 						& \mbox{if } \md \in \dx{}(Q_i) \\   
0, 						& \mbox{if } \md \in \dnx{}(Q_i) \\
\frac{1}{2}, 	& \mbox{if } \md \in \dz{}(Q_i) 
\end{cases} 
\end{equation} 
for $u(Q_i) = \true$ and 
\begin{equation}\label{eq:cond_query_prob2}
p_{\mD,prio}(Q_i=u(Q_i)\,|\,\md) = 
\begin{cases}
1, 						& \mbox{if } \md \in \dnx{}(Q_i) \\   
0, 						& \mbox{if } \md \in \dx{}(Q_i) \\
\frac{1}{2}, 	& \mbox{if } \md \in \dz{}(Q_i) 
\end{cases} 
\end{equation} 
for $u(Q_i) = \false$ where $\dx{}(Q_i)$, $\dnx{}(Q_i)$ and $\dz{}(Q_i)$ are computed w.r.t.\ the DPI $\langle\mo$, $\mb,\Tp\cup\Tp'',\Tn\cup\Tn''\rangle$ where $\Tp''$ and $\Tn''$, respectively, include all test cases collected \emph{prior} to $Q_i$, i.e.\ $\Tp''\cup\Tn'' = \setof{Q_1,\dots,Q_{i-1}}$ if queries are numbered chronologically. That is, if $\md$ predicted the answer $u(Q_i)$ to $Q_i$ given by the user, the probability is 1, zero if $\md$ predicted the converse answer $\lnot u(Q_i)$ and $\frac{1}{2}$ if $\md$ did not predict any answer to $Q_i$.

So, aside from the normalization factor (see above), $p_{\mD,prio}(Q_i=u(Q_i)\,|\,\md)$ is the factor by which the a-priori probability $p_{\mD,prio}(\md)$ must be multiplied to obtain the a-posteriori probability $p_{\mD}(\md)$ of a diagnosis $\md$ after a single query $Q_i$ has been answered and added as a test case to the DPI.

The intuitive explanation for the update by this factor
is that if $\md$ predicted (at least) one answer $u(Q)$ conversely as given by the user, then $\md$ is a-posteriori impossible since it has already been invalidated by the addition of test case $Q$.
In case a diagnosis has never predicted the wrong answer, but did not predict any answer for many queries so far, then it is a-posteriori more unlikely than a diagnosis that did predict a correct answer more often. That is, our a-posteriori degree of belief that $\md$ is the correct diagnosis is the higher, the more often $\md$ had predicted answers to queries that were later actually given by the user 
(cf.\ Section~\ref{sec:InterpretationOfQPartitions} for an explanation what we mean by ``predict''). 

The value of $p_{\mD,prio}(Q_i=u(Q_i)\,|\,\md)$ can be computed by use of $QA$ and the q-partitions $\Pt(Q_1)$, $\dots$, $\Pt(Q_{i-1})$ of the \emph{current} set of leading diagnoses $\mD_{\checkmark}$ (for which a-posteriori probabilities are to be computed) for all queries $Q_1,\dots,Q_{i-1}$ answered before query $Q_i$. Thereby, each $\Pt(Q_j)$ where $j\in\setof{1,\dots,i-1}$ must be computed for a DPI where only $Q_1,\dots,Q_{j-1}$ are incorporated as test cases.

Taking these thoughts into account, \textsc{getProbDist} (Algorithm~\ref{algo:inter_onto_debug_continued}) updates $p_{\mD,prio}(\md)$ for each diagnosis $\md\in\mD_{\checkmark}$ in that it runs through all query-answer pairs $\tuple{Q,u(Q)}$ in $QA$ chronologically and for each $\md\in\mD_{\checkmark}$ it multiplies $p_{\mD,prio}(\md)$ by $\frac{1}{2}$ if $\md \in \dz{}(Q)$ as per Formulas~\ref{eq:cond_query_prob1} and \ref{eq:cond_query_prob2}. For each check whether a diagnosis is in $\dz{}(Q)$ in lines~\ref{algoline:get_prob_dist:check_dz_1} and \ref{algoline:get_prob_dist:check_dz_2} a DPI is used that already incorporates all test cases $\Tp''$ and $\Tn''$ that have been added chronologically before $Q$ was asked. This is achieved by updating $\Tp''$ and $\Tn''$ successively (lines~\ref{algoline:get_prob_dist:update_Tp''} and \ref{algoline:get_prob_dist:update_Tn''}). After all elements of $QA$ have been processed, the updated diagnosis probabilities are finally normalized (line~\ref{algoline:get_prob_dist:normalize}, cf. Formula~\ref{eq:diag_prob_norm} on page \pageref{eq:diag_prob_norm}) and the resulting function $p_{\mD,prio}()$ is returned. 

\begin{remark}\label{rem:remarks_ad_function_getProbDist}
Note that the function \textsc{getProbDist} exploits the fact that all diagnoses in $\md_{\checkmark}$ are leading diagnoses w.r.t.\ the \emph{current} DPI $\langle\mo,\mb,\Tp\cup\Tp',\Tn\cup\Tn'\rangle_\RQ$ which guarantees that none of these diagnoses has been invalidated by any of the test cases in $\Tp'$ or in $\Tn'$ added throughout the execution of Algorithm~\ref{algo:inter_onto_debug}. 
Hence, it is clear that each $\md \in \mD_{\checkmark}$ must be in $\dx{}(Q) \cup \dz{}(Q)$ if $u(Q) = \true$ and in $\dnx{}(Q) \cup \dz{}(Q)$ if $u(Q) = \false$, and it is only tested whether $\md \notin \dx{}(Q)$ in the prior case (line~\ref{algoline:get_prob_dist:check_dz_1}) and whether $\md \notin \dnx{}(Q)$ in the latter (line~\ref{algoline:get_prob_dist:check_dz_2}). 
It must be further noted that, in case of $mode=dynamic$, diagnoses in $\mD_{\checkmark}$ are not necessarily minimal diagnoses w.r.t.\ the intermediate DPIs $\langle\mo$, $\mb,\Tp\cup\Tp'',\Tn\cup\Tn''\rangle$ that are used for the probability update. However, this is not problematic since any set of (minimal and/or non-minimal) diagnoses is partitioned into the three sets $\dx{}(Q)$, $\dnx{}(Q)$ and $\dz{}(Q)$ by a query $Q$ (cf.\ Remark~\ref{rem:query_partitions_any_set_of_diagnoses_into_dx_dnx_dz}) wherefore $\Pt(Q)$ exists for any set $\mD_{\checkmark}$. Thence, the correctness of \textsc{getProbDist} remains unaffected by the usage of the setting $mode=dynamic$.\qed
\end{remark}
\begin{remark}\label{rem:prob_update_obsolete_if_dz(Q)={}_for_all_queries}
We want to emphasize that an adaptation of $p_{\mD,prio}(\md)$ is only necessary in case $\md \in \dz{}(Q_j)$ for some query $Q_j$ answered so far during the execution of Algorithm~\ref{algo:inter_onto_debug} as otherwise a multiplication by 1 is required which does not change $p_{\mD,prio}(\md)$.

For the case of static debugging ($mode=static$), an immediate implication of this is the following: The restriction of asking the user \emph{only} queries $Q_j$ w.r.t.\ a DPI with the property that \emph{no minimal diagnosis} w.r.t.\ this DPI can be an element of $\dz{}(Q_j)$ makes the probability update for each diagnosis in $\mD_{\checkmark}$ equivalent to a multiplication by 1 and hence obsolete. This must be the case since each diagnosis in $\mD_{\checkmark}$ which is a subset of $\minD_{\langle\mo,\mb,\Tp,\Tn\rangle_\RQ} \cap \minD_{\langle\mo,\mb,\Tp\cup\Tp',\Tn\cup\Tn'\rangle_\RQ}$ (see 
Section~\ref{sec:Output}) must be a \emph{minimal} diagnosis w.r.t.\ each intermediate DPI (which includes a superset of the test cases in the input DPI $\langle\mo,\mb,\Tp,\Tn\rangle_\RQ$ and a subset of the test cases in the current DPI $\langle\mo,\mb,\Tp\cup\Tp',\Tn\cup\Tn'\rangle_\RQ$).
Consequently, such a scenario implicates that the order of diagnoses computed by \textsc{staticHS} corresponds to the best-first order also w.r.t.\ the a-posteriori diagnosis probabilities (cf.\ Remark~\ref{rem:absolute_vs_relative_fault_tolerance_AND_initial_vs_updated_probabilities}).

The approach of only using queries with this property is feasible, e.g.\ by using a \textsc{getEntailments} function in conformity with Proposition~\ref{prop:query_gen_explicit_entailments_dz_empty} for the generation of the query pool (\textsc{getPoolOfQueries}). Such a type of queries is also favorable from the discrimination point of view, as we pointed out in Section~\ref{sec:remarks_query_gen}. An improvement of static debugging with this type of queries is to deactivate the probability update, i.e.\ replace line~\ref{algoline:inter_onto_debug:getProbDist} in Algorithm~\ref{algo:inter_onto_debug} by line~\ref{algoline:get_prob_dist:getPrioDiagProbs} of Algorithm~\ref{algo:inter_onto_debug_continued}. This improvement is not shown in Algorithm~\ref{algo:inter_onto_debug}.

In a dynamic debugging session ($mode = dynamic$), on the contrary, the usage of such queries does not guarantee the triviality of the probability update. For, also if no minimal diagnosis w.r.t.\ the DPI (for which a query $Q_j$ is computed) can be an element of $\dz{}(Q_j)$, there may be some non-minimal one which is. For example, for any admissible DPI $\langle\mo,\mb,\Tp,\Tn\rangle_\RQ$ is holds that $\md := \mo$ is a diagnosis (cf.\ Proposition~\ref{prop:exist_diag} and Definition~\ref{def:admissible}), albeit in most cases a non-minimal one. In such a case, $(\mo \setminus \md) \cup \mb \cup U_{\Tp}$ which is equal to $\mb \cup U_{\Tp}$ cannot entail $Q_j$. Because, were this the case, then all minimal diagnoses $\md_i \in \minD_{\langle\mo,\mb,\Tp,\Tn\rangle_\RQ}$ would be elements of $\dx{}(Q_j)$ as each $\mo^*_i \supseteq \mb \cup U_{\Tp}$ and thus each $\mo^*_i \models Q_j$ by the monotonicity of $\mathcal{L}$. Hence, this would be a contradiction to the fact that $Q_j$ is a query w.r.t.\ $\langle\mo,\mb,\Tp,\Tn\rangle_\RQ$ by Corollary~\ref{cor:q-partition_dx_dnx}. On the other hand, $(\mo \setminus \md) \cup \mb \cup U_{\Tp} \cup Q_j = \mb \cup U_{\Tp} \cup Q_j$ cannot violate any $x \in \Tn\cup\RQ$. Since, if this were the case, then adding $Q_j$ to the positive test cases would lead to a non-admissible DPI $\langle\mo,\mb,\Tp\cup\setof{Q_j},\Tn\rangle_\RQ$. By Corollary~\ref{cor:query_leaves_valid_diag}, this would be a contradiction to the fact that $Q_j$ is a query w.r.t.\ $\langle\mo,\mb,\Tp,\Tn\rangle_\RQ$. Thence, $\md \in \dz{}(Q_j)$ must hold for the assumed non-minimal diagnosis $\md$. From that we conclude that the probability update in dynamic debugging cannot be made obsolete in general by the usage of such a type of queries.
\qed 
\end{remark}

\paragraph{Stop Criterion and Output.} The (a-posteriori) probability distribution $p_{\mD}()$ of leading diagnoses $\mD_{\checkmark}$ is then used in line~\ref{algoline:inter_onto_debug:getMode} of Algorithm~\ref{algo:inter_onto_debug} to compute the mode of this distribution, i.e.\ the one diagnosis $\md_{\max}\in\mD_{\checkmark}$ with maximum probability according to $p_{\mD}()$. 

In the sequel, $\md_{\max}$ is used to check the stop criterion (line~\ref{algoline:inter_onto_debug:stop_crit}), namely whether $\md_{\max}$ has a probability greater than or equal to $1-\sigma$. If this is the case and $mode = static$, the function \textsc{getSolKB} computes a maximal solution KB w.r.t.\ the input DPI as $(\mo \setminus \md_{\max}) \cup U_{\Tp}$ by means of the current DPI $\langle\mo,\mb,\Tp\cup\Tp',\Tn\cup\Tn'\rangle_\RQ$, $\Tp'$ and $\md_{\max}$. Given that $mode = dynamic$, \textsc{getSolKB} returns a maximal solution KB w.r.t.\ the current DPI as $(\mo \setminus \md_{\max}) \cup U_{\Tp\cup\Tp'}$ by means of the current DPI $\langle\mo,\mb,\Tp\cup\Tp',\Tn\cup\Tn'\rangle_\RQ$ and $\md_{\max}$.
This solution KB is then returned as an output of Algorithm~\ref{algo:inter_onto_debug}. If, on the other hand, the stop criterion is not met, the algorithm continues the execution with the computation of another query. 

\begin{remark}\label{rem:mode=static_=>_returned_solution_onto_extensible_to_be_solution_onto_w.r.t._current_DPI}
Notice that the returned maximal solution KB $(\mo \setminus \md_{\max}) \cup U_{\Tp}$ w.r.t.\ the input DPI in case $mode = static$ can be easily extended to constitute a maximal solution KB w.r.t.\ the current DPI, namely by extending it by $U_{\Tp'}$.  
If $mode = dynamic$, then the KB output in line~\ref{algoline:inter_onto_debug:return} is a maximal solution KB w.r.t.\ \emph{the current DPI}, but possibly a non-maximal solution KB w.r.t.\ the input DPI. 
\qed
\end{remark}

\paragraph{Query Computation and User Interaction.} In line~\ref{algoline:inter_onto_debug:calc_query}, the function \textsc{calcQuery} 
is applied to compute a query and the associated q-partition by means of the leading diagnoses $\mD_{\checkmark}$, (possibly) the collected data $qData$, the probability distribution $p_{\mD}()$ of the leading diagnoses, a query selection function $qsm()$ (which might exploit the function $p_{\widetilde{\mo}\cup\overline{\mo}}()$), a parameter $q$ determining the size of the computed query pool 
and the current DPI $\langle\mo,\mb,\Tp\cup\Tp',\Tn\cup\Tn'\rangle_\RQ$. 

As a first step within \textsc{calcQuery}, the function \textsc{getPoolOfQueries} computes a query pool $\QP$ as detailed in Section~\ref{sec:QueryGeneration} from $\mD_{\checkmark}$, $q$ and $\langle\mo,\mb,\Tp\cup\Tp',\Tn\cup\Tn'\rangle_\RQ$.
%
Then, the best tuple $\tuple{Q,\Pt(Q)} \in \QP$ according to the function $qsm()$ is searched for and finally returned as the output of \textsc{calcQuery}. During the query selection process, the evaluation of the query selection measure $qsm(Q) \in \mathbb{R}$ for queries $Q$ where $\tuple{Q,\Pt(Q)}\in\QP$ may require $qData$, the fault probabilities $p_{\mD}()$ of leading diagnoses as well as the fault probabilities $p_{\widetilde{\mo}\cup\overline{\mo}}()$ of syntactical elements in $\mo$. This depends on which concrete measure $qsm()$ is employed (see Section~\ref{sec:query_selection_measures} which presents some possible measures). 

As a next step, the query $Q$ of the best tuple $\tuple{Q,\Pt(Q)} \in \QP$ is presented to the interacting user in line~\ref{algoline:inter_onto_debug:user_interaction} which is the only place in Algorithm~\ref{algo:inter_onto_debug} where user interaction takes place. The user is modeled as a \emph{deterministic} function $u: \mQ_{\mD,\tuple{\mo,\mb,\Tp\cup\Tp',\Tn\cup\Tn'}} \rightarrow \setof{\true,\false}$ that allocates a positive ($\true$) or negative ($\false$) answer to each query w.r.t.\ any set of leading diagnoses $\mD$ for some current DPI $\tuple{\mo,\mb,\Tp\cup\Tp',\Tn\cup\Tn'}$. The answer $u(Q)$ given by the user is stored in the variable $answer$.

\begin{remark}\label{rem:remarks_to_the_user_function_for_answering_queries}
We want to point out that the algorithm can be easily adapted to allow a user to reject queries, e.g.\ if they are not sure how to answer. That is, the user function might be modeled as $u: \mQ_{\mD,\tuple{\mo,\mb,\Tp\cup\Tp',\Tn\cup\Tn'}} \rightarrow \setof{\true,\false, unknown}$ where $u(Q) = unknown$ signifies the rejection of query $Q$. In this case, an accordingly modified version of Algorithm~\ref{algo:inter_onto_debug} would calculate an alternative query w.r.t.\ $\mD$ and $\langle\mo$, $\mb,\Tp\cup\Tp',\Tn\cup\Tn'\rangle$, e.g.\ the second best one according to the query selection measure $qsm()$ among all tuples in $\QP$ (this potential feature is not shown in Algorithm~\ref{algo:inter_onto_debug}). In this vein, a total of $|\QP|-1$ queries can be dismissed per set of leading diagnoses $\mD$.

We want to accentuate that the presented interactive algorithm might be easily adapted to cope with queries whose answer is unknown to the user, but a definite assumption for the algorithm to return a correct solution is a user that does not give wrong answers. In other words, the algorithm does not provide inherent mechanisms that allow for the detection of wrong answers or for the debugging of the KB debugging procedure (keyword ``garbage in, garbage out''). So, we suppose the function $u()$ to be \emph{deterministic} which prohibits the situation that a user might change their mind at a later point in time. Of course, this is still a possible scenario in practice, but in case it arises, a user has to revise, i.e.\ delete or edit, specified test cases they disagree with by hand before a new debugging session using the modified DPI might be started.

Another remark at this place concerns the way a user might choose to answer the query. A ``minimal'' feedback of a user that we regard as an answer to a query $Q$ is to merely say $\true$, i.e.\ each formula in $Q$ (or the conjunction of formulas in $Q$) must be entailed by the correct KB, or $\false$, i.e.\ at least one formula in $Q$ (or the conjunction of formulas in $Q$) must not be entailed by the correct KB. The presented algorithm (Algorithm~\ref{algo:inter_onto_debug}) is designed to deal with exactly this kind of an answer. However, imagine a user being presented $Q$ and think of how they might proceed in order to come up with an answer to $Q$. The first observation is that, in order to respond by $\true$, a user must definitely scrutinize each single formula in $Q$ because otherwise they could never decide for sure whether the conjunction of all formulas in $Q$ is correct. Another observation is that a user might cease to go through the rest of the formulas in case they have already identified one that must not be an entailment of the desired KB. For, in this situation, the overall query $Q$ is already $\false$. This however indicates that at least one formula must be known to be correct or false whatever answer is given to $Q$. Therefore, we can usually expect a user to be able to give exactly this information, namely one formula in $Q$ that must be incorrect, additionally to answering by $\false$. This extra piece of information can be exploited to achieve better space and time efficiency in the context of diagnosis computation. Proposing more efficient algorithms that exploit this information is a topic for future work.
\qed
\end{remark}  

\paragraph{Incorporating the New Information.} The new information represented by the answer $answer$ to $Q$ is incorporated (lines~\ref{algoline:inter_onto_debug:param_update_start}-\ref{algoline:inter_onto_debug:param_update_end}) by updating values of all relevant parameters. First, by means of the function \textsc{append}, the tuple consisting of the answered query $Q$ and the corresponding answer $answer$ given by the user is added as a last element to the chronological list of queries and answers $QA$ that is used for the next probability update (line~\ref{algoline:inter_onto_debug:getProbDist}). 

Then, the subset $\mD_{out}$ of the leading diagnoses $\mD_{\checkmark}$ that gets invalidated after adding $Q$ to the 
positive or negative test cases of the DPI, respectively, is computed by the function \textsc{getInvalidDiags} that gets the q-partition $\Pt(Q) = \tuple{\dx{}(Q),\dnx{}(Q),\dz{}(Q)}$ of $Q$ and $answer$ as input arguments. $\mD_{out}$ then corresponds to the set $\dnx{}(Q)$ given that $answer$ is $\true$ and to $\dx{}(Q)$ otherwise 
(cf. Section~\ref{sec:InterpretationOfQPartitions}). 
Note that $\emptyset \subset \mD_{out} \subset \mD_{\checkmark}$ holds by Proposition~\ref{prop:query_dx_dnx} and since $Q$ is a query w.r.t.\ $\mD_{\checkmark}$ (since $\mD_{\checkmark}$ is given as an input to \textsc{calcQuery}).

As a next step, the data $qData$ is updated. As already pointed out in 
Section~\ref{sec:Variables}, the form of the variable $qData$ depends on the employed query selection measure $qsm()$ and so do the actions that are performed by \textsc{updateQData}.

In order to communicate the impact of the answered query to the hitting set tree algorithm (either \textsc{staticHS} or \textsc{dynamicHS}), the set of invalidated leading diagnoses $\mD_{out}$ is deleted from the leading diagnoses $\mD_{\checkmark}$ and added to $\mD_{\times}$. After this update, $\mD_{\checkmark}$ includes all diagnoses that have been computed by the hitting set tree algorithm so far that are minimal diagnoses w.r.t.\ the current DPI. 

Finally, the new test case $Q$ is added to the new positive test cases $\Tp'$ if $answer$ is $\true$ and to the new negative test cases $\Tn'$ in case of $answer = \false$.

\subsection{Query Selection Measures}
\label{sec:query_selection_measures}
In this section, we give a brief introduction to some query selection measures $qsm()$ that have been suggested and evaluated in literature within the scope of KB or ontology debugging \cite{Shchekotykhin2012,Rodler2013}. Such query selection measures, when used as a parameter in an interactive KB debugging algorithm such as the one described by Algorithm~\ref{algo:inter_onto_debug}, aim at solving the following optimization problems.
In Interactive Dynamic KB Debugging, the problem is defined as follows:\vspace{3pt}

\noindent\fcolorbox{black}{light-gray1}{\parbox[c][2em][c]{0.975\linewidth}{
\begin{prob_def}\label{prob_def:minimize_user_interact_dynamic}
The task is to solve the problem specified by Problem Definition~\ref{prob_def:dynamic} in a way that $|\Tp'|+|\Tn'|$ is minimal.
\end{prob_def}
}}

\vspace{3pt}
In Interactive Static KB Debugging, the problem is defined as follows:\vspace{3pt}

\noindent\fcolorbox{black}{light-gray1}{\parbox[c][2em][c]{0.975\linewidth}{
\begin{prob_def}\label{prob_def:minimize_user_interact_static}
The task is to solve the problem specified by Problem Definition~\ref{prob_def:static} in a way that $|\Tp'|+|\Tn'|$ is minimal.
\end{prob_def}
}}

\vspace{3pt}
That is, these optimization problems aim at the minimization of user effort during interactive KB debugging. In other words, the goal is the minimization of the number of queries required to be asked to a user in order to solve the Interactive Static KB Debugging or the Interactive Dynamic KB Debugging Problem, respectively.

In our previous work \cite{Shchekotykhin2012}, we have discussed entropy-based ($\mathsf{ENT}()$) and split-in-half ($\mathsf{SPL}()$) query selection measures. 

\paragraph{Entropy-Based Query Selection.} A best query $Q_{\mathsf{ENT}}$ according to $\mathsf{ENT}()$ has a maximal information gain among all queries $Q$ where $\tuple{Q,\Pt(Q)}\in\QP$. In other words, $Q_{\mathsf{ENT}}$ minimizes the expected entropy of the 
probability distribution
of the leading diagnoses $\mD_{\checkmark}$ after $Q_{\mathsf{ENT}}$ has been added as a test case to the DPI based on the user's answer $u(Q_{\mathsf{ENT}})$. As shown in \cite{dekleer1987}, this leads to the definition 
\begin{align*}
\mathsf{ENT}(Q) := \sum_{a\in\setof{\true,\false}} p(Q=a) \log p(Q=a) + p(\dz{}(Q))
\end{align*}
where $p()$ in the case of our algorithm corresponds to the leading diagnoses probability measure $p_{\mD}()$ computed in line~\ref{algoline:inter_onto_debug:getProbDist} in Algorithm~\ref{algo:inter_onto_debug} and 
\begin{align*}
p(Q=\true) &= p(\dx{}(Q))+ \frac{1}{2} p(\dz{}(Q)) \\
p(Q=\false) &= p(\dnx{}(Q))+ \frac{1}{2} p(\dz{}(Q))
\end{align*}
(cf.\ Section~\ref{sec:InterpretationOfQPartitions}) where
\begin{align*}
p(\dx{}(Q)) &= \sum_{\md \in \dx{}(Q)} p(\md) \\
p(\dnx{}(Q))&= \sum_{\md \in \dnx{}(Q)} p(\md) \\
p(\dz{}(Q)) &= \sum_{\md \in \dz{}(Q)} p(\md)
\end{align*} 
Then, the best query in a pool $\QP$ according to $qsm():=\mathsf{ENT}()$ is 
\begin{align*}
Q_{\mathsf{ENT}} = \argmin_{\setof{Q\,|\,\tuple{Q,\Pt(Q)}\in\QP}} \mathsf{ENT}(Q)
\end{align*}
So, theoretically optimal w.r.t.\ $\mathsf{ENT}()$ is a query $Q$ whose positive and negative answers are equally likely and for which $\dz{}(Q)$ is the empty set. In other words, the best query has the property that the \emph{sum of probabilities} of leading diagnoses predicting the positive answer as well as the \emph{sum of probabilities} of leading diagnoses predicting the negative answer is $50\%$.

\paragraph{Split-In-Half Query Selection.} For the selection criterion $qsm():=\mathsf{SPL}()$, on the other hand, the query 
\begin{align*}
Q_{\mathsf{SPL}} = \argmin_{\setof{Q\,|\,\tuple{Q,\Pt(Q)}\in\QP}} \mathsf{SPL}(Q)
\end{align*}
is preferred where 
\begin{align*}
\mathsf{SPL}(Q) := \left|\, |\dx{}(Q)| - |\dnx{}(Q)| \,\right| + |\dz{}(Q)|
\end{align*} 
Hence, this measure is optimized by queries $Q$ for which the \emph{number} of leading diagnoses predicting the positive answer is equal to the \emph{number} of leading diagnoses predicting the negative answer and for which $\dz{}(Q)$ is the empty set.

\paragraph{Risk-Optimized Query Selection.} For scenarios where a-priori probabilities are vague, we have presented another more complex query selection measure $\mathsf{RIO}()$ in \cite{Rodler2013} which uses a reinforcement learning strategy to constantly adapt some ``risk'' parameter that indicates the current amount of trust in the probabilities. Whereas $\mathsf{ENT}()$ and $\mathsf{SPL}()$ do not rely on $qData$, this learning strategy does so and requires the invalidation rate or ``performance'', i.e.\ $\frac{|\mD_{out}|}{|\mD_{\checkmark}|}$, of the previous iteration for the adaptation of the learning parameter. As long as the invalidation rate is ``good'', the trust in the current (a-posteriori) probabilities -- that strongly depend on the vague a-priori probabilities -- is high, but it is gradually decreased after observing ``worse'' performance, and so on. High trust in the probabilities means usage of $\mathsf{ENT}()$ which can exploit high quality fault information well as demonstrated in the experiments conducted in \cite{Shchekotykhin2012}, whereas low trust involves selection of queries that guarantee a higher worst case invalidation rate, i.e.\ have similar properties to queries $\mathsf{SPL}()$ would select. 
\begin{example}\label{example:query_selection_measures}
Let us reconsider the queries and associated q-partitions for the example DPI of Table~\ref{tab:example1} that are depicted by Table~\ref{tab:queries_partitions} on page~\pageref{tab:queries_partitions}. Let us denote by $Q_i \prec_{M} Q_j$ that $Q_i$ is preferred over $Q_j$ and by $Q_i \prec\succ_{M} Q_j$ that $Q_i$ is equally preferable as $Q_j$ if the query selection measure $qsm() := M$ is used. Furthermore, we make the assumption that the probability distribution $p_{\mD}$ of the (leading) diagnoses $\mD_{\checkmark} = \setof{\md_1,\dots,\md_4}$ is as shown in Table~\ref{tab:example:query_selection_measures--->diag_probs}. 

Then, we make the following observations: 
\begin{itemize}
	\item $Q_6$ is the theoretically optimal query w.r.t.\ $\mathsf{ENT}()$ since $p_{\mD}(\dx{}(Q_6)) = 0.5$, $p_{\mD}(\dnx{}(Q_6)) = 0.5$ and $\dz{}(Q_6) = \emptyset$, i.e.\ the positive and the negative answer have equal probabilities of $50\%$ and thus $Q_6$ the highest theoretically possible information gain of 1 (bit). This can be compared with one toss of a coin where the information gain of tossing the coin and checking whether it is head or tail is highest in a case where the coin is fair. For a coin that shows head with a probability of $0.95$, conversely, the information gain of tossing the coin is rather small since we are already quite sure about the result in advance.
	\item $Q_9 \prec_{M} Q_5$ as well as $Q_9 \prec_{M} Q_2$ for $M \in \setof{\mathsf{SPL}(),\mathsf{ENT}()}$ because both $Q_5$ and $Q_2$ share one set in $\setof{\dx{},\dnx{}}$ with $Q_9$, but exhibit a non-empty set $\dz{}$ whereas $\dz{}(Q_9) = \emptyset$. This shows that both split-in-half and entropy-based query selection penalize a query $Q$ if there are leading diagnoses that are definitely not discriminated by it, i.e.\ $\dz{}(Q) \neq \emptyset$. This is perfectly desirable as we discussed.
	\item $Q_4 \prec\succ_{M} Q_{10}$ for $M \in \setof{\mathsf{SPL}(),\mathsf{ENT}()}$ since their q-partitions differ just by commutation of the sets $\dx{}$ and $\dnx{}$. This is what one would expect of such a measure, i.e.\ that it does not matter whether the positive or negative answer is more probable if the probability values are the same (in case of $\mathsf{ENT}()$) and whether the number of diagnoses predicting the positive or negative answer is higher if the numbers are the same (in case of $\mathsf{SPL}()$). However, notice that $Q_4$ might be much easier to comprehend and answer for the interacting user. Therefore, $Q_4$ might be preferred in a scenario where some second measure $qsm_2()$ comes into play to identify a best query among equally preferable queries w.r.t.\ some $qsm_1()$ that is used as a primary measure. For, example some ``query-easiness'' measure $qsm_2()$ might be employed after $qsm_1() \in \setof{\mathsf{SPL}(),\mathsf{ENT}()}$ has filtered out an equally preferable set of queries; in this case let this set be $\setof{Q_4,Q_{10}}$. The measure $qsm_2()$ could be defined to simply count the logical connectives and quantifiers occurring in a query $Q$ 
and pick one for which this number is minimal. In this case, this number would be 0 for $Q_4$ and 7 for $Q_{10}$, wherefore $Q_4$ would be decisively better than $Q_{10}$ w.r.t.\ $qsm_2()$. 
	\item It holds that $Q_3 \prec_{\mathsf{ENT()}} Q_{10} \prec_{\mathsf{ENT()}} Q_1$, but $Q_3 \prec\succ_{\mathsf{SPL()}} Q_{10} \prec\succ_{\mathsf{SPL()}} Q_1$. The former holds since all three queries feature an empty set $\dz{}$, but the difference between $p(\dx{})$ and $p(\dnx{})$ is largest for $Q_1$ ($p(\dx{}(Q_1)) = 0.95$), second largest for $Q_{10}$ ($p(\dnx{}(Q_{10})) = 0.85$) and smallest for $Q_3$ ($p(\dx{}(Q_3)) = 0.7$).
	\item $Q_9$ is the second best query among those given in Table~\ref{tab:queries_partitions} because both answers of it are almost equally probable (positive answer has a probability of 0.55 and negative answer a probability of $0.45$).
	\item Queries $Q_7$, $Q_8$ and $Q_9$ are theoretically optimal w.r.t.\ the $\mathsf{SPL()}$ measure, since $\dz{} = \emptyset$ and $|\dx{}| = |\dnx{}|$ for all of them.
	\item Regarding the $\mathsf{RIO()}$ measure, queries $Q_7$, $Q_8$ and $Q_9$ are ``no risk'' queries since they feature the maximum possible worst case elimination rate of $50\%$. $Q_2$ and $Q_6$, for instance, have a ``higher risk'' as their minimal invalidation rate amounts to only $25\%$. That is, if $Q_2$ ($Q_6$) is answered positively (negatively), then only one of four leading diagnoses is invalidated.\qed
\end{itemize}
\end{example}

\begin{table}[tb]
	\centering
		\begin{tabular}{lcccc}
			$\md \in \mD_{\checkmark}$ & $\md_1$ & $\md_2$ & $\md_3$ & $\md_4$ \\\hline
			$p_{\mD}(\md)$             & 0.15  & 0.3  & 0.05  & 0.5  
		\end{tabular}
\caption[(Example~\ref{example:query_selection_measures}) Diagnoses Probabilities]{(Example~\ref{example:query_selection_measures}) Diagnosis probabilities for the example DPI given by Table~\ref{tab:example1}.}
\label{tab:example:query_selection_measures--->diag_probs}
\end{table}

\subsection[Correctness]{Interactive Debugging Algorithm: Correctness}
\label{sec:CorrectnessOfAlgorithmInterOntoDebug}
In this section we prove the correctness of Proposition~\ref{prop:correctness_of_interactive_KB_debugging_algo} on page~\pageref{prop:correctness_of_interactive_KB_debugging_algo} by using the results of Sections~\ref{sec:StaticAlgorithmWalkthrough} and \ref{sec:DynamicAlgorithmWalkthrough} which provide evidence for the correctness (soundness, completeness and optimality) of methods \textsc{staticHS} and \textsc{dynamicHS}:
\begin{proof}[Proof of Proposition~\ref{prop:correctness_of_interactive_KB_debugging_algo}]
First, we argue why Algorithm~\ref{algo:inter_onto_debug} must terminate. The function \textsc{getFormulaProbs} in line~\ref{algoline:inter_onto_debug:getAxiomProbs} terminates since it applies Formulas~\ref{eq:ax_prob_calc} and \ref{eq:adapt_ax_prob_to_get_min_diags} $|\mo|$ times and $|\mo|$ is finite by Definition~\ref{def:dpi}. If $mode=static$, then \textsc{staticHS} terminates due to Proposition~\ref{prop:static_hs_correctness}. If $mode=dynamic$, then \textsc{dynamicHS} terminates due to Proposition~\ref{prop:dynamic_hs_correctness}. \textsc{getProbDist} terminates since (1)~the number of already answered queries $|QA|$ is finite, (2)~$|\mD_{\checkmark}|$ is finite since diagnoses are subsets of $\mo$ and thus there is only a finite number of (minimal) diagnoses w.r.t.\ any DPI according to Definition~\ref{def:dpi} (since all sets included in the DPI are finite) and (3)~reasoning (\textsc{getEntailments} and \textsc{isKBValid}) is assumed to be decidable for the logic $\mathcal{L}$ over which the DPI is formulated as per Chapter~\ref{chap:basics}. Further, \textsc{getMode} clearly terminates due to the fact that $|\mD_{\checkmark}|$ is finite and returns the mode $\md_{\max}$ of the diagnoses probability distribution $p_{\mD}()$ over the diagnoses in $\mD_{\checkmark}$. Now, if the stop criterion $p_{\mD}(\md_{\max}) \geq 1 -\sigma$ is met, then \textsc{getSolKB} is called. \textsc{getSolKB} simply deletes the given diagnosis $\md_{\max}$ from the given KB $\mo$ and adds a finite set of formulas to it, and thence terminates.

If the stop criterion is not met, then $|\mD_{\checkmark}| \geq 2$ must hold as otherwise the single diagnosis $\md \in \mD_{\checkmark}$ would necessarily have fulfilled the stop criterion as its probability as per any probability measure over the sample space $\Omega := \mD_{\checkmark}$ must be equal to 1 and thus greater than or equal to $1 - \sigma$ where $\sigma \geq 0$. 

Due to $|\mD_{\checkmark}| \geq 2$, Proposition~\ref{prop:getPoolOfQueries_correctness} implies that
\textsc{getPoolOfQueries} (called within \textsc{calcQuery}) terminates and yields a non-empty query pool as output. \textsc{selectBestQuery} (also called within \textsc{calcQuery}) terminates as well since it simply selects one query from the pool according to the measure $qsm()$ (cf.\ Section~\ref{sec:query_selection_measures}). Since we assume the interacting user to answer to a query or to reject it within finite time, $u(Q)$ also terminates. It is clear that \textsc{append} terminates. \textsc{getInvalidDiags} simply extracts one entry of the given q-partition and thus terminates. Finally, \textsc{updateQData} also terminates by assumption (no $qsm()$ must be used for which \textsc{updateQData} might not terminate). As a consequence, all functions called in Algorithm~\ref{algo:inter_onto_debug} terminate. What remains to be proven is that the stop criterion must be met after a finite number of iterations, i.e.\ after a finite number of test cases have been added to the input DPI.

In $mode=static$ the stop criterion must be satisfied after a finite number of iterations due to the following argumentation: 
\begin{itemize}
	\item There is a finite set of minimal diagnoses w.r.t.\ the input DPI $\tuple{\mo,\mb,\Tp,\Tn}_{\RQ}$ since each (minimal) diagnosis w.r.t.\ this DPI is a subset of $\mo$ according to Definition~\ref{def:diagnosis} and since $|\mo|$ is finite by Definition~\ref{def:dpi}.
	\item In each iteration, one test case is added either to $\Tp'$ or $\Tn'$.
	\item Each test case added to whatever set $\Tp'$ or $\Tn'$ invalidates at least one minimal diagnosis w.r.t.\ the input DPI in the set $\mD_{\checkmark}$ by the definition of a query (Definition~\ref{def:query}) and since each query is computed w.r.t.\ the leading diagnoses $\mD_{\checkmark}$ by the correctness of \textsc{getPoolOfQueries} (cf.\ Proposition~\ref{prop:getPoolOfQueries_correctness}).
	\item $\mD_{\checkmark}$ contains only minimal diagnoses w.r.t.\ the input DPI by Proposition~\ref{prop:static_hs_correctness}.
	\item Also by Proposition~\ref{prop:static_hs_correctness}, no invalidated minimal diagnosis w.r.t.\ the input DPI can be an element of some subsequent set of leading diagnoses $\mD_{\checkmark}$.
	\item Therefore, unless the stop criterion is met before due to a sufficiently high probability of one of multiple leading diagnoses as per $p_{\mD}()$, Algorithm~\ref{algo:inter_onto_debug} in $mode=static$ must arrive at a point where $|\mD_{\checkmark}| = 1$ after a finite number of iterations. Note that $|\mD_{\checkmark}| = 0$ is impossible due to the definition of a query (Definition~\ref{def:query}) which ensures that each added test case leaves valid at least one minimal diagnosis in $\mD_{\checkmark}$.
\end{itemize}


Algorithm~\ref{algo:inter_onto_debug} terminates in $mode=dynamic$ since for any sequence $QA$ of queries that are added to the positive or negative test cases $\Tp'$ or $\Tn'$, respectively, there is a finite number $k_{QA}$ such that there is no more than one minimal diagnosis w.r.t.\ $\tuple{\mo,\mb,\Tp\cup\Tp',\Tn\cup\Tn'}_{\RQ}$ for $|\Tp'| + |\Tn'| = k_{QA}$ wherefore the stop criterion must be met.
Now, let us assume that the opposite holds. That is, there is a sequence $QA^*$ of queries that are added to the positive or negative test cases $\Tp'$ or $\Tn'$, respectively, and for all natural numbers $k$ there is more than one minimal diagnosis w.r.t.\ $\tuple{\mo,\mb,\Tp\cup\Tp',\Tn\cup\Tn'}_{\RQ}$ for $|\Tp'| + |\Tn'| = k$. Then we argue as follows to derive a contradiction:
\begin{itemize}
	\item There is a finite set of (minimal) diagnoses w.r.t.\ any DPI $\tuple{\mo,\mb,\Tp\cup\Tp',\Tn\cup\Tn'}_{\RQ}$ obtained from the input DPI by the addition of test cases. This is true since $|\mo|$ is finite by Definition~\ref{def:dpi} and since each (minimal) diagnosis w.r.t.\ $\tuple{\mo,\mb,\Tp\cup\Tp',\Tn\cup\Tn'}_{\RQ}$ is a subset of $\mo$ according to Definition~\ref{def:diagnosis}.
	\item In each iteration, one test case is added either to $\Tp'$ or $\Tn'$.
	\item Each test case added to whatever set $\Tp'$ or $\Tn'$ invalidates at least one minimal diagnosis w.r.t.\ the current DPI in the set $\mD_{\checkmark}$ by the definition of a query (Definition~\ref{def:query}) and since each query is computed w.r.t.\ the leading diagnoses $\mD_{\checkmark}$ by the correctness of \textsc{getPoolOfQueries} (cf.\ Proposition~\ref{prop:getPoolOfQueries_correctness}).
	\item If $DPI$ denotes the current DPI at the time \textsc{dynamicHS} is called, then the set $\mD_{\checkmark}$ returned by \textsc{dynamicHS} is a subset of or equal to $\minD_{DPI}$, i.e.\ $\mD_{\checkmark}$ contains only minimal diagnoses w.r.t.\ $DPI$ by Proposition~\ref{prop:dynamic_hs_correctness}.
	\item Let $\tuple{DPI_0, DPI_1,\dots}$ denote the sequence of DPIs encountered in the case of adding answered queries as test cases to the input DPI $DPI_0$ as per $QA^*$. Further, let $\tuple{\allD_0, \allD_1, \dots}$ be the sequence 
such that $\allD_i:= \allD_{DPI_i}, i=0,1,\dots$, i.e.\ $\allD_i$ is the set of all diagnoses w.r.t.\ $DPI_i$. Then $\allD_i \supset \allD_{i+1}$ for all $i \geq 0$. 
	\item As each query added as a test case to $DPI_i$ leaves valid at least one (minimal) diagnosis w.r.t.\ $DPI_i$ due to Definition~\ref{def:query}, we have that $\allD_k \supset \emptyset$ for $k = 0,1,\dots$.
	\item Since $\allD_i$ is finite, there must be some finite number $k^*$ such that $|\allD_{k^*}| = 1$ wherefore $|\minD_{k^*}| = 1$ must also be valid. This is a contradiction.
\end{itemize}
Thence, Algorithm~\ref{algo:inter_onto_debug} terminates in any mode $mode$. 
Now, we show that propositions (\ref{prop:correctness_of_interactive_KB_debugging_algo:stat1})-(\ref{prop:correctness_of_interactive_KB_debugging_algo:stat6}) of Proposition~\ref{prop:correctness_of_interactive_KB_debugging_algo} hold for (i)~$mode=static$ and (ii)~$mode=dynamic$.

(i): First, by the proof so far, we have that Algorithm~\ref{algo:inter_onto_debug} in $mode=static$ given the input DPI $\langle\mo,\mb,\Tp,\Tn\rangle_\RQ$ terminates. Since the only point where the algorithm can terminate is line~\ref{algoline:inter_onto_debug:return}, \textsc{getSolKB} is called with arguments $\tuple{\md_{\max}, \langle\mo,\mb,\Tp\cup\Tp',\Tn\cup\Tn'\rangle_\RQ, \Tp', static}$. By the definition of \textsc{getSolKB} (see Section~\ref{sec:Walkthrough}), we have that $(\mo \setminus \md_{\max}) \cup U_{\Tp}$ is returned by the algorithm.

Propositions~(\ref{prop:correctness_of_interactive_KB_debugging_algo:stat1}) and (\ref{prop:correctness_of_interactive_KB_debugging_algo:stat2}) follow from the specification of the \textsc{getMode} function which is called with arguments $\tuple{\mD_{\checkmark},p_{\mD}()}$. Proposition~(\ref{prop:correctness_of_interactive_KB_debugging_algo:stat3}) is true since \textsc{getSolKB} can never be reached without $p_{\mD}(\md_{\max}) \geq 1 -\sigma$ being fulfilled. 
$\mD_{\checkmark} \subseteq \minD_{\langle\mo,\mb,\Tp,\Tn\rangle_\RQ} \cap \minD_{\langle\mo,\mb,\Tp\cup\Tp',\Tn\cup\Tn'\rangle_\RQ}$ is true due to Proposition~\ref{prop:static_hs_correctness}, Remark~\ref{rem:p_D,prio()_same_order_for_diags_as_p_nodes()} and the fact that $\mD_{\checkmark}$ is obtained as an output of \textsc{staticHS}. Hence, Proposition~(\ref{prop:correctness_of_interactive_KB_debugging_algo:stat4}) holds. Proposition~(\ref{prop:correctness_of_interactive_KB_debugging_algo:stat5}) is implied by Remark~\ref{rem:p_D,prio()_same_order_for_diags_as_p_nodes()} and by the specification of the \textsc{getFormulaProbs} function which computes $p_{\mo}()$ from $p_{\widetilde{\mo} \cup \overline{\mo}}()$ as per Formulas~\ref{eq:ax_prob_calc} and \ref{eq:adapt_ax_prob_to_get_min_diags} in line~\ref{algoline:inter_onto_debug:getAxiomProbs}. Finally, Proposition~(\ref{prop:correctness_of_interactive_KB_debugging_algo:stat6}) is a consequence of the definition of the \textsc{getProbDist} function which accounts for the computation of $p_{\mD}()$ from $p_{\mo}()$, the input DPI, $\mD_{\checkmark}$ and the chronological sequence of all queries and associated answers $QA$ so far. Therefore, Proposition~\ref{prop:correctness_of_interactive_KB_debugging_algo} is true for $mode=static$.

(ii): First, by the proof so far, we have that Algorithm~\ref{algo:inter_onto_debug} in $mode=dynamic$ given the input DPI $\langle\mo,\mb,\Tp,\Tn\rangle_\RQ$ terminates. Since the only point where the algorithm can terminate is line~\ref{algoline:inter_onto_debug:return}, \textsc{getSolKB} is called with arguments $\tuple{\md_{\max}, \langle\mo,\mb,\Tp\cup\Tp',\Tn\cup\Tn'\rangle_\RQ, \Tp', dynamic}$. By the definition of \textsc{getSolKB} (see Section~\ref{sec:Walkthrough}), we have that $(\mo \setminus \md_{\max}) \cup U_{\Tp\cup\Tp'}$ is returned by the algorithm.

Propositions~(\ref{prop:correctness_of_interactive_KB_debugging_algo:dyn1}) and (\ref{prop:correctness_of_interactive_KB_debugging_algo:dyn2}) follow from the specification of the \textsc{getMode} function which is called with arguments $\tuple{\mD_{\checkmark},p_{\mD}()}$. Proposition~(\ref{prop:correctness_of_interactive_KB_debugging_algo:dyn3}) is true since \textsc{getSolKB} can never be reached without $p_{\mD}(\md_{\max}) \geq 1 -\sigma$ being fulfilled. 
$\mD_{\checkmark} \subseteq \minD_{\langle\mo,\mb,\Tp\cup\Tp',\Tn\cup\Tn'\rangle_\RQ}$ is true due to Proposition~\ref{prop:dynamic_hs_correctness}, Remark~\ref{rem:p_D,prio()_same_order_for_diags_as_p_nodes()} and the fact that $\mD_{\checkmark}$ is obtained as an output of \textsc{dynamicHS}. Hence, Proposition~(\ref{prop:correctness_of_interactive_KB_debugging_algo:dyn4}) holds. Proposition~(\ref{prop:correctness_of_interactive_KB_debugging_algo:dyn5}) is implied by Remark~\ref{rem:p_D,prio()_same_order_for_diags_as_p_nodes()} and by the specification of the \textsc{getFormulaProbs} function which computes $p_{\mo}()$ from $p_{\widetilde{\mo} \cup \overline{\mo}}()$ as per Formulas~\ref{eq:ax_prob_calc} and \ref{eq:adapt_ax_prob_to_get_min_diags} in line~\ref{algoline:inter_onto_debug:getAxiomProbs}. Finally, Proposition~(\ref{prop:correctness_of_interactive_KB_debugging_algo:dyn6}) is a consequence of the definition of the \textsc{getProbDist} function which accounts for the computation of $p_{\mD}()$ from $p_{\mo}()$, the input DPI, $\mD_{\checkmark}$ and the chronological sequence of all queries and associated answers $QA$ so far. Therefore, Proposition~\ref{prop:correctness_of_interactive_KB_debugging_algo} is true for $mode=dynamic$. 

Next, we show that the solution to Interactive Static KB Debugging is found for $\sigma = 0$ in case $mode = static$:
\begin{enumerate}[(s1)]
	\item $\mD_{\checkmark} \subseteq \minD_{\langle\mo,\mb,\Tp,\Tn\rangle_\RQ} \cap \minD_{\langle\mo,\mb,\Tp\cup\Tp',\Tn\cup\Tn'\rangle_\RQ}$ holds for the output of \textsc{staticHS} in each iteration by Proposition~\ref{prop:static_hs_correctness}. Therefore, $\mD_{\checkmark}$ comprises only minimal diagnoses w.r.t.\ the input DPI that comply with all specified test cases in $\Tp'$ and $\Tn'$.
	\item \label{proof_int_debug_correct:bullet_2} By $p_{\widetilde{\mo} \cup \overline{\mo}}(): \widetilde{\mo} \cup \overline{\mo} \rightarrow (0,1]$ we derive by Formula~\ref{eq:ax_prob_calc} that each formula in $\mo$ must have a probability greater than zero. Further, by Formula~\ref{eq:adapt_ax_prob_to_get_min_diags}, no formula in $\mo$ can have a probability greater than or equal to $0.5$ (i.e.\ in particular a probability of 1 is not possible for a formula). Hence, we have that $p_{\mo}: \mo \rightarrow (0,0.5)$ for the measure $p_{\mo}()$ computed by \textsc{getFormulaProbs} in line~\ref{algoline:inter_onto_debug:getAxiomProbs} in Algorithm~\ref{algo:inter_onto_debug}. Thence, by the definition of $p_{nodes}()$ in \textsc{staticHS} based on $p() := p_{\mo}()$ (cf.\ Definition~\ref{def:p_node()} on page~\pageref{def:p_node()}) due to the fact that $p_{\mo}()$ is given as an input argument to $\textsc{staticHS}$ in line~\ref{algoline:inter_onto_debug:staticHS}, we have that no diagnosis can have an (a-priori) probability of zero. Since the function \textsc{getProbDist} might only perform some multiplications of a diagnosis probability by $\frac{1}{2}$, also the a-posteriori  probability of each diagnosis must be greater than zero.
	\item \label{proof_int_debug_correct:bullet_3} Hence, due to $\sigma = 0$, it must be necessarily be true that $|\mD_{\checkmark}| = 1$ before the algorithm terminates.
	\item By Problem Definition~\ref{prob_def:static} and the specification of the \textsc{getSolKB} function, the output solution KB must be the solution to Interactive Static KB Debugging.
\end{enumerate}
That a solution found for $\sigma > 0$ in case $mode = static$ might be an approximate solution to Interactive Static KB Debugging is a direct consequence of the definition of approximate solution given in Remark~\ref{rem:approximate_solution}.
%
%

Finally, the proof that the solution to Interactive Dynamic KB Debugging is found for $\sigma = 0$ in case $mode = dynamic$ is analogue to the one for $mode = static$, just 
\begin{enumerate}[(d1)]
	\item $\mD_{\checkmark} \subseteq \minD_{\langle\mo,\mb,\Tp\cup\Tp',\Tn\cup\Tn'\rangle_\RQ}$ holds for the output of \textsc{dynamicHS} in each iteration by Proposition~\ref{prop:dynamic_hs_correctness}. Therefore, $\mD_{\checkmark}$ comprises only minimal diagnoses w.r.t.\ the current DPI.
	\item By (s\ref{proof_int_debug_correct:bullet_2}), (s\ref{proof_int_debug_correct:bullet_3}), Problem Definition~\ref{prob_def:dynamic} and the specification of the \textsc{getSolKB} function, the output solution KB must be the solution to Interactive Dynamic KB Debugging.
\end{enumerate}
That a solution found for $\sigma > 0$ in case $mode = dynamic$ might be an approximate solution to Interactive Dynamic KB Debugging is a direct consequence of the definition of approximate solution given in Remark~\ref{rem:approximate_solution}.

This completes the proof of Proposition~\ref{prop:correctness_of_interactive_KB_debugging_algo}. 
\end{proof}

\chapter{Iterative Diagnosis Computation}
\label{chap:IterativeDiagnosisComputation}
In this chapter we will introduce and discuss two methods, \textsc{staticHS} and \textsc{dynamicHS}, which are called in lines~\ref{algoline:inter_onto_debug:staticHS} and \ref{algoline:inter_onto_debug:dynamicHS} of Algorithm~\ref{algo:inter_onto_debug}, respectively. The former provides a method for solving the Interactive Static KB Debugging Problem (Problem Definition~\ref{prob_def:static}) whereas the latter aims at solving the Interactive Dynamic KB Debugging Problem (Problem Definition~\ref{prob_def:dynamic}). Both are methods for iterative diagnosis computation that are employed to compute a set of leading diagnoses in each iteration of the presented interactive KB debugging algorithm (Algorithm~\ref{algo:inter_onto_debug}). Each time a query has been answered by the interacting user and added to the respective set of test cases of the DPI, a subset of the leading diagnoses (and usually also a set of not-yet-computed minimal diagnoses) is invalidated. An iterative diagnosis computation method is then invoked to update the leading diagnoses set taking the new information into account that is given by the recently added test case.
That is, the $k \leq n_{\max}$ most probable ways of solving the Interactive Static (Dynamic) KB Debugging Problem in the light of the new evidence are extracted by \textsc{staticHS} (\textsc{dynamicHS}) after the search space has been suitably pruned. In this vein, if there is only one solution left, the (exact) solution of Interactive Static (Dynamic) KB Debugging has been found. 
%

\section[\textsc{staticHS}: A Static Iterative Diagnosis Computation Algorithm]{\textsc{staticHS}: A Static Iterative Diagnosis Computation Algorithm%
\sectionmark{Static Algorithm}}
\sectionmark{Static Algorithm}
\label{sec:StaticHSTree}
As the name already suggests, \textsc{staticHS} (Algorithm~\ref{algo:inter_stat_hs}) is a procedure that solves the problem of \emph{Interactive Static KB Debugging} defined by Problem Definition~\ref{prob_def:static} if used for leading diagnosis computation in Algorithm~\ref{algo:inter_onto_debug}. \textsc{staticHS} is sound, complete and optimal w.r.t.\ the set of solutions of the \emph{Interactive Static KB Debugging} problem. 
Optimality refers to the best-first computation of minimal diagnoses regarding a given probability measure.
  
\subsection{Overview and Intuition}
\label{sec:TheIntuition}
The \textsc{staticHS} algorithm is strongly related to the non-interactive hitting set algorithm \textsc{HS} (see Algorithm~\ref{algo:hs}) in that, at any stage during the execution of Algorithm~\ref{algo:inter_onto_debug}, the hitting set tree produced by \textsc{staticHS} corresponds to some part of the complete (non-interactive) wpHS-tree built-up by Algorithm~\ref{algo:hs}. This is achieved by the strategy to \emph{use new test cases only for the invalidation of diagnoses, and not for the computation of conflict sets (and thus diagnoses)}. That is, all minimal conflict sets are computed w.r.t.\ the input DPI. Thereby, the introduction of new diagnoses, i.e.\ ones that are not minimal diagnoses w.r.t.\ the input DPI, through addition of new test cases to the DPI is prohibited (cf.\ Proposition~\ref{prop:mindiag_mincs}).

So, what \textsc{staticHS} as a subroutine of Algorithm~\ref{algo:inter_onto_debug} does is gradually building up the standard (non-interactive) wpHS-tree in multiple phases. During each phase some new (not-yet-computed) minimal diagnoses w.r.t.\ the \emph{input DPI} are computed, in the order of their probability, most probable ones first. Before such a newly detected minimal diagnosis is added to the set of leading diagnoses ($\mD_{calc}\cup\mD_{\checkmark}$), a test is performed that verifies that this new diagnosis is consistent with all test cases added to the input DPI so far. In this vein, all answered queries so far not only serve to eliminate a subset of the set of leading diagnoses at the time when the respective query is answered, but also to eliminate incompatible minimal diagnoses w.r.t.\ the input DPI that are found at some later point in time. However, in order to be eliminated due to a specified test case, a minimal diagnosis must first be computed. That is, no partial diagnoses can be eliminated due to newly specified test cases.

Between each two phases of tree construction, a query computed on the basis of the current set of leading diagnoses is asked to the user (this is accomplished directly in Algorithm~\ref{algo:inter_onto_debug}). 
After incorporating the user's answer, some leading diagnoses are eliminated (this is granted by the definition of a query, see Definition~\ref{def:query}). Moreover, the ``state'' of the tree is maintained during the execution of Algorithm~\ref{algo:inter_onto_debug} until \textsc{staticHS} is again called in order to calculate further leading diagnoses. The state of the current partial wpHS-tree is stored by variables 
\begin{itemize}
	\item $\mD_{calc} \cup \mD_{\checkmark}$ -- computed minimal diagnoses w.r.t.\ the input DPI consistent with all test cases specified so far,
	\item $\Queue$ -- the list of open, non-labeled nodes,
	\item $\mC_{calc}$ -- minimal conflict sets w.r.t.\ the input DPI computed so far and
	\item $\mD_{\times}$ -- computed minimal diagnoses w.r.t.\ the input DPI not consistent with all test cases specified so far.
\end{itemize} 

Each time a tree construction phase, i.e.\ the computation of new leading diagnoses, is finished, a new diagnosis probability distribution is obtained by the diagnosis probability update as per Bayes' Theorem described in Section~\ref{sec:DetailedAlgorithmDescription}. Once this distribution involves one highly probable diagnosis (the probability of which exceeds a predefined threshold $1-\sigma$) and else just highly improbable ones, the algorithm terminates. The output is a solution KB w.r.t.\ the \emph{input DPI} built from this highly probable minimal diagnosis. 

\begin{remark}\label{rem:sigma_in_staticHS}
In case $\sigma$ has a predefined value of zero, the output is the (exact) solution to the problem of \emph{Interactive Static KB Debugging} for the input DPI. In a scenario where some fault tolerance $\sigma > 0$ is given, the solution KB returned by Algorithm~\ref{algo:inter_onto_debug} is an approximation of the (exact) solution to \emph{Interactive Static KB Debugging} for the input DPI where a better approximation can be expected for smaller values of $\sigma$ (cf.\ Remark~\ref{rem:approximate_solution}). ``Better'' in this context refers to the satisfaction of desired semantic properties of the KB returned by Algorithm~\ref{algo:inter_onto_debug}, i.e.\ desired entailments and desired non-entailments of the KB. The intuition is that the specification of additional test cases $T$ guarantees the output of a KB complying with these test cases, whereas accepting 
one -- albeit highly probable -- of multiple solution KBs without having incorporated $T$ leaves open the possibility for this KB to not fulfill $T$.

However, answering queries is effort for an interacting user. Therefore, the approach that involves the ``early'' termination of the algorithm after a solution KB has a sufficiently high probability (lower than 1) constitutes a trade-off between exactness of the output and the effort of the user and overall execution time of the interactive KB debugging algorithm, respectively.\qed
\end{remark}

\paragraph{Constant ``Convergence'' towards the Solution.} As said, each added test case is an answered \emph{query} and thus eliminates at least one minimal diagnosis w.r.t.\ the input DPI. And, only minimal diagnoses w.r.t.\ the input DPI are computed by \textsc{staticHS}. Hence, by the fact that a solution to Interactive Static KB Debugging can only be constructed from a minimal diagnosis w.r.t.\ the input DPI, it is guaranteed that the number of solutions to Interactive Static KB Debugging is strictly monotonically decreasing throughout the execution of Algorithm~\ref{algo:inter_onto_debug}. That is, the initial number of (all) minimal diagnoses (w.r.t.\ the input DPI) is ``static'' which means that no ``new'' minimal diagnoses can be introduced when the input DPI is extended by new test cases. 

As a consequence of this, it is reasonable to employ \textsc{staticHS} in a situation where the (complete) wpHS-tree produced by the standard (non-interactive) algorithm \textsc{HS} is believed to be as compact as to fit into the available system memory. In this case, \textsc{staticHS} is also guaranteed to not exceed the available memory, even if an exact solution ($\sigma = 0$) is intended. 

Unfortunately, however, it will be generally the case that a complete enumeration of all minimal diagnoses is intractable, especially due to an overwhelming space complexity. In such a case, 
Algorithm~\ref{algo:inter_onto_debug} using \textsc{staticHS} will definitely run out of memory (given that \textsc{staticHS} is called sufficiently often). The reason is that the space consumption of \textsc{staticHS} will sooner or later definitely reach the huge extent of the wpHS-tree produced by \textsc{HS}. Nevertheless, \textsc{staticHS} might be used to (possibly) find some (approximate) solution. This might work in a scenario where the given probabilistic information in terms of $p_{\widetilde{\mo}\cup\overline{\mo}}()$ provided as an input to Algorithm~\ref{algo:inter_onto_debug} is ``reasonable'' in that the desired diagnosis is assigned a rather high probability and is thus figured out early, before the available memory is exhausted. 

A possible modification of the stop criterion in \textsc{staticHS} in a way that new leading diagnoses are not computed until a desired number of such is detected or a timeout is reached, but rather until a predefined maximum space is consumed, would not mitigate space complexity issues very much. An explanation for this is that stopping \textsc{staticHS} on account of no more available memory implies that no further call of \textsc{staticHS} will be able to execute. That is because, as mentioned before, an added test case can only invalidate already computed diagnoses, no other branches in the wpHS-tree, and each invalidated minimal diagnosis cannot be discarded, but must be stored (in $\mD_{\times}$) to avoid the usage of leading diagnoses that are non-minimal w.r.t.\ the input DPI (cf.\ lines~\ref{algoline:slabel:non_min_crit_start}-\ref{algoline:slabel:non_min_crit_end} in Algorithm~\ref{algo:inter_stat_hs}). 
%

\paragraph{Poor Pruning.} As we explained before, the preservation of a constantly shrinking set of minimal diagnoses comes at the cost of being able to exploit new test cases only partially, i.e.\ only for the invalidation of already computed minimal diagnoses w.r.t.\ the input DPI and not for the computation of minimal conflict sets and thus minimal diagnoses. 
%
%
The incorporation of test cases into the DPI that is used to determine minimal conflict sets (line~\ref{algoline:slabel:qx} in Algorithm~\ref{algo:inter_stat_hs}) could, on the one hand, lead to new minimal conflict sets that are no minimal conflict sets w.r.t.\ the input DPI. As a consequence of this, minimal diagnoses might be determined by the algorithm which are no minimal diagnoses w.r.t.\ the input DPI, but w.r.t.\ the current DPI. Hence, the soundness of \textsc{staticHS} w.r.t.\ the set of solutions of the Interactive Static KB Debugging problem would be violated. Furthermore, such conflict sets 
could lead to the missing of some minimal diagnoses w.r.t.\ the input DPI, a violation of the completeness of \textsc{staticHS} w.r.t.\ the set of solutions of the Interactive Static KB Debugging problem. 

On the other hand, the exploitation of new test cases for conflict set generation might give rise to the possibility of pre-pruning of \emph{any} tree branches, not just branches that already correspond to diagnoses w.r.t.\ the input DPI. Such a ``dynamic'' strategy which exploits the new information given by a test case not just partially, but for the invalidation \emph{and computation} of diagnoses and conflict sets, will be implemented be \textsc{dynamicHS} which we will detail in Section~\ref{sec:DynamicHSTree}.

Put another way, in \textsc{staticHS} only the standard pruning rules for the construction of a wpHS-tree are applicable, namely the deletion of duplicate nodes and the elimination of non-minimal diagnoses (cf.\ Definition~\ref{def:weighted_pruned_hs_tree}). Newly defined test cases only facilitate the deletion of tree branches from the leading diagnoses set $\mD_{calc} \cup \mD_{\checkmark}$, but not from memory (as invalidated minimal diagnoses must be stored in $\mD_{\times}$, as pointed out before).  

To summarize, \textsc{staticHS} on the one hand makes sure to only consider relevant solutions 
of the problem of Interactive Static KB Debugging,
but on the other hand suffers from this conservative strategy in that tree pruning cannot be designed very effectively. So, on the positive side, uncontrolled growth of the produced wpHS-tree can be avoided, but, on the negative side, consultation of an interacting user cannot be taken advantage of in terms of reduction of the space complexity of \textsc{staticHS} compared to the construction of a wpHS-tree by a non-interactive procedure like Algorithm~\ref{algo:hs}.

\subsection{Algorithm Walkthrough}
\label{sec:StaticAlgorithmWalkthrough}
\paragraph{Input Parameters.} When \textsc{staticHS} (Algorithm~\ref{algo:inter_stat_hs}) is called for the first time in Algorithm~\ref{algo:inter_onto_debug}, the inputs $\mC_{calc}$, $\mD_{\checkmark}$, $\mD_{\times}$, $\Tp'$ and $\Tn'$ correspond to the empty set and $\Queue = [\emptyset]$ (cf.\ lines~\ref{algoline:inter_onto_debug:var_inst_start}-\ref{algoline:inter_onto_debug:var_inst_end} and \ref{algoline:inter_onto_debug:staticHS} in Algorithm~\ref{algo:inter_onto_debug}). Further on, $\mD_{calc}$ is defined to be the empty set at the beginning of each execution of \textsc{staticHS}. That is, \textsc{staticHS} starts the construction of the wpHS-tree from an initial tree consisting of a single unlabeled root node $\emptyset$ ($\in \Queue$). And, all collections that are later returned by \textsc{staticHS}, except for $\Queue$, are initially empty. Further input arguments are the DPI $\langle\mo,\mb,\Tp,\Tn\rangle_\RQ$ provided as an input to Algorithm~\ref{algo:inter_onto_debug}, the sets of positively ($\Tp'$) and negatively ($\Tn'$) answered queries since the start of Algorithm~\ref{algo:inter_onto_debug}, the leading diagnosis computation parameters $n_{\min},n_{\max},t$ (see description in Section~\ref{sec:UserInteraction} on page~\pageref{etc:leading_diag_params}) and the probability measure $p() := p_{\mo}()$ that assigns a probability in the interval $(0,0.5)$ to each formula in $\mo$ (cf.\ line~\ref{algoline:inter_onto_debug:getAxiomProbs} in Algorithm~\ref{algo:inter_onto_debug}).

\paragraph{The Main Loop.} During the repeat-loop, in each iteration the first node $\mathsf{node}$ in $\Queue$ is processed (\textsc{getFirst}, line~\ref{algoline:static:getfirst}). That is, $\mathsf{node}$ is deleted from $\Queue$ (\textsc{deleteFirst}, line~\ref{algoline:static_update_Q}) and the \textsc{sLabel} function is called given $\mathsf{node}$ (i.a.) as a parameter. Notice that elements are added to $\Queue$ (line~\ref{algoline:static:generate_nodes}) in a way that a sorting of $\Queue$ in descending order according to $p_{nodes}()$ (cf.\ Definition~\ref{def:p_node()}) is maintained throughout the execution of \textsc{staticHS}.

\paragraph{Computation of a Node Label.} The \textsc{sLabel} function processes $\mathsf{node}$ as follows. First, the \emph{non-minimality criterion} (lines~\ref{algoline:slabel:non_min_crit_start}-\ref{algoline:slabel:non_min_crit_end}) is checked. That is, among all nodes in $\mD_{(\times,\checkmark,calc)} = \md_{\times} \cup \mD_{\checkmark} \cup \mD_{calc}$ one is searched which is a subset of $\mathsf{node}$. If such a node $\mathsf{nd}$ is found, then $\mathsf{node}$ must be a non-minimal diagnosis ($\mathsf{nd} \subset \mathsf{node}$) or a duplicate diagnosis ($\mathsf{nd} = \mathsf{node}$) w.r.t.\ $\langle\mo,\mb,\Tp,\Tn\rangle_\RQ$ since all sets $\md_{\times}$, $\mD_{\checkmark}$ and $\mD_{calc}$ contain only minimal diagnoses w.r.t.\ $\langle\mo,\mb,\Tp,\Tn\rangle_\RQ$. In this case, the branch in the wpHS-tree corresponding to $\mathsf{node}$ can be dismissed which is taken account of by returning the label $closed$ for $\mathsf{node}$. 

In case the non-minimality criterion is not satisfied, the \emph{duplicate criterion} (lines~\ref{algoline:slabel:dup_crit_start}-\ref{algoline:slabel:dup_crit_end}) is checked next. Here, $\Queue$ is browsed for a node that is equal to $\mathsf{node}$. If such a one is found, $\mathsf{node}$ can be discarded because it suffices to consider only one tree branch among multiple tree branches in the wpHS-tree featuring one and the same set of edge labels. Hence, $closed$ is returned as a label for $\mathsf{node}$. Altogether, this means that only the last processed exemplar of a node corresponding to one and the same set of edge labels is labeled, all others are discarded.

If the duplicate criterion is not met, the \emph{reuse criterion} (lines~\ref{algoline:slabel:reuse_crit_start}-\ref{algoline:slabel:reuse_crit_end}) is checked next. That is, $\mC_{calc}$ is browsed for a set $\mc$ ($\mC_{calc}$ comprises only minimal conflict sets w.r.t.\ $\langle\mo,\mb,\Tp,\Tn\rangle_\RQ$) such that $\mc$ and $\mathsf{node}$ are disjoint sets. If such a $\mc$ is detected, then $\mc$ can be used to label $\mathsf{node}$ since the set of edge labels along the path in the wpHS-tree leading from the root node to $\mathsf{node}$ does not hit $\mc$. In this case, the label $\mc$ is returned for $\mathsf{node}$ by \textsc{sLabel}.

Given that the reuse criterion fails, $\scQX$ is called given the DPI $\langle\mo\setminus\mathsf{node},\mb,\Tp,\Tn\rangle_\RQ$ as an argument (line~\ref{algoline:slabel:qx}). If the output $L$ is equal to 'no conflict', then we know by Proposition~\ref{prop:qx_correctness} that $\mathsf{node}$ is a diagnosis w.r.t.\ $\langle\mo,\mb,\Tp,\Tn\rangle_\RQ$, wherefore the label $valid$ is returned for $\mathsf{node}$. Otherwise, the output $L$ must be a minimal conflict set w.r.t.\ $\langle\mo,\mb,\Tp,\Tn\rangle_\RQ$ that has an empty set-intersection with $\mathsf{node}$. Since the reuse criterion failed, i.e.\ there is no set in $\mC_{calc}$ that does not intersect with $\mathsf{node}$, $L$ must be a fresh minimal conflict set w.r.t.\ $\langle\mo,\mb,\Tp,\Tn\rangle_\RQ$ in the sense that $L \notin \mC_{calc}$ must hold. Therefore the label $L$ is first added to $\mC_{calc}$ and then returned by \textsc{sLabel} as a label for $\mathsf{node}$. 

\paragraph{Processing of a Node Label.} Back in the main procedure, $\mC_{calc}$ is updated (line~\ref{algoline:static:update_Ccalc}) and then the label $L$ returned by the \textsc{sLabel} function is processed as follows. If $L = valid$, then it is a fact that $\mathsf{node}$ is a minimal diagnosis w.r.t.\ $\langle\mo,\mb,\Tp,\Tn\rangle_\RQ$, but it is not certain that $\mathsf{node}$ also meets all positive test cases $\Tp'$ and all negative test cases $\Tn'$ that have been specified and added to $\langle\mo,\mb,\Tp,\Tn\rangle_\RQ$ so far. Thus, according to Proposition~\ref{prop:dpi_update}, the validity of the KB $\mo \setminus \mathsf{node}$ w.r.t.\ $\langle\cdot,\mb,\Tp\cup\Tp',\Tn\cup\Tn'\rangle_\RQ$ must still be checked (line~\ref{algoline:static:isOntValid}). If successful, $\mathsf{node}$ is added to the set $\mD_{calc}$ of calculated minimal diagnoses w.r.t.\ the input DPI that comply with all answered queries so far. Otherwise, $\mathsf{node}$ is added to the set $\mD_{\times}$  of minimal diagnoses w.r.t.\ the input DPI that have been invalidated by some answered query. 

Roughly, the minimality of diagnoses added to $\mD_{calc}$ is assured by the pruning rule (lines~\ref{algoline:slabel:non_min_crit_start}-\ref{algoline:slabel:non_min_crit_end}) which eliminates non-minimal nodes and the fact that $p_{nodes}()$ sorts a node $\mathsf{nd}'$ corresponding to a superset of some node $\mathsf{nd}$ behind $\mathsf{nd}$ in $\Queue$. 

If, on the other hand, $L=closed$ is the label returned by $\textsc{sLabel}$, then $\mathsf{node}$ must simply be removed from $\Queue$ which has already been executed in line~\ref{algoline:static_update_Q}. Thence, no actions are necessary (cf.\ line~\ref{algoline:static:if_L_closed}).

In the third case, if a minimal conflict set $L$ is returned by \textsc{sLabel}, then $L$ is a label for $\mathsf{node}$ meaning that $|L|$ successor nodes of $\mathsf{node}$, namely a node $\mathsf{node}\cup\setof{e}$ for all elements $e \in L$, need to be added to $\Queue$ in sorted order using the function $p_{nodes}()$ (\textsc{insertSorted}, line~\ref{algoline:static:generate_nodes}).
%

\paragraph{Stop Criterion.} The first criterion causing \textsc{staticHS} to terminate is $\Queue = []$ which means that the complete wpHS-tree has been constructed and no further nodes can be labeled. In this case, $\mD_{calc} \cup \mD_{\checkmark}$ comprises all minimal diagnoses w.r.t.\ $\langle\mo,\mb,\Tp,\Tn\rangle_\RQ$ that are compliant with all the specified positive and negative test cases $\Tp'$ and $\Tn'$.

If the first criterion is not met, then the second criterion is checked. That is, a test is performed which checks whether the number of leading minimal diagnoses w.r.t.\ $\langle\mo,\mb,\Tp,\Tn\rangle_\RQ$ in $\mD_{calc} \cup \mD_{\checkmark}$ amounts to at least $n_{\min}$ and either $|\mD_{calc} \cup \mD_{\checkmark}| = n_{\max}$ or more than $t$ time has passed since the start of the execution of \textsc{staticHS}. In the latter case, $n_{\min} \leq |\mD_{calc} \cup \mD_{\checkmark}| < n_{\max}$ holds. In the former case, $|\mD_{calc} \cup \mD_{\checkmark}| = n_{\max}$ is satisfied.

\paragraph{Processing of the Leading Diagnoses Returned by \textsc{staticHS}.} When a call of \textsc{staticHS} in Algorithm~\ref{algo:inter_onto_debug} returns $\tuple{\mD_{calc} \cup \mD_{\checkmark},\Queue, \mathbf{C}_{calc}, \mD_{\times}}$, the set $\mD_{calc} \cup \mD_{\checkmark}$ is stored in the variable $\mD_{\checkmark}$ in Algorithm~\ref{algo:inter_onto_debug}. Between two successive calls of \textsc{staticHS} in Algorithm~\ref{algo:inter_onto_debug}, only this set $\mD_{\checkmark}$ as well as $\mD_{\times}$ are modified. The list $\Queue$ and the set $\mC_{calc}$ remain unchanged until they are used as input parameters to the next call of \textsc{staticHS} in Algorithm~\ref{algo:inter_onto_debug}.

In case one diagnosis $\md_{\max}$ of the current leading diagnoses in $\mD_{\checkmark}$ has a probability greater or equal $1 - \sigma$ as per the probability measure $p_{\mD}()$ (see Section~\ref{sec:DetailedAlgorithmDescription}), the stop criterion of interactive KB debugging is met and a solution KB w.r.t.\ $\langle\mo,\mb,\Tp,\Tn\rangle_\RQ$ constructed from the input DPI $\langle\mo,\mb,\Tp,\Tn\rangle_\RQ$ as well as from $\md_{\max}$ is returned to the user. Thereafter, Algorithm~\ref{algo:inter_onto_debug} terminates and no more calls of \textsc{staticHS} take place.

Otherwise, if no leading diagnosis satisfies the stop criterion, a query $Q$ together with its q-partition $\Pt(Q)$ is computed as has been detailed in Sections~\ref{sec:QueryGeneration} and \ref{sec:DetailedAlgorithmDescription}. An answer $u(Q)$ to this query is submitted by the interacting user (line~\ref{algoline:inter_onto_debug:user_interaction} in Algorithm~\ref{algo:inter_onto_debug}). Then $u(Q)$ along with $\Pt(Q)$ is exploited to figure out the subset $\mD_{out}$ of $\mD_{\checkmark}$ that does not comply with $u(Q)$. This set $\mD_{out}$ is then deleted from $\mD_{\checkmark}$ and added to $\mD_{\times}$. Additionally, $Q$ is added to the positive test cases $\Tp'$ if $u(Q) = \true$ and to the negative test cases $\Tn'$ otherwise. Subsequently, \textsc{staticHS} is called again given 
\begin{itemize}
	\item the updated parameters $\mD_{\checkmark}$, $\mD_{\times}$, $\Tp'$ and $\Tn'$ (which are modified within and outside of \textsc{staticHS} during the execution of Algorithm~\ref{algo:inter_onto_debug}),
	\item the unchanged parameters $\Queue$, $\mC_{calc}$ (which are modified only within \textsc{staticHS} during the execution of Algorithm~\ref{algo:inter_onto_debug}) and
	\item the constant parameters $\langle\mo,\mb,\Tp,\Tn\rangle_\RQ$, $t$, $n_{\min}$, $n_{\max}$ and $p_{\mo}()$ (which are not modified within or outside of \textsc{staticHS} during the execution of Algorithm~\ref{algo:inter_onto_debug}).
\end{itemize}
The execution of this next and any subsequent call to \textsc{staticHS} runs in analogue way as described.  

\begin{remark}\label{rem:staticHS_query_computed_from_P_cup_P'_and_N_cup_N'}
We want to emphasize that queries are computed w.r.t.\ the current DPI $\langle\mo,\mb,\Tp\cup\Tp',\Tn\cup\Tn'\rangle_\RQ$ although \textsc{staticHS} focuses on solutions to the problem of Interactive Static KB Debugging which involves exclusively minimal diagnoses w.r.t.\ the input DPI $\langle\mo,\mb,\Tp,\Tn\rangle_\RQ$. However, a minimal diagnosis w.r.t.\ $\langle\mo,\mb,\Tp,\Tn\rangle_\RQ$ that satisfies all positive test cases $\Tp'$ as well as all negative test cases $\Tn'$ is also a minimal diagnosis w.r.t.\ $\langle\mo,\mb,\Tp\cup\Tp',\Tn\cup\Tn'\rangle_\RQ$. And, a minimal diagnosis w.r.t.\ $\langle\mo,\mb,\Tp,\Tn\rangle_\RQ$ that does not satisfy all positive test cases $\Tp'$ as well as all negative test cases $\Tn'$ is not a minimal diagnosis w.r.t.\ $\langle\mo,\mb,\Tp\cup\Tp',\Tn\cup\Tn'\rangle_\RQ$. 

Hence, it holds that
\begin{itemize}
\item $\md$ is a minimal diagnosis w.r.t.\ $\langle\mo,\mb,\Tp,\Tn\rangle_\RQ$ that satisfies $\Tp'\cup\setof{Q}$ as well as $\Tn'$ if and only if $\md$ is a minimal diagnosis w.r.t.\ $\langle\mo,\mb,\Tp\cup\Tp'\cup\setof{Q},\Tn\cup\Tn'\rangle_\RQ$ and
\item $\md$ is a minimal diagnosis w.r.t.\ $\langle\mo,\mb,\Tp,\Tn\rangle_\RQ$ that satisfies $\Tp'$ as well as $\Tn'\cup\setof{Q}$ if and only if $\md$ is a minimal diagnosis w.r.t.\ $\langle\mo,\mb,\Tp\cup\Tp',\Tn\cup\Tn'\cup\setof{Q}\rangle_\RQ$.
\end{itemize}
Therefore, each query constructed during Algorithm~\ref{algo:inter_onto_debug} with $mode = static$ must be a query w.r.t.\ the current set of leading diagnoses $\mD_{\checkmark}$ and the \emph{current} DPI $\langle\mo,\mb,\Tp\cup\Tp',\Tn\cup\Tn'\rangle_\RQ$ (cf.\ Equation~\ref{eq:sol_ont_candidate}, Definition~\ref{def:q-partition} and Proposition~\ref{prop:dpi_update} on pages~\pageref{eq:sol_ont_candidate}-\pageref{prop:dpi_update}).

As a consequence of this, no additional test is required in order to ascertain that each diagnosis in the set $\mD_{\checkmark}$ that is given as a parameter to the next call of \textsc{staticHS} does in fact satisfy all answered queries so far.\qed
\end{remark} 
The following proposition states the correctness of \textsc{staticHS} (a proof of this proposition is beyond the scope of this work):
\begin{proposition}[Correctness of \textsc{staticHS}]\label{prop:static_hs_correctness}
Any call to \textsc{staticHS} (given the inputs described in Algorithm~\ref{algo:inter_stat_hs}) within Algorithm~\ref{algo:inter_onto_debug} terminates and yields an output $\tuple{\mD,\Queue,\mathbf{C}_{calc},\mD_{\times}}$ where 
\begin{enumerate}[(1)]
\item it holds for $\mD$ that 
\begin{enumerate}[(a)]
\item $\mD \subseteq \minD_{\langle\mo,\mb,\Tp,\Tn\rangle_\RQ} \cap \minD_{\langle\mo,\mb,\Tp\cup\Tp',\Tn\cup\Tn'\rangle_\RQ}$ is the set of most probable 
 minimal diagnoses w.r.t.\ $\langle\mo,\mb,\Tp,\Tn\rangle_\RQ$ that satisfy all test cases $\Tp'$ and $\Tn'$ such that 
\begin{enumerate}[(i)]
\item $n_{\min} \leq |\mD| \leq n_{\max}$ and 
\item $\mD\supset\mD_{\checkmark}$,
\end{enumerate} 
if such a set $\mD$ exists, or 
\item $\mD$ is equal to the set of all minimal diagnoses $\minD_{\langle\mo,\mb,\Tp,\Tn\rangle_\RQ} \cap \minD_{\langle\mo,\mb,\Tp\cup\Tp',\Tn\cup\Tn'\rangle_\RQ}$, otherwise, 
\end{enumerate}
where ``most-probable'' refers to the probability measure $p_{nodes}()$ (cf.\ Definition~\ref{def:p_node()}) obtained from the given function $p()$;
\item $\Queue$ is the current queue of open (non-labeled) nodes of the produced (partial) wpHS-tree, 
\item $\mC_{calc}$ is the set of all minimal conflict sets w.r.t.\ $\langle\mo,\mb,\Tp,\Tn\rangle_\RQ$ computed so far and 
\item $\mD_{\times}$ is the set of all minimal diagnoses w.r.t.\ $\langle\mo,\mb,\Tp,\Tn\rangle_\RQ$ computed so far where each diagnosis in $\mD_{\times}$ does not satisfy all test cases $\Tp'$ and $\Tn'$.
\qed
\end{enumerate} 
\end{proposition}

\subsection[Examples]{\textsc{staticHS}: Examples}
\label{sec:TextscStaticHSExamples}
In this section we will give two examples of how interactive KB debugging using \textsc{staticHS} (Algorithm~\ref{algo:inter_onto_debug} with parameter $mode=static$) works. The first one will show the similarities and differences between the usage of \textsc{staticHS} (within Algorithm~\ref{algo:inter_onto_debug}) and \textsc{HS} (within Algorithm~\ref{algo:non_int_debug}) since it will depict the application of \textsc{staticHS} on the same example DPI (see Table~\ref{tab:example2}) that was used to show the functionality of \textsc{HS} in examples~\ref{example:non_interactive_debugging_with_tabExDpi2_and_without_probs} and \ref{example:non_interactive_debugging_with_tabExDpi2_and_probs}. At the same time, the first example will provide evidence that solving the problem of Interactive Static KB Debugging can be more efficient than solving the problem of Interactive Dynamic KB Debugging in terms of the number of query answers required from an interacting user. This will be discussed in more detail in Section~\ref{sec:TextscStaticHSVersusTextscDynamicHS}.

The second example is supposed to deepen the reader's understanding of the way \textsc{staticHS} works. To this end, the example DPI provided by Table~\ref{tab:example3} will be used which constitutes a significantly harder (interactive) debugging task than the DPI investigated in the first example. This example will involve the construction of a relatively large hitting set tree and thereby give a presentiment of the space and time complexity problems caused by the poor tree pruning inherent in the \textsc{staticHS} algorithm. In addition, this example will draw a reverse image of the first example in that it will stress the advantage of the decision to search for a solution of Interactive Dynamic KB Debugging rather than for a solution of Interactive Static KB Debugging (more on that in Section~\ref{sec:TextscStaticHSVersusTextscDynamicHS}).

\begin{example}\label{example:staticHS_simple_example_using_tabExDpi2}
In this example we assume that the author (called user throughout this example) of the (admissible) DPI $\langle\mo,\mb,\Tp,\Tn\rangle_\RQ$ given by Table~\ref{tab:example2} applies Algorithm~\ref{algo:inter_onto_debug} with $mode = static$ to interactively debug $\langle\mo,\mb,\Tp,\Tn\rangle_\RQ$. Further, suppose the following user requirements:

In order to guarantee a fast reaction time of the system (the time between two successive queries to the user), the user wants each query to be computed from the minimally necessary number of leading diagnoses. Thus, in each iteration exactly two leading diagnoses should be computed by \textsc{staticHS} (cf.\ Proposition~\ref{prop:q1}). This postulation is reflected by setting $n_{\min} = n_{\max} = 2$. Notice that the time limit $t$ is irrelevant in this case. 

Moreover, the user desires to get just any query, i.e.\ they do not demand any particular properties -- such as optimal information gain among a pool of queries -- to be satisfied by a query. This can be ensured by choosing $q := 1$ (cf.\ Section~\ref{sec:QueryGeneration}) and $qsm()$ equal to any query selection measure described in Section~\ref{sec:query_selection_measures}. 

The user is new to KB debugging and has neither an idea of faults they frequently make nor access to any kind of data that would indicate their tendency to certain types of faults. Thence, $p_{\mo}(\tax) := c < 0.5$ for all $\tax \in \mo$, i.e.\ all formula fault probabilities are specified to be equal (to some constant $c$). In such a case, if a formula fault probability measure $p_{\mo}()$ is given as an input to Algorithm~\ref{algo:inter_onto_debug}, then line~\ref{algoline:inter_onto_debug:getAxiomProbs} in Algorithm~\ref{algo:inter_onto_debug} is omitted. Please notice that this aspect is not shown in Algorithm~\ref{algo:inter_onto_debug}.

Finally, the user's intention is to get the (exact) solution to the problem of Interactive Static KB Debugging. This can be taken into account by specifying $\sigma := 0$.

The tree constructed and parameters computed and used by Algorithm~\ref{algo:inter_onto_debug} using \textsc{staticHS} are visualized by Figure~\ref{fig:example:inter_onto_debug_staticHS_TabExDpi2}. 
We use the same notation as in Figures~\ref{fig:example:non-interactive_onto_debug_auto=false+nmin=infty_and_auto=true} and \ref{fig:example:non-interactive_onto_debug_auto=false+nmin=2+nmax=4_with_probs} which is described in Examples~\ref{example:non_interactive_debugging_with_tabExDpi2_and_without_probs} and \ref{example:non_interactive_debugging_with_tabExDpi2_and_probs}. The only new notational element here is the $\Longrightarrow$ labeled by some designator of a query. That is, $\checkmark_{(\md_i)} \stackrel{Q_j}{\Longrightarrow} \checkmark$ means that $\md_i$ is still a minimal diagnosis after $Q_j$ has been answered and added to the respective set of test cases of the DPI. On the other hand, $\checkmark_{(\md_i)} \stackrel{Q_j}{\Longrightarrow} \times$ signifies that the minimal diagnosis $\md_i$ is invalidated through the addition of the answered query $Q_j$ to the respective set of test cases of the DPI. Please notice that $\Longrightarrow$ does not point at a node of the wpHS-tree. Instead, the label at which $\Longrightarrow$ points is to be understood as the new label of the node originally labeled by $\checkmark_{(\md_i)}$ from which the (first of possibly multiple) $\Longrightarrow$ goes out. This notation should help to keep track of the evolution of node labels in the wpHS-tree without needing to overload a single node by multiple different successive labels. 

In the first iteration, i.e.\ during the execution of the first call of \textsc{staticHS} during Algorithm~\ref{algo:inter_onto_debug}, the root node (initially the empty set) is labeled by the minimal conflict set $\tuple{1,2,5}$ w.r.t.\ $\langle\mo,\mb,\Tp,\Tn\rangle_\RQ$ and three successor nodes, namely $\setof{1}$, $\setof{2}$ as well as $\setof{5}$, are added to the queue of open nodes $\Queue$. Since all formulas have been assigned an equal fault probability, \textsc{staticHS} conducts a breadth-first tree construction (as displayed by the numbers \textcircled{\scriptsize i} that give the order of node labeling). That is, $\Queue$ in this case is a first-in-first-out queue. In this vein, first $[1]$ and then $[2]$ are identified as minimal diagnoses w.r.t.\ the given DPI. Since $\mD_{\checkmark}\cup\mD_{calc} = \emptyset \cup \setof{[1],[2]}$ has a cardinality of $n_{\min} = n_{\max} = 2$, the stop criterion of \textsc{staticHS} causes it to terminate and return $\tuple{\mD_{calc}\cup\mD_{\checkmark},\mC_{calc},\Queue,\mD_{\times}} = \tuple{\setof{[1],[2]},\setof{\tuple{1,2,5}},[\setof{5}],\emptyset}$ (because $\mD_{\checkmark}$ and $\mD_{\times}$ are initially empty sets), as shown in the upper right column in Figure~\ref{fig:example:inter_onto_debug_staticHS_TabExDpi2}. 

Then, in Algorithm~\ref{algo:inter_onto_debug}, outside of the \textsc{staticHS} procedure, the first query $Q_1 = \setof{E \rightarrow \lnot A}$ is computed from the leading diagnoses set $\setof{[1],[2]}$. The q-partition $\Pt(Q_1)$ associated with $Q_1$ is $\langle\setof{[1]},\setof{[2]}$, $\emptyset\rangle$. The user's answer $u(Q_1)$ to $Q_1$ is then $\false$. Thence, the set $\mD_{out}$ is calculated from $\Pt(Q_1)$ as $\dx{}(Q_1) = \setof{[1]}$ (due to negative answer, cf.\ Remark~\ref{rem:invalidated_sets_of_q-partition_for_query_answer}), deleted from $\mD_{\checkmark} := \mD_{\checkmark} \cup \mD_{calc}$ to yield $\mD_{\checkmark} = \setof{[2]}$ and added to $\mD_{\times}$ to yield $\mD_{\times} = \setof{[1]}$.
The set $\mD_{\checkmark}$ corresponds to the set of all already computed minimal diagnoses w.r.t.\ the input DPI that satisfy all queries answered so far. The set $\mD_{\times}$ comprises all already computed minimal diagnoses w.r.t.\ the input DPI that do not satisfy all queries answered so far. 
These sets $\mD_{\checkmark}$ and $\mD_{\times}$ along with the collections $\Queue$ and $\mC_{calc}$ which are unmodified outside of \textsc{staticHS} are used as input arguments for the second call of \textsc{staticHS}. Notice that, in the figure, the resulting values of operations performed within \textsc{staticHS} are given in the righthand column above the dashed line whereas values computed outside of \textsc{staticHS} are given below the dashed line. 

After the modifications caused by the addition of the query $Q_1$ to the negative test cases of $\langle\mo,\mb,\Tp$, $\Tn\rangle_\RQ$ have been taken into account in step \textcircled{\scriptsize 4}, 
the partial wpHS-tree built in iteration 1 is further constructed in iteration 2 resulting in the tree depicted by the middle picture in the lefthand column of Figure~\ref{fig:example:inter_onto_debug_staticHS_TabExDpi2}. Whereas the branches with edge labels $\setof{5,1}$ and $\setof{5,2}$ correspond to proper supersets of the minimal diagnoses $[1]$ and $[2]$, respectively, w.r.t.\ the input DPI $\tuple{\mo,\mb,\Tp,\Tn}_\RQ$ and are thus closed by the non-minimality criterion tested in the \textsc{sLabel} function, the branch with edge labels $\setof{5,7}$ is identified as a minimal diagnosis $\md_3 := [5,7]$ w.r.t.\ $\tuple{\mo,\mb,\Tp,\Tn}_\RQ$. However, $\md_3$ is not directly added to the set $\mD_{calc}$. In fact, the validity of the KB $\mo \setminus \md_3$ w.r.t.\ the \emph{current} DPI $\tuple{\mo,\mb,\Tp,\Tn\cup\setof{Q_1}}_\RQ$ is tested beforehand. As this test is successful, meaning that $\md_3 \in \minD_{\tuple{\mo,\mb,\Tp,\Tn}_\RQ} \cap \minD_{\tuple{\mo,\mb,\Tp,\Tn\cup\setof{Q_1}}_\RQ}$, $\md_3$ can be safely added to $\mD_{calc}$ implying the set of leading diagnoses $\mD_{\checkmark}\cup\mD_{calc} = \setof{\md_2,\md_3}$ with cardinality two. Due to $n_{\min} = n_{\max} = 2$, \textsc{staticHS} terminates.

After the second query $Q_2$ has been answered negatively involving the dismissal of the leading diagnosis $\md_2$, \textsc{staticHS} ends up with an empty queue $\Queue$ of open nodes in iteration 3 (see the tree in the lower left column of Figure~\ref{fig:example:inter_onto_debug_staticHS_TabExDpi2}). Hence, \textsc{staticHS} returns a singleton set including the leading diagnosis $\md_3$. Now, independently of the specified formula probabilities, $p_{\mD}(\md_3) = 1 \geq 1 - \sigma = 1$ is satisfied since the probability space considered by the probability measure $p_{\mD}()$ focuses on the sample space $\Omega = \setof{\md_3}$ (cf.\ Sections~\ref{sec:DiagnosisProbabilitySpace} and \ref{sec:DetailedAlgorithmDescription}). Thus, the stop condition of Algorithm~\ref{algo:inter_onto_debug} is met wherefore the solution KB $\mo_{sol} := (\mo \setminus \md_3) \cup U_{\Tp} = (\mo \setminus \md_3) \cup \emptyset = \mo \setminus \md_3$ is returned to the user. This solution KB $\mo_{sol}$ is the (exact) solution to Interactive Static KB Debugging given the DPI $\tuple{\mo,\mb,\Tp,\Tn}_\RQ$ of Table~\ref{tab:example2} as an input because $\md_3$ is the only minimal diagnosis w.r.t.\ $\tuple{\mo,\mb,\Tp,\Tn}_\RQ$ that conforms with all answered queries $Q_1 = \false$ and $Q_2 = \false$.

All in all, the execution of Algorithm~\ref{algo:inter_onto_debug} in this example performs 
\begin{itemize}
	\item 2 full $\scQX$ calls, i.e.\ calls of $\scQX$ that actually return a minimal conflict set (there are two minimal conflict sets labeled by $C$ in the picture at the bottom of the lefthand column in Figure~\ref{fig:example:inter_onto_debug_staticHS_TabExDpi2}) and
	\item 6 validity checks, i.e.\ calls of $\scQX$ that return 'no conflict' (one check for each of the three found minimal diagnoses; notice that $\scQX$ does only perform a single KB validity check by \textsc{isKBValid} in case it returns 'no conflict', see Algorithm~\ref{algo:qx}) or calls of \textsc{isKBValid} in line~\ref{algoline:static:isOntValid} in \textsc{staticHS} (one call for each of the three found minimal diagnoses),
\end{itemize}
computes
\begin{itemize}
	\item 3 minimal diagnoses w.r.t.\ the input DPI,
	\item 2 minimal conflict sets w.r.t.\ the input DPI and
	\item 2 queries and asks the user 2 logical formulas (1 per query)  
\end{itemize}
and stores
\begin{itemize}
	\item a maximum of 5 nodes (where node refers to the internal representation of a node in \textsc{staticHS} as a set of edge labels along a path from the root node to a leaf node; there are even more nodes in the sense of tree nodes in the picture at the bottom of the lefthand column in Figure~\ref{fig:example:inter_onto_debug_staticHS_TabExDpi2}).\qed
\end{itemize}
\end{example}

\begin{example}\label{example:staticHS_complex_example_using_tabExDpi3}
Let us now consider the (admissible) DPI $\langle\mo,\mb,\Tp,\Tn\rangle_\RQ$ given by Table~\ref{tab:example3}. We assume an expert (called user throughout this example) in the domain $Dom$ modeled by $\mo$ who wants to find a solution to Interactive Static KB Debugging for the given DPI $\langle\mo,\mb,\Tp,\Tn\rangle_\RQ$ by means of Algorithm~\ref{algo:inter_onto_debug} with $mode = static$. 
Further, we suppose the following requirements:

The user wants each query to be computed from three leading diagnoses. Thus, after each iteration of \textsc{staticHS}, the set $\mD_{\checkmark} \cup \mD_{calc}$ should comprise exactly three elements. This postulation is reflected by setting $n_{\min} = n_{\max} = 3$. Notice that the time limit $t$ is irrelevant in this case. 

Moreover, as in example~\ref{example:staticHS_simple_example_using_tabExDpi2}, we assume no demand for queries satisfying special properties which is reflected by choosing $q := 1$ (cf.\ Section~\ref{sec:QueryGeneration}) and $qsm()$ equal to any query selection measure described in Section~\ref{sec:query_selection_measures}.

Let there be several documentations of past debugging sessions (e.g.\ in terms of formula change logs) involving KBs in the domain $Dom$ of the author $auth$ of $\mo$ accessible to the user. Further, let the user have extracted term and logical construct probabilities $p_{\widetilde{\mo}\cup\overline{\mo}}(\tax)\in[0,1]$ for $\tax\in\mo$ for $auth$ from this data. This function $p_{\widetilde{\mo}\cup\overline{\mo}}: \widetilde{\mo}\cup\overline{\mo} \rightarrow [0,1]$ is then provided as an input to Algorithm~\ref{algo:inter_onto_debug}. 

Finally, the user's intention is to get the (exact) solution to the problem of Interactive Static KB Debugging. This can be taken into account by specifying $\sigma := 0$.

The tree constructed and parameters computed and used by Algorithm~\ref{algo:inter_onto_debug} using \textsc{staticHS} are visualized by Figures~\ref{fig:example:inter_onto_debug_staticHS_TabExDpi3} as well as \ref{fig:example:inter_onto_debug_staticHS_TabExDpi3_continued}. 
We use the same notation as in Figures~\ref{fig:example:non-interactive_onto_debug_auto=false+nmin=infty_and_auto=true}, \ref{fig:example:non-interactive_onto_debug_auto=false+nmin=2+nmax=4_with_probs} and \ref{fig:example:inter_onto_debug_staticHS_TabExDpi2} which is described in Examples~\ref{example:non_interactive_debugging_with_tabExDpi2_and_without_probs}, \ref{example:non_interactive_debugging_with_tabExDpi2_and_probs} and \ref{example:staticHS_simple_example_using_tabExDpi2}. 

After the initialization of variables, Algorithm~\ref{algo:inter_onto_debug} calls the function \textsc{getFormulaProbs} in line~\ref{algoline:inter_onto_debug:getAxiomProbs} which exploits $p_{\widetilde{\mo}\cup\overline{\mo}}()$ to calculate the function $p_{\mo}()$ giving the fault probabilities of formulas in $\mo$ (cf.\ Sections~\ref{sec:prob_space_construction}, \ref{sec:DetailedAlgorithmDescription} and Example~\ref{example:ax_prob_calc}). Let the resulting probabilities be as depicted by Table~\ref{tab:example:staticHS_complex--->axiom_probs}.
\begin{table}[tb]
	\centering
		\begin{tabular}{lccccccc}
			$\tax \in \mo$ & 1 & 2 & 3 & 4 & 5 & 6 & 8 \\\hline
			$p_{\mo}(\tax)$ & 0.26 & 0.18 & 0.21 & 0.41 & 0.18 & 0.40 & 0.18 
		\end{tabular}
\caption[(Example~\ref{example:staticHS_complex_example_using_tabExDpi3}) Formula Fault Probabilities]{(Example~\ref{example:staticHS_complex_example_using_tabExDpi3}) Computed formula fault probabilities for the example DPI given by Table~\ref{tab:example3}.}
\label{tab:example:staticHS_complex--->axiom_probs}
\end{table}

Then, \textsc{staticHS} is called for the first time, resulting in the wpHS-tree given in the first picture in Figure~\ref{fig:example:inter_onto_debug_staticHS_TabExDpi3}. Contrary to Example~\ref{example:staticHS_simple_example_using_tabExDpi2}, where the tree was built up in breadth-first order, in this example the formula probabilities $p() := p_{\mo}()$ given by Table~\ref{tab:example:staticHS_complex--->axiom_probs} are used to assign a probability $p_{nodes}(\mathsf{n})$ to each path $\mathsf{n}$ in the wpHS-tree starting from the root node (cf.\ Formula~\ref{eq:path_prob_calc} and Definition~\ref{def:p_node()}). In this vein, as outlined by the numbers \textcircled{\scriptsize i} indicating when a node is labeled, after the root node has been labeled by $\mc_1 := \tuple{1,2,5}$, the node corresponding to the outgoing edge of $\mc_1$ labeled by the formula with the largest fault probability among all formulas in $\mc_1$ is labeled first. That is, the node $\setof{1}$ with $p_{nodes}(\setof{1}) = 0.41$ (as opposed to the nodes $\setof{2}$ and $\setof{5}$ with $0.25$ each) is labeled first. The \textsc{sLabel} procedure, after checking whether $\setof{1}$ is a non-minimal diagnosis w.r.t.\ $\langle\mo,\mb,\Tp,\Tn\rangle_\RQ$ or a duplicate of some other node in $\Queue$ (both checks negative), computes another minimal conflict set $\mc_2 := \tuple{2,4,6}$ such that $\setof{1}\cap\mc_2 = \emptyset$ ($\mc_2$ is not hit by the node $\setof{1}$) to constitute a label for node $\setof{1}$. The successor nodes $\setof{1,2}$, $\setof{1,4}$ and $\setof{1,6}$ of $\setof{1}$ are generated and added to the list $\Queue$ in a way that the sorting of $\Queue$ in descending order of $p_{nodes}()$ is maintained.  

Since $\setof{1,4}$ (0.28) as well as $\setof{1,6}$ (0.27) have a larger probability (as per $p_{nodes}()$) than the nodes $\setof{2}$ (0.25) and $\setof{5}$ (0.25), $\Queue$ is given by $[\setof{1,4},\setof{1,6},\setof{2},\setof{5},\setof{1,2}]$ when it comes to the processing of the next node. Since \textsc{staticHS} always treats the first node of $\Queue$ next, it identifies the first minimal diagnosis $\md_1 := [1,4]$ w.r.t.\ $\langle\mo,\mb,\Tp,\Tn\rangle_\RQ$ in step \textcircled{\scriptsize 3}. In steps \textcircled{\scriptsize 4} and \textcircled{\scriptsize 8}, two further minimal diagnoses $\md_2 := [1,6]$ and $\md_3 := [5,4]$ are detected. Altogether, the union of $\mD_{\checkmark}$ (initially the empty set) and $\mD_{calc}$ (comprising the three computed diagnoses) now contains $3 = n_{\min} = n_{\max}$ elements wherefore \textsc{staticHS} terminates and outputs the tuple $\tuple{\mD_{calc}\cup\mD_{\checkmark},\mC_{calc},\Queue,\mD_{\times}}$ where the sets in this tuple are given under the wpHS-tree of iteration 1 in Figure~\ref{fig:example:inter_onto_debug_staticHS_TabExDpi3}.

From this set of leading diagnoses $\mD_{\checkmark} := \mD_{\checkmark} \cup \mD_{calc}$, the probability measure $p_{\mD}: \mD_{\checkmark} \rightarrow [0,1]$ is computed by the function \textsc{getProbDist} (cf.\ Algorithm~\ref{algo:inter_onto_debug_continued} and Section~\ref{sec:DetailedAlgorithmDescription}). The result is $\tuple{p_{\mD}(\md_1),p_{\mD}(\md_2),p_{\mD}(\md_3)} = \tuple{0.38,0.37,0.25}$. The mode $\md_{\max} := \md_1$ of this probability distribution is then computed by \textsc{getMode}. As $\sigma = 0$, $p_{\mD}(\md_{\max}) = 0.38 \not\geq 1$ wherefore the stop criterion of Algorithm~\ref{algo:inter_onto_debug} is not satisfied. 

Consequently, Algorithm~\ref{algo:inter_onto_debug} proceeds to generate the first query $Q_1 = \setof{B \sqsubseteq K}$ (based on the current set of leading diagnoses $\mD_{\checkmark}$) along with its associated q-partition $\Pt(Q_1) = \tuple{\setof{\md_1,\md_2},\setof{\md_3},\emptyset}$. The diagnosis $\md_1$ is in $\dx{}(Q_1)$ because $\ot_1 = (\mo\setminus\md_1) \cup \mb \cup U_{\Tp}$ (recall Formula~\ref{eq:sol_ont_candidate} for a definition of $\ot_i$) comprises formulas 2, 3, 5, 6, 7, 8 and 9 as well as $\tp_1$ (cf.\ Table~\ref{tab:example3}) wherefore $\ot_1 \models \setof{B \sqsubseteq K} = Q_1$ (due to the set of formulas $\setof{2,3} = \setof{B\sqsubseteq G, G \sqsubseteq K}$). That $\md_2$ belongs to $\dx{}(Q_1)$ as well follows analogously. On the other hand, $\md_3 \in \dnx{}(Q_1)$ must be true since $\ot_3 \cup Q_1$ includes i.a.\ $A \sqsubseteq B$ (formula 1) and $B \sqsubseteq K$ ($\in Q_1$) wherefore $\setof{A \sqsubseteq K} = \tn_1$ is an entailment of $\ot_3$. Thus, the negative test case $\tn_1$ is violated.    

The positive user answer $u(Q_1) = \true$ is incorporated in that $Q_1$ is appended to the set of positive test cases $\Tp$ yielding $\Tp \cup \setof{Q_1} = \setof{\setof{r(x,y)},\setof{B \sqsubseteq K}}$. Step~\textcircled{\scriptsize 9} shows the impact of this test case addition on the set of leading diagnoses, i.e.\ all diagnoses in the set $\mD_{out} = \dnx{}(Q_1) = \setof{\md_3}$ (due to positive answer, cf.\ Remark~\ref{rem:invalidated_sets_of_q-partition_for_query_answer}) are re-labeled by $\times$ whereas all other leading diagnoses ($\md_1, \md_2$) are still labeled by $\checkmark$.

In the same fashion, further node labelings are conducted in iteration 2 until $|\mD_{\checkmark} \cup \mD_{calc}| = |\setof{\md_1, \md_2} \cup \setof{[2,1]}| = 3 = n_{\min} = n_{\max}$ holds again. These actions are displayed by the tree at the bottom of Figure~\ref{fig:example:inter_onto_debug_staticHS_TabExDpi3}. 

Notice that, after step~\textcircled{\tiny 12}, two nodes corresponding to the same set are elements of the list $\Queue$. At step~\textcircled{\tiny 13}, the duplicate criterion checked by \textsc{sLabel} comes into play. Since the node $\setof{1,2}$ (the leftmost branch in the tree) is ranked first in $\Queue$ (we assume a first-in-first-out ordering of nodes corresponding to equal sets of edge labels in $\Queue$), the \textsc{sLabel} procedure is called given $\mathsf{node} := \setof{1,2}$ as an argument and detects the node $\setof{2,1}$ (the fourth leftmost branch in the tree) in $\Queue$. Hence, $\mathsf{node} = \setof{1,2}$ is closed as a duplicate node which finds expression in the label $\times_{(dup)}$. When $\setof{2,1}$ (which must have the same probability as $\setof{1,2}$ due to set-equality) is processed at step~\textcircled{\tiny 14}, it is discovered to be a minimal diagnosis ($\md_5$) w.r.t.\ $\langle\mo,\mb,\Tp,\Tn\rangle_\RQ$.

Moreover, we want to point out that another minimal diagnosis ($\md_4 = [2,4,6]$) is found in iteration 2 before $\md_5$ is detected. However, $\md_4$ is immediately ruled out and added to $\mD_{\times}$ (cf.\ line~\ref{algoline:static:add_to_Dtimes} in \textsc{staticHS}) due to the fact that $\mo \setminus \md_4$ is invalid w.r.t.\ the \emph{current} DPI $\langle\cdot,\mb,\Tp\cup\setof{Q_1},\Tn\rangle_\RQ$ (cf.\ Definition~\ref{def:valid_onto}). The explanation why this holds is as follows: 

By Definition~\ref{def:valid_onto}, $\mo\setminus\md_4$ is valid w.r.t.\ $\langle\cdot,\mb,\Tp\cup\setof{Q_1},\Tn\rangle_\RQ$ iff $\ot_4 = (\mo \setminus \md_4) \cup \mb \cup U_{(\Tp\cup\setof{Q_1})}$ (recall Formula~\ref{eq:sol_ont_candidate} for a definition of $\ot_i$) does not violate any $r\in\RQ = \setof{\text{consistency},\text{coherency}}$ and does not entail any $\tn \in \Tn = \setof{\tn_1,\tn_2} = \setof{\setof{A \sqsubseteq K},\setof{L \sqsubseteq \exists r.F, B(x), G \sqsubseteq K}}$. Applying the diagnosis $\md_4$ to $\mo$ yields $\mo \setminus \md_4 = \setof{1,3,5,8}$ which includes in particular formula $1$ which is equal to $A \sqsubseteq B$ (see Table~\ref{tab:example3}). However, there is also the negative test case $\tn_1$ indicating that $A \sqsubseteq K$ must not be entailed by $\ot_4$. That is, $B \sqsubseteq K \in \ot_4$ (due to $Q_1$) and $A \sqsubseteq B \in \ot_4$ which implies that $\ot_4 \models \setof{A \sqsubseteq K} = \tn_1$ wherefore $\ot_4$ is invalid w.r.t.\ $\langle\cdot,\mb,\Tp\cup\setof{Q_1},\Tn\rangle_\RQ$.  

Such a direct dismissal of a discovered diagnosis $\md_i$ due to a newly added test case $Q_j$ is indicated by $\textcircled{\scriptsize k} \checkmark_{(\md_i)} \stackrel{Q_j}{\Longrightarrow}  \textcircled{\scriptsize k}\times$, i.e.\ the step number \textcircled{\scriptsize k} at the shaft of the $\Longrightarrow$ is equal to the step number at the head of $\Longrightarrow$. In case of the invalidation of a \emph{leading} diagnosis (i.e.\ one that was utilized in the computation of $Q_j$), on the contrary, the step number at the shaft is lower than the step number at the arrow head.

As shown at the top of Figure~\ref{fig:example:inter_onto_debug_staticHS_TabExDpi3_continued}, the second query $Q_2$ computed from the leading diagnosis set $\mD_{\checkmark} \cup \mD_{calc} = \setof{\md_1,\md_2,\md_5}$ is then answered by $u(Q_2) = \true$ as well, wherefore the leading diagnoses $\md_2, \md_5$ are ruled out and added to $\mD_{\times}$. So, the input argument $\mD_{\checkmark}$ given to the next call of \textsc{staticHS} in Algorithm~\ref{algo:inter_onto_debug} consists of the single diagnosis $\md_1$. 

In the third iteration (see the picture given in Figure~\ref{fig:example:inter_onto_debug_staticHS_TabExDpi3_continued}), \textsc{staticHS} again executes in order to complete the leading diagnosis set to contain three elements. However, as we can say in advance, $\md_1$ is the only minimal diagnosis w.r.t.\ the input DPI $\langle\mo,\mb,\Tp,\Tn\rangle_\RQ$ which is also a diagnosis w.r.t.\ the current DPI $\langle\mo,\mb,\Tp\cup\setof{Q_1,Q_2},\Tn\rangle_\RQ$. Nevertheless, \textsc{staticHS} continues expanding the wpHS-tree until it has verified that this is the case ($\Queue = []$). This is equivalent to finishing the construction of the non-interactive wpHS-tree that is generated by \textsc{HS} with parameters $n_{\min} = n_{\max} = \infty$. We want to stress that the construction of the entire wpHS-tree w.r.t.\ $\langle\mo,\mb,\Tp,\Tn\rangle_\RQ$ and $p() := p_{\mo}()$ is inevitable in a debugging scenario where the (exact) solution to the Interactive Static KB Debugging problem is sought (the probability w.r.t.\ $p_{\mD}()$ of a diagnosis can only be equal to 1 if there is only a single leading diagnosis returned by \textsc{staticHS}). 

In fact, there are five further diagnoses $\md_6,\dots,\md_{10}$ w.r.t.\ $\langle\mo,\mb,\Tp,\Tn\rangle_\RQ$ that are detected in iteration 3 and directly dismissed (added to $\mD_{\times}$) after the validity check in line~\ref{algoline:static:isOntValid} of \textsc{staticHS}. All other tree branches are closed due to the non-minimality (label $\times_{(\supset \md_i)}$) or duplicate criterion (label $\times_{(dup)}$).  
Due to $\sigma = 0$ and the associated necessity to grow the wpHS-tree until all leaf nodes are labeled, the final tree (19 labeled leaf nodes) depicted in Figure~\ref{fig:example:inter_onto_debug_staticHS_TabExDpi3_continued} is relatively large in comparison to the small size $|\mo| = 7$. 

This example might already give an idea of the potential explosion of the wpHS-tree produced by \textsc{staticHS} in case the (exact) solution to the Interactive Static KB Debugging problem is desired. This is why it will usually make sense in practice to specify a fault tolerance $\sigma > 0$ which enables Algorithm~\ref{algo:inter_onto_debug} with $mode = static$ to escape from the generally intractable complexity of the complete investigation of all minimal diagnoses w.r.t.\ the input DPI (full construction of the wpHS-tree). However, in this concrete example, allowing a small fault tolerance $\sigma$ has no effect either. Actually, $\sigma \geq 0.56$ is necessary to achieve a premature termination of the tree construction. This holds due to the fact that the probability distributions of leading diagnoses are $\tuple{p_{\mD}(\md_1),p_{\mD}(\md_2),p_{\mD}(\md_3)} = \tuple{0.38,0.37,0.25}$ (after iteration 1) and $\tuple{p_{\mD}(\md_1),p_{\mD}(\md_2),p_{\mD}(\md_5)} = \tuple{0.44,0.42,0.14}$ (after iteration 2). Now, given say $\sigma := 0.6$, the stop criterion of Algorithm~\ref{algo:inter_onto_debug} would be met after iteration 2 because $p_{\mD}(\md_{\max}) = p_{\mD}(\md_1) = 0.44 \geq 0.4 = 1 - 0.6 = 1-\sigma$. Nate that, in this case, the same (exact) solution would be returned as for the setting $\sigma := 0$. The (significant) difference is just that the final tree in this case has only 14 leaf nodes, of which only 7 are labeled (the labeling of a node is in general significantly more costly than the mere generation of a node). As opposed to this, the full tree comprises 19 labeled nodes. On the other side of the coin, choosing a value of $\sigma > 0.5$, for example, means that -- from the point of view of the knowledge at the time Algorithm~\ref{algo:inter_onto_debug} terminates -- a solution to Interactive Static KB Debugging is returned by Algorithm~\ref{algo:inter_onto_debug} which has a higher probability of not being the (exact) solution than of being the (exact) solution.

All in all, the execution of Algorithm~\ref{algo:inter_onto_debug} in this example performs 
\begin{itemize}
	\item 4 full $\scQX$ calls, i.e.\ calls of $\scQX$ that actually return a minimal conflict set (there are four minimal conflict sets labeled by $C$ in the tree in Figure~\ref{fig:example:inter_onto_debug_staticHS_TabExDpi3_continued}) and
	\item 20 validity checks, i.e.\ calls of $\scQX$ that return 'no conflict' (one check for each of the 10 found minimal diagnoses; notice that $\scQX$ does only perform a single KB validity check by \textsc{isKBValid} in case it returns 'no conflict', see Algorithm~\ref{algo:qx}) or calls of \textsc{isKBValid} in line~\ref{algoline:static:isOntValid} in \textsc{staticHS} (one call for each of the 10 found minimal diagnoses),
\end{itemize}
computes
\begin{itemize}
	\item 10 minimal diagnoses w.r.t.\ the input DPI,
	\item 4 minimal conflict sets w.r.t.\ the input DPI and
	\item 2 queries and asks the user 2 logical formulas (1 per query)  
\end{itemize}
and stores
\begin{itemize}
	\item a maximum of 19 nodes (where node refers to the internal representation of a node in \textsc{staticHS} as a set of edge labels along a path from the root node to a leaf node; there are even more nodes in the sense of tree nodes in the picture in Figure~\ref{fig:example:inter_onto_debug_staticHS_TabExDpi3_continued}).\qed
\end{itemize}
%
\end{example}

\newgeometry{margin=2cm}

\begin{figure}[t]
\begin{minipage}[c]{0.472\textwidth} 
\small
\xygraph{
!{<0cm,0cm>;<1.7cm,0cm>:<0cm,1.2cm>::}
!{(1,4)}*+{\textcircled{\scriptsize 1}\tuple{1,2,5}^C}="c1c"
!{(0,3) }*+{\textcircled{\scriptsize 2}\checkmark_{(\md_1)}}="d1" 
!{(1,3) }*+{\textcircled{\scriptsize 3}\checkmark_{(\md_2)}}="d2" 
!{(3,3) }*+{?}="c2c"
"c1c":"d1"_{1}
"c1c":"d2"^{2}
"c1c":"c2c"^{5}
}
\vspace{5pt}
\begin{center}
\small Iteration 1
\end{center}
\end{minipage}
\begin{minipage}[c]{20pt}
\hspace{3pt} $\Bigg>$ 
\end{minipage}
\begin{minipage}[c]{0.41\textwidth}
\small
\begin{tabular}{l}                                          
		$\mD_{\checkmark}\cup\mD_{calc} = \emptyset \cup \setof{\md_1,\md_2} = \setof{[1],[2]}$ \\
		$\Queue = [\setof{5}]$ \\
		$\mC_{calc} = \setof{\tuple{1,2,5}}$ \\
		$\mD_{\times} = \emptyset$ \vspace{-8pt} \\
    \hspace{-8pt}\hdashrule{1\textwidth}{0.5pt}{2mm}\vspace{-4pt} \\
		$\tuple{Q_1,\Pt(Q_1)} = \tuple{\setof{E \rightarrow \lnot A},\tuple{\setof{\md_1},\setof{\md_2},\emptyset}}$ \\
		$u(Q_1) = \false$ \\
		$\mD_{\checkmark} = \setof{\md_2}$ \\
		$\mD_{out} = \mD_{\times} = \setof{\md_1}$
\end{tabular}
\end{minipage}
\begin{minipage}[c]{10pt}
$\Bigg>$ 
\end{minipage} 

\vspace{10pt}
\begin{minipage}[c]{0.45\textwidth} 
\small
\xygraph{
!{<0cm,0cm>;<1.7cm,0cm>:<0cm,1.2cm>::}
!{(1,4)}*+{\textcircled{\scriptsize 1}\tuple{1,2,5}^C}="c1c"
!{(0,3) }*+{\textcircled{\scriptsize 2}\checkmark_{(\md_1)}}="d1"
!{(1,3) }*+{\textcircled{\scriptsize 3}\checkmark_{(\md_2)}}="d2" 
!{(3,3) }*+{\textcircled{\scriptsize 5}\tuple{1,2,7}^C}="c2c"
!{(0,2) }*+{\textcircled{\scriptsize 4}\times}="inv_d1_q1"
!{(1,2) }*+{\textcircled{\scriptsize 4}\checkmark}="val_d2_q1"
!{(2,2) }*+{\textcircled{\scriptsize 6}\times_{(\supset\md_1)}}="nonmin"
!{(3,2) }*+{\textcircled{\scriptsize 7}\times_{(\supset\md_2)}}="nonmin1"
!{(4,2) }*+{\textcircled{\scriptsize 8}\checkmark_{(\md_3)}}="d3"
"c1c":"d1"_{1}
"c1c":"d2"^{2}
"c1c":"c2c"^{5}
"d1":@2{->}"inv_d1_q1"^{Q_1}
"d2":@2{->}"val_d2_q1"^{Q_1}
"c2c":"nonmin"_{1}
"c2c":"nonmin1"^{2}
"c2c":"d3"^{7}
}
\vspace{5pt}
\begin{center}
\small Iteration 2
\end{center}
\end{minipage}
\begin{minipage}[c]{15pt}
$\Bigg>$ 
\end{minipage}
\begin{minipage}[c]{0.41\textwidth}
\small\vspace{7pt}
\begin{tabular}{l}            
		%
		%
		$\mD_{\checkmark}\cup\mD_{calc} = \setof{\md_2} \cup \setof{\md_3} = \setof{[2],[5,7]}$ \\
		$\Queue = []$ \\
		$\mC_{calc} = \setof{\tuple{1,2,5}, \tuple{1,2,7}}$ \\
		$\mD_{\times} = \setof{\md_1} = \setof{[1]}$ \vspace{-8pt} \\
    \hspace{-8pt}\hdashrule{1\textwidth}{0.5pt}{2mm}\vspace{-4pt} \\
		$\tuple{Q_2,\Pt(Q_2)} = \tuple{\setof{Y \rightarrow \lnot A},\tuple{\setof{\md_2},\setof{\md_3},\emptyset}}$ \\
		$u(Q_2) = \false$ \\
		$\mD_{\checkmark} = \setof{\md_3}$ \\
		$\mD_{out} = \setof{\md_2}$ \\
		$\mD_{\times} = \setof{\md_1,\md_2}$
		\end{tabular}
\end{minipage}
\begin{minipage}[c]{10pt}
$\Bigg>$ 
\end{minipage} 

\vspace{10pt}
\begin{minipage}[c]{0.45\textwidth} 
\small
\xygraph{
!{<0cm,0cm>;<1.7cm,0cm>:<0cm,1.2cm>::}
!{(1,4)}*+{\textcircled{\scriptsize 1}\tuple{1,2,5}^C}="c1c"
!{(0,3) }*+{\textcircled{\scriptsize 2}\checkmark_{(\md_1)}}="d1"
!{(1,3) }*+{\textcircled{\scriptsize 3}\checkmark_{(\md_2)}}="d2" 
!{(3,3) }*+{\textcircled{\scriptsize 5}\tuple{1,2,7}^C}="c2c"
!{(0,2) }*+{\textcircled{\scriptsize 4}\times}="inv_d1_q1"
!{(1,2) }*+{\textcircled{\scriptsize 4}\checkmark}="val_d2_q1"
!{(2,2) }*+{\textcircled{\scriptsize 6}\times_{(\supset\md_1)}}="nonmin"
!{(3,2) }*+{\textcircled{\scriptsize 7}\times_{(\supset\md_2)}}="nonmin1"
!{(4,2) }*+{\textcircled{\scriptsize 8}\checkmark_{(\md_3)}}="d3"
!{(1,1) }*+{\textcircled{\scriptsize 9}\times}="inv_d2_q2"
!{(4,1) }*+{\textcircled{\scriptsize 9}\checkmark}="val_d3_q2"
"c1c":"d1"_{1}
"c1c":"d2"^{2}
"c1c":"c2c"^{5}
"d1":@2{->}"inv_d1_q1"^{Q_1}
"d2":@2{->}"val_d2_q1"^{Q_1}
"c2c":"nonmin"_{1}
"c2c":"nonmin1"^{2}
"c2c":"d3"^{7}
"val_d2_q1":@2{->}"inv_d2_q2"^{Q_2}
"d3":@2{->}"val_d3_q2"^{Q_2}
}
\vspace{5pt}
\begin{center}
\small Iteration 3
\end{center}
\end{minipage}
\begin{minipage}[c]{15pt}
$\Bigg>$ 
\end{minipage}
\begin{minipage}[c]{0.45\textwidth}
\small
\begin{tabular}{l}                                          
		$\mD_{\checkmark}\cup\mD_{calc} = \setof{\md_3} \cup \emptyset = \setof{[5,7]}$ \\
		$\Queue = []$ \\
		$\mC_{calc} = \setof{\tuple{1,2,5}, \tuple{1,2,7}}$ \\
		$\mD_{\times} = \setof{\md_1,\md_2} = \setof{[1],[2]}$\vspace{-8pt} \\
    \hspace{-8pt}\hdashrule{0.95\textwidth}{0.5pt}{2mm}\vspace{-4pt} \\
		$p_{\mD}(\md_3) = 1 $ \\
		$\Rightarrow\quad$ return the solution KB $(\mo\setminus\md_3) \qed$
		\end{tabular}
\end{minipage}

\vspace{10pt}
\caption[(Example~\ref{example:staticHS_simple_example_using_tabExDpi2}) Solving the Problem of Interactive Static KB Debugging]{(Example~\ref{example:staticHS_simple_example_using_tabExDpi2}) Solving the problem of Interactive Static KB Debugging (Problem Definition~\ref{prob_def:static}) for the example DPI given by Table~\ref{tab:example2} by means of Algorithm~\ref{algo:inter_onto_debug} and \textsc{staticHS}.} 
\label{fig:example:inter_onto_debug_staticHS_TabExDpi2}
\end{figure}
\restoregeometry

\newgeometry{margin=2cm}

\begin{sidewaysfigure*}
\begin{minipage}[c]{0.95\textwidth} 
\xygraph{
!{<0cm,0cm>;<1.5cm,0cm>:<0cm,1.5cm>::}
!{(4,4)}*+{\textcircled{\footnotesize 1}\tuple{1,2,5}^C}="c1c"
!{(1,3) }*+{\textcircled{\footnotesize 2}\tuple{2,4,6}^C}="c2c" 
!{(4,3) }*+{\textcircled{\footnotesize 5}\tuple{1,3,4}^C}="c3c" 
!{(11,3) }*+{\textcircled{\footnotesize 6}\tuple{2,4,6}^R}="c2r"
!{(0,2) }*+{?}="d4"
!{(1,2) }*+{\textcircled{\footnotesize 3}\checkmark_{(\md_1)}}="d1"
!{(2,2) }*+{\textcircled{\footnotesize 4}\checkmark_{(\md_2)}}="d2"
!{(3,2) }*+{?}="dup1"
!{(5,2) }*+{?}="?2-2"
!{(8,2) }*+{\textcircled{\footnotesize 7}\tuple{1,5,6,8}^C}="c4c"
!{(10,2) }*+{?}="?2-3"
!{(12,2) }*+{\textcircled{\footnotesize 8}\checkmark_{(\md_3)}}="d3"
!{(14,2) }*+{?}="?2-4"
!{(7,1) }*+{?}="?3-1"
!{(8,1) }*+{?}="?3-2"
!{(9,1) }*+{?}="?3-3"
!{(10,1) }*+{?}="?3-4"
"c1c":"c2c"_{1}^(0.65){0.41}
"c1c":"c3c"_{2}^(0.7){0.25}
"c1c":"c2r"_{5}^(0.87){0.25}
"c2c":"d4"_{2}_(0.9){0.09}
"c2c":"d1"_{4}^(0.75){0.28}
"c2c":"d2"^{6}^(0.85){0.27}
"c3c":"dup1"_{1}^(0.6){0.09}
"c3c":"?2-2"_{3}^(0.88){0.07}
"c3c":"c4c"_{4}^(0.75){0.18}
"c2r":"?2-3"_{2}^(0.7){0.06}
"c2r":"d3"_(0.4){4}^(0.8){0.18}
"c2r":"?2-4"^{6}^(0.88){0.17}
"c4c":"?3-1"_{1}_(0.85){0.06}
"c4c":"?3-2"_{5}_(0.78){0.04}
"c4c":"?3-3"_{6}^(0.9){0.11}
"c4c":"?3-4"^{8}^(0.9){0.04}
}
\vspace{5pt}
\begin{center}
\small Iteration 1
\end{center}
\end{minipage}
\begin{minipage}[c]{20pt}
$\Bigg>$ 
\end{minipage}

\vspace{20pt}
\begin{minipage}[c]{0.95\textwidth}
\small  
\begin{tabular}{l}                                          
		$\mD_{\checkmark}\cup\mD_{calc} = \emptyset\cup\setof{\md_1, \md_2, \md_3} = \setof{[1,4],[1,6],[5,4]}$, 
		$\quad\Queue = [\setof{5,6},\setof{2,4,6},\setof{1,2},\setof{2,1},\setof{2,3}, 
		\setof{5,2},\setof{2,4,1},\setof{2,4,5},\setof{2,4,8}]$, 
		$\quad\mD_{\times} = \emptyset$\\
		$\mC_{calc} = \setof{\tuple{1,2,5},\tuple{2,4,6},\tuple{1,3,4},\tuple{1,5,6,8}}$, \vspace{-8pt} \\
    \hspace{-8pt}\hdashrule{0.95\textwidth}{0.5pt}{2mm}\vspace{-4pt} \\
		$\tuple{Q_1,\Pt(Q_1)} = \tuple{\setof{B \sqsubseteq K},\tuple{\setof{\md_1,\md_2},\setof{\md_3},\emptyset}}$, 
    $\quad u(Q_1) = \true$, 
		$\quad\mD_{\checkmark} = \setof{\md_1,\md_2}$,  $\quad \mD_{out} = \mD_{\times} = \setof{\md_3}$
		\end{tabular}
\end{minipage}
\begin{minipage}[c]{10pt}
$\Bigg>$ 
\end{minipage} 

\vspace{10pt}
\begin{minipage}[c]{0.95\textwidth} 
\xygraph{
!{<0cm,0cm>;<1.45cm,0cm>:<0cm,1.5cm>::}
!{(4,4)}*+{\textcircled{\footnotesize 1}\tuple{1,2,5}^C}="c1c"
!{(1,3) }*+{\textcircled{\footnotesize 2}\tuple{2,4,6}^C}="c2c" 
!{(4,3) }*+{\textcircled{\footnotesize 5}\tuple{1,3,4}^C}="c3c" 
!{(11,3) }*+{\textcircled{\footnotesize 6}\tuple{2,4,6}^R}="c2r"
!{(0,2) }*+{\textcircled{\tiny 13}\times_{(dup)}}="dup1"
!{(1,2) }*+{\textcircled{\footnotesize 3}\checkmark_{(\md_1)}}="d1"
!{(2,2) }*+{\textcircled{\footnotesize 4}\checkmark_{(\md_2)}}="d2"
!{(3,2) }*+{\textcircled{\tiny 14}\checkmark_{(\md_5)}}="d5"
!{(5,2) }*+{?}="?2-2"
!{(8,2) }*+{\textcircled{\footnotesize 7}\tuple{1,5,6,8}^C}="c4c"
!{(10,2) }*+{?}="?2-3"
!{(12,2) }*+{\textcircled{\footnotesize 8}\checkmark_{(\md_3)}}="d3"
!{(14,2) }*+{\textcircled{\tiny 10}\tuple{1,3,4}^R}="c3r"
!{(7,1) }*+{?}="?3-1"
!{(8,1) }*+{?}="?3-2"
!{(10,1) }*+{?}="?3-4"
!{(9,1) }*+{\textcircled{\tiny 11}\checkmark_{(\md_4)}}="d4"
!{(12,1) }*+{\textcircled{\footnotesize 9}\times}="inv_d3_q1"
!{(1,1) }*+{\textcircled{\footnotesize 9}\checkmark}="val_d1_q1"
!{(2,1) }*+{\textcircled{\footnotesize 9}\checkmark}="val_d2_q1"
!{(13,1) }*+{?}="?3-5"
!{(14,1) }*+{?}="?3-6"
!{(15,1) }*+{\textcircled{\tiny 12}\times_{(\supset\md_3)}}="sup_d3"
!{(9,0) }*+{\textcircled{\tiny 11}\times}="inv_d4_q1"
"c1c":"c2c"_{1}^(0.65){0.41}
"c1c":"c3c"_{2}^(0.7){0.25}
"c1c":"c2r"_{5}^(0.87){0.25}
"c2c":"dup1"_{2}_(0.85){0.09}
"c2c":"d1"_{4}^(0.75){0.28}
"c2c":"d2"^{6}^(0.85){0.27}
"c3c":"d5"_{1}^(0.6){0.09}
"c3c":"?2-2"_{3}^(0.88){0.07}
"c3c":"c4c"_{4}^(0.75){0.18}
"c2r":"?2-3"_{2}^(0.7){0.06}
"c2r":"d3"_{4}^(0.8){0.18}
"c2r":"c3r"^{6}^(0.88){0.17}
"c4c":"?3-1"_{1}_(0.85){0.06}
"c4c":"?3-2"_{5}_(0.78){0.04}
"c4c":"?3-4"^{8}^(0.9){0.04}
"c4c":"d4"_{6}^(0.9){0.11}
"d3":@2{->}"inv_d3_q1"^{Q_1}
"d1":@2{->}"val_d1_q1"^{Q_1}
"d2":@2{->}"val_d2_q1"^{Q_1}
"c3r":"?3-5"_{1}_(0.75){0.06}
"c3r":"?3-6"_{3}^(0.75){0.04}
"c3r":"sup_d3"^(0.55){4}^(0.85){0.11}
"d4":@2{->}"inv_d4_q1"^{Q_1}
}
\vspace{5pt}
\begin{center}
\small Iteration 2
\end{center}
\end{minipage}
\begin{minipage}[c]{20pt}
$\Bigg>$ 
\end{minipage}

\vspace{10pt}
\caption[(Example~\ref{example:staticHS_complex_example_using_tabExDpi3}) Solving the Problem of Interactive Static KB Debugging]{(Example~\ref{example:staticHS_complex_example_using_tabExDpi3}) Solving the problem of Interactive Static KB Debugging (Problem Definition~\ref{prob_def:static}) for the example DPI given by Table~\ref{tab:example3} by means of \newline Algorithm~\ref{algo:inter_onto_debug} and \textsc{staticHS}.} 
\label{fig:example:inter_onto_debug_staticHS_TabExDpi3}
\end{sidewaysfigure*}

\begin{sidewaysfigure*}
\begin{minipage}[c]{0.95\textwidth}
\small  
\begin{tabular}{l}                                          
		$\mD_{\checkmark} \cup \mD_{calc} = \setof{\md_1, \md_2} \cup \setof{\md_5} = \setof{[1,4],[1,6],[2,1]}$, 
		$\quad\Queue = [\setof{2,4,6},\setof{2,3},\setof{5,2},\setof{2,4,1},\setof{5,6,1},\setof{2,4,5},\setof{5,6,3}]$,
		$\quad\mD_{\times} = \setof{\md_3,\md_4} = \setof{[5,4],[2,4,6]}$   \\
		$\mC_{calc} = \setof{\tuple{1,2,5},\tuple{2,4,6},\tuple{1,3,4},\tuple{1,5,6,8}}$, \vspace{-8pt} \\
    \hspace{-8pt}\hdashrule{0.95\textwidth}{0.5pt}{2mm}\vspace{-4pt} \\		
		$\tuple{Q_2,\Pt(Q_2)} = \tuple{\setof{B \sqsubseteq \exists r.F},\tuple{\setof{\md_1},\setof{\md_2,\md_5},\emptyset}}$, 
    $\quad u(Q_2) = \true$,
		$\quad \mD_{\checkmark} = \setof{\md_1}$, $\quad\mD_{out} = \setof{\md_2,\md_5}$, $\quad\mD_{\times} = \setof{\md_3,\md_4,\md_2,\md_5}$
		\end{tabular}
\end{minipage}
\begin{minipage}[c]{10pt}
$\Bigg>$ 
\end{minipage} 

\vspace{10pt}
\begin{minipage}[c]{0.95\textwidth} 
\xygraph{
!{<0cm,0cm>;<1.27cm,0cm>:<0cm,1.5cm>::}
!{(4,4)}*+{\textcircled{\footnotesize 1}\tuple{1,2,5}^C}="c1c"
!{(1,3) }*+{\textcircled{\footnotesize 2}\tuple{2,4,6}^C}="c2c" 
!{(6,1) }*+{\textcircled{\footnotesize 5}\tuple{1,3,4}^C}="c3c" 
!{(13,4) }*+{\textcircled{\footnotesize 6}\tuple{2,4,6}^R}="c2r"
!{(0,2) }*+{\textcircled{\tiny 13}\times_{(dup)}}="dup1"
!{(1,2) }*+{\textcircled{\footnotesize 3}\checkmark_{(\md_1)}}="d1"
!{(2,2) }*+{\textcircled{\footnotesize 4}\checkmark_{(\md_2)}}="d2"
!{(3,0) }*+{\textcircled{\tiny 14}\checkmark_{(\md_5)}}="d5"
!{(7,0) }*+{\textcircled{\tiny 16}\tuple{1,5,6,8}^R}="c4r"
!{(13,0) }*+{\textcircled{\footnotesize 7}\tuple{1,5,6,8}^C}="c4c"
!{(9,3) }*+{\textcircled{\tiny 17}\tuple{1,3,4}^R}="c3r1"
!{(12,3) }*+{\textcircled{\footnotesize 8}\checkmark_{(\md_3)}}="d3"
!{(15,3) }*+{\textcircled{\tiny 10}\tuple{1,3,4}^R}="c3r"
!{(10,-1) }*+{\textcircled{\tiny 18}\times_{(\supset\md_5)}}="sup_d5"
!{(12,-1) }*+{\textcircled{\tiny 20}\times_{(\supset\md_3)}}="sup_d3_1"
!{(16,-1) }*+{\textcircled{\tiny 21}\checkmark_{(\md_6)}}="d6"
!{(14,-1) }*+{\textcircled{\tiny 11}\checkmark_{(\md_4)}}="d4"
!{(12,2) }*+{\textcircled{\footnotesize 9}\times}="inv_d3_q1"
!{(1,1) }*+{\textcircled{\footnotesize 9}\checkmark}="val_d1_q1"
!{(2,1) }*+{\textcircled{\footnotesize 9}\checkmark}="val_d2_q1"
!{(3,-1) }*+{\textcircled{\tiny 15}\times}="inv_d5_q2"
!{(13,2) }*+{\textcircled{\tiny 19}\times_{(\supset\md_2)}}="sup_d2"
!{(15,2) }*+{\textcircled{\tiny 22}\checkmark_{(\md_7)}}="d7"
!{(17,2) }*+{\textcircled{\tiny 12}\times_{(\supset\md_3)}}="sup_d3"
!{(7,2) }*+{\textcircled{\tiny 28}\times_{(\supset\md_5)}}="sup_d5_1"
!{(11,2) }*+{\textcircled{\tiny 24}\times_{(\supset\md_3)}}="sup_d3_2"
!{(3,-2) }*+{\textcircled{\tiny 25}\times_{(\supset\md_5)}}="sup_d5_2"
!{(5,-2) }*+{\textcircled{\tiny 26}\times_{(dup)}}="dup2"
!{(7,-2) }*+{\textcircled{\tiny 23}\checkmark_{(\md_8)}}="d8"
!{(9,-2) }*+{\textcircled{\tiny 27}\checkmark_{(\md_{9})}}="d9"
!{(9,2) }*+{\textcircled{\tiny 29}\checkmark_{(\md_{10})}}="d10"
!{(14,-2) }*+{\textcircled{\tiny 11}\times}="inv_d4_q1"
!{(15,1) }*+{\textcircled{\tiny 22}\times}="inv_d7_q1"
!{(1,0) }*+{\textcircled{\tiny 15}\checkmark}="val_d1_q2"
!{(2,0) }*+{\textcircled{\tiny 15}\times}="inv_d2_q2"
!{(16,-2) }*+{\textcircled{\tiny 21}\times}="inv_d6_q1"
!{(7,-3) }*+{\textcircled{\tiny 23}\times}="inv_d8_q2"
!{(9,-3) }*+{\textcircled{\tiny 27}\times}="inv_d9_q1"
!{(9,1) }*+{\textcircled{\tiny 29}\times_{(\md_{10})}}="inv_d10_q1"
"c1c":"c2c"_{1}^(0.65){0.41}
"c1c":"c3c"_{2}^(0.9){0.25}
"c1c":"c2r"_{5}^(0.9){0.25}
"c2c":"dup1"_{2}_(0.8){0.09}
"c2c":"d1"_{4}^(0.75){0.28}
"c2c":"d2"^{6}^(0.85){0.27}
"c3c":"d5"_{1}^(0.7){0.09}
"c3c":"c4r"_{3}^(0.7){0.07}
"c3c":"c4c"_{4}^(0.84){0.18}
"c2r":"c3r1"_{2}^(0.7){0.06}
"c2r":"d3"_{4}^(0.6){0.18}
"c2r":"c3r"_(0.4){6}^(0.75){0.17}
"c4c":"sup_d5"_{1}^(0.68){0.06}
"c4c":"sup_d3_1"^{5}_(0.85){0.04}
"c4c":"d6"^{8}^(0.8){0.04}
"c4c":"d4"_{6}^(0.87){0.11}
"d3":@2{->}"inv_d3_q1"^{Q_1}
"d1":@2{->}"val_d1_q1"^{Q_1}
"d2":@2{->}"val_d2_q1"^{Q_1}
"c3r":"sup_d2"_{1}^(0.6){0.06}
"c3r":"d7"_{3}^(0.75){0.04}
"c3r":"sup_d3"_{4}^(0.75){0.11}
"c4r":"sup_d5_2"_{1}^(0.75){0.02}
"c4r":"dup2"_{5}^(0.8){0.02}
"c4r":"d8"_{6}^(0.85){0.04}
"c4r":"d9"_{8}^(0.88){0.02}
"c3r1":"sup_d3_2"_{4}^(0.75){0.04}
"c3r1":"sup_d5_1"_{1}^(0.6){0.02}
"c3r1":"d10"_{3}^(0.75){0.02}
"d4":@2{->}"inv_d4_q1"^{Q_1}
"d5":@2{->}"inv_d5_q2"^{Q_2}
"val_d1_q1":@2{->}"val_d1_q2"^{Q_2}
"val_d2_q1":@2{->}"inv_d2_q2"^{Q_2}
"d6":@2{->}"inv_d6_q1"^{Q_1}
"d8":@2{->}"inv_d8_q2"^{Q_2}
"d7":@2{->}"inv_d7_q1"^{Q_1}
"d9":@2{->}"inv_d9_q1"^{Q_1}
"d10":@2{->}"inv_d10_q1"^{Q_1}
}
\vspace{5pt}
\begin{center}
\small Iteration 3
\end{center}
\end{minipage}
\begin{minipage}[c]{20pt}
$\Bigg>$ 
\end{minipage}

\vspace{10pt}
\begin{minipage}[c]{0.95\textwidth}
\small  
\begin{tabular}{l}                        
    $\mD_{\checkmark} \cup \mD_{calc} = \setof{\md_1} \cup \emptyset = \setof{\md_1}$, 
		$\quad \Queue = []$, 
		$\quad \mC_{calc} = \setof{\tuple{1,2,5},\tuple{2,4,6},\tuple{1,3,4},\tuple{1,5,6,8}}$, \\
		$\mD_{\times} = \setof{\md_3,\md_4,\md_2,\md_5,\md_6,\md_7,\md_8,\md_9,\md_{10}} = \setof{[5,4],[2,4,6],[1,6],[2,1],[2,4,8],[5,6,3],[2,3,6],[2,3,8],[5,2,3]}$, \vspace{-8pt} \\
    \hspace{-8pt}\hdashrule{0.95\textwidth}{0.5pt}{2mm}\vspace{-4pt} \\
		$p_{\mD}(\md_1) = 1  \quad\Rightarrow\quad$ return the solution KB $(\mo\setminus\md_1) \cup p_1 \quad$ ($p_1$: cf.\ Table~\ref{tab:example3}) $\qed$
\end{tabular}
\end{minipage}
\caption[(Example~\ref{example:staticHS_complex_example_using_tabExDpi3} continued) Solving the Problem of Interactive Static KB Debugging]{(Example~\ref{example:staticHS_complex_example_using_tabExDpi3} continued) Solving the problem of Interactive Static KB Debugging (Problem Definition~\ref{prob_def:static}) for the example DPI given by Table~\ref{tab:example3} by means of \newline Algorithm~\ref{algo:inter_onto_debug} and \textsc{staticHS}.} 
\label{fig:example:inter_onto_debug_staticHS_TabExDpi3_continued}
\end{sidewaysfigure*}
\restoregeometry

\newgeometry{margin=2cm}

\begin{algorithm*}
\small
\caption{Iterative Construction of a Static Hitting Set Tree} \label{algo:inter_stat_hs}
\begin{algorithmic}[1]
\Require 
a tuple $\tuple{ \langle\mo,\mb,\Tp,\Tn\rangle_\RQ, \Queue, t, n_{\min}, n_{\max}, \mC_{calc}, \mD_{\checkmark}, \mD_{\times}, p(), \Tp', \Tn' }$ consisting of
\begin{itemize}
	\item the DPI $\langle\mo,\mb,\Tp,\Tn\rangle_\RQ$ given as input to Algorithm~\ref{algo:inter_onto_debug},
	\item the overall sets of positively ($\Tp'$) and negatively ($\Tn'$) answered queries added as test cases to $\langle\mo,\mb,\Tp,\Tn\rangle_\RQ$ so far, 
	\item the current queue $\Queue$ of open (non-labeled) nodes of a (partial) wpHS-tree,
	\item some desired computation timeout $t$,
	\item a desired minimal ($n_{\min}\geq2$) and maximal ($n_{\max}$) number of minimal diagnoses to be returned, 
	\item the set $\mC_{calc}$ of all minimal conflict sets w.r.t.\ $\langle\mo,\mb,\Tp,\Tn\rangle_\RQ$ computed so far,
	\item the set $\mD_{\checkmark}$ of all minimal diagnoses w.r.t.\ $\langle\mo,\mb,\Tp,\Tn\rangle_\RQ$ computed so far that satisfy all test cases $\Tp'$ and $\Tn'$,
	\item the set $\mD_{\times}$ of all minimal diagnoses w.r.t.\ $\langle\mo,\mb,\Tp,\Tn\rangle_\RQ$ computed so far 
	that do not satisfy all test cases $\Tp'$ and $\Tn'$.
	\item a function $p: \mo \rightarrow (0,0.5)$.
\end{itemize}
\Ensure 
a tuple $\tuple{\mD,\Queue, \mathbf{C}_{calc}, \mD_{\times}}$ where
\begin{itemize}
\item $\mD$ is the current set of leading diagnoses such that
\begin{enumerate}[(a)]
\item $\mD \subseteq \minD_{\langle\mo,\mb,\Tp,\Tn\rangle_\RQ} \cap \minD_{\langle\mo,\mb,\Tp\cup\Tp',\Tn\cup\Tn'\rangle_\RQ}$ is the set of most probable 
 minimal diagnoses w.r.t.\ $\langle\mo,\mb,\Tp,\Tn\rangle_\RQ$ that satisfy all test cases $\Tp'$ and $\Tn'$ such that 
\begin{enumerate}[(i)]
\item $n_{\min} \leq |\mD| \leq n_{\max}$ and 
\item $\mD\supset\mD_{\checkmark}$,
\end{enumerate} 
if such a set $\mD$ 
 exists,
or 
\item $\mD$ is equal to the set of all minimal diagnoses $\minD_{\langle\mo,\mb,\Tp,\Tn\rangle_\RQ} \cap \minD_{\langle\mo,\mb,\Tp\cup\Tp',\Tn\cup\Tn'\rangle_\RQ}$, otherwise,
\end{enumerate}
where ``most-probable'' refers to the probability measure $p_{nodes}()$ (cf.\ Definition~\ref{def:p_node()}) obtained from the given function $p()$;
\item $\Queue$ is the current queue of open (non-labeled) nodes of the produced (partial) wpHS-tree, 
\item $\mC_{calc}$ is the set of all minimal conflict sets w.r.t.\ $\langle\mo,\mb,\Tp,\Tn\rangle_\RQ$ computed so far and 
\item $\mD_{\times}$ comprises those minimal diagnoses w.r.t.\ $\langle\mo,\mb,\Tp,\Tn\rangle_\RQ$ computed so far that do not satisfy all test cases $\Tp'$ and $\Tn'$.
\end{itemize}  
\vspace{10pt}
\Procedure{staticHS}{$\langle\mo,\mb,\Tp,\Tn\rangle_\RQ, \Queue, t, n_{\min}, n_{\max}, \mathbf{C}_{calc}, \mD_{\checkmark}, \mD_{\times}, p(), \Tp', \Tn'$}
\State $t_{start} \gets \Call{getTime}{ }$
\State $\mD_{calc} \gets \emptyset$
\Repeat
\State $\mathsf{node} \gets \Call{getFirst}{\Queue}$ \label{algoline:static:getfirst}
\State $\Queue \gets \Call{deleteFirst}{\Queue}$\label{algoline:static_update_Q}
\State $\tuple{L,\mathbf{C}} \gets \Call{sLabel}{\langle\mo,\mb,\Tp,\Tn\rangle_\RQ, \mathsf{node},\mathbf{C}_{calc},\mD_{\times} \cup \mD_{\checkmark} \cup \mD_{calc}, \Queue}$ \label{algoline:static:slabel}
\State $\mathbf{C}_{calc} \gets \mathbf{C}$ \label{algoline:static:update_Ccalc}
\If{$L = valid$} \label{algoline:static:if_L_valid}     \Comment{$\mathsf{node}$ is minimal diagnosis w.r.t.\ $\langle\mo,\mb,\Tp,\Tn\rangle_\RQ$}
	\If{\Call{isKBValid}{$\mo\setminus\mathsf{node}, \langle\cdot,\mb,\Tp\cup\Tp',\Tn\cup\Tn'\rangle_\RQ$}}\label{algoline:static:isOntValid}  \Comment{\textsc{isKBValid} (see Algorithm~\ref{algo:qx})}
		\State $\mD_{calc} \gets \mD_{calc} \cup \setof{\mathsf{node}}$\label{algoline:static:add_to_Dcalc}  \Comment{$\mathsf{node}$ does satisfy all test cases $\Tp'$ and $\Tn'$}
	\Else
		\State $\mD_{\times} \gets \mD_{\times} \cup \setof{\mathsf{node}}$\label{algoline:static:add_to_Dtimes} \Comment{$\mathsf{node}$ does not satisfy all test cases $\Tp'$ and $\Tn'$}
	\EndIf
\ElsIf{$L = closed$} \label{algoline:static:if_L_closed}
				\Comment{do nothing, no need to store non-minimal diagnoses}
\Else 	\Comment{$L$ must be a minimal conflict set}
	\For{$e \in L$}
		\State $\Queue \gets \Call{insertSorted}{ \mathsf{node} \cup \setof{e}, \Queue, p_{nodes}(), descending}$  \label{algoline:static:generate_nodes}
\EndFor
\EndIf
\Until{$\Queue=[] \lor [|\mD_{calc}| \neq \emptyset \land \left|\mD_{calc} \cup \mD_{\checkmark}\right| \geq n_{\min} \land ( |\mD_{calc} \cup \mD_{\checkmark}| = n_{\max} \lor \Call{getTime}{ } - t_{start} > t)]$}\label{algoline:static:until}
\State \Return $\tuple{\mD_{calc} \cup \mD_{\checkmark},\Queue, \mathbf{C}_{calc}, \mD_{\times}}$
\EndProcedure
\vspace{10pt}
\Procedure{\textsc{sLabel}}{$\langle\mo,\mb,\Tp,\Tn\rangle_\RQ,\mathsf{node},\mathbf{C}_{calc}, \mD_{(\times,\checkmark,calc)}, \Queue$}
\For{$\mathsf{nd} \in \mD_{(\times,\checkmark,calc)}$}\label{algoline:slabel:non_min_crit_start}
	\If{$\mathsf{node} \supseteq \mathsf{nd}$}    \Comment{$\mathsf{node}$ is a non-minimal diagnosis}
			\State \Return $\tuple{closed,\mathbf{C}_{calc}}$
	\EndIf
\EndFor\label{algoline:slabel:non_min_crit_end}
\For{$\mathsf{nd} \in \Queue$}\label{algoline:slabel:dup_crit_start}
	\If{$\mathsf{node} = \mathsf{nd}$}      \Comment{$\mathsf{node}$ is a duplicate node}
			\State \Return $\tuple{closed,\mathbf{C}_{calc}}$
	\EndIf
\EndFor\label{algoline:slabel:dup_crit_end}
\For{$\mc \in \mathbf{C}_{calc}$}   \label{algoline:slabel:reuse_crit_start}
	\If{$\mc \cap \mathsf{node} = \emptyset$}    \Comment{the minimal conflict set $\mc$ can be reused to label $\mathsf{node}$}
		\State \Return $\tuple{\mc,\mathbf{C}_{calc}}$  \label{algoline:slabel:reuse_crit_end}
	\EndIf
\EndFor
\State $L\gets \Call{QX}{\langle\mo\setminus\mathsf{node},\mb,\Tp,\Tn\rangle_\RQ}$\label{algoline:slabel:qx} \Comment{see Algorithm~\ref{algo:qx} (page~\pageref{algo:qx})}
\If{$L$ = \text{'no conflict'}}						\Comment{$\mathsf{node}$ is a diagnosis}
	\State \Return $\tuple{valid,\mathbf{C}_{calc}}$\label{algoline:slabel:return_valid}
\Else						\Comment{$L$ is a \emph{new} minimal conflict set ($\notin \mathbf{C}_{calc}$)}
	\State $\mathbf{C}_{calc} \gets \mathbf{C}_{calc} \cup \setof{L}$
	\State \Return $\tuple{L,\mathbf{C}_{calc}}$
\EndIf
\EndProcedure
\end{algorithmic}
\normalsize
\end{algorithm*}
\restoregeometry

\section[\textsc{dynamicHS}: A Dynamic Iterative Diagnosis Computation Algorithm]{\textsc{dynamicHS}: A Dynamic Iterative Diagnosis Computation Algorithm%
\sectionmark{Dynamic Algorithm}}
\sectionmark{Dynamic Algorithm}
\label{sec:DynamicHSTree}
As the name already suggests, \textsc{dynamicHS} (Algorithm~\ref{algo:inter_dyn_hs}) is a procedure that solves the problem of \emph{Interactive Dynamic KB Debugging} defined by Problem Definition~\ref{prob_def:dynamic} if used for leading diagnosis computation in Algorithm~\ref{algo:inter_onto_debug}. \textsc{dynamicHS} is sound, complete and optimal w.r.t.\ the set of solutions of the \emph{Interactive Dynamic KB Debugging} problem. 
Optimality refers to the best-first computation of minimal diagnoses regarding a given probability measure.

\subsection{Overview and Intuition}
\label{sec:OverviewAndIntuition}
\paragraph{Synoptic View of the Algorithm.} \textsc{dynamicHS} (Algorithm~\ref{algo:inter_dyn_hs}) is employed as a subroutine in Algorithm~\ref{algo:inter_onto_debug} with $mode = dynamic$ to build up a hitting set tree iteratively. That is, each time \textsc{dynamicHS} is called in Algorithm~\ref{algo:inter_onto_debug}, it expands the existing tree only to a sufficient extent in order to determine a desired number of new leading diagnoses used for the generation of the next query. Then, the leading diagnoses set is returned. 

Outside of the \textsc{dynamicHS} method in Algorithm~\ref{algo:inter_onto_debug}, a new diagnosis probability distribution is obtained by the diagnosis probability update (cf.\ Section~\ref{sec:DetailedAlgorithmDescription}). Once this distribution involves one diagnosis, the probability of which exceeds a predefined threshold $1-\sigma$, the algorithm terminates. The output is a solution KB w.r.t.\ the \emph{current DPI} built from this highly probable minimal diagnosis.

\begin{remark}\label{rem:sigma_in_dynamicHS}
In case $\sigma$ has a predefined value of zero, the output is the (exact) solution to the problem of \emph{Interactive Dynamic KB Debugging} for the input DPI. In a scenario where some fault tolerance $\sigma > 0$ is given, the solution KB returned by Algorithm~\ref{algo:inter_onto_debug} is an approximation of the (exact) solution to \emph{Interactive Dynamic KB Debugging} for the input DPI where a better approximation can be expected for smaller values of $\sigma$ (cf.\ Remark~\ref{rem:approximate_solution}). ``Better'' in this context refers to the satisfaction of desired semantic properties of the KB returned by Algorithm~\ref{algo:inter_onto_debug}, i.e.\ desired entailments and desired non-entailments of the KB. The intuition is that specification of additional test cases $T$ guarantees the output of a KB complying with these test cases, whereas accepting 
one -- albeit highly probable -- of multiple solution KBs without having incorporated $T$ leaves open the possibility for this KB to not fulfill $T$.

However, answering queries is effort for an interacting user. Therefore, the approach that involves the ``early'' termination of the algorithm after a solution KB has a sufficiently high probability (lower than 1) constitutes a trade-off between exactness of the output and the effort of the user and overall execution time of the interactive KB debugging algorithm, respectively.\qed
\end{remark}

In case there is no highly probable leading diagnosis, a query constructed from the current set of leading diagnoses is asked to the user. The user's answer is incorporated into the current DPI resulting in a new DPI. Thereafter, \textsc{dynamicHS} is invoked again given this new DPI 
as an argument.

\paragraph{Storage of the Tree.} Between each two calls of \textsc{dynamicHS} in Algorithm~\ref{algo:inter_onto_debug}, the ``state'' of the current hitting set tree is stored by variables 
\begin{itemize}
	\item $\mD_{calc}$ -- computed minimal diagnoses w.r.t.\ the current DPI,
	\item $\Queue$ -- the list of open, non-labeled nodes,
	\item $\mC_{calc}$ -- (not necessarily minimal) conflict sets w.r.t.\ the current DPI computed so far,
	\item $\mD_{\supset}$ -- non-minimal diagnoses w.r.t.\ the current DPI computed so far,
	\item $\Queue_{dup}$ -- non-labeled duplicate nodes (i.e.\ nodes corresponding to tree branches with the same set of edge labels as branches that are already present in the tree)
	\item $\mD_{\times}$ -- the empty set (is filled up during Algorithm~\ref{algo:inter_onto_debug} between two calls of \textsc{dynamicHS} with diagnoses from $\mD_{calc}$ that have been invalidated by an answered query)
\end{itemize}
where nodes in the tree again store (among others) the edge labels on the path from the root node to themselves. 

\paragraph{Tree Update.} It is immediately apparent from the enumeration given above that, in comparison to \textsc{staticHS}, additional collections, i.e.\ $\mD_{\supset}$ as well as $\Queue_{dup}$, need to be maintained in order to ``remember'' the current tree while Algorithm~\ref{algo:inter_onto_debug} is processing outside of the method \textsc{dynamicHS}. The cause for these additional variables is the \emph{tree update} necessary after each addition of a test case to a DPI. For, each iteration of \textsc{dynamicHS} considers a different DPI in terms of the test cases. And, any two different DPIs in general lead to a different hitting set tree and to different sets of minimal diagnoses and conflict sets. Hence, the idea of the tree update is the following: Reuse the partial hitting set tree $T$ (stored by the variables described above) constructed before the new test case was added to the current DPI $DPI_j$ and perform suitable modifications to $T$ 
in order to obtain a tree $T'$ such that the further expansion of $T'$ allows to identify \emph{all} minimal diagnoses w.r.t.\ the new DPI $DPI_{j+1}$ resulting from the addition of the new test case to $DPI_j$. In other words, the tree update seeks to establish a tree that is equivalent to one built by execution of \textsc{dynamicHS} using the new DPI $DPI_{j+1}$ starting from an empty tree.

\paragraph{Node Storage.} Notice that, unlike in \textsc{staticHS} or \textsc{HS}, it is crucial to store nodes not as sets in \textsc{dynamicHS}, but as \emph{ordered lists of formulas}. That is, each node $\mathsf{nd}$ stores a list of all the edge labels along the (directed) path in the hitting set tree from the root node to $\mathsf{nd}$ where the order of formulas in the list is given by the order of traversing the edge labels along this path. Additionally, \textsc{dynamicHS} stores the attribute $\mathsf{nd.cs}$ for each node $\mathsf{nd}$ which is an ordered list including the node labels, i.e.\ the conflict sets, along the path from the root node to $\mathsf{nd}$ in analogous way. Associating a node with these two lists instead of one set is necessary from the point of view of the tree update.
Because this facilitates the differentiation between two nodes corresponding to an equal (partial) diagnosis. For example, there could be some node $\mathsf{nd}_1$ that is ``redundant'' after some query $Q$ has been answered, but there is a set-equal node $\mathsf{nd}_2$ which is still ``relevant'' (set-equality refers to equal \emph{sets}, not lists, of edge labels stored by two nodes). 
In this case, the algorithm should get rid of $\mathsf{nd}_1$ (in order to save time and space) while preserving node $\mathsf{nd}_2$ (in order to maintain completeness). Associating set-equal nodes with each other might thus either lead to unnecessary tree expansion steps (if none is deleted) or incompleteness of the algorithm concerning the consideration of all minimal diagnoses (in case both are deleted).
%
%
%
%
%

\paragraph{Addition of a Test Case Changes Set of Solutions.} Unlike the \textsc{staticHS} algorithm, which is strongly related to the non-interactive hitting set algorithm \textsc{HS} (Algorithm~\ref{algo:hs}) as outlined in Section~\ref{sec:TheIntuition},
the hitting set tree produced by \textsc{dynamicHS} will usually differ significantly from the non-interactive hitting set tree produced by \textsc{HS}. 
The reason for this is that in \textsc{dynamicHS} the initial DPI $DPI_0$ is not fixed (in that conflict sets and diagnoses are calculated only w.r.t.\ $DPI_0$), but \emph{new test cases are also used for the computation of minimal conflict sets (and thus minimal diagnoses) and not only for the invalidation of diagnoses}. Hence, every time a query has been answered and a respective test case has been incorporated into the DPI, the minimal conflict sets computed for the old DPI $DPI_j$ might not be minimal conflict sets w.r.t.\ the current DPI $DPI_{j+1}$ anymore (see Examples~\ref{example:dynamicHS_small_example_using_tabExDpi2} and \ref{example:dynamicHS_large_example_using_tabExDpi3}). 
On the one hand, a minimal conflict set $\mc$ w.r.t.\ $DPI_j$ might be a non-minimal conflict set w.r.t.\ $DPI_{j+1}$ (since there is a new minimal conflict set $\mc' \subset \mc$ w.r.t.\ $DPI_{j+1}$). On the other hand, there might be also ``completely new'' minimal conflict sets $\mc_k$ w.r.t.\ $DPI_{j+1}$ which are in no set-relationship with any minimal conflict set w.r.t.\ $DPI_j$. 

Due to this changing set of minimal conflict sets, the set of minimal diagnoses is variable as well (cf.\ Proposition~\ref{prop:mindiag_mincs}). To see this, let $\md$ be a minimal diagnosis w.r.t.\ $DPI_j$. Then $\md$ hits all minimal conflict sets $\mc_k$ in $\minC_{DPI_j}$. Now, assume that $\md$ comprises (only) the element $\tax$ from $\mc_k$, but there is a minimal conflict set $\mc'_k$ in $\minC_{DPI_{j+1}}$ such that $\mc'_k \subseteq \mc_k \setminus \setof{\tax}$. In this case, $\md$ is not a (minimal) hitting set of all minimal conflict sets in $\minC_{DPI_{j+1}}$ (since $\md$ does not hit $\mc'_k$), i.e.\ $\md$ is not a (minimal) diagnosis w.r.t.\ $DPI_{j+1}$. That means, $\md$ needs to be extended (by a hitting set of all minimal conflict sets in $\minC_{DPI_{j+1}}$ it does not hit) in order to become a diagnosis w.r.t.\ $DPI_{j+1}$. After extending $\md$, both situations might arise, either that $\md$ is a minimal diagnosis w.r.t.\ $DPI_{j+1}$ or that $\md$ is a non-minimal diagnosis w.r.t.\ $DPI_{j+1}$. When the latter case occurs, \textsc{dynamicHS} might often be able to figure out that (the tree branch corresponding to) $\md$ is simply \emph{redundant} (w.r.t.\ the new DPI $DPI_{j+1}$) and does not need to be considered during the further expansion of the hitting set tree (which searches for minimal diagnoses w.r.t.\ $DPI_{j+1}$ and not w.r.t.\ $DPI_{j}$). That is, such redundant tree branches are unnecessary in order to explore all minimal diagnoses w.r.t.\ $DPI_{j+1}$. 

As a consequence, the nice property of \textsc{staticHS} that the set of minimal diagnoses that needs to be taken into account given $DPI_{j+1}$ is a proper subset of the minimal diagnoses set that needed to be considered given $DPI_{j}$ in no longer valid for \textsc{dynamicHS}. That is, the set of remaining solution candidates in \textsc{dynamicHS} is not guaranteed to ``converge'' \emph{constantly} towards a singleton comprising only one solution. The DPI, the minimal conflict sets as well as the minimal diagnoses are ``dynamic''. What holds for both \textsc{dynamicHS} and \textsc{staticHS} is the guarantee that the set of all (i.e.\ minimal and non-minimal) diagnoses is constantly shrinking, i.e.\ $\allD_{DPI_j} \supset \allD_{DPI_{j+1}}$. 

\paragraph{Tree Pruning.} Let $T$ be the hitting set tree produced in the $j$-th iteration of \textsc{dynamicHS} (i.e.\ $T$ is the tree that was used to search for minimal diagnoses w.r.t.\ $DPI_{j}$). Then, after a new test case has been added to $DPI_{j}$, there are often redundant subtrees in $T$ that can be pruned. The resulting tree $T'$ can then be used in the $(j+1)$-th iteration of \textsc{dynamicHS} to identify minimal diagnoses w.r.t.\ the new DPI $DPI_{j+1}$. Using $T$ instead of $T'$ might lead to a significant time and (more severely) space overhead, due to the unnecessary expansion of redundant branches that are known to give no new information at all. Another approach could be to simply discard the entire tree $T$ and start to construct a new one w.r.t.\ $DPI_{j+1}$ from scratch. This strategy, however, will usually also suffer from a non-negligible time overhead since most of the tree $T$ can be safely reused in iteration $j+1$ and only parts of it must be revised. In particular, this strategy would potentially involve many additional calls of $\scQX$ (which internally calls an expensive reasoner) as, in the worst case (when no pruning is possible), the entire existing tree might be rebuilt. 

As Remark~\ref{rem:finding_all_redundant_nodes_makes_algo_inefficient}
and Examples~\ref{example:dynamicHS_small_example_using_tabExDpi2} as well as \ref{example:dynamicHS_large_example_using_tabExDpi3} shall indicate, the overhead in terms of (expensive) calls to a reasoner (i.e.\ calls of $\scQX$) due to tree pruning (compared to its impact on the tree) seems absolutely reasonable. In fact, only one call of a ``fast version'' of $\scQX$ 
might already lead to the deletion of $75\%$ of the tree branches as one can see in the first pruning step in Example~\ref{example:dynamicHS_large_example_using_tabExDpi3}.
%

The evolution of the hitting set tree produced by Algorithm~\ref{algo:inter_onto_debug} using \textsc{dynamicHS} is thus characterized by \emph{alternating expansion and pruning phases}. Also for very complex problems, in case that expansion phases are ``short enough'' such that tree pruning can take place ``often enough'', one might be able to keep the hitting set tree ``small enough'' to handle it efficiently. The extent of the expansion phase can be steered by the specification of the leading diagnosis parameters $n_{\min}$, $n_{\max}$ and $t$ (cf.\ Section~\ref{sec:DetailedAlgorithmDescription}). In the extreme case, these can be defined in a way ($n_{\min}=n_{\max}=2$) the algorithm will allow only the computation of a single further minimal diagnosis (in the first expansion phase: two diagnoses) before \textsc{dynamicHS} (i.e.\ the tree expansion phase) terminates and a further pruning phase might take place. 

However, it is not automatically warranted that tree pruning is possible after each expansion phase. Similarly, no certainty is given that the transition from $DPI_j$ to $DPI_{j+1}$ just causes the deletion of parts of the tree and no additional expansion of the tree. 
In fact, this depends on certain properties of the test case that is added after an expansion phase (i.e.\ properties of the generated query).

\paragraph{Test Cases Affect Tree Pruning.} Some added test case might give rise to some pruning steps as well as it might induce the construction of new subtrees (where ``new'' means that these would be no subtress of a hitting set tree w.r.t.\ the previous DPI $DPI_j$). The latter situation occurs when ``completely new'' minimal conflict sets (see above) are introduced by the addition of a test case. If this is the only impact of a test case, then this test case has only a negative influence on the time and space complexity. In other words, none of the invalidated minimal diagnoses (and no other nodes in the tree) are redundant; but all of them must additionally hit the set of ``completely new'' minimal conflict sets (in order to become diagnoses w.r.t.\ $DPI_{j+1}$). Hence, in this case, the transition from $DPI_j$ to $DPI_{j+1}$ results only in monotonic growth of the tree. If possible, such ``negative-impact test cases'' must be avoided. On the other hand, one must strive for the usage of ``positive-impact test cases'', i.e.\ those that only trigger tree pruning, but no tree expansion.  
Defining and studying properties that constitute such ``positive-impact test cases'' and developing specialized algorithms for extracting exactly those types of queries that enable as substantial and effective pruning as possible is a topic of future research. 

An idea pertinent to this issue could for example be to attempt to extract a query by means of the conflict set $\mc$ that labels the root node of the tree. More concretely, if any answer to a query yields a new test case that leads to the introduction of a minimal conflict set that is a proper subset of $\mc$, then it is for sure that significant pruning can take place (since \emph{entire subtrees starting from the root of the tree} can be deleted). For instance, the first query $Q_1$ in Example~\ref{example:dynamicHS_large_example_using_tabExDpi3} features this property. Roughly, the reasons for that are that $Q_1$ is an entailment of a proper subset $\mc_{sub}$ of $\mc$ (i.e.\ $\mc_{sub}$ is a justification of $Q_1$, cf.\ Section~\ref{sec:ConflictSetsVersusJustifications}) and $Q_1$ 
is ``relevant'' for this conflict set $\mc$ to be a conflict set. In other words, the latter means that $Q_1$ can be used to ``replace'' the part $\mc_{sub}$ of $\mc$, i.e.\ $(\mc \setminus \mc_{sub}) \cup Q_1$ is invalid w.r.t.\ the given DPI. 
That is, addition of $Q_1$ to the positive test cases asserts the correctness of one part of $\mc$, namely $\mc_{sub}$ (cf.\ Example~\ref{example:dynamicHS_large_example_using_tabExDpi3}), wherefore the other part must be incorrect (because some part of a conflict set must be definitely incorrect). On the other hand, assignment of $Q_1$ to the negative test cases asserts exactly the incorrectness of $\mc_{sub}$ wherefore the formulas $\mc \setminus \mc_{sub}$ become obsolete in the minimal conflict set $\mc$ yielding the new minimal conflict set $\mc' := \mc_{sub}$.
Another desirable property of $Q_1$ is that addition of $Q_1$ to either set of test cases does not imply the origination of any ``completely new'' conflict sets (see above) which result in additional growth of the tree.

That is, in its original form (without assuring only the usage of ``positive-impact test cases''), the time and
space complexity of \textsc{dynamicHS} is a function of the generated queries. There is a potential to perform significant pruning, but also the risk of significant tree growth. In case mostly ``positive-impact queries'' are generated and asked to the user, the performance might be very nice and significantly superior to the one of \textsc{staticHS}. In the reverse case, the performance might be also worse than the one of \textsc{staticHS}. In the case of \textsc{staticHS}, there is no chance for significant pruning, but also no chance for a tree growth that goes beyond the size of the non-interactive tree produced by \textsc{HS}.  

In \textsc{staticHS}, there are only expansion phases (in case the tree pruning described by Definition~\ref{def:pruned_hs_tree} is considered part of an expansion phase) which means that the tree constructed by \textsc{staticHS} will constantly grow (apart from the deleted duplicate nodes and non-minimal diagnoses). All the user can do is hope that Algorithm~\ref{algo:inter_onto_debug} applying \textsc{staticHS} will not run out of memory (cf.\ Section~\ref{sec:TheIntuition}). 

The idea is now to be able to use \textsc{dynamicHS} instead of \textsc{staticHS} particularly if the latter runs out of memory soon. If the leading diagnosis parameters are specified small enough to prevent the hitting set tree produced during one expansion phase from becoming too large and test cases are not chosen unfavorably, the \textsc{dynamicHS} method should be able to outperform \textsc{staticHS} significantly, as Examples~\ref{example:staticHS_complex_example_using_tabExDpi3} and \ref{example:dynamicHS_large_example_using_tabExDpi3} suggest.

%
%
%
%
%
%
%
%
%
%
%
%
%
%
%
%
%
%
%

\subsection{Algorithm Walkthrough}
\label{sec:DynamicAlgorithmWalkthrough}
\paragraph{Input Parameters.} When \textsc{dynamicHS} (Algorithm~\ref{algo:inter_dyn_hs}) is called for the first time in Algorithm~\ref{algo:inter_onto_debug}, the inputs $\mC_{calc}$, $\mD_{\checkmark}$, $\mD_{\times}$, $\Tp'$ and $\Tn'$ correspond to the empty set and $\Queue = [\emptyset]$ (cf.\ lines~\ref{algoline:inter_onto_debug:var_inst_start}-\ref{algoline:inter_onto_debug:var_inst_end} and \ref{algoline:inter_onto_debug:dynamicHS} in Algorithm~\ref{algo:inter_onto_debug}). Further on, $\mD_{calc}$ is defined to be the empty set at the beginning of each execution of \textsc{dynamicHS}. That is, \textsc{dynamicHS} starts the construction of the hitting set tree from an initial tree consisting of a single unlabeled root node $\emptyset$ ($\in \Queue$). And, all collections that are later returned by \textsc{dynamicHS} in line~\ref{algoline:dyn:return}, except for $\Queue$, are initially empty. Further input arguments are the DPI $\langle\mo,\mb,\Tp,\Tn\rangle_\RQ$ provided as an input to Algorithm~\ref{algo:inter_onto_debug}, the sets of positively ($\Tp'$) and negatively ($\Tn'$) answered queries since the start of Algorithm~\ref{algo:inter_onto_debug} (both sets initially empty), the leading diagnosis computation parameters $n_{\min},n_{\max},t$ (see description in Section~\ref{sec:UserInteraction} on page~\pageref{etc:leading_diag_params}) and the probability measure $p() := p_{\mo}()$ that assigns a probability in the interval $(0,0.5)$ to each formula in $\mo$ (see line~\ref{algoline:inter_onto_debug:getAxiomProbs} in Algorithm~\ref{algo:inter_onto_debug}).

\paragraph{Tree Update during First Iteration of \textsc{dynamicHS}.} Before the repeat-loop in \textsc{dynamicHS} is entered, the \textsc{updateTree} function is called (line~\ref{algoline:dyn:update_tree}), but has no effect. This holds since \textsc{updateTree} first iterates over all elements in $\mD_{\times}$, then over all elements in $\mD_{\supset}$ and finally over all elements in $\mD_{\checkmark}$ where $\mD_{\times}=\mD_{\supset}=\mD_{\checkmark}=\emptyset$, as pointed out before.

\paragraph{The Main Loop.} During the repeat-loop, in each iteration the first node $\mathsf{node}$ in the queue $\Queue$ of open (non-labeled) nodes is processed (\textsc{getFirst}, line~\ref{algoline:dyn:get_first}). Notice that, anywhere throughout \textsc{dynamicHS}, nodes are added to $\Queue$ in a way that a sorting of $\Queue$ in descending order according to $p_{nodes}()$ (cf.\ Definition~\ref{def:p_node()}) is maintained (cf.\ \textsc{insertSorted} in lines~\ref{algoline:static:generate_nodes}, \ref{algoline:update:insert_sorted_0}, \ref{algoline:update:insert_sorted_0.5}, \ref{algoline:update:insert_sorted_1}, \ref{algoline:prune:insert_alternative_equal_node_into_S'} and \ref{algoline:prune:insert_same_node_into_S'}). Hence, the most probable node (according to $p_{nodes}()$) is always processed next.

So, when $\mathsf{node}$ is processed, it is first deleted from $\Queue$ (\textsc{deleteFirst}, line~\ref{algoline:dyn:delete_from_queue}). Then a test is performed whether $\mathsf{node} \in \mD_{\checkmark}$, i.e.\ whether $\mathsf{node}$ is already known to be a minimal diagnosis w.r.t.\ the current DPI $\langle\mo,\mb,\Tp\cup\Tp',\Tn\cup\Tn'\rangle_\RQ$. In case this test is positive, $\mathsf{node}$ is directly added to $\mD_{calc}$, the set of leading diagnoses that will be output by the current call of \textsc{dynamicHS}. Otherwise, the \textsc{dLabel} function is called given $\mathsf{node}$ (i.a.) as a parameter (line~\ref{algoline:dyn:dlabel}). 

\paragraph{Computation of a Node Label.} The \textsc{dLabel} function processes $\mathsf{node}$ as follows. First, the \emph{non-minimality criterion} (lines~\ref{algoline:dlabel:non-min_crit_start}-\ref{algoline:dlabel:non-min_crit_end}) is checked. That is, among all nodes in $\mD_{calc}$, one is searched which is a proper subset of $\mathsf{node}$. If such a node $\mathsf{nd}$ is found, then $\mathsf{node}$ must be a non-minimal diagnosis w.r.t.\ the current DPI since, anytime throughout the execution of \textsc{dynamicHS}, 
$\mD_{calc}$ contains only minimal diagnoses w.r.t.\ the current DPI $\langle\mo,\mb,\Tp\cup\Tp',\Tn\cup\Tn'\rangle_\RQ$. 
In this case, unlike in \textsc{staticHS}, the branch in the hitting set tree corresponding to $\mathsf{node}$ cannot be simply discarded, but needs to be still stored (in the set $\mD_{\supset}$). It is necessary to store non-minimal diagnoses as these might become minimal diagnoses w.r.t.\ the new DPI obtained after the subsequent addition of a new test case to the current DPI. 

In case the non-minimality criterion is not satisfied, the \emph{reuse criterion} (lines~\ref{algoline:dlabel:reuse_start}-\ref{algoline:dlabel:reuse_end}) is checked next. That is, the set $\mC_{calc}$ containing (not necessarily minimal) conflict sets w.r.t.\ the current DPI is browsed for a set $\mc$ such that $\mc$ and $\mathsf{node}$ are disjoint sets. If such a set $\mc$ is found, there must be some set $X \subseteq \mc$ which is a \emph{minimal} conflict set w.r.t.\ the current DPI. This minimal conflict set $X$ can then be 
used to label $\mathsf{node}$ since the set of edge labels along the path in the tree leading from the root node to $\mathsf{node}$ does not hit $X$ (because it does not hit $\mc$). 

The minimality of $\mc$ is verified by a call of $\scQX(\langle\mc,\mb,\Tp\cup\Tp',\Tn\cup\Tn'\rangle_\RQ)$ that yields $X$, a minimal conflict set w.r.t.\ the current DPI. 
In case $X \subset \mc$ (line~\ref{algoline:dlabel:if_X=C}), before $X$ is returned as a 
label for $\mathsf{node}$, the following tree pruning steps are performed:
\begin{itemize}
	\item All the conflict sets $\mc_i$ used as node labels in the hitting set tree or in duplicate tree branches so far (i.e.\ $\mc_i \in \mathsf{nd.cs}$ for a node $\mathsf{nd} \in \Queue \cup \mD_{\supset} \cup \Queue_{dup}$) such that $X \subset \mc_i$ are replaced by $X$ (\textsc{pruneQdup} and \textsc{prune} in lines~\ref{algoline:dlabel:call_prune_Qdup}-\ref{algoline:dlabel:call_prune_Dsupset}),
	\item any subtree is pruned if its root node is linked to a node now labeled by $X$ (replacing some $\mc_i \supset X$) by an edge with label $\tax$ where $\tax$ is in $\mc_i \setminus X$ (\textsc{pruneQdup} and \textsc{prune} in lines~\ref{algoline:dlabel:call_prune_Qdup}-\ref{algoline:dlabel:call_prune_Dsupset}) and
	\item for each pruned node $\mathsf{nd}$, if there is a non-pruned node in $\Queue_{dup}$ suited to construct a node $\mathsf{nd}'$ that can replace $\mathsf{nd}$, $\mathsf{nd}'$ is added to the collection of nodes from which $\mathsf{nd}$ was deleted (\textsc{pruneQdup} and \textsc{prune} in lines~\ref{algoline:dlabel:call_prune_Qdup}-\ref{algoline:dlabel:call_prune_Dsupset}),
	%
	\item all the conflict sets $\mc_i \in \mC_{calc}$ that are proper supersets of $X$ are deleted from $\mC_{calc}$ and $X$ is added to $\mC_{calc}$ (\textsc{addSetDelSupsets} in line~\ref{algoline:dlabel:add_set_del_supset}). 
\end{itemize}
Otherwise, $\mc$ ($= X$) is directly returned by \textsc{dLabel} without performing any tree pruning because the reused conflict set $\mc$ is (still) a \emph{minimal} conflict set w.r.t.\ the current DPI $\langle\mo,\mb,\Tp\cup\Tp',\Tn\cup\Tn'\rangle_\RQ$ (notice that each element of $\mC_{calc}$ was added to $\mC_{calc}$ as a minimal conflict set w.r.t.\ some DPI $\langle\mo,\mb,\Tp\cup\Tp'',\Tn\cup\Tn''\rangle_\RQ$ where $\Tp'' \subseteq \Tp'$ and $\Tn'' \subseteq \Tn'$ during the execution of this or a previous call of \textsc{dynamicHS}). 

\begin{remark}\label{rem:no_tree_pruning_in_first_iteration_of_dynamicHS}
During the execution of the first call of \textsc{dynamicHS} in Algorithm~\ref{algo:inter_onto_debug}, no tree pruning can take place (neither within the scope of \textsc{dLabel} nor anywhere else) since all elements of $\mC_{calc}$ (initially the empty set) must be minimal conflict sets w.r.t.\ the input DPI which is at the same time the current DPI. Pruning of the hitting set tree is only possible in case some non-leaf nodes of the tree are labeled by conflict sets that are \emph{not minimal} w.r.t.\ the current DPI.\qed
\end{remark}

Given that the reuse criterion fails, $\scQX$ is called given the \emph{current} DPI $\langle\mo\setminus\mathsf{node},\mb,\Tp\cup\Tp',\Tn\cup\Tn'\rangle_\RQ$ as an argument (line~\ref{algoline:dlabel:qx_2}). If the output $L$ is equal to 'no conflict', then we know by Proposition~\ref{prop:qx_correctness} that $\mathsf{node}$ is a diagnosis w.r.t.\ the current DPI, wherefore the label $valid$ is returned for $\mathsf{node}$. Otherwise, the output $L$ must be a minimal conflict set w.r.t.\ $\langle\mo,\mb,\Tp\cup\Tp',\Tn\cup\Tn'\rangle_\RQ$ that has an empty set-intersection with $\mathsf{node}$. Since the reuse criterion failed, i.e.\ there is no set in $\mC_{calc}$ that does not intersect with $\mathsf{node}$, $L$ must be a fresh minimal conflict set w.r.t.\ $\langle\mo,\mb,\Tp\cup\Tp',\Tn\cup\Tn'\rangle_\RQ$ in the sense that $L \notin \mC_{calc}$ must hold. Therefore the label $L$ is first added to $\mC_{calc}$ and then returned by \textsc{dLabel} as a label for $\mathsf{node}$. 

\begin{remark}\label{rem:qx_to_compute_min_cs_key_difference_between_staticHS_and_dynamicHS}
Please notice that this call of $\scQX$ to label a node is one of the key differences between \textsc{staticHS} and \textsc{dynamicHS}. Whereas the former uses $\scQX$ exclusively for the computation of minimal conflict sets w.r.t.\ the (static) input DPI exploiting just the initial sets of positive and negative test cases $\Tp$ and $\Tn$, respectively, the latter employs $\scQX$ to compute minimal conflict sets w.r.t.\ the (dynamic) current DPI which includes all new test cases ($\Tp'$ and $\Tn'$) resulting from answered queries in the ongoing interactive debugging session so far.\qed
\end{remark}

\paragraph{Processing of a Node Label.} Back in the main procedure, the label $L$ returned by the \textsc{dLabel} function is processed as follows. If $L = valid$, then it is a fact that $\mathsf{node}$ is a minimal diagnosis w.r.t.\ the current DPI 
wherefore $\mathsf{node}$ is added to the set $\mD_{calc}$. Otherwise, if $nonmin$ is the returned label for $\mathsf{node}$, $\mathsf{node}$ is added to the set $\mD_{\supset}$ of non-minimal diagnoses w.r.t.\ the current DPI. Otherwise, i.e.\ if $L \notin \setof{valid,nonmin}$, then $L$ must be a minimal conflict set w.r.t.\ the current DPI (see the description of node label computation above). In this case, $|L|$ successor nodes of $\mathsf{node}$ are generated (lines~\ref{algoline:dyn:add_ax_to_node} and \ref{algoline:dyn:add_cs_to_node.cs}). For each logical formula $e \in L$, a new node is computed from $\mathsf{node}$ (and $\mathsf{node.cs}$) as $\mathsf{node}_e := \textsc{add}(\mathsf{node},e)$ and $\mathsf{node}_e.\mathsf{cs} := \textsc{add}(\mathsf{node.cs},L)$ which means that $e$ is appended to the end of the list $\mathsf{node}$ and $L$ is appended to the end of the list $\mathsf{node.cs}$.

If there is already a node $\mathsf{nd}\in\Queue$ such that $\mathsf{nd} = \mathsf{node}_e$ (line~\ref{algoline:dyn:check_node_already_in_Q}), where '$=$' applied to these lists means that the list $\mathsf{nd}$ \emph{interpreted as a set} is equal to the list $\mathsf{node}_e$ \emph{interpreted as a set}, 
then there is already a branch in the existing tree which includes the same set of edge labels as the new node $\mathsf{node}_e$. Note that the tree branch corresponding to $\mathsf{nd}$ will differ from the one corresponding to $\mathsf{node}_e$ in terms of the order of edge labels or (the order of) the node labels visited when traversed starting from the root node. As it makes no sense to expand two branches with equal sets of edge labels in a hitting set tree (cf.\ rule~\ref{def:pruned_hs_tree:rule6} in Definition~\ref{def:pruned_hs_tree}) for time and space complexity reasons and the fact that the sought diagnoses are sets -- and not lists -- of edge labels in the tree, such a duplicate node $\mathsf{node}_e$ is stored in the separate list $\Queue_{dup}$. This list $\Queue_{dup}$ is always kept sorted by ascending node-cardinality (\textsc{insertSorted} in line~\ref{algoline:dyn:add_to_Qdup}). 

The purpose of storing and not deleting such nodes is the possibility that the now ``active'' branch $\mathsf{nd}$ might be pruned after the addition of some test case whereas $\mathsf{node}_e$ might be unaffected by that pruning step. In this case, $\mathsf{node}_e$, given it meets certain properties (see Algorithm~\ref{algo:prune}), can be reactivated and incorporated into the tree in order to replace $\mathsf{nd}$. Had $\mathsf{node}_e$ just been discarded instead of being stored, the completeness of Algorithm~\ref{algo:inter_onto_debug} with $mode = dynamic$ would be violated in general. That is, we would not have any guarantee that all minimal diagnoses w.r.t.\ the current DPI are actually explored by the algorithm.

Otherwise, if there is no node in $\Queue$ that is set-equal to $\mathsf{node}_e$, then $\mathsf{node}_e$ is added to the $k$-th position in $\Queue$ (\textsc{insertSorted} in line~\ref{algoline:dyn:generate_nodes}) if there are (exactly) $k-1$ nodes in $\Queue$ that have a probability as per $p_{nodes}()$ that is greater than or equal to $p_{nodes}(\mathsf{node}_e)$.  

\paragraph{Stop Criterion.} The repeat-loop of \textsc{dynamicHS} is executed until the stop criterion in line~\ref{algoline:dyn:until} is satisfied. The first criterion causing \textsc{dynamicHS} to terminate is $\Queue = []$ which means that the complete hitting set tree has been constructed and no further nodes can be labeled. In this case, $\mD_{calc}$ comprises all minimal diagnoses w.r.t.\ the current DPI $\langle\mo,\mb,\Tp\cup\Tp',\Tn\cup\Tn'\rangle_\RQ$. 

If the first criterion is not met, then the second criterion is checked. That is, a test is performed which checks first whether there is at least one \emph{new} diagnosis w.r.t.\ the current DPI in $\mD_{calc}$ which was not returned by the last-but-one call of \textsc{dynamicHS} (i.e.\ which is not an element of $\mD_{\checkmark}$). Notice that this criterion or $\Queue = []$ will be definitely met after finite execution time of \textsc{dynamicHS} since either new nodes in $\Queue$ will be processed (and labeled) until there is some new diagnosis w.r.t.\ the current DPI identified or the $\Queue$ will become empty.

Additionally, the second criterion involves a test that checks whether the cardinality of $\mD_{calc}$ amounts to at least $n_{\min}$ and either $|\mD_{calc}| = n_{\max}$ or more than $t$ time has passed since the start of the execution of \textsc{dynamicHS}. In the latter case, $n_{\min} \leq |\mD_{calc}| < n_{\max}$ holds. In the former case, $|\mD_{calc}| = n_{\max}$ is satisfied.

\paragraph{Processing of the Leading Diagnoses Returned by \textsc{dynamicHS}.} When a call of \textsc{dynamicHS} in Algorithm~\ref{algo:inter_onto_debug} returns $\tuple{\mD_{calc} ,\Queue, \mathbf{C}_{calc}, \mD_{\times}, \mD_{\supset}, \Queue_{dup}}$, the set $\mD_{calc}$ is stored in the variable $\mD_{\checkmark}$ in Algorithm~\ref{algo:inter_onto_debug}. Between two successive calls of \textsc{dynamicHS} in Algorithm~\ref{algo:inter_onto_debug}, only this set $\mD_{\checkmark}$ as well as $\mD_{\times}$ are modified. The collections $\Queue$, $\mC_{calc}$, $\mD_{\supset}$ as well as $\Queue_{dup}$ remain unchanged until they are used as input parameters when it comes to the next call of \textsc{dynamicHS} in Algorithm~\ref{algo:inter_onto_debug}.

In case one diagnosis $\md_{\max}$ of the current leading diagnoses in $\mD_{\checkmark}$ has a probability greater than or equal to $1 - \sigma$ as per the probability measure $p_{\mD}()$ (see Section~\ref{sec:DetailedAlgorithmDescription}), the stop criterion of interactive KB debugging is met and the solution KB $(\mo\setminus\md_{\max}) \cup U_{\Tp \cup \Tp'}$ w.r.t.\ the current DPI $\langle\mo,\mb,\Tp\cup\Tp',\Tn\cup\Tn'\rangle_\RQ$ 
is returned to the user (\textsc{getSolKB} in line~\ref{algoline:inter_onto_debug:return}, cf.\ Section~\ref{sec:DetailedAlgorithmDescription}). Thereafter, Algorithm~\ref{algo:inter_onto_debug} terminates and no more calls of \textsc{dynamicHS} take place.

Otherwise, if no leading diagnosis satisfies the stop criterion, a query $Q$ together with its q-partition $\Pt(Q)$ is computed as has been detailed in Sections~\ref{sec:QueryGeneration} and \ref{sec:DetailedAlgorithmDescription}. An answer $u(Q)$ to this query is submitted by the interacting user (line~\ref{algoline:inter_onto_debug:user_interaction} in Algorithm~\ref{algo:inter_onto_debug}). Then $u(Q)$ along with $\Pt(Q)$ is exploited to figure out the subset $\mD_{out}$ of $\mD_{\checkmark}$ that does not comply with $u(Q)$. This set $\mD_{out}$ is then deleted from $\mD_{\checkmark}$ and added to $\mD_{\times}$. Additionally, $Q$ is added to the positive test cases $\Tp'$ if $u(Q) = \true$ and to the negative test cases $\Tn'$ otherwise. Subsequently, \textsc{dynamicHS} is called again given 
\begin{itemize}
	\item the \emph{updated parameters} $\mD_{\checkmark}$, $\mD_{\times}$, $\Tp'$ and $\Tn'$ (which are modified within and outside of \textsc{dynamicHS} during the execution of Algorithm~\ref{algo:inter_onto_debug}),
	\item the \emph{unchanged parameters} $\Queue$, $\mC_{calc}$, $\mD_{\supset}$ and $\Queue_{dup}$ (which are modified only within \textsc{dynamicHS} during the execution of Algorithm~\ref{algo:inter_onto_debug}) and
	\item the \emph{constant parameters} $\langle\mo,\mb,\Tp,\Tn\rangle_\RQ$, $t$, $n_{\min}$, $n_{\max}$ and $p_{\mo}()$ (which are not modified within or outside of \textsc{dynamicHS} during the execution of Algorithm~\ref{algo:inter_onto_debug}).
\end{itemize}
The execution of this next and any subsequent call to \textsc{dynamicHS} runs in analogue way as described so far, except for the effect of the \textsc{updateTree} function called at the very beginning of each execution of \textsc{dynamicHS} (recall that the execution of \textsc{updateTree} had no effect during the \emph{first} execution of \textsc{dynamicHS}). We shall now explicate how this function works in all other executions of \textsc{dynamicHS}, except for the first one.

\paragraph{Tree Update.} Between line~\ref{algoline:update:process_Dtimes_start} and line~\ref{algoline:update:process_Dtimes_end}, \textsc{updateTree} goes through all nodes $\mathsf{nd}\in\mD_{\times}$ (recall that $\mD_{\times}$ includes exactly these diagnoses that have been ruled out by the most recently answered query) and first performs the \emph{Quick Redundancy Check} (QRC, lines~\ref{algoline:update:qx}-\ref{algoline:update:quickPC_end}) for $\mathsf{nd}$. If the QRC is not successful, it additionally performs the \emph{Complete Redundancy Check} (CRC, lines~\ref{algoline:update:completePC_start}-\ref{algoline:update:completePC_end}) for $\mathsf{nd}$. 

The QRC 
aims at identifying whether $\mathsf{nd}$ is redundant and can be pruned, i.e.\ it attempts to find a witness of redundancy of $\mathsf{nd}$. Informally, a \emph{redundant node} in (redundant subtree of) the tree is a node (subtree) such that the further expansion of the current tree without this node (subtree) still yields to the detection of all minimal diagnoses w.r.t.\ the current DPI. A \emph{witness of redundancy of $\mathsf{nd}$} is a minimal conflict set $\mc'$ w.r.t.\ the current DPI such that a superset $\mc \supset \mc'$ was used as a node label on the tree path $\mathsf{nd}$ represents (that is, 
there is some $i \leq |\mathsf{nd.cs}|$ such that $\mc$ is the $i$-th element of $\mathsf{nd.cs}$, i.e.\ $\mc = \mathsf{nd.cs}[i]$)
and the label ($\mathsf{nd}[i]$) of the outgoing edge of $\mc$ on the path represented by $\mathsf{nd}$ is an element not in $\mc'$ (that is, an element in $\mc\setminus \mc'$). 

To this end, the QRC involves the call of $\scQX(\tuple{U_{\mathsf{nd.cs}}\setminus \mathsf{nd},\mb,\Tp\cup\Tp',\Tn\cup\Tn'}_\RQ)$ which returns $X$. If $X$ is a set (and not 'no conflict'), then $X$ is a minimal conflict set w.r.t.\ the current DPI $\langle\mo,\mb,\Tp\cup\Tp',\Tn\cup\Tn'\rangle_\RQ$ (as $U_{\mathsf{nd.cs}}\setminus \mathsf{nd} \subseteq \mo$, cf.\ Proposition~\ref{prop:qx_correctness}). To check if $X$ is in fact a witness of redundancy of $\mathsf{nd}$, $X \subset \mc$ (line~\ref{algoline:update:X_subset_C_(QRC)}) is tested for all $\mc \in \mathsf{nd.cs}$. If such a $\mc$ is located, $X$ is a witness of redundancy of $\mathsf{nd}$ and the QRC is successful (expressed by $quickRC \gets \true$ in line~\ref{algoline:update:qrc_gets_true}). In this case, the execution is resumed at line~\ref{algoline:update:if_qrc_or_crc_true}.

The QRC bears its name due to the fact that it requires \emph{at most one call of $\scQX$} (which internally performs expensive calls to a reasoner). Moreover, it passes to $\scQX$ a (DPI including a) KB of a size that is generally significantly smaller than $|\mo|$ where $|\mo|$ is roughly the size of the KB used in the (more expensive) calls of $\scQX$ made in the \textsc{dLabel} function. Hence, the QRC will be usually very fast (cf.\ Proposition~\ref{prop:qx_complexity}).

Otherwise, since the negative outcome of the QRC (which is sound, but not complete w.r.t.\ the finding of a witness of redundancy of $\mathsf{nd}$) does not imply the non-existence of a witness of redundancy of $\mathsf{nd}$, the CRC 
must be performed. As the name already suggests, the CRC is sound \emph{and complete} and will therefore be positive and yield a witness of redundancy if and only if there is some. The CRC involves multiple calls of $\scQX(\tuple{\mathsf{nd.cs}[i]\setminus \setof{\mathsf{nd}[i]},\mb,\Tp\cup\Tp',\Tn\cup\Tn'}_\RQ)$, one for each conflict set $\mathsf{nd.cs}[i]$ in $\mathsf{nd.cs}$. 
It is straightforward from the characterization of a witness of redundancy given before that, given the CRC returns a set $X$, $X$ is a witness of redundancy of $\mathsf{nd}$. 

If $\mathsf{nd}$ is non-redundant, there cannot be any witness of redundancy of $\mathsf{nd}$. Hence, the complete and sound method CRC will not find such a one. Therefore, $quickRC = \false$ and $completeRC = \false$ must hold in line~\ref{algoline:update:if_qrc_or_crc_true}. In this case, the for-loop in line~\ref{algoline:update:process_Dtimes_start} continues with the next node in $\mD_{\times}$.

On the other hand, if $\mathsf{nd}$ is redundant, due to the completeness of CRC, either $quickRC = \true$ or $completeRC = \true$ must hold when it comes to the execution of the if-statement in line~\ref{algoline:update:if_qrc_or_crc_true}. At this point, it is guaranteed that the variable $X$ stores a witness of redundancy of $\mathsf{nd}$.

The CRC, contrary to the QRC, generally requires \emph{multiple} (at most $|\mathsf{nd}|$) \emph{calls of $\scQX$} (which internally performs expensive calls to a reasoner). But, like the QRC, it passes to $\scQX$ a (DPI including a) KB of a size that is generally significantly smaller than $|\mo|$. Furthermore, at most one call of $\scQX$ will involve more than one call of \textsc{isKBValid} (see Algorithm~\ref{algo:qx}), i.e.\ the function that calls the reasoner. This must be true since CRC only requires an additional call of $\scQX$ if a witness of redundancy has not yet been found. And, each call of $\scQX$ that does not find a witness of redundancy of $\mathsf{nd}$ returns 'no conflict' which necessitates only a single invocation of \textsc{isKBValid}. Hence, each execution of the CRC will be very fast in general as well (cf.\ Proposition~\ref{prop:qx_complexity}).

What comes next is the pruning of all redundant nodes in the tree for which $X$ is a witness of redundancy. Essentially, the same pruning steps are performed here as in the \emph{reuse criterion} described in 'Computation of a node label' above. 

Notice that a redundant node is guaranteed to be a redundant node in any further iteration of \textsc{dynamicHS} (using a new current DPI that incorporates new test cases). 
So, nodes pruned by \textsc{prune} or \textsc{pruneQdup} can be deleted for good and do not need to be stored any longer. Moreover, it should be noted that only redundant nodes are pruned at any pruning step in \textsc{dynamicHS}. For, as long as a node in \textsc{dynamicHS} is not known to be redundant, 
some successor node of this node might be a minimal diagnosis w.r.t.\ the current DPI. Thus, the deletion of such a node could perhaps prevent the algorithm from finding a particular minimal diagnosis which would implicate the algorithm's incompleteness. 

\begin{remark}\label{rem:Dtimes_might_change_in_for_loop_in_updateTree} 
Since the removal of a node from a collection $S \in \setof{\mD_{\times},\Queue,\Queue_{dup},\mD_{\supset}}$ within the scope of \textsc{prune} or \textsc{pruneQdup} can be followed by the re-addition to $S$ of a suitable duplicate node constructed from a node stored in $\Queue_{dup}$, 
$\mD_{\times}$ might be changed both in that nodes are deleted from it and added to it during the for-loop (line~\ref{algoline:update:process_Dtimes_start}). Therefore, the '$\mathbf{for}\; \mathsf{nd}\in\mD_{\times}$'-statement must be read as 'if $\mathsf{nd}$ is a node in the \emph{current} set $\mD_{\times}$ which has not yet been processed'. For a better code readability, we abstained from using a programmatically precise representation of this issue in Algorithm~\ref{algo:update_tree}.\qed 
\end{remark}

Due to the soundness and completeness of QRC paired with CRC concerning the identification of a witness of redundancy for a given node and the accomplished pruning of (at least) all nodes in $\mD_{\times}$ for which a witness of redundancy has been extracted, all nodes that are in $\mD_{\times}$ when the algorithm reaches line~\ref{algoline:update:reinsert_D_of_Dx_to_Q} are \emph{non-redundant} nodes. 
Consequently, there is no evidence to exclude the remaining nodes in $\mD_{\times}$ from the further search for minimal diagnoses. For this reason, each of these nodes is reinserted into $\Queue$ by \textsc{insertSorted} in line~\ref{algoline:update:insert_sorted_0} such that the sorting of $\Queue$ in descending order of $p_{nodes}()$ is maintained. Then these nodes are deleted from $\mD_{\times}$. Thus, $\mD_{\times} = \emptyset$ holds after each execution of \textsc{updateTree}. 

So, in \textsc{dynamicHS}, unlike in \textsc{staticHS}, diagnoses (and nodes in general) are not ruled out due to the fact that they contradict an answered query, but only if they are (found to be) redundant. Nevertheless, a diagnosis that contradicts an answered query is a ``hot candidate'' for finding some witness of redundancy. For that reason, \textsc{updateTree} searches for witnesses of redundancy (only) by means of $\mD_{\times}$ which includes the most ``suspicious'' nodes. Namely, it comprises those nodes that were minimal diagnoses w.r.t.\ the last-but-one DPI, but have been invalidated by the most recently answered query. The two possible reasons for a diagnosis $\mathsf{nd}$ to be invalidated are its redundancy as defined above or that it does not hit a \emph{new} minimal conflict set (which is not a subset of one in $\mathsf{nd.cs}$) that has been introduced by the addition of the test case resulting from the user's query answer. Thus, it is likely to detect witnesses of redundancy by investigating nodes in $\mD_{\times}$, as the QRC and the CRC do. Throughout the pruning steps performed in lines~\ref{algoline:update:call_prune_Qdup}-\ref{algoline:update:call_prune_Dsupset}, witnesses of redundancy extracted from nodes in $\mD_{\times}$ are exploited to remove redundant nodes in the other collections $\Queue_{dup}$, $\mD_{\supset}$ and $\Queue$ as well.

\begin{remark}\label{rem:finding_all_redundant_nodes_makes_algo_inefficient}
It should be noted that the collections $\Queue$ as well as $\mD_{\supset}$ are not necessarily cleaned from all redundant nodes after all pruning steps in \textsc{updateTree} are finished. At this point, all those redundant nodes are still elements of these collections for which no witness of redundancy was found (there might exist one, though) throughout the redundancy checks (QRC and CRC) performed. 

Assuring the non-existence of redundant nodes in $\Queue$ and $\mD_{\supset}$ might involve extensive usage of the (expensive) reasoner. In the worst case, one call of $\scQX$ for each non-leaf node along each path from the root node to a leaf node labeled by $nonmin$ or to a leaf node that has no label would be necessary. However, the number of these non-leaf nodes is generally \emph{exponential} in the maximum length of such a path in the tree. In comparison, the number of calls of $\scQX$ for investigating all nodes in $\mD_{\times}$ by QRC and CRC is \emph{polynomial (linear)} in the maximum length of a tree path labeled by $\times$. For, the number of $\scQX$-calls cannot get larger than $(n_{\max}-1) (|\mathsf{nd}_{\max}|+1)$ where the constant $n_{\max}$ is the maximum number of desired leading diagnoses predefined by the user and $|\mathsf{nd}_{\max}|$ is the maximum cardinality of some $\mathsf{nd} \in \mD_{\times}$. This holds since $|\mD_{\times}| \leq n_{\max} - 1$ (cf.\ Corollary~\ref{cor:query_leaves_valid_diag}) and QRC requires at most one and CRC at most $|\mathsf{nd}_{\max}|$ $\scQX$-calls. 
 
Other than that, the chance of locating new witnesses of redundancy by means of investigating nodes in $\Queue$ and $\mD_{\supset}$ can be assumed to be smaller than for nodes in $\mD_{\times}$ since there is no indication or evidence that these nodes might be redundant. So, cleaning $\Queue$ and $\mD_{\supset}$ from all redundant nodes might be significant effort with negligible impact. Therefore, \textsc{dynamicHS} is designed to focus the search for witnesses of redundancy only on the ``suspicious nodes'' in $\mD_{\times}$.\qed   
%
\end{remark}

As mentioned above, when the execution arrives at line~\ref{algoline:update:process_Dsupset_start}, only nodes that are definitely redundant (because they were deleted due to some witness of redundancy) have been deleted from the sets $\Queue$, $\mD_{\times}$, $\mD_{\supset}$ and $\Queue_{dup}$. 

In lines~\ref{algoline:update:process_Dsupset_start}-\ref{algoline:update:process_Dsupset_end}, each node $\mathsf{nd}\in\mD_{\supset}$ which has not been deleted throughout the pruning operations in line~\ref{algoline:update:call_prune_Dsupset} is processed as follows: If there is no minimal diagnosis $\md\in\mD_{\checkmark}$ such that $\mathsf{nd} \supset \md$, then $\mathsf{nd}$ is removed from $\mD_{\supset}$ and reinserted into $\Queue$ (lines~\ref{algoline:update:insert_sorted_0.5} and \ref{algoline:update:delete_from_Dsupset}) in a way the sorting of $\Queue$ in descending order according to $p_{nodes}()$ is maintained (\textsc{insertSorted}). This re-insertion is plausible since there is no more evidence 
of $\mathsf{nd}$ (which is a non-minimal diagnosis w.r.t.\ the last-but-one DPI) being a non-minimal diagnosis w.r.t.\ the current DPI (non-minimal diagnoses might become minimal diagnoses by the addition of test cases).

Otherwise, $\mathsf{nd}$ remains an element of the set of non-minimal diagnoses $\mD_{\supset}$ w.r.t.\ the current DPI as $\mD_{\checkmark}$ comprises exclusively minimal diagnoses w.r.t.\ the current DPI and one of these is a proper subset of $\mathsf{nd}$. 

In lines~\ref{algoline:update:process_Dcheckmark_start}-\ref{algoline:update:process_Dcheckmark_end}, all elements in $\mD_{\checkmark}$, each of which is a minimal diagnosis w.r.t.\ the current DPI, are added to $\Queue$ in a way the sorting of $\Queue$ in descending order according to $p_{nodes}()$ is maintained. 

\begin{remark}\label{rem:elements_of_Dcheckmark_are_added_to_Q}
Please notice that the elements of $\mD_{\checkmark}$, although they are known to be minimal diagnoses w.r.t.\ the current DPI, are not directly added to the set of found leading diagnoses $\mD_{calc}$ w.r.t.\ the current DPI, but to $\Queue$. The reason for this is that there might be (not-yet-found) minimal diagnoses w.r.t.\ the current DPI (nodes in $\Queue$ or successor nodes thereof) which were not minimal diagnoses w.r.t.\ the last-but-one DPI (and thus are no elements of $\mD_{\checkmark}$) that have a higher probability as per $p_{nodes}()$ than elements of $\mD_{\checkmark}$. For instance, such diagnoses might have been added to $\Queue$ from the set $\mD_{\supset}$ in line~\ref{algoline:update:insert_sorted_0.5}. 

In this way, since always the first (and most probable) node in $\Queue$ is processed next, a guarantee is given that $\mD_{calc}$ always comprises the $|\mD_{calc}|$ most probable minimal diagnoses w.r.t.\ the current DPI as per $p_{nodes}()$. 
The knowledge of the validity of minimal diagnoses in $\mD_{\checkmark}$ w.r.t.\ the current DPI is however not forgotten, but exploited in line~\ref{algoline:dyn:if_L_valid} (i.e.\ no call of \textsc{dLabel} and $\scQX$ is necessary for a node in $\mD_{\checkmark}$ to be added to $\mD_{calc}$), as elucidated in 'The main loop' above.\qed
\end{remark}
The next proposition asserts that \textsc{dynamicHS} works correctly, i.e.\ terminates and yields an output complying with the assertions given in Algorithm~\ref{algo:inter_dyn_hs} (a proof of this proposition is beyond the scope of this work):
\begin{proposition}[Correctness of \textsc{dynamicHS}]\label{prop:dynamic_hs_correctness}
Any call to \textsc{dynamicHS} (given the inputs described in Algorithm~\ref{algo:inter_dyn_hs}) within Algorithm~\ref{algo:inter_onto_debug} terminates and yields an output $\tuple{\mD_{calc},\Queue, \mathbf{C}_{calc}, \mD_{\times}, \mD_{\supset},\Queue_{dup}}$ where
\begin{enumerate}[(1)]
\item $\mD_{calc}$ is the current set of leading diagnoses such that
\begin{enumerate}[(a)]
\item $\mD_{calc} \subseteq \minD_{\langle\mo,\mb,\Tp\cup\Tp',\Tn\cup\Tn'\rangle_\RQ}$ is the set of most probable minimal diagnoses w.r.t.\ $\langle\mo,\mb,\Tp\cup\Tp',\Tn\cup\Tn'\rangle_\RQ$ such that 
\begin{enumerate}[(i)]
\item $n_{\min} \leq |\mD_{calc}| \leq n_{\max}$ and 
\item $\mD_{calc}\setminus\mD_{\checkmark} \neq \emptyset$,
\end{enumerate}
if such a set $\mD_{calc}$ exists;
or 
\item $\mD_{calc}$ is equal to the set of all minimal diagnoses $\minD_{\langle\mo,\mb,\Tp\cup\Tp',\Tn\cup\Tn'\rangle_\RQ}$, otherwise;
\end{enumerate}
where ``most-probable'' refers to the probability measure $p_{nodes}()$ given by Definition~\ref{def:p_node()} and obtained from the function $p()$ given as an input argument to \textsc{dynamicHS}.
\item $\Queue$ is the current queue of open (non-labeled) nodes of the produced hitting set tree,
\item $\mathbf{C}_{calc}$ is a set of conflict sets w.r.t.\ the current DPI $\langle\mo,\mb,\Tp\cup\Tp',\Tn\cup\Tn'\rangle_\RQ$,
\item $\mD_{\times} = \emptyset$, 
\item $\mD_{\supset}$ is the set of all processed nodes so far throughout the execution of Algorithm~\ref{algo:inter_onto_debug} that are non-minimal diagnoses w.r.t.\ the current DPI $\langle\mo,\mb,\Tp\cup\Tp',\Tn\cup\Tn'\rangle_\RQ$ and
\item $\Queue_{dup}$ is the updated set of stored (duplicate) nodes $\mathsf{nd}$ that can be used when it comes to constructing a replacement node of a pruned node $\mathsf{nd}' \supseteq \mathsf{nd}$ after tree pruning. 
\qed
\end{enumerate}
%
%
%
\end{proposition}

\subsection[Examples]{\textsc{dynamicHS}: Examples}
\label{sec:TextscDynamicHSExamples}
In this section we will give two examples of how interactive KB debugging using \textsc{dynamicHS} (Algorithm~\ref{algo:inter_onto_debug} with parameter $mode=dynamic$) works. The first one will show the similarities and differences between the usage of \textsc{dynamicHS} (within Algorithm~\ref{algo:inter_onto_debug}) and \textsc{HS} (within Algorithm~\ref{algo:non_int_debug}) since it will depict the application of \textsc{staticHS} on the same example DPI (see Table~\ref{tab:example2}) that was used to show the functionality of \textsc{HS} in examples~\ref{example:non_interactive_debugging_with_tabExDpi2_and_without_probs} and \ref{example:non_interactive_debugging_with_tabExDpi2_and_probs}. At the same time, the first example will provide evidence that solving the problem of Interactive Dynamic KB Debugging can be less efficient than solving the problem of Interactive Static KB Debugging in terms of the number of query answers required from an interacting user. This will be discussed in more detail in Section~\ref{sec:TextscStaticHSVersusTextscDynamicHS}.

The second example is supposed to deepen the reader's understanding of the way \textsc{dynamicHS} works. To this end, the example DPI provided by Table~\ref{tab:example3} will be used which constitutes a significantly harder (interactive) debugging task than the DPI investigated in the first example. This example will involve the construction of a relatively large hitting set tree in the first iteration of \textsc{dynamicHS} (which behaves very similarly to \textsc{staticHS} as well as \textsc{HS} and constructs the same wpHS-tree as these methods), but will then show the power of the tree pruning that can be exploited in Interactive Dynamic KB Debugging in that the tree will shrink rapidly after the addition of test cases. Hence, this example will emphasize the advantage of the decision to search for a solution of Interactive Dynamic KB Debugging rather than for a solution of Interactive Static KB Debugging (more on that in Section~\ref{sec:TextscStaticHSVersusTextscDynamicHS}).

Notice that, in the following examples, whenever some tuple or list occurs in an expression using set operators, it is interpreted as a set.
\begin{example}\label{example:dynamicHS_small_example_using_tabExDpi2}
In this example we assume that the author (called user throughout this example) of the (admissible) DPI $\langle\mo,\mb,\Tp,\Tn\rangle_\RQ$ given by Table~\ref{tab:example2} applies Algorithm~\ref{algo:inter_onto_debug} with $mode = dynamic$ to interactively debug $\langle\mo,\mb,\Tp,\Tn\rangle_\RQ$. Further, the same scenario and parameter settings as in Example~\ref{example:staticHS_simple_example_using_tabExDpi2} are supposed. That is, $n_{\min} = n_{\max} = 2$ (notice that the time limit $t$ is irrelevant in this case), $q := 1$ (cf.\ Section~\ref{sec:QueryGeneration}), $qsm()$ is equal to any query selection measure described in Section~\ref{sec:query_selection_measures}, $p_{\mo}(\tax) := c < 0.5$ for all $\tax \in \mo$, i.e.\ all formula fault probabilities are specified to be equal (to some constant $c$) and $\sigma := 0$.

The tree constructed and parameters computed and used by Algorithm~\ref{algo:inter_onto_debug} using \textsc{dynamicHS} are visualized by Figures~\ref{fig:example:inter_onto_debug_dynamicHS_TabExDpi2} and \ref{fig:example:inter_onto_debug_dynamicHS_TabExDpi2_continued}. 
We use the same notation as in Figures~\ref{fig:example:non-interactive_onto_debug_auto=false+nmin=infty_and_auto=true}, \ref{fig:example:non-interactive_onto_debug_auto=false+nmin=2+nmax=4_with_probs}, \ref{fig:example:inter_onto_debug_staticHS_TabExDpi2}, \ref{fig:example:inter_onto_debug_staticHS_TabExDpi3} and \ref{fig:example:inter_onto_debug_staticHS_TabExDpi3_continued} which is described in Examples~\ref{example:non_interactive_debugging_with_tabExDpi2_and_without_probs}, \ref{example:non_interactive_debugging_with_tabExDpi2_and_probs}, \ref{example:staticHS_simple_example_using_tabExDpi2} and \ref{example:staticHS_complex_example_using_tabExDpi3}. 
%

In the first iteration, i.e.\ during the execution of the first call of \textsc{dynamicHS} during Algorithm~\ref{algo:inter_onto_debug}, the root node (initially the empty set) is labeled by the minimal conflict set $\tuple{1,2,5}$ w.r.t.\ $\langle\mo,\mb,\Tp,\Tn\rangle_\RQ$ and three successor nodes, namely $\mathsf{nd}_1 := [1]$, $\mathsf{nd}_2 := [2]$ as well as $\mathsf{nd}_3 := [5]$ with $\mathsf{nd}_1.\mathsf{cs} = \mathsf{nd}_2.\mathsf{cs} = \mathsf{nd}_3.\mathsf{cs} = [\tuple{1,2,5}]$, are added to the queue of open nodes $\Queue$. Since all formulas have been assigned an equal fault probability, \textsc{dynamicHS} conducts a breadth-first tree construction (as displayed by the numbers \textcircled{\scriptsize i} that give the order of node labeling). That is, $\Queue$ in this case is a first-in-first-out queue. In this vein, first $[1]$ and then $[2]$ are identified as minimal diagnoses w.r.t.\ the given DPI. 

Since $\mD_{calc} = \setof{[1],[2]}$ has a cardinality of $n_{\min} = n_{\max} = 2$, the stop criterion of \textsc{dynamicHS} causes it to terminate and return $\tuple{\mD_{calc},\Queue,\mC_{calc},\Queue,\mD_{\times},\mD_{\supset},\Queue_{dup}} = \langle\ \setof{[1],[2]}$, $[[5]]$, $\setof{\tuple{1,2,5}}$, $\emptyset,\emptyset,[]\rangle$, as shown in the upper right column in Figure~\ref{fig:example:inter_onto_debug_dynamicHS_TabExDpi2}. 

Then, in Algorithm~\ref{algo:inter_onto_debug}, outside of the \textsc{dynamicHS} procedure, the first query $Q_1 = \setof{E \rightarrow \lnot A}$ is computed from the leading diagnoses set $\setof{[1],[2]}$. The q-partition $\Pt(Q_1)$ associated with $Q_1$ is $\tuple{\setof{[1]},\setof{[2]},\emptyset}$. The user's answer $u(Q_1)$ to $Q_1$ is then $\false$. Thence, the set $\mD_{out}$ is calculated from $\Pt(Q_1)$ as $\dx{}(Q_1) = \setof{[1]}$ (due to negative answer, cf.\ Remark~\ref{rem:invalidated_sets_of_q-partition_for_query_answer}), deleted from $\mD_{\checkmark} := \mD_{\checkmark} \cup \mD_{calc}$ to yield $\mD_{\checkmark} = \setof{[2]}$ and added to $\mD_{\times}$ to yield $\mD_{\times} = \setof{[1]}$.
Now, the set $\mD_{\checkmark}$ corresponds to the set of all computed (i.e.\ added to $\mD_{calc}$) minimal diagnoses w.r.t.\ the last-but-one DPI $\langle\mo,\mb,\Tp,\Tn\rangle_\RQ$ that are minimal diagnoses w.r.t.\ current DPI $\langle\mo,\mb,\Tp,\Tn\cup\setof{Q_1}\rangle_\RQ$, i.e.\ that satisfy the most recently answered query $Q_1$. The set $\mD_{\times}$ comprises all
computed (i.e.\ added to $\mD_{calc}$) minimal diagnoses w.r.t.\ the last-but-one DPI $\langle\mo,\mb,\Tp,\Tn\rangle_\RQ$ that are not minimal diagnoses w.r.t.\ current DPI $\langle\mo,\mb,\Tp,\Tn\cup\setof{Q_1}\rangle_\RQ$, i.e.\ that do not satisfy the most recently answered query $Q_1$. 

These sets $\mD_{\checkmark}$ and $\mD_{\times}$ along with the collections $\Queue$, $\Queue_{dup}$, $\mD_{\supset}$ and $\mC_{calc}$ which are unmodified outside of \textsc{dynamicHS} are used as input arguments for the second call of \textsc{dynamicHS}. Notice that, in Figures~\ref{fig:example:inter_onto_debug_dynamicHS_TabExDpi2} and \ref{fig:example:inter_onto_debug_dynamicHS_TabExDpi2_continued}, the resulting values of operations performed within \textsc{dynamicHS} are given in the righthand column above the dashed line whereas values computed outside of \textsc{dynamicHS} are given below the dashed line. 

The execution of the second call of \textsc{dynamicHS} starts with a call of the \textsc{updateTree} function. The purpose of this function is to transform the hitting set tree $T$ that was constructed by the first call of \textsc{dynamicHS} into an updated hitting set tree $T'$. Whereas the tree $T$ was used to locate minimal diagnoses w.r.t.\ the last-but-one DPI $\langle\mo,\mb,\Tp,\Tn\rangle_\RQ$, the modified tree $T'$ should serve to generate minimal diagnoses w.r.t.\ the current DPI $\langle\mo,\mb,\Tp,\Tn\cup\setof{Q_1}\rangle_\RQ$. The parameters $\mD_{\checkmark}$, $\mD_{\times}$, $\Queue$, $\Queue_{dup}$, $\mD_{\supset}$ and $\mC_{calc}$ that represent the tree $T$ (given at the top of the lefthand column in Figure~\ref{fig:example:inter_onto_debug_dynamicHS_TabExDpi2}), where $\mD_{\checkmark} \cup \mD_{\times}$ is equal to the set $\mD_{calc}$ produced by the first call of \textsc{dynamicHS}, are i.a.\ given as input arguments to the \textsc{updateTree} function. 

As a first step within \textsc{updateTree}, a redundancy check is performed for each diagnosis in $\mD_{\times}$. In this case $\mD_{\times} = \setof{\md_1}$ since $\md_1$ is the only minimal diagnosis that has been ruled out by the most recently added negative test case $Q_1$. The purpose of the redundancy check is to figure out whether $\md_1$ is redundant w.r.t.\ the current DPI and must be pruned or whether it might be extended to become a minimal diagnosis w.r.t.\ the current DPI.
  
First, the Quick Redundancy Check (QRC) $\scQX(\tuple{\setof{2,5},\mb,\Tp,\Tn\cup\setof{Q_1}}_\RQ) = \tuple{2,5}$ (line~\ref{algoline:update:qx} in \textsc{dynamicHS}) is executed for $\md_1$ which detects (line~\ref{algoline:update:X_subset_C_(QRC)} in \textsc{dynamicHS}) that $\md_1$ (and possibly some further nodes) is redundant and can be pruned. This holds since the minimal conflict set $\tuple{1,2,5}$ w.r.t.\ the last-but-one DPI $\langle\mo,\mb,\Tp,\Tn\rangle_\RQ$ is not a minimal conflict set w.r.t.\ the current DPI $\langle\mo,\mb,\Tp,\Tn\cup\setof{Q_1}\rangle_\RQ$ because $\tuple{2,5}$ returned by $\scQX$ is already a minimal conflict set w.r.t.\ the current DPI (cf.\ Proposition~\ref{prop:qx_correctness}). We call the minimal conflict set $\tuple{2,5}$ a \emph{witness of redundancy} for $\md_1$. Hence, all branches in the hitting set tree starting from the outgoing edge of $\tuple{1,2,5}$ labeled by $1$ can be safely deleted from all collections representing the new tree $T'$ (warranted that all minimal diagnoses w.r.t.\ the current DPI can still be generated from the pruned tree $T'$).

Please notice that the QRC involves only a single call of $\scQX$ using a KB of a size (here: 2) that is generally significantly smaller than $|\mo|$ (here: 7) which is roughly the size of the KB used in calls of $\scQX$ made in the \textsc{dLabel} function. Hence, the QRC will be usually very fast.

An illustration why $\tuple{2,5}$ ``replaces'' $\tuple{1,2,5}$ as a minimal conflict set w.r.t.\ the current DPI can be given as follows: First, $\tuple{1,2,5}$ is a minimal conflict set w.r.t.\ $\langle\mo,\mb,\Tp,\Tn\rangle_\RQ$ as it is a set-minimal subset of $\mo$ that entails $\setof{\lnot A} = \tn_1 \in \Tn$, there is no other negative test case in $\Tn$ except for $\tn_1$ and there is no proper subset $\mc'$ of $\tuple{1,2,5}$ where $\mc' \cup \mb \cup U_{\Tp}$ violates any $r \in \RQ$ (see example~\ref{example:analysis_TabExDpi2} for a detailed explanation). Second, formula $2$ implies in particular $E \rightarrow Y$ which, along with formula $5$ ($Y \rightarrow \lnot A$), yields $E \rightarrow \lnot A$. As the negative answer to $Q_1$ is equivalent to postulating that $\setof{E \rightarrow \lnot A}$ must not be entailed by the KB desired by the user, we have that $\tuple{2,5}$ is a conflict set w.r.t.\ $\langle\mo,\mb,\Tp,\Tn\cup\setof{Q_1}\rangle_\RQ$. As neither $\setof{2}$ nor $\setof{5}$ is a invalid KB w.r.t.\ $\langle\cdot,\mb,\Tp,\Tn\cup\setof{Q_1}\rangle_\RQ$ (cf.\ Corollary~\ref{cor:validonto_cs} and Definition~\ref{def:cs}), we have that $\tuple{2,5}$ is a \emph{minimal} conflict set w.r.t.\ $\langle\mo,\mb,\Tp,\Tn\cup\setof{Q_1}\rangle_\RQ$.

Because the QRC has been successful, yielding some witness of redundancy of $\md_1$, the Complete Redundancy Check (CRC) is no more necessary and the collections $\Queue_{dup}$, $\Queue$, $\mD_{\times}$ as well as $\mD_{\supset}$ are processed by the \textsc{prune} and \textsc{pruneQdup} functions, respectively, which involve the removal of all nodes in these collections that are redundant due to the witness $\tuple{2,5}$. In other words, all nodes are eliminated which correspond to a path in the tree that includes a node label $\mc_{old} \supset \tuple{2,5}$ and the label $e$ of the outgoing edge of $\mc_{old}$ on this path is an element of $\mc_{old}\setminus\tuple{2,5}$. Moreover, all the supersets of $\tuple{2,5}$ in $\mC_{calc}$ (here, only $\tuple{1,2,5}$) are replaced by $\tuple{2,5}$ since they are not minimal conflict sets anymore (\textsc{addSetDelSupsets}).

The pruning of nodes is expressed by dashed arrows in the pictures labeled by 'Updated Tree' in Figures~\ref{fig:example:inter_onto_debug_dynamicHS_TabExDpi2} and \ref{fig:example:inter_onto_debug_dynamicHS_TabExDpi2_continued} where the location of cutting a branch is marked by a crossline at the shaft of a dashed arrow. Furthermore, the elements of ``old'' minimal conflict sets that are no more elements of known (i.e.\ already computed) current minimal conflict sets are crossed out. As shown by the picture 'Updated Tree' in the righthand column of Figure~\ref{fig:example:inter_onto_debug_dynamicHS_TabExDpi2}, $\md_1$ is the only removed node during the pruning steps using the witness of redundancy $\tuple{2,5}$. 

Since $\mD_{\supset} = \emptyset$, \textsc{updateTree} directly jumps to the last three lines where all elements of $\mD_{\checkmark}$ are re-added to $\Queue$ in sorted order (but at the same time remain elements of $\mD_{\checkmark}$). In the figure, this is displayed by the $\stackrel{Q_1}{\Longrightarrow}$ pointing to a question mark (which stands for an open node) instead of a checkmark as in the case of the \textsc{staticHS} algorithm. Notice that, although it is a fact that all elements of $\mD_{\checkmark}$ are minimal diagnoses w.r.t.\ the current DPI, this step is necessary in order to make sure the set $\mD_{calc}$ returned by any call of \textsc{dynamicHS} actually comprises the $|\mD_{calc}|$ most probable minimal diagnoses w.r.t.\ the current DPI. For, there might be, for instance, some node that is a non-minimal diagnosis w.r.t.\ the last-but-one DPI (and is thus not an element of $\mD_{\checkmark}$), but becomes a minimal diagnosis w.r.t.\ the current DPI and has a higher probability than some node in $\mD_{\checkmark}$. Additionally, we want to point out that no calls of the \textsc{dLabel} procedure are needed for diagnoses in $\mD_{\checkmark}$ as we know their label must be $valid$. This is reflected by the test in line~\ref{algoline:dyn:node_in_Dcheckmark} in \textsc{dynamicHS}.

In the figure, all the updated collections $\mD_{\supset}$, $\mC_{calc}$, $\Queue$ as well as $\Queue_{dup}$, after being processed by \textsc{updateTree} are shown at the bottom of fields labeled by \textsc{updateTree}. We want to remark that $\mD_{\times}$ is always the empty set at the end of the execution of \textsc{updateTree} since each node in $\mD_{\times}$ gets either pruned or is reinserted into $\Queue$ as an open node. These updated collections represent the new pruned hitting set tree that can be further constructed in order to detect all and only minimal diagnoses w.r.t.\ the current DPI $\langle\mo,\mb,\Tp,\Tn\cup\setof{Q_1}\rangle_\RQ$. Note that the actions carried out by \textsc{updateTree} take place between steps \textcircled{\scriptsize 4} and \textcircled{\scriptsize 5}.

The expansion of this tree during the repeat-loop in \textsc{dynamicHS} is depicted by the picture named 'Iteration 2' in Figure~\ref{fig:example:inter_onto_debug_dynamicHS_TabExDpi2}. Namely, first (step \textcircled{\scriptsize 5}) the node $[2]$ is directly labeled by $valid$ (line~\ref{algoline:dyn:node_in_Dcheckmark}) since it is a known minimal diagnosis w.r.t.\ the current DPI (as explained before). In the sixth step, $[5]$ is labeled by the minimal conflict set $\tuple{1,2,7}$ w.r.t.\ the current DPI and three further nodes ($[5,1]$, $[5,2]$ and $[5,7]$, all with $\mathsf{nd.cs} = [\tuple{2,5}, \tuple{1,2,7}]$) are generated as successor nodes of $[5]$ and are added to $\Queue$. Now, $[5,1]$ (first-in-first-out) is the foremost node in $\Queue$ and is thus processed next and found to be a minimal diagnosis w.r.t.\ the current DPI. Therefore, \textsc{dynamicHS} terminates and returns i.a.\ the new set of leading diagnoses $\mD_{calc} = \setof{[2],[5,1]}$.

Please notice the difference here to Example~\ref{example:staticHS_simple_example_using_tabExDpi2} where the node $\setof{5,1}$ never became part of $\Queue$ in \textsc{staticHS} due to the existence of a minimal diagnosis $[1]$ w.r.t.\ the input DPI $\langle\mo,\mb,\Tp,\Tn\rangle_\RQ$ which is a proper subset of this node (and due to the fact that \textsc{staticHS} must only consider minimal diagnoses \emph{w.r.t.\ the input DPI}). In the current example, this node can only become relevant \emph{w.r.t.\ the current DPI} if all (known) diagnoses (here, only $[1]$) that are proper subsets of it have already been pruned. It should now be clear to the reader why non-minimal nodes cannot be deleted for good as in \textsc{staticHS} and why the set $\mD_{\supset}$ is necessary in \textsc{dynamicHS}.

This leading diagnosis $[5,1]$ is also the reason why the second query $Q_2 = \setof{E \rightarrow G}$ is different from the second query ($Y \rightarrow \lnot A$) calculated in Example~\ref{example:staticHS_simple_example_using_tabExDpi2}.

The execution of the algorithm continues in an analogue manner as explained so far. In the following, we just want to explain some interesting aspects in the rest of its execution:
\begin{itemize}
	\item After the query $Q_3 = \setof{Y \rightarrow \lnot A}$ (the same query as the second query in Example~\ref{example:staticHS_simple_example_using_tabExDpi2}) is answered negatively and $Q_3$ is added to $\Tn'$ yielding the current DPI $\langle\mo,\mb,\Tp,\Tn\cup\setof{Q_1,Q_2,Q_3}\rangle_\RQ$, the \textsc{updateTree} function not only prunes $[2] = \md_2 \in \mD_{\times}$ and adds $[5,7] = \md_4 \in \mD_{\checkmark}$ to $\Queue$ as we delineated above for the first query $Q_1$, but adds $[5,2] \in \mD_{\supset}$ to $\Queue$ as well. The reason for that is the deletion of the minimal diagnosis $[2]$ w.r.t.\ the last-but-one DPI $\langle\mo,\mb,\Tp,\Tn\cup\setof{Q_1,Q_2}\rangle_\RQ$ wherefore the last evidence for the non-minimality of node $[5,2]$ has been deleted. Hence, the status of $[5,2]$ as a non-minimal diagnosis is no more justified wherefore it must be added to the queue to preserve the completeness of the algorithm w.r.t.\ the finding of all minimal diagnoses w.r.t.\ the current DPI. And, indeed, $[5,2]$ is identified as minimal diagnosis ($\md_5$) in iteration 4.
	\item For each element of $\mD_{\times}$ during each execution of \textsc{updateTree} throughout the execution of Algorithm~\ref{algo:inter_onto_debug}, the Quick Redundancy Check (QRC) is successful. That is, each witness of redundancy used for pruning throughout the entire runtime of the algorithm could be determined very fast. Namely, as it is easy to see from line~\ref{algoline:update:qx} in \textsc{dynamicHS}, the KB used in the call of $\scQX$ in the QRC for some node $\mathsf{nd}$ has a size in $O((|\mathsf{nd}|-1) |\mc_{\max}|)$ where $\mc_{\max}$ is the minimal conflict set of maximum cardinality in $\mC_{calc}$. In most of the cases, $|\mathsf{nd}| \ll |\mo|$ as well as $|\mc_{\max}| \ll |\mo|$ will hold. The (usually more expensive) Complete Redundancy Check (CRC), which requires $O(|\mathsf{nd}|)$ calls to $\scQX$ with a KB of size $O(|\mc_{\max}|-1)$, is thus never employed.
	\item In this example, the same minimal diagnosis $[5,7]$ is used to compute the finally returned solution KB as in Example~\ref{example:staticHS_simple_example_using_tabExDpi2}. The only difference between both outputs is that the KB $(\mo \setminus [5,7]) \cup Q_4$ returned by \textsc{dynamicHS} in this example contains the new positive test case $Q_4 \in \Tp'$. The output by \textsc{staticHS} in	Example~\ref{example:staticHS_simple_example_using_tabExDpi2} does not contain any newly specified positive test case in $\Tp'$ (cf.\ Remark~\ref{rem:mode=static_=>_returned_solution_onto_extensible_to_be_solution_onto_w.r.t._current_DPI}), just the union of the ``original'' positive test cases in $\Tp$ (apart from that, there is not even a newly specified positive test case in Example~\ref{example:staticHS_simple_example_using_tabExDpi2}).
	\item In spite of finding the same solution diagnosis, \textsc{staticHS} requires fewer queries than \textsc{dynamicHS}. Notably, \textsc{dynamicHS} even needs a proper superset of the queries asked by \textsc{staticHS} ($Q_1, Q_2$ in Example~\ref{example:staticHS_simple_example_using_tabExDpi2} are equal to $Q_1, Q_3$ in our current example) in this case. Such a proposition however cannot be made in general since the queries formulated by \textsc{staticHS} generally differ from those formulated by \textsc{dynamicHS}. In this vein, it might just as well be the case that it takes \textsc{dynamicHS} fewer queries to finish than it takes \textsc{staticHS}, due to its advantages in tree pruning. 
\end{itemize}
%
%
 %
%

All in all, the execution of Algorithm~\ref{algo:inter_onto_debug} in this example performs 
\begin{itemize}
	\item 2 full $\scQX$ calls, i.e.\ calls of $\scQX$ using the KB $\mo \setminus \mathsf{node}$ for a node $\mathsf{node}$ that actually return a minimal conflict set (there are two minimal conflict sets labeled by $C$ in Figures~\ref{fig:example:inter_onto_debug_dynamicHS_TabExDpi2} and \ref{fig:example:inter_onto_debug_dynamicHS_TabExDpi2_continued} which do not result from QRC, CRC or the minimality test of a conflict set in line~\ref{algoline:dlabel:qx_1} of \textsc{dynamicHS}),
	\item 4 fast $\scQX$ calls, i.e.\ executions of $\scQX$ within the scope of the QRC (one call of $\scQX$ each for the QRC of $\md_1$, $\md_3$, $\md_2$ and $\md_5$), 
	\item 5 validity checks, i.e.\ calls of $\scQX$ that return 'no conflict' (one check for each of the five found minimal diagnoses where the identification of diagnoses $\md_2$ at step \textcircled{\scriptsize 5}, $\md_2$ at step \textcircled{\scriptsize 9}, $\md_4$ at step \textcircled{\tiny 14} and $\md_4$ at step \textcircled{\tiny 16} does not require any call to a reasoning service by means of $\mD_{\checkmark}$, see line~\ref{algoline:dyn:node_in_Dcheckmark} in \textsc{dynamicHS}; notice that $\scQX$ does only perform a single KB validity check by \textsc{isKBValid} in case it returns 'no conflict', see Algorithm~\ref{algo:qx}) and
	\item 4 tree update processes involving 4 pruned nodes (1 per tree update), 
\end{itemize}
computes
\begin{itemize}
	\item 5 minimal diagnoses ($\md_1$, $\md_2$, $\md_4$ w.r.t.\ the input DPI and $\md_3$ and $\md_5$ w.r.t.\ some DPI resulting from the input DPI by addition of new test cases),
	\item 6 minimal conflict sets ($\tuple{1,2,5}$ as well as $\tuple{1,2,7}$ w.r.t.\ the input DPI and the subsets thereof $\tuple{2,5}$, $\tuple{2,7}$, $\tuple{5}$ and $\tuple{7}$ w.r.t.\ some DPI resulting from the input DPI by addition of new test cases) and 
	\item 4 queries and asks the user 4 logical formulas (1 per query)  
\end{itemize}
and stores
\begin{itemize}
	\item a maximum of 4 nodes (where node refers to the internal representation of a node $\mathsf{nd}$ in \textsc{dynamicHS} as a list of edge labels ($\mathsf{nd}$) and a list of node labels ($\mathsf{nd.cs}$) along a path from the root node to a leaf node).\qed
\end{itemize}	
\end{example}

\newgeometry{margin=2cm}

\begin{figure*}
\begin{minipage}[c]{0.45\textwidth} 
\small
\xygraph{
!{<0cm,0cm>;<1.8cm,0cm>:<0cm,1.2cm>::}
!{(1,4)}*+{\textcircled{\scriptsize 1}\tuple{1,2,5}^C}="c1c"
!{(0,3) }*+{\textcircled{\scriptsize 2}\checkmark_{(\md_1)}}="d1" 
!{(1,3) }*+{\textcircled{\scriptsize 3}\checkmark_{(\md_2)}}="d2" 
!{(3,3) }*+{?}="c2c"
"c1c":"d1"_{1}
"c1c":"d2"^{2}
"c1c":"c2c"^{5}
}
\vspace{5pt}
\begin{center}
\small Iteration 1
\end{center}
\end{minipage}
\begin{minipage}[c]{20pt}
$\Bigg>$ 
\end{minipage}
\begin{minipage}[c]{0.45\textwidth}
\small  
\begin{tabular}{l}                                          
		$\mD_{calc} = \setof{\md_1,\md_2} = \setof{[1],[2]}$ \\
		$\Queue = [[5]]$ \\
		$\mC_{calc} = \setof{\tuple{1,2,5}}$ \\
		$\mD_{\supset} = \emptyset$ \\
		$\Queue_{dup} = []$\vspace{-7pt} \\
    \hspace{-8pt}\hdashrule{0.95\textwidth}{0.5pt}{2mm}\vspace{-4pt} \\
		$\tuple{Q_1,\Pt(Q_1)} = \tuple{\setof{E \rightarrow \lnot A},\tuple{\setof{\md_1},\setof{\md_2},\emptyset}}$ \\
    $u(Q_1) = \false$ \\
		$\mD_{\checkmark} = \setof{\md_2}$, $\mD_{\times} = \setof{\md_1}$
		\end{tabular}
\end{minipage}
\begin{minipage}[c]{10pt}
$\Bigg>$ 
\end{minipage} 

\vspace{10pt}
\begin{minipage}[c]{0.45\textwidth}
\small  
\begin{tabular}{l}                        
		\textsc{updateTree}: \\
		QRC ($\md_1$): 
		$\scQX(\tuple{\setof{2,5},\mb,\Tp,\Tn\cup\setof{Q_1}}) = \tuple{2,5}$ \\
		$\Rightarrow\;$ \textsc{prune}: $\tuple{1,2,5} \rightarrow \tuple{2,5}$ \\
		$\quad \bullet\;$ prune all subtrees starting  
		from nodes $\tuple{1,2,5}$ \\ 
		$\quad\;\;\;$ by outgoing edge with label $1$  \\ 
		$\quad \bullet\;$ replace by $\tuple{2,5}$ all node labels in the tree \\ 
		$\quad\;\;\;$ that are proper supersets of $\tuple{2,5}$ \\
		$\Rightarrow\;$
		$\mD_{\supset} = \emptyset$, 
		$\mC_{calc} = \setof{\tuple{2,5}}$, \\
		$\quad\;$ $\Queue = [[2],[5]]$, 
		$\Queue_{dup} = []$,
\end{tabular}
\end{minipage}
\begin{minipage}[c]{20pt}
$\Bigg>$ 
\end{minipage}
\begin{minipage}[c]{0.45\textwidth}
\small 
\xygraph{
!{<0cm,0cm>;<1.8cm,0cm>:<0cm,1.2cm>::}
!{(1,4)}*+{\textcircled{\scriptsize 1}\tuple{\xcancel{1},2,5}^C}="c1c"
!{(0,3) }*+{\textcircled{\scriptsize 2}\checkmark_{(\md_1)}}="d1" 
!{(1,3) }*+{\textcircled{\scriptsize 3}\checkmark_{(\md_2)}}="d2" 
!{(3,3) }*+{?}="c2c"
!{(0,2) }*+{\textcircled{\scriptsize 4}\times}="inv_d1_q1" 
!{(1,2) }*+{\textcircled{\scriptsize 4} ?}="val_d2_q1"
"c1c":@{|.>}"d1"_{1}
"c1c":"d2"^{2}
"c1c":"c2c"^{5}
"d1":@2{.>}"inv_d1_q1"^{Q_1}
"d2":@2{->}"val_d2_q1"^{Q_1}
}
\vspace{5pt}
\begin{center}
\small Updated Tree
\end{center}
\end{minipage}
\begin{minipage}[c]{10pt}
$\Bigg>$ 
\end{minipage} 

\vspace{10pt}
\begin{minipage}[c]{0.45\textwidth}
\small 
\xygraph{
!{<0cm,0cm>;<1.8cm,0cm>:<0cm,1.2cm>::}
!{(1,4)}*+{\textcircled{\scriptsize 1}\tuple{2,5}^C}="c1c"
!{(1,3) }*+{\textcircled{\scriptsize 3}\checkmark_{(\md_2)}}="d2" 
!{(3,3) }*+{\textcircled{\scriptsize 6}\tuple{1,2,7}^C}="c2c"
!{(1,2) }*+{\textcircled{\scriptsize 5}\checkmark_{(\md_2)}}="val_d2_q1"
!{(2,2) }*+{\textcircled{\scriptsize 7}\checkmark_{(\md_3)}}="d3"
!{(3,2) }*+{?}="nonmin1"
!{(4,2) }*+{?}="d4"
"c1c":"d2"^{2}
"c1c":"c2c"^{5}
"d2":@2{->}"val_d2_q1"^{Q_1}
"c2c":"d3"_{1}
"c2c":"nonmin1"^{2}
"c2c":"d4"^{7}
}
\vspace{5pt}
\begin{center}
\small Iteration 2
\end{center}
\end{minipage}
\begin{minipage}[c]{20pt}
$\Bigg>$ 
\end{minipage}
\begin{minipage}[c]{0.45\textwidth}
\small
\begin{tabular}{l}                                          
		$\mD_{calc} = \setof{\md_2,\md_3} = \setof{[2],[5,1]}$ \\
		$\Queue = [[5,2], [5,7]]$ \\
		$\mC_{calc} = \setof{\tuple{2,5}, \tuple{1,2,7}}$ \\
		$\mD_{\supset} = \emptyset$ \\
		$\Queue_{dup} = []$ \vspace{-7pt} \\
    \hspace{-8pt}\hdashrule{0.95\textwidth}{0.5pt}{2mm}\vspace{-4pt} \\
    $\tuple{Q_2,\Pt(Q_2)} = \tuple{\setof{E \rightarrow G},\tuple{\setof{\md_3},\setof{\md_2},\emptyset}}$ \\
    $u(Q_2) = \false$ \\
		$\mD_{\checkmark} = \setof{\md_2}$, $\mD_{\times} = \setof{\md_3}$
		\end{tabular}
\end{minipage}
\begin{minipage}[c]{10pt}
$\Bigg>$ 
\end{minipage} 

\vspace{10pt}
\begin{minipage}[c]{0.45\textwidth}
\small
\begin{tabular}{l}                                          
		\textsc{updateTree}: \\
		QRC ($\md_3$): 
		$\scQX(\tuple{\setof{2,7},\mb,\Tp,\Tn\cup\setof{Q_1,Q_2}}) = \tuple{2,7}$ \\
		$\Rightarrow\;$ \textsc{prune}: $\tuple{1,2,7} \rightarrow \tuple{2,7}$ \\
		$\Rightarrow\;$
		$\mD_{\supset} = \emptyset$, 
		$\mC_{calc} = \setof{\tuple{2,5}, \tuple{2,7}}$ \\
		$\quad\;$ $\Queue = [[2], [5,2], [5,7]]$, 
		$\Queue_{dup} = []$
\end{tabular}
\end{minipage}
\begin{minipage}[c]{20pt}
$\Bigg>$ 
\end{minipage}
\begin{minipage}[c]{0.45\textwidth}
\small 
\xygraph{
!{<0cm,0cm>;<1.8cm,0cm>:<0cm,1.2cm>::}
!{(1,4)}*+{\textcircled{\scriptsize 1}\tuple{2,5}^C}="c1c"
!{(1,3) }*+{\textcircled{\scriptsize 3}\checkmark_{(\md_2)}}="d2" 
!{(3,3) }*+{\textcircled{\scriptsize 6}\tuple{\xcancel{1},2,7}^C}="c2c"
!{(1,2) }*+{\textcircled{\scriptsize 5}\checkmark_{(\md_2)}}="val_d2_q1"
!{(2,2) }*+{\textcircled{\scriptsize 7}\checkmark_{(\md_3)}}="d3"
!{(3,2) }*+{?}="nonmin1"
!{(4,2) }*+{?}="d4"
!{(1,1) }*+{\textcircled{\scriptsize 8} ?}="val_d2_q2"
!{(2,1) }*+{\textcircled{\scriptsize 8}\times}="inv_d3_q2"
"c1c":"d2"^{2}
"c1c":"c2c"^{5}
"d2":@2{->}"val_d2_q1"^{Q_1}
"c2c":@{|.>}"d3"_{1}
"c2c":"nonmin1"^{2}
"c2c":"d4"^{7}
"val_d2_q1":@2{->}"val_d2_q2"^{Q_2}
"d3":@2{.>}"inv_d3_q2"^{Q_2}
}
\vspace{5pt}
\begin{center}
\small Updated Tree
\end{center}
\end{minipage}
\begin{minipage}[c]{10pt}
$\Bigg>$ 
\end{minipage} 

\vspace{10pt}
\begin{minipage}[c]{0.45\textwidth}
\small 
\xygraph{
!{<0cm,0cm>;<1.8cm,0cm>:<0cm,1.2cm>::}
!{(1,4)}*+{\textcircled{\scriptsize 1}\tuple{2,5}^C}="c1c"
!{(1,3) }*+{\textcircled{\scriptsize 3}\checkmark_{(\md_2)}}="d2" 
!{(3,3) }*+{\textcircled{\scriptsize 6}\tuple{2,7}^C}="c2c"
!{(1,2) }*+{\textcircled{\scriptsize 5}\checkmark_{(\md_2)}}="val_d2_q1"
!{(3,2) }*+{\textcircled{\tiny 10}\times_{(\supset\md_2)}}="nonmin1"
!{(4,2) }*+{\textcircled{\tiny 11}\checkmark_{(\md_4)}}="d4"
!{(1,1) }*+{\textcircled{\scriptsize 9}\checkmark_{(\md_2)}}="val_d2_q2"
"c1c":"d2"^{2}
"c1c":"c2c"^{5}
"d2":@2{->}"val_d2_q1"^{Q_1}
"c2c":"nonmin1"^{2}
"c2c":"d4"^{7}
"val_d2_q1":@2{->}"val_d2_q2"^{Q_2}
}
\vspace{5pt}
\begin{center}
\small Iteration 3
\end{center}
\end{minipage}
\begin{minipage}[c]{20pt}
$\Bigg>$ 
\end{minipage}
\begin{minipage}[c]{0.45\textwidth}
\small
\begin{tabular}{l}                                          
		$\mD_{calc} = \setof{\md_2,\md_4} = \setof{[2],[5,7]}$ \\
		$\Queue = []$ \\
		$\mC_{calc} = \setof{\tuple{2,5}, \tuple{2,7}}$ \\
		$\mD_{\supset} = \setof{[5,2]}$ \\
		$\Queue_{dup} = []$ \vspace{-7pt} \\
    \hspace{-8pt}\hdashrule{0.95\textwidth}{0.5pt}{2mm}\vspace{-4pt} \\
    $\tuple{Q_3,\Pt(Q_3)} = \tuple{\setof{Y \rightarrow \lnot A},\tuple{\setof{\md_2},\setof{\md_4},\emptyset}}$ \\
    $u(Q_3) = \false$ \\
		$\mD_{\checkmark} = \setof{\md_4}$, $\mD_{\times} = \setof{\md_2}$
		\end{tabular}
\end{minipage}
\begin{minipage}[c]{10pt}
$\Bigg>$ 
\end{minipage} 

\vspace{10pt}
\caption[(Example~\ref{example:dynamicHS_small_example_using_tabExDpi2}) Solving the Problem of Interactive Dynamic KB Debugging]{(Example~\ref{example:dynamicHS_small_example_using_tabExDpi2}) Solving the problem of Interactive Dynamic KB Debugging (Problem Definition~\ref{prob_def:dynamic}) for the example DPI given by Table~\ref{tab:example2} by means of Algorithm~\ref{algo:inter_onto_debug} and \textsc{dynamicHS}.} 
\label{fig:example:inter_onto_debug_dynamicHS_TabExDpi2}
\end{figure*}

\begin{figure*}
\begin{minipage}[c]{0.45\textwidth}
\small
\begin{tabular}{l}                                          
		\textsc{updateTree}: \\
		QRC ($\md_2$): 
		$\scQX(\tuple{\setof{5},\mb,\Tp,\Tn\cup\setof{Q_1,Q_2,Q_3}}) = \tuple{5}$ \\
		$\Rightarrow\;$ \textsc{prune}: $\tuple{2,5} \rightarrow \tuple{5}$ \\
		$\Rightarrow\;$
		$\mD_{\supset} = \emptyset$, 
		$\mC_{calc} = \setof{\tuple{5}, \tuple{2,7}}$ \\
		$\quad\;$ $\Queue = [[5,2],[5,7]]$, 
		$\Queue_{dup} = []$
\end{tabular}
\end{minipage}
\begin{minipage}[c]{20pt}
$\Bigg>$ 
\end{minipage}
\begin{minipage}[c]{0.45\textwidth}
\small 
\xygraph{
!{<0cm,0cm>;<1.8cm,0cm>:<0cm,1.2cm>::}
!{(1,4)}*+{\textcircled{\scriptsize 1}\tuple{\xcancel{2},5}^C}="c1c"
!{(1,3) }*+{\textcircled{\scriptsize 3}\checkmark_{(\md_2)}}="d2" 
!{(3,3) }*+{\textcircled{\scriptsize 6}\tuple{2,7}^C}="c2c"
!{(1,2) }*+{\textcircled{\scriptsize 5}\checkmark_{(\md_2)}}="val_d2_q1"
!{(3,2) }*+{\textcircled{\tiny 10}\times_{(\supset\md_2)}}="nonmin1"
!{(4,2) }*+{\textcircled{\tiny 11}\checkmark_{(\md_4)}}="d4"
!{(1,1) }*+{\textcircled{\scriptsize 9}\checkmark_{(\md_2)}}="val_d2_q2"
!{(3,1) }*+{\textcircled{\tiny 12}?}="d5"
!{(4,1) }*+{\textcircled{\tiny 12}?}="val_d4_q3"
!{(1,0) }*+{\textcircled{\tiny 12}\times}="inv_d2_q3"
"c1c":@{|.>}"d2"^{2}
"c1c":"c2c"^{5}
"d2":@2{.>}"val_d2_q1"^{Q_1}
"c2c":"nonmin1"^{2}
"c2c":"d4"^{7}
"val_d2_q1":@2{.>}"val_d2_q2"^{Q_2}
"nonmin1":@2{->}"d5"^{Q_3}
"d4":@2{->}"val_d4_q3"^{Q_3}
"val_d2_q2":@2{.>}"inv_d2_q3"^{Q_3}
}
\vspace{5pt}
\begin{center}
\small Updated Tree
\end{center}
\end{minipage}
\begin{minipage}[c]{10pt}
$\Bigg>$ 
\end{minipage} 

\vspace{10pt}
\begin{minipage}[c]{0.45\textwidth}
\small 
\xygraph{
!{<0cm,0cm>;<1.8cm,0cm>:<0cm,1.2cm>::}
!{(1,4)}*+{\textcircled{\scriptsize 1}\tuple{5}^C}="c1c"
!{(3,3) }*+{\textcircled{\scriptsize 6}\tuple{2,7}^C}="c2c"
!{(3,2) }*+{\textcircled{\tiny 10}\times_{(\supset\md_2)}}="nonmin1"
!{(4,2) }*+{\textcircled{\tiny 11}\checkmark_{(\md_4)}}="d4"
!{(3,1) }*+{\textcircled{\tiny 13}\checkmark_{(\md_5)}}="d5"
!{(4,1) }*+{\textcircled{\tiny 14}\checkmark_{(\md_4)}}="val_d4_q3"
"c1c":"c2c"^{5}
"c2c":"nonmin1"^{2}
"c2c":"d4"^{7}
"nonmin1":@2{->}"d5"^{Q_3}
"d4":@2{->}"val_d4_q3"^{Q_3}
}
\vspace{5pt}
\begin{center}
\small Iteration 4
\end{center}
\end{minipage}
\begin{minipage}[c]{20pt}
$\Bigg>$ 
\end{minipage}
\begin{minipage}[c]{0.45\textwidth}
\small
\begin{tabular}{l}                                          
		$\mD_{calc} = \setof{\md_4,\md_5} = \setof{[5,7],[5,2]}$ \\
		$\Queue = []$ \\
		$\mC_{calc} = \setof{\tuple{5}, \tuple{2,7}}$ \\
		$\mD_{\supset} = \emptyset$ \\
		$\Queue_{dup} = []$ \vspace{-7pt} \\
    \hspace{-8pt}\hdashrule{0.95\textwidth}{0.5pt}{2mm}\vspace{-4pt} \\
    $\tuple{Q_4,\Pt(Q_4)} = \tuple{\setof{E \rightarrow Z},\tuple{\setof{\md_4},\setof{\md_5},\emptyset}}$ \\
    $u(Q_4) = \true$ \\
		$\mD_{\checkmark} = \setof{\md_4}$, $\mD_{\times} = \setof{\md_5}$
		\end{tabular}
\end{minipage}
\begin{minipage}[c]{10pt}
$\Bigg>$ 
\end{minipage} 

\vspace{10pt}
\begin{minipage}[c]{0.45\textwidth}
\small
\begin{tabular}{l}                                          
		\textsc{updateTree}: \\
		QRC ($\md_5$): \\
		$\scQX(\tuple{\setof{7},\mb,\Tp\cup\setof{Q_4},\Tn\cup\setof{Q_1,Q_2,Q_3}}) = \tuple{7}$ \\
		$\Rightarrow\;$ \textsc{prune}: $\tuple{2,7} \rightarrow \tuple{7}$ \\
		$\Rightarrow\;$
		$\mD_{\supset} = \emptyset$, 
		$\mC_{calc} = \setof{\tuple{5}, \tuple{7}}$ \\
		$\quad\;$ $\Queue = [[5,7]]$, 
		$\Queue_{dup} = []$
\end{tabular}
\end{minipage}
\begin{minipage}[c]{20pt}
$\Bigg>$ 
\end{minipage}
\begin{minipage}[c]{0.45\textwidth} 
\small
\xygraph{
!{<0cm,0cm>;<1.8cm,0cm>:<0cm,1.2cm>::}
!{(1,4)}*+{\textcircled{\scriptsize 1}\tuple{5}^C}="c1c"
!{(3,3) }*+{\textcircled{\scriptsize 6}\tuple{\xcancel{2},7}^C}="c2c"
!{(3,2) }*+{\textcircled{\tiny 10}\times_{(\supset\md_2)}}="nonmin1"
!{(4,2) }*+{\textcircled{\tiny 11}\checkmark_{(\md_4)}}="d4"
!{(3,1) }*+{\textcircled{\tiny 13}\checkmark_{(\md_5)}}="d5"
!{(4,1) }*+{\textcircled{\tiny 14}\checkmark_{(\md_4)}}="val_d4_q3"
!{(3,0) }*+{\textcircled{\tiny 15}\times}="inv_d5_q4"
!{(4,0) }*+{\textcircled{\tiny 15} ?}="val_d4_q4"
"c1c":"c2c"^{5}
"c2c":@{|.>}"nonmin1"^{2}
"c2c":"d4"^{7}
"nonmin1":@2{.>}"d5"^{Q_3}
"d4":@2{->}"val_d4_q3"^{Q_3}
"d5":@2{.>}"inv_d5_q4"^{Q_4}
"val_d4_q3":@2{->}"val_d4_q4"^{Q_4}   
}
\vspace{5pt}
\begin{center}
\small Updated Tree
\end{center}
\end{minipage}
\begin{minipage}[c]{10pt}
$\Bigg>$ 
\end{minipage} 

\vspace{10pt}
\begin{minipage}[c]{0.45\textwidth} 
\small
\xygraph{
!{<0cm,0cm>;<1.8cm,0cm>:<0cm,1.2cm>::}
!{(1,4)}*+{\textcircled{\scriptsize 1}\tuple{5}^C}="c1c"
!{(3,3) }*+{\textcircled{\scriptsize 6}\tuple{7}^C}="c2c"
!{(4,2) }*+{\textcircled{\tiny 11}\checkmark_{(\md_4)}}="d4"
!{(4,1) }*+{\textcircled{\tiny 14}\checkmark_{(\md_4)}}="val_d4_q3"
!{(4,0) }*+{\textcircled{\tiny 16}\checkmark_{(\md_4)}}="val_d4_q4"
"c1c":"c2c"^{5}
"c2c":"d4"^{7}
"d4":@2{->}"val_d4_q3"^{Q_3}
"val_d4_q3":@2{->}"val_d4_q4"^{Q_4}   
}
\vspace{5pt}
\begin{center}
\small Iteration 5
\end{center}
\end{minipage}
\begin{minipage}[c]{20pt}
$\Bigg>$ 
\end{minipage}
\begin{minipage}[c]{0.45\textwidth}
\small
\begin{tabular}{l}                                          
		$\mD_{calc} = \setof{\md_4} = \setof{[5,7]}$ \\
		$\Queue = []$ \\
		$\mC_{calc} = \setof{\tuple{5}, \tuple{7}}$ \\
		$\mD_{\supset} = \emptyset$ \\
		$\Queue_{dup} = []$ \vspace{-7pt} \\
    \hspace{-8pt}\hdashrule{0.95\textwidth}{0.5pt}{2mm}\vspace{-4pt} \\
    $p_{\mD}(\md_4) = 1 $ \\ 
		$\Rightarrow\quad$ return the solution KB $(\mo\setminus\md_4) \cup Q_4 \quad \qed$
		\end{tabular}
\end{minipage}

\vspace{10pt}
\caption[(Example~\ref{example:dynamicHS_small_example_using_tabExDpi2} continued) Solving the Problem of Interactive Dynamic KB Debugging]{(Example~\ref{example:dynamicHS_small_example_using_tabExDpi2} continued) Solving the problem of Interactive Dynamic KB Debugging (Problem Definition~\ref{prob_def:dynamic}) for the example DPI given by Table~\ref{tab:example2} by means of Algorithm~\ref{algo:inter_onto_debug} and \textsc{dynamicHS}.} 
\label{fig:example:inter_onto_debug_dynamicHS_TabExDpi2_continued}
\end{figure*}
\restoregeometry

\begin{example}\label{example:dynamicHS_large_example_using_tabExDpi3}
Let us now consider the (admissible) DPI $\langle\mo,\mb,\Tp,\Tn\rangle_\RQ$ given by Table~\ref{tab:example3}. We assume an expert (called user throughout this example) in the domain $Dom$ modeled by $\mo$ who wants to find a solution to Interactive Dynamic KB Debugging for the given DPI $\langle\mo,\mb,\Tp,\Tn\rangle_\RQ$ by means of Algorithm~\ref{algo:inter_onto_debug} with $mode = dynamic$.
Further, the same scenario and parameter settings as in Example~\ref{example:staticHS_complex_example_using_tabExDpi3} are supposed. That is, $n_{\min} = n_{\max} = 3$ (notice that the time limit $t$ is irrelevant in this case), $q := 1$ (cf.\ Section~\ref{sec:QueryGeneration}), $qsm()$ is equal to any query selection measure described in Section~\ref{sec:query_selection_measures}, $p_{\widetilde{\mo}\cup\overline{\mo}}: \widetilde{\mo}\cup\overline{\mo} \rightarrow [0,1]$ is given such that $p_{\mo}(\tax)$ for $\tax \in \mo$ resulting from the application of \textsc{getAxiomsProbs} is as given by Table~\ref{tab:example:staticHS_complex--->axiom_probs} and $\sigma := 0$.

The tree constructed and parameters computed and used by Algorithm~\ref{algo:inter_onto_debug} using \textsc{dynamicHS} are visualized by Figures~\ref{fig:example:inter_onto_debug_dynamicHS_TabExDpi3} and \ref{fig:example:inter_onto_debug_dynamicHS_TabExDpi3_continued}. 
We use the same notation as in Figures~\ref{fig:example:non-interactive_onto_debug_auto=false+nmin=infty_and_auto=true}, \ref{fig:example:non-interactive_onto_debug_auto=false+nmin=2+nmax=4_with_probs}, \ref{fig:example:inter_onto_debug_staticHS_TabExDpi2}, \ref{fig:example:inter_onto_debug_staticHS_TabExDpi3}, \ref{fig:example:inter_onto_debug_staticHS_TabExDpi3_continued}, \ref{fig:example:inter_onto_debug_dynamicHS_TabExDpi2} and \ref{fig:example:inter_onto_debug_dynamicHS_TabExDpi2_continued} which is described in Examples~\ref{example:non_interactive_debugging_with_tabExDpi2_and_without_probs}, \ref{example:non_interactive_debugging_with_tabExDpi2_and_probs}, \ref{example:staticHS_simple_example_using_tabExDpi2}, \ref{example:staticHS_complex_example_using_tabExDpi3} and \ref{example:dynamicHS_small_example_using_tabExDpi2}. 
%

After the initialization of variables, Algorithm~\ref{algo:inter_onto_debug} calls the function \textsc{getFormulaProbs} in line~\ref{algoline:inter_onto_debug:getAxiomProbs} which exploits $p_{\widetilde{\mo}\cup\overline{\mo}}()$ to calculate the function $p_{\mo}()$ giving the fault probabilities of formulas in $\mo$ (cf.\ Sections~\ref{sec:prob_space_construction}, \ref{sec:DetailedAlgorithmDescription} and Example~\ref{example:ax_prob_calc}).

Then, \textsc{dynamicHS} is called for the first time, resulting in the hitting set tree given in the first picture in Figure~\ref{fig:example:inter_onto_debug_dynamicHS_TabExDpi3}. 
As outlined by the numbers \textcircled{\scriptsize i} indicating at which point in time a node is labeled, the root node (initially the empty set) is labeled first by $\mc_1 := \tuple{1,2,5}$ and three successor nodes, namely $\mathsf{nd}_1 := [1]$, $\mathsf{nd}_2 := [2]$ as well as $\mathsf{nd}_3 := [5]$ with $\mathsf{nd}_1.\mathsf{cs} = \mathsf{nd}_2.\mathsf{cs} = \mathsf{nd}_3.\mathsf{cs} = [\tuple{1,2,5}]$, are added to the queue of open nodes $\Queue$.
Contrary to Example~\ref{example:dynamicHS_small_example_using_tabExDpi2}, where the tree was built up in breadth-first order, in this example the formula probabilities $p() := p_{\mo}()$ given by Table~\ref{tab:example:staticHS_complex--->axiom_probs} are used to assign a probability $p_{nodes}(\mathsf{n})$ to each path $\mathsf{n}$ in the tree starting from the root node (cf.\ Formula~\ref{eq:path_prob_calc} and Definition~\ref{def:p_node()}). 
In this vein, the node corresponding to the outgoing edge of $\mc_1$ labeled by the formula with the largest fault probability among all formulas in $\mc_1$ is processed next. That is, the node $[1]$ with $p_{nodes}([1]) = 0.41$ (as opposed to the nodes $[2]$ and $[5]$ with $0.25$ each) is labeled next. The \textsc{dLabel} procedure, after checking whether $[1]$ is a non-minimal diagnosis w.r.t.\ $\langle\mo,\mb,\Tp,\Tn\rangle_\RQ$ (check is negative), computes another minimal conflict set $\mc_2 := \tuple{2,4,6}$ such that $[1]\cap\mc_2 = \emptyset$ ($\mc_2$ is not hit by the node $[1]$) to constitute a label for node $[1]$. The successor nodes $[1,2]$, $[1,4]$ and $[1,6]$ of $[1]$ are generated and added to the list $\Queue$ in a way that the sorting of $\Queue$ in descending order of $p_{nodes}()$ is maintained. 

Since $[1,4]$ (0.28) as well as $[1,6]$ (0.27) have a larger probability (as per $p_{nodes}()$) than the nodes $[2]$ (0.25) and $[5]$ (0.25), $\Queue$ is given by $[[1,4],[1,6],[2],[5],[1,2]]$ when it comes to the processing of the next node. Since \textsc{dynamicHS} always treats the first node of $\Queue$ next, it identifies the first minimal diagnoses $\md_1 := [1,4]$ and $\md_2 := [1,6]$ w.r.t.\ $\langle\mo,\mb,\Tp,\Tn\rangle_\RQ$ at steps \textcircled{\scriptsize 3} and \textcircled{\scriptsize 4}, respectively. At step \textcircled{\scriptsize 5}, when node $[2]$ is processed, a minimal conflict set $\mc_3 := \tuple{1,3,4}$ is computed and set as a label for $[2]$, giving rise to the generation of three further nodes $[2,1]$, $[2,3]$ and $[2,4]$, all with $\mathsf{nd}_i.\mathsf{cs} = [\tuple{1,2,5},\tuple{1,3,4}]$. 

However, notice that not all of these new nodes are added to $\Queue$, contrary to \textsc{staticHS} (cf.\ Example~\ref{example:staticHS_complex_example_using_tabExDpi3}). For, there is already a node $[1,2]$ corresponding to the set $\setof{1,2}$ in $\Queue$. Due to the test performed in line~\ref{algoline:dyn:check_node_already_in_Q}, this duplicate node $[2,1]$ is assigned to the list $\Queue_{dup}$ which is expressed in the figure by $dup$. Since diagnoses are sets, not lists, $[1,2,\tax_1,\dots,\tax_k]$ and $[2,1,\tax_1,\dots,\tax_k]$ constitute one and the same diagnosis and it is irrelevant whether the one or the other is found. Hence, the nodes $[1,2]$ and $[2,1]$ are regarded as duplicates. Nevertheless, $\mathsf{nd}_i := [2,1]$ (with $\mathsf{nd}_i.\mathsf{cs} = [\tuple{1,2,5},\tuple{1,3,4}]$) must not be completely deleted as it might be the case that (some successor node of) $\mathsf{nd}_j := [1,2]$ (with $\mathsf{nd}_j.\mathsf{cs} = [\tuple{1,2,5},\tuple{2,4,6}]$) becomes redundant due to the eventual addition of some test case. For example, in case the reason for the redundancy of $\mathsf{nd}_j$ is given (only) by a witness of redundancy that is a subset of $\tuple{2,4,6}$, $\mathsf{nd}_j$ is pruned and replaced by the node $\mathsf{nd}_i$ which is still non-redundant.

Thence, only $[2,3]$ and $[2,4]$ are added to $\Queue$ as successor nodes of the processed node $[2]$. Next, the minimal conflict set $\mc_2 = \tuple{2,4,6}$ is reused (lines~\ref{algoline:dlabel:reuse_start}-\ref{algoline:dlabel:reuse_end} in \textsc{dLabel}) as a label for node $[5]$ with $p_{nodes}([5]) = 0.25$ and the three new nodes $[5,2]$, $[5,4]$ as well as $[5,6]$ are generated and assigned to $\Queue$ at step \textcircled{\scriptsize 7}. Then, the fourth minimal conflict set $\mc_4 := \tuple{1,5,6,8}$ is computed to label the node $[2,4]$ with $p_{nodes}([2,4]) = 0.18$ and the four new nodes $[2,4,1]$, $[2,4,5]$, $[2,4,6]$ as well as $[2,4,8]$ are generated and assigned to $\Queue$ st step \textcircled{\scriptsize 8}. At step \textcircled{\scriptsize 9}, the third minimal diagnosis $\md_3 := [5,4]$ w.r.t.\ $\langle\mo,\mb,\Tp,\Tn\rangle_\RQ$ is eventually found and added to $\mD_{calc}$ which now has reached a cardinality of $3 = n_{\min} = n_{\max}$ wherefore \textsc{dynamicHS} stops and returns i.a.\ the set of leading diagnoses $\mD_{calc} = \setof{[1,4],[1,6],[5,4]}$. The returned values are given in the lefthand column in Figure~\ref{fig:example:inter_onto_debug_dynamicHS_TabExDpi3}.

As in Example~\ref{example:staticHS_complex_example_using_tabExDpi3}, where a debugging session for the same DPI using \textsc{staticHS} is presented, the first query $Q_1$ is computed as $\setof{B \sqsubseteq K}$ and answered by $\true$ by the user. The assignment of $Q_1$ to the positive test cases of the DPI $\langle\mo,\mb,\Tp,\Tn\rangle_\RQ$ brings the opportunity to perform some significant pruning actions (within the function \textsc{updateTree} called at the beginning of the second call of \textsc{dynamicHS}). These are shown in the tree with the caption 'Updated Tree' and in the righthand column in Figure~\ref{fig:example:inter_onto_debug_dynamicHS_TabExDpi3}.

As a first step within \textsc{updateTree}, a redundancy check is performed for each diagnosis in $\mD_{\times}$. In this case $\mD_{\times} = \setof{\md_3} = \setof{[5,4]}$ since $\md_3$ is the only minimal diagnosis that has been ruled out by the most recently added positive test case $Q_1$. The purpose of the redundancy check is to figure out whether $\md_3$ is redundant w.r.t.\ the current DPI and must be pruned or whether it might be extended to become a minimal diagnosis w.r.t.\ the current DPI.
  
First, the Quick Redundancy Check (QRC) $\scQX(\tuple{\setof{1,2,6},\mb,\Tp\cup\setof{Q_1},\Tn}) = \tuple{1}$ (line~\ref{algoline:update:qx} in \textsc{dynamicHS}) is executed for $\md_3$ where the KB $\setof{1,2,6}$ used in this call of $\scQX$ is obtained by deletion of $\mathsf{node} := \md_3$ from the union of all conflict sets (the elements of $\mathsf{node.cs}$) along the path that corresponds to $\md_3$, i.e.\ $\setof{1,2,6} = (\tuple{1,2,5} \cup \tuple{2,4,6}) \setminus [5,4]$. By means of the QRC it is figured out (line~\ref{algoline:update:X_subset_C_(QRC)} in \textsc{dynamicHS}) that $\md_3$ (and possibly some further nodes) is redundant and can be pruned. This holds since the minimal conflict set $\tuple{1,2,5}$ w.r.t.\ the last-but-one DPI $\langle\mo,\mb,\Tp,\Tn\rangle_\RQ$ is not a minimal conflict set w.r.t.\ the current DPI $\langle\mo,\mb,\Tp\cup\setof{Q_1},\Tn\rangle_\RQ$ because $\tuple{1}$ returned by $\scQX$ is already a minimal conflict set w.r.t.\ the current DPI (cf.\ Proposition~\ref{prop:qx_correctness}). We call this minimal conflict set $\tuple{1}$ a \emph{witness of redundancy} for $\md_3$. Hence, all branches in the hitting set tree starting from an outgoing edge of $\tuple{1,2,5}$ labeled by $2$ or by $5$ can be safely deleted from all collections storing nodes in \textsc{dynamicHS}.

An illustration why $\tuple{1}$ ``replaces'' $\tuple{1,2,5}$ as a minimal conflict set w.r.t.\ the current DPI can be given as follows: First, $\tuple{1,2,5}$ is a minimal conflict set w.r.t.\ $\langle\mo,\mb,\Tp,\Tn\rangle_\RQ$ as it is a set-minimal subset of $\mo$ that entails $\setof{A \sqsubseteq K} = \tn_1 \in \Tn$ and there is no proper subset $\mc'$ of $\tuple{1,2,5}$ where $\mc' \cup \mb \cup U_{\Tp}$ violates any $r \in \RQ$ or entails any $\tn \in \Tn$ (see example~\ref{example:analysis_TabExDpi3} for a detailed explanation). Second, considering the current DPI $\langle\mo,\mb,\Tp\cup\setof{Q_1},\Tn\rangle_\RQ$, we have that $\tuple{1,2,5} \cup \mb \cup U_{\Tp\cup\setof{Q_1}} \models \tn_1$, too. However, $\setof{2,5} = \setof{B \sqsubseteq G, G \sqsubseteq K} \models \setof{B \sqsubseteq K} = Q_1$ implies that $\mb \cup U_{\Tp \cup \setof{Q_1}} \supseteq Q_1$ can replace the subset $\setof{2,5}$ of the conflict set $\tuple{1,2,5}$. For, formula $1$ ($A \sqsubseteq B$) along with $Q_1$ ($B \sqsubseteq K$) already entails $\tn_1$. Further, $\mb \cup U_{\Tp \cup \setof{Q_1}}$ cannot violate any negative test case $\tn_i \in \Tn$ or requirement $r_j \in \RQ$ by the admissibility of the input DPI $\langle\mo,\mb,\Tp,\Tn\rangle_\RQ$, the fact that $Q_1$ is a query, Corollary~\ref{cor:query_leaves_valid_diag}, Definition~\ref{def:admissible} and Proposition~\ref{prop:exist_diag}. Thus, by Definition~\ref{def:cs}, $\tuple{1}$ is in fact a minimal conflict set w.r.t.\ the current DPI $\langle\mo,\mb,\Tp\cup\setof{Q_1},\Tn\rangle_\RQ$.

Now, the first nice thing at this point is that $\tuple{1}$ is not only a witness of redundancy of nodes $\mathsf{nd}$ where $\tuple{1,2,5} \in \mathsf{nd.cs}$, but of each $\mathsf{nd}$ (in the tree or in the set $\Queue_{dup}$ of duplicate nodes) where $\mathsf{nd.cs}$ contains a conflict set that is a proper superset of $\tuple{1}$.
That is, $\tuple{1}$ also replaces $\tuple{1,3,4}$ as well as $\tuple{1,5,6,8}$. This implicates that two outgoing edges (those labeled by $2$ or $5$) of $\tuple{1,2,5}$, two outgoing edges (those labeled by $3$ or $4$) of $\tuple{1,3,4}$ and three outgoing edges (those labeled by $5$, $6$ or $8$) of $\tuple{1,5,6,8}$ can be pruned.

The second nice thing that has an even more significant bearing on tree pruning than the first thing is that $\tuple{1}$ is a witness of redundancy of the conflict set that labels the root node. That is, pruning can take place at the very top of the tree and two of three subtrees rooted at successor nodes of the root node can be pruned.
That is, for instance, \emph{within} the rightmost subtree of the root node in the picture with caption 'Updated Tree' in Figure~\ref{fig:example:inter_onto_debug_dynamicHS_TabExDpi3} no pruning is possible at all since the conflict set $\tuple{2,4,6}$ labels the root node of this subtree and $\tuple{1}$ is not a subset of $\tuple{2,4,6}$. However, this subtree is still redundant since it is connected with the root node by a ``redundant'' edge labeled by $5$. As a consequence, we can observe the pruning of a total of 9 nodes (of altogether 12 nodes in the tree) in only one execution of \textsc{updateTree}. 

Now, to receive an impression of the power of tree pruning in \textsc{dynamicHS}, the reader is invited to compare the trees used in iterations 2 and 3 in the current example (the bottom left pictures in Figure~\ref{fig:example:inter_onto_debug_dynamicHS_TabExDpi3} and Figure~\ref{fig:example:inter_onto_debug_dynamicHS_TabExDpi3_continued}) with the trees used in iterations 2 and 3 in Example~\ref{example:staticHS_complex_example_using_tabExDpi3} (the bottom picture in Figure~\ref{fig:example:inter_onto_debug_staticHS_TabExDpi3} and the picture in Figure~\ref{fig:example:inter_onto_debug_staticHS_TabExDpi3_continued}) which deals with the debugging of the same DPI (just by means of \textsc{staticHS} instead of \textsc{dynamicHS}), uses the same sets of leading diagnoses in each iteration, thus the same queries, and of course the same user (that gives the same answers in both examples).

After all diagnoses of $\mD_{\checkmark}$ are added to $\Queue$ as a final action within \textsc{updateTree}, the repeat-loop of the second iteration of \textsc{dynamicHS} is entered. Here, the minimal diagnoses $\md_1$ ($p_{nodes}(\md_1) = 0.28$, step \textcircled{\tiny 11}), $\md_2$ ($0.27$, \textcircled{\tiny 12}) and $\md_4$ ($0.09$, \textcircled{\tiny 13}) are found and assigned to the empty set $\mD_{calc}$ before \textsc{dynamicHS} terminates again. Notice that only one call of the \textsc{dLabel} procedure is required in the second iteration (for node $[1,2]$) due to the test in line~\ref{algoline:dyn:node_in_Dcheckmark} of \textsc{dynamicHS} which is positive for $\md_1$ and $\md_2$ (since $\md_1, \md_2 \in \mD_{\checkmark}$). 

Once the second query $Q_2 = \setof{B \sqsubseteq \exists r.F}$ is added to the positive test cases resulting in the DPI $\langle\mo,\mb,\Tp\cup\setof{Q_1,Q_2},\Tn\rangle_\RQ$, the \textsc{updateTree} function causes the pruning of two further nodes ($\md_2 = [1,6]$ and $\md_4 = [1,2]$) leading to the continuance of only a single node ($\md_1 = [1,4]$) in the memory of \textsc{dynamicHS} (see the picture with caption 'Updated Tree' in Figure~\ref{fig:example:inter_onto_debug_dynamicHS_TabExDpi3_continued}). The reason for this is that $Q_2$ can ``replace'' the part $\setof{2,6} = \setof{B \sqsubseteq G, G \sqsubseteq \exists r.F}$ (which entails $Q_2$) of the minimal conflict set $\tuple{2,4,6}$ w.r.t.\ the last-but-one DPI $\langle\mo,\mb,\Tp\cup\setof{Q_1},\Tn\rangle_\RQ$ such that $\tuple{2,4,6} \setminus \setof{2,6} = \tuple{4}$ is already a minimal conflict set w.r.t.\ the current DPI $\langle\mo,\mb,\Tp\cup\setof{Q_1,Q_2},\Tn\rangle_\RQ$ (cf.\ the analysis of the minimal conflict set $\mc_2 = \tuple{2,4,6}$ in Example~\ref{example:analysis_TabExDpi3}). 

Since, by now, all minimal conflict sets $\tuple{1,2,5}$, $\tuple{2,4,6}$, $\tuple{1,5,6,8}$ as well as $\tuple{1,3,4}$ w.r.t.\ the input DPI $\langle\mo,\mb,\Tp,\Tn\rangle_\RQ$ have ``shrunk'' as much as to constitute only two different set-minimal sets $\tuple{1}$ and $\tuple{4}$, it is clear by Proposition~\ref{prop:mindiag_mincs} that there can be only a single minimal diagnosis $[1,4]$ w.r.t.\ the current DPI $\langle\mo,\mb,\Tp\cup\setof{Q_1,Q_2},\Tn\rangle_\RQ$. Therefore, the third iteration of \textsc{dynamicHS} terminates due to $\Queue = []$ and returns the singleton set $\mD_{calc} = \setof{[1,4]}$. Consequently, the probability $p_{\mD}([1,4]) = 1$ wherefore Algorithm~\ref{algo:inter_onto_debug} also stops executing and returns $(\mo \setminus [1,4]) \cup \tp_1 \cup Q_1 \cup Q_2$ as the (exact) solution to the Interactive Dynamic KB Debugging problem for the DPI $\langle\mo,\mb,\Tp,\Tn\rangle_\RQ$.

The advantage of \textsc{dynamicHS} in this example over \textsc{staticHS} in Example~\ref{example:staticHS_complex_example_using_tabExDpi3} in iterations 2 and 3 is that the pruning of nodes lets the algorithm automatically focus on the still relevant (i.e.\ non-redundant) parts of the tree. \textsc{staticHS}, on the other hand, is doomed to spend most of the execution time for investigating nodes that turn out to be already invalidated by some specified test case(s). As already mentioned in Example~\ref{example:staticHS_complex_example_using_tabExDpi3}, the inability of \textsc{staticHS} to ``early-prune'' incomplete branches of the tree is especially unfavorable in the last iteration of \textsc{staticHS} in case $\sigma = 0$ since all irrelevant minimal diagnoses w.r.t.\ the input DPI must first be computed before they can be ruled out.
%
%

This immense upside of \textsc{dynamicHS} over \textsc{staticHS} (see the analysis in the end of Example~\ref{example:staticHS_complex_example_using_tabExDpi3}) also finds expression in the quantitative analysis of this example given next. 
All in all, the execution of Algorithm~\ref{algo:inter_onto_debug} in this example performs 
\begin{itemize}
	\item 4 full $\scQX$ calls, i.e.\ calls of $\scQX$ using the KB $\mo \setminus \mathsf{node}$ for a node $\mathsf{node}$ that actually return a minimal conflict set (there are four minimal conflict sets labeled by $C$ in Figures~\ref{fig:example:inter_onto_debug_dynamicHS_TabExDpi3} and \ref{fig:example:inter_onto_debug_dynamicHS_TabExDpi3_continued} which do not result from QRC, CRC or the minimality test of a conflict set in line~\ref{algoline:dlabel:qx_1} of \textsc{dynamicHS}),
	\item 2 fast $\scQX$ calls, i.e.\ executions of $\scQX$ within the scope of the QRC (one call of $\scQX$ each for the QRC of $\md_3$ and $\md_2$), 
	\item 4 validity checks, i.e.\ calls of $\scQX$ that return 'no conflict' (one check for each of the four found minimal diagnoses where the identification of diagnoses $\md_1$ at step \textcircled{\tiny 11}, $\md_2$ at step \textcircled{\tiny 12} and $\md_1$ at step \textcircled{\tiny 15} does not require any call to a reasoning service by means of $\mD_{\checkmark}$, see line~\ref{algoline:dyn:node_in_Dcheckmark} in \textsc{dynamicHS}; notice that $\scQX$ does only perform a single KB validity check by \textsc{isKBValid} in case it returns 'no conflict', see Algorithm~\ref{algo:qx}) and
	\item 2 tree update processes involving 11 pruned nodes (9 nodes during the first update between steps \textcircled{\tiny 10} and \textcircled{\tiny 11} and 2 nodes during the second between steps \textcircled{\tiny 14} and \textcircled{\tiny 15}), 
\end{itemize}
computes
\begin{itemize}
	\item 4 minimal diagnoses ($\md_1$, $\md_2$, $\md_3$ and $\md_4$, all w.r.t.\ the input DPI),
	\item 6 minimal conflict sets ($\tuple{1,2,5}$, $\tuple{2,4,6}$, $\tuple{1,3,4}$ and $\tuple{1,5,6,8}$ w.r.t.\ the input DPI and the subsets thereof $\tuple{1}$ and $\tuple{4}$ w.r.t.\ some DPI resulting from the input DPI by addition of new test cases) and 
	\item 2 queries and asks the user 2 logical formulas (1 per query)  
\end{itemize}
and stores
\begin{itemize}
	\item a maximum of 12 nodes (where node refers to the internal representation of a node $\mathsf{nd}$ in \textsc{dynamicHS} as a list of edge labels ($\mathsf{nd}$) and a list of node labels ($\mathsf{nd.cs}$) along a path from the root node to a leaf node).
\end{itemize}
Finally, we want to emphasize that, in all executions of \textsc{updateTree} throughout this example, the usually very efficient QRC was successful right off and the usually more time-consuming CRC was never required.\qed
\end{example}


\newgeometry{margin=1.8cm}

\begin{figure*}
\begin{minipage}[c]{0.95\textwidth} 
\small
\xygraph{
!{<0cm,0cm>;<2.0cm,0cm>:<0cm,1.0cm>::}
!{(4,4)}*+{\textcircled{\scriptsize 1}\tuple{1,2,5}^C}="c1c"
!{(1,3) }*+{\textcircled{\scriptsize 2}\tuple{2,4,6}^C}="c2c" 
!{(4,3) }*+{\textcircled{\scriptsize 5}\tuple{1,3,4}^C}="c3c" 
!{(7,3) }*+{\textcircled{\scriptsize 7}\tuple{2,4,6}^R}="c2r"
!{(0,2) }*+{?}="d4"
!{(1,2) }*+{\textcircled{\scriptsize 3}\checkmark_{(\md_1)}}="d1"
!{(2,2) }*+{\textcircled{\scriptsize 4}\checkmark_{(\md_2)}}="d2"
!{(3,2) }*+{\textcircled{\scriptsize 6}\, dup}="dup1"
!{(4,2) }*+{?}="?2-2"
!{(5,2) }*+{\textcircled{\scriptsize 8}\tuple{1,5,6,8}^C}="c4c"
!{(6,2) }*+{?}="?2-3"
!{(7,2) }*+{\textcircled{\scriptsize 9}\checkmark_{(\md_3)}}="d3"
!{(8,2) }*+{?}="?2-4"
!{(3,0) }*+{?}="?3-1"
!{(4,0) }*+{?}="?3-2"
!{(5,0) }*+{?}="?3-3"
!{(6,0) }*+{?}="?3-4"
"c1c":"c2c"_{1}^(0.75){0.41}
"c1c":"c3c"_{2}^(0.5){0.25}
"c1c":"c2r"_{5}^(0.75){0.25}
"c2c":"d4"_{2}^(0.6){0.09}
"c2c":"d1"_{4}^(0.6){0.28}
"c2c":"d2"_{6}^(0.75){0.27}
"c3c":"dup1"_{1}^(0.6){0.09}
"c3c":"?2-2"_{3}^(0.6){0.07}
"c3c":"c4c"_{4}^(0.75){0.18}
"c2r":"?2-3"_{2}^(0.6){0.06}
"c2r":"d3"_{4}^(0.6){0.18}
"c2r":"?2-4"_{6}^(0.75){0.17}
"c4c":"?3-1"_{1}^(0.75){0.06}
"c4c":"?3-2"_{5}^(0.82){0.04}
"c4c":"?3-3"_{6}^(0.91){0.11}
"c4c":"?3-4"_{8}^(0.93){0.04}
}
\vspace{5pt}
\begin{center}
\small Iteration 1
\end{center}
\end{minipage}
\begin{minipage}[c]{20pt}
$\Bigg>$ 
\end{minipage}

\vspace{20pt}
\begin{minipage}[c]{0.45\textwidth}
\small  
\begin{tabular}{l}                                          
		$\mD_{calc} = \setof{\md_1, \md_2, \md_3} = \setof{[1,4],[1,6],[5,4]}$ \\
		$\Queue = [[5,6],[2,4,6],[1,2],[2,3],[5,2], $\\
		$\quad\quad\;\;\, [2,4,1],[2,4,5],[2,4,8]]$ \\
		$\mC_{calc} = \setof{\tuple{1,2,5},\tuple{2,4,6},\tuple{1,3,4},\tuple{1,5,6,8}}$ \\
		$\mD_{\supset} = \emptyset$ \\
		$\Queue_{dup} = [[2,1]]$ \\
    $\tuple{Q_1,\Pt(Q_1)} = \tuple{\setof{B \sqsubseteq K},\tuple{\setof{\md_1,\md_2},\setof{\md_3},\emptyset}}$ \\
    $u(Q_1) = \true$ \\
		$\mD_{\checkmark} = \setof{\md_1,\md_2}$, $\mD_{\times} = \setof{\md_3}$
		\end{tabular}
\end{minipage}
\begin{minipage}[c]{10pt}
$\Bigg>$ 
\end{minipage} 
%
\begin{minipage}[c]{0.45\textwidth}
\small  
\begin{tabular}{l}                        
		\textsc{updateTree}: \\
		QRC ($\md_3$): 
		$\scQX(\tuple{\setof{1,2,6},\mb,\Tp\cup\setof{Q_1},\Tn}) = \tuple{1}$ \\
		$\Rightarrow\;$ \textsc{pruneQdup}/\textsc{prune}: $\tuple{1,2,5} \rightarrow \tuple{1}$, $\tuple{1,3,4} \rightarrow \tuple{1}$ \\
		$\quad \bullet\;$ prune all subtrees starting  
		from nodes $\tuple{1,2,5}$ \\
		$\quad\;\;\;$ by an outgoing edge with label $2$ or $5$  \\ 
		$\quad \bullet\;$ prune all subtrees starting  
		from nodes $\tuple{1,3,4}$\\
		$\quad\;\;\;$ by an outgoing edge with label $3$ or $4$  \\ 
		$\quad \bullet\;$ replace by $\tuple{1}$ all node labels in the tree \\ 
		$\quad\;\;\;$ that are proper supersets of $\tuple{1}$ \\
		$\Rightarrow\;$
		$\mD_{\supset} = \emptyset$, 
		$\mC_{calc} = \setof{\tuple{1},\tuple{2,4,6}}$, \\
		$\quad\;$ $\Queue = [[1,4],[1,6],[1,2]]$, $\Queue_{dup} = []$
\end{tabular}
\end{minipage}
\begin{minipage}[c]{20pt}
$\Bigg>$ 
\end{minipage}

\vspace{20pt}
\begin{minipage}[c]{0.95\textwidth} 
\small
\xygraph{
!{<0cm,0cm>;<2.0cm,0cm>:<0cm,1.0cm>::}
!{(4,4)}*+{\textcircled{\scriptsize 1}\tuple{1,\xcancel{2},\xcancel{5}}^C}="c1c"
!{(1,3) }*+{\textcircled{\scriptsize 2}\tuple{2,4,6}^C}="c2c" 
!{(4,3) }*+{\textcircled{\scriptsize 5}\tuple{1,\xcancel{3},\xcancel{4}}^C}="c3c" 
!{(7,3) }*+{\textcircled{\scriptsize 7}\tuple{2,4,6}^R}="c2r"
!{(0,2) }*+{?}="d4"
!{(1,2) }*+{\textcircled{\scriptsize 3}\checkmark_{(\md_1)}}="d1"
!{(2,2) }*+{\textcircled{\scriptsize 4}\checkmark_{(\md_2)}}="d2"
!{(3,2) }*+{\textcircled{\scriptsize 6}\, dup}="dup1"
!{(4,2) }*+{?}="?2-2"
!{(5,2) }*+{\textcircled{\scriptsize 8}\tuple{1,\xcancel{5},\xcancel{6},\xcancel{8}}^C}="c4c"
!{(6,2) }*+{?}="?2-3"
!{(7,2) }*+{\textcircled{\scriptsize 9}\checkmark_{(\md_3)}}="d3"
!{(8,2) }*+{?}="?2-4"
!{(1,1) }*+{\textcircled{\tiny 10} ?}="val_d1_q1"
!{(2,1) }*+{\textcircled{\tiny 10} ?}="val_d2_q1"
!{(7,1) }*+{\textcircled{\tiny 10} \times}="inv_d3_q1"
!{(3,0) }*+{?}="?3-1"
!{(4,0) }*+{?}="?3-2"
!{(5,0) }*+{?}="?3-3"
!{(6,0) }*+{?}="?3-4"
"c1c":"c2c"_{1}^(0.75){0.41}
"c1c":@{|.>}"c3c"_{2}^(0.5){0.25}
"c1c":@{|.>}"c2r"_{5}^(0.75){0.25}
"c2c":"d4"_{2}^(0.6){0.09}
"c2c":"d1"_{4}^(0.6){0.28}
"c2c":"d2"_{6}^(0.75){0.27}
"c3c":"dup1"_{1}^(0.6){0.09}
"c3c":@{|.>}"?2-2"_{3}^(0.6){0.07}
"c3c":@{|.>}"c4c"_{4}^(0.75){0.18}
"c2r":"?2-3"_{2}^(0.6){0.06}
"c2r":"d3"_{4}^(0.6){0.18}
"c2r":"?2-4"_{6}^(0.75){0.17}
"d1":@2{->}"val_d1_q1"^{Q_1}
"d2":@2{->}"val_d2_q1"^{Q_1}
"d3":@2{->}"inv_d3_q1"^{Q_1}
"c4c":"?3-1"_{1}^(0.75){0.06}
"c4c":@{|.>}"?3-2"_{5}^(0.82){0.04}
"c4c":@{|.>}"?3-3"_{6}^(0.91){0.11}
"c4c":@{|.>}"?3-4"_{8}^(0.93){0.04}
}
\vspace{5pt}
\begin{center}
\small Updated Tree
\end{center}
\end{minipage}
\begin{minipage}[c]{10pt}
$\Bigg>$ 
\end{minipage} 

\vspace{10pt}

\vspace{10pt}
\begin{minipage}[c]{0.45\textwidth} 
\small
\xygraph{
!{<0cm,0cm>;<2.0cm,0cm>:<0cm,1.0cm>::}
!{(3,4)}*+{\textcircled{\scriptsize 1}\tuple{1}^C}="c1c"
!{(1,3) }*+{\textcircled{\scriptsize 2}\tuple{2,4,6}^C}="c2c" 
!{(0,2) }*+{\textcircled{\tiny 13}\checkmark_{(\md_4)}}="d4"
!{(1,2) }*+{\textcircled{\scriptsize 3}\checkmark_{(\md_1)}}="d1"
!{(2,2) }*+{\textcircled{\scriptsize 4}\checkmark_{(\md_2)}}="d2"
!{(1,1) }*+{\textcircled{\tiny 11} \checkmark_{(\md_1)}}="val_d1_q1"
!{(2,1) }*+{\textcircled{\tiny 12} \checkmark_{(\md_2)}}="val_d2_q1"
"c1c":"c2c"_{1}^(0.6){0.41}
"c2c":"d4"_{2}^(0.6){0.09}
"c2c":"d1"_{4}^(0.6){0.28}
"c2c":"d2"_{6}^(0.75){0.27}
"d1":@2{->}"val_d1_q1"^{Q_1}
"d2":@2{->}"val_d2_q1"^{Q_1}
}
\vspace{5pt}
\begin{center}
\small Iteration 2
\end{center}
\end{minipage}
\begin{minipage}[c]{20pt}
$\Bigg>$ 
\end{minipage}
\begin{minipage}[c]{0.45\textwidth}
\small
\begin{tabular}{l}                                          
		$\mD_{calc} = \setof{\md_1, \md_2, \md_4} = \setof{[1,4],[1,6],[1,2]}$ \\
		$\Queue = []$ \\
		$\mC_{calc} = \setof{\tuple{1},\tuple{2,4,6}}$ \\
		$\mD_{\supset} = \emptyset$ \\
		$\Queue_{dup} = []$ \\
    $\tuple{Q_2,\Pt(Q_2)} = \tuple{\setof{B \sqsubseteq \exists r.F},\tuple{\setof{\md_1},\setof{\md_2,\md_4},\emptyset}}$ \\
    $u(Q_2) = \true$ \\
		$\mD_{\checkmark} = \setof{\md_1}$, $\mD_{\times} = \setof{\md_2,\md_4}$
		\end{tabular}
\end{minipage}
\begin{minipage}[c]{10pt}
$\Bigg>$ 
\end{minipage}
\vspace{20pt}
\caption[(Example~\ref{example:dynamicHS_large_example_using_tabExDpi3}) Solving the Problem of Interactive Dynamic KB Debugging]{(Example~\ref{example:dynamicHS_large_example_using_tabExDpi3}) Solving the problem of Interactive Dynamic KB Debugging (Problem Definition~\ref{prob_def:dynamic}) for the example DPI given by Table~\ref{tab:example3} by means of Algorithm~\ref{algo:inter_onto_debug} and \textsc{dynamicHS}.}  
\label{fig:example:inter_onto_debug_dynamicHS_TabExDpi3}
\end{figure*}
\restoregeometry

\newgeometry{margin=2cm}

\begin{figure*}
\begin{minipage}[c]{0.45\textwidth}
\small
\begin{tabular}{l}                                          
		\textsc{updateTree}: \\
		QRC ($\md_2$): 
		$\scQX(\tuple{\setof{2,4},\mb,\Tp\cup\setof{Q_1,Q_2},\Tn}) = \tuple{4}$ \\
		$\Rightarrow\;$ \textsc{prune}: $\tuple{2,4,6} \rightarrow \tuple{4}$ \\
		$\Rightarrow\;$
		$\mD_{\supset} = \emptyset$, 
		$\mC_{calc} = \setof{\tuple{1}, \tuple{4}}$ \\
		$\quad\;$ $\Queue = [[1,4]]$, 
		$\Queue_{dup} = []$
\end{tabular}
\end{minipage}
\begin{minipage}[c]{20pt}
$\Bigg>$ 
\end{minipage}
\begin{minipage}[c]{0.45\textwidth} 
\small
\xygraph{
!{<0cm,0cm>;<2.0cm,0cm>:<0cm,1.0cm>::}
!{(3,4)}*+{\textcircled{\scriptsize 1}\tuple{1}^C}="c1c"
!{(1,3) }*+{\textcircled{\scriptsize 2}\tuple{\xcancel{2},4,\xcancel{6}}^C}="c2c" 
!{(0,2) }*+{\textcircled{\tiny 13}\checkmark_{(\md_4)}}="d4"
!{(1,2) }*+{\textcircled{\scriptsize 3}\checkmark_{(\md_1)}}="d1"
!{(2,2) }*+{\textcircled{\scriptsize 4}\checkmark_{(\md_2)}}="d2"
!{(1,1) }*+{\textcircled{\tiny 11} \checkmark_{(\md_1)}}="val_d1_q1"
!{(2,1) }*+{\textcircled{\tiny 12} \checkmark_{(\md_2)}}="val_d2_q1"
!{(0,1) }*+{\textcircled{\tiny 14} \times}="inv_d4_q2"
!{(1,0) }*+{\textcircled{\tiny 14} ?}="val_d1_q2"
!{(2,0) }*+{\textcircled{\tiny 14} \times}="inv_d2_q2"
"c1c":"c2c"_{1}^(0.6){0.41}
"c2c":@{|.>}"d4"_{2}^(0.6){0.09}
"c2c":"d1"_{4}^(0.6){0.28}
"c2c":@{|.>}"d2"_{6}^(0.75){0.27}
"d1":@2{->}"val_d1_q1"^{Q_1}
"d2":@2{.>}"val_d2_q1"^{Q_1}
"d4":@2{.>}"inv_d4_q2"^{Q_2}
"val_d1_q1":@2{->}"val_d1_q2"^{Q_2}
"val_d2_q1":@2{.>}"inv_d2_q2"^{Q_2}
}
\vspace{5pt}
\begin{center}
\small Updated Tree
\end{center}
\end{minipage}
\begin{minipage}[c]{10pt}
$\Bigg>$ 
\end{minipage} 

\vspace{20pt}
\begin{minipage}[c]{0.45\textwidth}
\small 
\xygraph{
!{<0cm,0cm>;<2.0cm,0cm>:<0cm,1.0cm>::}
!{(3,4)}*+{\textcircled{\scriptsize 1}\tuple{1}^C}="c1c"
!{(1,3) }*+{\textcircled{\scriptsize 2}\tuple{4}^C}="c2c" 
!{(1,2) }*+{\textcircled{\scriptsize 3}\checkmark_{(\md_1)}}="d1"
!{(1,1) }*+{\textcircled{\tiny 11} \checkmark_{(\md_1)}}="val_d1_q1"
!{(1,0) }*+{\textcircled{\tiny 15} \checkmark_{(\md_1)}}="val_d1_q2"
"c1c":"c2c"_{1}^(0.6){0.41}
"c2c":"d1"_{4}^(0.6){0.28}
"d1":@2{->}"val_d1_q1"^{Q_1}
"val_d1_q1":@2{->}"val_d1_q2"^{Q_2}
}
\vspace{5pt}
\begin{center}
\small Iteration 3
\end{center}
\end{minipage}
\begin{minipage}[c]{20pt}
$\Bigg>$ 
\end{minipage}
\begin{minipage}[c]{0.45\textwidth}
\small
\begin{tabular}{l}                                          
		$\mD_{calc} = \setof{\md_1} = \setof{[1,4]}$ \\
		$\Queue = []$ \\
		$\mC_{calc} = \setof{\tuple{1}, \tuple{4}}$ \\
		$\mD_{\supset} = \emptyset$ \\
		$\Queue_{dup} = []$ \\
    $p_{\mD}(\md_1) = 1$  \\
		$\Rightarrow\quad$ return the solution KB $(\mo\setminus\md_1) \cup p_1 \cup Q_1 \cup Q_2$ \\
		$\qquad$ ($p_1$: cf.\ Table~\ref{tab:example3}) $\qed$
		\end{tabular}
\end{minipage}

\vspace{20pt}
\caption[(Example~\ref{example:dynamicHS_large_example_using_tabExDpi3} continued) Solving the Problem of Interactive Dynamic KB Debugging]{(Example~\ref{example:dynamicHS_large_example_using_tabExDpi3} continued) Solving the problem of Interactive Dynamic KB Debugging (Problem Definition~\ref{prob_def:dynamic}) for the example DPI given by Table~\ref{tab:example3} by means of Algorithm~\ref{algo:inter_onto_debug} and \textsc{dynamicHS}.} \label{fig:example:inter_onto_debug_dynamicHS_TabExDpi3_continued}
\end{figure*}
\restoregeometry
\newgeometry{margin=2cm}

\begin{algorithm*}
\small
\caption[Iterative Construction of a Dynamic Hitting Set Tree 1]{Iterative Construction of a Dynamic Hitting Set Tree} \label{algo:inter_dyn_hs}
\begin{algorithmic}[1]
\Require a tuple $\tuple{ \langle\mo,\mb,\Tp,\Tn\rangle_\RQ, \Queue, \Queue_{dup}, t, n_{\min}, n_{\max}, \mathbf{C}_{calc}, \mD_{\checkmark}, \mD_{\times}, p(), \Tp', \Tn', \mD_{\supset} }$ consisting of
\begin{itemize}
	\item the DPI $\langle\mo,\mb,\Tp,\Tn\rangle_\RQ$ given as input to Algorithm~\ref{algo:inter_onto_debug},
	\item the overall sets of positively ($\Tp'$) and negatively ($\Tn'$) answered queries added as test cases to $\langle\mo,\mb,\Tp,\Tn\rangle_\RQ$ so far, 
	\item a queue $\Queue$ of open (non-labeled) nodes, 
	\item some computation timeout $t$,
	\item a desired minimal ($n_{\min}\geq2$) and maximal ($n_{\max}$) number of minimal diagnoses to be returned, 
	\item a set $\mathbf{C}_{calc}$ of conflict sets w.r.t.\ the current DPI $\langle\mo,\mb,\Tp\cup\Tp',\Tn\cup\Tn'\rangle_\RQ$,
	\item a set $\mD_{\checkmark}$ of minimal diagnoses w.r.t.\ the current DPI $\langle\mo,\mb,\Tp\cup\Tp',\Tn\cup\Tn'\rangle_\RQ$,
	\item a set $\mD_{\times}$ of minimal diagnoses w.r.t.\ the last-but-one DPI that are invalidated by the most recently added test case, 
	\item a function $p: \mo \rightarrow (0,0.5)$,  
	\item a set $\mD_{\supset}$ of non-minimal diagnoses w.r.t.\ the last-but-one DPI and 
	\item a set $\Queue_{dup}$ of stored (duplicate) nodes $\mathsf{nd}$ that can be used when it comes to constructing a replacement node of a pruned node $\mathsf{nd}' \supseteq \mathsf{nd}$ after tree pruning.
\end{itemize}
 
%

%

%
 
%

%
%

%

%

%

%
\Ensure
a tuple $\tuple{\mD_{calc},\Queue, \mathbf{C}_{calc}, \mD_{\times}, \mD_{\supset},\Queue_{dup}}$ where
\begin{itemize}
\item $\mD_{calc}$ is the current set of leading diagnoses such that
\begin{enumerate}[(a)]
\item $\mD_{calc} \subseteq \minD_{\langle\mo,\mb,\Tp\cup\Tp',\Tn\cup\Tn'\rangle_\RQ}$ is the set of most probable minimal diagnoses w.r.t.\ $\langle\mo,\mb,\Tp\cup\Tp',\Tn\cup\Tn'\rangle_\RQ$ such that 
\begin{enumerate}[(i)]
\item $n_{\min} \leq |\mD_{calc}| \leq n_{\max}$ and 
\item $\mD_{calc}\setminus\mD_{\checkmark} \neq \emptyset$,
\end{enumerate}
if such a set $\mD_{calc}$ 
 exists,
or 
\item $\mD_{calc}$ is equal to the set of all minimal diagnoses $\minD_{\langle\mo,\mb,\Tp\cup\Tp',\Tn\cup\Tn'\rangle_\RQ}$, otherwise,
\end{enumerate}
where ``most-probable'' refers to the probability measure $p_{nodes}()$ (cf.\ Definition~\ref{def:p_node()}) obtained from the given function $p()$;
%
\item $\Queue$ is the current queue of open (non-labeled) nodes of the hitting set tree,
\item $\mathbf{C}_{calc}$ is a set of conflict sets w.r.t.\ the current DPI $\langle\mo,\mb,\Tp\cup\Tp',\Tn\cup\Tn'\rangle_\RQ$,
\item $\mD_{\times} = \emptyset$, 
\item $\mD_{\supset}$ is the set of all processed nodes so far throughout the execution of Algorithm~\ref{algo:inter_onto_debug} that are non-minimal diagnoses w.r.t.\ the current DPI $\langle\mo,\mb,\Tp\cup\Tp',\Tn\cup\Tn'\rangle_\RQ$ and
\item $\Queue_{dup}$ is the updated set of stored (duplicate) nodes $\mathsf{nd}$ that can be used when it comes to constructing a replacement node of a pruned node $\mathsf{nd}' \supseteq \mathsf{nd}$ after tree pruning.
\end{itemize}
\vspace{10pt}
\Procedure{dynamicHS}{$\langle\mo,\mb,\Tp,\Tn\rangle_\RQ, \Queue, \Queue_{dup}, t, n_{\min}, n_{\max}, \mathbf{C}_{calc}, \mD_{\checkmark}, \mD_{\times}, p(), \Tp', \Tn', \mD_{\supset}$}
\State $t_{start} \gets \Call{getTime}{ }$
\State $\mD_{calc} \gets \emptyset$\label{algoline:dyn:Dcalc_gets_emptyset}
\State $\tuple{\Queue, \mD_{\times}, \mD_{\supset}, \mC_{calc}, \Queue_{dup}} \gets \Call{updateTree}{\langle\mo,\mb,\Tp,\Tn\rangle_\RQ, \mD_{\times}, \Queue, \Queue_{dup}, \mD_{\supset},\mD_{\checkmark}, \mC_{calc}, p(), \Tp', \Tn'}$\label{algoline:dyn:update_tree}     
\Repeat	\label{algoline:dyn:repeat}																																	\Comment{\textsc{updateTree} (see Algorithm~\ref{algo:update_tree})}
\State $\mathsf{node} \gets \Call{getFirst}{\Queue}$\label{algoline:dyn:get_first}			  \Comment{$\mathsf{node}$ is processed}		
\State $\Queue \gets \Call{deleteFirst}{\Queue}$\label{algoline:dyn:delete_from_queue}
\If{$\mathsf{node} \in \mD_{\checkmark}$}\label{algoline:dyn:node_in_Dcheckmark}  \Comment{$\mD_{\checkmark}$ includes only minimal diagnoses w.r.t.\ current DPI}
	\State $L \gets valid$\label{algoline:dyn:set_L_to_valid}               
\Else
	\State $\tuple{L,\mathbf{C}_{calc},\Queue_{dup}} \gets \Call{dLabel}{\langle\mo,\mb,\Tp,\Tn\rangle_\RQ, \mathsf{node}, \mathbf{C}_{calc}, \mD_{calc}, \Queue, \Queue_{dup}, p(), \Tp', \Tn'}$\label{algoline:dyn:dlabel}
\EndIf
\If{$L = valid$}\label{algoline:dyn:if_L_valid}  \Comment{\textsc{dLabel} (see Algorithm~\ref{algo:update_tree})}
	\State $\mD_{calc} \gets \mD_{calc} \cup \setof{\mathsf{node}}$\label{algoline:dyn:add_to_Dcalc}      \Comment{$\mathsf{node}$ is a minimal diagnosis w.r.t.\ current DPI}
\ElsIf{$L = nonmin$}								
	\State $\mD_{\supset} \gets \mD_{\supset} \cup \setof{\mathsf{node}}$ \label{algoline:dyn:add_to_Dsupset}  \Comment{$\mathsf{node}$ is a non-minimal diagnosis w.r.t.\ current DPI}
\Else 	
	\For{$e \in L$}\label{algoline:dyn:for_e_in_L}            \Comment{$L$ is a minimal conflict set w.r.t.\ current DPI}
		\State $\mathsf{node}_e \gets \Call{add}{\mathsf{node},e}$ \label{algoline:dyn:add_ax_to_node}       \Comment{$\mathsf{node}_e$ is generated}   
		\State $\mathsf{node}_{e}.\mathsf{cs} \gets \Call{add}{\mathsf{node.cs},L}$ \label{algoline:dyn:add_cs_to_node.cs}
		\If{$\mathsf{node}_e \in \Queue$}   \label{algoline:dyn:check_node_already_in_Q}                      \Comment{$\mathsf{node}_e$ is a (set-equal) duplicate of a node in $\Queue$}
			\State $\Queue_{dup} \gets \Call{insertSorted}{ \mathsf{node}_e, \Queue_{dup}, cardinality, ascending}$ \label{algoline:dyn:add_to_Qdup}
		\Else
			\State $\Queue \gets \Call{insertSorted}{ \mathsf{node}_e, \Queue, p_{nodes}(), descending}$\label{algoline:dyn:generate_nodes}
		\EndIf
			%
%
	\EndFor
\EndIf
\Until{$\Queue= [] \lor [\mD_{calc}\setminus\mD_{\checkmark} \neq \emptyset \land \left|\mD_{calc}\right| \geq n_{\min} \land ( |\mD_{calc}| = n_{\max} \lor \Call{getTime}{ } - t_{start} > t)]$}\label{algoline:dyn:until}
\State \Return $\tuple{\mD_{calc} ,\Queue, \mathbf{C}_{calc}, \mD_{\times}, \mD_{\supset}, \Queue_{dup}}$ \label{algoline:dyn:return}
\EndProcedure
\algstore{save_algo_dyn_hs}
\end{algorithmic}
\normalsize
\end{algorithm*}

\begin{algorithm*}
\small
\caption[Iterative Construction of a Dynamic Hitting Set Tree 2]{Iterative Construction of a Dynamic Hitting Set Tree (continued)} \label{algo:update_tree}
\begin{algorithmic}[1]
\algrestore{save_algo_dyn_hs}
\Procedure{\textsc{dLabel}}{$\langle\mo,\mb,\Tp,\Tn\rangle_\RQ,\mathsf{node},\mathbf{C}_{calc},\mD_{calc}, \Queue, \Queue_{dup}, p(), \Tp', \Tn'$} 
\For{$\mathsf{nd} \in \mD_{calc}$}\label{algoline:dlabel:non-min_crit_start}
	\If{$\mathsf{node} \supset \mathsf{nd}$}    \Comment{$\mathsf{node}$ is a non-minimal diagnosis}
			\State \Return $\tuple{nonmin,\mathbf{C}_{calc},\Queue_{dup}}$
	\EndIf
\EndFor\label{algoline:dlabel:non-min_crit_end}
\For{$\mc \in \mathbf{C}_{calc}$}\label{algoline:dlabel:reuse_start} \Comment{$\mC_{calc}$ includes only conflict sets w.r.t.\ current DPI}
	\If{$\mc \cap \mathsf{node} = \emptyset$}\label{algoline:dlabel:if_C_cap_node=emptyset}    \Comment{reuse (a subset of) $\mc$ to label $\mathsf{node}$} 
		\State $X \gets \Call{\scQX}{\langle\mc,\mb,\Tp\cup\Tp',\Tn\cup\Tn'\rangle_\RQ}$\label{algoline:dlabel:qx_1} \Comment{Algorithm~\ref{algo:qx} (page~\pageref{algo:qx}) to test if $\mc$ is minimal w.r.t.\ current DPI}
		\If{$X = \mc$} \label{algoline:dlabel:if_X=C}
			\State \Return $\tuple{\mc,\mathbf{C}_{calc},\Queue_{dup}}$\label{algoline:dlabel:return_C}
		\Else      \Comment{$X \subset \mc$}
			\State $\Queue_{dup} \gets \Call{pruneQdup}{X,\Queue_{dup}}$\label{algoline:dlabel:call_prune_Qdup}     \Comment{\textsc{pruneQdup} (see Algorithm~\ref{algo:prune})}
			\State $\Queue \gets \Call{prune}{X,\Queue,\Queue_{dup},p_{nodes}()}$\label{algoline:dlabel:call_prune_Q}     \Comment{\textsc{prune} (see Algorithm~\ref{algo:prune})}
			\State $\mD_{\supset} \gets \Call{prune}{X,\mD_{\supset},\Queue_{dup},\emptyset}$\label{algoline:dlabel:call_prune_Dsupset}
			%
			%
			\State $\mathbf{C}_{calc} \gets \Call{addSetDelSupsets}{X, \mathbf{C}_{calc}}$\label{algoline:dlabel:add_set_del_supset} \Comment{add $X$ to $\mC_{calc}$ and delete all its supersets from $\mC_{calc}$}
			\State \Return $\tuple{X,\mathbf{C}_{calc},\Queue_{dup}}$ \label{algoline:dlabel:return_X}
		\EndIf
	\EndIf
\EndFor\label{algoline:dlabel:reuse_end}
\State $L\gets \Call{QX}{\langle\mo\setminus\mathsf{node},\mb,\Tp\cup\Tp',\Tn\cup\Tn'\rangle_\RQ}$\label{algoline:dlabel:qx_2} \Comment{Algorithm~\ref{algo:qx} (page~\pageref{algo:qx}) to test if $\mathsf{node}$ is a diagnosis}
\If{$L$ = \text{'no conflict'}}						\Comment{$\mathsf{node}$ is a diagnosis}
	\State \Return $\tuple{valid,\mathbf{C}_{calc},\Queue_{dup}}$\label{algoline:dlabel:return_valid}
\Else						\Comment{$L$ is a \emph{new} minimal conflict set ($\notin \mathbf{C}_{calc}$)}
	\State $\mathbf{C}_{calc} \gets \mathbf{C}_{calc} \cup \setof{L}$\label{algoline:dlabel:add_new_cs}
	\State \Return $\tuple{L,\mathbf{C}_{calc},\Queue_{dup}}$\label{algoline:dlabel:return_new_cs}
\EndIf
\EndProcedure
\vspace{10pt}
\Procedure{\textsc{updateTree}}{$\langle\mo,\mb,\Tp,\Tn\rangle_\RQ,\mD_{\times}, \Queue, \Queue_{dup}, \mD_{\supset}, \mD_{\checkmark}, \mC_{calc}, p(), \Tp', \Tn'$}
\For{$\mathsf{nd} \in \mD_{\times}$} \label{algoline:update:process_Dtimes_start}
		\State $quickRC, completeRC \gets \false$
		\State $X \gets \Call{QX}{\langle U_{\mathsf{nd.cs}}\setminus\mathsf{nd},\mb,\Tp\cup\Tp',\Tn\cup\Tn'\rangle_\RQ}$\label{algoline:update:qx} 
		\For{$\mc \in \mathsf{nd.cs}$} 
			\If{$X \subset \mc$}\label{algoline:update:X_subset_C_(QRC)}												
				\State $quickRC \gets \true$ \label{algoline:update:qrc_gets_true}
				\State \textbf{break}  
			\EndIf
		\EndFor\label{algoline:update:quickPC_end} 
		\If{$quickRC = \false$}											
			\For{$i \leftarrow 1,\dots,|\mathsf{nd}|$}\label{algoline:update:completePC_start} 
				\State $X \gets \Call{QX}{\langle \mathsf{nd.cs}[i]\setminus \setof{\mathsf{nd}[i]},\mb,\Tp\cup\Tp',\Tn\cup\Tn'\rangle_\RQ}$\label{algoline:update:qx1}  
				\If{$X \neq \text{'no conflict'}$}      
					\State $completeRC \gets \true$     
					\State \textbf{break} 
				\EndIf
			\EndFor\label{algoline:update:completePC_end} 
		\EndIf
		\If{$quickRC = \true \;\lor\; completeRC = \true$}\label{algoline:update:if_qrc_or_crc_true}  \Comment{condition $\true$ iff $\mathsf{nd}$ redundant w.r.t.\ current DPI}
			\State $\Queue_{dup} \gets \Call{pruneQdup}{X,\Queue_{dup}}$\label{algoline:update:call_prune_Qdup}     \Comment{\textsc{pruneQdup} (see Algorithm~\ref{algo:prune})}
			\State $\Queue \gets \Call{prune}{X,\Queue,\Queue_{dup},p_{nodes}()}$\label{algoline:update:call_prune_Q} \Comment{\textsc{prune} (see Algorithm~\ref{algo:prune})}
			\State $\mD_{\times} \gets \Call{prune}{X,\mD_{\times},\Queue_{dup},\emptyset}$\label{algoline:update:call_prune_Dtimes}
			\State $\mD_{\supset} \gets \Call{prune}{X,\mD_{\supset},\Queue_{dup},\emptyset}$\label{algoline:update:call_prune_Dsupset}
			%
			%
			\State $\mC_{calc} \gets \Call{addSetDelSupsets}{X, \mathbf{C}_{calc}}$\label{algoline:update:add_set_del_supset} \Comment{add $X$ to $\mC_{calc}$ and delete all its supersets from $\mC_{calc}$} 
				
		\EndIf
	\EndFor
\For{$\mathsf{nd} \in \mD_{\times}$}\label{algoline:update:reinsert_D_of_Dx_to_Q}
	\Comment{add all (non-pruned) nodes in $\mD_{\times}$ to $\Queue$}
		\State $\Queue \gets \Call{insertSorted}{ \mathsf{nd}, \Queue, p_{nodes}(), descending}$\label{algoline:update:insert_sorted_0}
		\State $\mD_{\times} \gets \mD_{\times} \setminus \setof{\mathsf{nd}}$ \label{algoline:update:delete_from_Dtimes}
\EndFor \label{algoline:update:process_Dtimes_end}
\For{$\mathsf{nd} \in \mD_{\supset}$}\label{algoline:update:process_Dsupset_start} \Comment{update $\mD_{\supset}$: add all nodes to $\Queue$ which are not proper supersets of a diagnosis in $\mD_{\checkmark}$}
	\State $nonmin \gets \false$
	\For{$\mathsf{nd}' \in \mD_{\checkmark}$}
		\If{$\mathsf{nd} \supset \mathsf{nd}'$}    
			\State $nonmin \gets \true$
			\State \textbf{break} 
		\EndIf
	\EndFor
	\If{$nonmin = \false$}
		\State $\Queue \gets \Call{insertSorted}{ \mathsf{nd}, \Queue, p_{nodes}(), descending}$\label{algoline:update:insert_sorted_0.5}
		\State $\mD_{\supset} \gets \mD_{\supset} \setminus \setof{\mathsf{nd}}$\label{algoline:update:delete_from_Dsupset}
	\EndIf
\EndFor \label{algoline:update:process_Dsupset_end}
\For{$\md \in \mD_{\checkmark}$}\label{algoline:update:process_Dcheckmark_start}  \Comment{reinsert known minimal diagnoses to $\Queue$ to find diagnoses in order of descending $p_{nodes}()$}
	\State $\Queue \gets \Call{insertSorted}{ \md, \Queue, p_{nodes}(), descending}$\label{algoline:update:insert_sorted_1}
\EndFor \label{algoline:update:process_Dcheckmark_end}
\State \Return $\tuple{\Queue, \mD_{\times}, \mD_{\supset}, \mC_{calc}, \Queue_{dup}}$
\EndProcedure
\vspace{10pt}
%
\algstore{save_algo_dyn_hs1}
\end{algorithmic}
\normalsize
\end{algorithm*}

\begin{algorithm*}
\small
\caption[Iterative Construction of a Dynamic Hitting Set Tree 3]{Iterative Construction of a Dynamic Hitting Set Tree (continued)} \label{algo:prune}
\begin{algorithmic}[1]
\algrestore{save_algo_dyn_hs1}
\Procedure{\textsc{prune}}{$X,S,Dup,sort\_measure$}    
\If{$S$ is a list}
	\State $S' \gets []$
\Else
	\State $S' \gets \emptyset$
\EndIf
\For{$\mathsf{nd} \in S$}\label{algoline:prune:for_nd_in_S}
	\State $k \gets 0$ \label{algoline:prune:k_gets_0}
	\For{$i=1$ to $|\mathsf{nd.cs}|$}\label{algoline:prune:for_cs_in_nd.cs}
		\If{$\mathsf{nd.cs}[i]\supset X$}\label{algoline:prune:redundancy_check_part1}     \Comment{check first redundancy criterion} 
			\If{$\mathsf{nd}[i] \in \mathsf{nd.cs}[i]\setminus X$}\label{algoline:prune:redundancy_check_part2}    \Comment{check second redundancy criterion} 
				\State $k \gets i$\label{algoline:prune:k_gets_i}
			\Else
				\State $\mathsf{nd.cs}[i] \gets X$\label{algoline:prune:nd.cs[i]_gets_X}          \Comment{replace each superset of $X$ in $\mathsf{nd.cs}$ by $X$}
			\EndIf
		\EndIf
	\EndFor
	\If{$k > 0$}\label{algoline:prune:if_k>0}                                    \Comment{$\mathsf{nd}$ is redundant}
		\For{$\mathsf{node} \gets Dup[1],\dots,Dup[|Dup|]$}\label{algoline:prune:check_for_alternative_paths_start}
				\If{$|\mathsf{node}| \geq k \land \mathsf{nd}[1..|\mathsf{node}|] = \mathsf{node}$}\label{algoline:prune:check_if_alternative_subnode}
					\State $\mathsf{nd}_{new} \gets \Call{add}{\mathsf{node},\mathsf{nd}[|\mathsf{node}|+1..|\mathsf{nd}|]}$\label{algoline:prune:construct_alternative_equal_node}     \Comment{construct replacement node $\mathsf{nd}_{new}$ of $\mathsf{nd}$}
					\State $\mathsf{nd}_{new}.cs \gets \Call{add}{\mathsf{node.cs},\mathsf{nd.cs}[|\mathsf{node}|+1..|\mathsf{nd}|]}$\label{algoline:prune:construct_alternative_equal_node.cs}
					\State $S' \gets \Call{insertSorted}{ \mathsf{nd}_{new}, S', sort\_measure, descending}$\label{algoline:prune:insert_alternative_equal_node_into_S'}
					\State \textbf{break}\label{algoline:prune:break1}
				\EndIf
		\EndFor \label{algoline:prune:check_for_alternative_paths_end}
	\Else     \Comment{$X$ is not a witness of redundancy of $\mathsf{nd}$}
		\State $S' \gets \Call{insertSorted}{ \mathsf{nd}, S', sort\_measure, descending}$\label{algoline:prune:insert_same_node_into_S'}
	\EndIf
\EndFor
\State \Return $S'$ 
\EndProcedure
\vspace{10pt}
\Procedure{\textsc{pruneQdup}}{$X,Dup$}    
\State $Dup_{new} \gets []$
\For{$i \gets 1$ to $|Dup|$}\label{algoline:pruneQdup:for_i_in_Dup}
	\State $\mathsf{ndi} \gets Dup[i]$\label{algoline:pruneQdup:ndi_gets_Dup[i]}
	\State $k \gets 0$\label{algoline:pruneQdup:k_gets_0}
	\For{$m\gets 1$ to $|\mathsf{ndi.cs}|$}\label{algoline:pruneQdup:for_m_gets_1_to_ndi.cs_start}
		\If{$\mathsf{ndi.cs}[m]\supset X$}\label{algoline:pruneQdup:redundancy_check_part1}    \Comment{check first redundancy criterion} 
			\If{$\mathsf{ndi}[m] \in \mathsf{ndi.cs}[m]\setminus X$}\label{algoline:pruneQdup:redundancy_check_part2}    \Comment{check second redundancy criterion} 
				\State $k \gets m$\label{algoline:pruneQdup:k_gets_m}
			\Else
				\State $\mathsf{ndi.cs}[m] \gets X$\label{algoline:pruneQdup:ndi.cs[m]_gets_X}    \Comment{replace each superset of $X$ in $\mathsf{ndi.cs}$ by $X$}
			\EndIf
		\EndIf
	\EndFor\label{algoline:pruneQdup:endfor_m_gets_1_to_ndi.cs}
	\If{$k > 0$}\label{algoline:pruneQdup:if_k>0}                          \Comment{$\mathsf{ndi}$ is redundant}
		\For{$\mathsf{ndj} \in Dup_{new}$}\label{algoline:pruneQdup:for_ndj_in_Dupnew}
			\If{$|\mathsf{ndj}| \geq k \land \mathsf{ndi}[1..|\mathsf{ndj}|] = \mathsf{ndj}$}\label{algoline:pruneQdup:check_if_alternative_subnode}
				\State $\mathsf{ndi}_{new} \gets \Call{add}{\mathsf{ndj},\mathsf{ndi}[|\mathsf{ndj}|+1..|\mathsf{ndi}|]}$\label{algoline:pruneQdup:construct_alternative_equal_node}   \Comment{construct replacement node $\mathsf{ndi}_{new}$ of $\mathsf{ndi}$}
				\State $\mathsf{ndi}_{new}.cs \gets \Call{add}{\mathsf{ndj.cs},\mathsf{ndi.cs}[|\mathsf{ndj}|+1..|\mathsf{ndi}|]}$\label{algoline:pruneQdup:construct_alternative_equal_node.cs}
				\State $Dup_{new} \gets \Call{insertSorted}{ \mathsf{ndi}_{new}, Dup_{new}, cardinality, ascending}$\label{algoline:pruneQdup:insert_alternative_equal_node_into_Dupnew}
				\State \textbf{break}\label{algoline:pruneQdup:break}
			\EndIf
		\EndFor
	\Else                  \Comment{$X$ is not a witness of redundancy of $\mathsf{ndi}$}
	\State $Dup_{new} \gets \Call{insertSorted}{ \mathsf{ndi}, Dup_{new}, cardinality, ascending}$\label{algoline:pruneQdup:insert_node_into_Dupnew}
	\EndIf
\EndFor
\State \Return $Dup_{new}$
\EndProcedure
\end{algorithmic}
\normalsize
\end{algorithm*}
\restoregeometry

\newgeometry{margin=2cm}

\begin{table}[h]
\small
\centering
\rowcolors[]{2}{gray!8}{gray!16}
\begin{tabular}{p{0.23\textwidth} p{0.33\textwidth} p{0.33\textwidth}}
\rowcolor{gray!40}
\toprule\addlinespace[0pt] 
&\textsc{staticHS} & \textsc{dynamicHS} \\
\hline
\textbf{is used to solve} &
Interactive Static KB Debugging problem (Problem Definition~\ref{prob_def:static}) & 
Interactive Dynamic KB Debugging problem (Problem Definition~\ref{prob_def:dynamic})\\
\textbf{soundness} &
yes & 
yes \\
\textbf{completeness} &
yes & 
yes \\
\textbf{optimality} &
yes & 
yes \\
%
%
\textbf{number of solutions that must be considered (but not necessarily computed)} &
\vspace{-12pt}
\begin{itemize}
	\item initially fixed
	\item upper bound: $|\minD_{inputDPI}|$\vspace{-12pt}
\end{itemize}
&
\vspace{-12pt}
\begin{itemize}
	\item not initially fixed, depends on specified test cases (answered queries)
	\item upper bound: $|\allD_{inputDPI}|$\vspace{-10pt}
\end{itemize}
 \\ \hline
\textbf{diagnoses} &
considers only minimal diagnoses w.r.t.\ the input DPI which satisfy all answered queries added as test cases so far & 
considers only minimal diagnoses w.r.t.\ the current DPI\\
\textbf{conflict sets} &
computes only minimal conflict sets w.r.t.\ the input DPI  & 
computes minimal conflict sets w.r.t.\ the current DPI\\
\textbf{computes} &
a set $\mD$ including the $|\mD| \leq n_{\max}$ (a-priori) most probable minimal diagnoses w.r.t.\ the input DPI which satisfy all answered queries added as test cases so far & 
a set $\mD$ including the $|\mD| \leq n_{\max}$ (a-priori) most probable minimal diagnoses w.r.t.\ the current DPI\\
\textbf{purpose of test cases} &
differentiation between minimal diagnoses of fixed DPI 
& 
obtaining a new DPI with fewer minimal diagnoses
\\
\textbf{set of all minimal diagnoses upon addition of a test case} &
is reduced to a proper subset 
& 
some are invalidated, some new ones might be introduced 
\\
\textbf{set of all diagnoses upon addition of a test case} &
is reduced to a proper subset  & 
is reduced to a proper subset \\
\textbf{set of all minimal conflict sets upon addition of a test case} &
constant  & 
some minimal conflict sets are reduced to smaller sets and/or some new minimal conflict sets (in no set-relation with existing ones) are introduced \\ \hline
\textbf{constructed tree: \newline comparison to non-interac-tive wpHS-tree (Alg.~\ref{algo:hs})} &
equivalent (except for labels of leaf nodes) & 
might differ significantly \\
\textbf{non-leaf-node labels} &
only minimal conflict sets w.r.t.\ the input DPI  & 
(not necessarily minimal) conflict sets w.r.t.\ the current DPI  \\
\textbf{non-minimal and duplicate tree paths} &
deleted  & 
stored \\
\textbf{evolution of produced tree} &
only expansion (except for deletion of non-minimal and duplicate tree paths)  & 
alternating tree expansion and pruning phases \\
\textbf{pre-pruning (deletion of partial diagnoses)} &
only duplicate tree paths  & 
any \\
\textbf{post-pruning (deletion of complete diagnoses)} &
only non-minimal diagnoses (all invalidated minimal diagnoses are stored)  & 
any \\
\textbf{overall tree pruning} &
poor  & 
significant \\
%
\textbf{tree construction: worst case time and space complexity} &
\vspace{-12pt}
\begin{itemize}
	\item independent of specified test cases
	\item upper and lower bound is time and space required by non-interactive wpHS-tree (Alg.~\ref{algo:hs}) \vspace{-12pt}
\end{itemize}
&
\vspace{-12pt}
\begin{itemize}
	\item a function of the specified test cases and the leading diagnosis computation parameters $n_{\min}, n_{\max}, t$
	\item best case: significant savings compared to \textsc{staticHS}
	\item worst case: significant overhead compared to \textsc{staticHS} \vspace{-10pt}
\end{itemize}
 \\ 
\textbf{query generation} &
w.r.t.\ the current DPI  & 
w.r.t.\ the current DPI \\
%
\addlinespace[0pt]\bottomrule 
\end{tabular}
\caption{Comparison: \textsc{staticHS} versus \textsc{dynamicHS}.}
\label{tab:comparison_static_vs_dynamic}
\vspace{-10pt}
\end{table}

\restoregeometry

\section[Discussion of Iterative Diagnosis Computation]{Discussion of Iterative Diagnosis Computation%
\sectionmark{Discussion}}
\sectionmark{Discussion}
\label{sec:TextscStaticHSVersusTextscDynamicHS}
In this section we want to summarize properties of and differences between \textsc{staticHS} and \textsc{dynamicHS} that we already pointed out in previous sections and, additionally, we want to shed light on some further interesting aspects of these iterative diagnosis computation methods in the scope of interactive KB debugging (Algorithm~\ref{algo:inter_onto_debug}). Table~\ref{tab:comparison_static_vs_dynamic} provides an overview of what we did discuss or will discuss below.

\paragraph{First Segment of Table~\ref{tab:comparison_static_vs_dynamic} -- Addressed Problem and Properties w.r.t.\ Solutions.} The first row of the table has been proven by Proposition~\ref{prop:correctness_of_interactive_KB_debugging_algo} on page~\pageref{prop:correctness_of_interactive_KB_debugging_algo}. Results given by the second up to the fourth row of the table are substantiated by Propositions~\ref{prop:static_hs_correctness} (\textsc{staticHS}) and \ref{prop:dynamic_hs_correctness} (\textsc{dynamicHS}). We have discussed in Section~\ref{sec:TheIntuition} that Algorithm~\ref{algo:inter_onto_debug} with $mode = static$ can artificially fix the search space for possible solutions initially. This is an inherent property of the Interactive Static KB Debugging Problem which the algorithm aims to solve in static mode. For, a minimal diagnosis w.r.t.\ the input DPI which satisfies all answered queries added as test cases throughout the debugging session must be detected (see left column of category ``diagnoses'' in Table~\ref{tab:comparison_static_vs_dynamic}). Hence, the solution space is given by $|\minD_{inputDPI}|$. ``Initially fixed search space'' in this case means that, given the fault tolerance $\sigma = 0$, Algorithm~\ref{algo:inter_onto_debug} in static mode must compute all minimal diagnoses w.r.t.\ the input DPI, i.e.\ the entire set $\minD_{inputDPI}$. In case of dynamic mode, on the other hand, the solution space (i.e.\ minimal diagnoses w.r.t.\ the current DPI, see right column of Table~\ref{tab:comparison_static_vs_dynamic} in category ``diagnoses'') that needs to be explored by Algorithm~\ref{algo:inter_onto_debug} for a given value of zero for $\sigma$ is not known in advance. It rather depends on which test cases are specified or, respectively, which queries the user is asked. In case of the usage of mainly ``positive-impact queries'', the search space might have significantly smaller cardinality than $\minD_{inputDPI}$ whereas it might grow significantly beyond the cardinality of $\minD_{inputDPI}$ in a scenario where many unfavorable ``negative-impact queries'' are generated (cf.\ Section~\ref{sec:OverviewAndIntuition}). The maximum theoretically possible cardinality of the search space for \textsc{dynamicHS} is given by $|\allD_{inputDPI}|$. 

\paragraph{Second Segment of Table~\ref{tab:comparison_static_vs_dynamic} -- Impact of New Test Cases and Computation Focus.} The properties given in the category ``computes'' in Table~\ref{tab:comparison_static_vs_dynamic} are confirmed by Propositions~\ref{prop:static_hs_correctness} (\textsc{staticHS}) and \ref{prop:dynamic_hs_correctness} (\textsc{dynamicHS}). Hence, other than \textsc{dynamicHS} which analyzes the \emph{current DPI} in terms of minimal conflict sets and diagnoses in each iteration, \textsc{staticHS} must only consider minimal conflict sets w.r.t.\ the \emph{input DPI} (see categories ``diagnoses'' and ``conflict sets'' in Table~\ref{tab:comparison_static_vs_dynamic}). This is sufficient for the exploration of all minimal diagnoses w.r.t.\ the input DPI by Proposition~\ref{prop:mindiag_mincs}. 
In this vein, new test cases in static KB debugging are not taken into account in the computation of minimal conflict sets. Instead, new test cases are just exploited to invalidate \emph{already computed} minimal diagnoses w.r.t.\ the input DPI. 
Thus, test cases specified \emph{during} static KB debugging are treated somewhat inferior to test cases already present in the input DPI. Because, the newly gained information given by these test cases is not utilized to reveal new faults in the KB or to lay the focus on just the \emph{now} relevant parts of existing faults, but only for the purpose of constraining the search space for minimal diagnoses w.r.t.\ the input DPI $\langle\mo,\mb,\Tp,\Tn\rangle_\RQ$. We might thus call test cases added during the execution of Algorithm~\ref{algo:inter_onto_debug} with $mode = static$ pure \emph{differentiation test cases} (see category ``purpose of test cases'' in Table~\ref{tab:comparison_static_vs_dynamic}). 

Of course, seen from the point of view of a current DPI, i.e.\ the input DPI extended by differentiation test cases, \textsc{staticHS} does not guarantee completeness w.r.t.\ this current DPI, but only w.r.t.\ the initial one. This however does not mean that, after the (exact) solution $\ot := (\mo\setminus\md) \cup U_{\Tp}$ of the Interactive Static KB Debugging problem has been localized by means of \textsc{staticHS}, the differentiation test cases ($\Tp'$ and $\Tn'$) cannot be simply added to the DPI. In this case, $\ot$ is \emph{still} a maximal solution KB w.r.t.\ the extended input DPI $\langle\mo,\mb,\Tp\cup\Tp',\Tn\cup\Tn'\rangle_\RQ$. In other words, there is no conflict set (and thus no diagnosis) w.r.t.\ $\langle\mo\setminus\md,\mb,\Tp\cup\Tp',\Tn\cup\Tn'\rangle_\RQ$ and $\mo\setminus\md$ is valid w.r.t.\ $\langle\cdot,\mb,\Tp\cup\Tp',\Tn\cup\Tn'\rangle_\RQ$. However, in spite of using the (exact) solution KB of the Interactive Static KB Debugging problem, it is not ensured that this solution is the optimal one w.r.t.\ the \emph{extended DPI}, i.e.\ of the Interactive Dynamic KB Debugging problem. This is because user interaction is just exploited to the extent that the best solution w.r.t.\ the input DPI is crystallized out. It is not used to have the solution verified by the user in the light of the extended DPI. 

On the other hand, test cases assigned throughout dynamic KB debugging by means of Algorithm~\ref{algo:inter_onto_debug} with $mode = dynamic$ are treated equally as test cases already given in the input DPI. They are used to prune the search space and to pinpoint new faults that arise from added test cases resulting from answered queries. The dynamic algorithm assists the user in filtering out a solution and verifying in a thorough manner that this solution is the desired one w.r.t.\ the extended DPI, among \emph{all} existing solutions w.r.t.\ the extended DPI. Due to these aspects we might regard Algorithm~\ref{algo:inter_onto_debug} with mode $mode = dynamic$ as the \emph{standard method for Interactive KB Debugging}.

W.r.t.\ the impact of new test cases (answered queries) added to the DPI on the set of minimal (all) diagnoses 
it can be shown for \textsc{staticHS} that (for arbitrary iteration $i$ of Algorithm~\ref{algo:inter_onto_debug}) $\minD_i \supset \minD_{i+1}$ and $\allD_i \supset \allD_{i+1}$ where $\minD_i$ and $\allD_i$ denote the set of all minimal diagnoses and the set of all diagnoses, respectively, that are relevant (for the DPI considered) during iteration $i$. That is, the set of minimal as well as the set of all diagnoses (w.r.t.\ the input DPI) is reduced to a proper subset after a new test case has been added. For \textsc{dynamicHS}, (for arbitrary iteration $i$ of Algorithm~\ref{algo:inter_onto_debug}) we have that generally $\minD_i \not\supset \minD_{i+1}$, but still $\allD_i \supset \allD_{i+1}$, where $\minD_i$ and $\allD_i$ are defined as above.
That is, not only might some minimal diagnoses (w.r.t.\ the last-but-one DPI) be invalidated, but also some new ones (w.r.t.\ the current DPI) might originate from the incorporation of the information given by a query answer.

Concerning minimal conflict sets, the set of all (or: relevant) minimal conflict sets does not change throughout a debugging session by means of \textsc{staticHS}, i.e.\ $\minC_i = \minC_{i+1}$ (for arbitrary iteration $i$ of Algorithm~\ref{algo:inter_onto_debug}) where $\minC_i$ is the set of minimal conflict sets relevant (for the DPI considered) during iteration $i$. This holds since the minimal conflict sets w.r.t.\ the input DPI are artificially fixed (see above). 
On the contrary, the assignment of a new test case using \textsc{dynamicHS} involves the reduction of some minimal conflict sets (w.r.t.\ the last-but-one DPI) to smaller subset conflict sets (w.r.t.\ the current DPI) and/or the introduction of some ``completely new'' minimal conflict sets (which are in no subset-relation with existing ones, cf.\ Section~\ref{sec:OverviewAndIntuition}). These results are summarized by the categories ``set of all $X$ upon addition of a test case'' in Table~\ref{tab:comparison_static_vs_dynamic}.  

\paragraph{Third Segment of Table~\ref{tab:comparison_static_vs_dynamic} -- Hitting Set Tree Construction, Pruning, Complexity and Query Generation.} Regarding the constructed hitting set tree, we have explained that \textsc{staticHS} builds a wpHS-tree (see Definition~\ref{def:weighted_pruned_hs_tree} on page~\pageref{def:weighted_pruned_hs_tree}) 
just as the \textsc{HS} method which is employed for diagnosis computation in the presented non-interactive KB debugging scenario (Algorithm~\ref{algo:non_int_debug}). The main differences between Algorithm~\ref{algo:inter_onto_debug} in static mode and Algorithm~\ref{algo:non_int_debug} are, first, that the former constructs the wpHS-tree step-by-step in multiple phases. Between each two phases a query is generated and presented to the user. The latter, by contrast, finishes the tree construction (to the extent as prescribed by the given parameters $n_{\min}$, $n_{\max}$ and $t$, see Section~\ref{sec:non_int_debug_procedure}) before a single most probable automatically selected solution or a set of solutions is displayed to the user. Second, the tree constructed by the interactive static algorithm exhibits a different labeling of leaf nodes than the one built up be the non-interactive algorithm. In the former, some leaf nodes might be labeled by $\times$ indicating that the path to this node is a minimal diagnosis w.r.t.\ the input DPI, but one which is not in accordance with all answered queries. Notice that such invalidated diagnoses cannot be simply deleted in favor of memory savings, but must be stored in order for the non-minimality criterion (lines~\ref{algoline:slabel:non_min_crit_start}-\ref{algoline:slabel:non_min_crit_end}) to function properly which is necessary to preserve the property of \textsc{staticHS} to compute only \emph{minimal} diagnoses. 
In the non-interactive wpHS-tree, on the other hand, all minimal diagnoses w.r.t.\ the input DPI 
are labeled by $\checkmark$. 

What the interactive static and the non-interactive tree have in common is the usage of only minimal conflict sets w.r.t.\ the input DPI as labels of internal (i.e.\ non-leaf) nodes and the adherence to the ``standard'' pruning rules \cite{Reiter87} as per Definition~\ref{def:pruned_hs_tree} on page~\pageref{def:pruned_hs_tree}, i.e.\ the immediate deletion of non-minimal and duplicate tree paths. Except for the standard pruning actions that take place during tree expansion, no separate pruning phases are performed by \textsc{staticHS}. The reason for this is the fixation of the minimal conflict sets, i.e.\ the consideration of only minimal conflict sets w.r.t.\ the input DPI. Incorporation of new minimal conflict sets resulting from answered queries would generally negate completeness of \textsc{staticHS} w.r.t.\ the exploration of all minimal diagnoses w.r.t.\ the input DPI. Integration of new conflict sets that are subsets of existing ones, however, is the key to more substantial pruning actions carried out by \textsc{dynamicHS}. 

Due to the more or less equivalent construction of both the tree built up by \textsc{staticHS} and the one constructed by the \textsc{HS} method in the non-interactive algorithm, it is straightforward to recognize that the worst case time and space complexity of both \emph{tree} computations (without taking into the account other actions performed by the interactive algorithm like probability updates and query generations) are equal. By worst case complexity we refer to the complexity of the search for the (exact) solution solution of the Interactive Static KB Debugging Problem on the one hand and the complexity of enumerating all minimal diagnoses w.r.t.\ the input DPI on the other hand. In particular, the complexity of tree construction in static KB debugging is independent of given parameters such as the ones for leading diagnoses computation ($n_{\min}$, $n_{\max}$ and $t$) and of the test cases that are classified positively or negatively, respectively, during the debugging session. 

To sum up, due to the artificial fixation of the solution set, there is no possibility of tree pruning in static KB debugging except for the standard pruning rules and hence no way to escape the generally immense worst case complexity for diagnosis search in case $\sigma = 0$.

The hitting set tree constructed by \textsc{dynamicHS}, on the other hand, might differ significantly from the wpHS-tree produced by the non-interactive algorithm. First, it uses minimal conflict sets w.r.t.\ the \emph{current DPI} to label internal nodes in the tree during each expansion stage. Since minimal conflict sets can only ``shrink'' and not ``grow'' due to the integration of test cases into a DPI, 
the finding that by now a subset of a former minimal conflict set (w.r.t.\ some previous DPI) is already a minimal conflict set (w.r.t.\ the current DPI) gives rise to very powerful ways of tree pruning, as we 
illustrated by Example~\ref{example:dynamicHS_large_example_using_tabExDpi3}. In this vein, the evolution of the tree produced by \textsc{dynamicHS} can be characterized by alternating expansion and pruning stages. A pruning stage takes place after a test case has been added to the last-but-one DPI in order to modify the tree $T_i$ used to search for minimal diagnoses w.r.t.\ the last-but-one DPI to obtain a tree $T_{i+1}$ that enables the discovery of all minimal diagnoses w.r.t.\ the current DPI. Concretely, both pre-pruning as well as post-pruning is possible during a pruning phase. Pre-pruning refers to the deletion of tree paths ending in an open leaf node, i.e.\ paths corresponding to partial diagnoses, and post-pruning refers to the deletion of tree paths ending in a closed node, i.e.\ paths corresponding to (minimal or non-minimal) diagnoses. Both pre- and post-pruning are not possible in \textsc{staticHS}. The ability for significant tree pruning comes at the cost of not being able to exploit the standard pruning rules as \textsc{staticHS} does. For, non-minimal diagnoses and duplicate tree paths must be stored to guarantee the proper working of tree pruning and in further consequence the completeness of minimal diagnoses search for each current DPI. 

As we pointed out in Section~\ref{sec:OverviewAndIntuition}, the test cases specified during the dynamic debugging session and the defined leading diagnoses computation parameters $n_{\min}$, $n_{\max}$ and $t$ might have a material influence on the extent of possible tree pruning on the one hand and the extent of undesired tree growth on the other. Thence, worst case time and space complexity of the tree generation by means of \textsc{dynamicHS} cannot be initially (at least theoretically) quantified as in the case of \textsc{staticHS}. Consequently, significant savings as well as a substantial overhead compared to \textsc{staticHS} are possible. Careful ``control'' of certain properties of asked queries (added test cases) might help to keep considerable unwanted tree growth within bounds, as we touched upon in Section~\ref{sec:OverviewAndIntuition} and will elaborate on in future work. 

Nevertheless, we want to mention a shortcoming of \textsc{staticHS} compared to \textsc{dynamicHS}. Namely, for $\sigma = 0$, \textsc{staticHS} must \emph{enumerate all minimal diagnoses w.r.t.\ the input DPI} (otherwise no diagnosis can have a probability of 1, see the proof of Proposition~\ref{prop:correctness_of_interactive_KB_debugging_algo} in Section~\ref{sec:CorrectnessOfAlgorithmInterOntoDebug}) whereas \textsc{dynamicHS} might be able to obtain some extended DPI (by the addition of test cases) soon for which only one minimal diagnosis exists. This might require the computation of only a small fraction of the number of $|\minD_{inputDPI}|$ minimal diagnoses that \textsc{staticHS} must determine and therefore might be substantially more time and space saving than figuring out all minimal diagnoses w.r.t.\ some DPI. This is quite well illustrated by Examples~\ref{example:staticHS_complex_example_using_tabExDpi3} and \ref{example:dynamicHS_large_example_using_tabExDpi3}. 

Regarding query generation, we explained in Remark~\ref{rem:staticHS_query_computed_from_P_cup_P'_and_N_cup_N'} on page~\pageref{rem:staticHS_query_computed_from_P_cup_P'_and_N_cup_N'} that queries in \textsc{staticHS} are computed w.r.t.\ the current DPI albeit only minimal diagnoses w.r.t.\ the input DPI (which are at the same time minimal diagnoses w.r.t.\ the current DPI, cf.\ bullet (\ref{etc:staticHS_output_bullet_a}) on page~\pageref{etc:staticHS_output_bullet_a}) are considered and calculated by Algorithm~\ref{algo:inter_onto_debug} with $mode = static$. In the case of dynamic debugging it is clear that queries are computed w.r.t.\ the current DPI since only minimal diagnoses w.r.t.\ the current DPI are taken into account.

%% file: related.tex
\chapter{Related Work}
\label{chap:RelatedWork}
To the best of our knowledge no interactive KB debugging methods that ask a user automatically selected queries have been proposed to repair faulty (monotonic) KBs so far (except for our own previous works~\cite{ksgf2010, Shchekotykhin2012, Rodler2013, Shchekotykhin2014}).

Non-interactive debugging methods for KBs (ontologies) are introduced in~\cite{Schlobach2007,Kalyanpur.Just.ISWC07,friedrich2005gdm}. Ranking of diagnoses and proposing a ``best'' diagnosis is presented in~\cite{Kalyanpur2006}. This method uses a number of measures such as (a)~the frequency with which a formula appears in conflict sets, (b)~the impact on the KB in terms of its ``lost'' entailments when some formula is modified or removed, (c)~provenance information about the formula and (d)~syntactic relevance of a formula. All these measures are evaluated for each formula in a conflict set. The scores are then combined in a rank value which is associated with the corresponding formula. These ranks are then used by a modified hitting set tree algorithm that identifies diagnoses with a minimal rank. 
In this work no query generation and selection strategy is proposed if the intended diagnosis cannot be determined reliably with the given a-priori knowledge. In our work additional information is acquired until the minimal diagnosis with the intended semantics can be identified \emph{with confidence}. 
In general, the work of~\cite{Kalyanpur2006} can be combined with the approaches presented in our work as ranks of logical formulas can be taken into account together with other observations for calculating the prior probabilities of minimal diagnoses (see Section~\ref{sec:prob_space_construction}).

The idea of selecting the next query based on certain query selection measures was exploited in the generation of decisions trees \cite{Quinlan1986} and for selecting measurements in the model-based diagnosis of circuits \cite{dekleer1987} (in both works, the minimal expected entropy measure was used). We extended these methods to query selection in the domain of KB debugging \cite{ksgf2010} and devised further query selection measures \cite{Shchekotykhin2012,Rodler2013}.  

An approach for the debugging of faulty aligned KBs (ontologies) was proposed by \cite{meilicke2011}. An aligned KB is the union of two KBs $\mo_1$ and $\mo_2$ and an alignment $A_{1,2}$ (which is properly formatted as a set of logical formulas, cf.\ Definition~18 in \cite{meilicke2011}). $A_{1,2}$ is
a set of correspondences (each with an associated automatically computed confidence value) produced by an automatic system (an ontology matcher) given $\mo_1$ and $\mo_2$ as inputs where each correspondence represents a (possible) semantic relationship between a term occurring in $\mo_1$ and a term occurring in $\mo_2$.
The goal of a debugging system for faulty aligned KBs is usually the determination of a subset of the alignment $A'_{1,2} \subset A_{1,2}$ such that the aligned KB using $A'_{1,2}$ is not faulty. In terms of our approaches, this corresponds to the setting $\mo := A_{1,2}$ and $\mb := \mo_1 \cup \mo_2$. We have already shown in \cite{Rodler2012OM,Shchekotykhin2012b} that our systems can also be applied for fault localization in aligned KBs. The work of \cite{meilicke2011} describes approximate algorithms for computing a ``local optimal diagnosis'' and complete methods to discover a ``global optimal diagnosis''. Optimality in this context refers to the maximum sum of confidences in the resulting repaired alignment $A'_{1,2}$. In contrast to our framework, diagnoses are determined automatically without support for user interaction. Instead, \cite{meilicke2011} demonstrates techniques for the manual revision of the alignment as a procedure \emph{independent} from debugging. 
Another difference to our approach is the way of detecting sources of faults. We rely on a divide-and-conquer algorithm \cite{junker04} for the identification of a minimal conflict set $C \subseteq A_{1,2}$ (in \cite{meilicke2011} $C$ is called a MIPS, cf.\ \cite{friedrich2005gdm,Schlobach2007}). In the worst case the method we use exhibits only $O(|C|*\log(|A_{1,2}|/|C|))$ calls of some function that performs a check for faults in a KB and internally uses a reasoner (in our case \textsc{isKBValid}, see Algorithm~\ref{algo:qx}). The ``shrink'' strategy applied in \cite{meilicke2011} (which is similar to the ``expand-and-shrink'' method used in \cite{Kalyanpur.Just.ISWC07}), on the other hand, requires a worst case number of $O(|A_{1,2}|)$ calls to such a function. Empirical evaluations and a theoretical analysis of the best and worst case complexity of the ``expand-and-shrink'' method compared to the divide-and-conquer method performed in \cite{Shchekotykhin2008} revealed that the latter is preferable over the former. It should be noted that a similar divide-and-conquer method as used in our work could most probably be also plugged into the system in \cite{meilicke2011} instead of the ``shrink'' method.

There are some ontology matchers which incorporate alignment repair features: 
CODI \cite{Huber2011}, YAM++ \cite{Ngo2012}, ASMOV \cite{Jean-Mary2009} and KOSIMap \cite{Reul2010}, for instance, 
employ logic-based techniques to search for a set of predefined ``anti-patterns'' which must not occur in the aligned ontology, either to avoid inconsistencies or incoherencies or to eliminate unwanted or redundant entailments. In case such a pattern is revealed, it is resolved by eliminating from the alignment some correspondences responsible for its occurrence.  
All the techniques incorporated in these matchers are distinct from the presented approaches in that they implement incomplete or approximate methods of alignment repair, i.e. not all alternative solutions to the alignment debugging problem are taken into account. As a consequence of this, on the one hand, the final alignment produced by these systems may still trigger faults in the aligned KB. On the other hand, a suboptimal solution may be found, e.g.\ in terms of the user-intended semantics w.r.t. the aligned ontology or other criteria such as alignment confidence or cardinality.

Another ontology matcher, LogMap~2 \cite{Jimenez-Ruiz2012a}, provides integrated debugging features and the opportunity for a user to interact during this process. However, the system is not really comparable with ours since it is very specialized and dedicated to the goal of producing a fault-free alignment. Concretely, there are at least two differences to our approach. First, LogMap 2 uses incomplete reasoning mechanisms in order to speed up the matching process. Hence, the output is not guaranteed to be fault-free. Second, the option for user interaction aims in fact at the revision of a set of correspondences, i.e.\ the sequential assessing of single correspondences as 'faulty' or 'correct'. Our approach, on the contrary, asks the user queries (i.e.\ \emph{entailments} of non-faulty parts of the KB).  

An interactive technique similar to our approaches was presented in~\cite{Nikitina2011}, where a user is successively asked single KB formulas (ontology axioms) in order to 
obtain a partition of a given ontology into a set of desired or correct and a set of undesired or incorrect formulas. 
Whereas our strategies aim at finding a parsimonious solution involving minimal change to the given faulty KB in order to repair it, the method proposed in~\cite{Nikitina2011} pursues a (potentially) more invasive approach to KB quality assurance, namely a (reasoner-supported) exhaustive manual inspection of (parts of) a KB.  
Given an inconsistent/incoherent KB, this technique starts from an empty set of desired formulas aiming at adding to this set only correct formulas of the KB which preserve consistency and coherency. Our approach, on the other hand, works its way forward the other way round in that it starts from the complete KB aiming at finding a minimal set of formulas to be deleted or modified which are responsible for the violation of the pre-specified requirements.
Another difference of our approach compared to the one suggested in \cite{Nikitina2011} is the type of queries asked to the user and the way these are selected. Our method allows for the generation of queries which are not explicit formulas in the KB, but implicit consequences of non-faulty parts of the KB. Besides, the set of selectable queries in our approach differs from one iteration to the next due to the changing set of leading diagnoses whereas queries (i.e.\ KB formulas) in \cite{Nikitina2011} are known in advance and the challenge is to figure out the best ordering of formulas to be assessed by the user. Whereas we apply mostly information theoretic measures (e.g.\ the minimal expected entropy in the set of leading diagnoses after a query has been answered), the authors in \cite{Nikitina2011} employ ``impact measures'' which, roughly speaking, indicate the number of automatically classifiable formulas in case of positive and, respectively, negative classification of a query (i.e.\ a particular formula).  

%% file: conclusion.tex
\chapter{Summary and Future Work}
\label{chap:conclusion}

\subsubsection*{Summary}
\label{sec:Summary}
In this work we motivated why appropriate tool assistance is a must when it comes to 
repairing faulty KBs. For, KBs that do not satisfy some minimal quality criteria such as logical consistency can make artificial intelligence applications relying on the domain knowledge modeled by this KB completely useless. In such a case, no meaningful reasoning or answering of queries about the domain is possible.

Non-interactive debugging systems published in research literature often cannot localize all possible faults (\emph{incompleteness}), suggest the deletion or modification of unnecessarily large parts of the KB (\emph{non-minimality}), return incorrect solutions which lead to a repaired KB not satisfying the imposed quality requirements (\emph{unsoundness}) or suffer from \emph{poor scalability} due to the inherent complexity of the KB debugging problem \cite{Stuckenschmidt2008}. Even if a system is complete and sound and considers only minimal solutions, there are generally exponentially many solution candidates to select one from. However, any two repaired KBs obtained from these candidates differ in their semantics in terms of entailments and non-entailments. Selection of just any of these repaired KBs might result in unexpected entailments, the loss of desired entailments or unwanted changes to the KB which in turn might cause unexpected new faults during the further development or application of the repaired KB. Also, manual inspection of a large set of solution candidates can be time-consuming (if not practically infeasible), tedious and error-prone since human beings are normally not capable of fully realizing the semantic consequences of deleting a set of formulas from a KB. 

To account for this issue, we evolved a comprehensive theory on which provably complete, sound and optimal (in terms of given probability information) interactive KB debugging systems can be built which suggest only minimal changes to repair a present KB. Interaction with a user is realized by asking the user queries. That is, a conjunction of logical formulas must be classified either as an intended or a non-intended entailment of the correct KB. To construct a query, only a minimal set of two solution candidates must be available. After the answer to a query is known, the search space for solutions is pruned. Iteration of this process until 
there is only a single solution candidate left yields a (repaired) solution KB which features exactly the semantics desired and expected by the user.

We presented algorithms for the computation of minimal conflict sets, i.e.\ irreducible faulty subsets of the KB, and for the computation of minimal diagnoses, i.e.\ irreducible sets of KB formulas that must be properly modified or deleted in order to repair the KB. We combined these algorithms with methods that derive probabilities of diagnoses from meta information about faults (e.g.\ the outcome of a statistical analysis) to constitute a non-interactive debugging system for monotonic KBs which computes minimal diagnoses in best-first order. Building on the idea of this non-interactive method, we devised a complete and sound best-first algorithm for the interactive debugging of monotonic KBs that allows a user to take part in the debugging process in order to figure out the best solution.  

In order to integrate the new information collected by successive consultations of the user, the diagnoses computation in an interactive system must be regularly stopped. That is, there must be alternating phases, on the one hand for the further exploration of the solution space in order to gain new evidence for query generation and on the other hand for user interaction. To this end, we proposed two new strategies for the iterative computation of minimal diagnoses that exactly serve this purpose.
The first strategy, \textsc{staticHS}, takes advantage of an artificial fixation of the solution set which guarantees the monotonic reduction of the solution space independently of the asked queries, the given answers or other parameters of the algorithm. In this vein, the complexity of this algorithm is initially known and the maximum overhead compared to the non-interactive algorithm is polynomially bound.\footnote{This holds under the reasonable assumption that, in practice, a debugging session will involve only a polynomial number of queries to an interacting user. Recall that a user can abort the debugging session at any time and select the currently most probable diagnosis as their solution to the debugging problem.} On the downside, \textsc{staticHS} cannot optimally exploit the information given by the answered queries and thus cannot employ powerful methods that enable a more efficient pruning of the solution search space. Such powerful methods can be incorporated by the second suggested strategy, \textsc{dynamicHS}, the performance of which can be orders of magnitude better than the (initially fixed) performance of \textsc{staticHS} in the best case. That is, the ability to fully incorporate the information gained from user interaction might lead to a modified problem instance for which only a single (best) solution exists with only a small fraction of the time, space and user effort needed by \textsc{staticHS}. Moreover, the (exact) solution located by means of an interactive debugging session applying \textsc{dynamicHS} is generally a better (verified) solution than the (exact) solution found by use of \textsc{staticHS}.  
However, the complexity of \textsc{dynamicHS} depends to a great degree on which queries are generated and which input parameters are chosen and the worst case complexity is not initially bound as in case of \textsc{staticHS}. 

We want to point out that this work is unique in that it provides an in-depth theoretical workup of the topic of interactive (monotonic) KB debugging which (to the best of our knowledge) cannot be found in such a detailed fashion in other works. Furthermore, this is the first work that gives precise definitions of the problems addressed in interactive KB debugging. Additionally, it is unique in that it features (new) algorithms that provably solve these interactive KB debugging problems. To account for a tradeoff between solution quality and execution time, these algorithms are equipped with a feature to compute approximate solutions where the goodness of the approximation can be steered by the user. Another unique characteristic of this work is that it \emph{deals with an entire system of algorithms that are required for the interactive debugging of monotonic KBs}, considers and details all algorithms separately, 
proves their correctness and demonstrates how all these algorithms are orchestrated to make up a full-fledged and provably correct interactive KB debugging system.
\vspace{10pt}

\subsubsection*{Topics for Future Work}
	\label{sec:future_work}
This work has given rise to several questions we will elaborate on in our future work: 

\paragraph{Query Generation and Selection.}
Our discussions of the presented query generation methods have revealed some drawbacks (cf.\ Section~\ref{sec:QueryGeneration}). Albeit being a fixed-parameter tractable problem as argued, the exponential time complexity regarding the number of leading diagnoses $|\mD|$ in case an optimal query 
	is desired is clearly an aspect that should be improved. This high complexity arises from the paradigm of computing an optimal query w.r.t.\ some measure $qsm()$ by calculating a (generally exponentially large) pool $\QP$ of queries in a first stage, whereupon the best query in $\QP$ according to $qsm()$ is filtered out in a second stage.
	
	A key to solving this issue is the use of a different paradigm that does not rely on the computation of the pool $\QP$. Instead, qualitative measures can be derived from quantitative measures that have been used in interactive debugging scenarios~\cite{Shchekotykhin2012, Rodler2013, ksgf2010}. These qualitative measures provide a way to estimate the $qsm()$ value of \emph{partial q-partitions}, i.e.\ ones where not all leading diagnoses have been assigned to the respective set in the q-partition yet. In this way a \emph{direct} search for a query with (nearly) optimal properties is possible. A similar strategy called CKK has been employed in~\cite{Shchekotykhin2012} for the information gain measure $qsm() := \mathsf{ENT()}$ (see Section~\ref{sec:query_selection_measures}). From such a technique we can expect to save a high number of reasoner calls. Because usually only a small subset of q-partitions included in a query pool (of exponential cardinality) is required to find a query with desirable properties if the search is implemented by means of a \emph{heuristic} that involves the exploration of seemingly favorable (potential) queries and (partial) q-partitions, respectively, first. 

Another shortcoming of the paradigm of query pool generation and subsequent selection of the best query is the extensive use of reasoning services which may be computationally expensive (depending on the given DPI). Instead of computing a set of common entailments $Q$ of a set of KBs $\mo_i^*$ first and consulting a reasoner to fill up the (q-)partition for $Q$ in order to test whether $Q$ is a query at all (see Section~\ref{sec:QueryGeneration}), the idea enabling a significant reduction of reasoner dependence is to compute some kind of \emph{canonical query} without a reasoner and use simple set comparisons to decide whether the associated partition is a q-partition. Guided by qualitative properties mentioned before, a search for such q-partition with desirable properties can be accomplished \emph{without reasoning at all}. Also, a set-minimal version of the optimal canonical query can be computed without reasoning aid. Only for the optional enrichment of the identified optimal canonical query by additional entailments and for the subsequent minimization of the enriched query, the reasoner may be employed. We will present strategies accounting for these ideas in the near future.

Another aspect that can be improved is that \emph{only one} minimized version of each query is computed by Algorithm~\ref{algo:query_gen}. That is, per q-partition $\Pt$, there might be some set-minimal queries which do not occur in the output set $\QP$. From the point of view of how well a query might be understood by an interacting user, of course not all minimized queries can be assumed equally good in general. For instance, consider the minimized queries $Q_4$ and $Q_{10}$ in Table~\ref{tab:queries_partitions} on page~\pageref{tab:queries_partitions}. Both are equally good regarding their q-partitions (just the sets $\dx{}$ and $\dnx{}$ are commuted), but most people will probably agree that $Q_4$ is much easier to comprehend from the logical point of view and thus much easier to answer.

Hence, in order to avoid a situation where a potentially best-understood query w.r.t.\ $\Pt$ is not included in $\QP$, the query minimization process (see Section~\ref{sec:minimization_of_queries}) might be adapted to take into account some information about faults the interacting user is prone to. This could be exploited to estimate how well this user might be able to understand and answer a query. For instance, given that the user frequently has problems to apply $\exists$ in a correct manner to express what they intend to express, but has never made any mistakes in formulating implications $\rightarrow$, then the query $Q_1 = \setof{\forall X\,p(X) \rightarrow q(X), r(a)}$ might be better comprehended than $Q_2 = \setof{\forall X \exists Y s(X,Y)}$. One way to achieve the finding of a well-understood query for some q-partition $\Pt$ is to run the query minimization \textsc{minQ} more than once, each time with a modified input (using a hitting set tree to accomplish this in a systematic manner -- cf.\ Chapter~\ref{chap:DiagnosisComputation}, where an analogue idea is used to compute different minimal conflict sets w.r.t.\ a DPI). In this way, different set-minimal queries for $\Pt$ can be identified and the process can be stopped when a suitable query is found. 

In order to come up with such a strategy, however, one must first gain insight into how well a user might understand certain logical formalisms and what properties make a query easy to comprehend from the logical perspective. It is planned to gather corresponding data about different users in the scope of a user study and to utilize the results to achieve a model of ``query hardness'' (by sticking to a similar overall methodology as used in \cite{Horridge2011b}) in order to come up with strategies for the determination of minimal queries that are easily understood. Note that such a model could also act as a guide how to specify the initial fault probabilities of syntactical elements that are used to obtain diagnoses probabilities (see Section~\ref{sec:DiagnosisProbabilitySpace}).

\paragraph{Usage of ``Positive-Impact'' Queries in Combination with \textsc{dynamicHS}.}	
	As we discussed in Section~\ref{sec:OverviewAndIntuition} in the context of Algorithm~\ref{algo:inter_onto_debug} in dynamic mode, an added test case might give rise to some pruning steps as well as it might induce the construction of new subtrees (where ``new'' means that these would be no subtress of a hitting set tree w.r.t.\ the DPI not including this test case). The latter situation occurs when ``completely new'' minimal conflict sets (those that are in no subset-relationship with existing ones) are introduced by the addition of a test case. If this is the only impact of a test case, then this test case has only a negative influence on the time and space complexity of Algorithm~\ref{algo:inter_onto_debug} using \textsc{dynamicHS}. In other words, none of the invalidated minimal diagnoses (and no other nodes in the tree) are redundant, but all of them must additionally hit the set of ``completely new'' minimal conflict sets (in order to become diagnoses w.r.t.\ new DPI). Hence, in this case, the transition from one DPI to another including this test case results only in monotonic growth of the tree. If possible, such ``negative-impact test cases'' must be avoided. On the other hand, one must strive for the usage of ``positive-impact test cases'', i.e.\ those that only trigger tree pruning, but no tree expansion.  
Defining and studying properties that constitute such ``positive-impact test cases'' and ``negative-impact test cases'', respectively, and developing specialized algorithms for extracting exactly those types of queries that enable as substantial and effective pruning as possible in the context of \textsc{dynamicHS} is part of our already ongoing research. Note that a rough intuition of which properties make out a ``positive-impact test case'' is illustrated on the basis of an example in Section~\ref{sec:OverviewAndIntuition}.

\paragraph{Finding the Right Expert to Answer a Query in a Collaborative KB Development Setting.}	As we mentioned in Chapter~\ref{chap:intro}, there are collaborative KB development projects such as the OBO Project\footnote{http://obo.sourceforge.net} and the NCI Thesaurus\footnote{http://nciterms.nci.nih.gov/ncitbrowser}, where many different people contribute to the specification of their knowledge in large KBs. In such a setting, it may be hard to decide who is the person that has the highest chance of being able to answer a concrete query correctly. The idea in such a scenario could be to use a combination of different measures such as educational level (e.g.\ professor versus PhD student) or hierarchy of contributors (e.g.\ senior user versus regular user), statistical information about past faults of a contributor (e.g., how many of the formulas originally authored by a person have been corrected by other persons of higher educational level) or provenance information regarding terms occurring in the query (who has authored most of the formulas in which these terms occur?) in order to learn an ``expert model'' and use it to devise some kind of recommender system \cite{jannach2010} that suggests which person to ask a particular query. 

Once established, such an expert model together with provenance information of KB formulas and other types of information discussed in Section~\ref{sec:prob_space_construction} could also be exploited when it comes to the definition of the fault information provided as input to our debugging system. An example of a system which enables the remote collaborative development of KBs (ontologies) and also provides logs of interesting usage data such as formula change logs and provenance information is Web Prot\'{e}g\'{e}~\cite{Tudorache2013}.